\documentclass[fleqn,10pt,x11names]{wlscirep}
\usepackage[utf8]{inputenc}
\usepackage[T1]{fontenc}
\usepackage{subfig}
\usepackage{graphicx}
\usepackage{float}
\usepackage{caption}
\usepackage{tabularx}
\usepackage{array}
\newcolumntype{P}[1]{>{\centering\arraybackslash}p{#1}}
\usepackage{multirow}
\usepackage{comment}
\usepackage{tikz}
\usepackage{amsmath}
\usepackage{xcolor}
\usepackage{lscape}
\usepackage{lineno}
\usetikzlibrary{arrows.meta, shadows, fadings,shapes.arrows,positioning}
\tikzset{My Arrow Style/.style={single arrow, fill=white!50, anchor=base, align=center,text width=2.8cm, draw=red,line width=1pt}}

\newcolumntype{M}[1]{>{\centering\arraybackslash}m{#1}}

\title{Classification of datasets with imputed missing values: does imputation quality matter?}

\author[1,+]{Tolou Shadbahr}
\author[2,3,+,*]{Michael Roberts}
\author[2,+]{Jan Stanczuk}
\author[2,+]{Julian Gilbey}
\author[3,+]{Philip Teare}
\author[2,14]{S\"oren Dittmer}
\author[4]{Matthew Thorpe}
\author[5]{Ramon Vi\~nas Torn\'e}
\author[6]{Evis Sala}
\author[5]{Pietro Li\'o}
\author[3,7]{Mishal Patel}
\author[8]{AIX-COVNET Collaboration}
\author[9]{James H. F. Rudd}
\author[1,10,11]{Tuomas Mirtti}
\author[1,11,12]{Antti Sakari Rannikko}
\author[13]{John A. D. Aston}
\author[1]{Jing Tang}
\author[2]{Carola-Bibiane Sch\"onlieb}

\affil[1]{Research Program in Systems Oncology, Faculty of Medicine, University of Helsinki, Helsinki, Finland}
\affil[2]{Department of Applied Mathematics and Theoretical Physics, University of Cambridge, Cambridge, UK}
\affil[3]{Data Science \& Artificial Intelligence, AstraZeneca, Cambridge, UK}
\affil[4]{Department of Mathematics, University of Manchester, Manchester, UK}
\affil[5]{Department of Computer Science and Technology, University of Cambridge, Cambridge, UK}
\affil[6]{Department of Radiology, University of Cambridge, Cambridge, UK}

\affil[7]{Clinical Pharmacology \& Safety Sciences, AstraZeneca, Cambridge, UK}

\affil[8]{A list of authors and their affiliations appears at the end of the paper}
\affil[9]{Department of Medicine, University of Cambridge, Cambridge, UK}
\affil[10]{Department of Pathology, University of Helsinki and Helsinki University Hospital, Finland.}
\affil[11]{iCAN-Digital Precision Cancer Medicine Flagship, Helsinki, Finland.}
\affil[12]{Department of Urology, University of Helsinki and Helsinki University Hospital, Helsinki, Finland}
\affil[13]{Department of Pure Mathematics and Mathematical Statistics, University of Cambridge, Cambridge, UK}
\affil[14]{ZeTeM, University of Bremen, Bremen, Germany}

\affil[*]{corresponding author, email: michael.roberts@maths.cam.ac.uk}
\affil[+]{these authors contributed equally to this work}

\begin{abstract}
Classifying samples in incomplete datasets is a common aim for machine learning practitioners, but is non-trivial. Missing data is found in most real-world datasets and these missing values are typically imputed  using established methods, followed by classification of the now complete, imputed, samples. The focus of the machine learning researcher is then to optimise the downstream classification performance.
In this study, we highlight that it is imperative to consider the quality of the imputation. We demonstrate how the commonly used measures for assessing quality are flawed and propose a new class of discrepancy scores which focus on how well the method recreates the overall distribution of the data. 
To conclude, we highlight the compromised interpretability of classifier models trained using poorly imputed data. All code and data used in this paper are also released publicly at \textbf{[inserted upon publication]}.
\end{abstract}

\begin{document}

\flushbottom
\maketitle

\thispagestyle{empty}

\section{Introduction}

%\textbf{\textcolor{red}{[1058 words]}}

Datasets with missing values are ubiquitous in many applications, and feature values can be missing for many reasons. This may be due to incomplete or inadequate data collection, corruption of the dataset, or differences in the recording of data between different cohorts or sources. Furthermore, these reasons do not always remain constant or consistent across data sources. Each of these sources of missingness can also introduce different types of missingness, e.g.\ missing completely at random (MCAR), missing at random (MAR) and missing not at random (MNAR) \cite{eekhout_missing_2012}.

Training machine learning classification models is non-trivial if the underlying data is incomplete, as many methods require complete data \cite{emmanuel_survey_2021}. Simply excluding incomplete samples could lead to both a significant reduction in statistical power and also risks introducing a bias if the cause of the missingness is related to the outcome.
The typical solution to this problem is to follow a two-stage process, firstly imputing the missing values and then using a machine learning method to classify the now-complete dataset. 
There is an extensive literature discussing imputation methods, most of which is focussed around imputation of data at a single time-point (as we focus on). However, more nuanced scenarios have also been considered, in particular for imputation of longitudinal data \cite{luo_evaluating_2022, huque_comparison_2018}, imputation of decentralised datasets using a federated approach \cite{chang_multiple_2020} and imputation of variables which have a hierarchical (multi-level) relationship to one another \cite{van_buuren_flexible_2018}.

Despite the two-stage methods forming the bedrock of approaches to making predictions from incomplete data \cite{roberts_common_2021, wynants_prediction_2020, li_predicting_2021, score2_working_group_and_esc_cardiovascular_risk_collaboration_score2_2021} , it is still unclear which approaches perform `best', and how this should be measured. In particular, it is unclear how the classification model's performance is influenced by the underlying imputation method and how performance is affected by levels of missingness in the data. When fitting a model to incomplete data using a two-stage approach, the primary aim of many studies is to optimise the downstream classification performance, rather than carefully assessing if the imputed data reflects the underlying feature distribution. However, the latter is profoundly important, as a model trained using poorly imputed data could assign spurious importance to particular features. When we artificially induce missingness into a complete dataset, the quality of an imputation method is typically measured by comparing the imputed values with the ground truth using common metrics such as the (root / normalised) mean square error (MSE) \cite{chang_multiple_2020, deng_multiple_2016, schmitt_comparison_2015, muzellec_missing_2020, lin_missing_2020, platias_comparison_2020, armina_review_2017, thurow_goodness_2021, jager_benchmark_2021, emmanuel_survey_2021}, mean absolute (percentage) error \cite{lin_missing_2020, muzellec_missing_2020, platias_comparison_2020, emmanuel_survey_2021} or  (root / normalised) square deviance \cite{luo_evaluating_2022, chang_multiple_2020, deng_multiple_2016}. 
However, as other authors have commented \cite{van_buuren_flexible_2018}, optimal results for some metrics are achieved even when the distribution of the imputed data is far from the true distribution, see Figure~\ref{fig:AB}. 

There are a wide range of discrepancy scores for measuring imputation quality considered in the literature, with a distinction made for those used with categorical data and those used for continuous data. For example, for categorical data, one could consider the proportion of false classifications and Cram\'er's V metric. For continuous data there are many additional metrics and discrepancy scores such as, the two-sample Kolmogorov--Smirnov statistic, the (2-)Wasserstein distance (Mallows' $L^{2}$) and the Kullback--Leibler divergence. In particular, the paper of Thurrow et al.\cite{thurow_goodness_2021} considers imputation quality in detail, reporting many of these discrepancy scores, on a feature-by-feature basis to identify distributional differences between the imputed and true missing values.

In this paper, we systematically and carefully address two open research questions. 
Firstly, for two-stage methods, how should one choose the optimal configuration (e.g.\ imputation method and classifier) for a particular dataset, to ensure optimal classifying performance for the incomplete data? To address this, we systematically evaluate the performance of several two-stage methods for classifying incomplete data. Using multi-factor ANOVA analysis, we quantify how the downstream classification performance is influenced by the imputation method, classification method and data missingness rate.
Secondly, how faithfully do different data imputation methods reproduce the distribution of the underlying the dataset? This is a crucial question and requires careful and extensive evaluation. We assess imputation quality using standard metrics, such as RMSE, MAE and the coefficient of determination ($R^{2})$, and also introduce a class of discrepancy scores inspired by the sliced Wasserstein distance \cite{kantorovich_mathematical_1960} for evaluating how well imputed data faithfully reconstructs the overall distribution of feature values. We demonstrate how the new proposed class of measures is more appropriate for assessing imputation quality than existing popular discrepancy statistics.
We explore the link between imputation quality and downstream classification performance and show the remarkable result that a classifier built on poor imputation quality can actually give satisfactory downstream performance. We postulate that this could be due to the ability of powerful methods to overcome issues in the imputed data, or that a poor imputation is equivalent to injecting noise into the dataset. Training machine learning-based models on such noisy data can be viewed as a form of data augmentation known to improve generalisability and robustness of the models \cite{goodfellow_deep_2016}.

The stability of different imputation methods is explored and we found that the algorithms consisting of neural network components are susceptible to local minima and can give highly variable imputation results. In adidition, we find that the popular discrepancy measures use to assess imputation quality are uncorrelated from the downstream model performance, whereas the measures which consider distributional discrepancies do show a correlation. Finally, we demonstrate how high-performing classification models trained using poorly imputed data assign spurious importances to particular features in the dataset.
We release a codebase to the community, with example datasets, at \texttt{github.com/}\textbf{[shared upon publication]} which provides a framework for practitioners to allow for easy, reproducible benchmarking of imputation method performance, for evaluating a wide range of classifiers along with assessing imputation quality in a completely transparent way. 

\section{Materials and Methods}

In this section we briefly introduce the datasets used in the study, the benchmarking exercise and the methods for assessing imputation quality.

\subsection{Datasets}

In this study, we focus on four datasets denoted as \textbf{Breast Cancer}, \textbf{MIMIC-III}, \textbf{NHSX COVID-19}, and \textbf{Simulated}. The first three of these are derived from real clinical datasets, while the final one is synthetic. The \textbf{MIMIC-III} and \textbf{Simulated} datasets are complete (i.e.\ do not contain missing values), giving us control over the induced missingness type and rate. The \textbf{Breast Cancer} and \textbf{\textbf{NHSX COVID-19}} datasets exhibit their natural missingness.

\begin{itemize}
\item \textbf{Simulated} is a synthetic dataset created using the \texttt{scikit-learn} \cite{pedregosa_scikit-learn_2011} function \verb+make_classification+, giving a dataset with 1000 samples and 25 informative features. This allows us to perform a simulation study to determine which factors can potentially influence classification after imputation.

\item \textbf{MIMIC-III}. The Medical Information Mart for Intensive Care (MIMIC) dataset \cite{johnson_mimic-iii_2016} is a large, freely-available database comprising deidentified health-related data from patients who were admitted to the critical care units of the Beth Israel Deaconess Medical Center in the years 2001--2012. Following the preprocessing detailed in the Supplementary Materials, we obtain the \textbf{MIMIC-III} dataset used in this paper. This contains data for 7214 unique patients with 14 clinical features and a survival outcome recorded for each patient.

\item \textbf{Breast Cancer}. This dataset is derived from an oncology dataset collected at Memorial Sloan Kettering Cancer Center between April 2014 and March 2017 \cite{razavi_genomic_2018}. The dataset contains genomic profiling of 1918 tumor samples from 1756 patients with detailed clinical variables and outcomes for each patient and the therapy administrated over the time of treatment. The \textbf{Breast Cancer} dataset is obtained after the preprocessing described in the Supplementary Materials and has 16 features for 1756 patients with the natural missingness of the data retained.

\item \textbf{NHSX COVID-19}. This dataset is derived from the NHSX National COVID-19 Chest Imaging Database (NCCID) \cite{cushnan_towards_2021} which contains clinical variables and outcomes for COVID-19 patients admitted in many hospitals around the UK. The dataset is continually updated by the NHSX; the download we performed was on 5 August 2020 which contained data for 851 unique patients. The \textbf{NHSX COVID-19} dataset contains 23 features for 851 patients.

\end{itemize}

\noindent
\textbf{Outcome variables.}
For the \textbf{MIMIC-III}, \textbf{Breast Cancer} and \textbf{NHSX COVID-19} datasets we use survival status as the outcome of interest, the \textbf{Simulated} dataset outcome is a binary variable generated at the data synthesis stage by the \verb+make_classification+ 
function in \texttt{scikit-learn} \cite{pedregosa_scikit-learn_2011}.

\subsection{Dataset partitioning}

In our experiments, we simultaneously compare many combinations of imputation and classification methods, each with several hyperparameters that can be tuned. Therefore, we must be careful to avoid overfitting. To address this, we partition each dataset at two levels. At the first level, we identify the development and holdout cohorts and at the second, we partition the development sets into training and validation cohorts (see the Supplementary material for more details.)\\

\noindent
\textbf{Induced missingness.}
The \textbf{Simulated} and \textbf{MIMIC-III} datasets are complete and, therefore, we can induce different missingness rates in the development and holdout datasets. We randomly removed 25\% and 50\% of entries from each of the development and holdout datasets. This allows us to compare 4 different scenarios for development-holdout missingness rates: 25\%-25\%, 25\%-50\%, 50\%-25\% and 50\%-50\%.

\subsection{Imputation methods}

Data imputation is the process of substituting missing values in a dataset with new values that are, ideally, close to the true values which would have been recorded if they had been observed. To formalise the notation, we consider a dataset $\mathcal{D}$, consisting of $N$~samples $\mathbf{x}_i\in\mathbb{R}^{d}$ drawn from some (unknown) distribution. Some of the elements in these samples may be missing and we impute them to give the complete samples~$\hat{\mathbf{x}}_i$.
Whilst it is not possible to be certain about the true value of the missing entries, it is desirable that the uncertainty in the imputed values be considered any downstream task which relies on the imputed data \cite{little_statistical_2019}. 

In general, imputation methods fall into two categories, namely `single imputation' and `multiple imputation'. In the case of single imputation, plausible values are imputed in place of the missing values just once, whereas for multiple imputation methods \cite{rubin_overview_1988, rubin_multiple_1987}, imputation is performed multiple times to generate a series of imputed values for each missing item. For imputation methods that have some stochastic nature, this allows for insight into the uncertainty of the imputed values. 
For multiple imputation methods, there are two approaches for performing a downstream classification task. The first approach involves pooling the multiple imputation results and creating a single, summary, dataset from which we perform the classification task. The second approach, which we follow in this paper, is to perform the classification task on each of the multiple imputation results separately, before pooling the predictions of the classification model. The final prediction is made using e.g.\ averaging or majority vote. See van Buuren\cite{van_buuren_flexible_2018}, for a comparison of the approaches and justification for the latter being preferable.
 
In our study, we consider five popular imputation methods from the literature, namely mean imputation, multivariate imputation by chained equations (MICE) \cite{van_buuren_flexible_1999, van_buuren_mice_2011}, MissForest \cite{stekhoven_missforestnon-parametric_2012}, generative adversarial imputation networks (GAIN) \cite{yoon_gain_2018}, and the missing data importance-weighted autoencoder (MIWAE) \cite{mattei_miwae_2019}.
Among the four datasets we consider, \textbf{Simulated} and \textbf{MIMIC-III} contain only numerical features whereas \textbf{Breast Cancer} and \textbf{NHSX COVID-19} contain numerical, categorical (single and multi-level) and ordinal features. In preprocessing, the categorical features are one-hot encoded and the ordinal features are coded with integer values. 

\subsection{Classification methods}

Classification is the process of grouping items with similar characteristics into specific classes. In our case, the classification methods are machine learning-based algorithms which take an input sample  $\hat{\mathbf{x}}_i\in\mathbb{R}^{d}$ and output a particular class label $\ell_{i}\in\{0,1\}$. The classification methods we consider in this paper are logistic regression, Random Forest, XGBoost, NGBoost and an artificial neural network (for more details see the Supplementary Materials.) 

\subsection{Hyperparameter optimisation}
\label{sec:hyperparam-opt}

Each of the classifiers that we consider (except for logistic regression) have several tuneable hyperparameters, with classifier performance significantly dependent on them. In order to identify the best classifier for each dataset, we perform an exhaustive grid search over a wide range of hyperparameters.
For each configuration of dataset, train/test missingness level, imputation method, holdout set, validation set and each of the multiple imputations, we fit a classifier, exhaustively, over a large range of hyperparameters as detailed in Supplementary Table~\ref{tab:hparams}. To identify the optimal hyperparameter choice for each of the these configurations, we evaluate the model on each of the five validation folds. The hyperparameter choice which results in the best average performance over these validation folds is selected as the optimal model configuration i.e.\ for each holdout set H1--H3 (Fig.~\ref{fig:dataset_splitting}) we obtain an optimal model.
This exhaustive grid search required millions of experiments to ensure the comparisons are fair. 

\subsection{ANOVA analysis}

One of the key aims of this paper is to identify and quantify the influence of key factors on the performance of the downstream classification. To quantify the impact, we used a generalized linear model (binomial logistic regression-based) to perform multi-factor ANOVA for the holdout AUC of the optimal classifiers identified in \S\ref{sec:hyperparam-opt}. 
Firstly, to establish the factors most influencing the models built on the real clinical models, we pool the results for the \textbf{MIMIC-III}, \textbf{Breast Cancer}, and \textbf{NHSX COVID-19} datasets and perform a logistic ANOVA. 
Additionally, each of the datasets are assessed individually.
After constructing the ANOVA model including all factors and their interaction, we excluded factors not significant at the 1\% level using backward elimination (see the Supplementary Materials for more details). 

\subsection{Measuring the quality of imputation}
\label{sec:measuring-quality}

In addition to determining how the imputation method, missingness rates and datasets affect the downstream classification performance, we are also interested in exploring how the quality of the imputation affects the downstream classification performance. However, there is no widely accepted approach for measuring the quality of an imputation.

In the literature, many authors, such as Jadha et. al.\cite{jadhav_comparison_2019} and Thurow et. al. \cite{thurow_goodness_2021}, assess imputation quality by taking a complete dataset, introducing missingness artificially and imputing the resulting incomplete dataset; we follow this approach here. For this purpose, we induced missingness completely at random (MCAR) with rate 25\% and 50\% into the \textbf{MIMIC-III} and \textbf{Simulated} datasets. In order to quantitatively assess how well an imputation method reconstructs the missing values, we must define a score that has the desirable property that it has a low value when the distribution of imputed values closely resembles that of the true values. Explicitly, our aim is to compute a distance between the original samples $\mathcal{D} = \{ \mathbf{x}_i \}_{i=1}^{N}$ and the imputed samples $\hat{\mathcal{D}} = \{ \hat{\mathbf{x}}_i \}_{i=1}^{N}$ that will indicate the quality of the imputation.

In this paper, we consider many of the popular discrepancy statistics used in the literature for measuring imputation quality. 
These typically fall into two classes: (A) measures of discrepancy between the imputed and true values of individual samples, and (B) measures of discrepancy in distributions of individual features for the imputed and true data.
The class of measures which would be of most practical value to practitioners is (C) measures of discrepancy for imputed and true data across the whole data distribution. In the literature, we were unable to find any examples of discrepancy measures of this type and we propose such a class in this paper.

\textbf{A. Sample-wise discrepancy.} 
In much of the literature, the quality of imputation is determined by measuring the discrepancy in the real and imputed values sample-by-sample, then summarising over all samples in the dataset. In this paper, we consider the three discrepancy statistics, Root mean square error (RMSE), mean absolute error (MAE), and the coefficient of determination ($\mathbf{R^2}$). These statistics compare the imputed values explicitly to the true values. 

\textbf{B. Feature-wise distribution discrepancy.} 
In the literature, some authors have considered discrepancy measures to quantify for how faithfully the distributions of individual features are reconstructed. In particular, Thurow et al.\cite{thurow_goodness_2021}\ consider several distribution measures on a feature-by-feature basis, including the Kullback--Leibler (KL) divergence, the two-sample Kolmogorov--Smirnov (KS) statistic and 2-Wasserstein (2W) distance. We report results for all of these in this study.
As these discrepancies are measured feature-by-feature, we get many scores for each dataset and report the minimum, maximum and median discrepancy for each distance over all features. More details about these statistics, including definitions and the implementations used, can be found in the Supplementary Materials. 

\begin{figure}[htb!]
    \centering
    \begin{tabular}{ccc}
      \includegraphics[width=4cm]{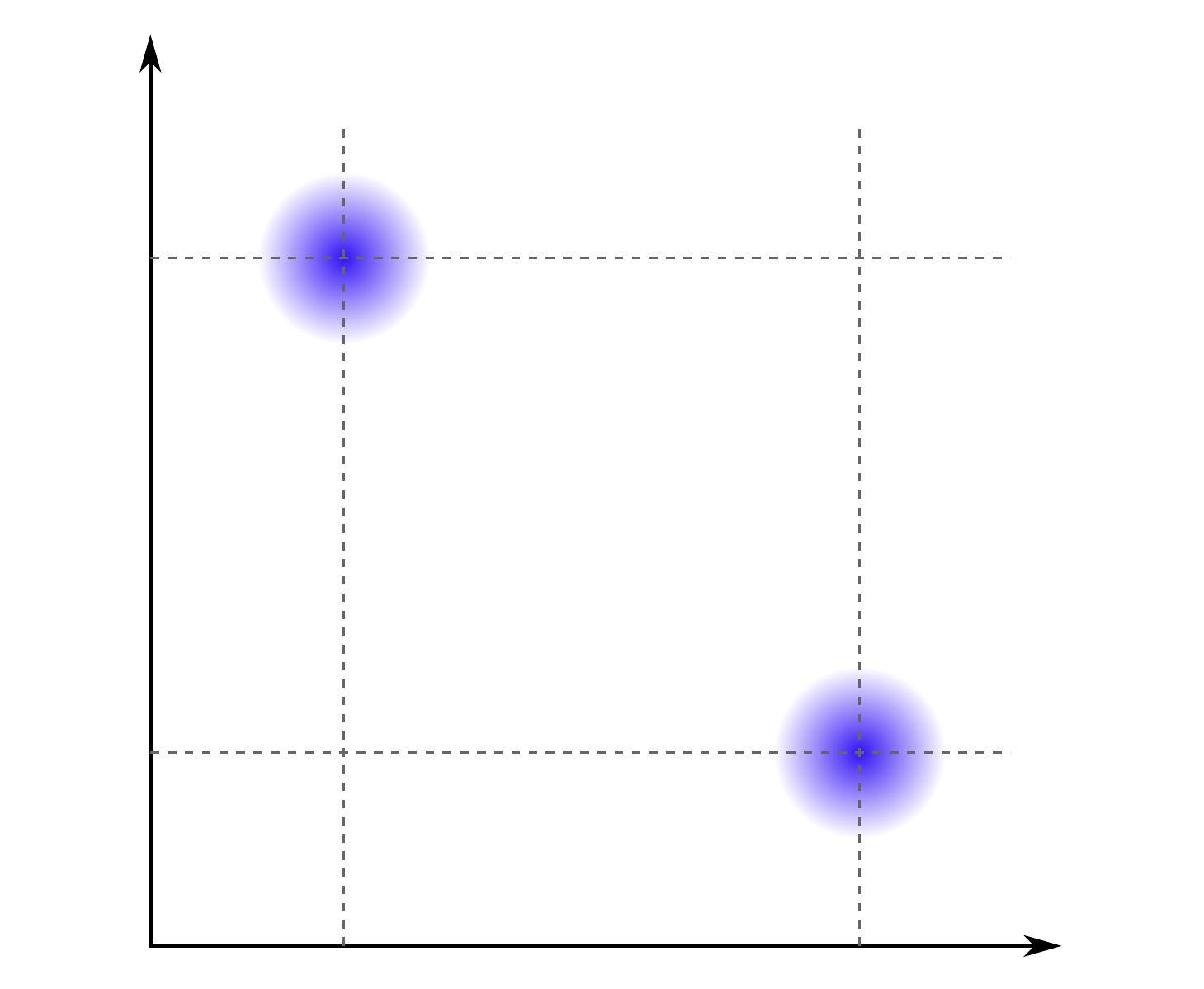}&
      \includegraphics[width=4cm]{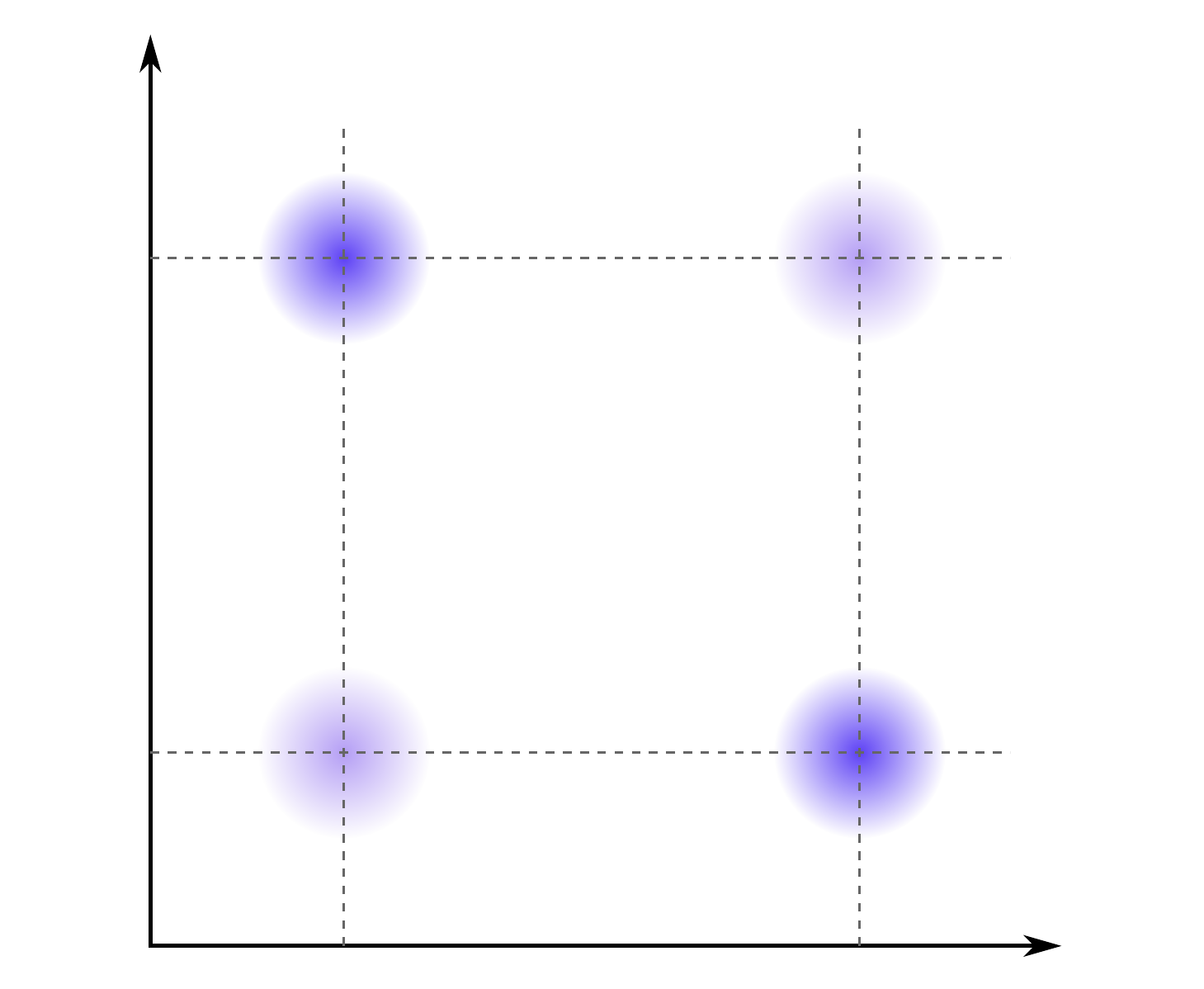}&
      \includegraphics[width=4cm]{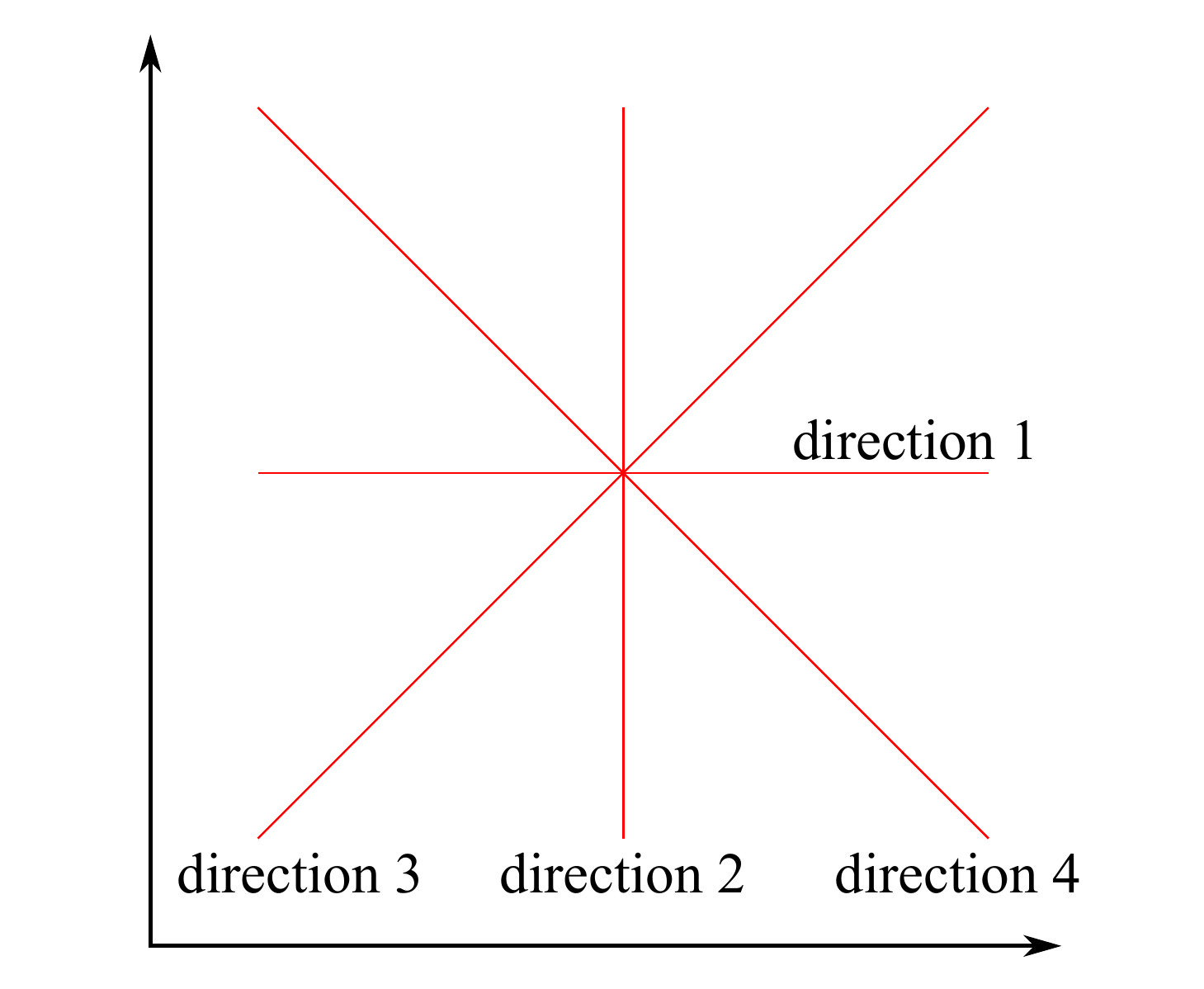}\\
      (a) Original & (b) Imputed & (c) Some directions \\[\bigskipamount]
      \includegraphics[width=4cm]{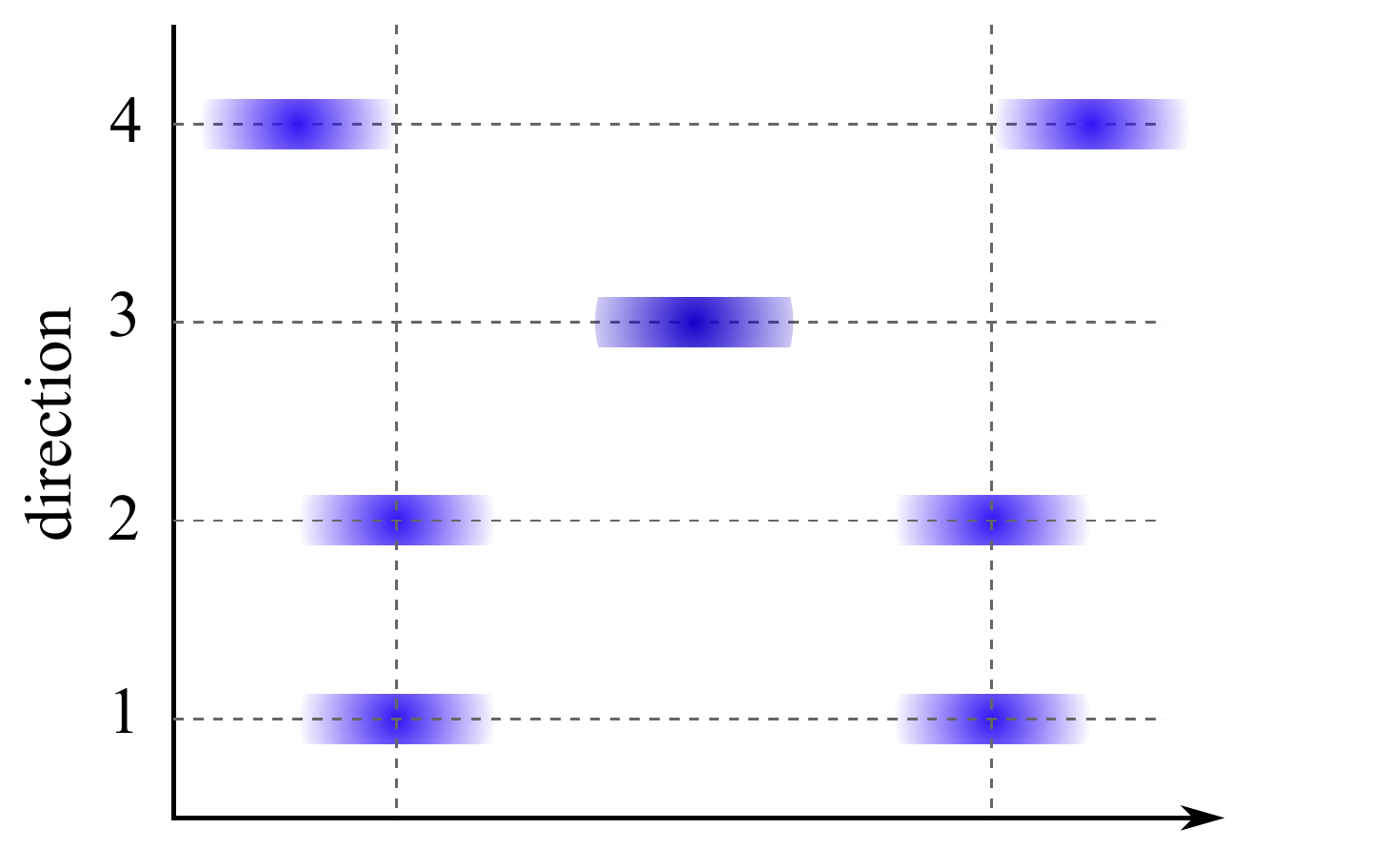}&
      \includegraphics[width=4cm]{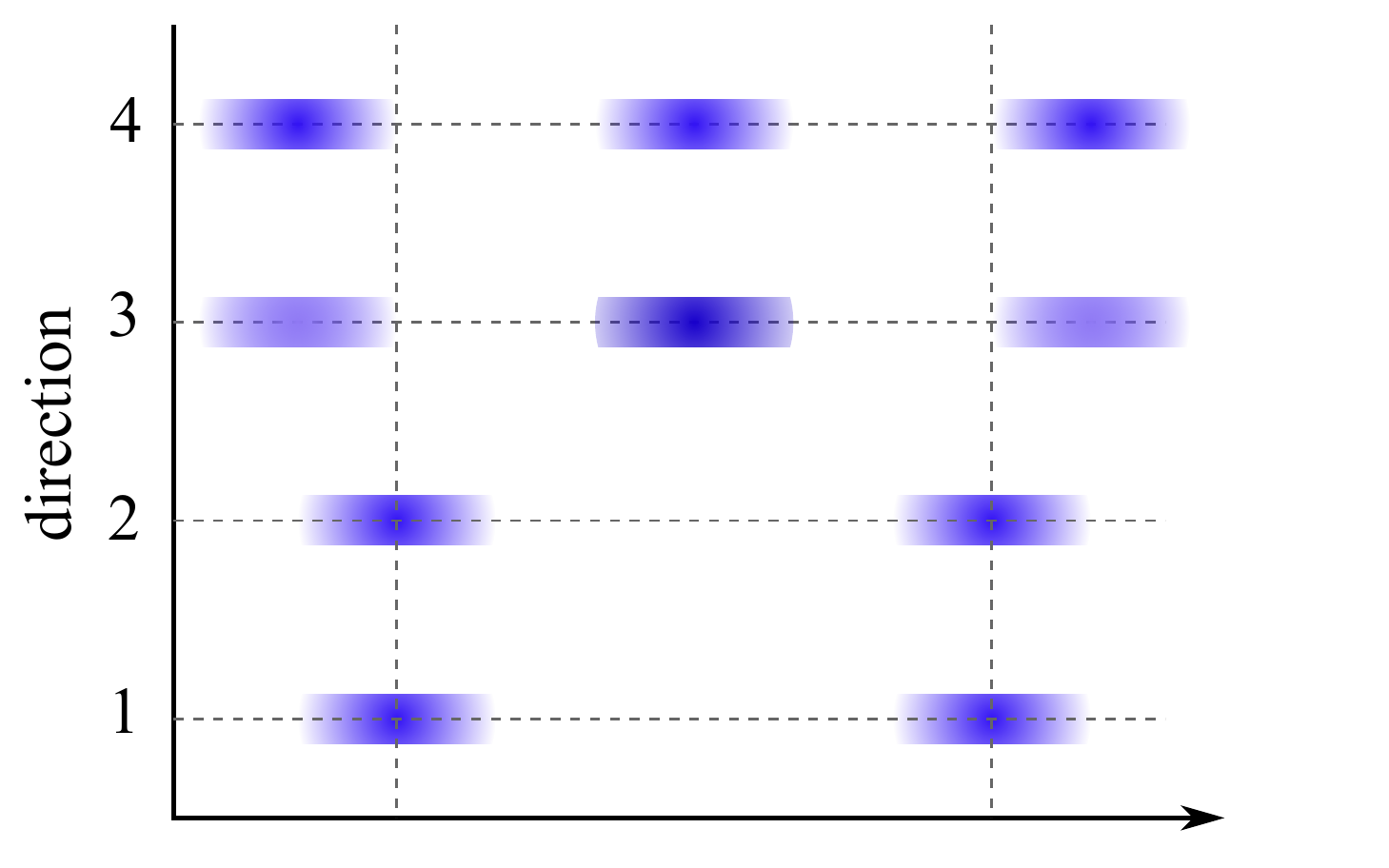} \\
      (d) Projections of original distribution & (e) Projections of imputed distribution
    \end{tabular}
    \caption{The effects of projecting imputed data in multiple directions. (a)~The underlying 2-dimensional data distribution; (b)~the distribution of imputed data; (c) some example directions: 1 and 2 are in the direction of the features, directions 3 and~4 are not; (d)~and~(e) show the original and imputed data distributions projected onto the four directions shown in~(c): their marginals (directions 1 and~2) are indistinguishable but the distributions are clearly different when projected in directions 3 and~4.}
    \label{fig:ABC}
\end{figure}

\textbf{C. Sliced Wasserstein distance.}
In Figure~\ref{fig:ABC}, we see that simply considering the feature-by-feature marginal distributions is not sufficient to quantify how well a high-dimensional data structure has been imputed. The marginal distributions of the imputed data (directions 1 and 2) match that of the original data perfectly but do not identify the discrepancy of the distributions shown in Figures~\ref{fig:ABC}(a) and (b). This motivates us to consider a new measure, which harnesses the multi-dimensional nature of the data to better identify distribution differences like this.

\noindent 
Modelling the discrepancy of imputed and true data in high dimensions is challenging for two key reasons. Firstly, the curse of dimensionality results in computations which are infeasible for high-dimensional datasets. Secondly, high-dimensional datasets are sparse unless there are an unrealistically large number of samples. We address both of these issues by repeatedly projecting the entire data distribution to random one-dimensional subspaces, thereby increasing the density of the data while considering more axes than those simply defined by the features themselves. We then consider the distance of the imputed data from the original data in the random direction, commonly known as the sliced Wasserstein distance\cite{rabin2011wasserstein,bonneel2015sliced}; this gives a distribution of distances across all randomly chosen directions. Initially, we perform the following steps:

\begin{itemize}
    \item Step 1: Determine partitions and random directions.
    
    Choose $M$ random unit vectors (directions) $\mathbf{n}_r\in\mathbb{R}^d$, $r=1$, \dots,~$M$, where $M\ge d$. Choose $P$~random partitions of the index set $\{1, 2, \dots, N\}$ into subsets $I_p$ and~$J_p$ for $p=1$, \dots,~$P$ (where $N$~is the number of samples).  If $N$~is even, then $I_p$ and $J_p$ are the same size, while if $N$~is odd, these subsets have sizes $(N+1)/2$ and $(N-1)/2$ respectively. To account for the dimensionality differences between \textbf{MIMIC-III} and \textbf{Simulated}, we used $M=50$ and $M=90$ respectively. We set $P=10$ throughout.

    \item Step 2: Calculate projections of data.

    For each $r$, we project all of the data onto the one-dimensional subspace of~$\mathbb{R}^d$ spanned by~$\mathbf{n}_r$; this gives the projected original data $\mathbf{x}_i.\mathbf{n}_r$ and projected imputed data $\hat{\mathbf{x}}_i.\mathbf{n}_r$.

    \item Step 3: Calculate sliced Wasserstein Distances.

    For each pair $(r,p)$, with $r\in\{1,\dots,M\}$ and $p\in\{1,\dots,P\}$, we calculate (2-)Wasserstein distances between the projected original and imputed data. The data are normalised by dividing through by the standard deviation, $s$, of the projected data $\mathbf{x}_i.\mathbf{n}_r$ for $i\in I_p$. The data points in $I_p$ are taken as the `true' distribution and we determine the distance $w(r, p)$ of the remaining data in $J_p$ from this. This is our baseline distance inherent to the data.
    We then calculate $\hat w(r, p)$, the 2-Wasserstein distance between $\mathbf{x}_i.\mathbf{n}_r/s$ for $i\in I_p$ and the imputed data $\hat{\mathbf{x}}_{j}.\mathbf{n}_r/s$ for $j\in J_p$.
\end{itemize}

\begin{figure}[htb!]
    \centering
    \begin{tabular}{p{7.2cm} c p{5.7cm}}
    \includegraphics[width=7cm]{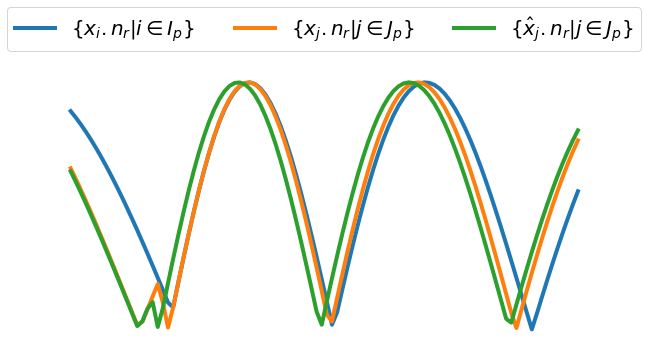}
    &
    \begin{tikzpicture}[scale=0.6]
    \node[My Arrow Style]{Taken over all partitions, random directions, validation sets and multiple imputations};
    \end{tikzpicture}&
      \includegraphics[width=5.5cm]{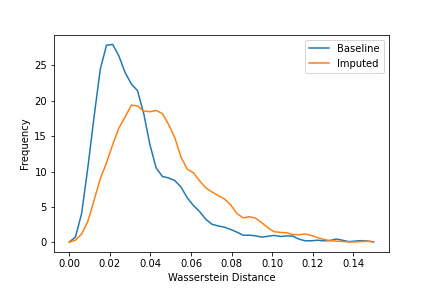} \\
      (a) Example density plot of the projection onto $\mathbf{n}_r$ of the original data in $I_p$ (blue) and $J_p$ (orange), and the imputed data in $J_p$ (green) & & (b) Density plot of the sliced Wasserstein distances for the original and imputed data.
    \end{tabular}
    \caption{Procedure for deriving the sliced Wasserstein discrepancy statistics.}
    \label{fig:projections}
\end{figure}
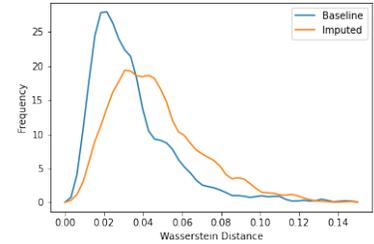

\noindent 
Over all $(r,p)$ pairs, these steps result in two distributions of distances, for $w(r, p)$ and $\hat w(r, p)$, as illustrated in Figure~\ref{fig:projections}. Firstly, these can be regarded as probability distributions, allowing us to compute the same class B feature-wise discrepancy scores discussed previously.
Secondly, we evaluate the relative change in the Wasserstein distance due to imputation for each $r$ and~$p$, namely $\hat w(r,p)/w(r,p)$, allowing us to quantify how much different imputation methods induce discrepancy in the data distribution. Finally, we assess the stability of the different imputation methods by exploring the variance in the induced sliced Wasserstein distance across repeated imputations. 

\begin{comment}
\begin{figure}[htb!]
\centering
\captionsetup[subfigure]{oneside,margin={-1cm,0cm}}
%\subfloat[Classifier dependence]{
   \includegraphics[width=0.80\textwidth]{figs/Projected_w_dist_flowchart.pdf}
    \caption{\textbf{Work flow of calculating the projected Wasserstein distances}}
         \label{fig:projections}%}
\end{figure}
\end{comment}

\subsection{Downstream effects of imputation quality}

\noindent
A critical question that this paper aimed to answer is: in what way does the quality of the data imputation affect the downstream classification performance?
Firstly, we determined whether there is a correlation between the discrepancy scores in classes A--C and the performance of the predictive model. Secondly, we investigated whether a classification model, trained using poorly imputed data, could use unexpected features to make its predictions. To answer this, we analyse the models fit to the \textbf{Simulated} dataset. This is an ideal dataset, as the features are of equal importance by design, the clusters of values within each feature are normally distributed, centered at vertices of a hypercube and separated\cite{pedregosa_scikit-learn_2011}. 

For each classifier, we identify two models which perform well at the classification task but where one is trained using poorly imputed data and the other is fit to data which is imputed well. See the Supplementary Materials for details on the model selection. To assess the feature importance for each of the trained models, we used the popular Shapley value approach of Lundberg et al.\cite{lundberg_unified_2017}. For each feature, this assigns an importance to every value, identifying how the prediction is changed on the basis of the value of this feature. A positive importance value indicates the value influences the model towards a positive prediction. Due to the design of the \textbf{Simulated} dataset, we expect the distribution of the Shapley values to be symmetric, as the clusters are separated by a fixed distance and are normally distributed within those clusters. Using the skewness, we measure the symmetry of the feature importances for each feature. The Python package, for computing the Shapley values, cannot currently support neural network and logistic regression models, but we consider them for all other models.

\section{Results}

\begin{figure}[htb!]
\centering
\captionsetup[subfigure]{oneside,margin={-1cm,0cm}}
\subfloat[Classifier dependence]{
   \includegraphics[width=0.8\textwidth]{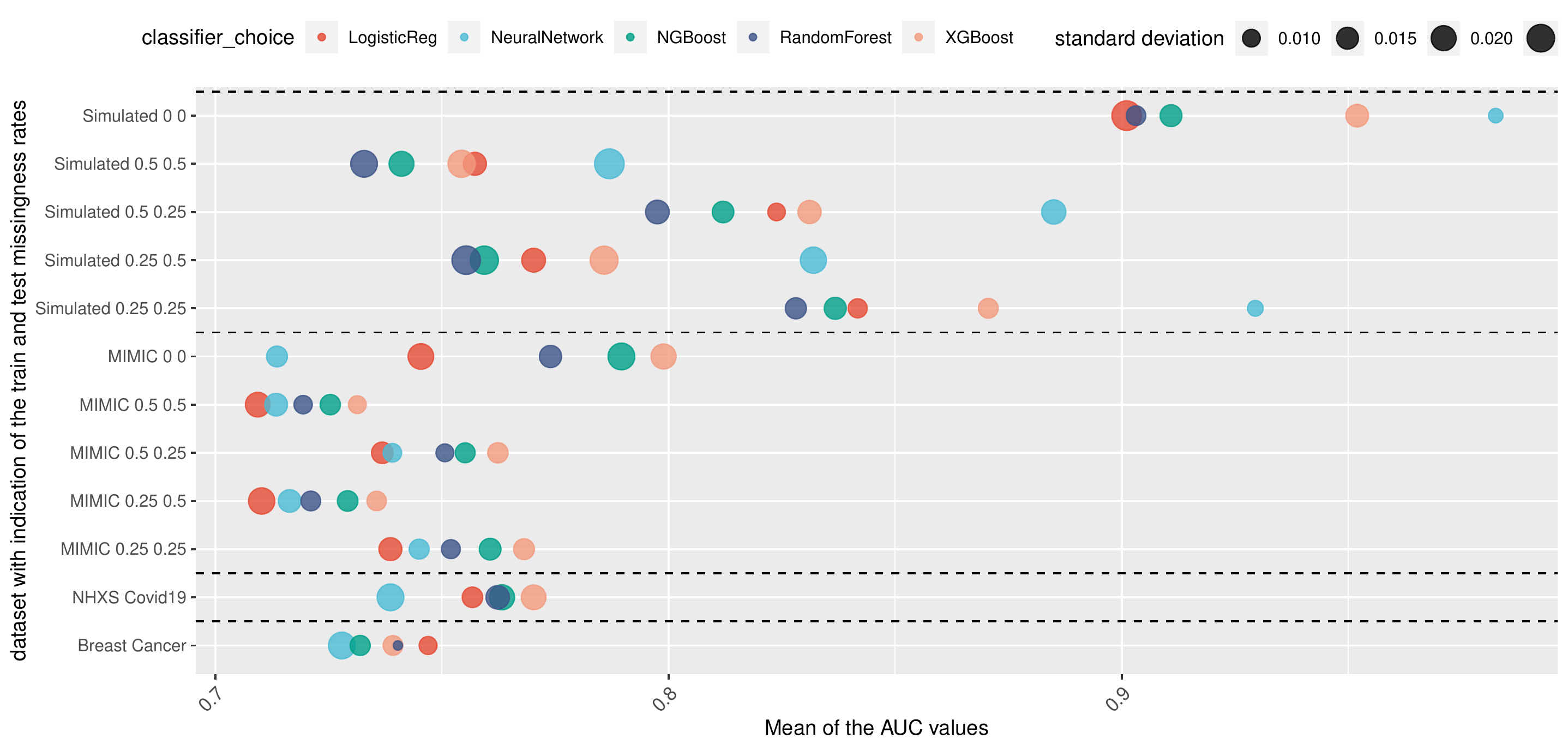}
   %{figs/buble_plt_cls_complete_dt_corrected.pdf}
         \label{result1a}}
        % \newline
     %\end{subfigure}
     \qquad
\subfloat[Imputation dependence]{
    \centering
       \hspace{-0.8cm}\includegraphics[width=0.8\textwidth]{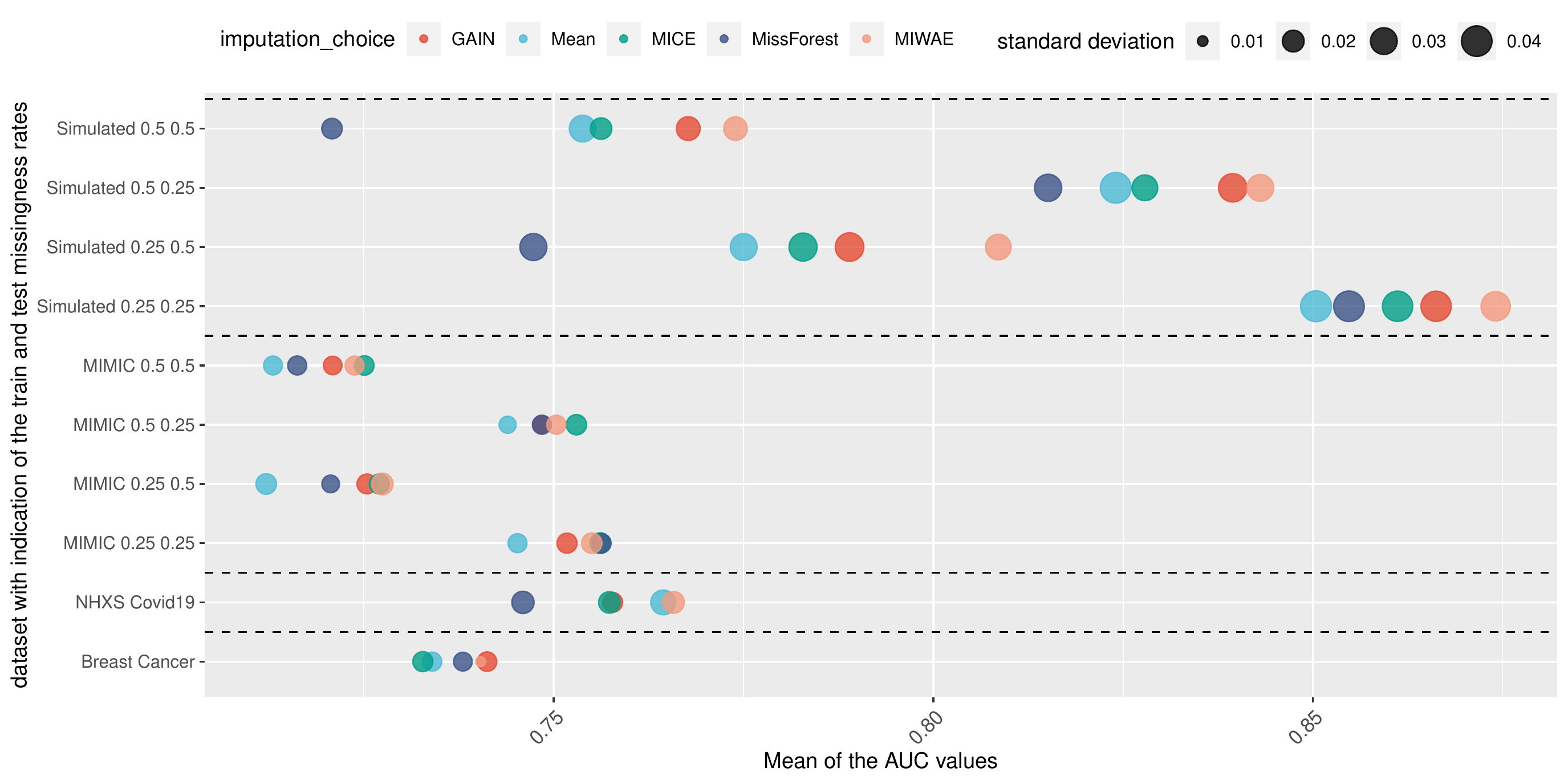}%{figs/Buble_plot_imputation.png}
         %\caption{$y=x$}
         \label{result1b}}

 \caption{\textbf{Dependence of downstream classification performance on classification and imputation methods.} These plots show the dependence of the downstream performance on the classification and imputation methods. For the \textbf{Simulated} and \textbf{MIMIC-III} datasets we show performance for 25\% and 50\% levels of missingness in the development and test data. The size of each marker corresponds to the standard deviation\label{fig:results1}}
\end{figure}

\noindent
\textbf{Classifier influence on downstream performance.}
In Figure~\ref{fig:results1}(a), we show how the classifier affects the downstream performance for each dataset across different train and test missingness rates.
For the \textbf{MIMIC-III} and \textbf{Simulated} datasets, the classifier trained using the complete original dataset has performance which always exceeds that of those built on imputed data (with the sole exception of the Neural Network for \textbf{MIMIC-III}).
The performances for the \textbf{Simulated} dataset are much higher than for the real datasets as, by design, there is a direct link between outcome and the feature values. For the \textbf{Simulated} and \textbf{MIMIC-III} datasets, we see that increasing the train and test missingness rates leads to a decline in performance. 
For all classifiers, with a fixed train missingness rate, a change in the test missingness rate affects performance very significantly compared to changing the train missingness rate for a fixed test missingness rate. 
For each imputed dataset, we see that when varying the missingness rates of the development and test data, the ranking of each classifier's performance is almost consistent, e.g.\ for the \textbf{Simulated} dataset the Neural Network always performs the best while Random Forest performs the worst. The XGBoost classifier performs best for the \textbf{MIMIC-III} and \textbf{NHSX COVID-19} datasets, with consistency in performance ranking at all missingness rates. 
The \textbf{Breast Cancer} dataset follows the trend of the other real-world datasets, with the exception that the logistic regression classifier performs the best. We see that increasing the test missingness rate leads to an increase in the standard deviation. For most datasets, the variance of the classifier performance is similar across the classifiers, the exception being the \textbf{Breast Cancer} dataset for which the Random Forest has a small variance and Neural Network a relatively large variance.\\

\noindent
\textbf{Imputation influence on downstream performance.}
In Figure~\ref{fig:results1}(b), we show the dependence of the downstream classification performance on the imputation methods used to generate the complete datasets. For the \textbf{Simulated} and \textbf{NHSX COVID-19} datasets, MIWAE imputation gives the best downstream performance for all levels of development and test missingness rates. For the \textbf{Simulated} dataset, as the missingness rate increases, the performance of models trained using MissForest imputed data significantly declines.
In the real-world datasets, there is no `best' imputation method that consistently leads to a model which outperforms the others. For the \textbf{MIMIC-III} dataset, we see that the simple mean imputation method gives the worst downstream performance for all missingness rates but for \textbf{NHSX COVID-19} is competitive with the best performing MIWAE method. For the \textbf{MIMIC-III} dataset, MICE imputation leads to the best performance for most missingness rates but is the worst for the \textbf{Breast Cancer} dataset. The variance in the performances is consistent across the real-world datasets, in the range $[0.01, 0.02]$, but is significantly higher for all levels of missingness in the \textbf{Simulated} dataset, in the range of $[0.02, 0.04]$.\\

\begin{figure}[htb!]
     \centering
\subfloat[Pooled ANOVA analysis]{
         \centering
       \includegraphics[width=0.36\textwidth]{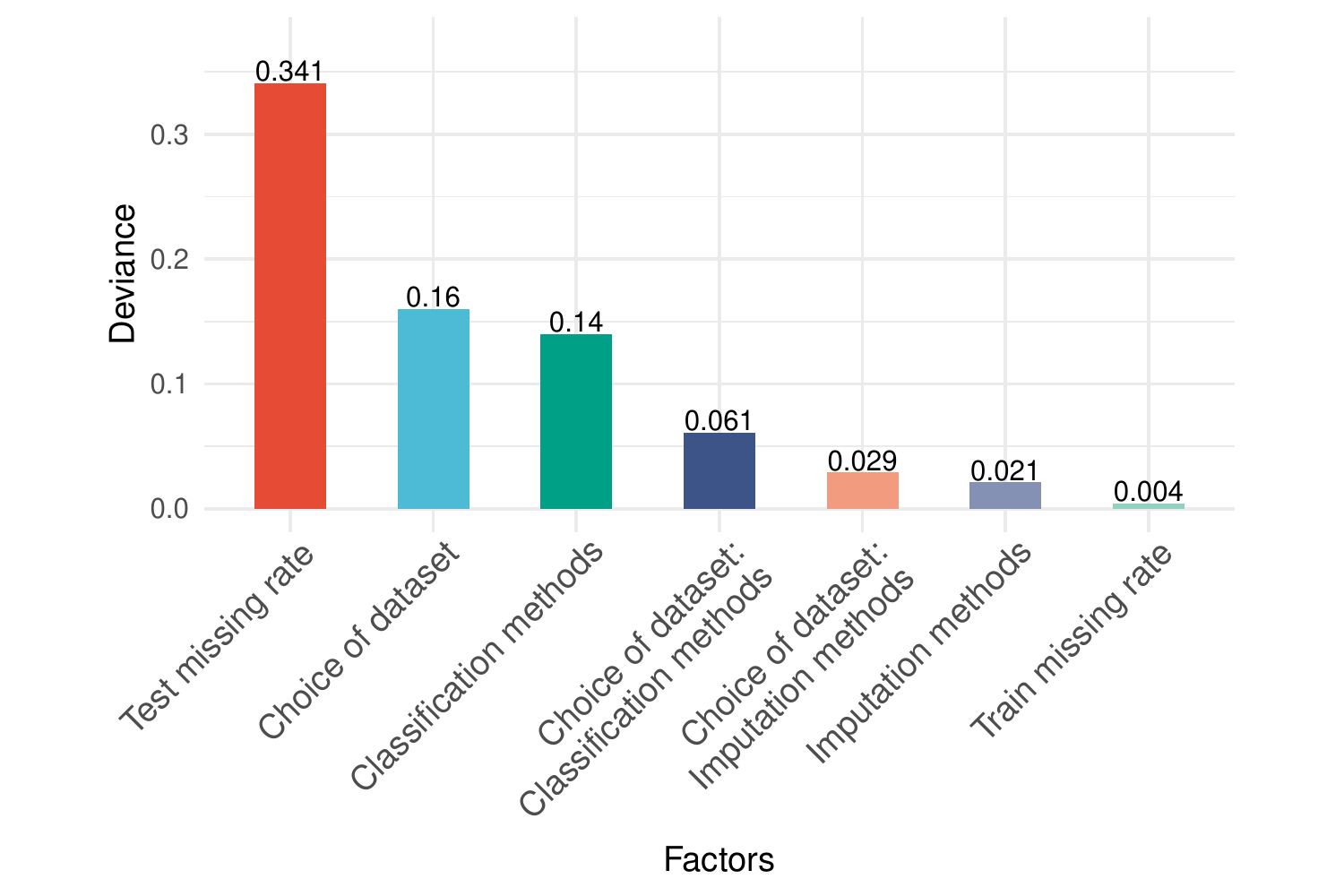}
         } \qquad
\subfloat[\textbf{Simulated} dataset]{
         \centering
       \includegraphics[width=0.46\textwidth]{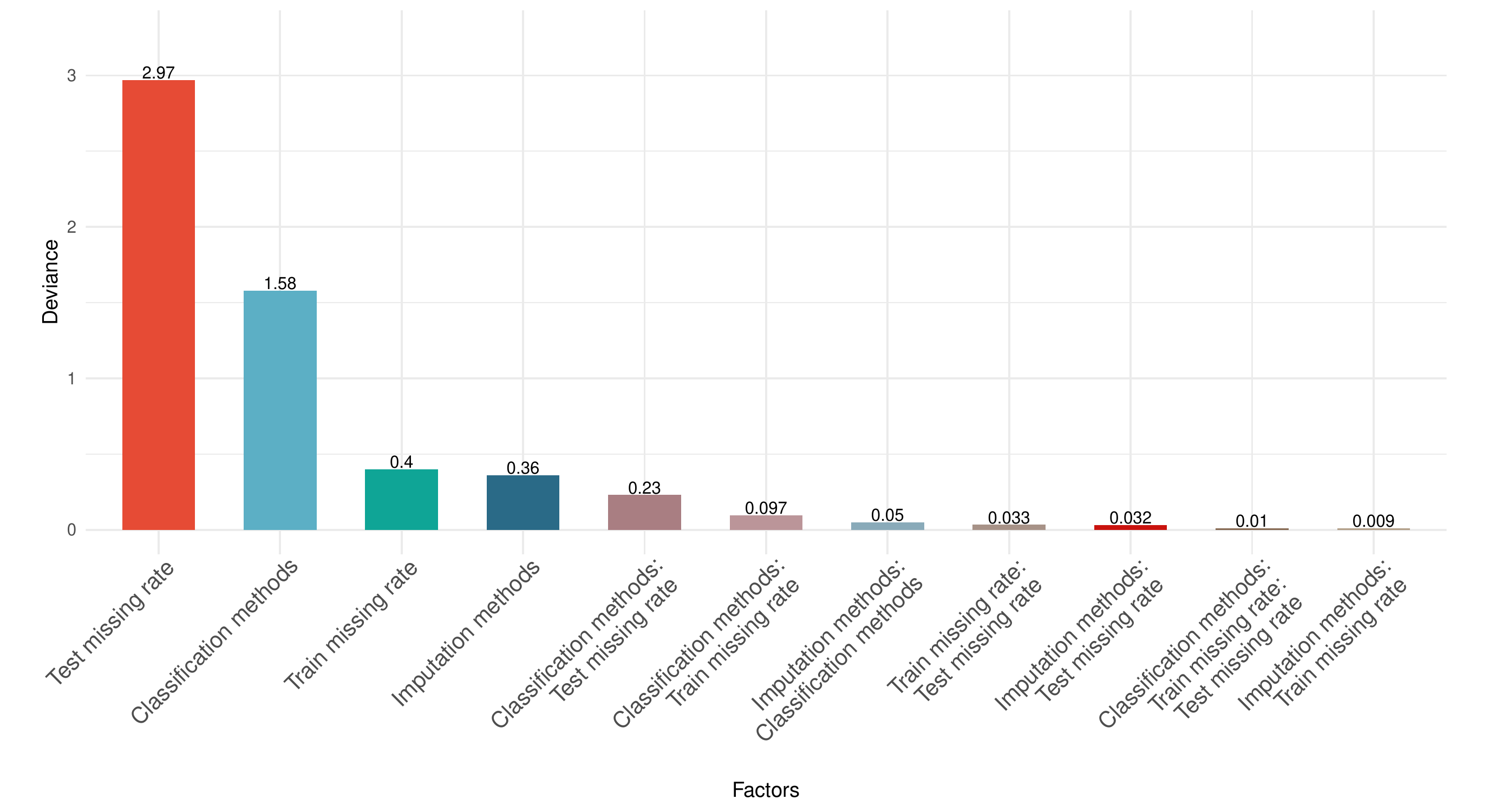}
         } \qquad
\subfloat[\textbf{MIMIC-III} dataset]{
   \includegraphics[width=0.46\textwidth]{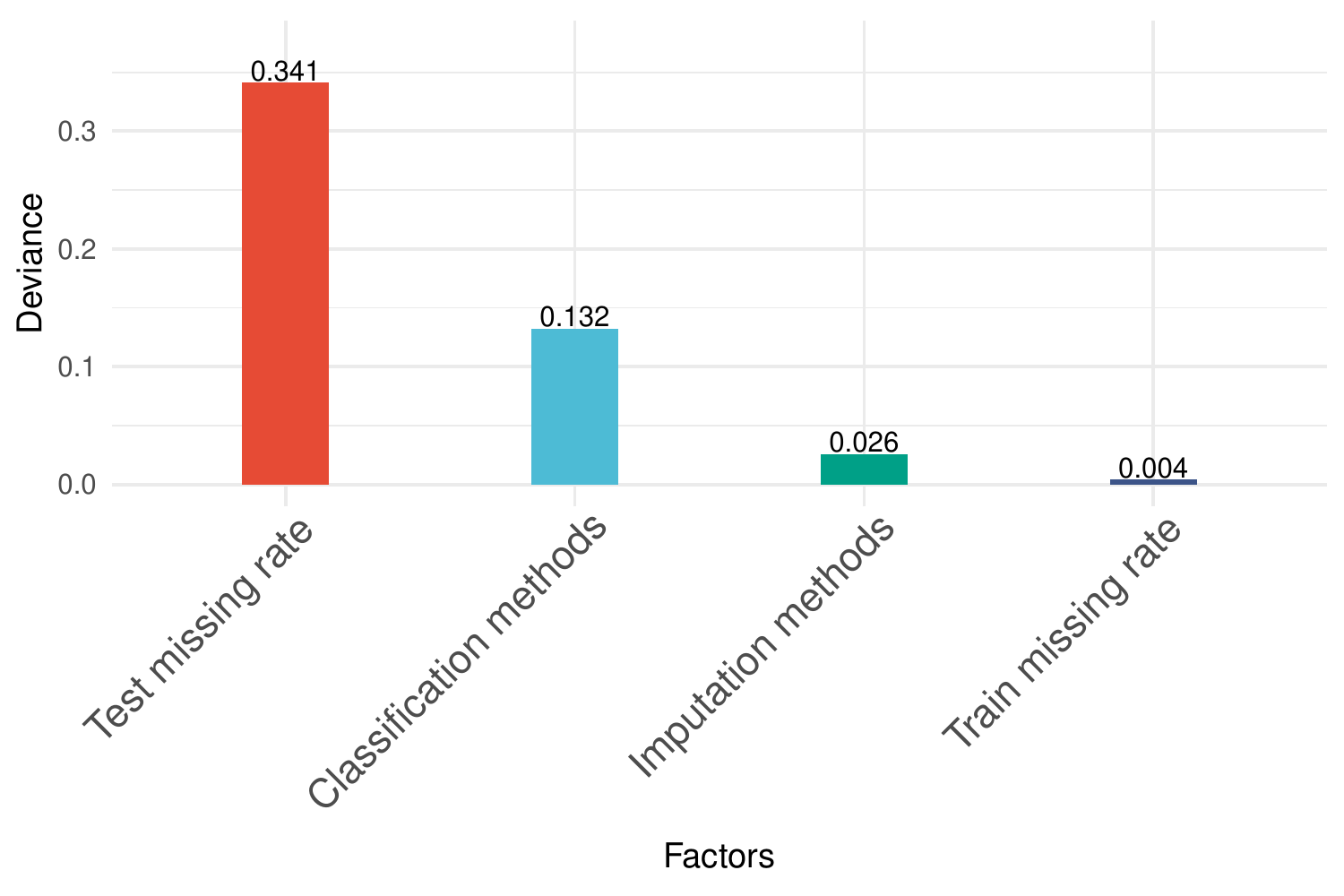}
   }\qquad
\subfloat[\textbf{NHSX COVID-19} dataset]{
         \centering
       \includegraphics[width=0.46\textwidth]{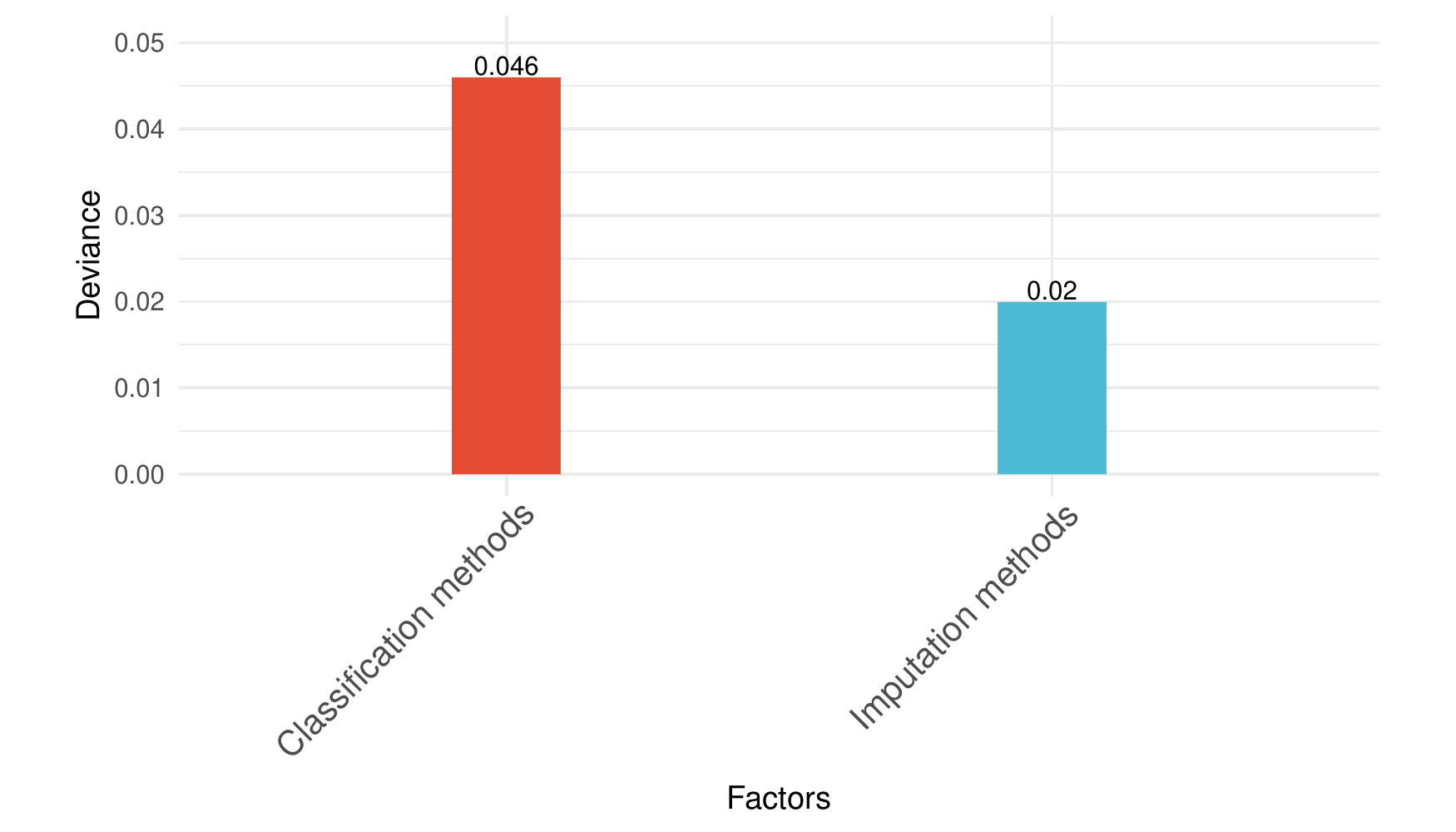}
         %\caption{$y=x$}
         %\label{fig:result3b}
         }
 \caption{\textbf{Pooled and dataset-segregated ANOVA analysis.} In these plots we show the significant factors in the ANOVA analysis for the pooled and segregated datasets (we do not display for \textbf{Breast Cancer}, as the choice of the classifier is the only significant factor.). \label{fig:results3}}
\end{figure}

\noindent
\textbf{ANOVA for downstream classification performance.}
In Figure~\ref{fig:results3}, we show the significant factors identified for the pooled data and individual datasets.
For the pooled data, it is clear that the missingness rate of the test set explains most of the deviance in the results of our predictive models, confirming the observations previously derived from Figure~\ref{fig:results1}. The dataset under consideration and the classification method used are the next most important factors.
The ANOVA is also performed for the individual datasets, shown in Figure~\ref{fig:results3}(b)--(d). For the \textbf{Simulated} dataset, we see that many factors affect the classification performance. 
Only the \textbf{Simulated} and \textbf{MIMIC-III} datasets have induced missingness, and we see that the test set missingness rate is the most significant source of deviance in the downstream classification performance, followed by the choice of classification method. For both the \textbf{NHSX COVID-19} and \textbf{Breast Cancer} datasets, the classification method is the primary source of deviance. Indeed for the \textbf{Breast Cancer} dataset, it is the only significant factor at the 1\% level.\\

\noindent
\textbf{Comparing imputation quality}\\

\begin{figure}[htb!]
    \centering
    \begin{tabular}{M{5cm} | M{5cm} | M{5cm}}
     \textbf{RMSE} & \textbf{MAE} & \textbf{R2} \\
    \hline
      \includegraphics[height=4cm,width=5cm]{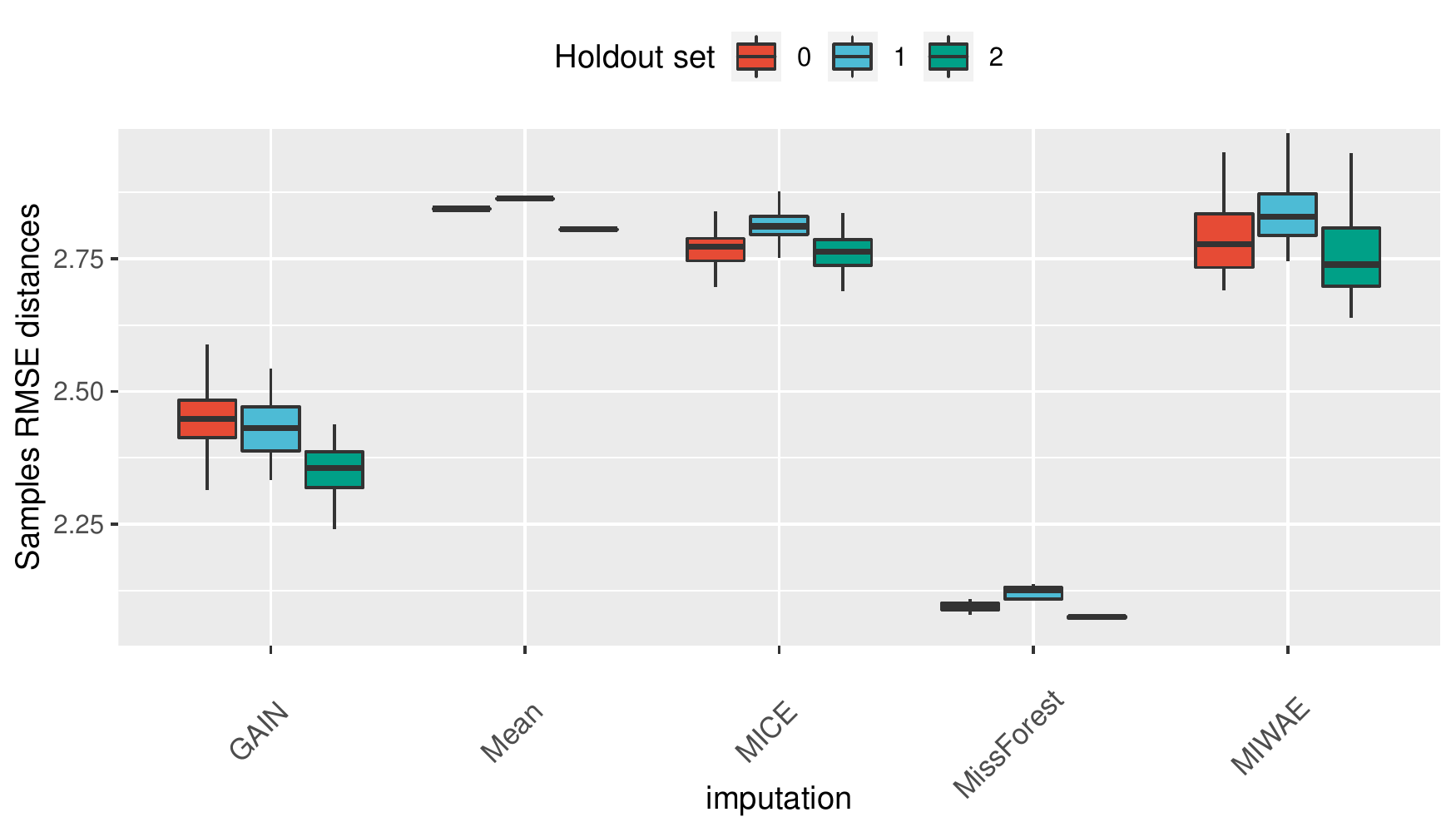}&
      \includegraphics[height=4cm,width=5cm]{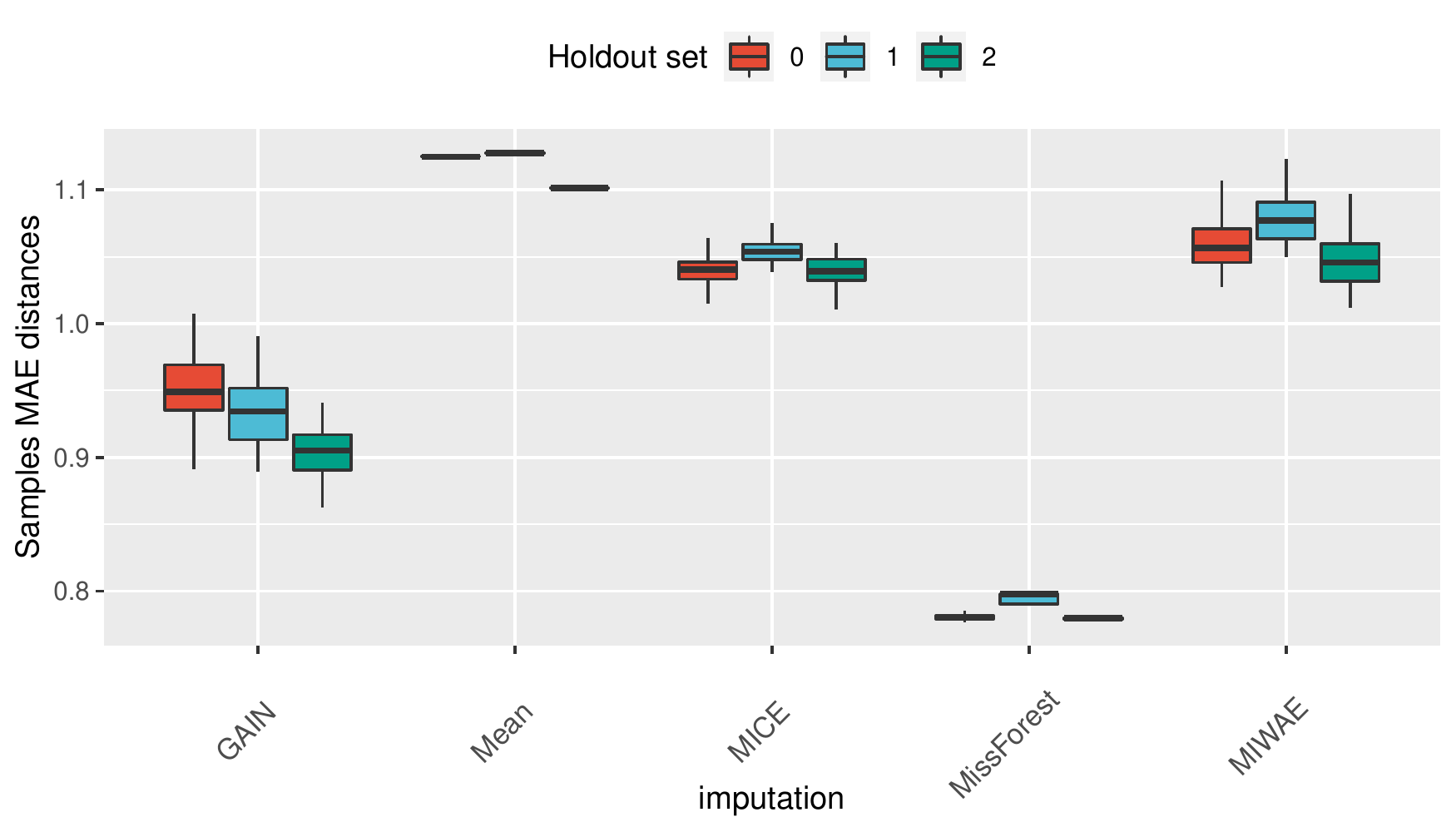}&
      \includegraphics[height=4cm,width=5cm]{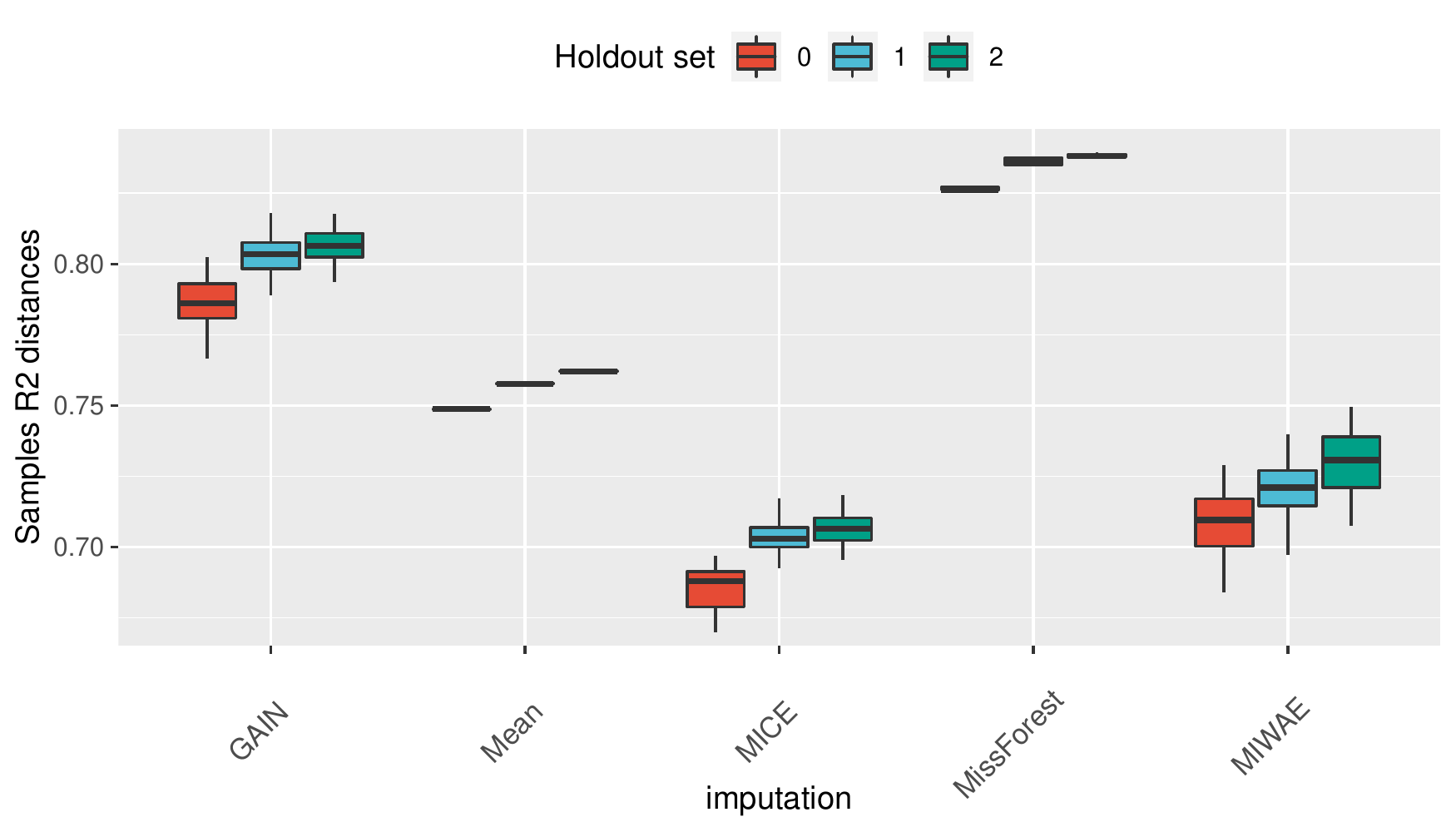}\\
    \end{tabular}
    \caption{The sample-wise statistics for the \textbf{MIMIC-III} dataset with 25\% train and test missingness rates.}
    \label{fig:samplewise}
\end{figure}

\begin{figure}[htb!]
    \centering
    \begin{tabular}{m{0.2in} | M{5cm} | M{5cm} | M{5cm}}
    & \textbf{RMSE} & \textbf{MAE} & \textbf{R2} \\
    \hline
    \parbox[c][][c]{0.5in}{\rotatebox[origin=t]{90}{Train 25 \%, Test 25\%}} &
      \includegraphics[height=4cm,width=5cm]{MIMIC_samplewise_dist/MIMIC_0.25_0.25_RMSE.pdf}&
      \includegraphics[height=4cm,width=5cm]{MIMIC_samplewise_dist/MIMIC_0.25_0.25_MAE.pdf}&
      \includegraphics[height=4cm,width=5cm]{MIMIC_samplewise_dist/MIMIC_0.25_0.25_R2.pdf}\\
    \hline
        \parbox[c][][c]{0.5in}{\rotatebox[origin=t]{90}{Train 50 \%, Test 25\%}} &
      \includegraphics[height=4cm,width=5cm]{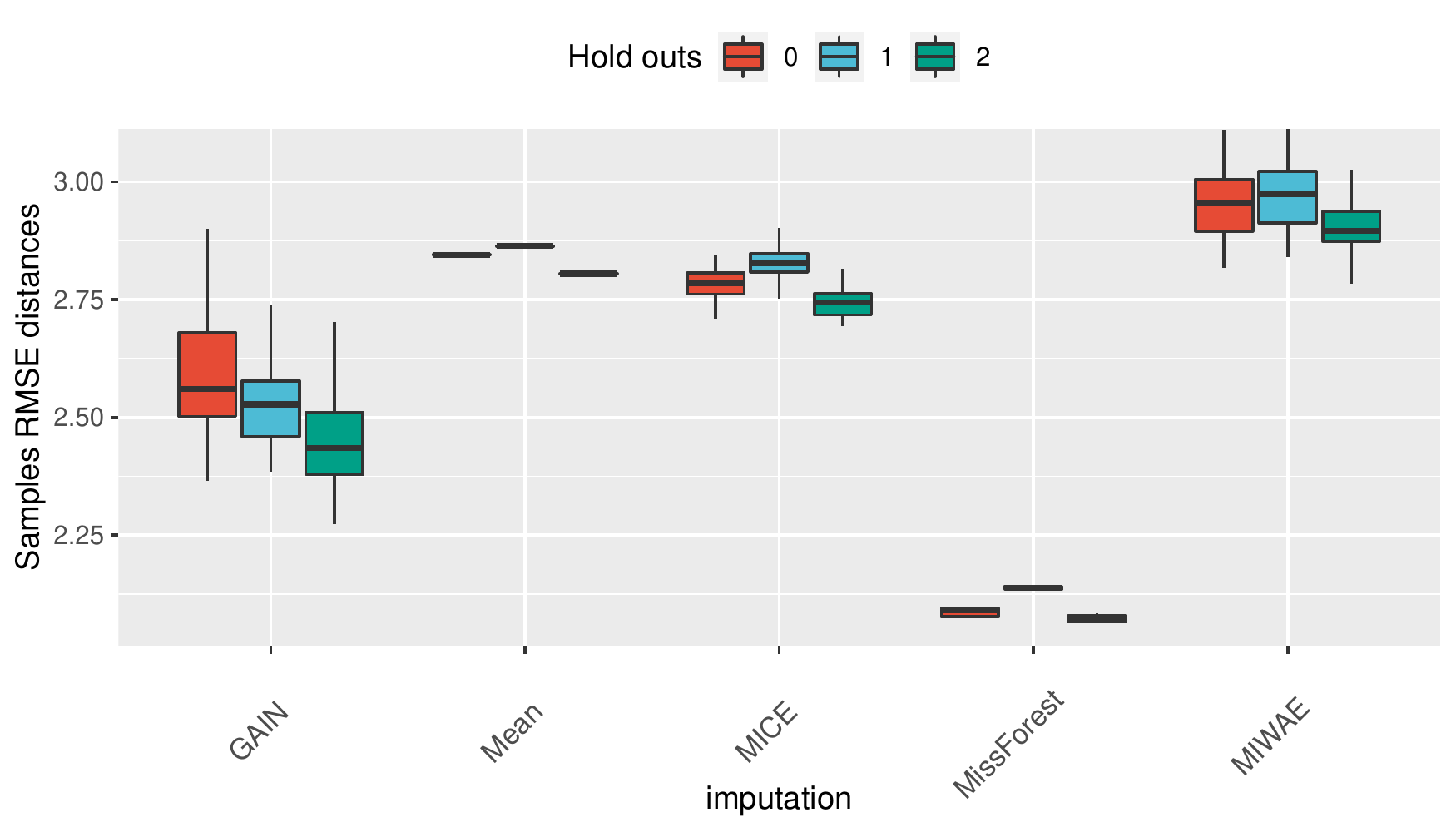}&
      \includegraphics[height=4cm,width=5cm]{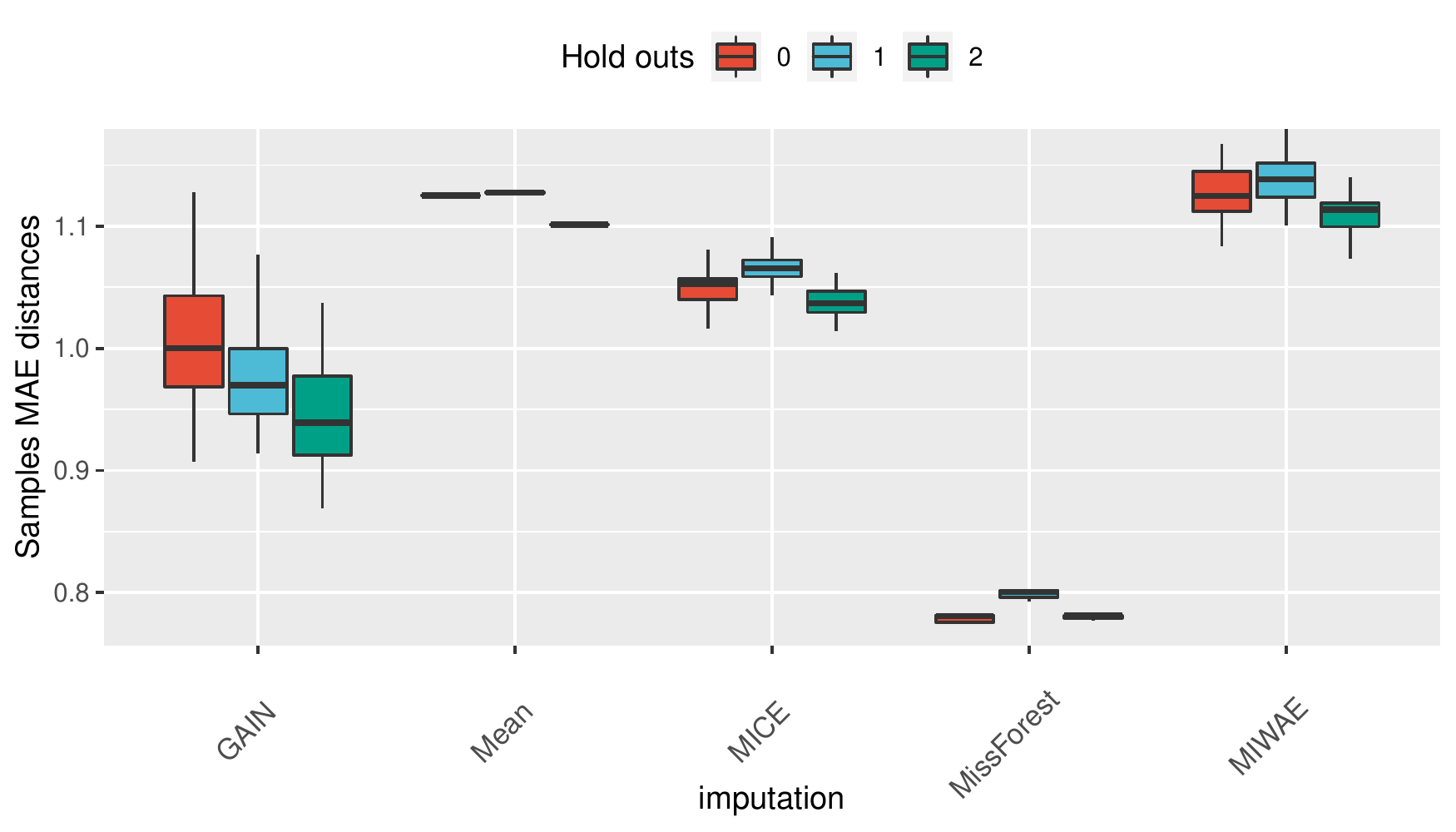}&
      \includegraphics[height=4cm,width=5cm]{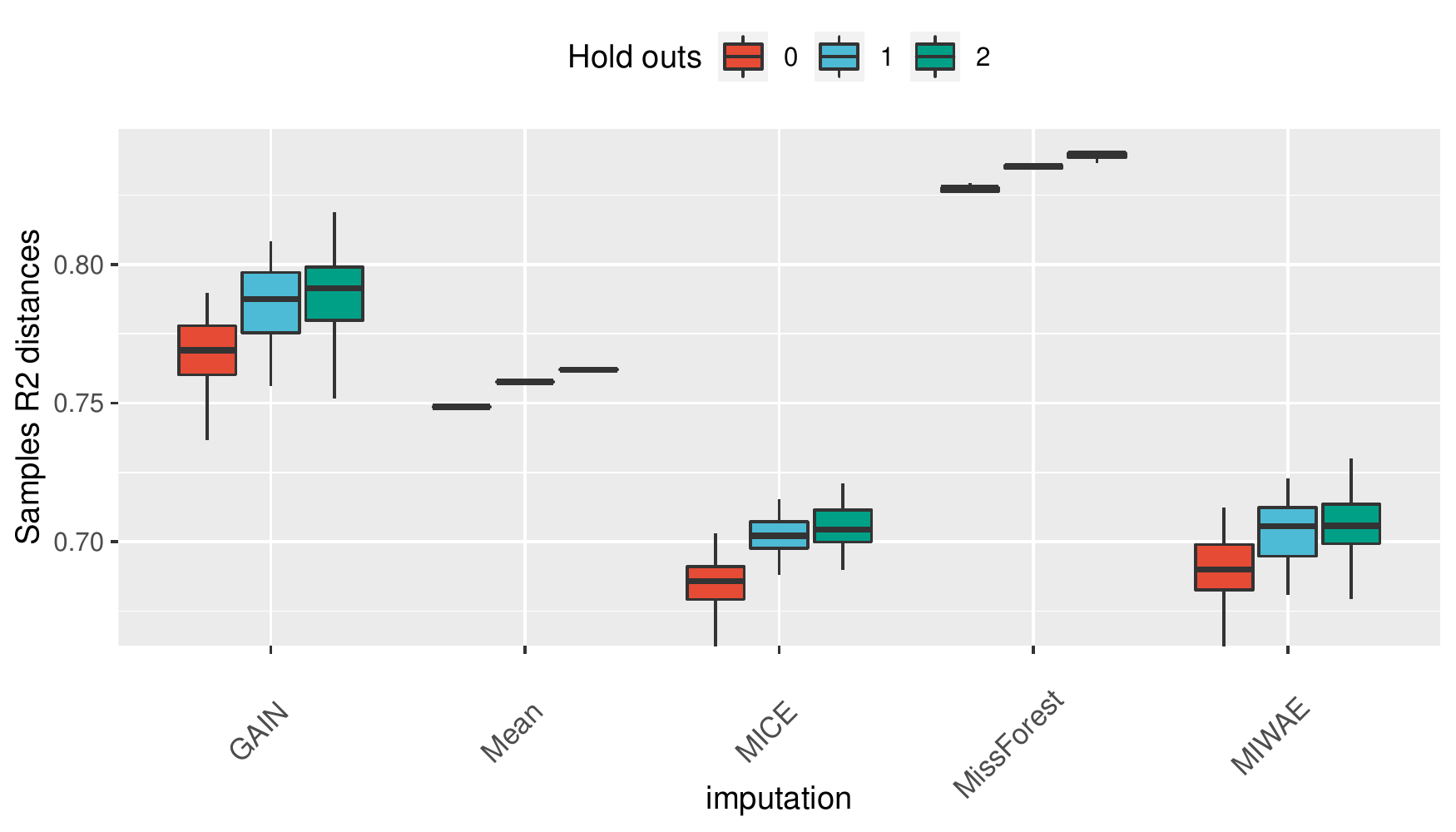}\\
    \hline
    \parbox[c][][c]{0.5in}{\rotatebox[origin=t]{90}{Train 25 \%, Test 50\%}} &
      \includegraphics[height=4cm,width=5cm]{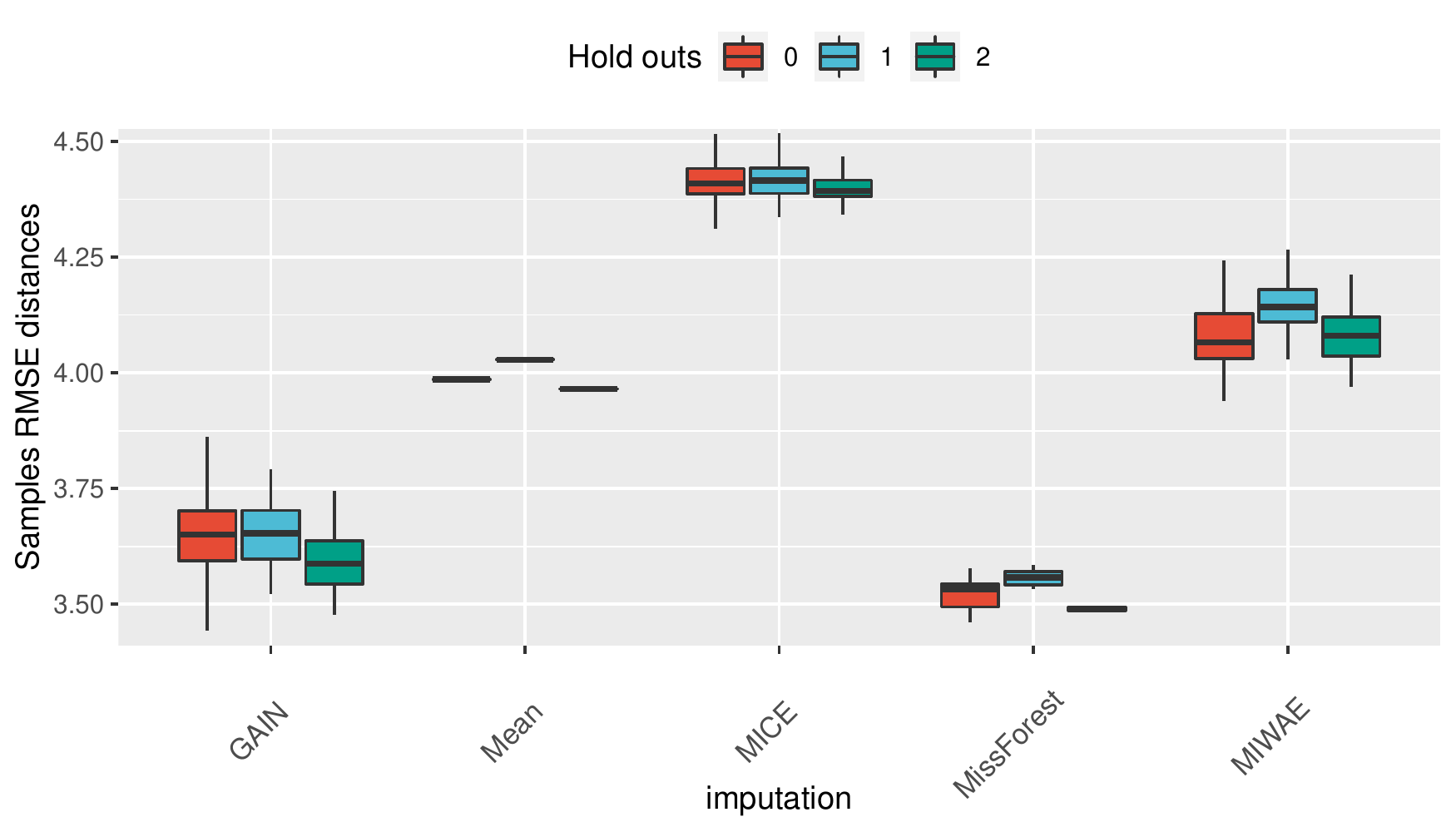}&
      \includegraphics[height=4cm,width=5cm]{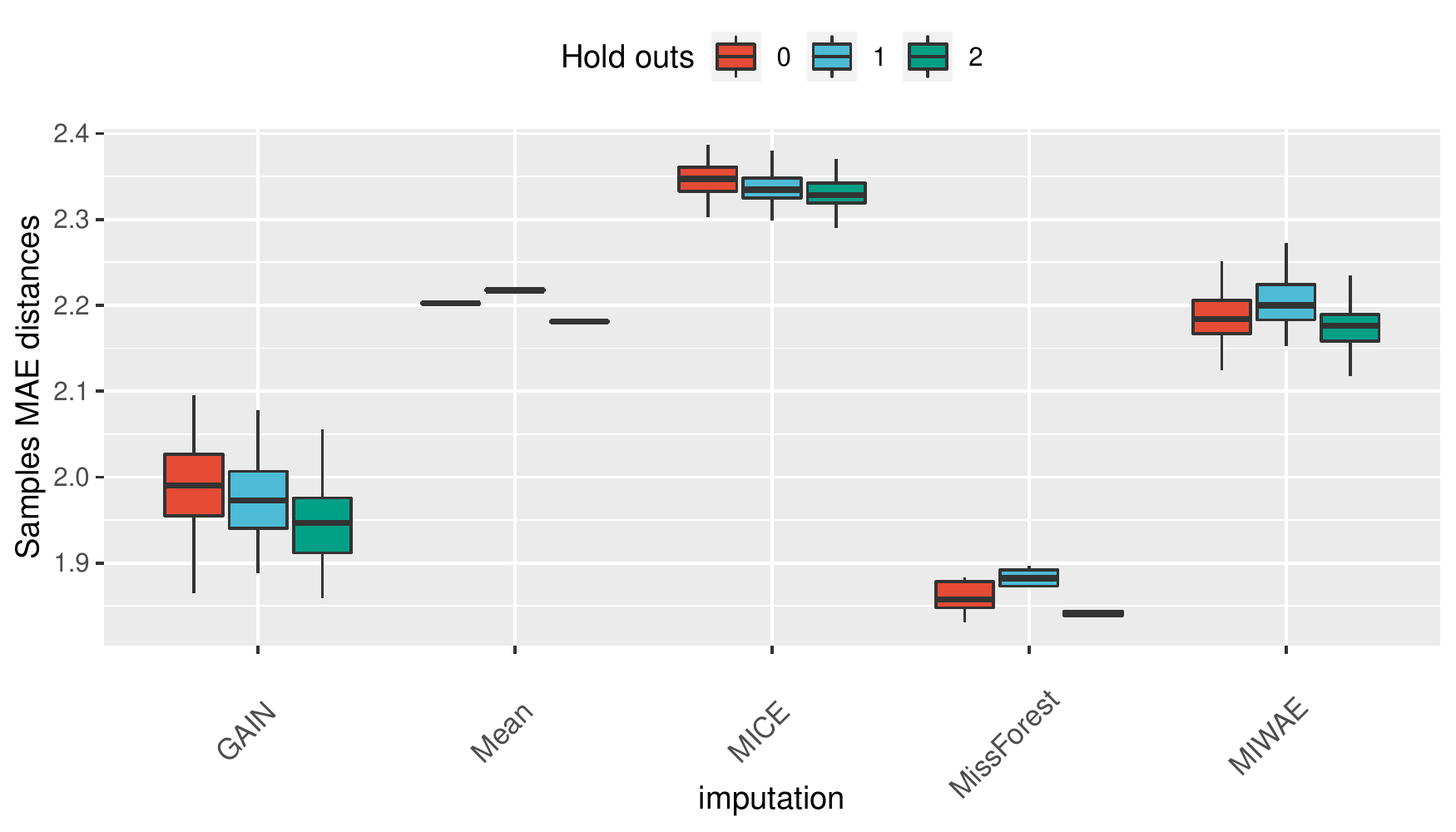}&
      \includegraphics[height=4cm,width=5cm]{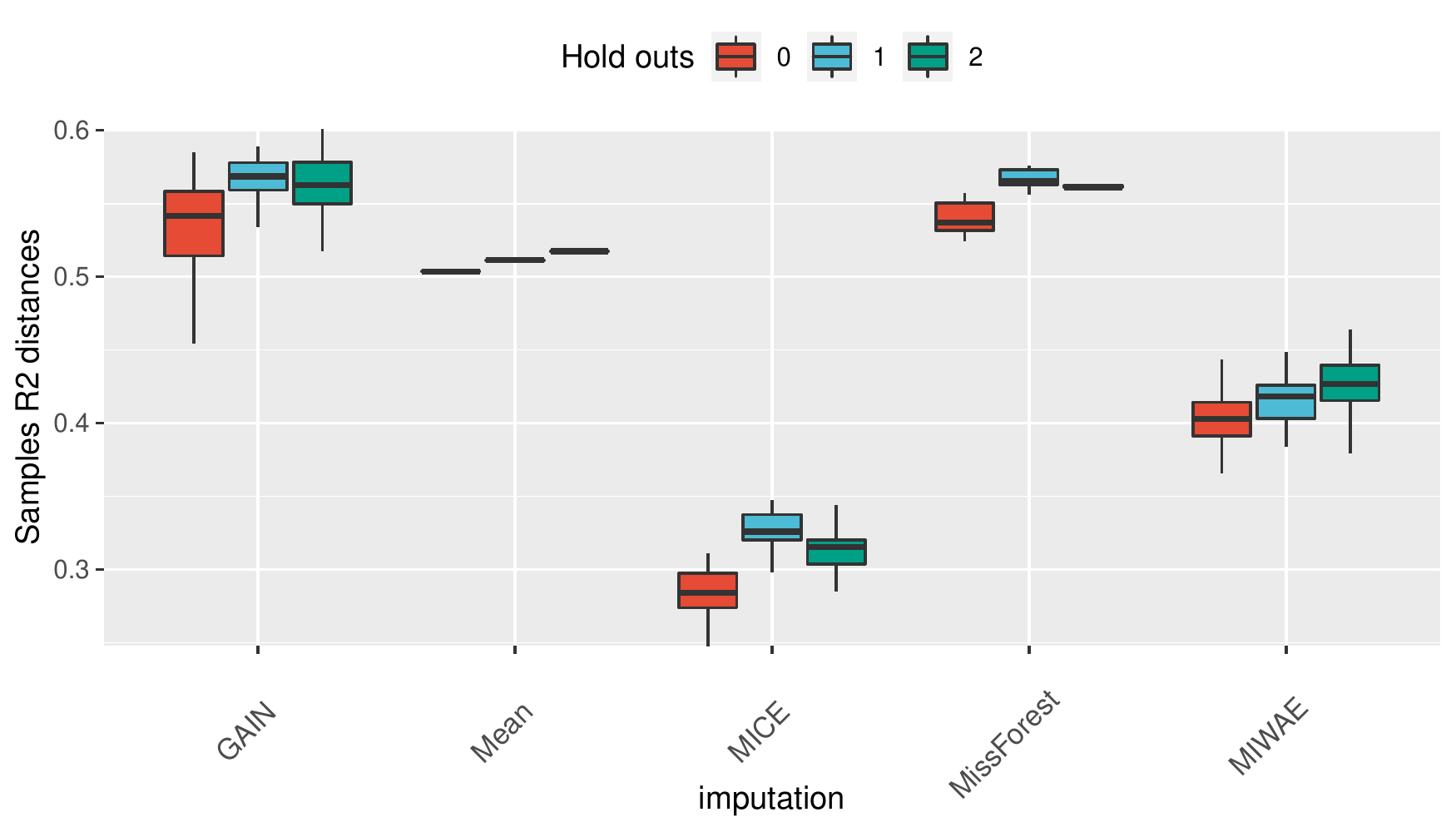}\\
    \hline
    \parbox[c][][c]{0.5in}{\rotatebox[origin=t]{90}{Train 50 \%, Test 50\%}} &
      \includegraphics[height=4cm,width=5cm]{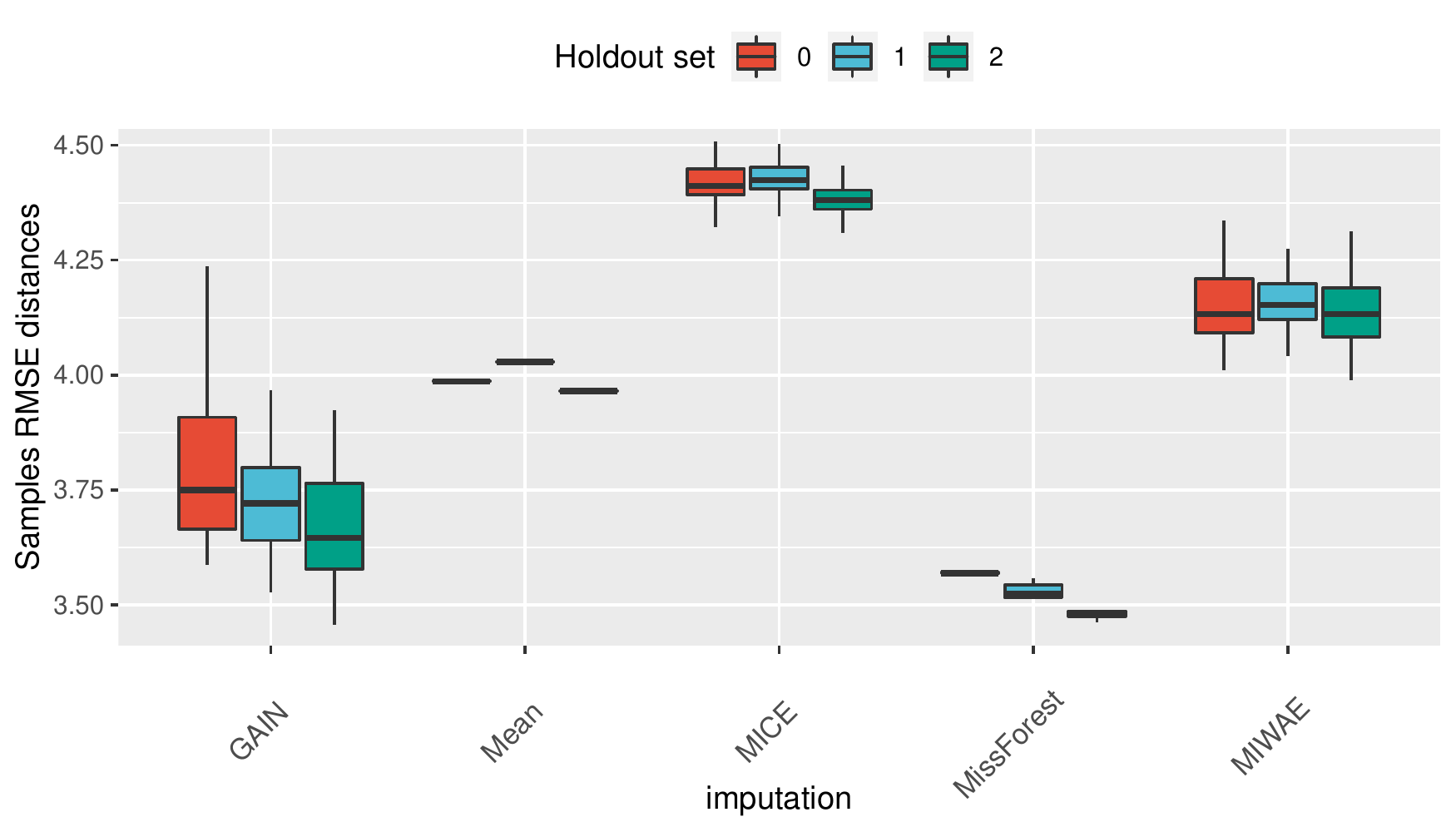}&
      \includegraphics[height=4cm,width=5cm]{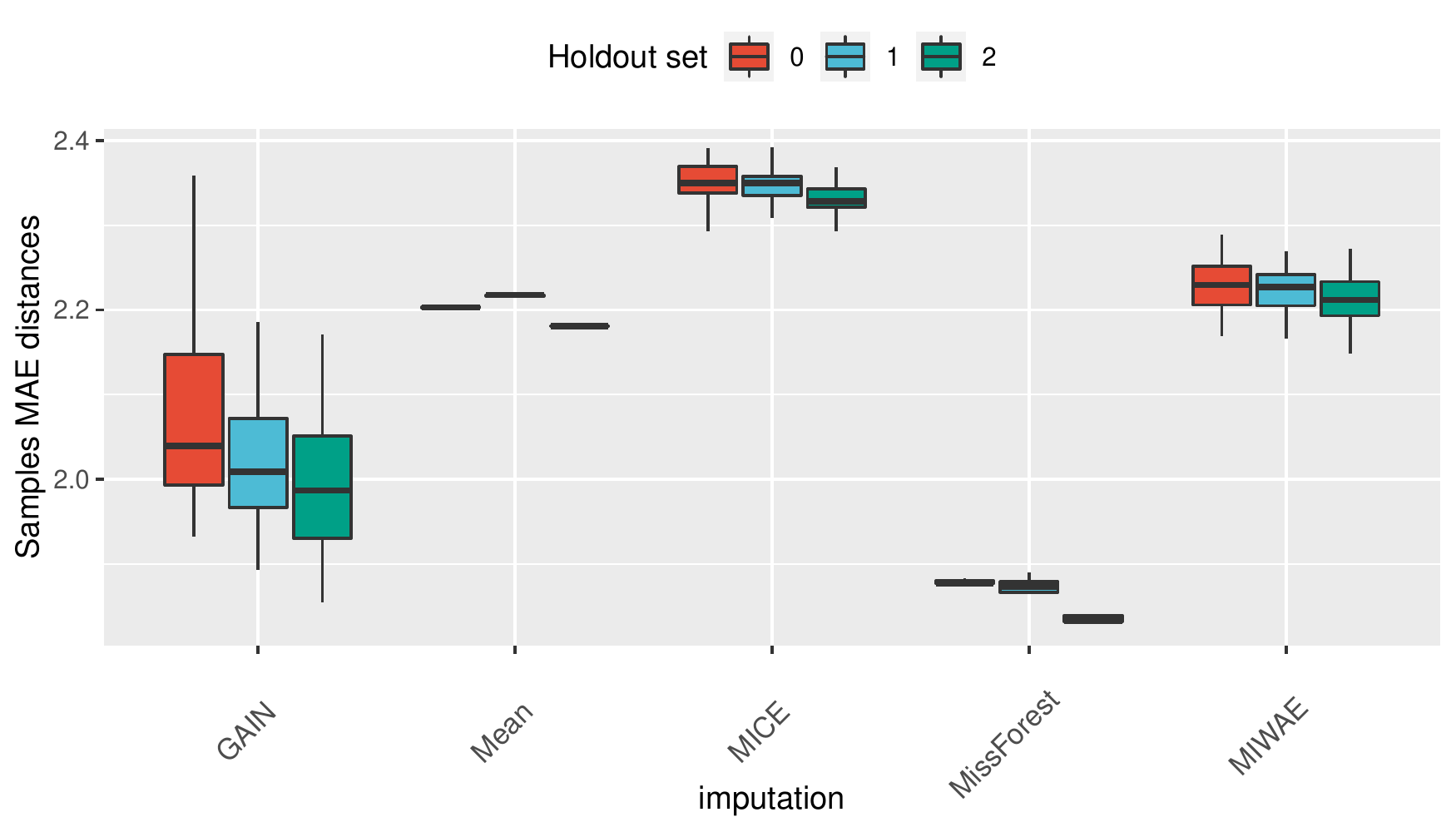}&
      \includegraphics[height=4cm,width=5cm]{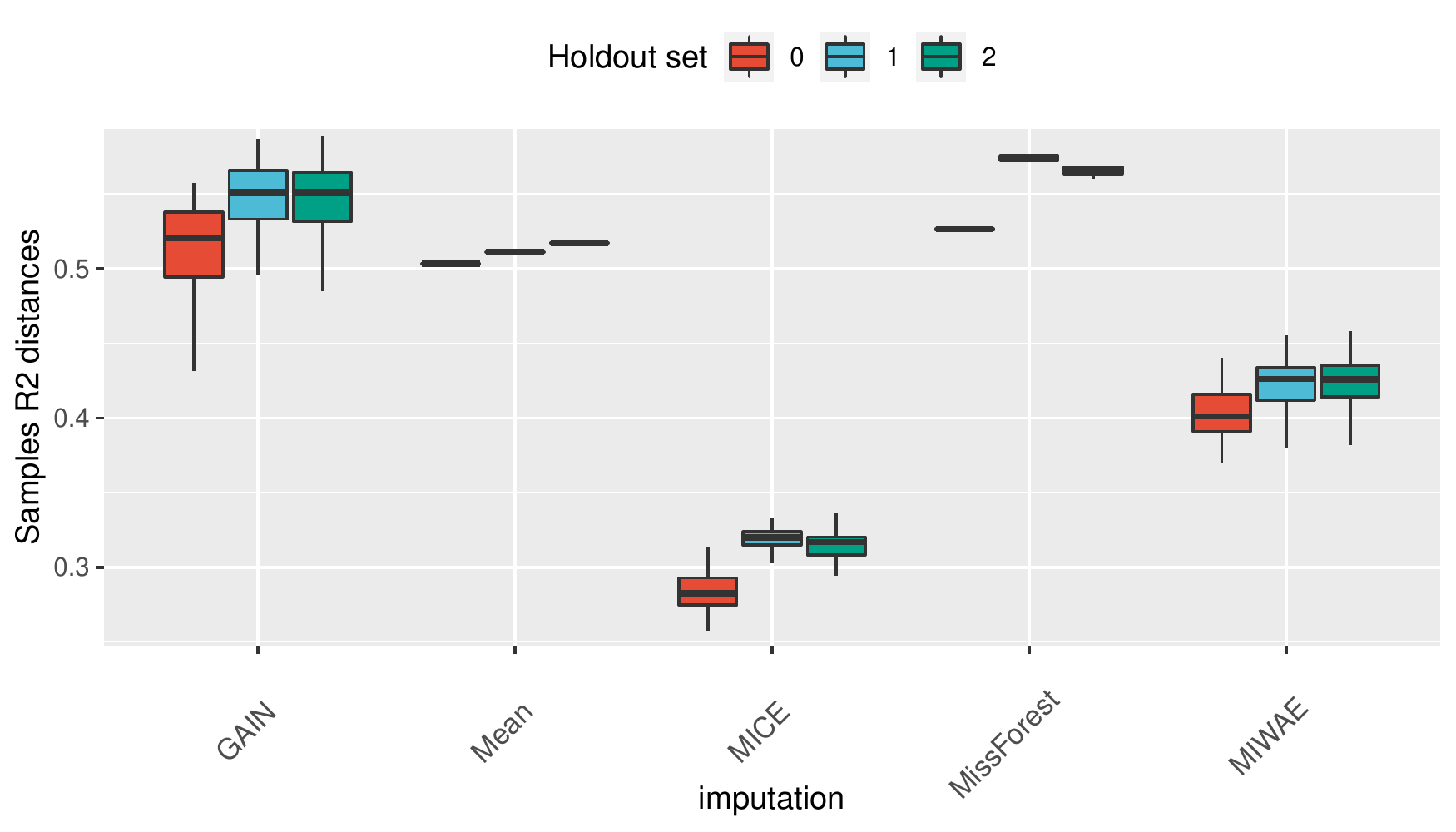}\\
    \end{tabular}
    \caption{The sample-wise statistics for the \textbf{MIMIC-III} dataset at the different train and test missingness rates considered.}
    \label{fig:samplewise_mimic}
\end{figure}

\begin{figure}[htb!]
    \centering
    \begin{tabular}{m{0.2in} | M{5cm} | M{5cm} | M{5cm}}
    & \textbf{RMSE} & \textbf{MAE} & \textbf{R2} \\
    \hline
    \parbox[c][][c]{0.5in}{\rotatebox[origin=t]{90}{Train 25 \%, Test 25\%}} &
      \includegraphics[height=4cm,width=5cm]{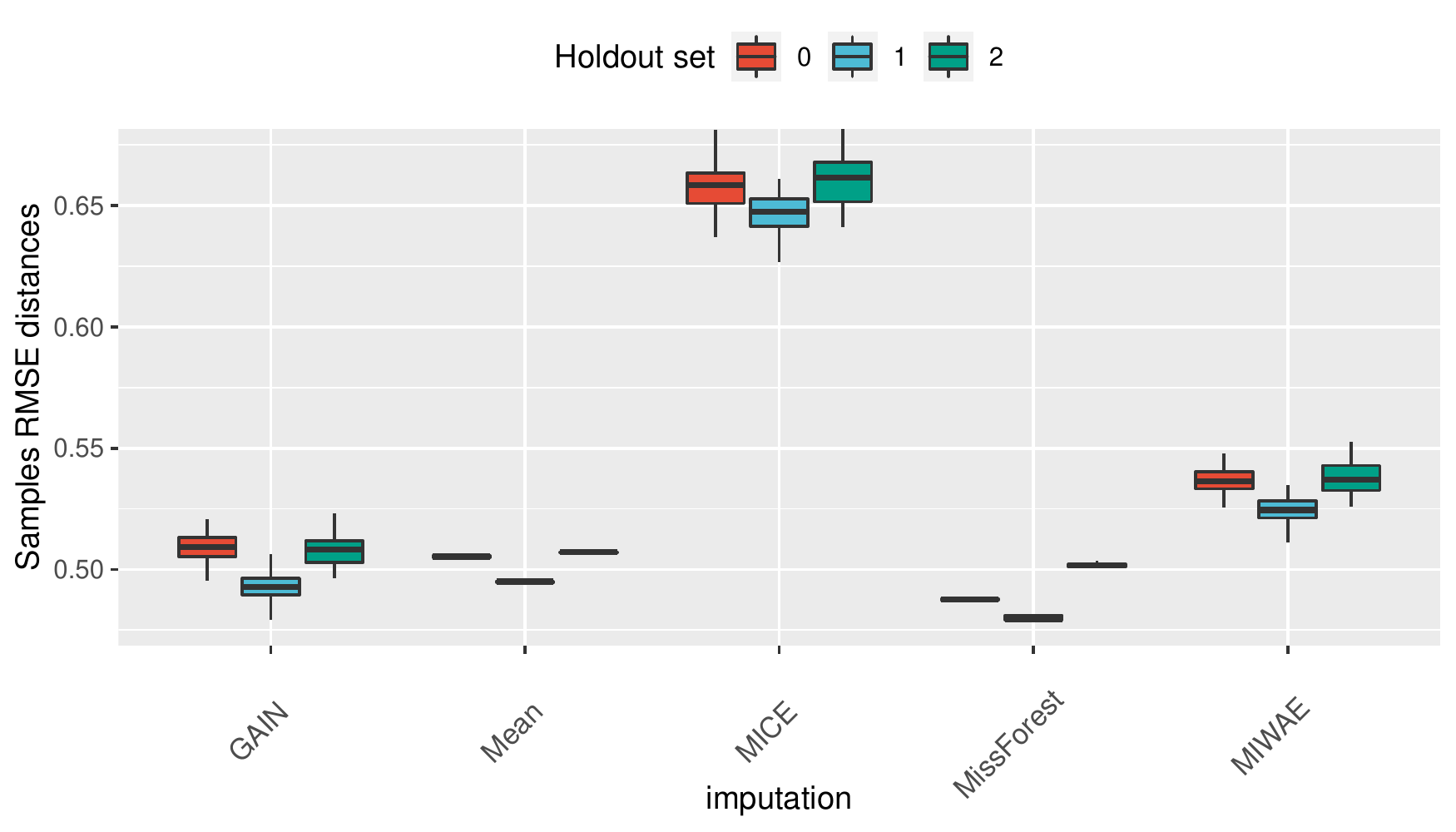}&
      \includegraphics[height=4cm,width=5cm]{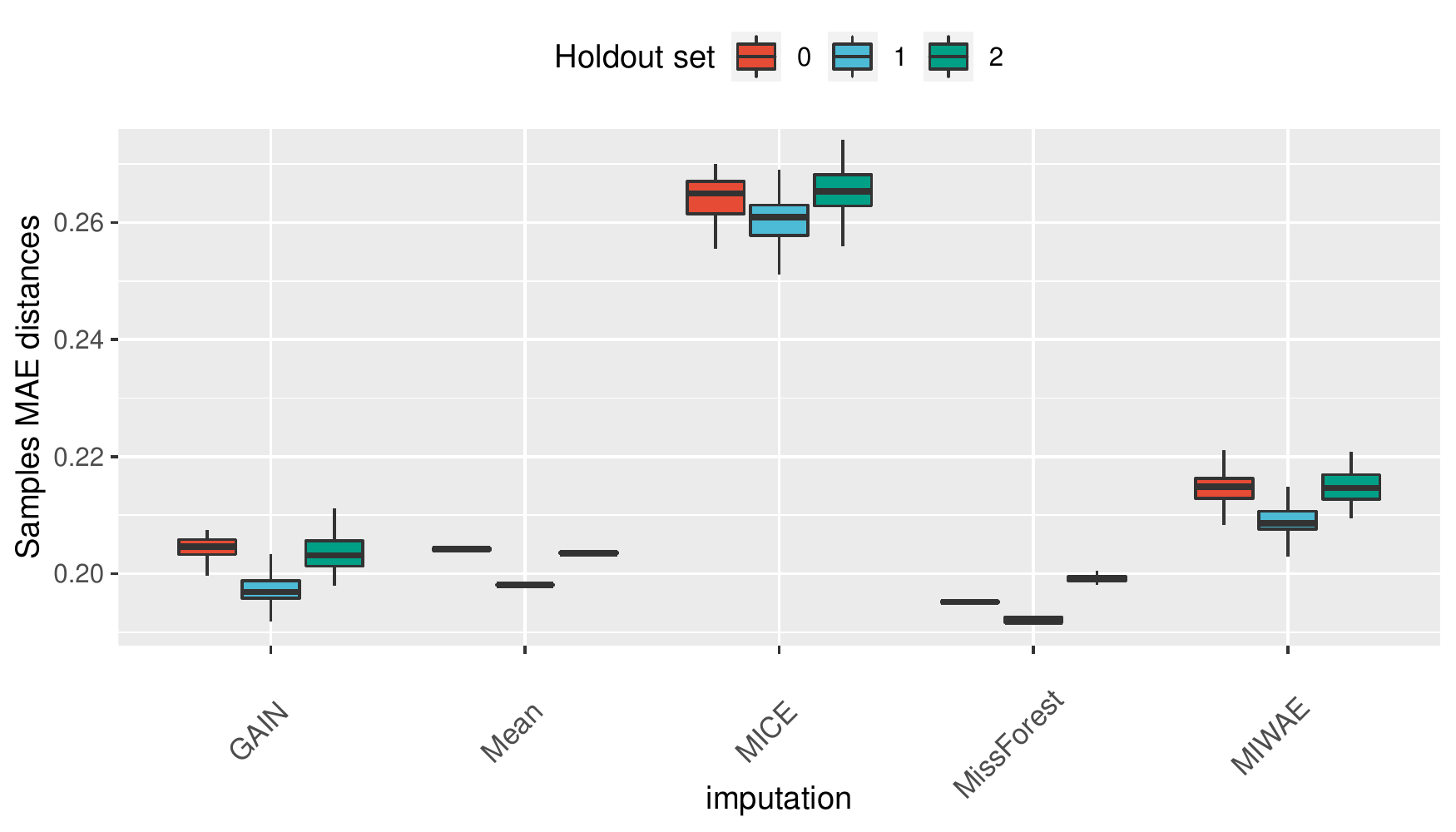}&
      \includegraphics[height=4cm,width=5cm]{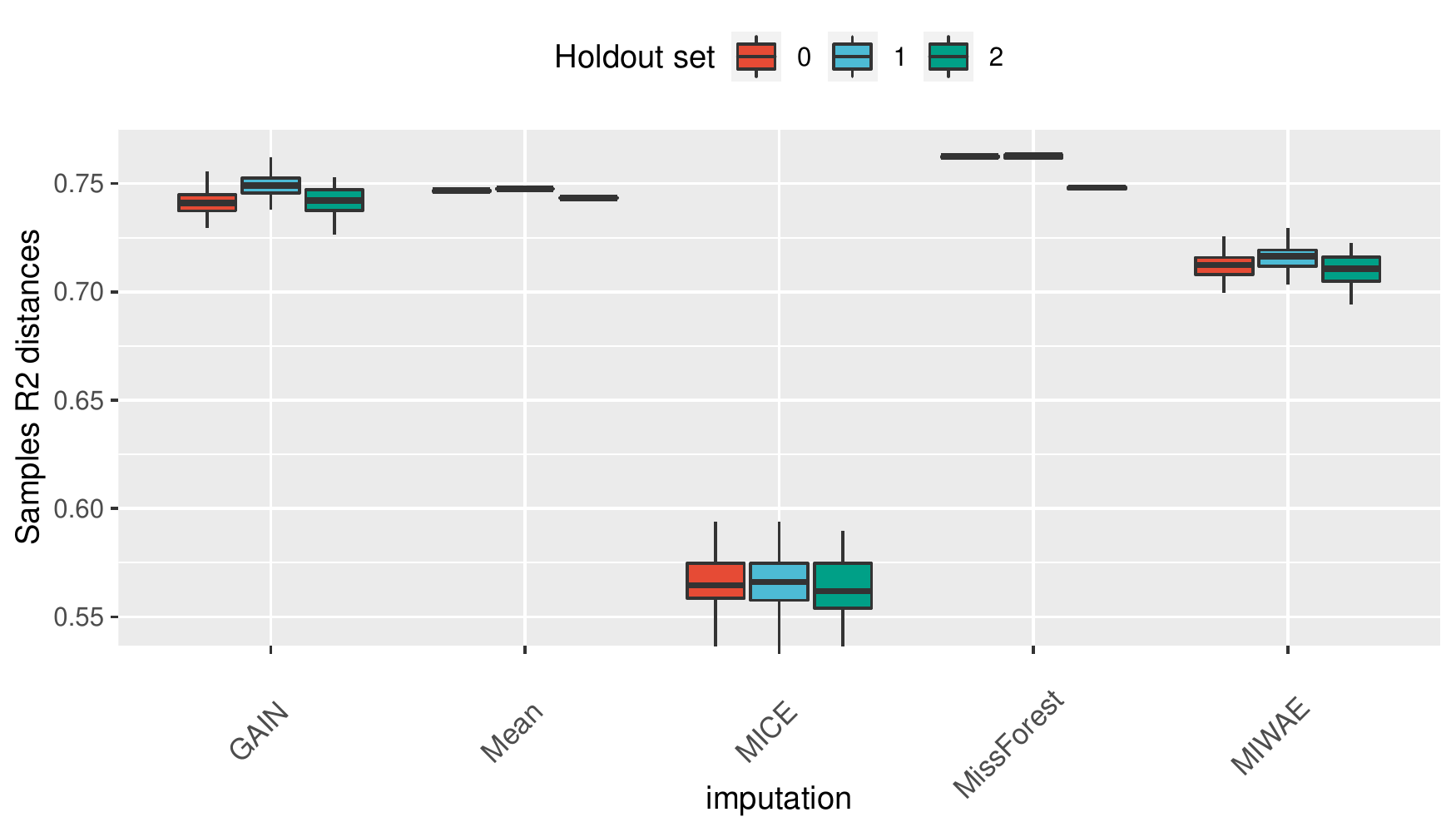}\\
    \hline
        \parbox[c][][c]{0.5in}{\rotatebox[origin=t]{90}{Train 50 \%, Test 25\%}} &
      \includegraphics[height=4cm,width=5cm]{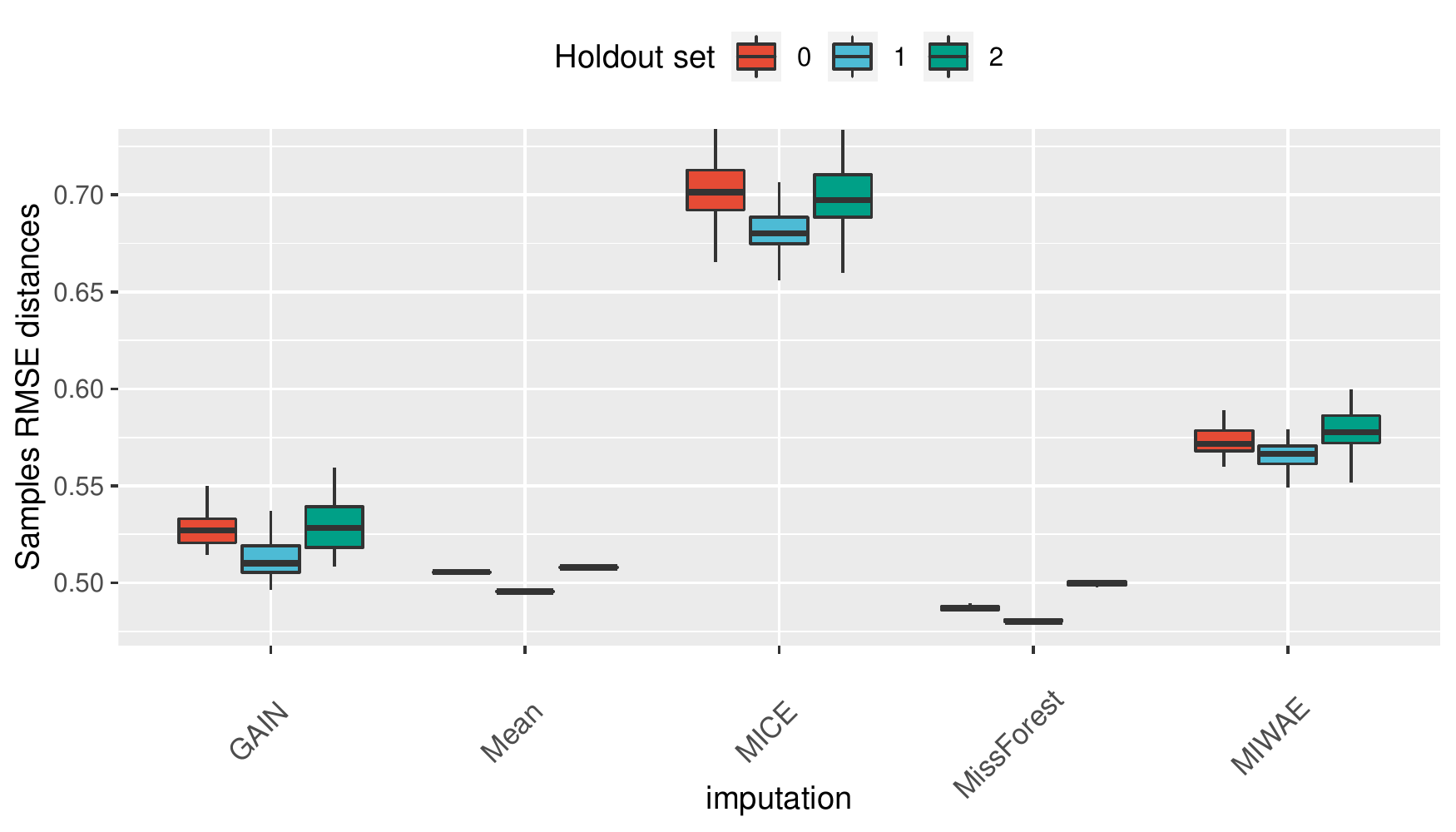}&
      \includegraphics[height=4cm,width=5cm]{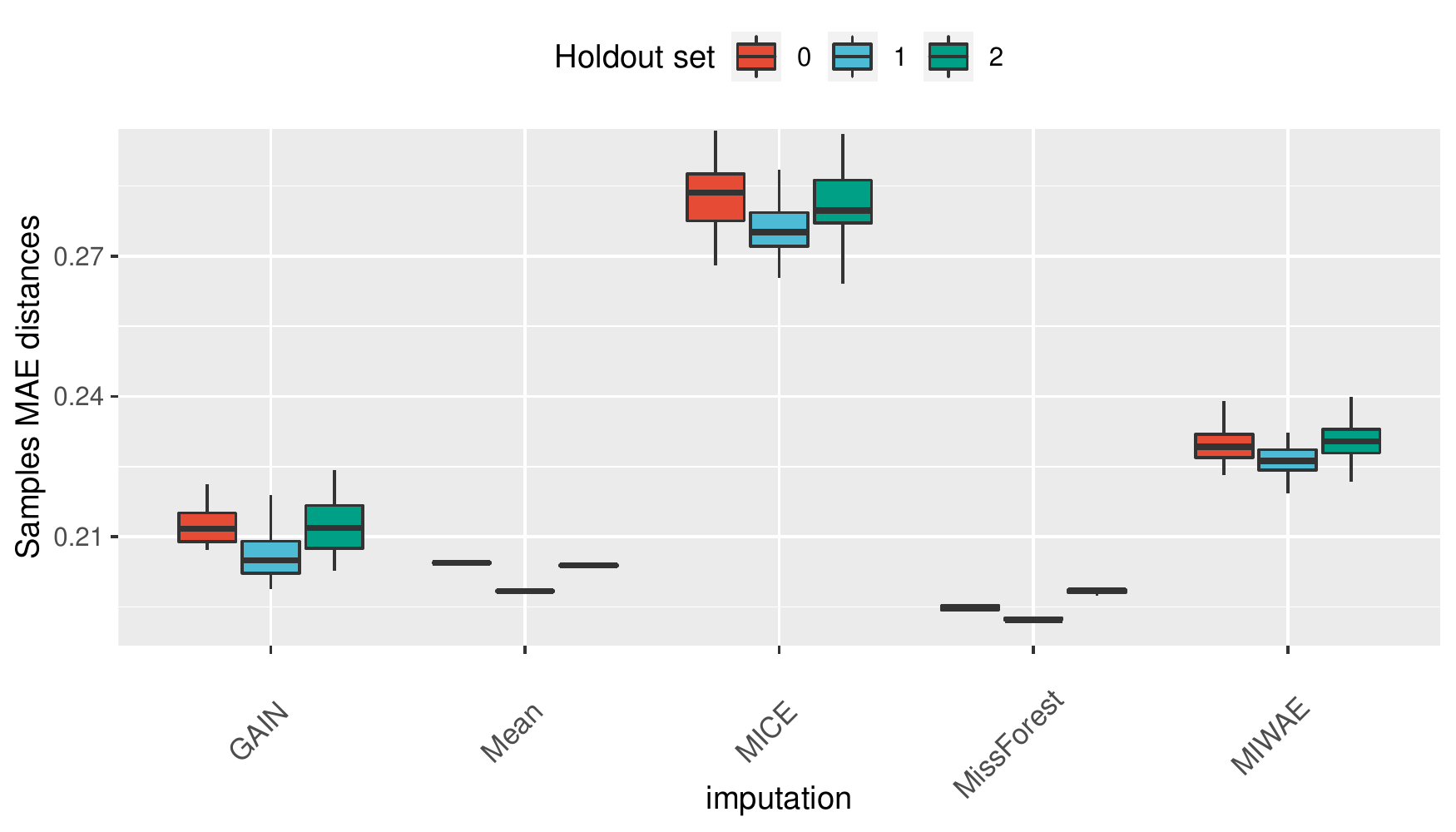}&
      \includegraphics[height=4cm,width=5cm]{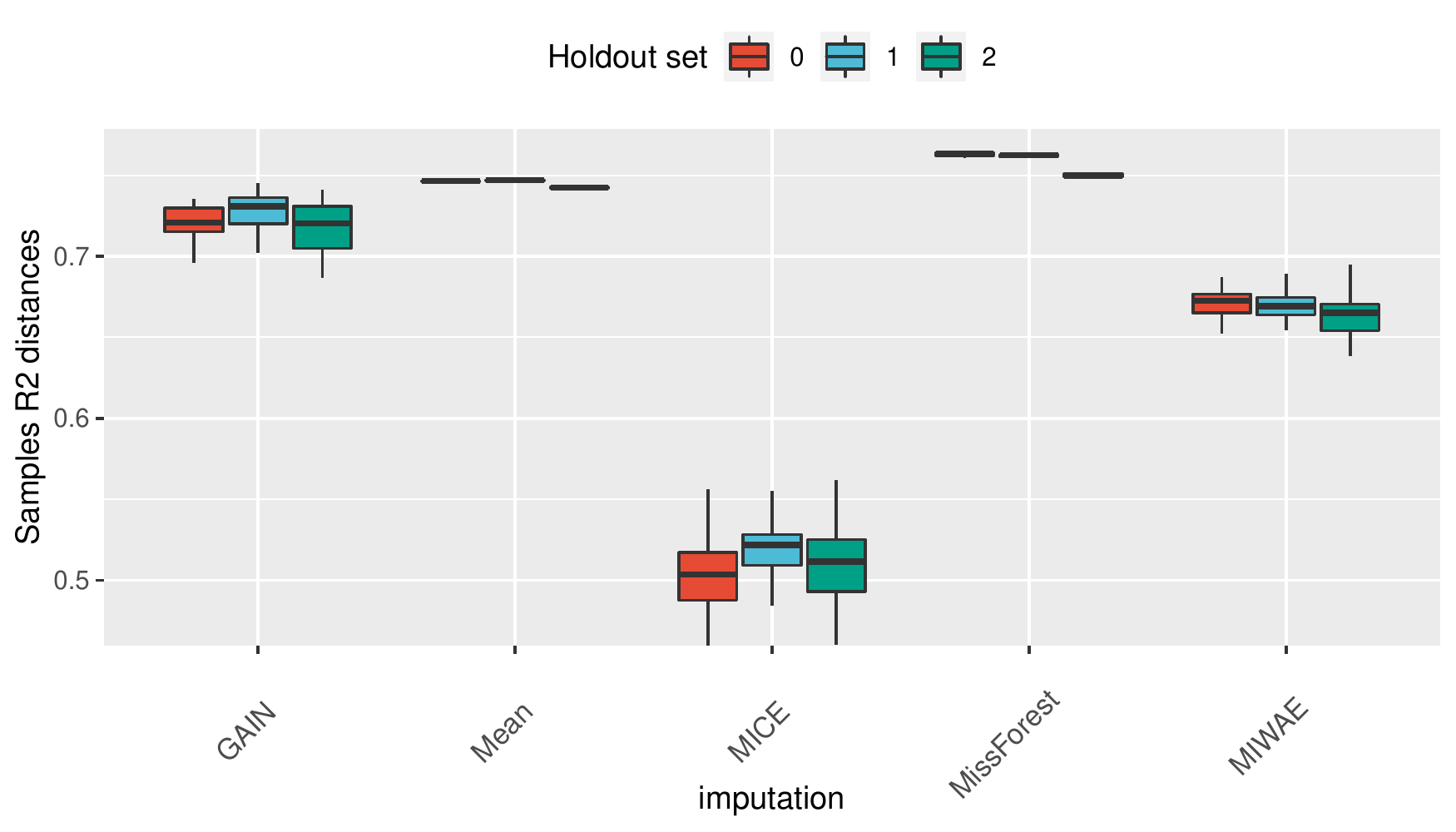}\\
    \hline
    \parbox[c][][c]{0.5in}{\rotatebox[origin=t]{90}{Train 25 \%, Test 50\%}} &
      \includegraphics[height=4cm,width=5cm]{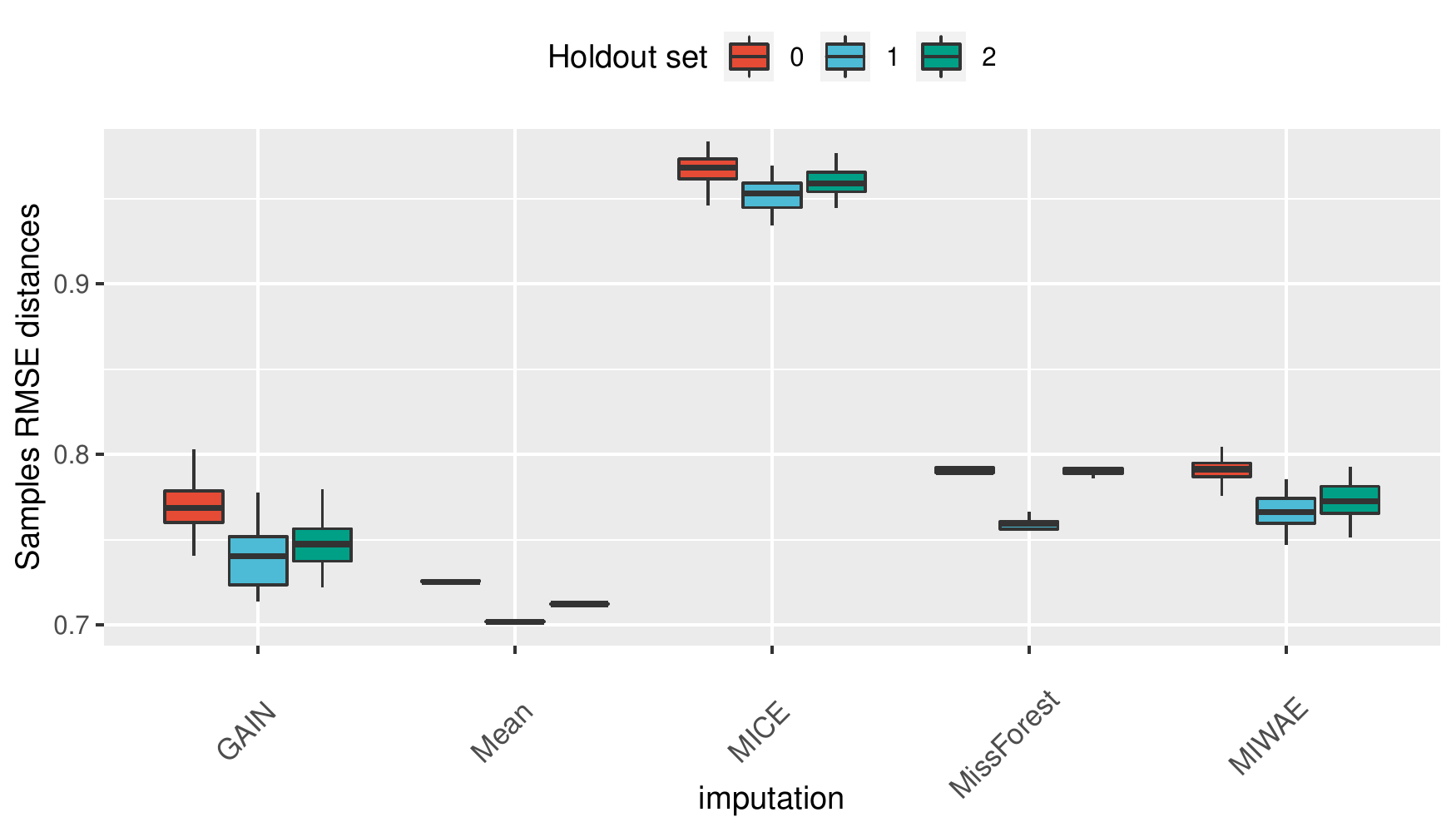}&
      \includegraphics[height=4cm,width=5cm]{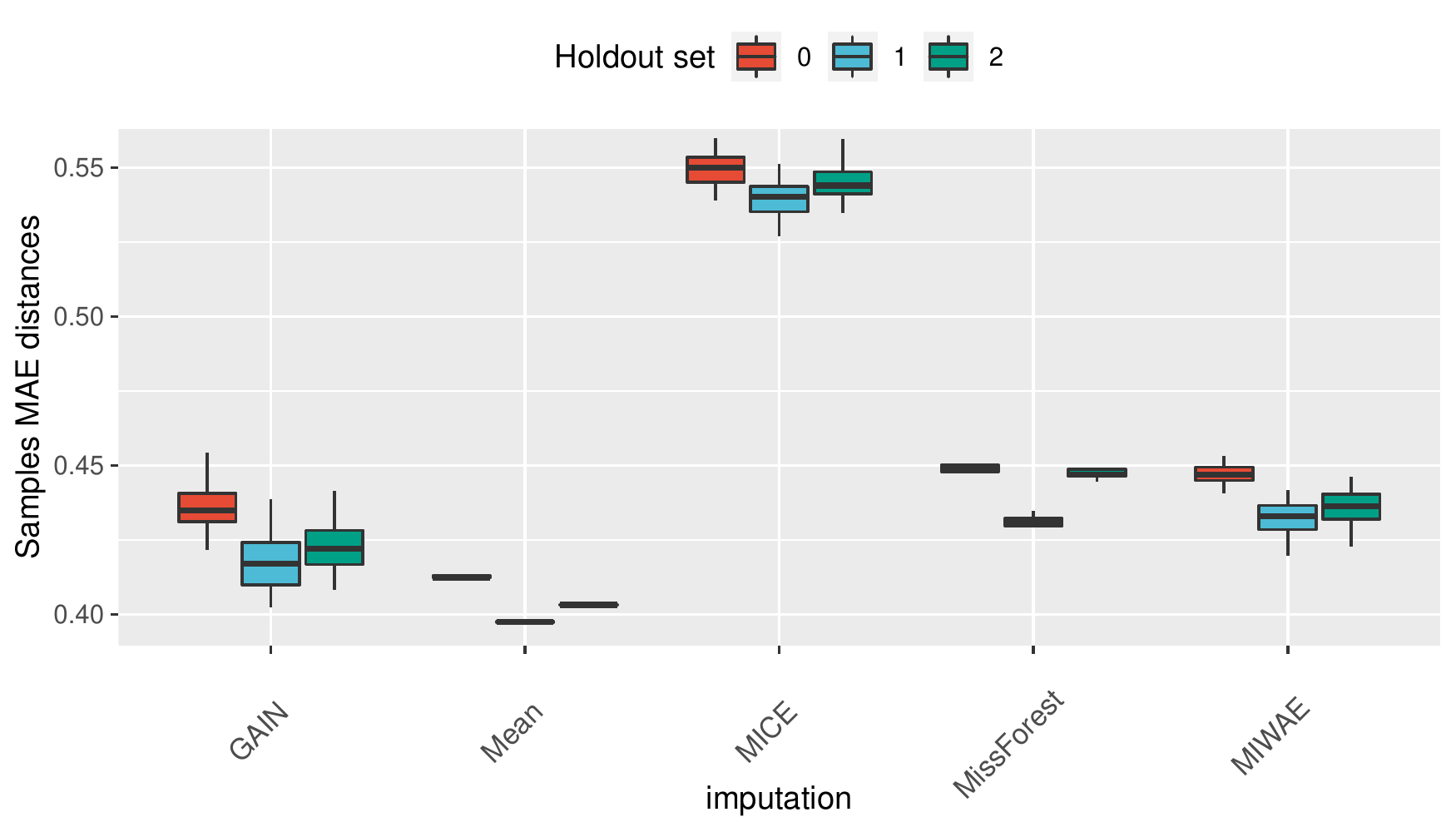}&
      \includegraphics[height=4cm,width=5cm]{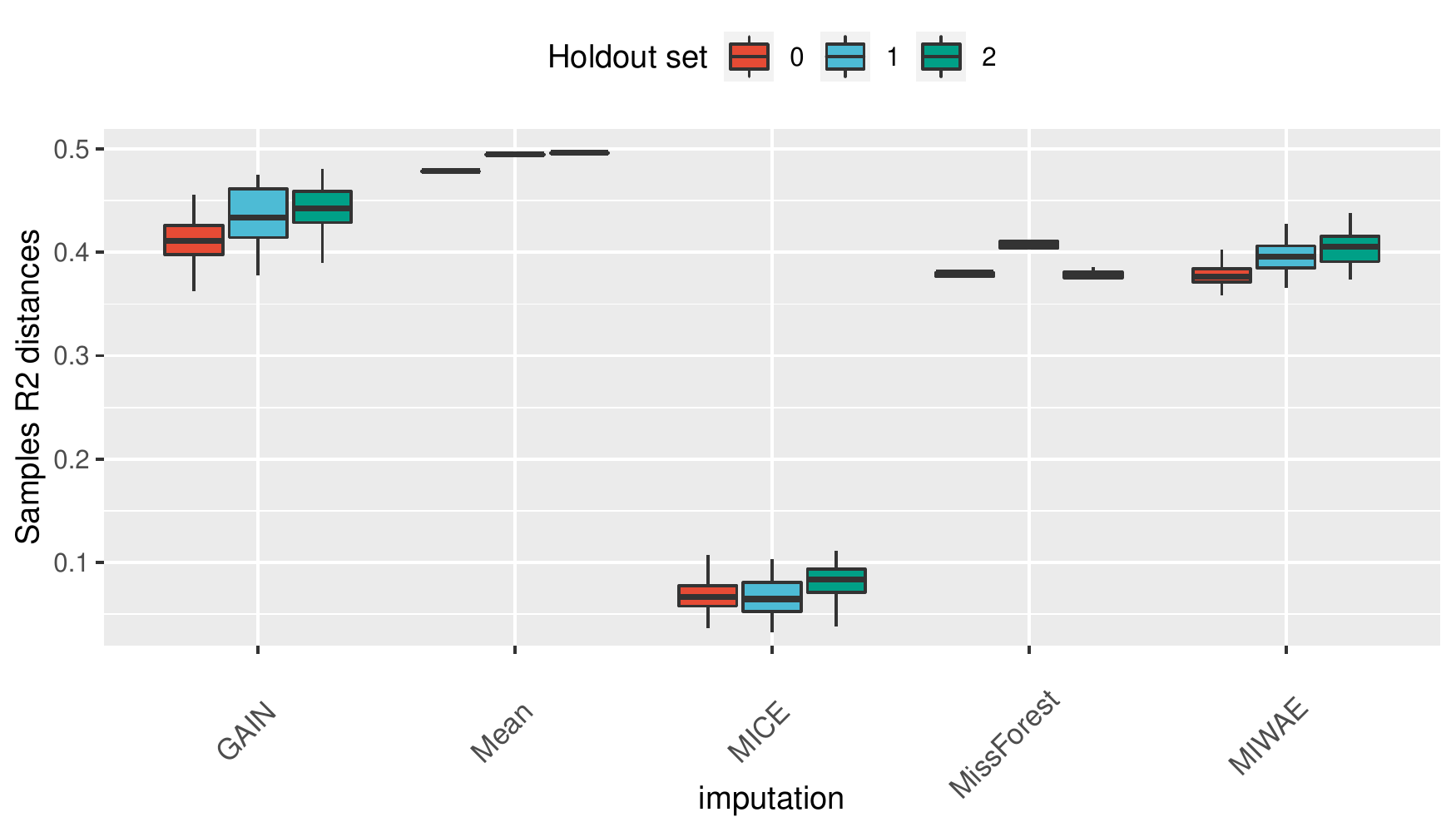}\\
    \hline
    \parbox[c][][c]{0.5in}{\rotatebox[origin=t]{90}{Train 50 \%, Test 50\%}} &
      \includegraphics[height=4cm,width=5cm]{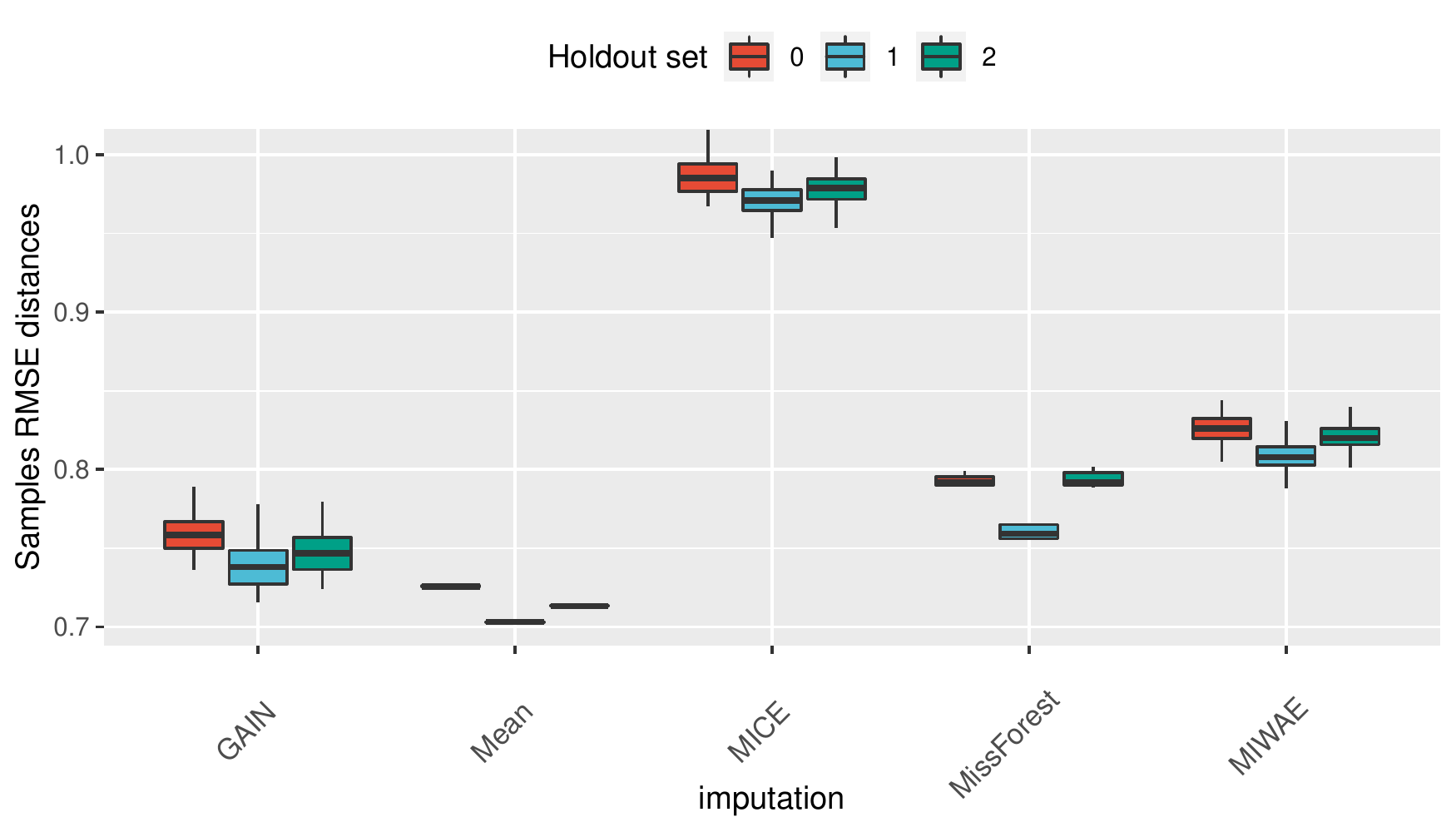}&
      \includegraphics[height=4cm,width=5cm]{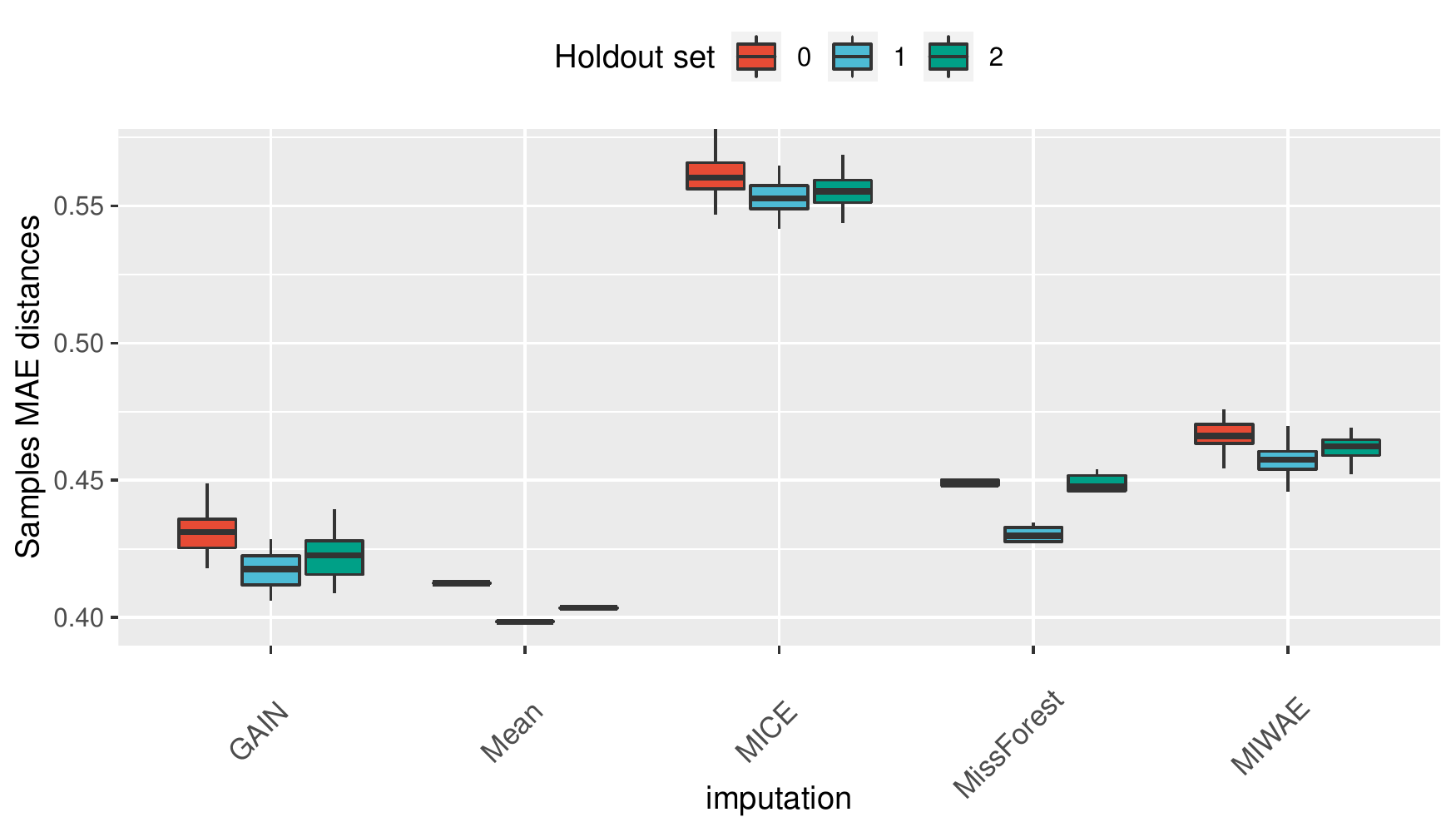}&
      \includegraphics[height=4cm,width=5cm]{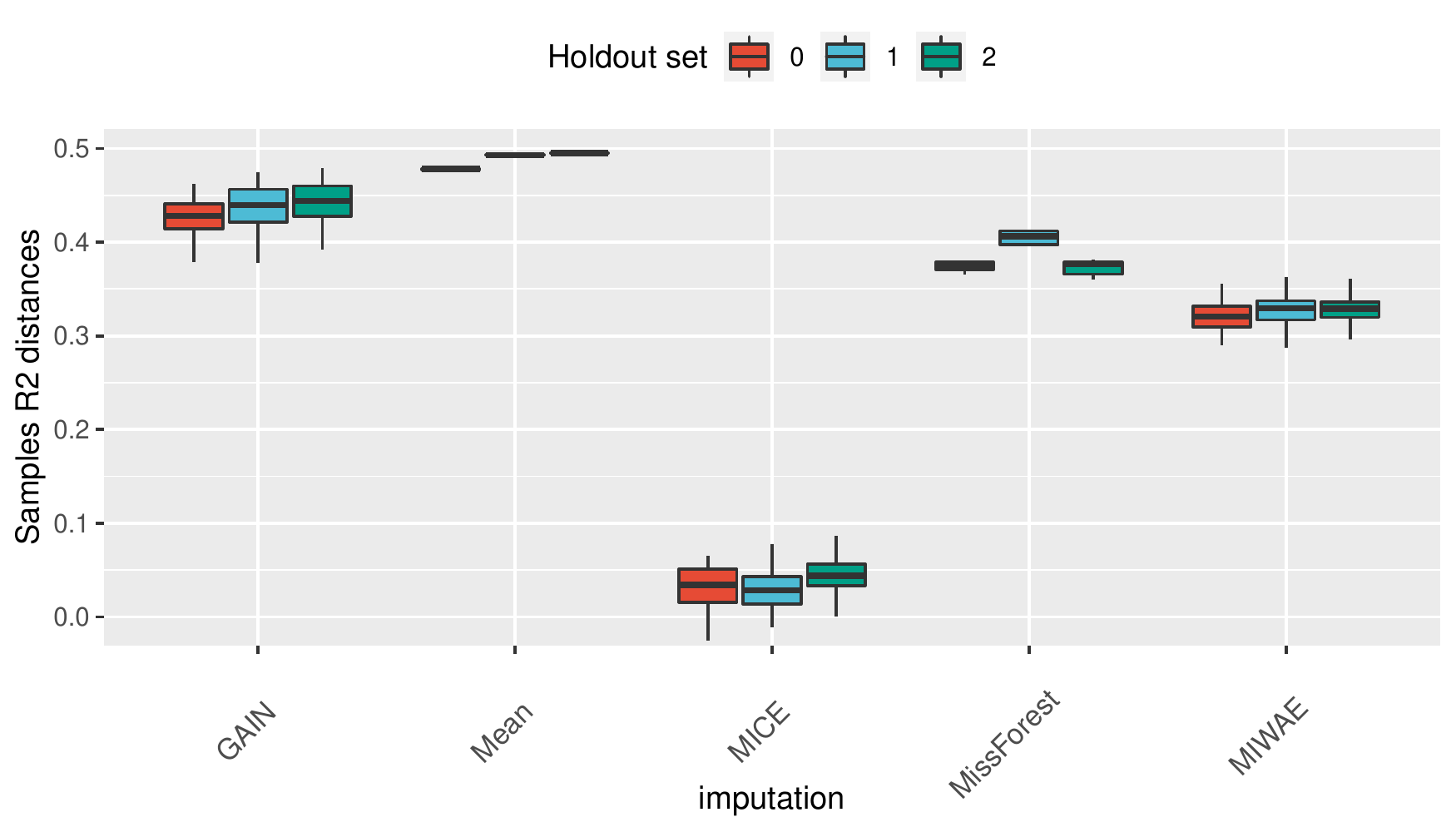}\\
    \end{tabular}
    \caption{The sample-wise statistics for the \textbf{Simulated} dataset at the different train and test missingness rates considered.}
    \label{fig:samplewise_syn}
\end{figure}

\noindent
{\emph{A. Sample-wise statistics.}}
In Figure~\ref{fig:samplewise} and Supplementary Figures~\ref{fig:samplewise_mimic}--\ref{fig:samplewise_syn} the sample-wise discrepancy scores are shown for the \textbf{MIMIC-III} and \textbf{Simulated} datasets for different train and test missingness levels. RMSE and MAE are generally consistent across the different imputation methods for fixed train and test missingness rates. There is a minimal performance difference between holdout sets.
Using these sample-wise measures for the \textbf{MIMIC-III} dataset, the MissForest imputation method performs the best overall, followed by GAIN. The gap between the performance of MissForest and GAIN narrows as the train and test missingness rates increase. Mean imputation performs worst at 25\% test missingness, whilst at 50\%, MICE is the worst.
For the \textbf{Simulated} dataset, the best performing methods are MissForest and mean imputation whilst MICE is the worst.
MICE generally attains the lowest $R^{2}$ score at all missingness rates for both datasets. The poor performance of MICE by the RMSE metric can also be seen in Figure~\ref{fig:AB} although, qualitatively, the distribution is better recreated using this imputation method. \\

\begin{figure}[htb!]
    \centering
    \begin{tabular}{m{0.2in} | M{5cm} | M{5cm} | M{5cm}}
    & \textbf{Minimum} & \textbf{Median} & \textbf{Maximum} \\
    \hline
     \parbox[c][][c]{0.5in}{\rotatebox[origin=t]{90}{B1: Kullback-Leibler}} &
      \includegraphics[height=4cm,width=5cm]{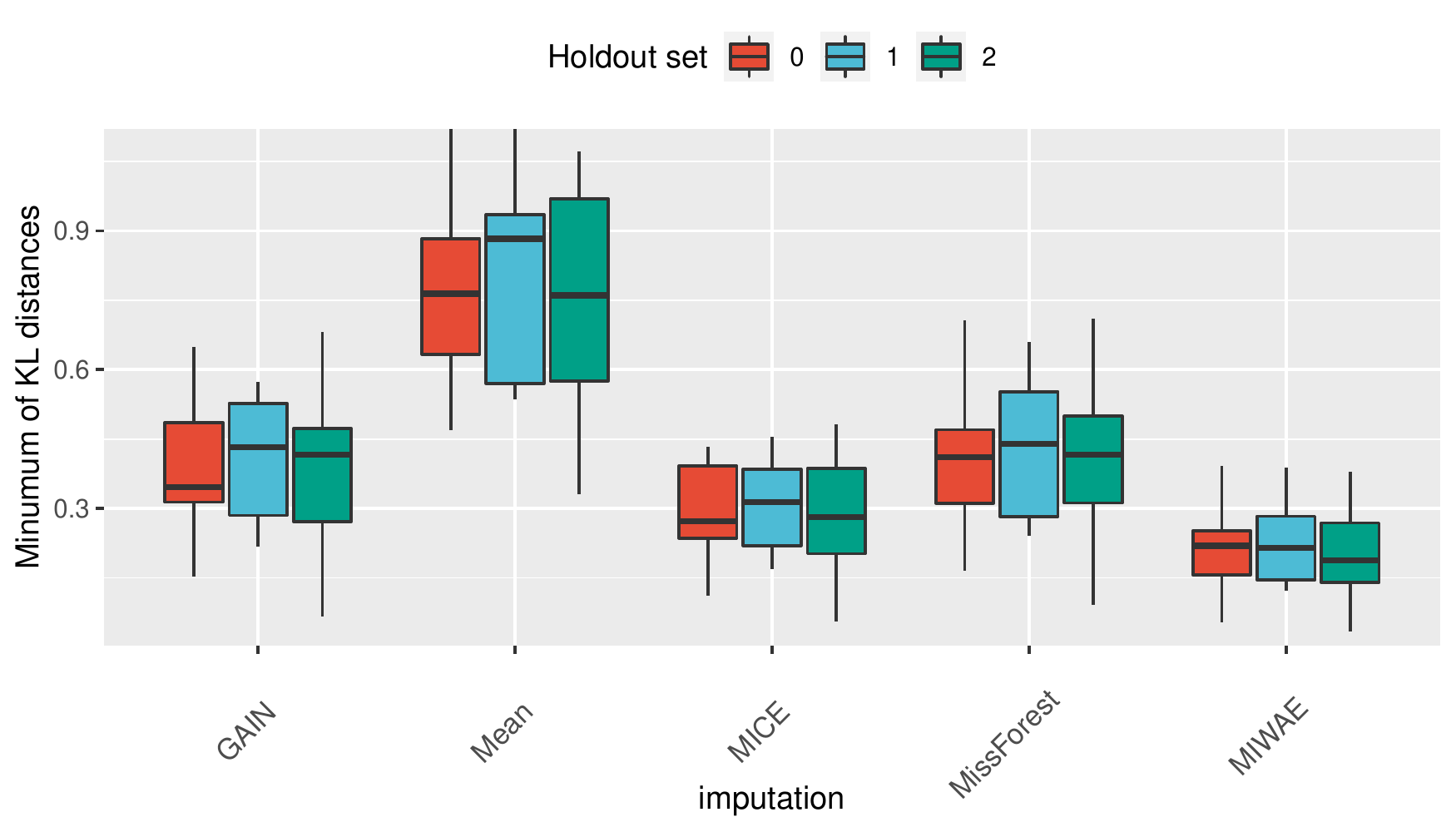}&
      \includegraphics[height=4cm,width=5cm]{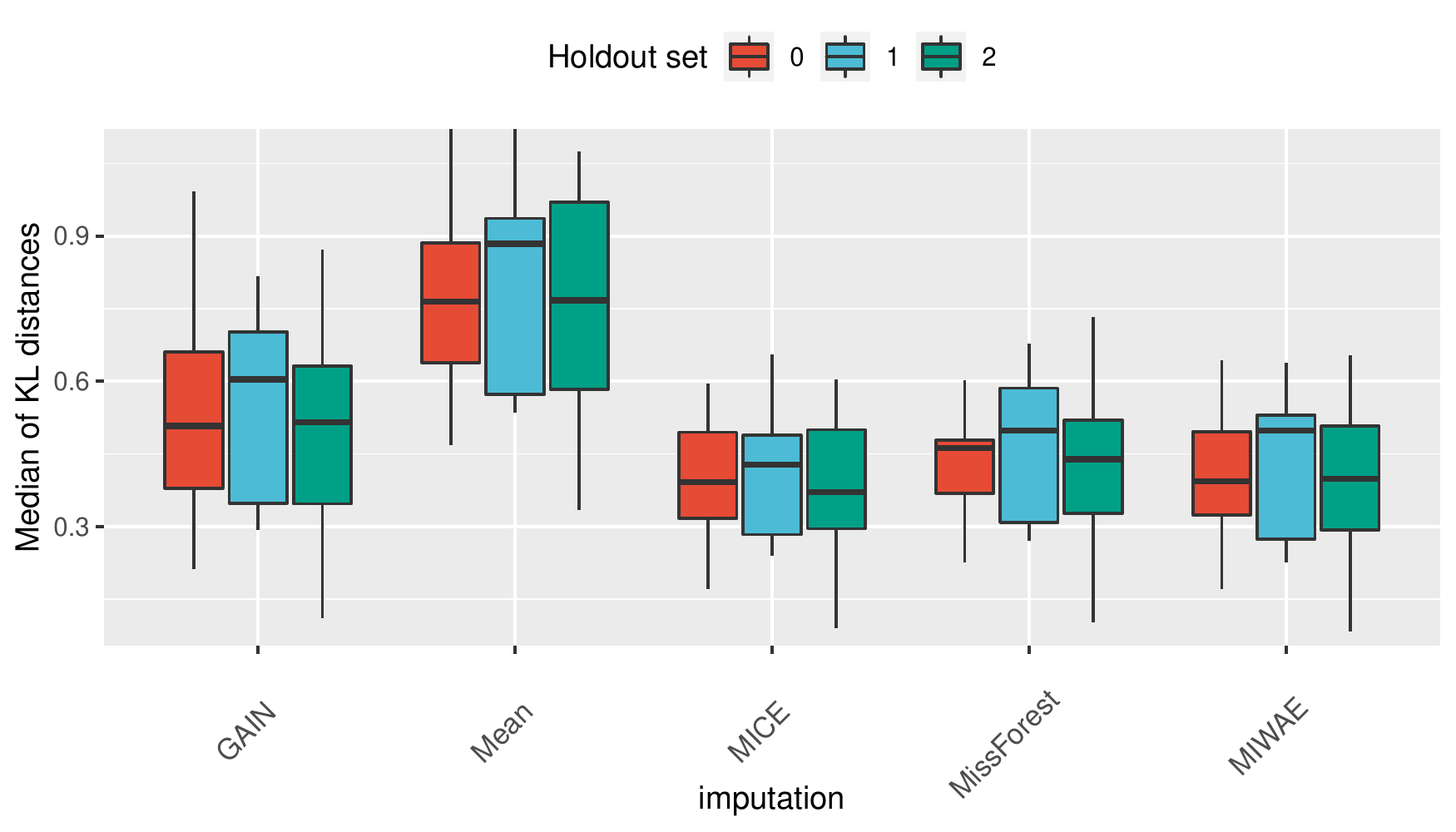}&
      \includegraphics[height=4cm,width=5cm]{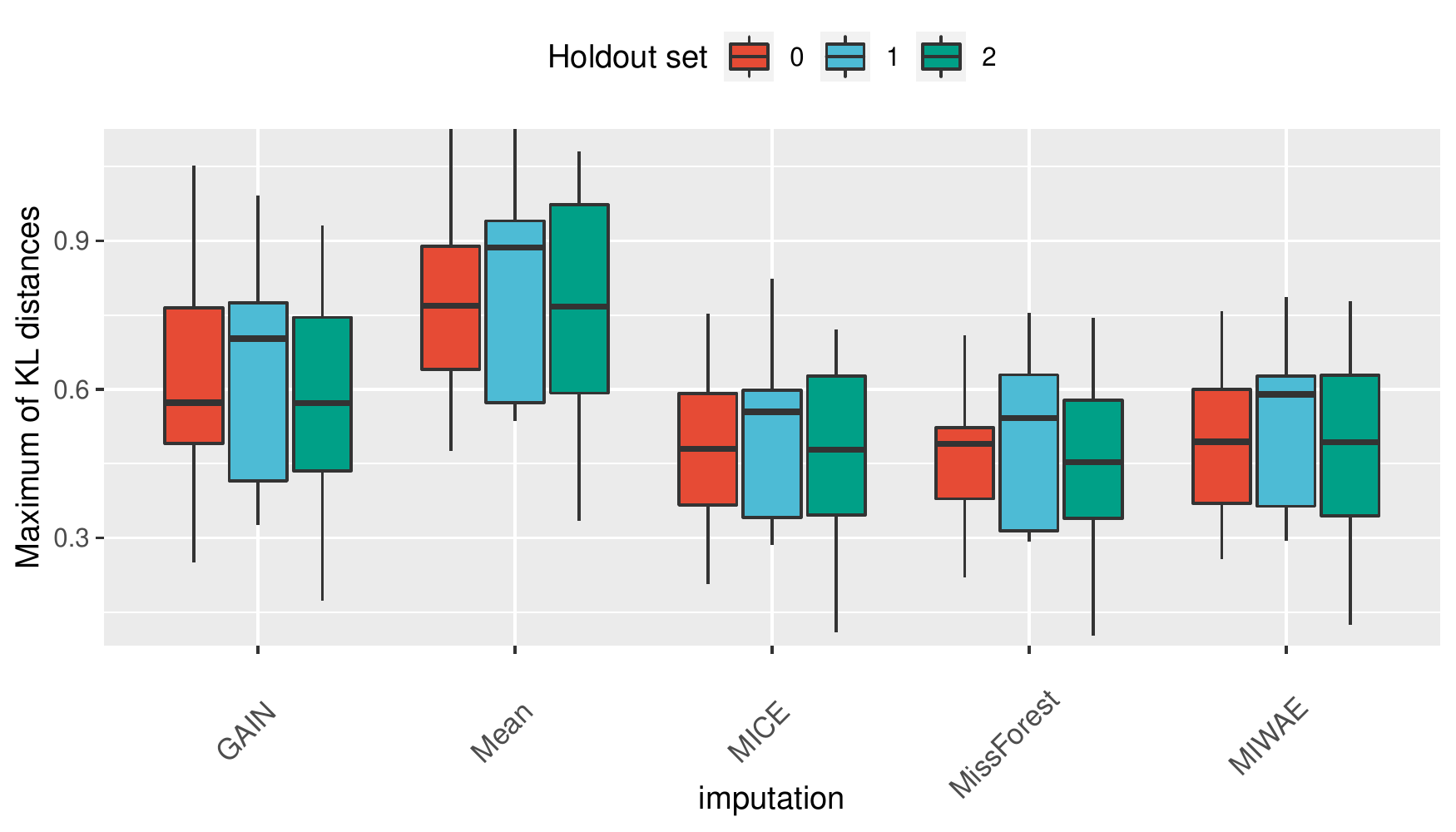}\\
      \hline
      \parbox[c][][c]{0.5in}{\rotatebox[origin=c]{90}{B2: Kolmogorov-Smirnov}} &
      \includegraphics[height=4cm,width=5cm]{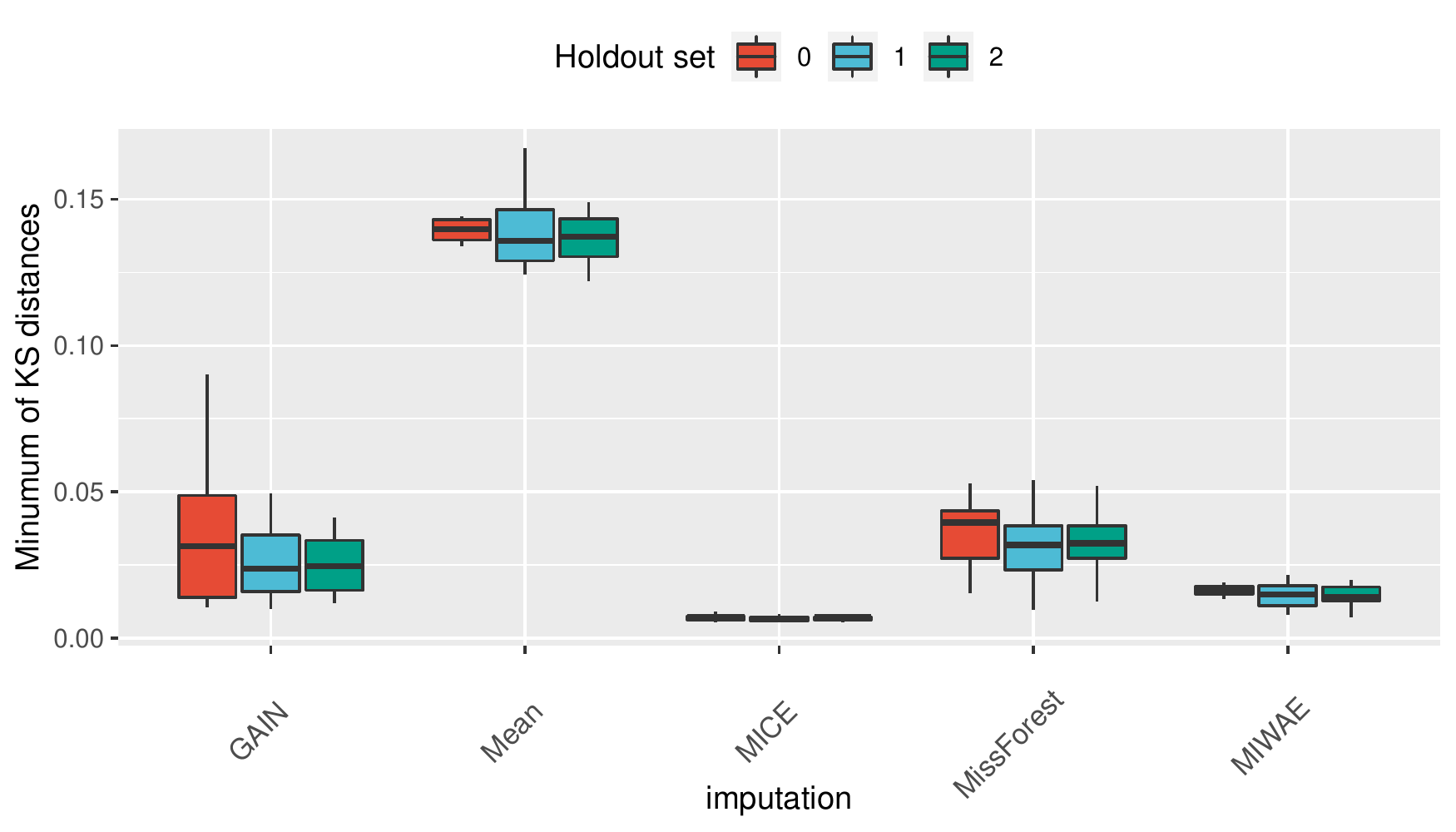}&
      \includegraphics[height=4cm,width=5cm]{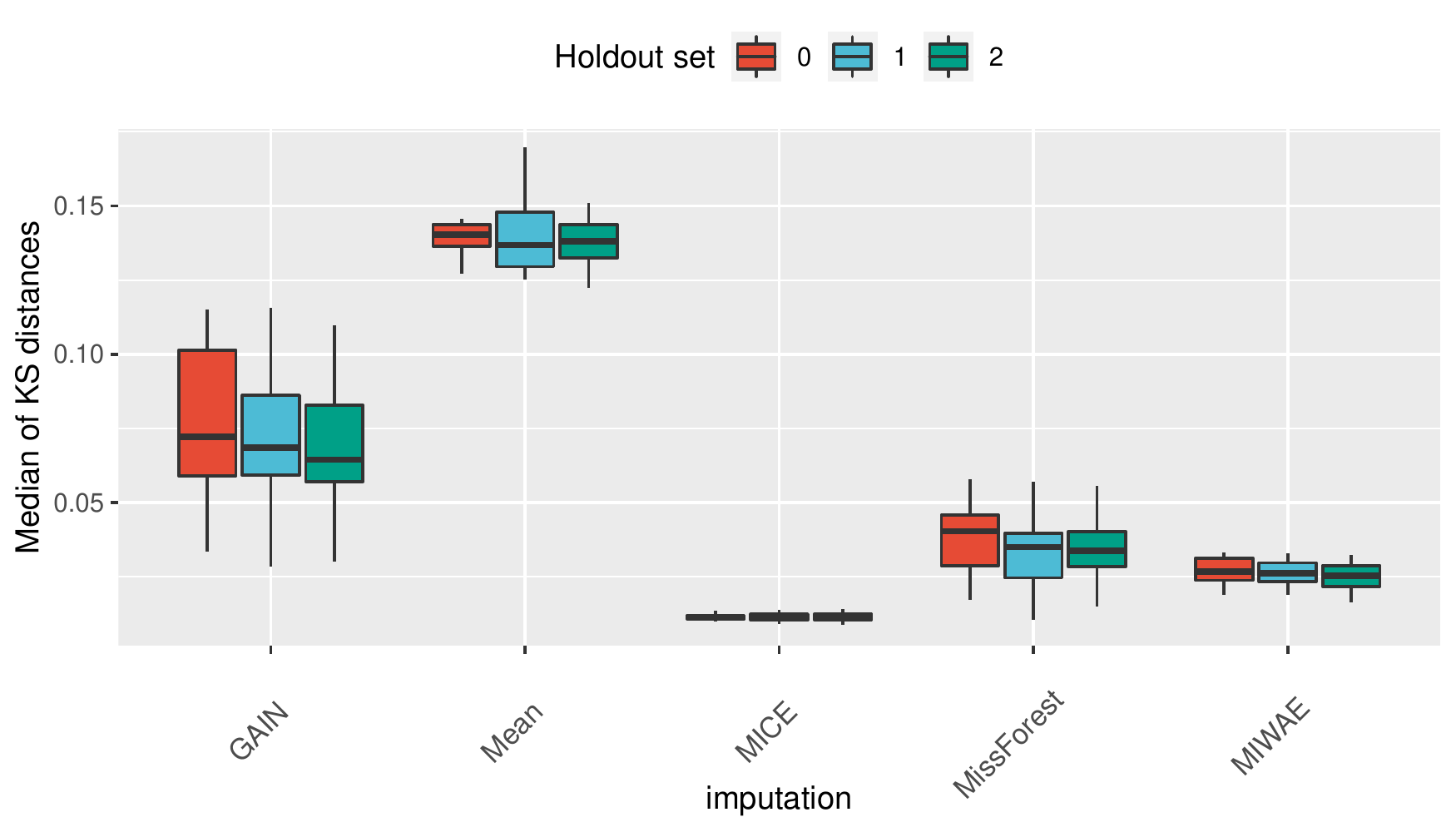}&
      \includegraphics[height=4cm,width=5cm]{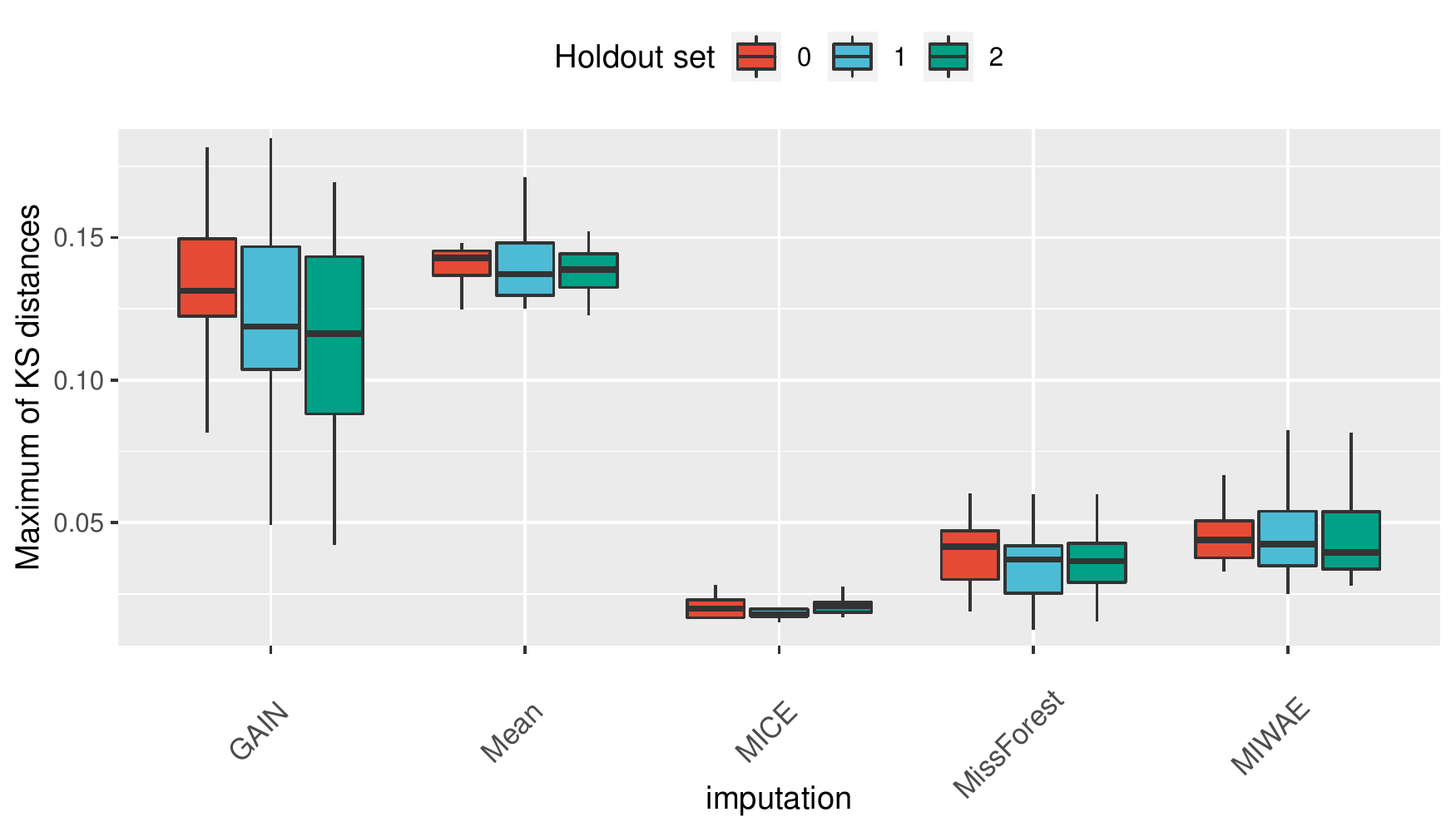}\\
      \hline
      \parbox[c][][c]{0.5in}{\rotatebox[origin=c]{90}{B3: Wasserstein}} &
      \includegraphics[height=4cm,width=5cm]{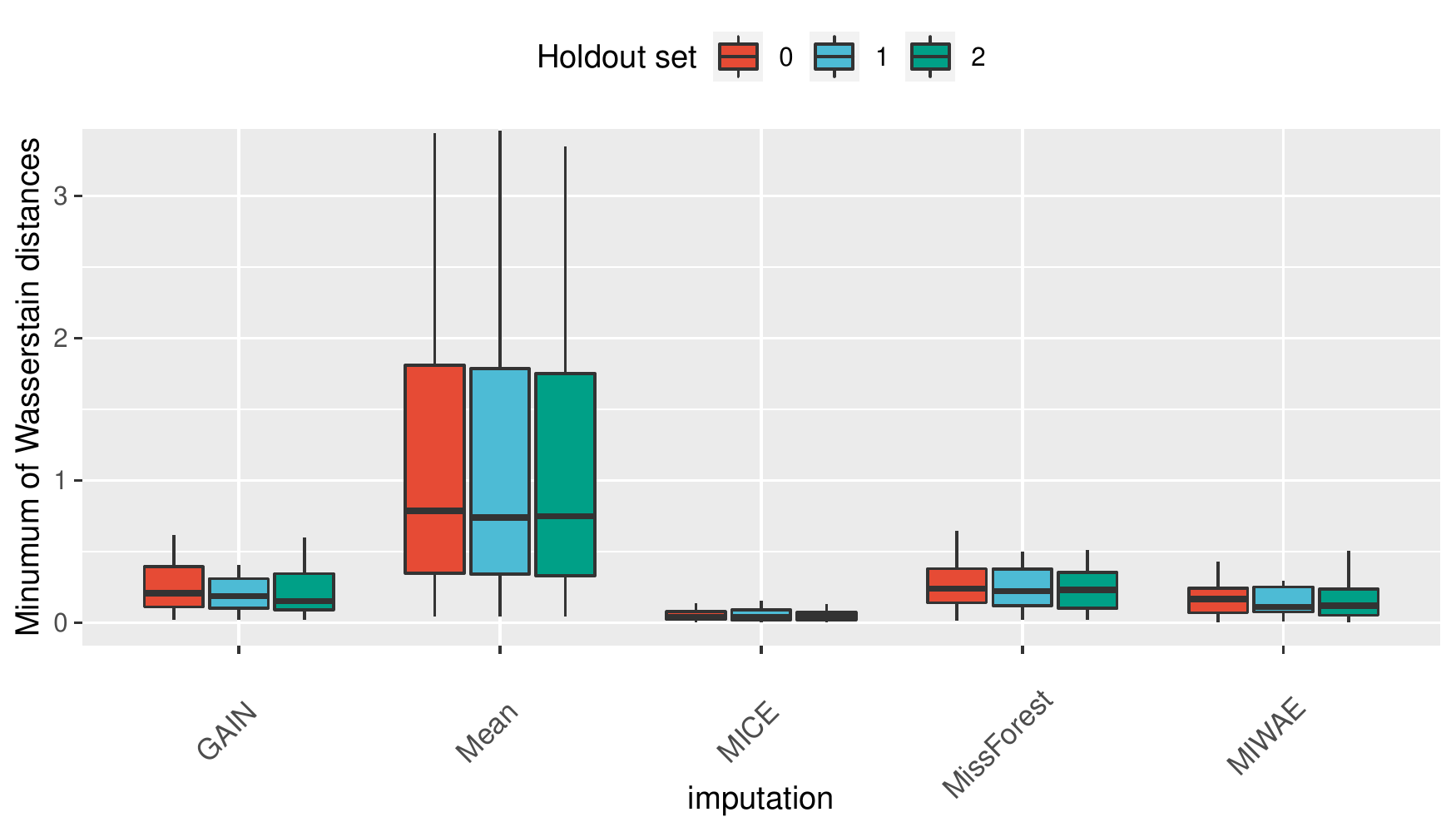}&
      \includegraphics[height=4cm,width=5cm]{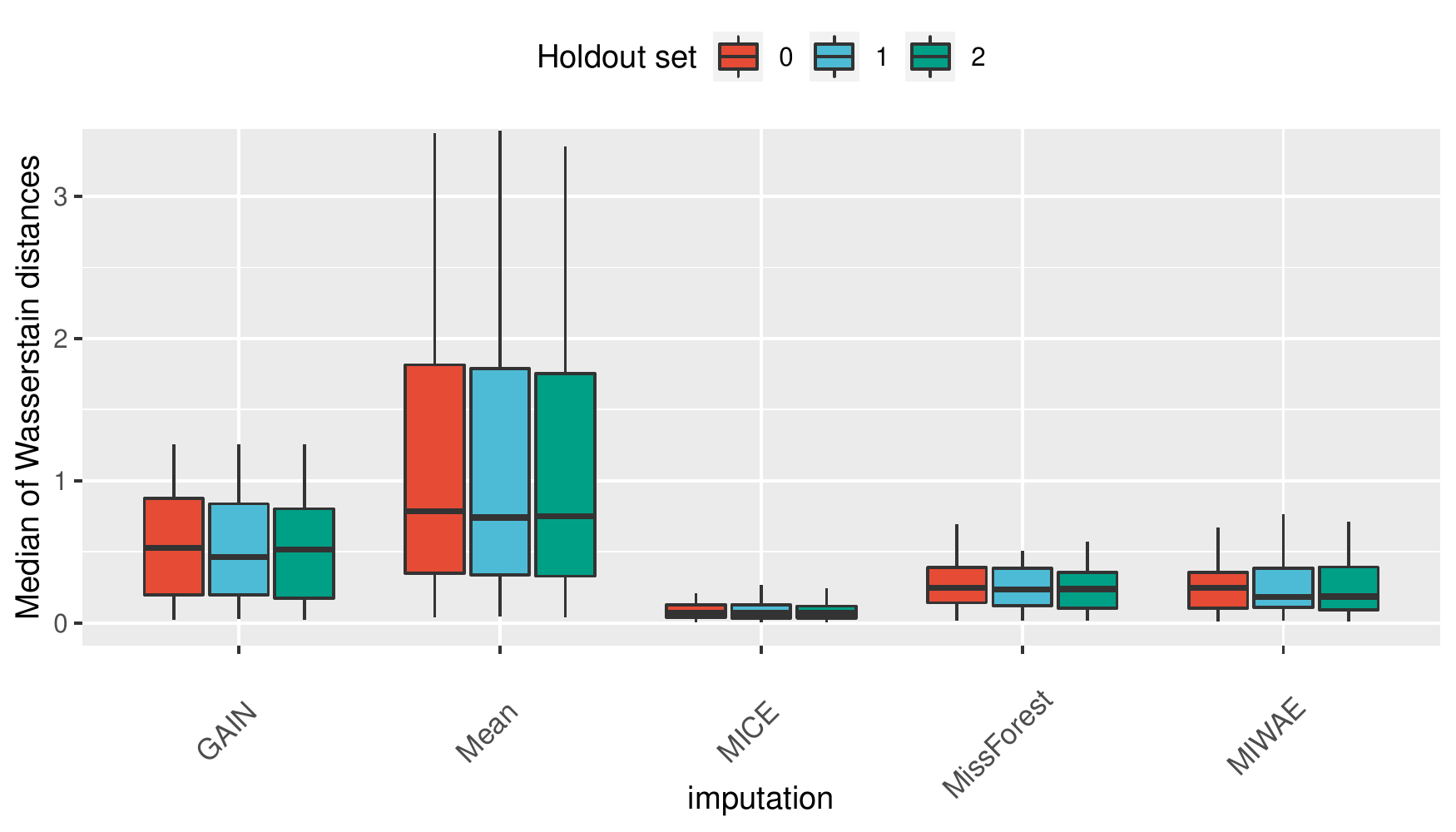}&
      \includegraphics[height=4cm,width=5cm]{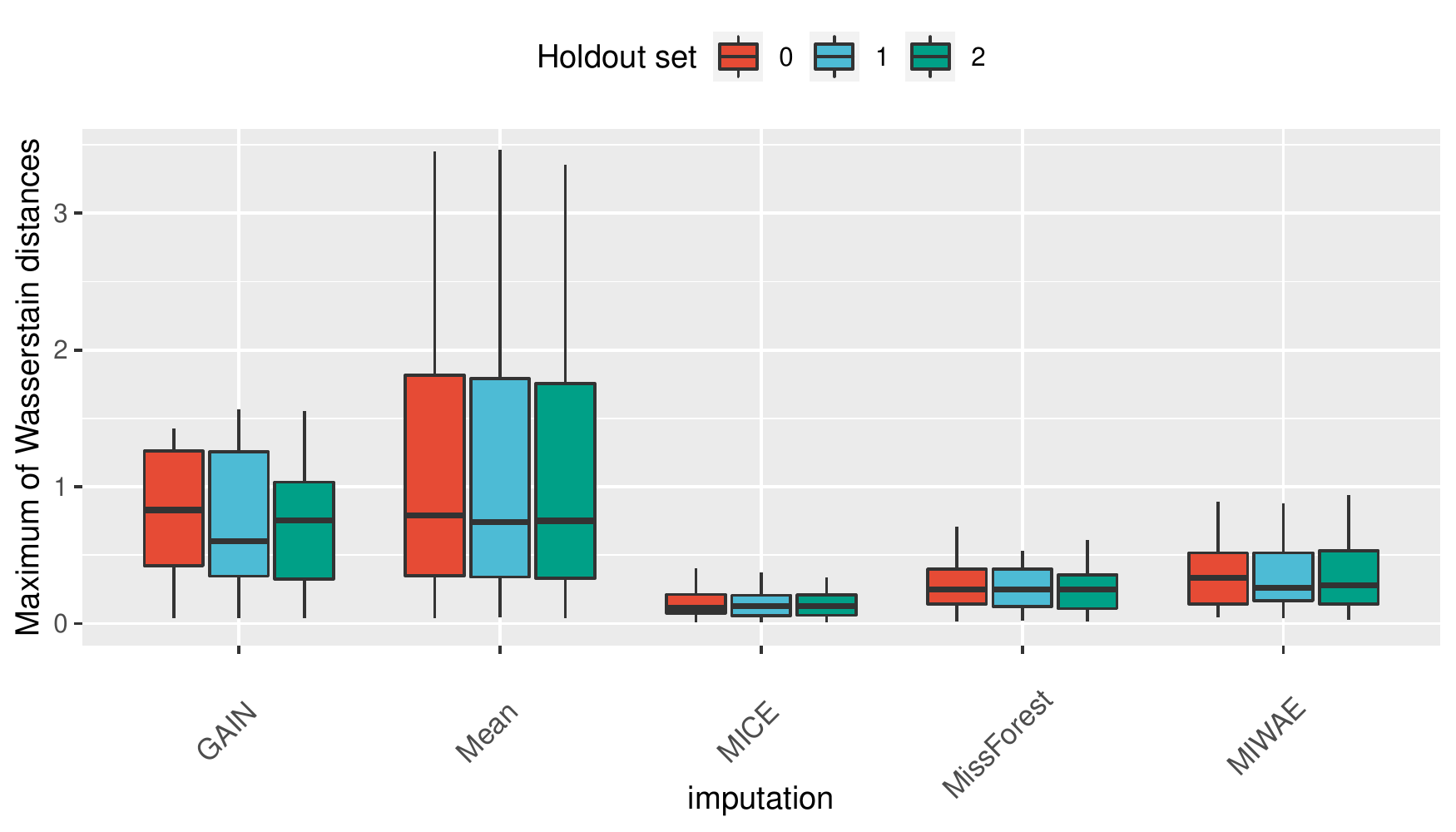}\\
    \end{tabular}
    \caption{The feature-wise statistics for the \textbf{MIMIC-III} dataset with 25\% train and test missingness rates.}
    \label{fig:featurewise_mimic_25_25}
\end{figure}

\begin{figure}[htb!]
    \centering
    \begin{tabular}{m{0.2in} | M{5cm} | M{5cm} | M{5cm}}
    & \textbf{Minimum} & \textbf{Median} & \textbf{Maximum} \\
    \hline
     \parbox[c][][c]{0.5in}{\rotatebox[origin=t]{90}{B1: Kullback-Leibler}} &
      \includegraphics[height=4cm,width=5cm]{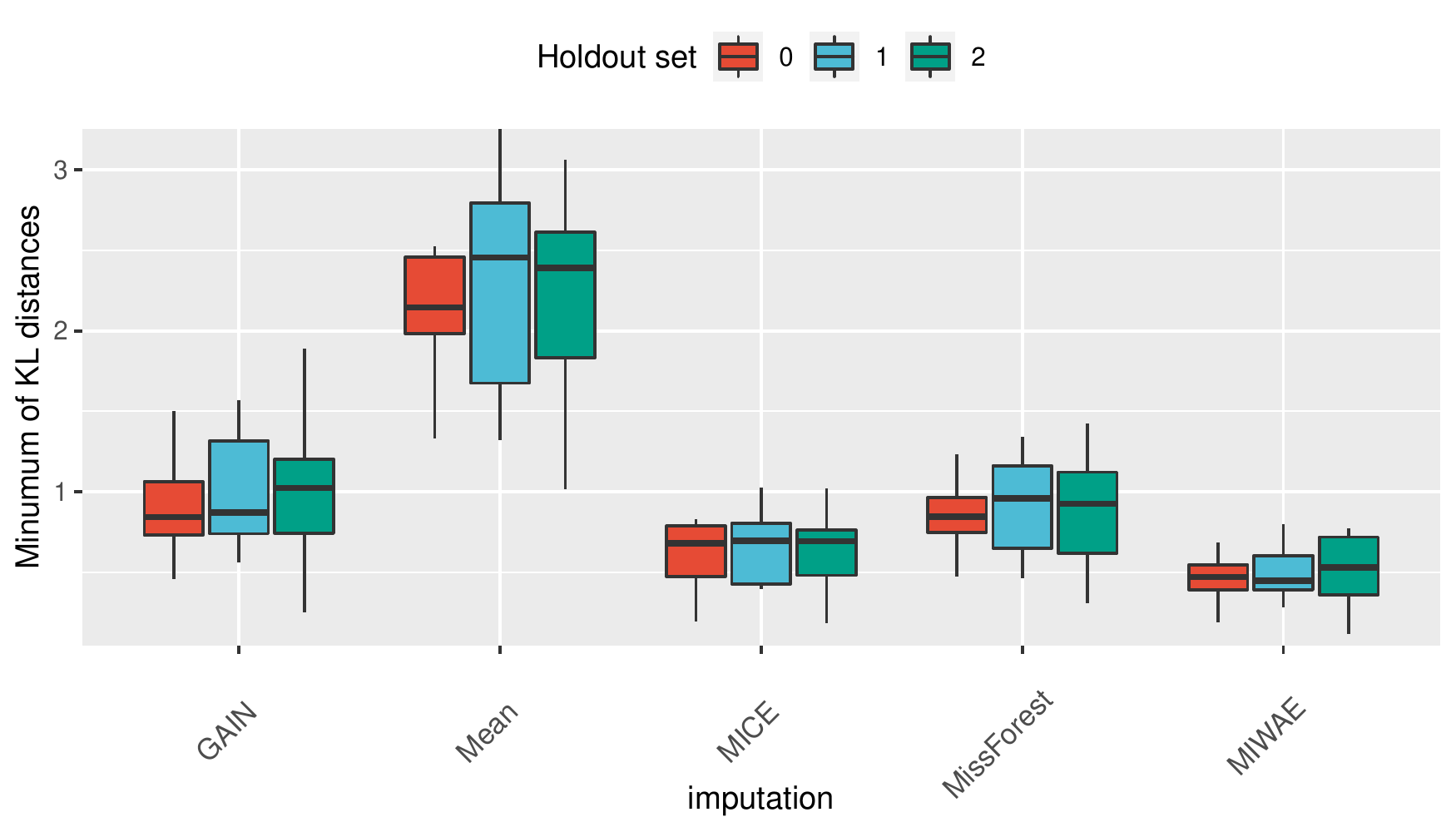}&
      \includegraphics[height=4cm,width=5cm]{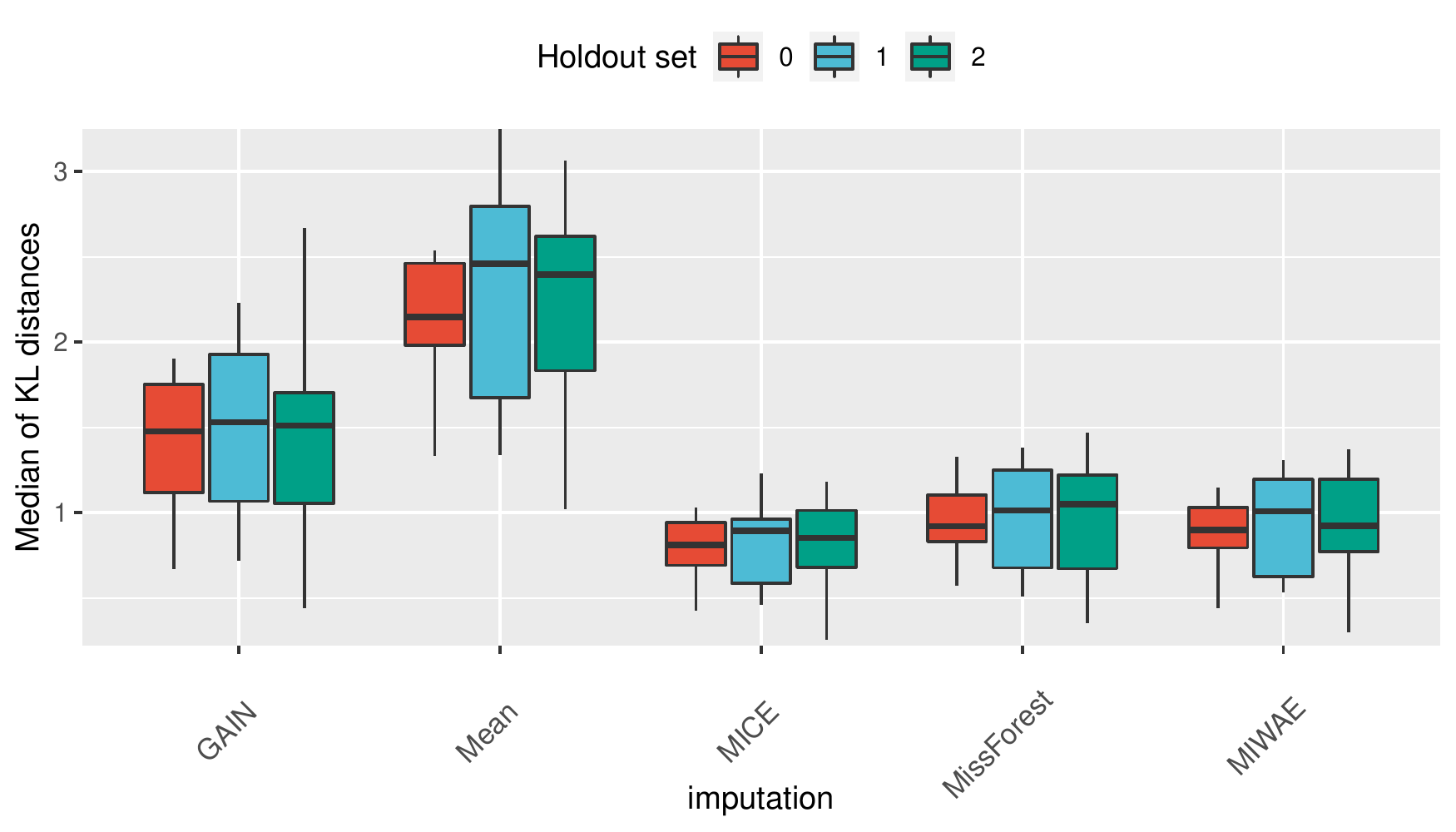}&
      \includegraphics[height=4cm,width=5cm]{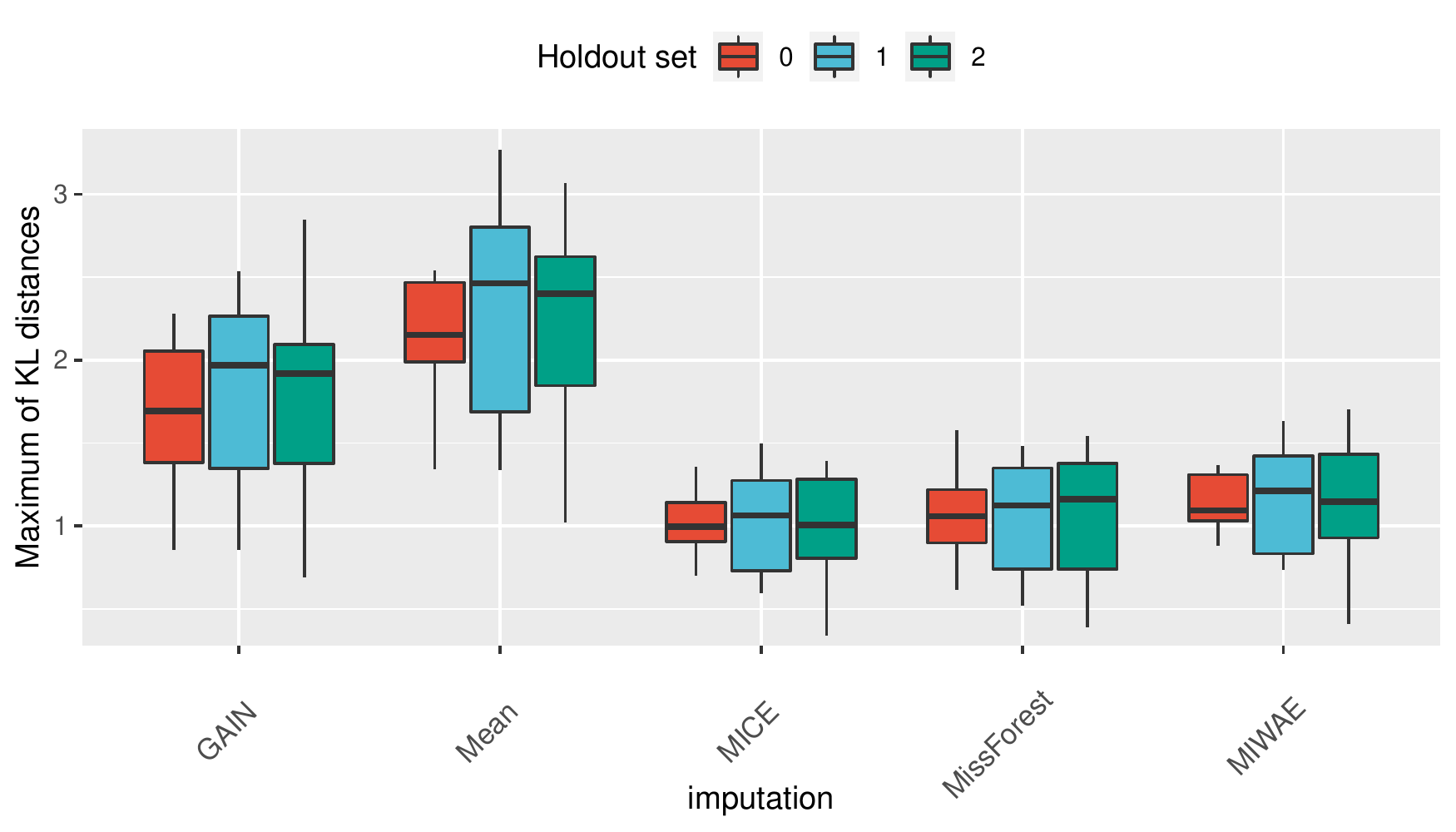}\\
      \hline
      \parbox[c][][c]{0.5in}{\rotatebox[origin=c]{90}{B2: Kolmogorov-Smirnov}} &
      \includegraphics[height=4cm,width=5cm]{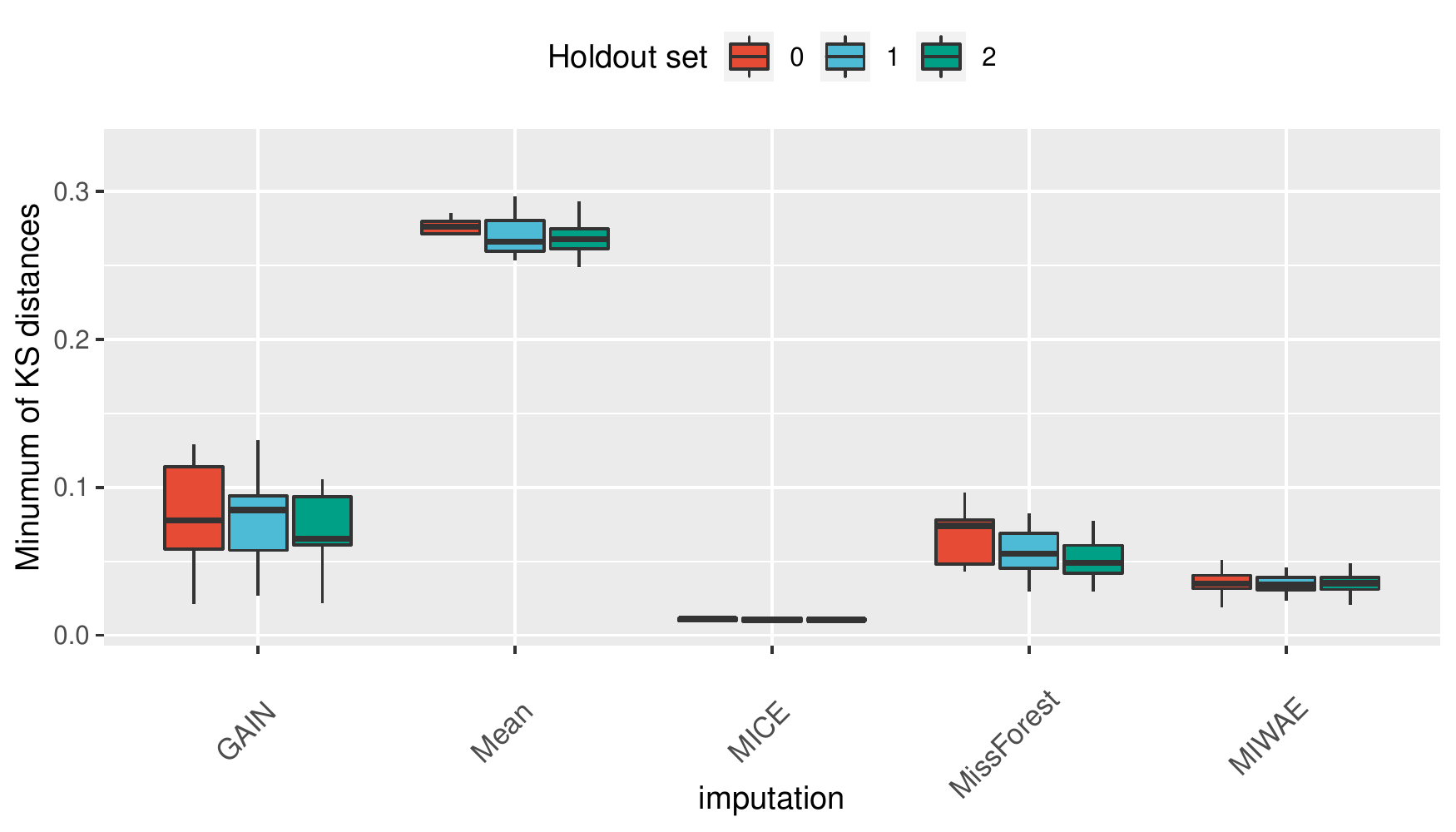}&
      \includegraphics[height=4cm,width=5cm]{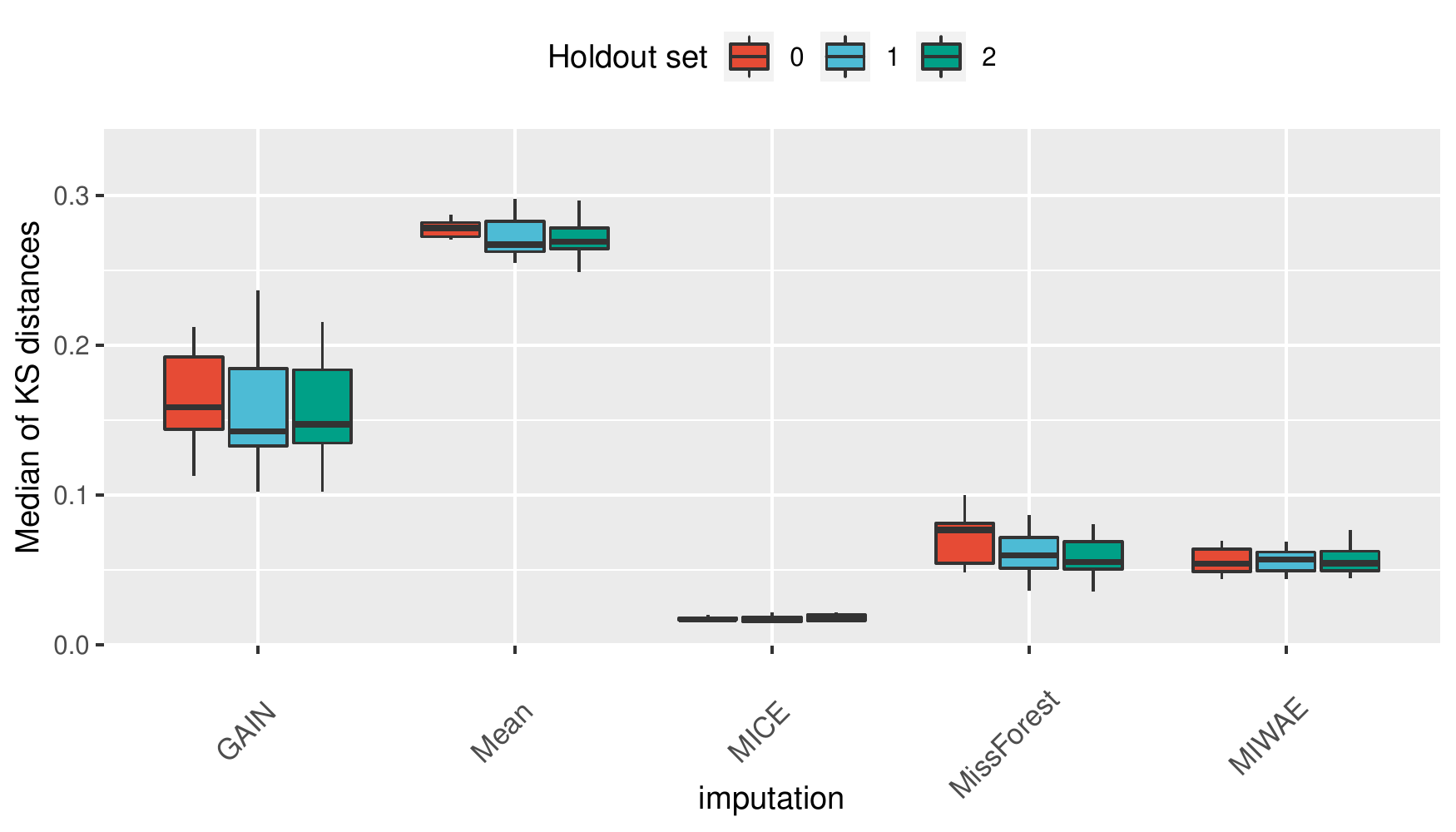}&
      \includegraphics[height=4cm,width=5cm]{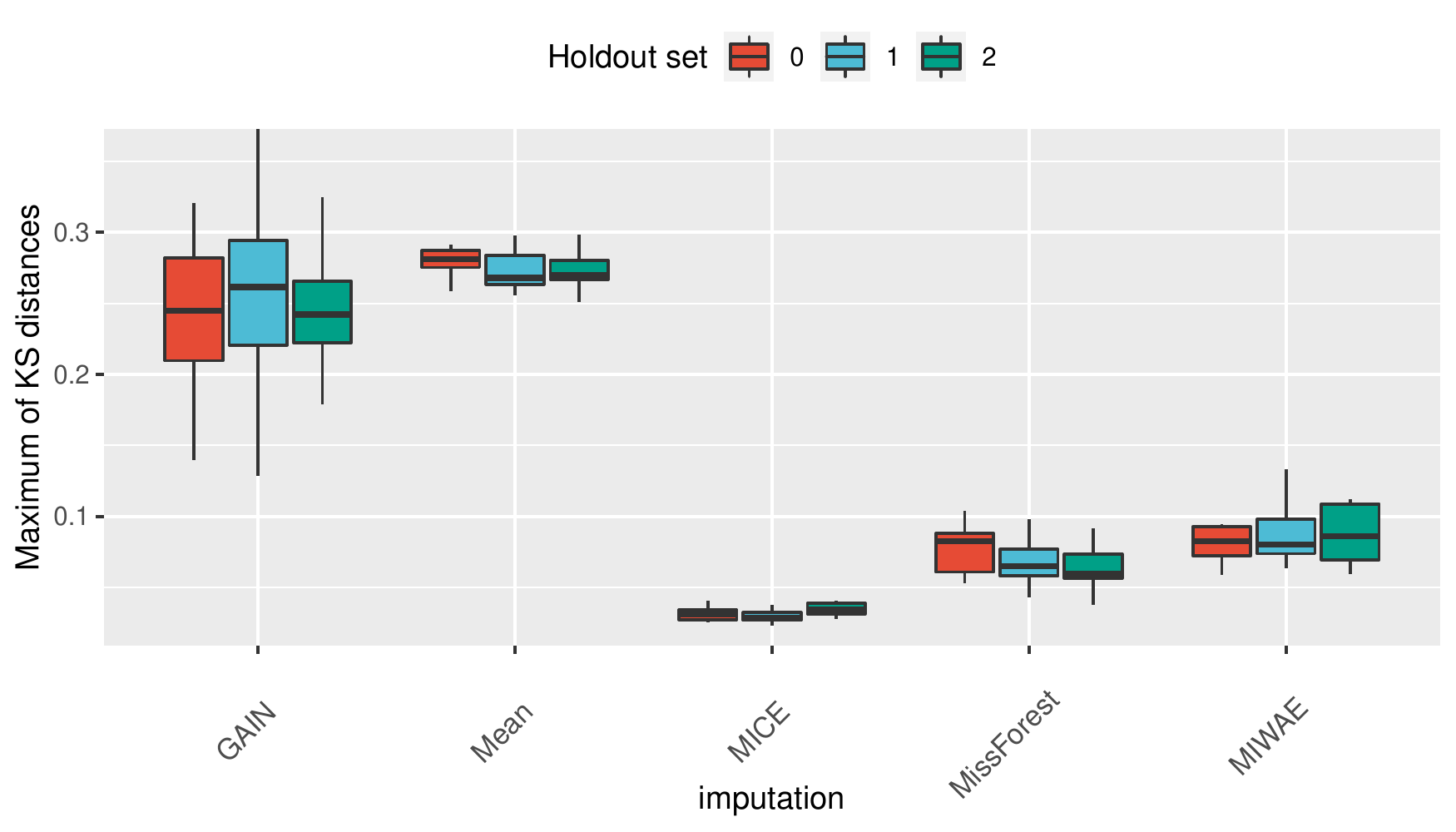}\\
      \hline
      \parbox[c][][c]{0.5in}{\rotatebox[origin=c]{90}{B3: Wasserstein}} &
      \includegraphics[height=4cm,width=5cm]{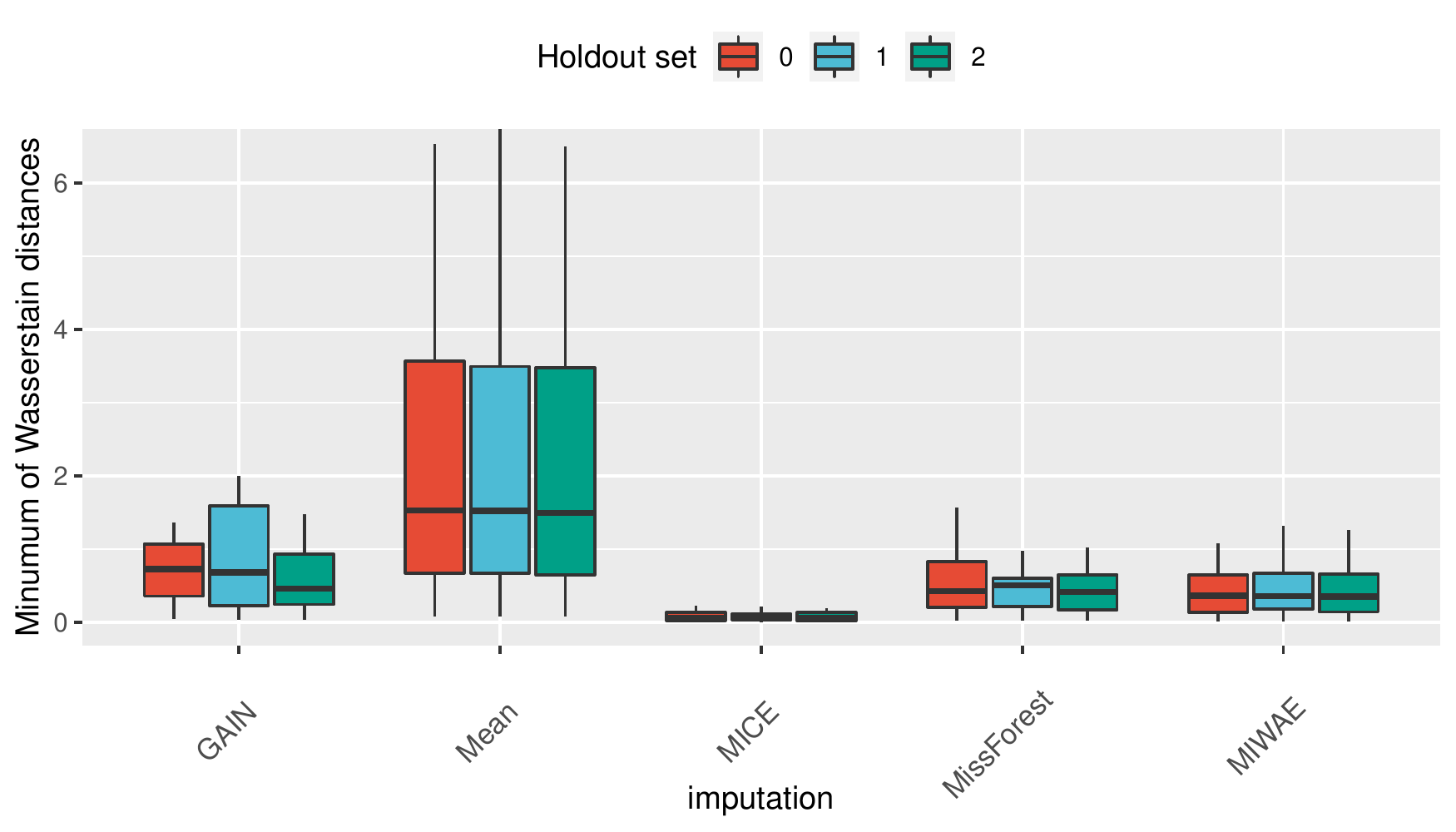}&
      \includegraphics[height=4cm,width=5cm]{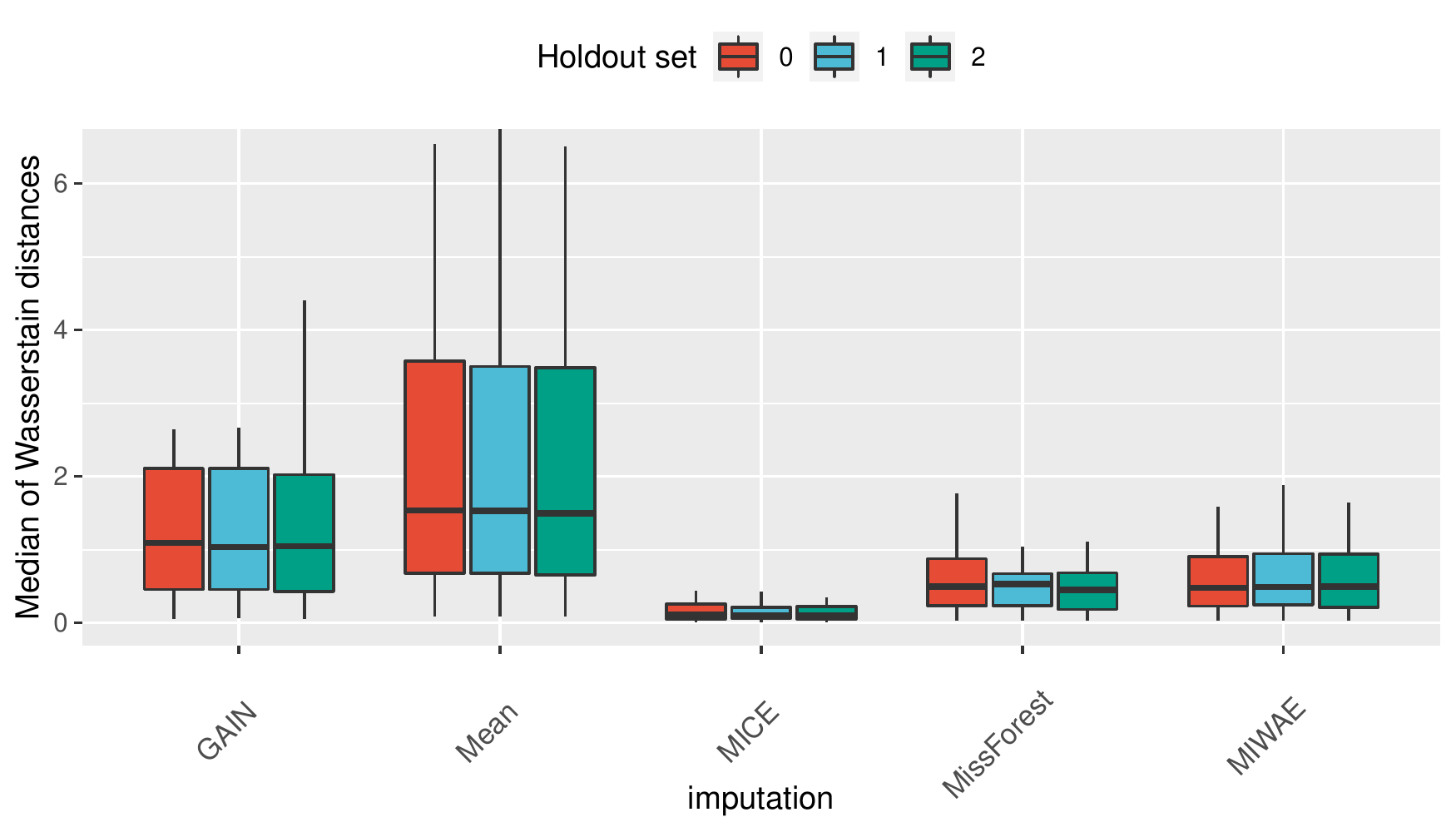}&
      \includegraphics[height=4cm,width=5cm]{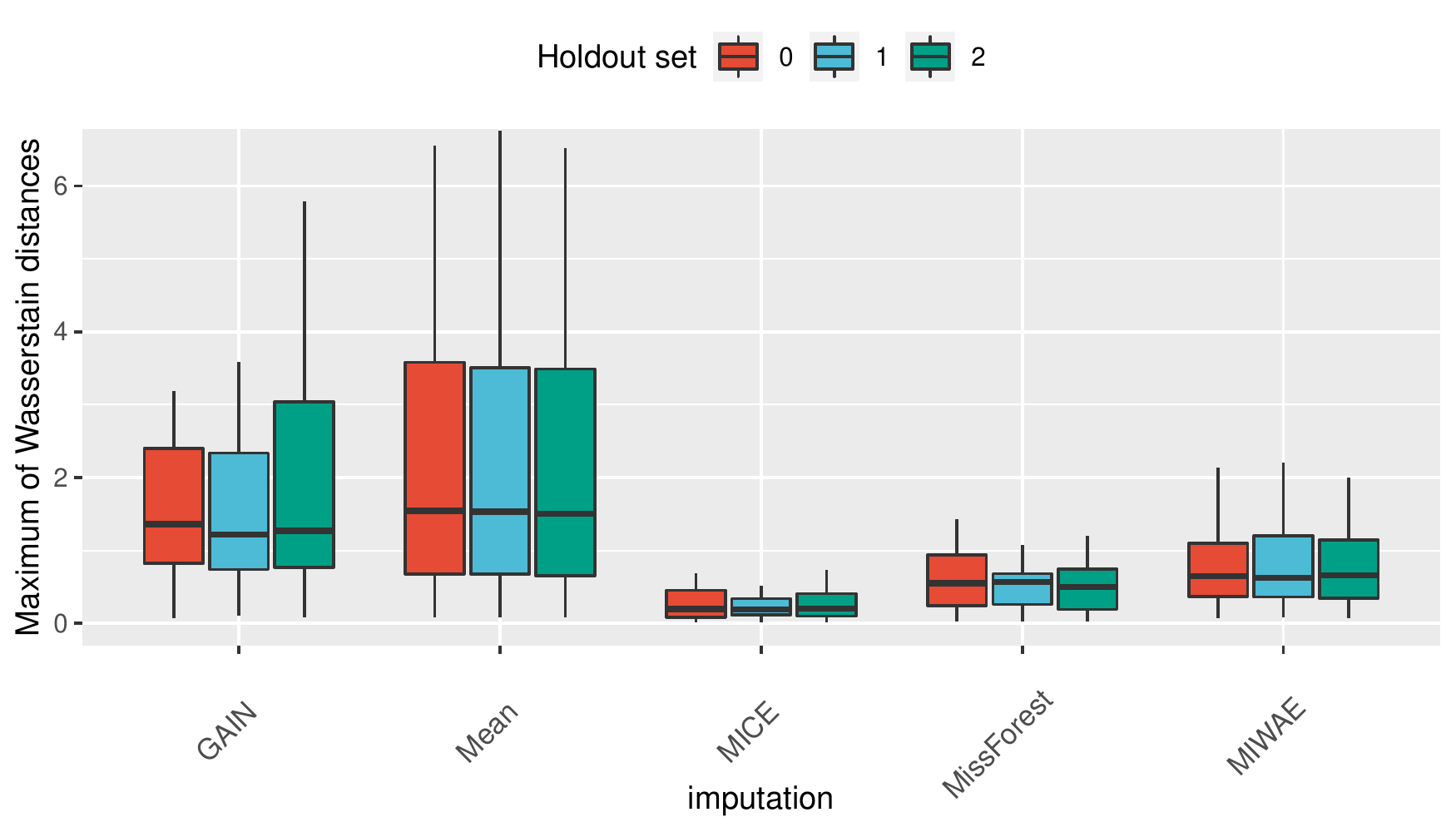}\\
      \hline
      \parbox[c][][c]{0.5in}{\rotatebox[origin=c]{90}{B3 excl. Mean}} &
      \includegraphics[height=4cm,width=5cm]{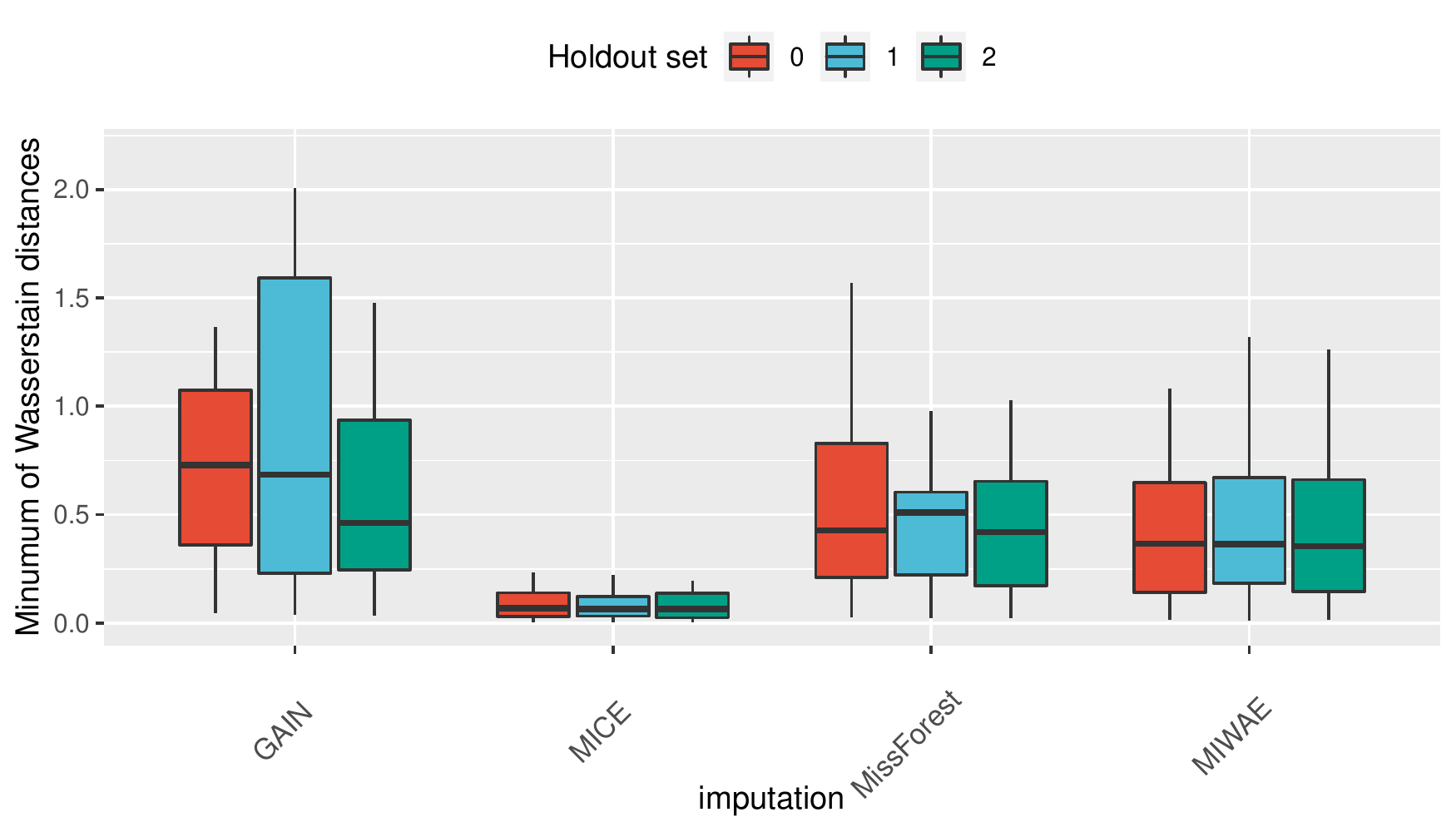}&
      \includegraphics[height=4cm,width=5cm]{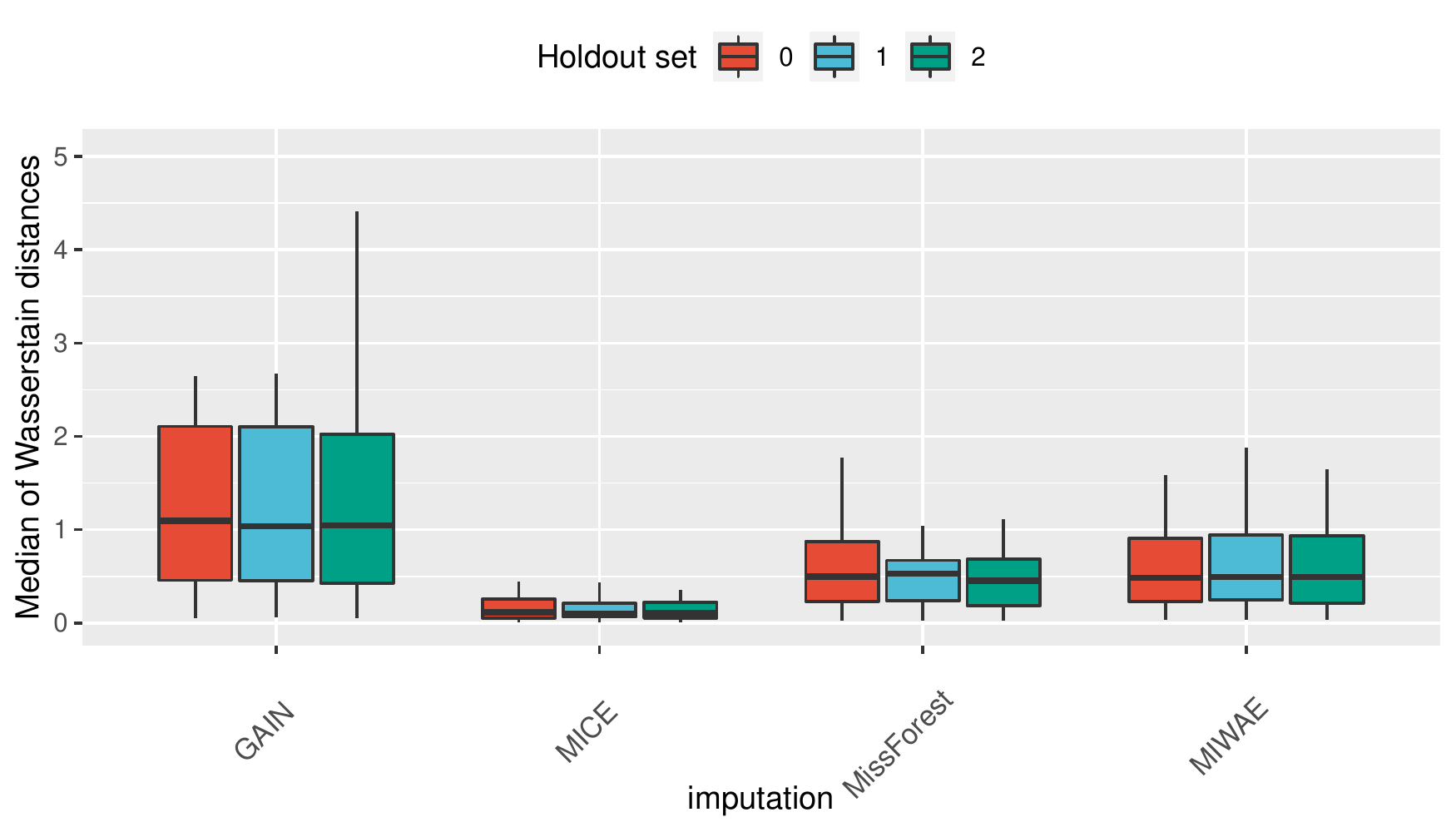}&
      \includegraphics[height=4cm,width=5cm]{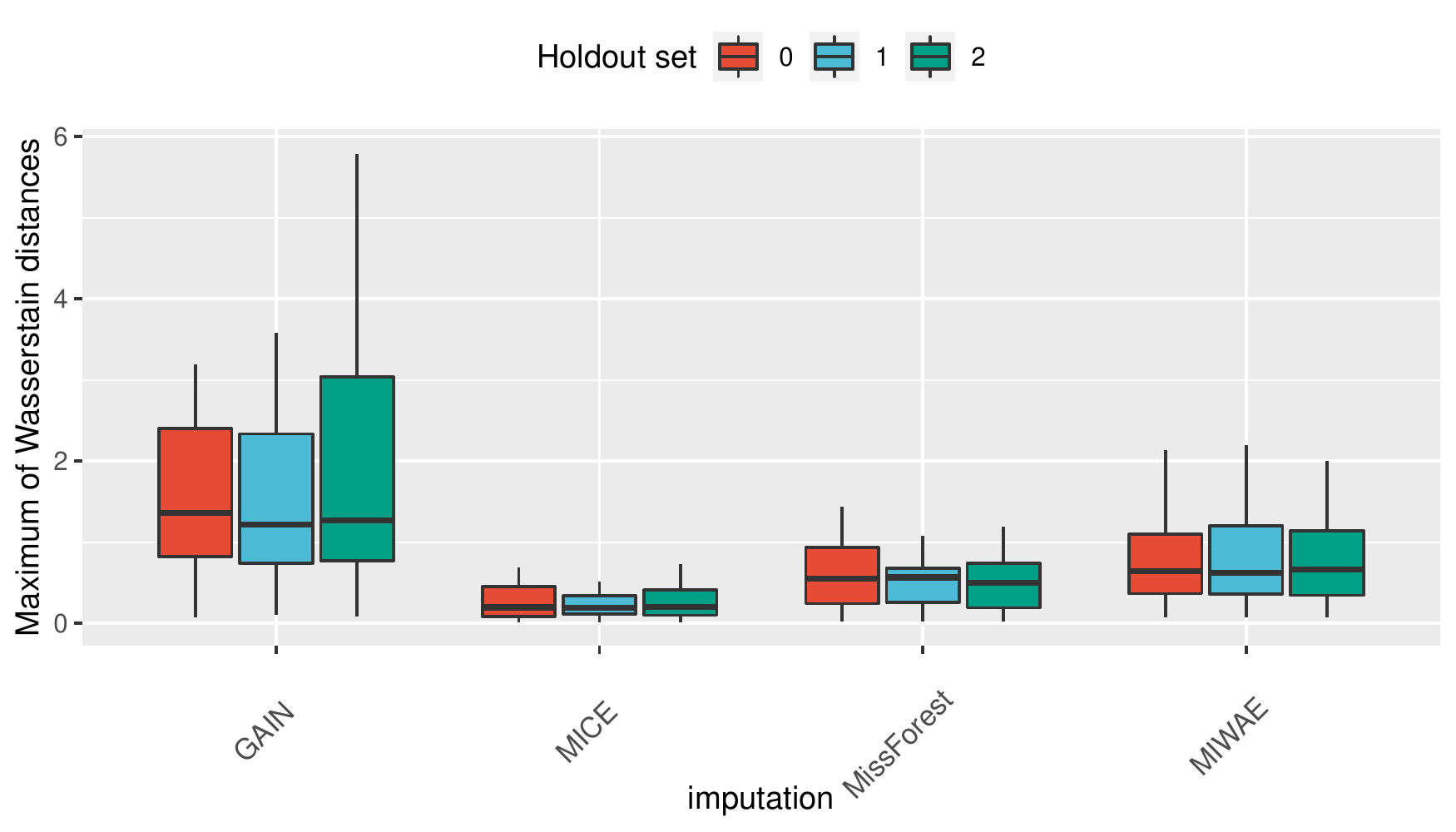}\\
    \end{tabular}
    \caption{Feature-wise 25\% train missingness and 50\% test missingness.}
    \label{fig:featurewise_mimic_25_50}
\end{figure}

\begin{figure}[htb!]
    \centering
    \begin{tabular}{m{0.2in} | M{5cm} | M{5cm} | M{5cm}}
    & \textbf{Minimum} & \textbf{Median} & \textbf{Maximum} \\
    \hline
     \parbox[c][][c]{0.5in}{\rotatebox[origin=t]{90}{B1: Kullback-Leibler}} &
      \includegraphics[height=4cm,width=5cm]{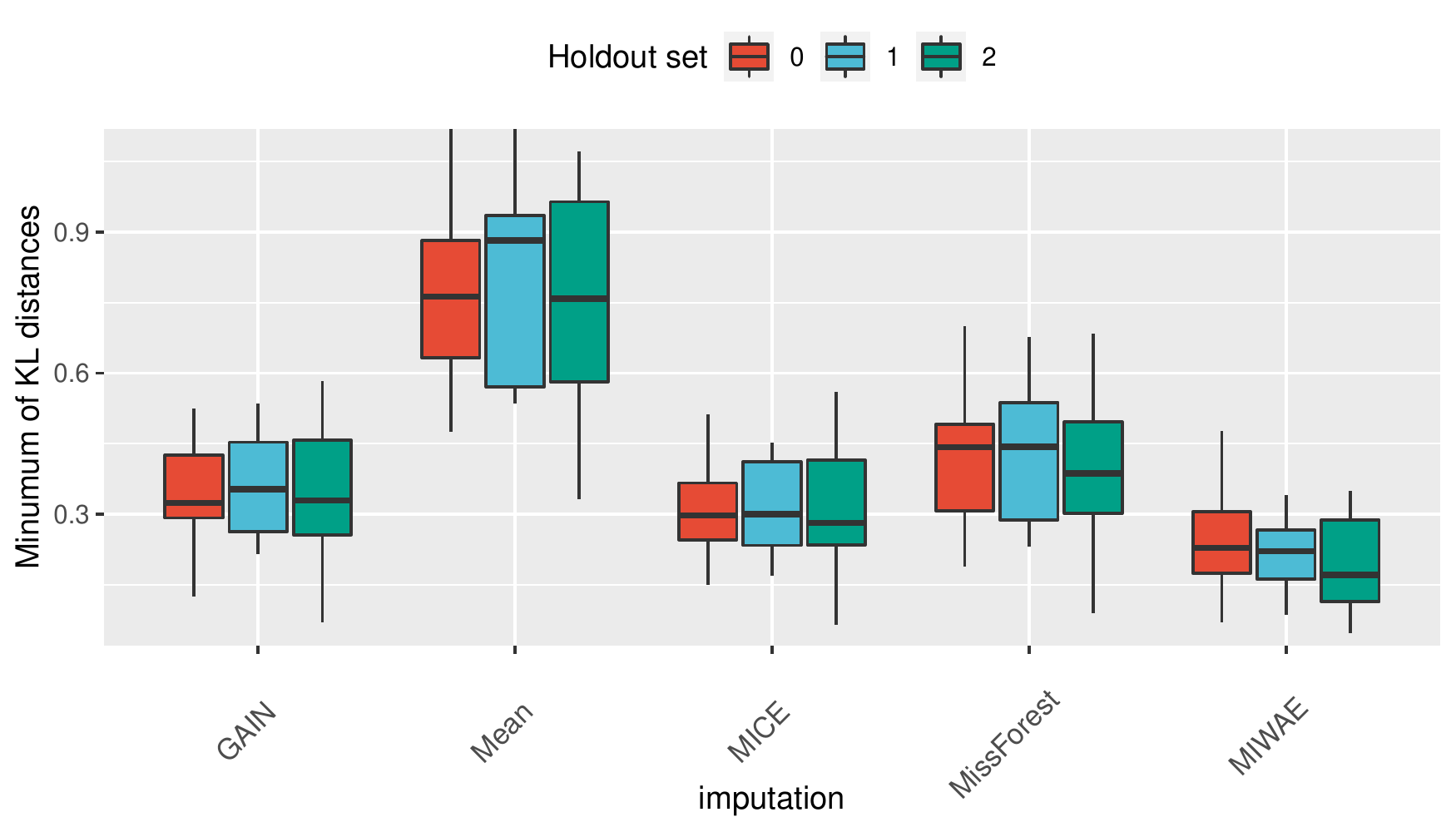}&
      \includegraphics[height=4cm,width=5cm]{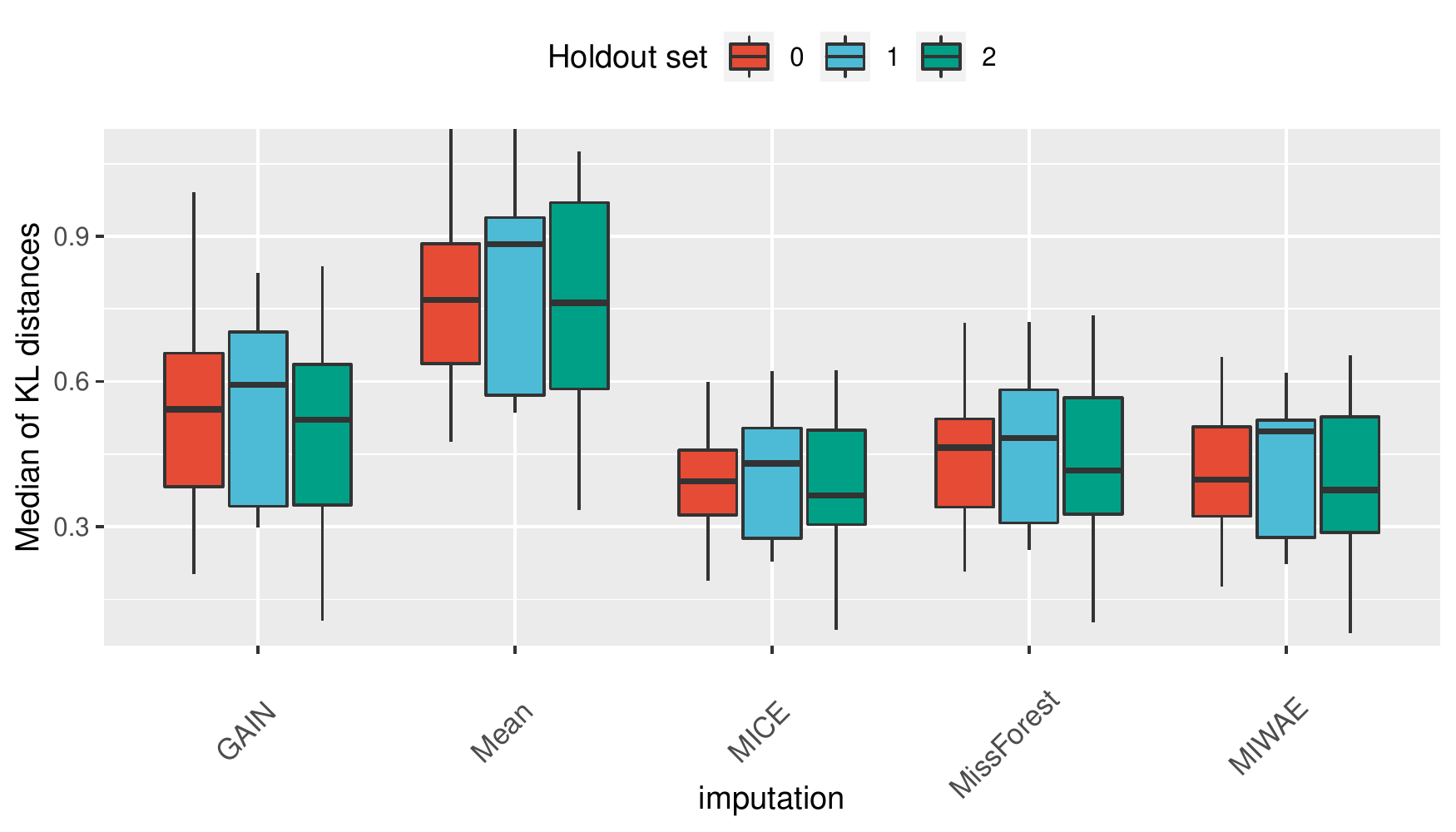}&
      \includegraphics[height=4cm,width=5cm]{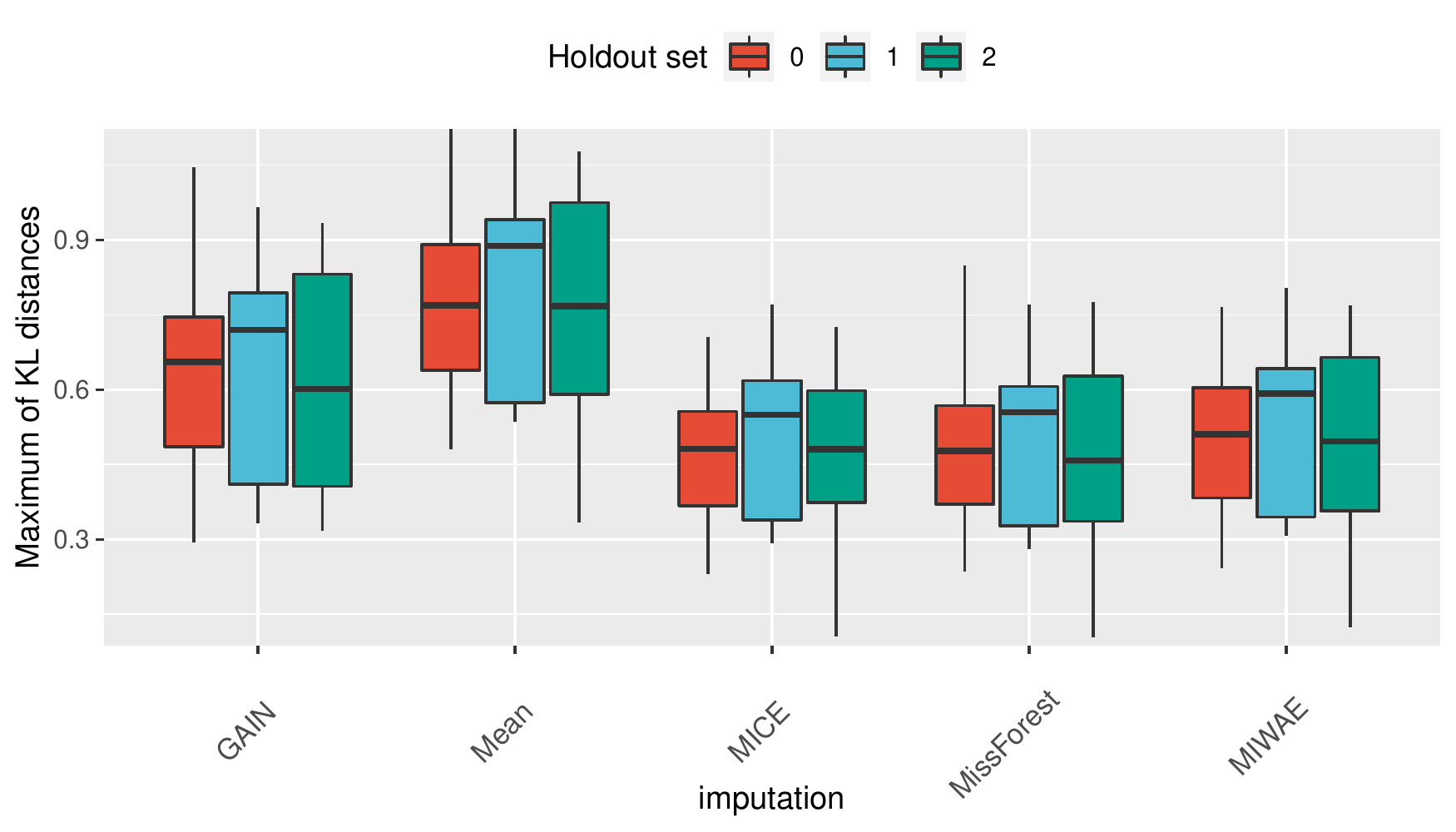}\\
      \hline
      \parbox[c][][c]{0.5in}{\rotatebox[origin=c]{90}{B2: Kolmogorov-Smirnov}} &
      \includegraphics[height=4cm,width=5cm]{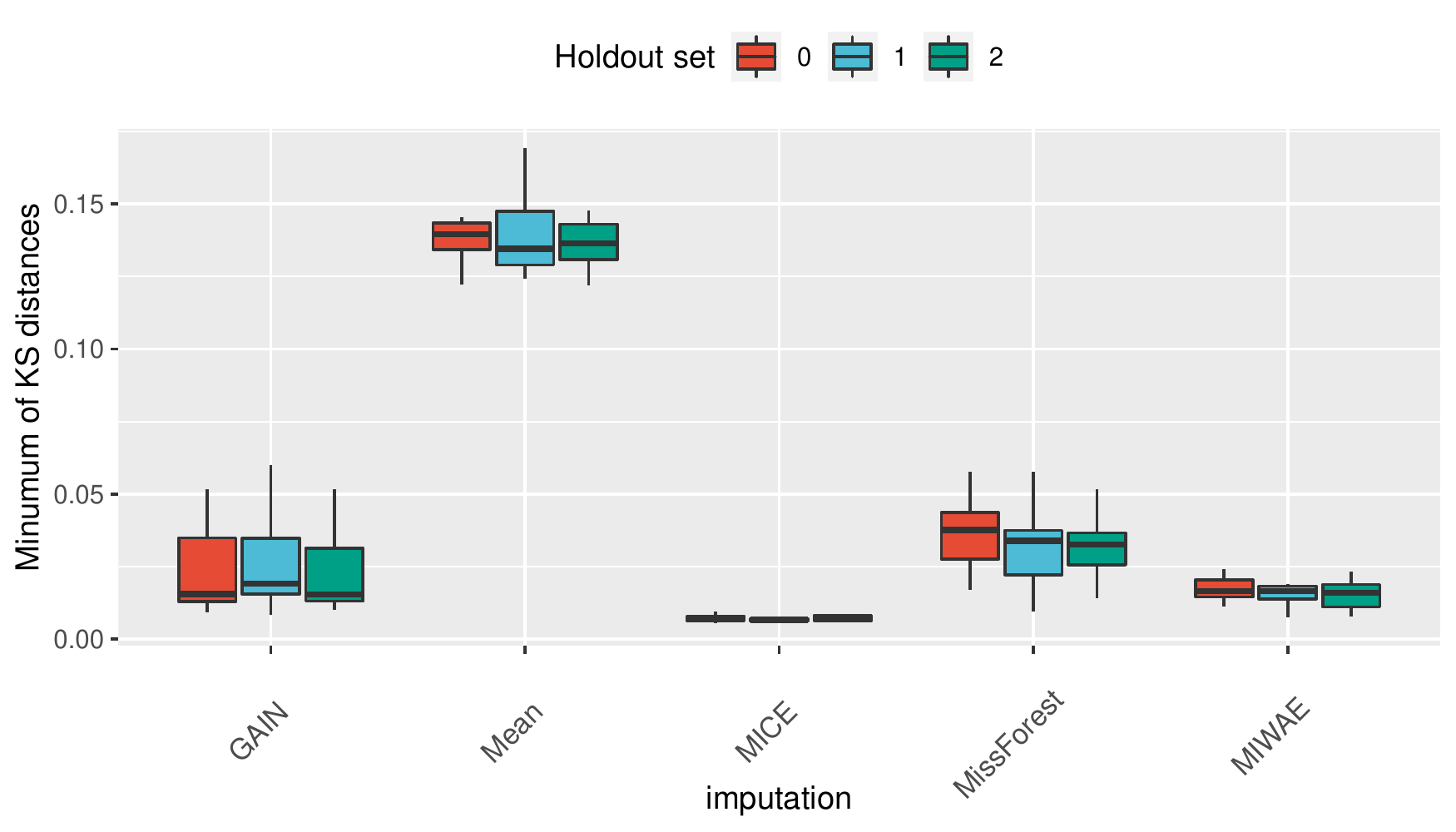}&
      \includegraphics[height=4cm,width=5cm]{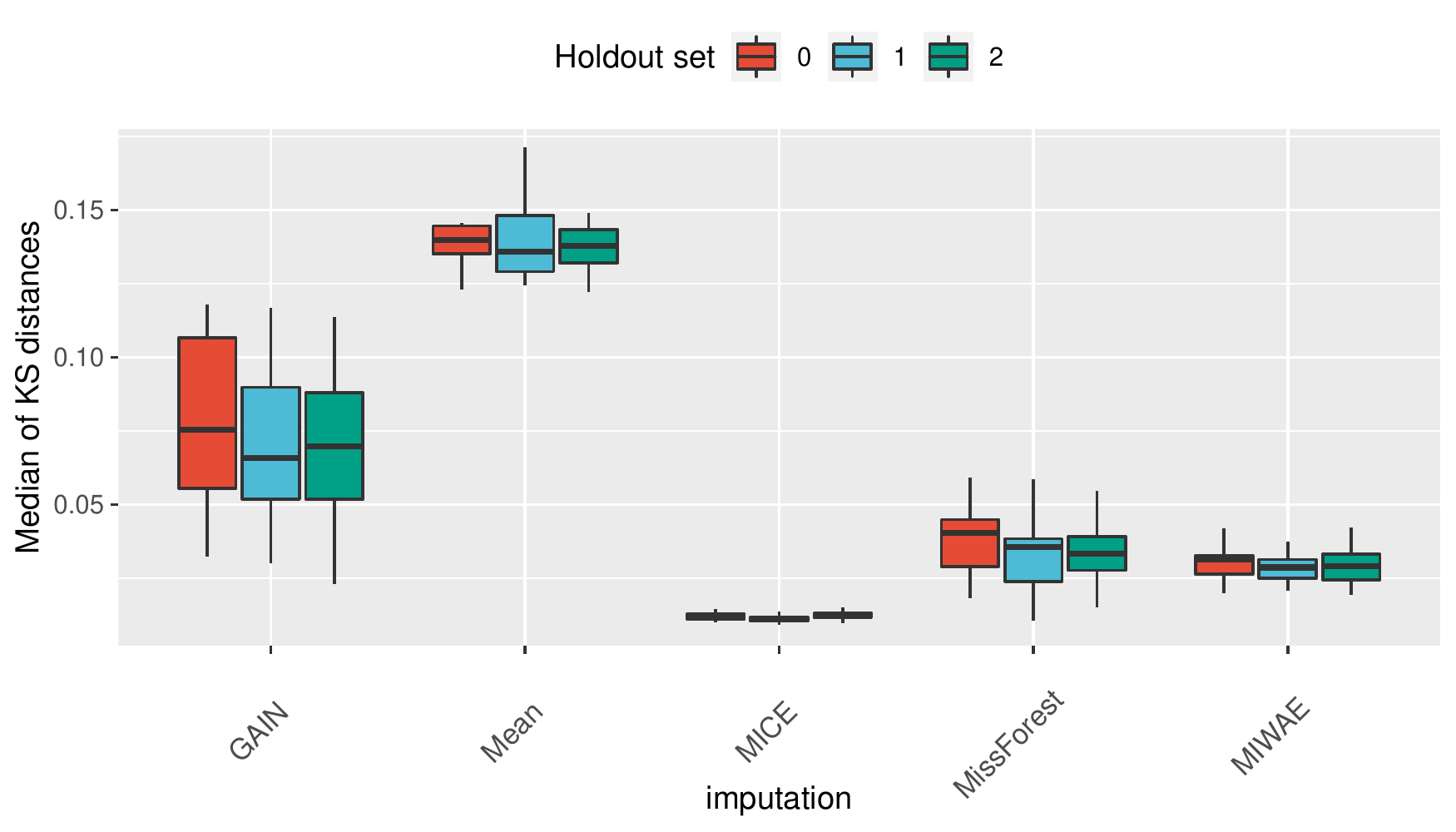}&
      \includegraphics[height=4cm,width=5cm]{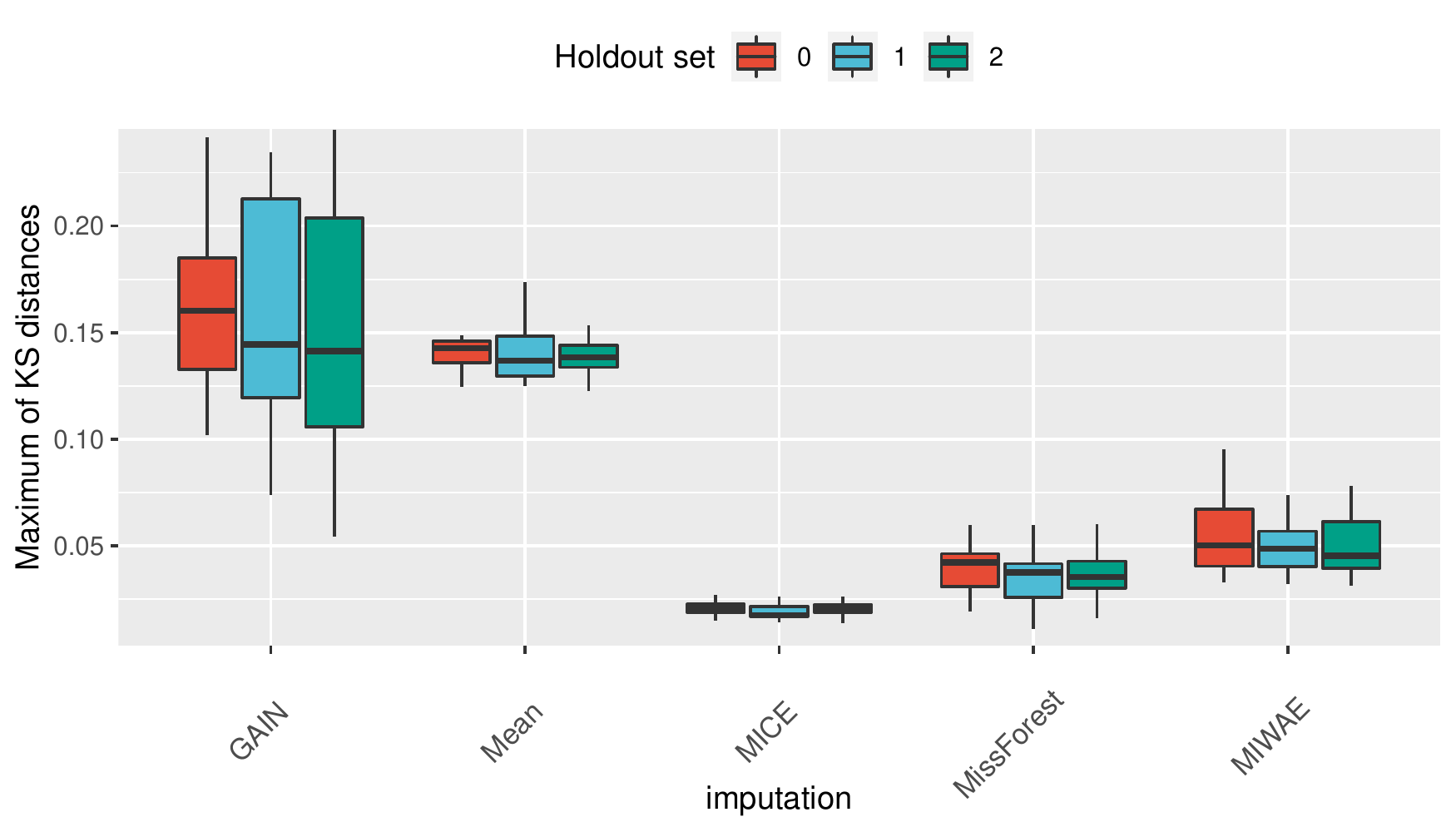}\\
      \hline
      \parbox[c][][c]{0.5in}{\rotatebox[origin=c]{90}{B3: Wasserstein}} &
      \includegraphics[height=4cm,width=5cm]{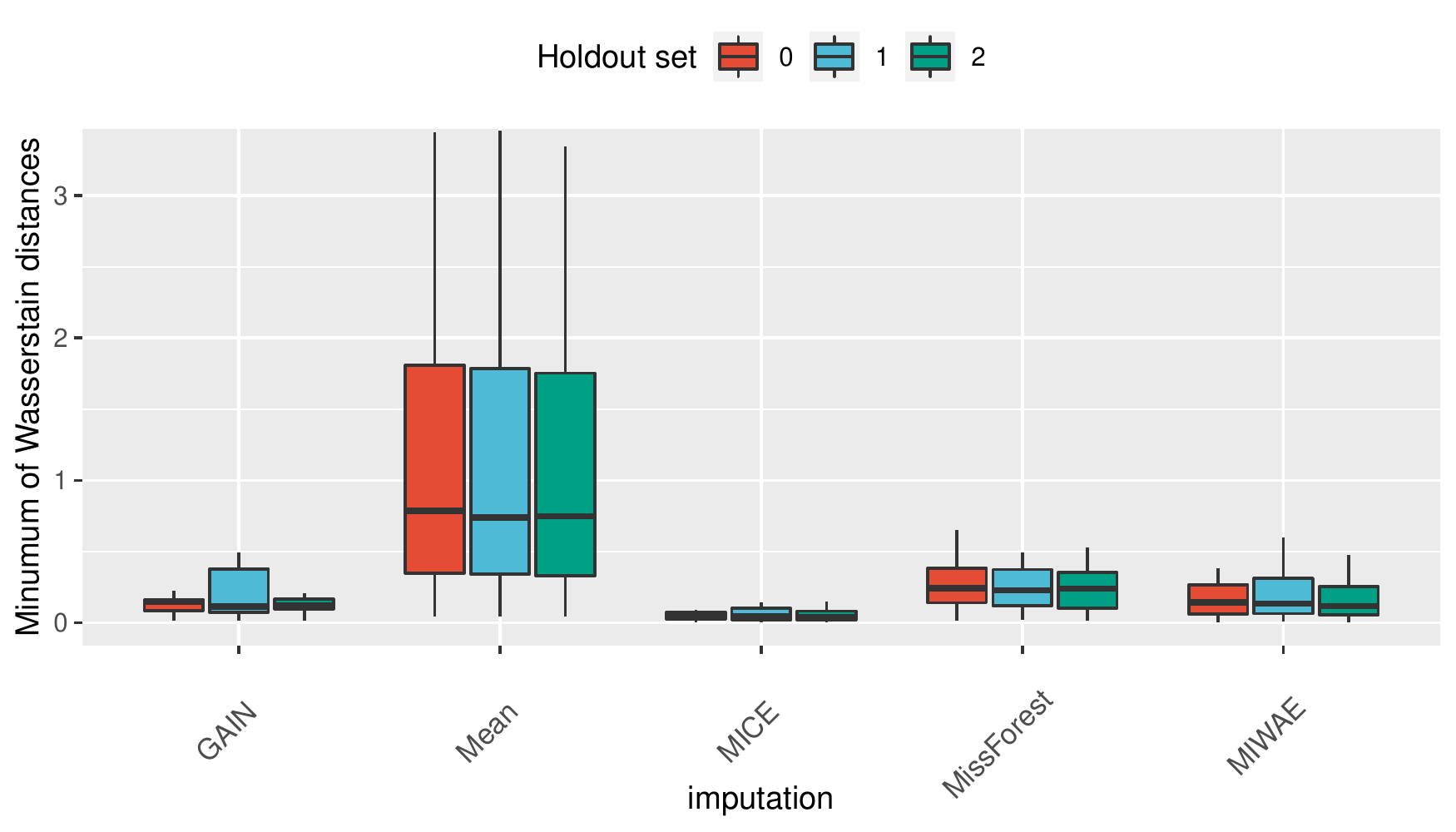}&
      \includegraphics[height=4cm,width=5cm]{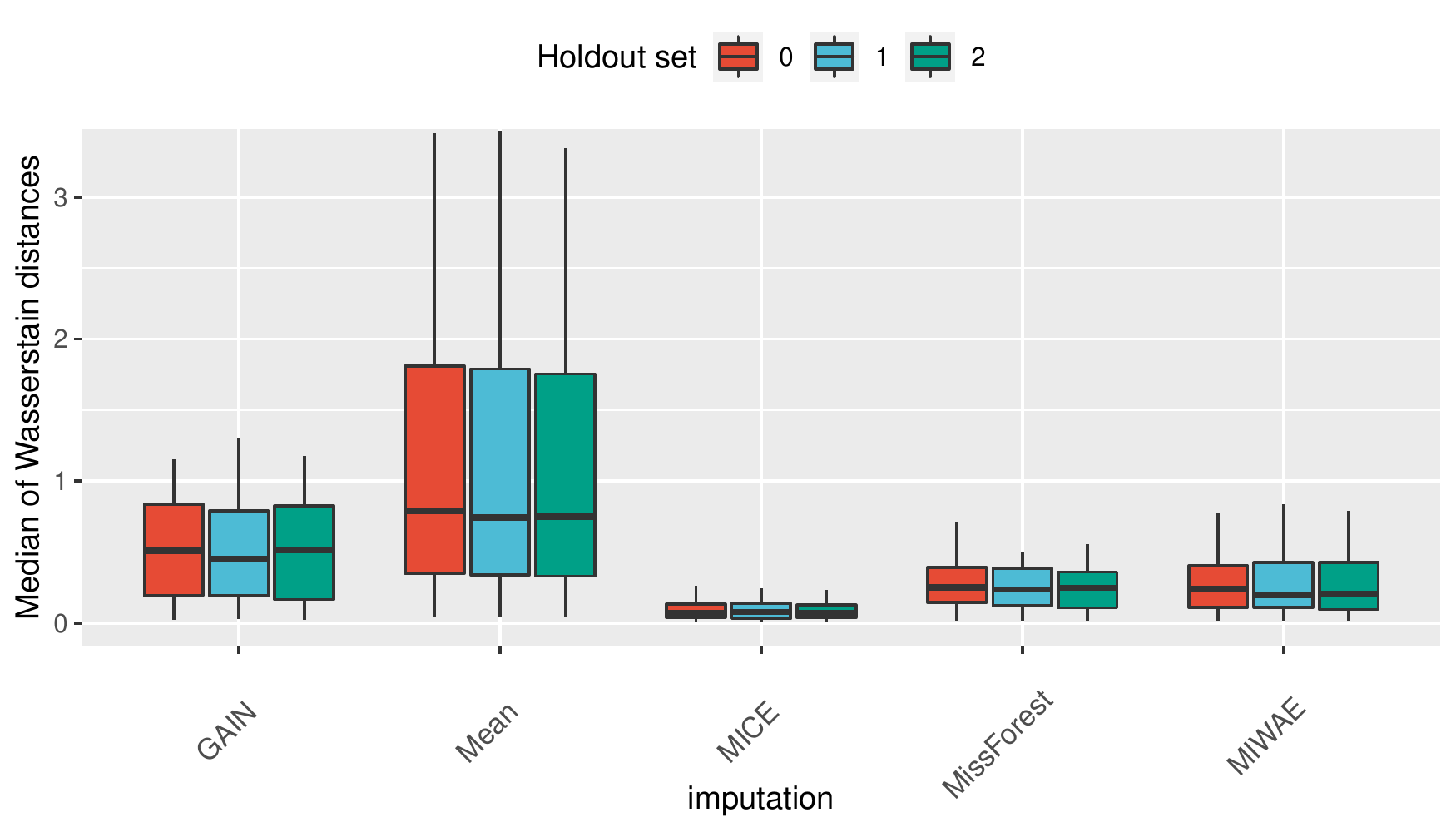}&
      \includegraphics[height=4cm,width=5cm]{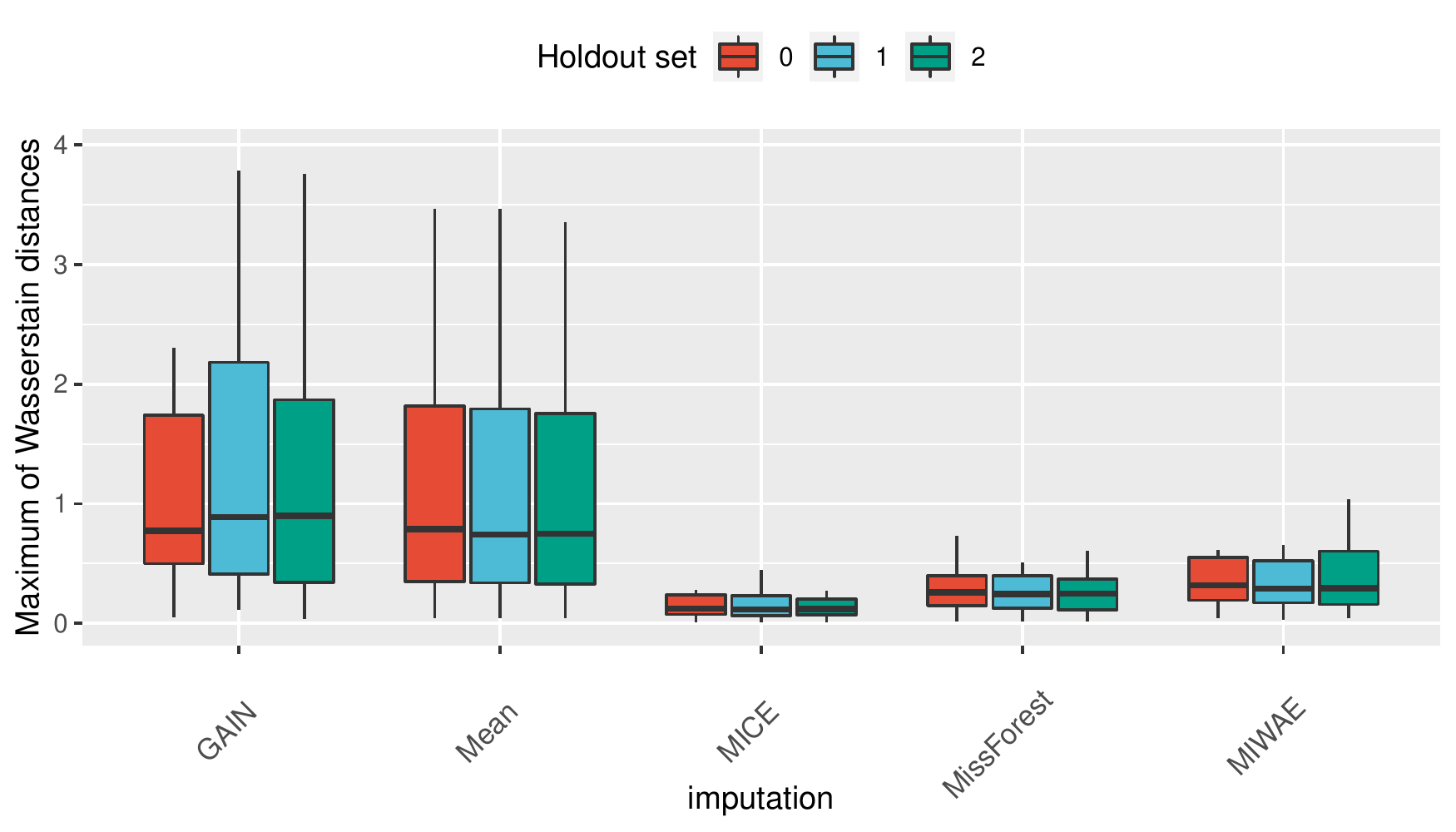}\\
      \hline
      \parbox[c][][c]{0.5in}{\rotatebox[origin=c]{90}{B3 excl. Mean}} &
      \includegraphics[height=4cm,width=5cm]{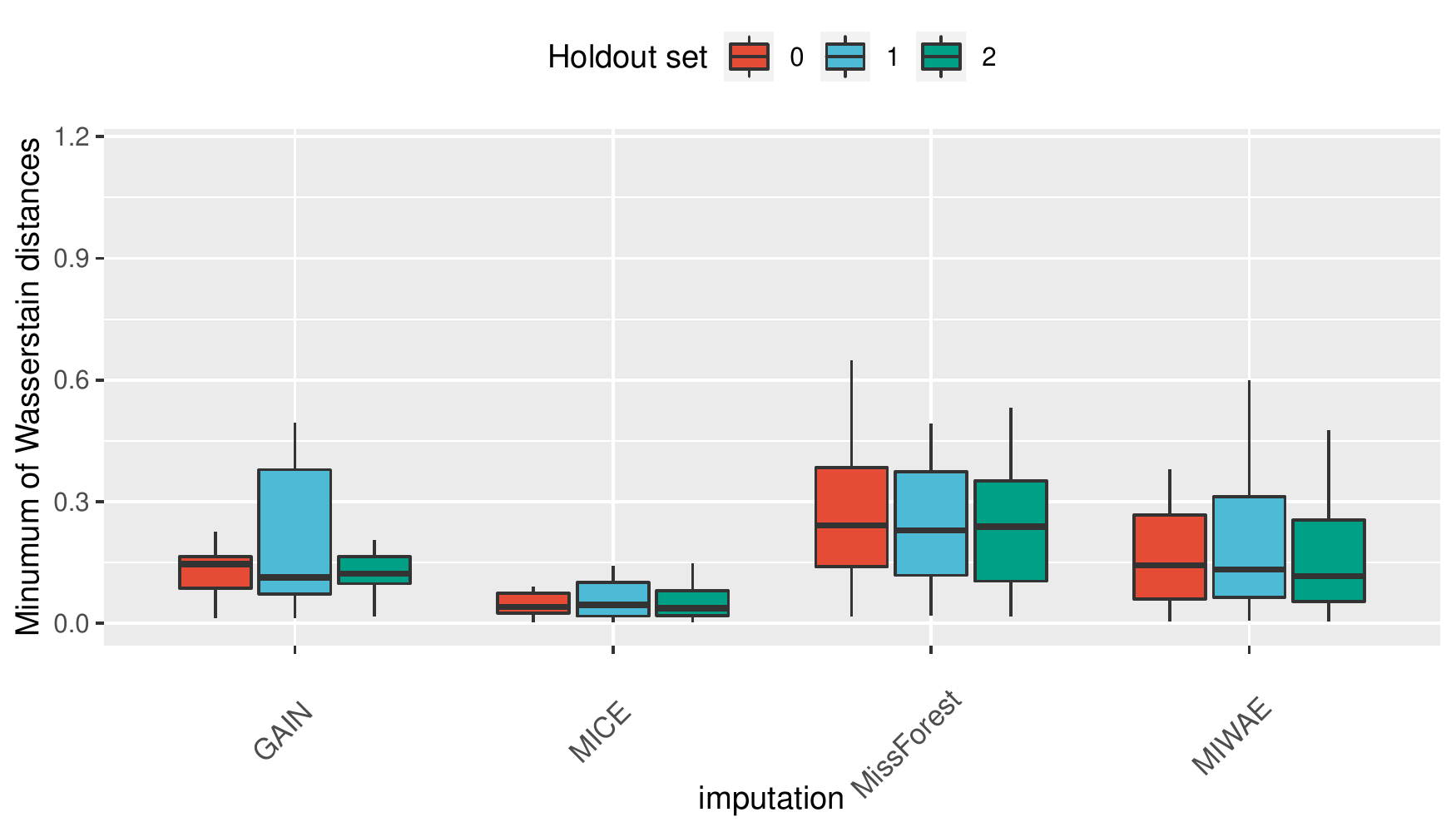}&
      \includegraphics[height=4cm,width=5cm]{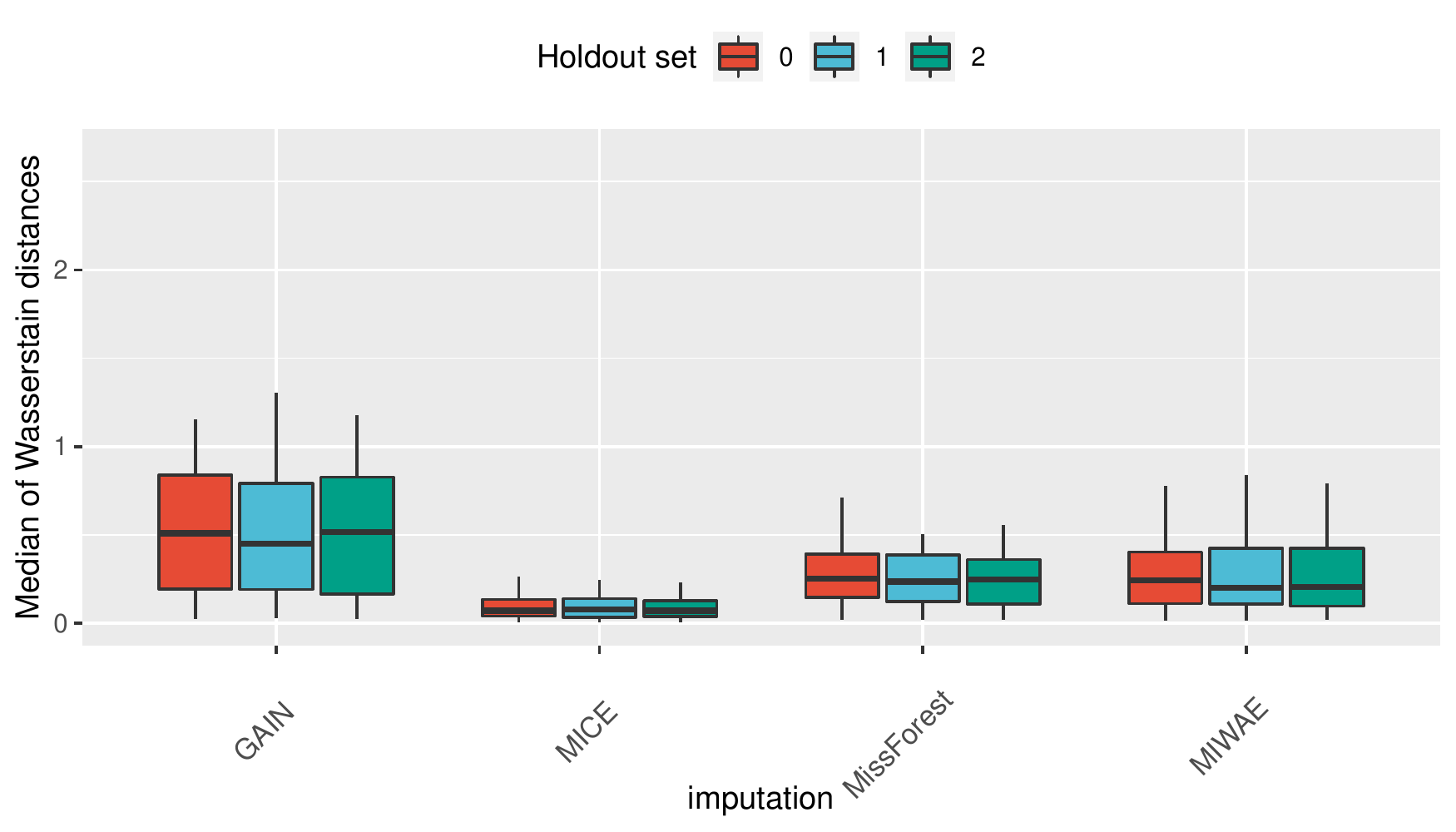}&
      \includegraphics[height=4cm,width=5cm]{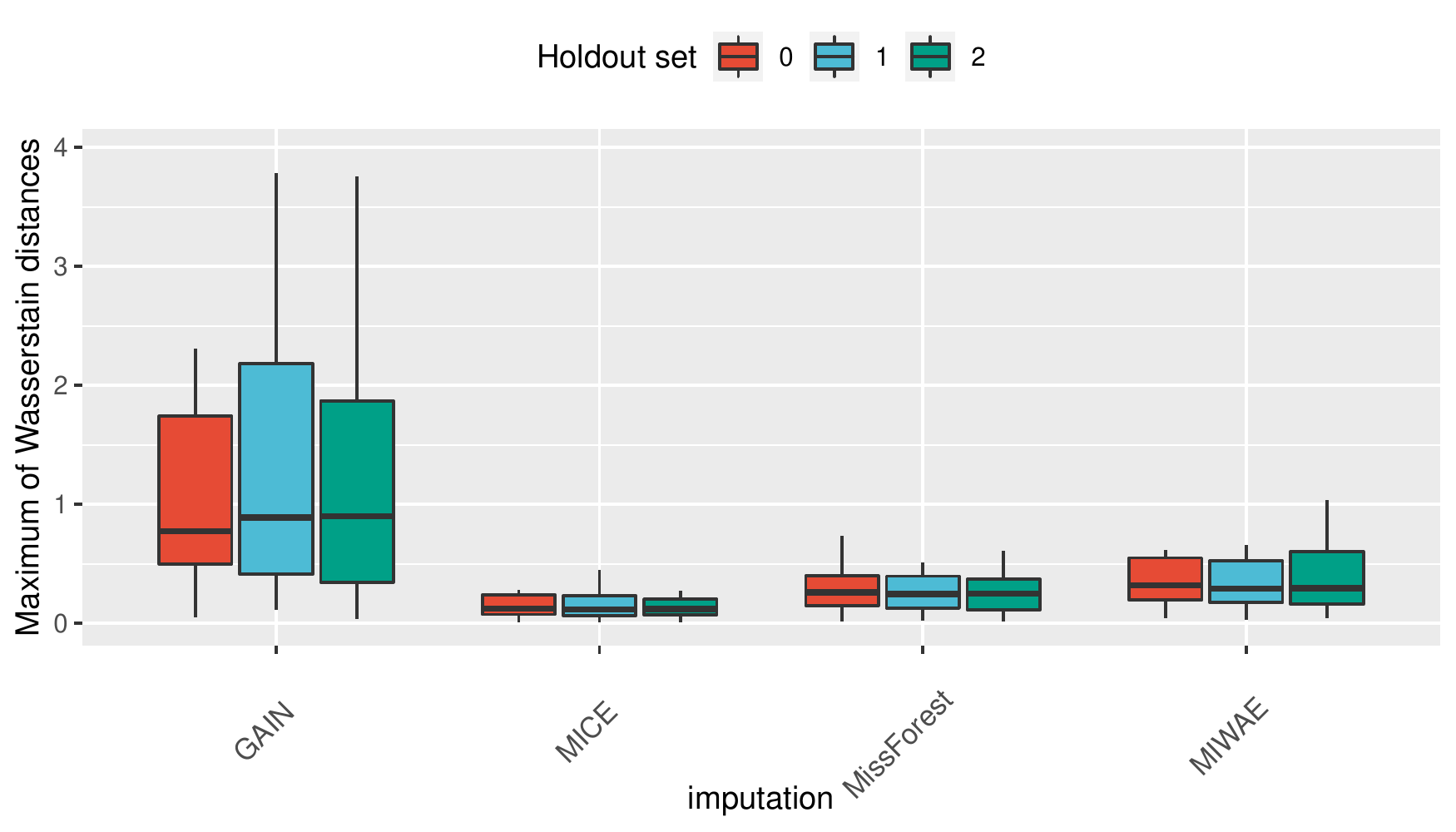}\\
    \end{tabular}
    \caption{Feature-wise 50\% train missingness and 25\% test missingness.}
    \label{fig:featurewise_mimic_50_25}
\end{figure}

\begin{figure}[htb!]
    \centering
    \begin{tabular}{m{0.2in} | M{5cm} | M{5cm} | M{5cm}}
    & \textbf{Minimum} & \textbf{Median} & \textbf{Maximum} \\
    \hline
     \parbox[c][][c]{0.5in}{\rotatebox[origin=t]{90}{B1: Kullback-Leibler}} &
      \includegraphics[height=4cm,width=5cm]{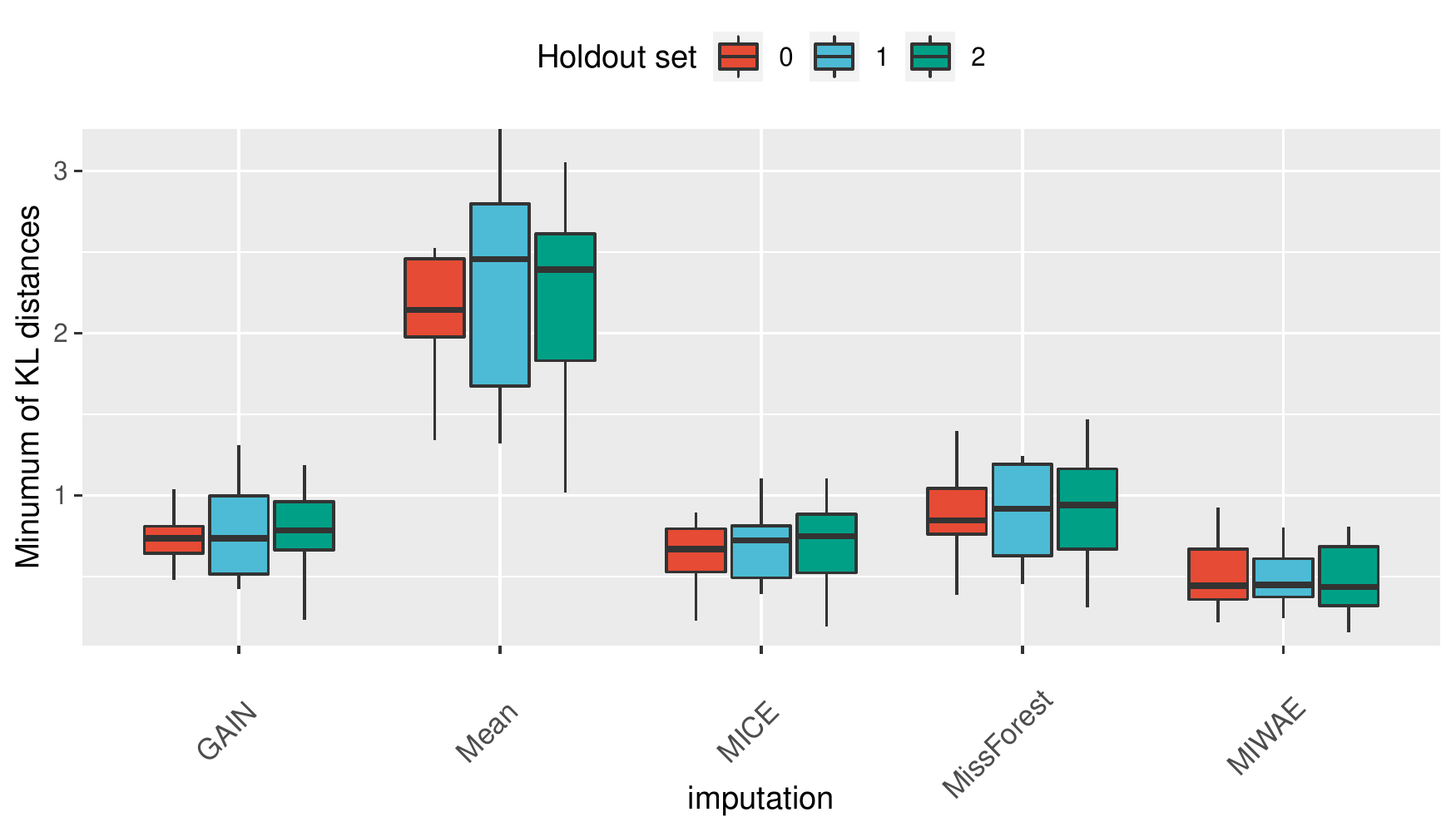}&
      \includegraphics[height=4cm,width=5cm]{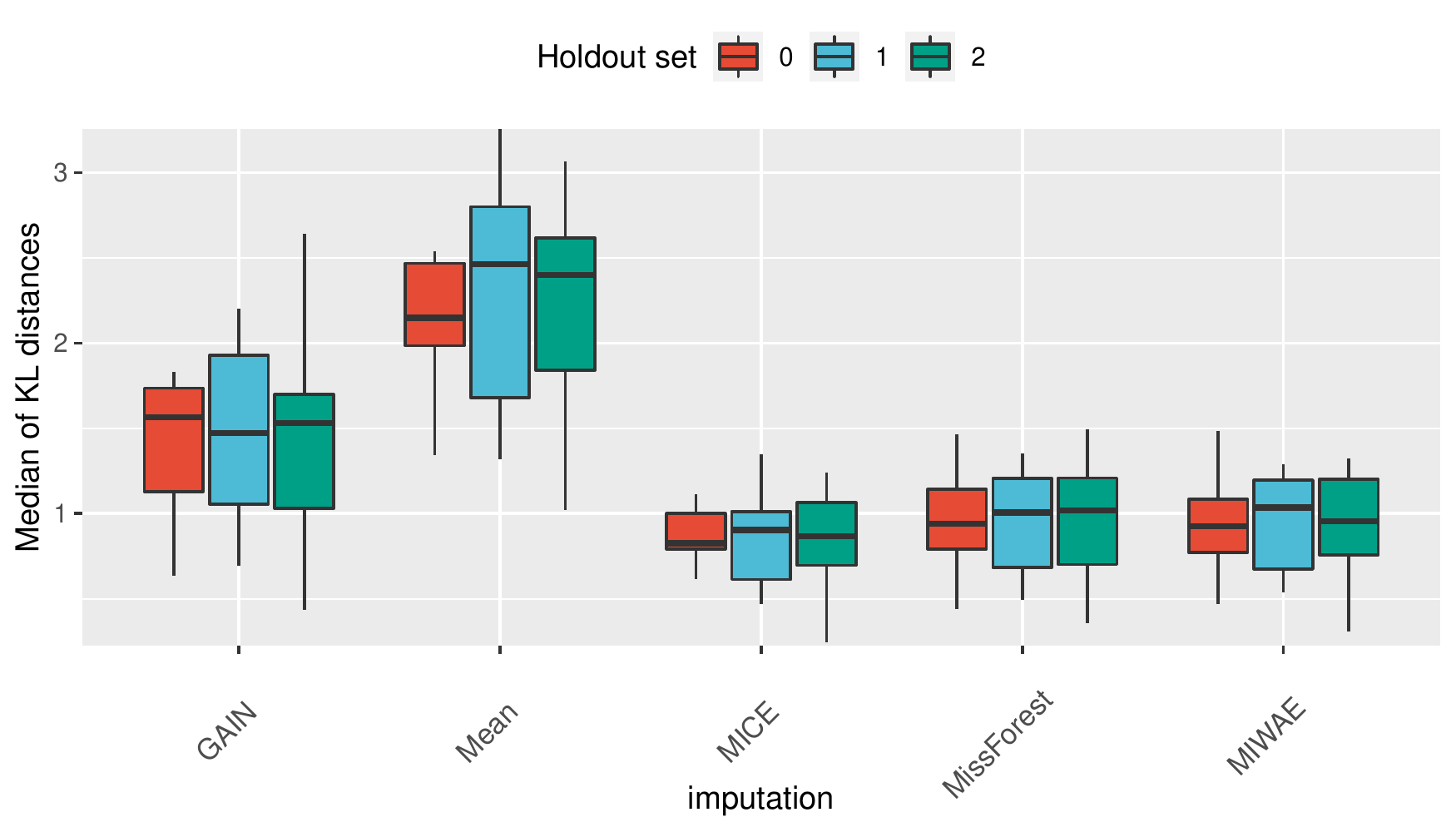}&
      \includegraphics[height=4cm,width=5cm]{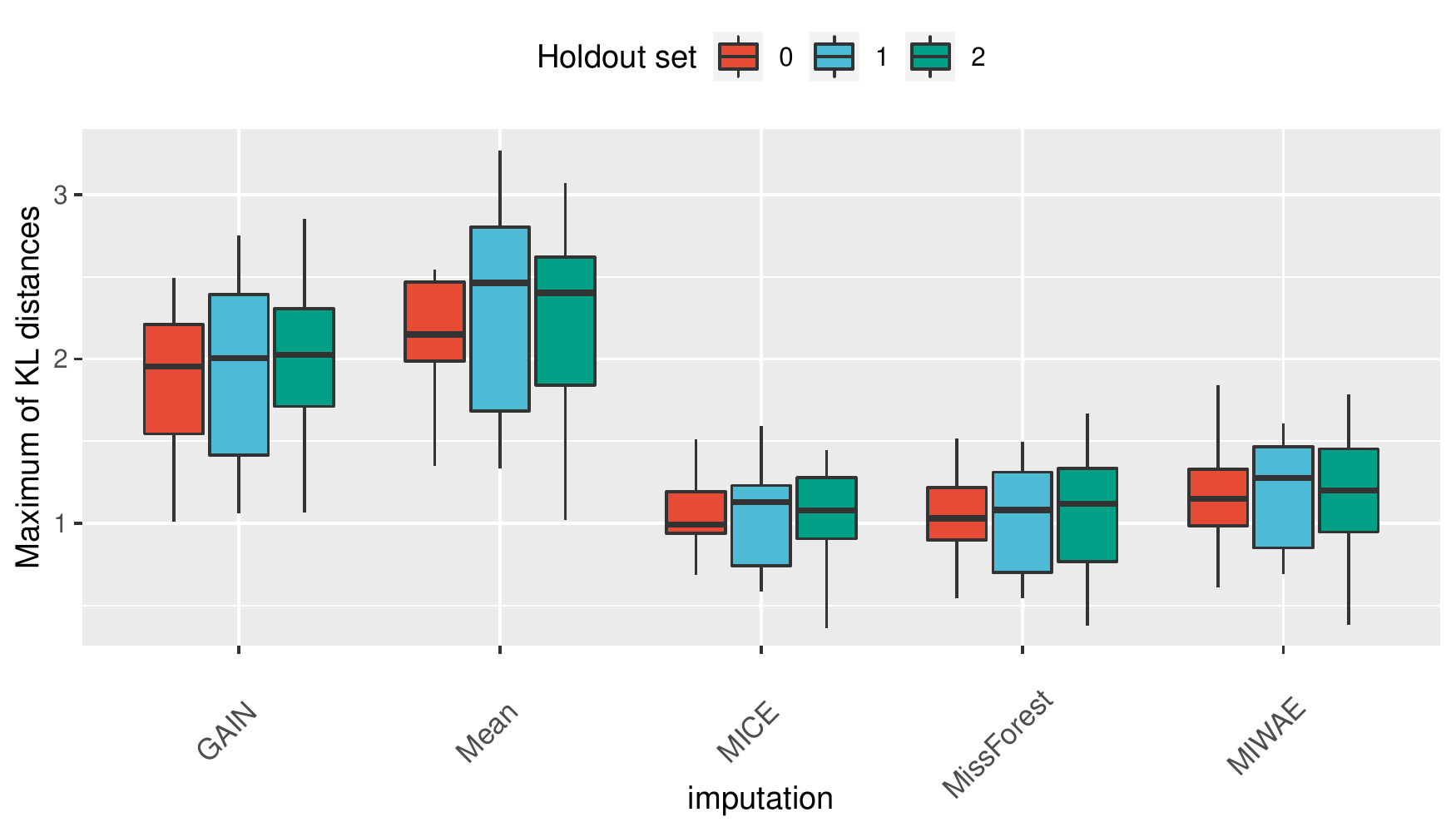}\\
      \hline
      \parbox[c][][c]{0.5in}{\rotatebox[origin=c]{90}{B2: Kolmogorov-Smirnov}} &
      \includegraphics[height=4cm,width=5cm]{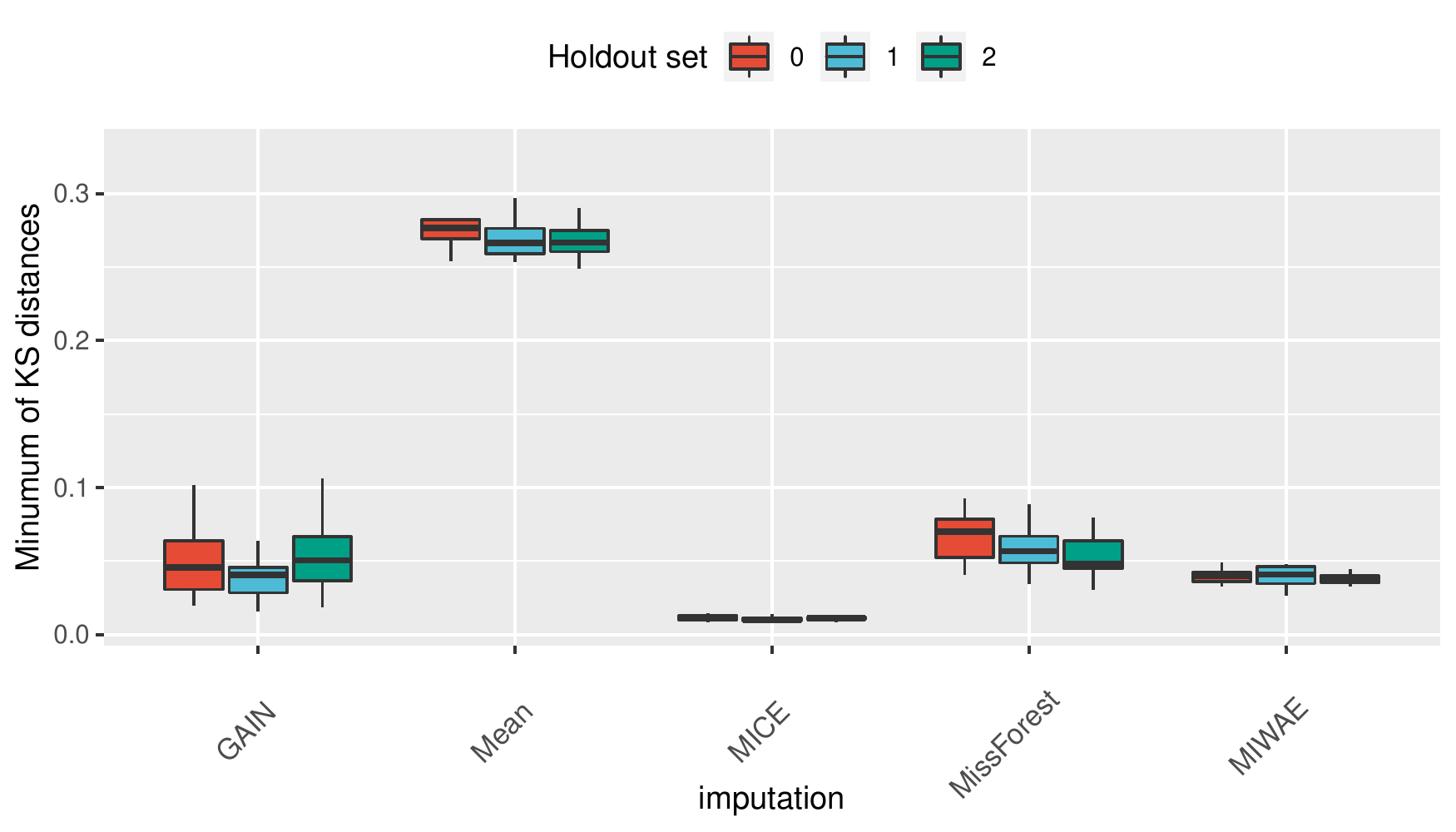}&
      \includegraphics[height=4cm,width=5cm]{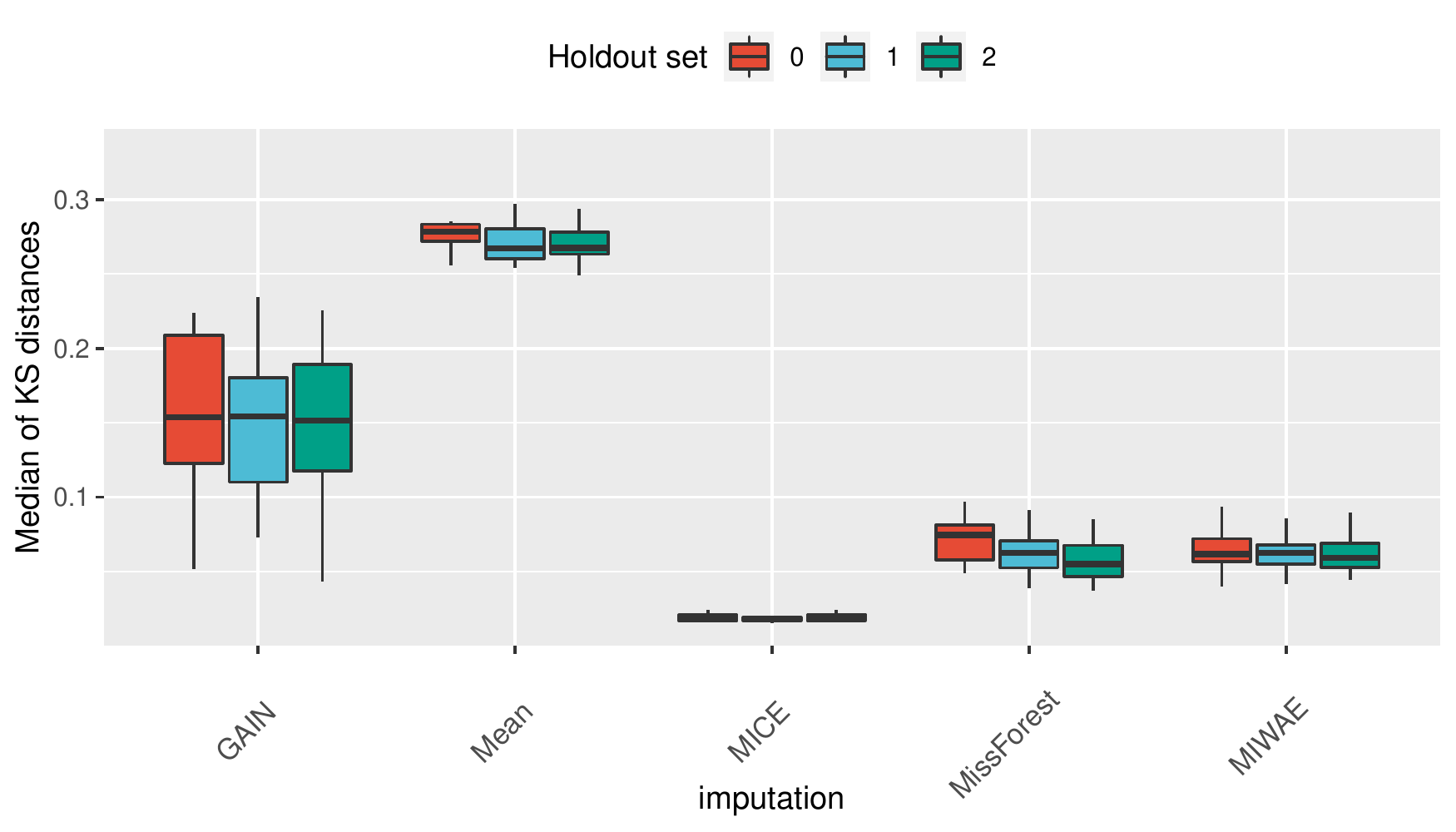}&
      \includegraphics[height=4cm,width=5cm]{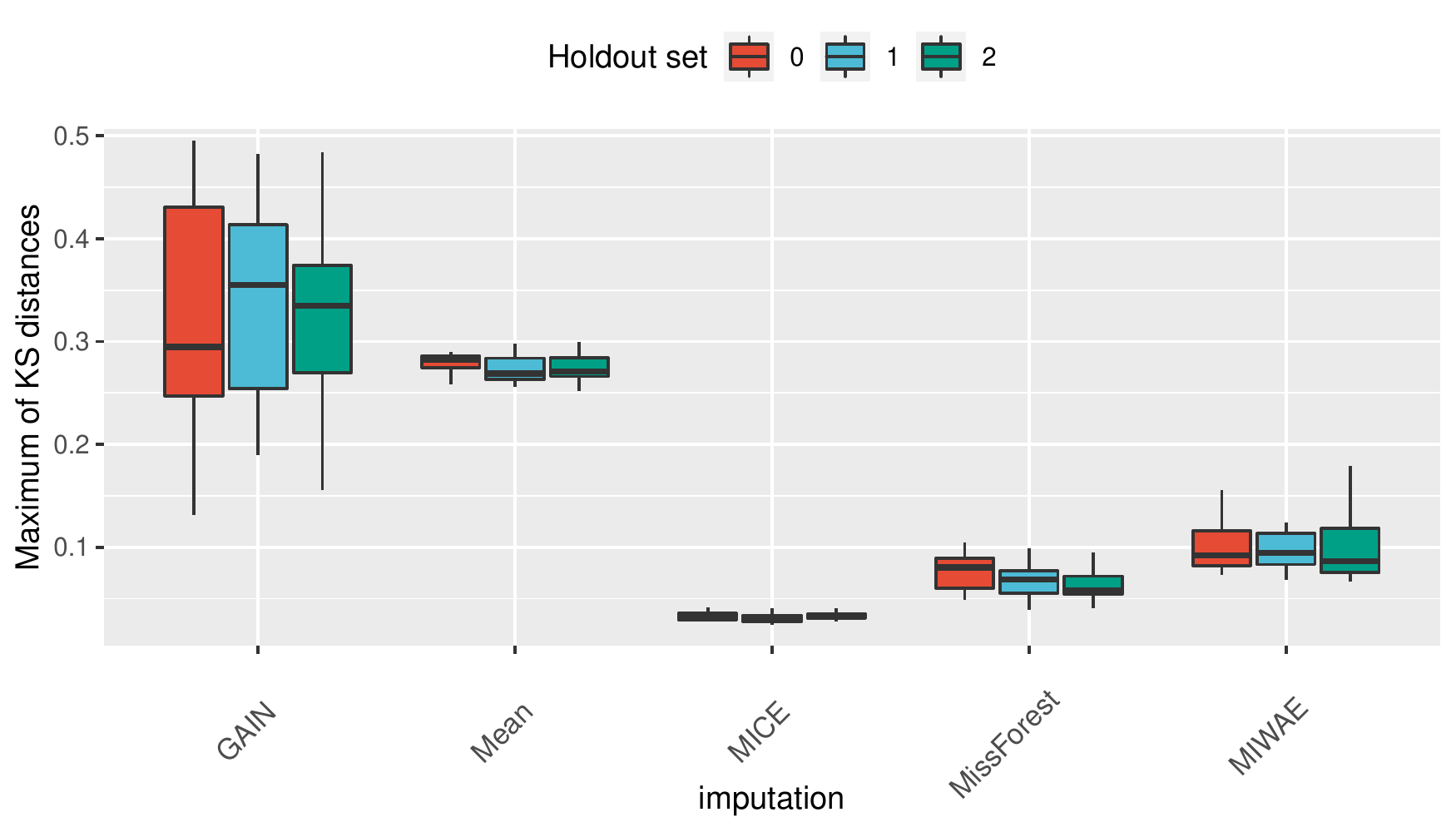}\\
      \hline
      \parbox[c][][c]{0.5in}{\rotatebox[origin=c]{90}{B3: Wasserstein}} &
      \includegraphics[height=4cm,width=5cm]{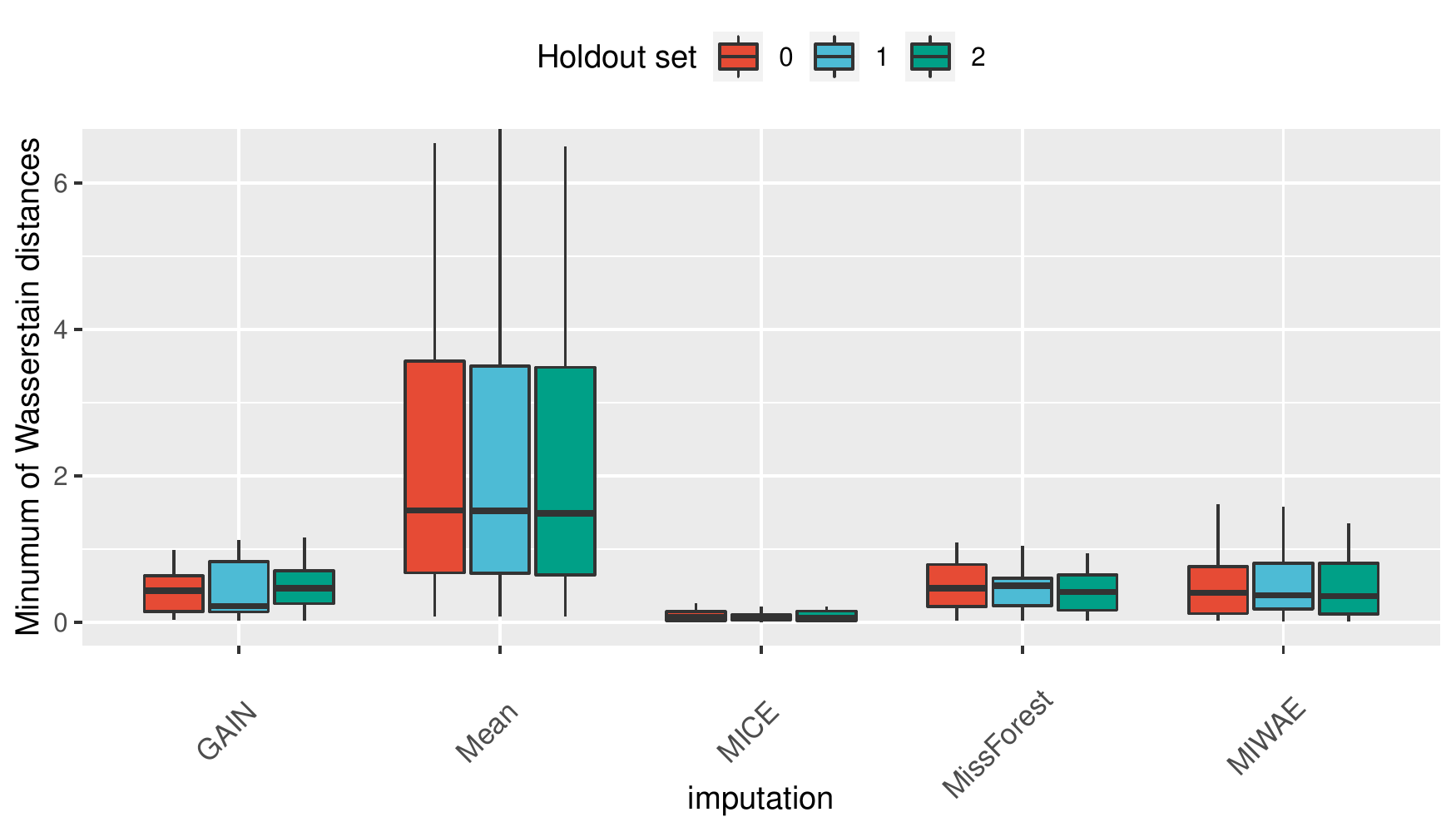}&
      \includegraphics[height=4cm,width=5cm]{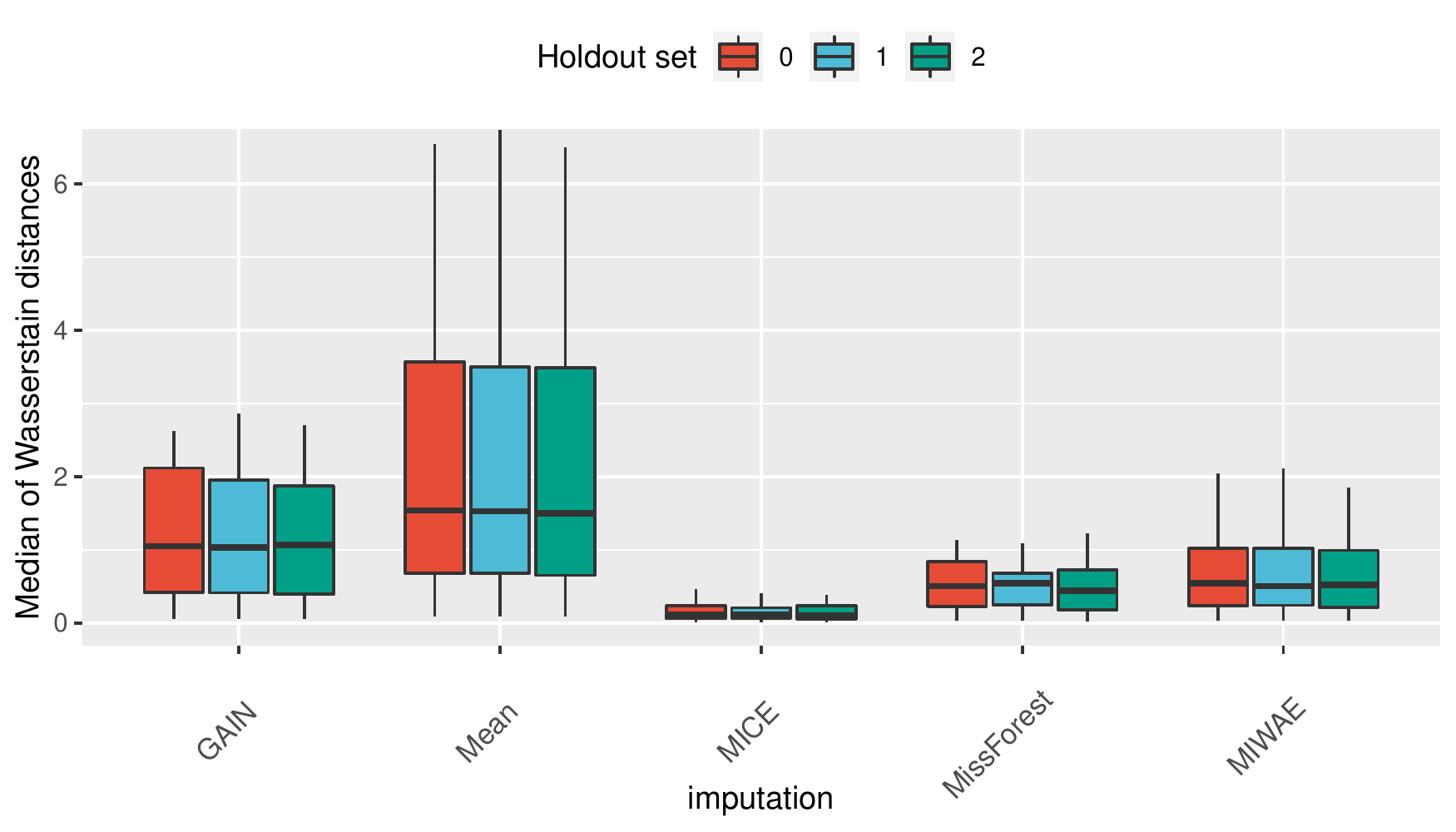}&
      \includegraphics[height=4cm,width=5cm]{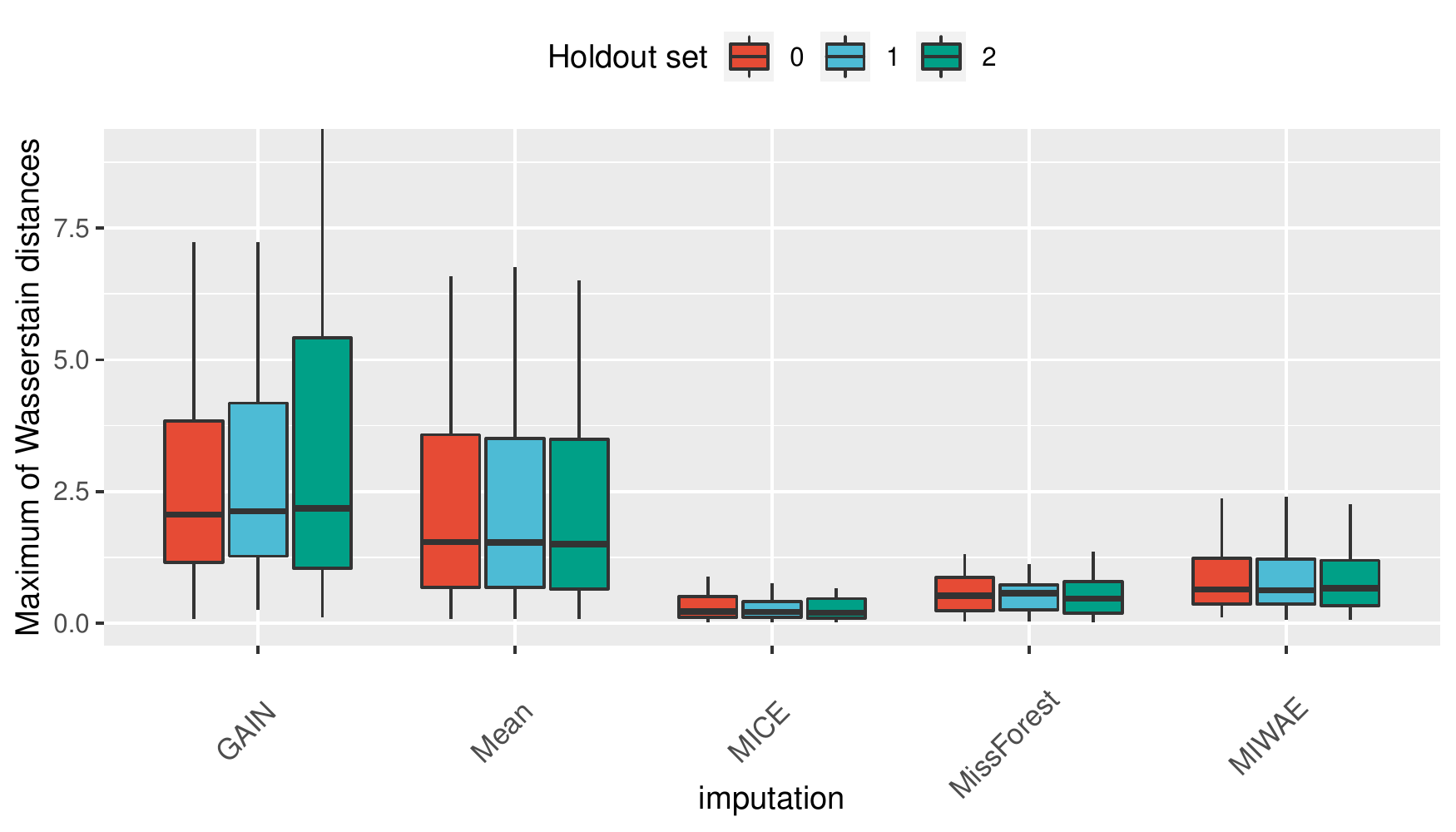}\\
      \hline
      \parbox[c][][c]{0.5in}{\rotatebox[origin=c]{90}{B3 excl. Mean}} &
      \includegraphics[height=4cm,width=5cm]{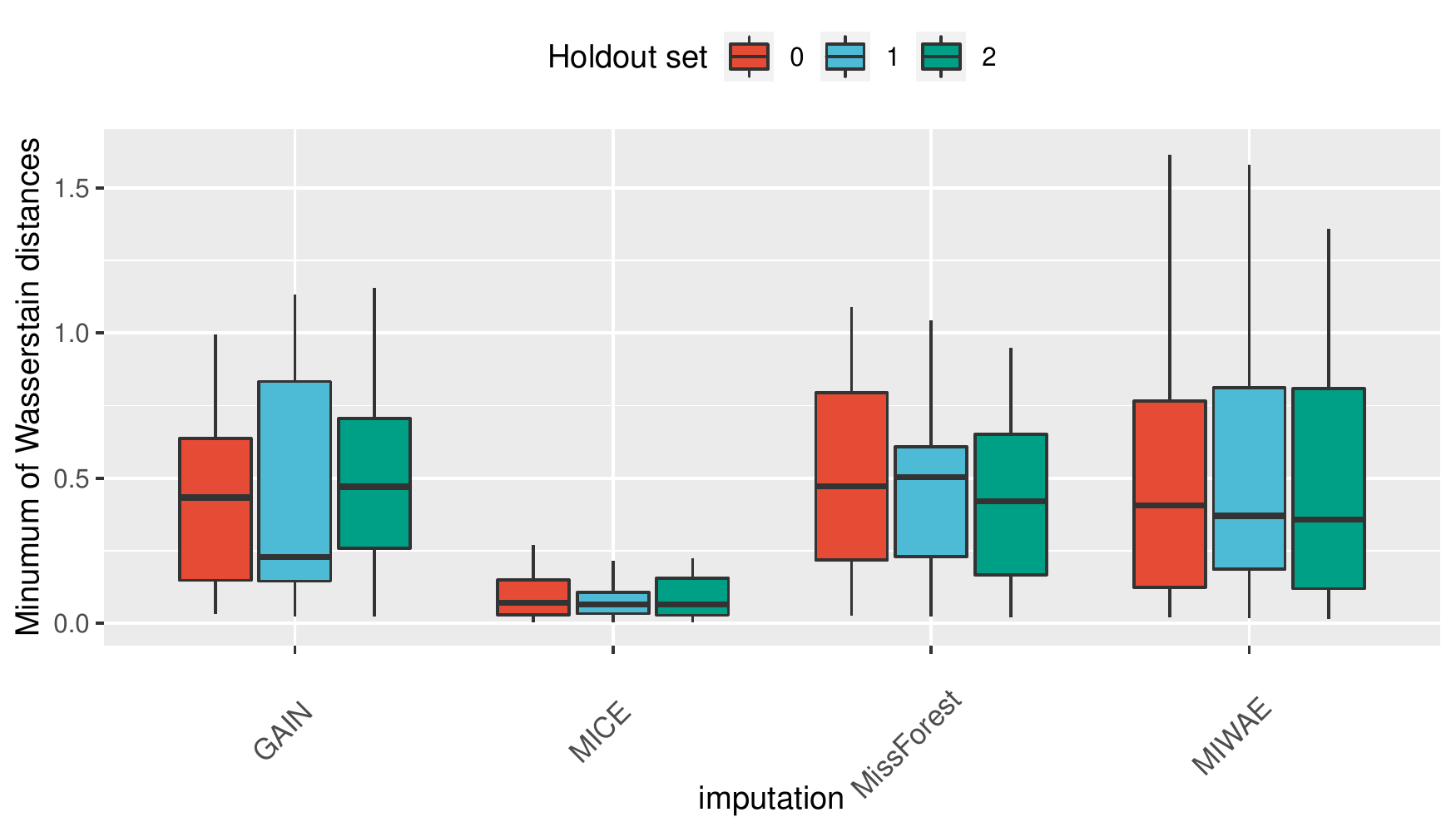}&
      \includegraphics[height=4cm,width=5cm]{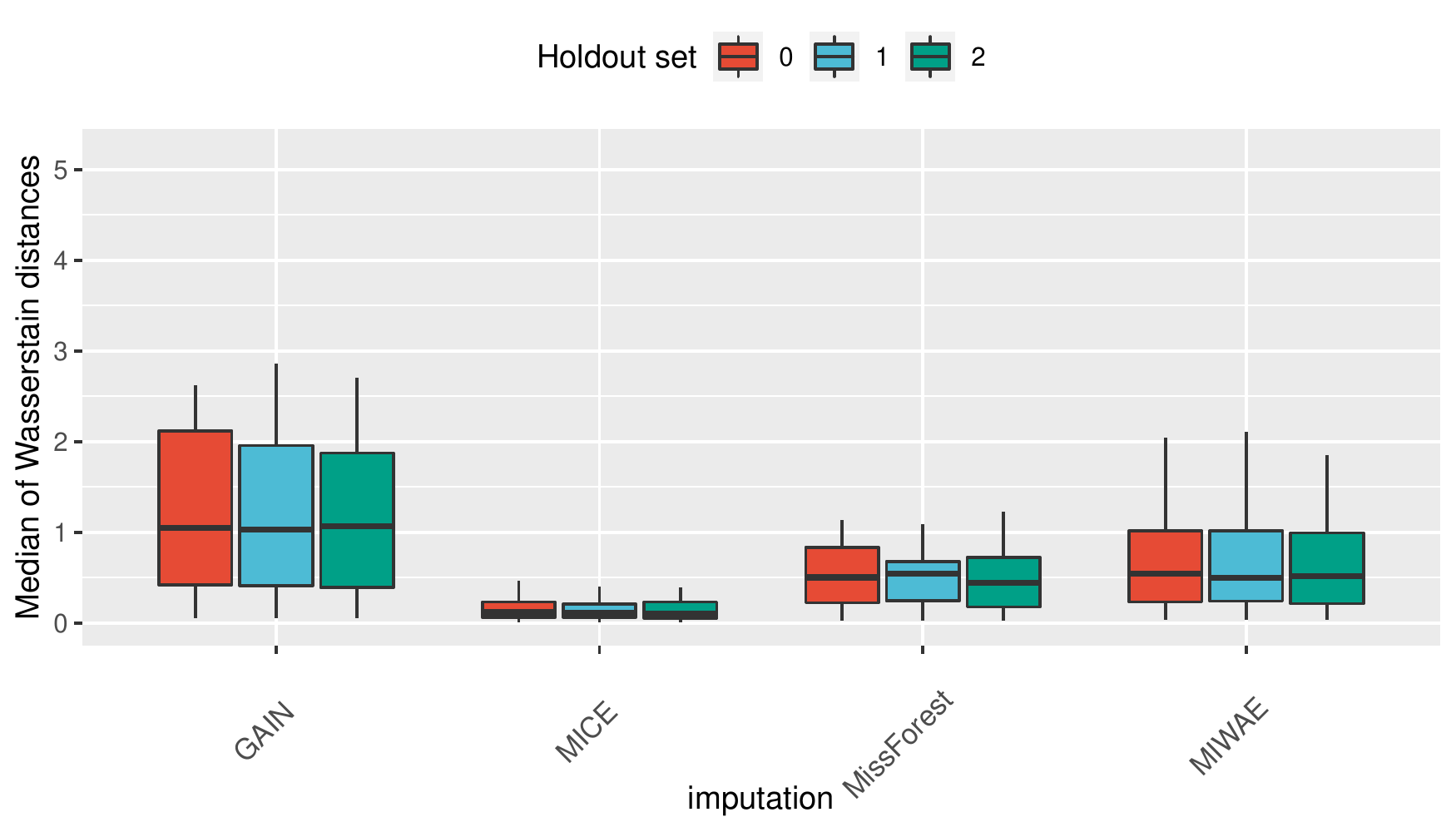}&
      \includegraphics[height=4cm,width=5cm]{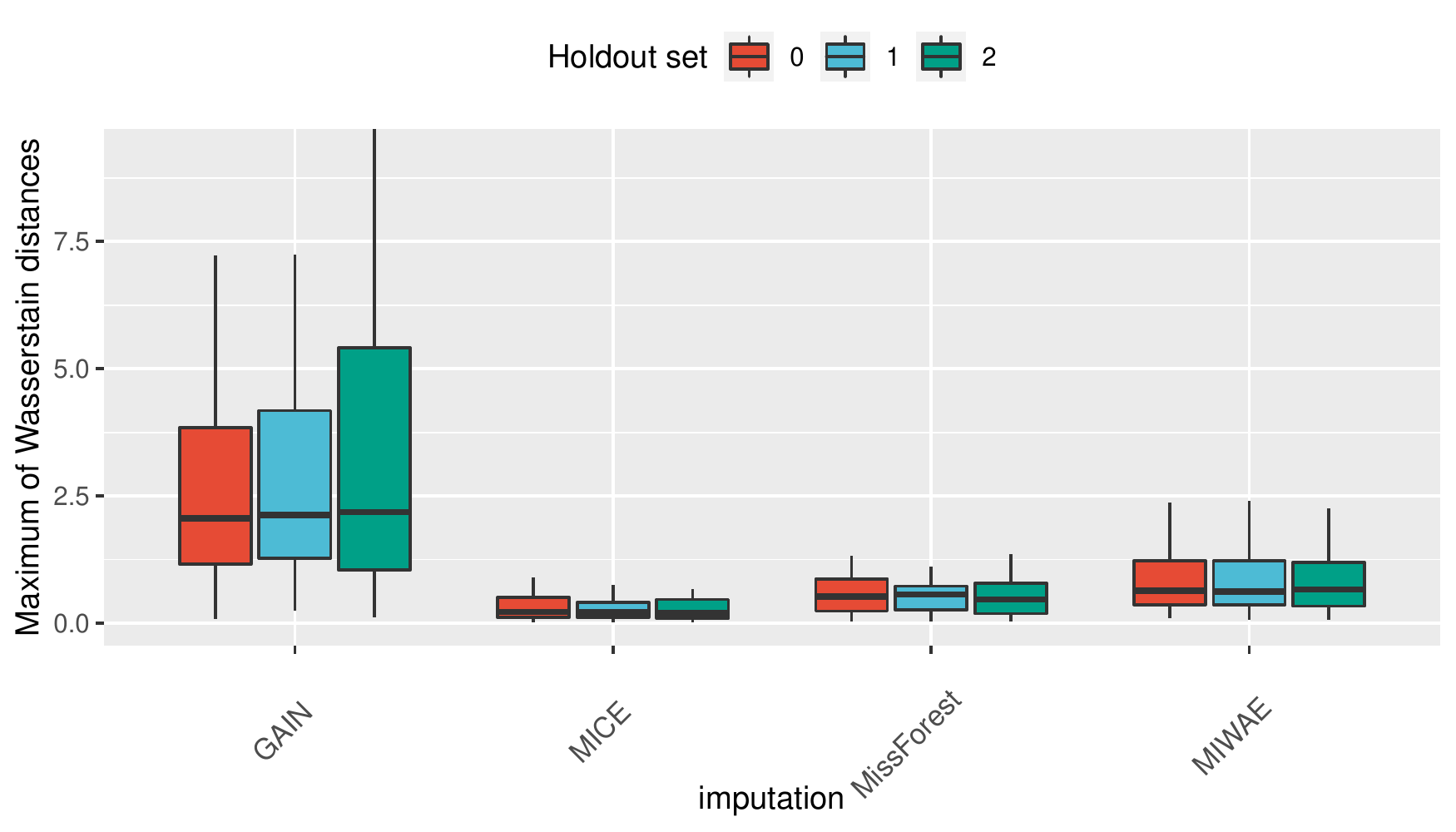}\\
    \end{tabular}
    \caption{Feature-wise 50\% train missingness and 50\% test missingness.}
    \label{fig:featurewise_mimic_50_50}
\end{figure}

\begin{figure}[htb!]
    \centering
    \begin{tabular}{m{0.2in} | M{5cm} | M{5cm} | M{5cm}}
    & \textbf{Minimum} & \textbf{Median} & \textbf{Maximum} \\
    \hline
     \parbox[c][][c]{0.5in}{\rotatebox[origin=t]{90}{B1: Kullback-Leibler}} &
      \includegraphics[height=4cm,width=5cm]{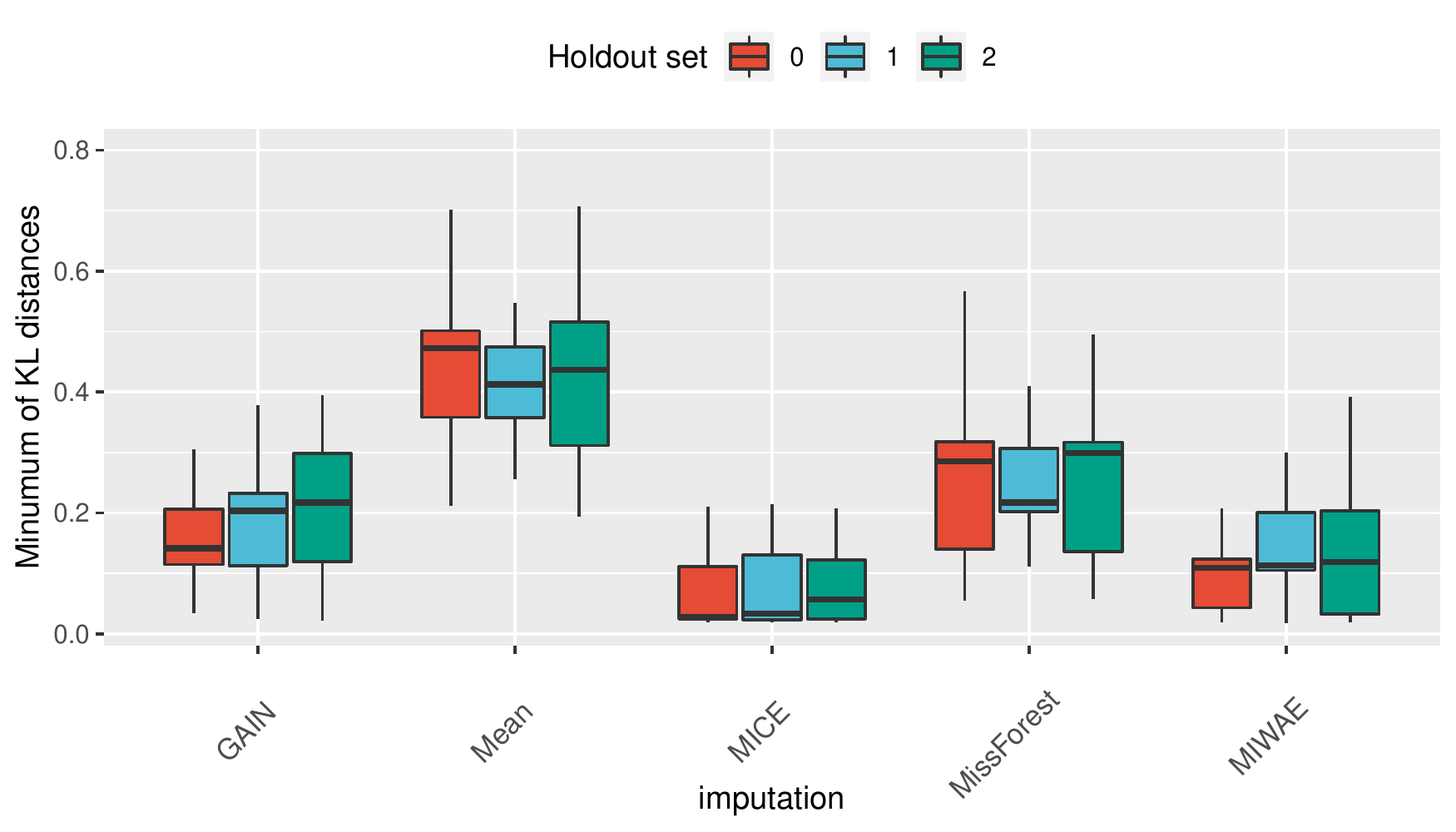}&
      \includegraphics[height=4cm,width=5cm]{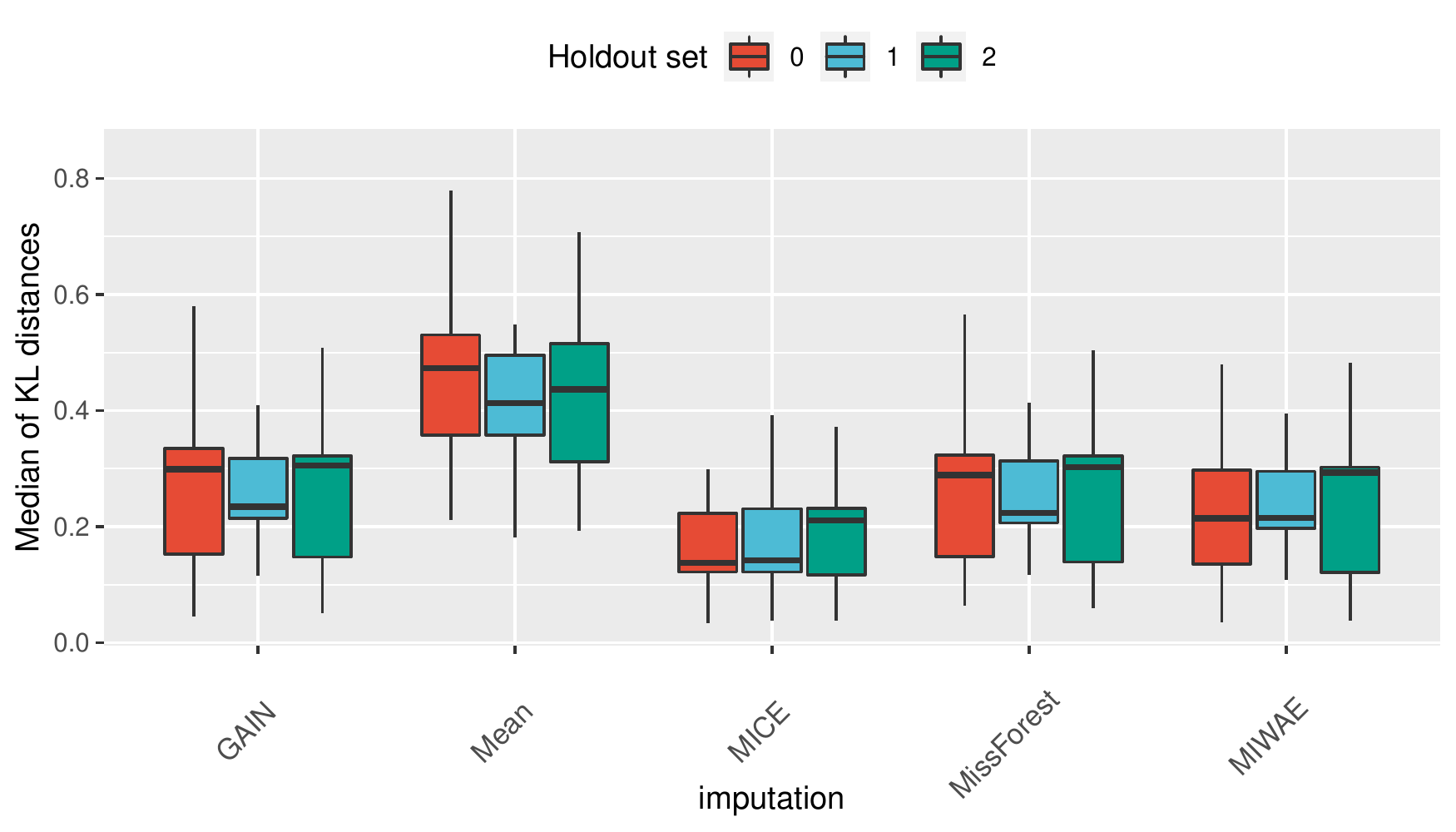}&
      \includegraphics[height=4cm,width=5cm]{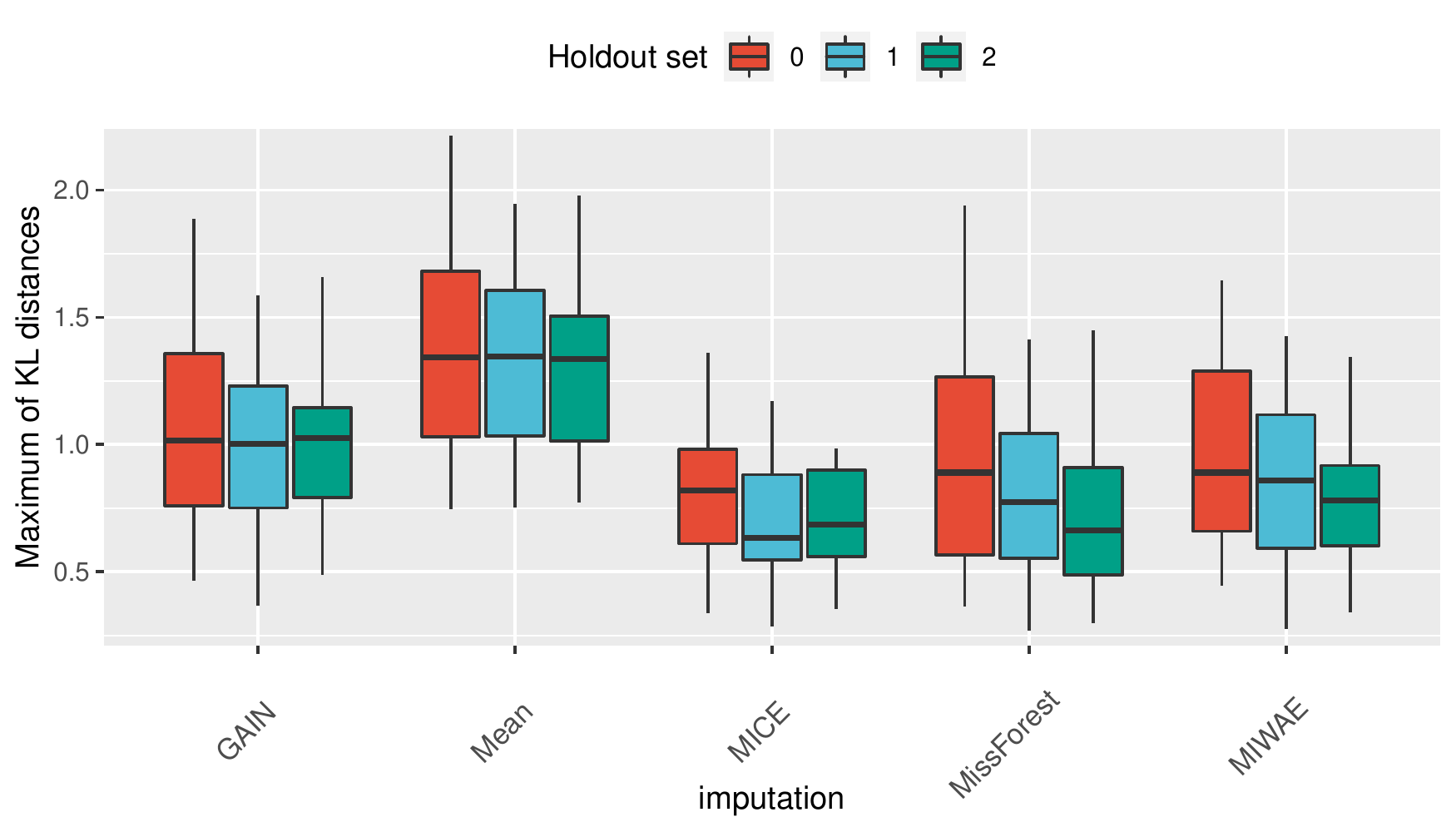}\\
      \hline
      \parbox[c][][c]{0.5in}{\rotatebox[origin=c]{90}{B2: Kolmogorov-Smirnov}} &
      \includegraphics[height=4cm,width=5cm]{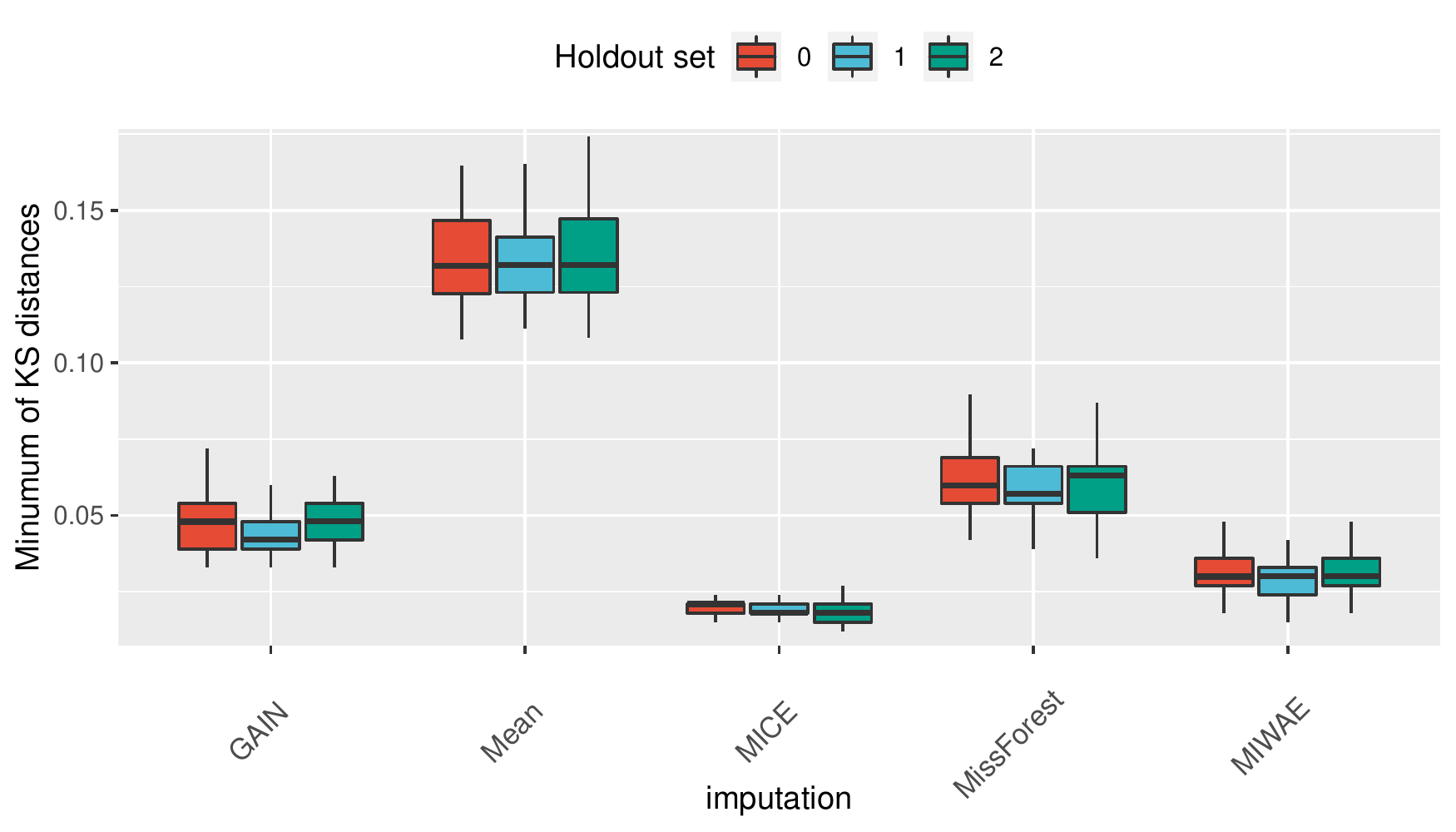}&
      \includegraphics[height=4cm,width=5cm]{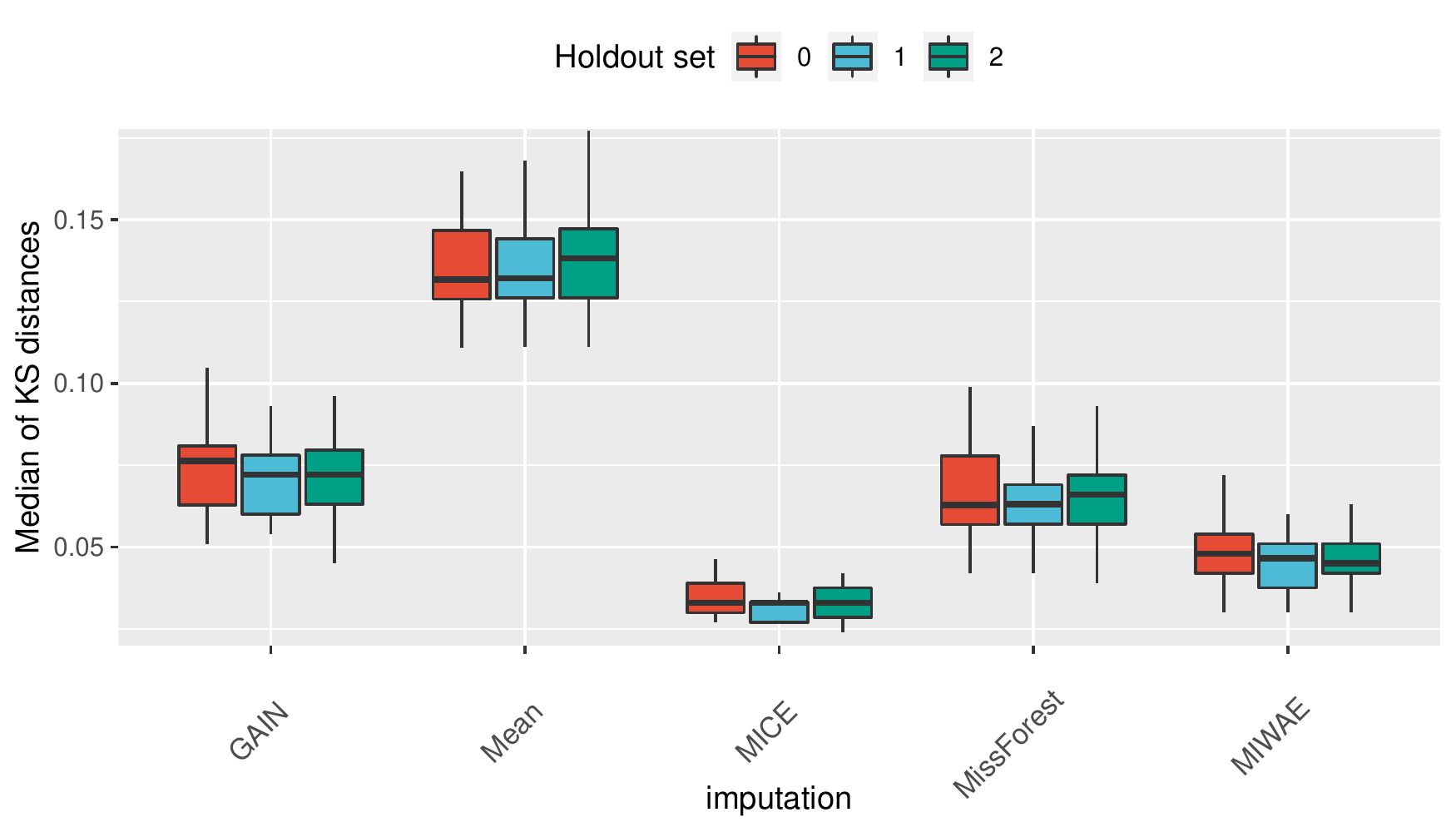}&
      \includegraphics[height=4cm,width=5cm]{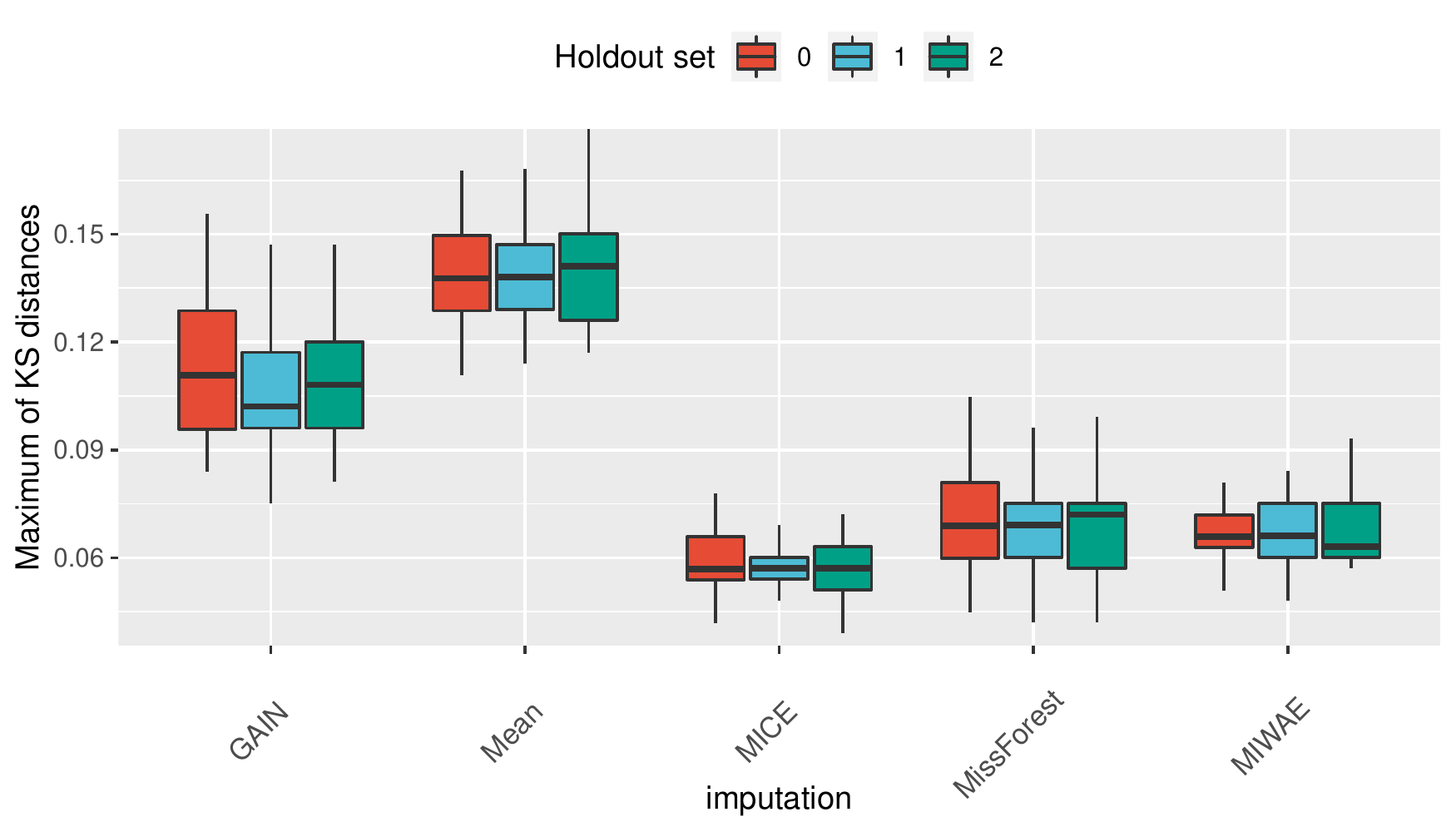}\\
      \hline
      \parbox[c][][c]{0.5in}{\rotatebox[origin=c]{90}{B3: Wasserstein}} &
      \includegraphics[height=4cm,width=5cm]{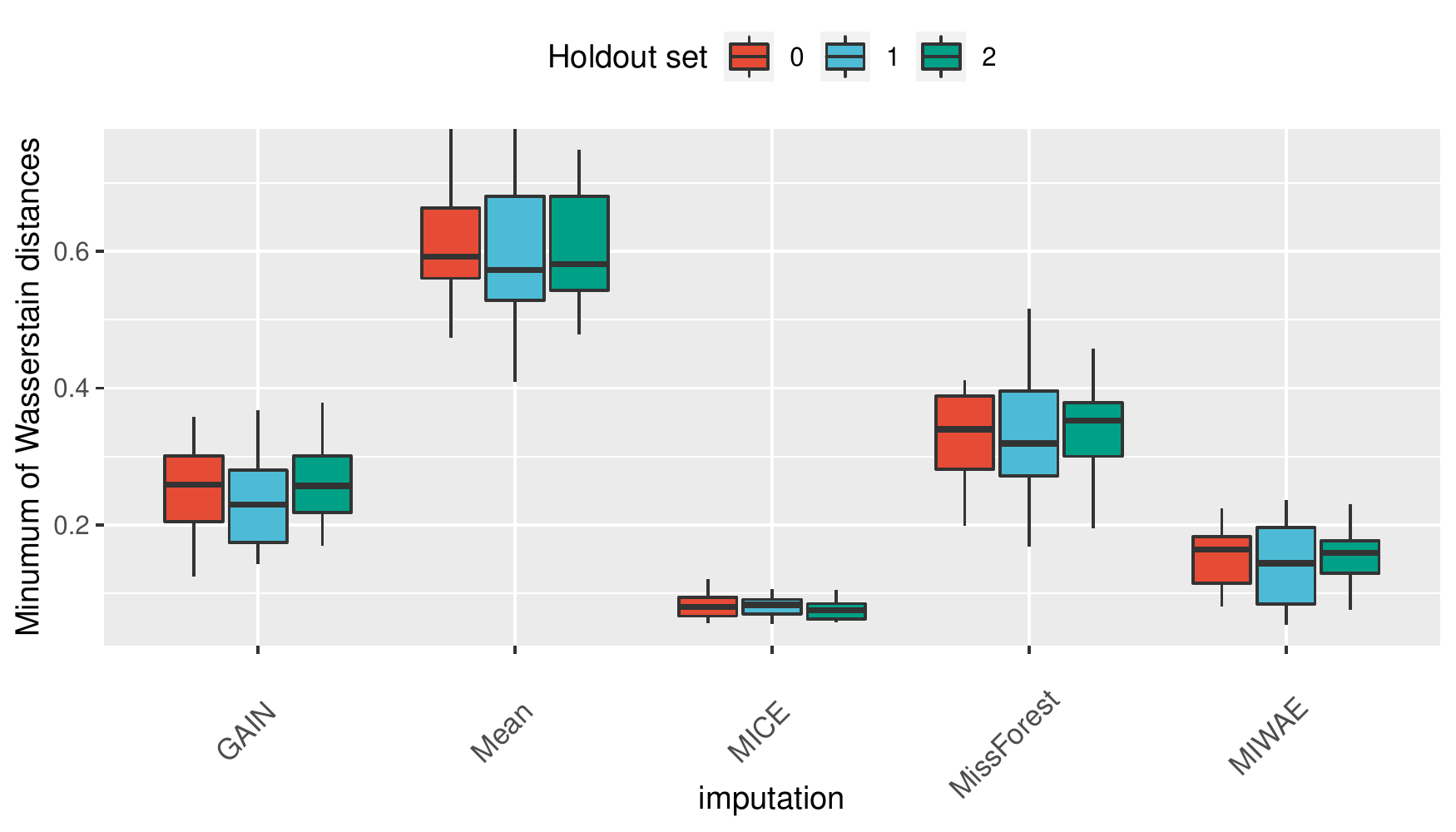}&
      \includegraphics[height=4cm,width=5cm]{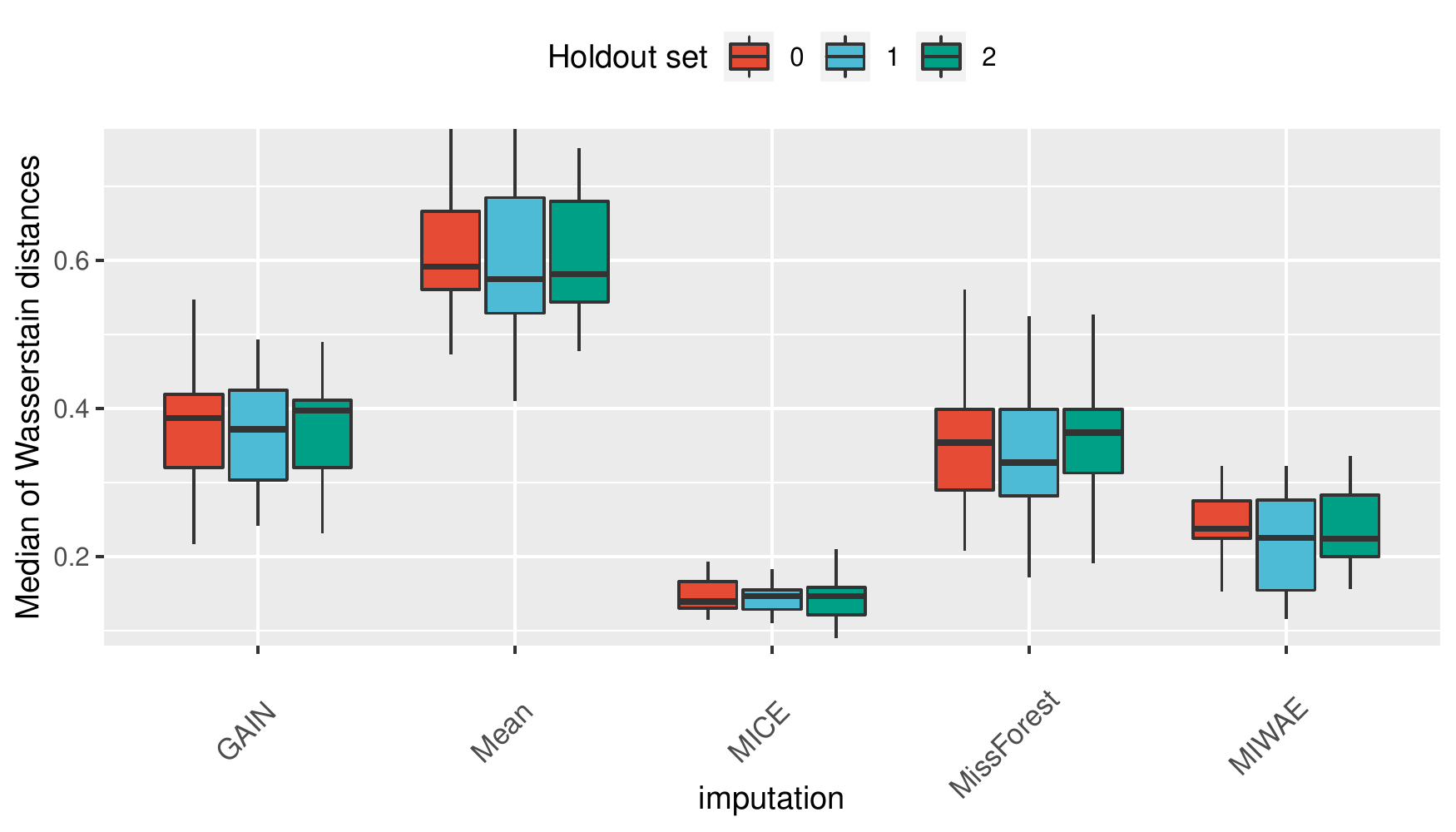}&
      \includegraphics[height=4cm,width=5cm]{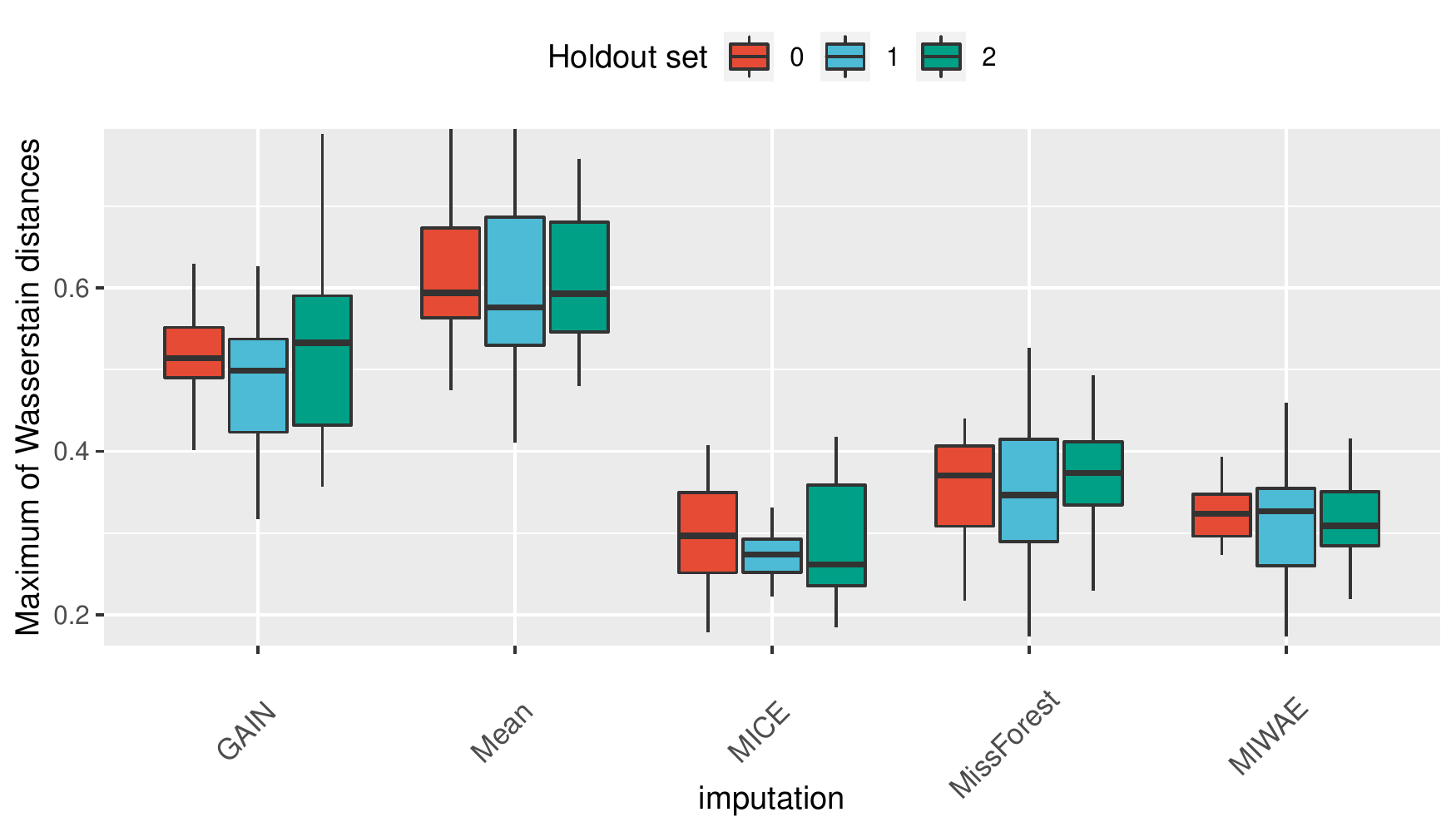}\\
      \hline
      \parbox[c][][c]{0.5in}{\rotatebox[origin=c]{90}{B3 excl. Mean}} &
      \includegraphics[height=4cm,width=5cm]{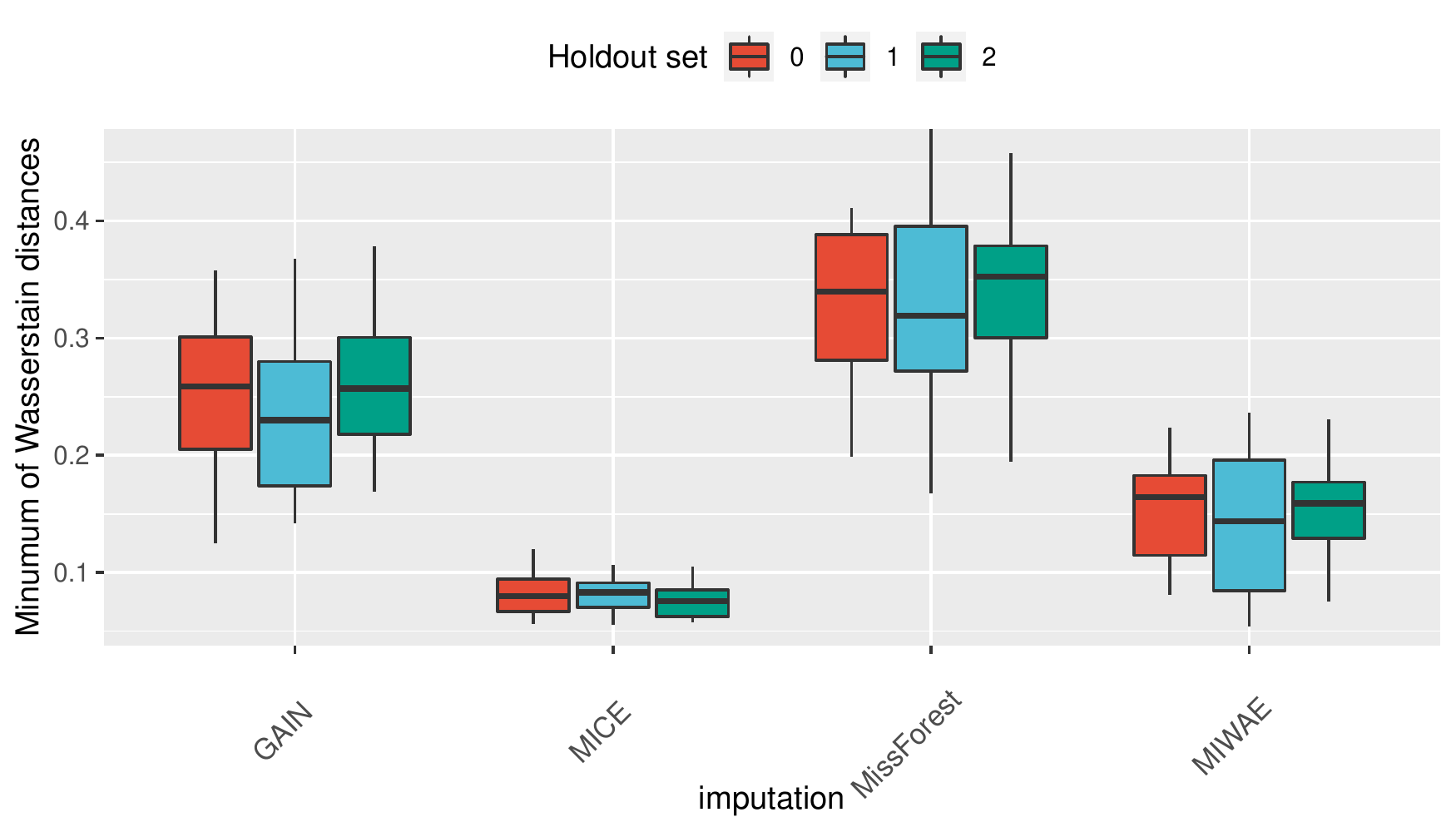}&
      \includegraphics[height=4cm,width=5cm]{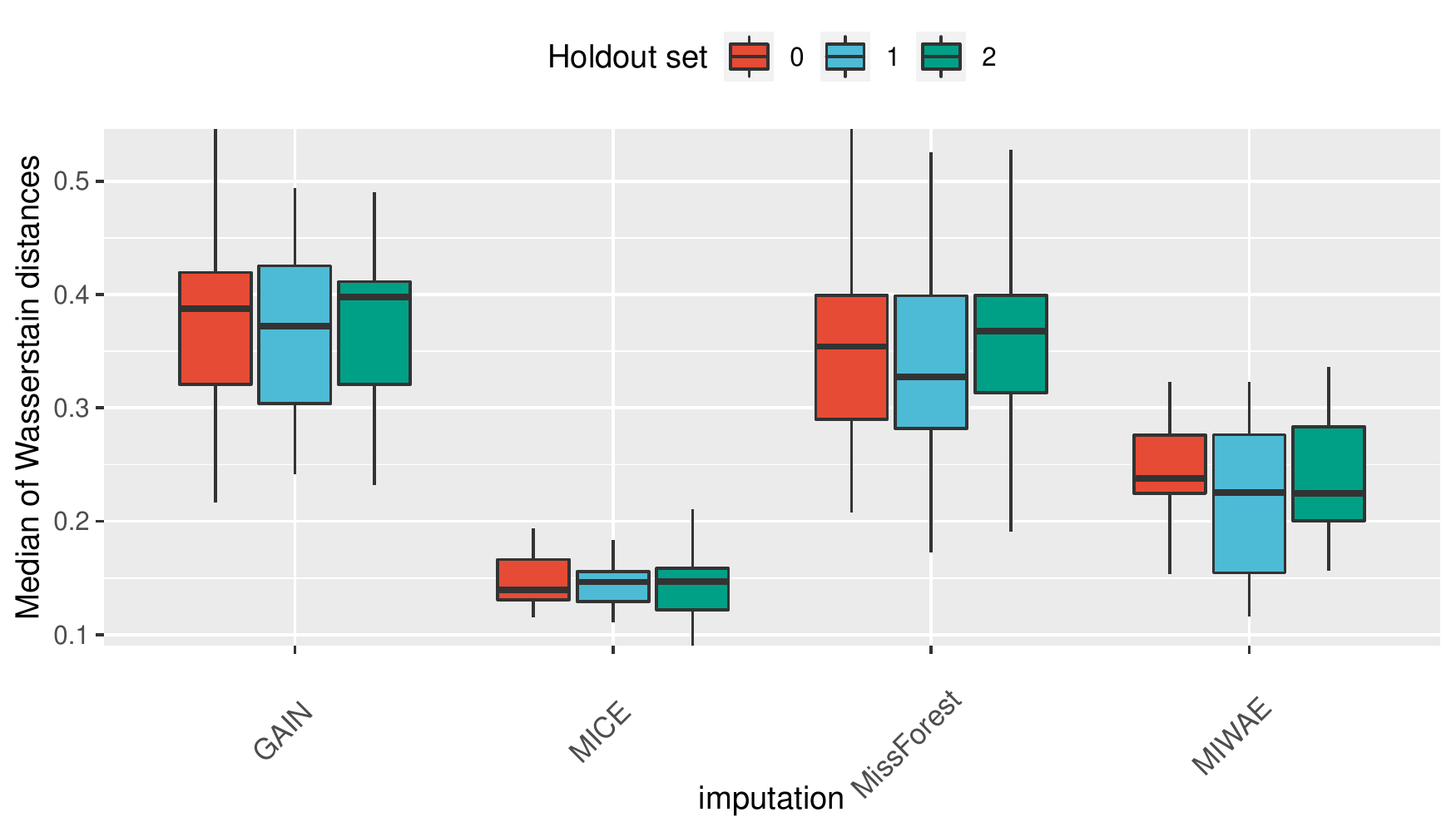}&
      \includegraphics[height=4cm,width=5cm]{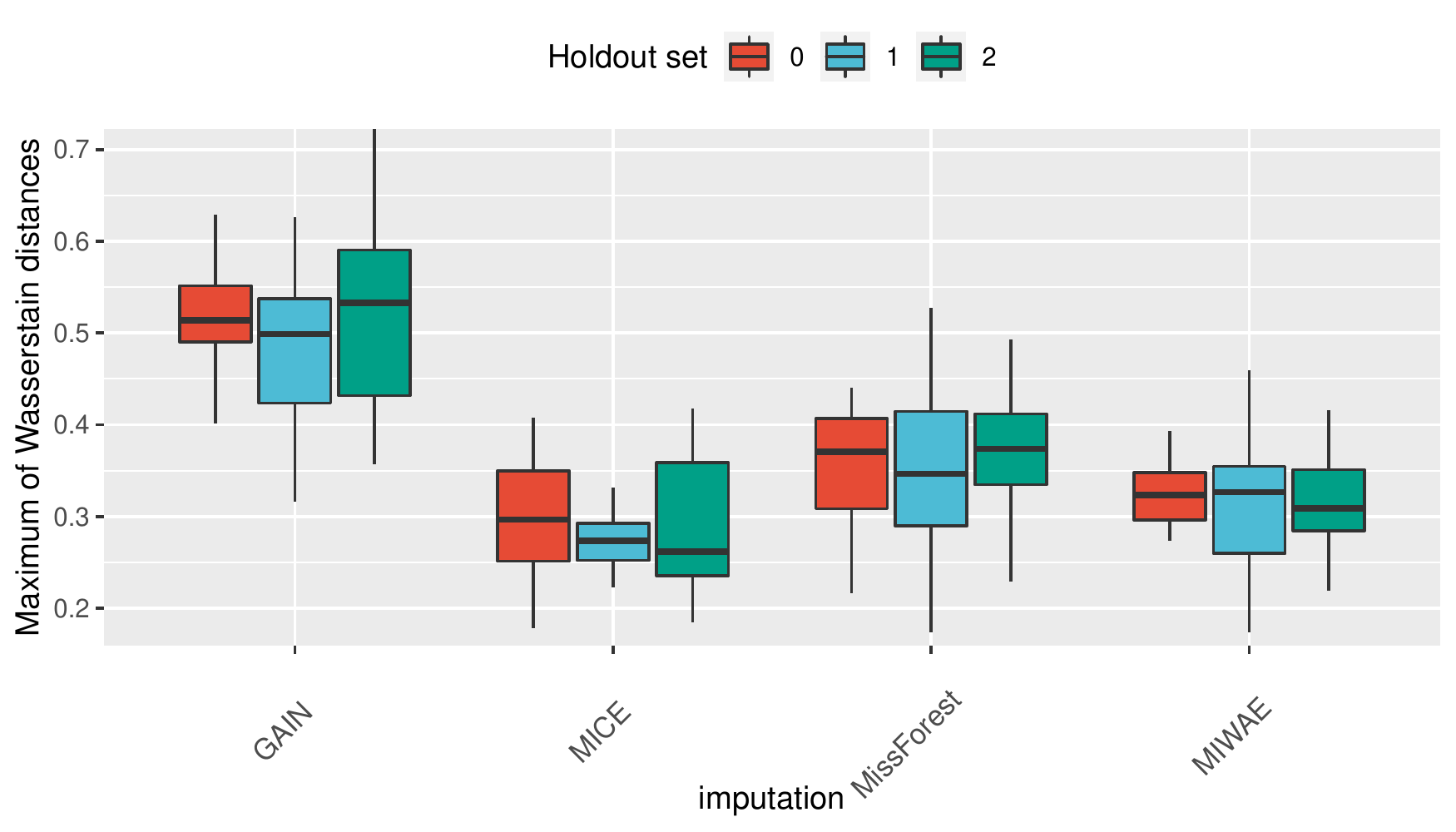}\\
    \end{tabular}
    \caption{Feature-wise 25\% train missingness and 25\% test missingness.}
    \label{fig:featurewise_syn_25_25}
\end{figure}

\begin{figure}[htb!]
    \centering
    \begin{tabular}{m{0.2in} | M{5cm} | M{5cm} | M{5cm}}
    & \textbf{Minimum} & \textbf{Median} & \textbf{Maximum} \\
    \hline
     \parbox[c][][c]{0.5in}{\rotatebox[origin=t]{90}{B1: Kullback-Leibler}} &
      \includegraphics[height=4cm,width=5cm]{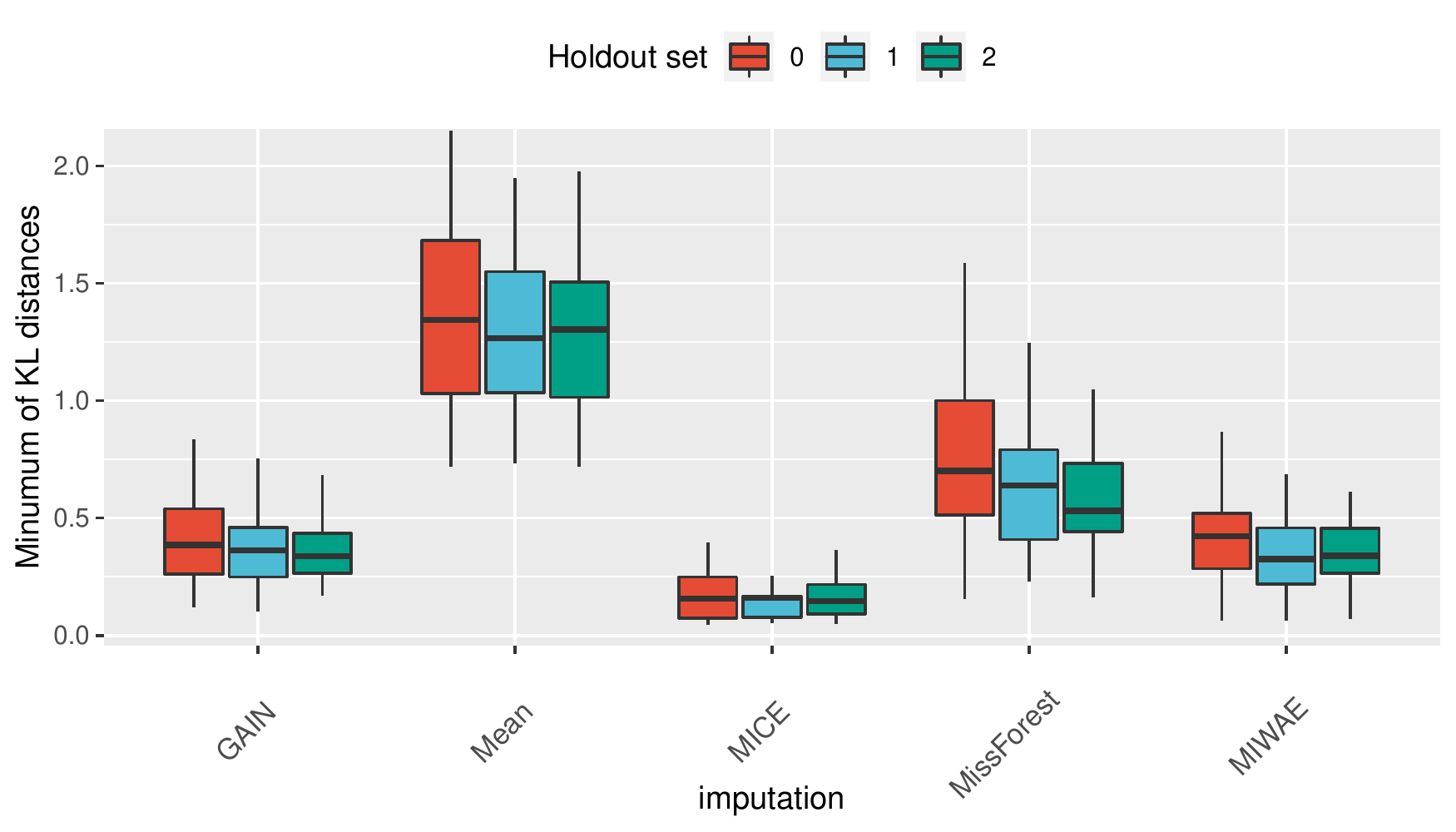}&
      \includegraphics[height=4cm,width=5cm]{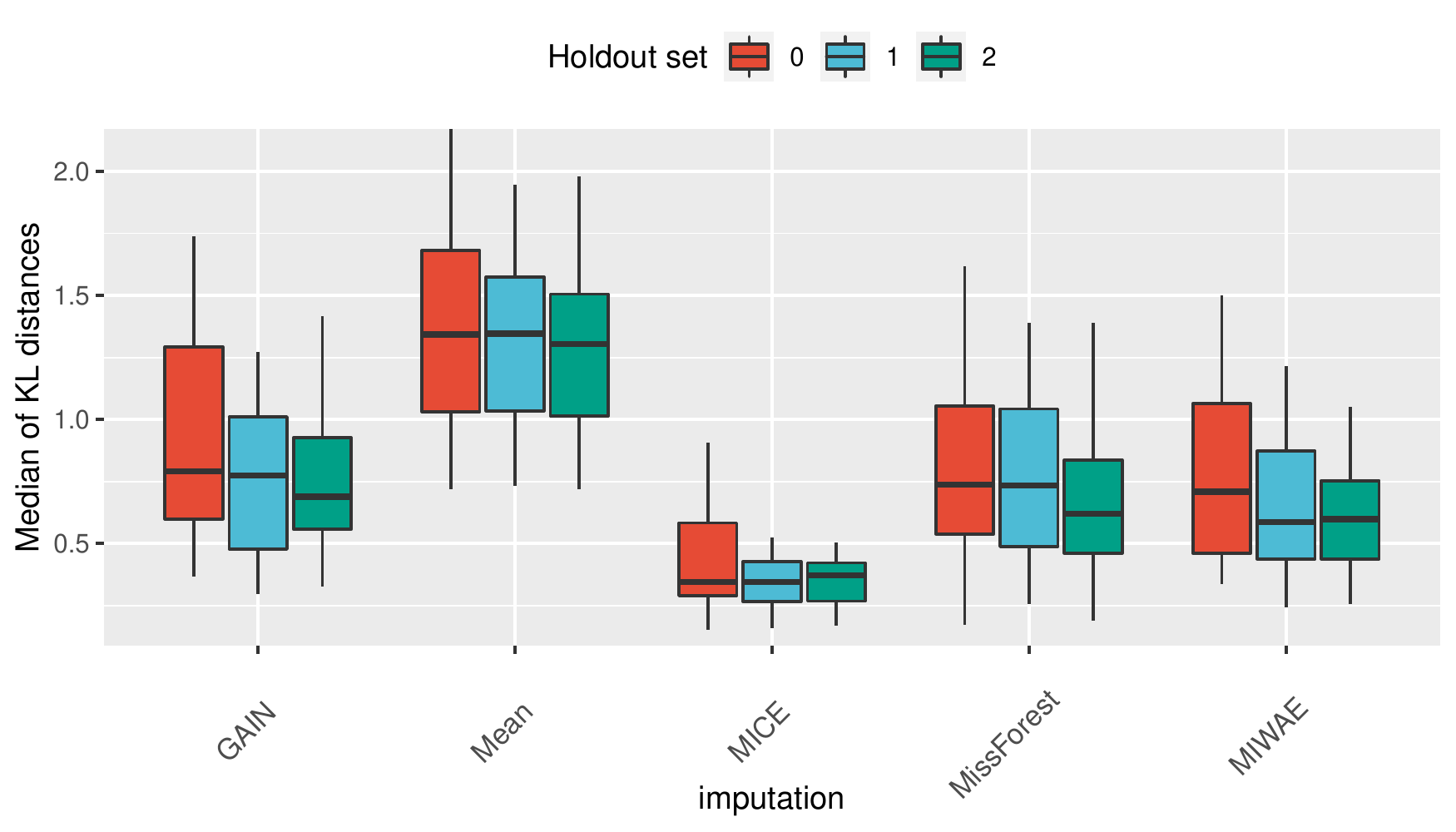}&
      \includegraphics[height=4cm,width=5cm]{Supp_synthetic_plots/synthetic_boxplot/feature-wise/SYN_0.25_0.5_kl_max.pdf}\\
      \hline
      \parbox[c][][c]{0.5in}{\rotatebox[origin=c]{90}{B2: Kolmogorov-Smirnov}} &
      \includegraphics[height=4cm,width=5cm]{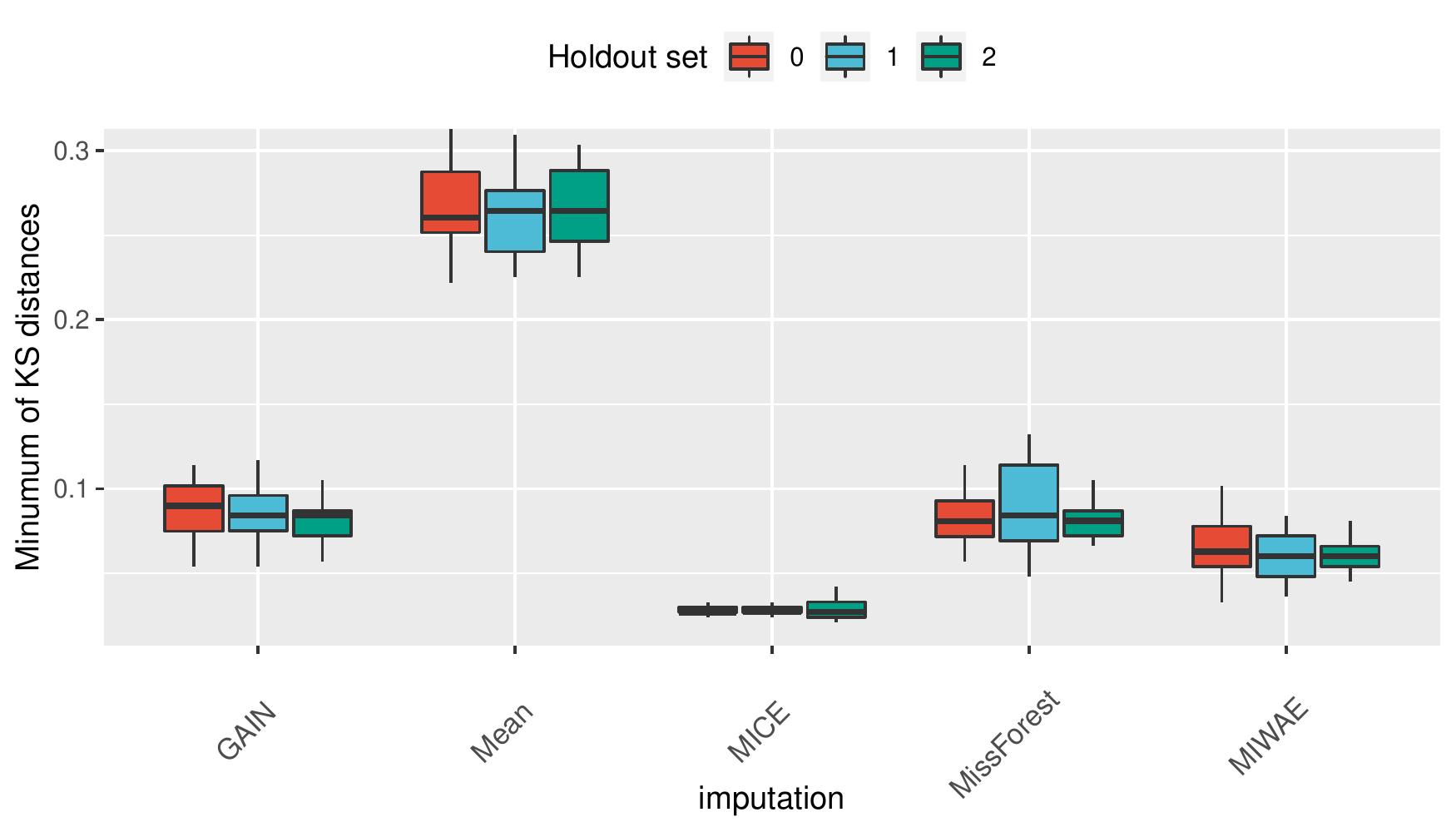}&
      \includegraphics[height=4cm,width=5cm]{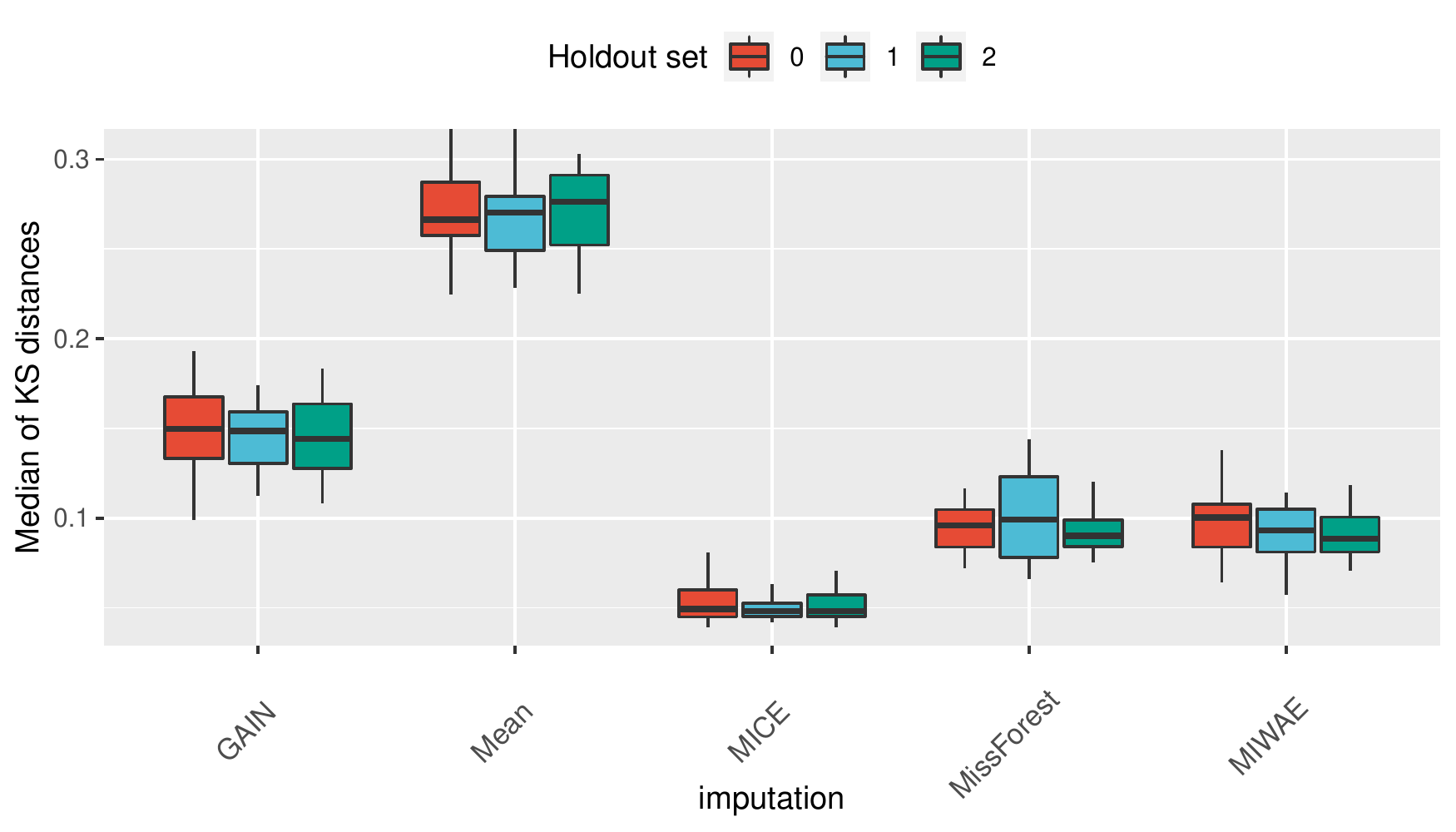}&
      \includegraphics[height=4cm,width=5cm]{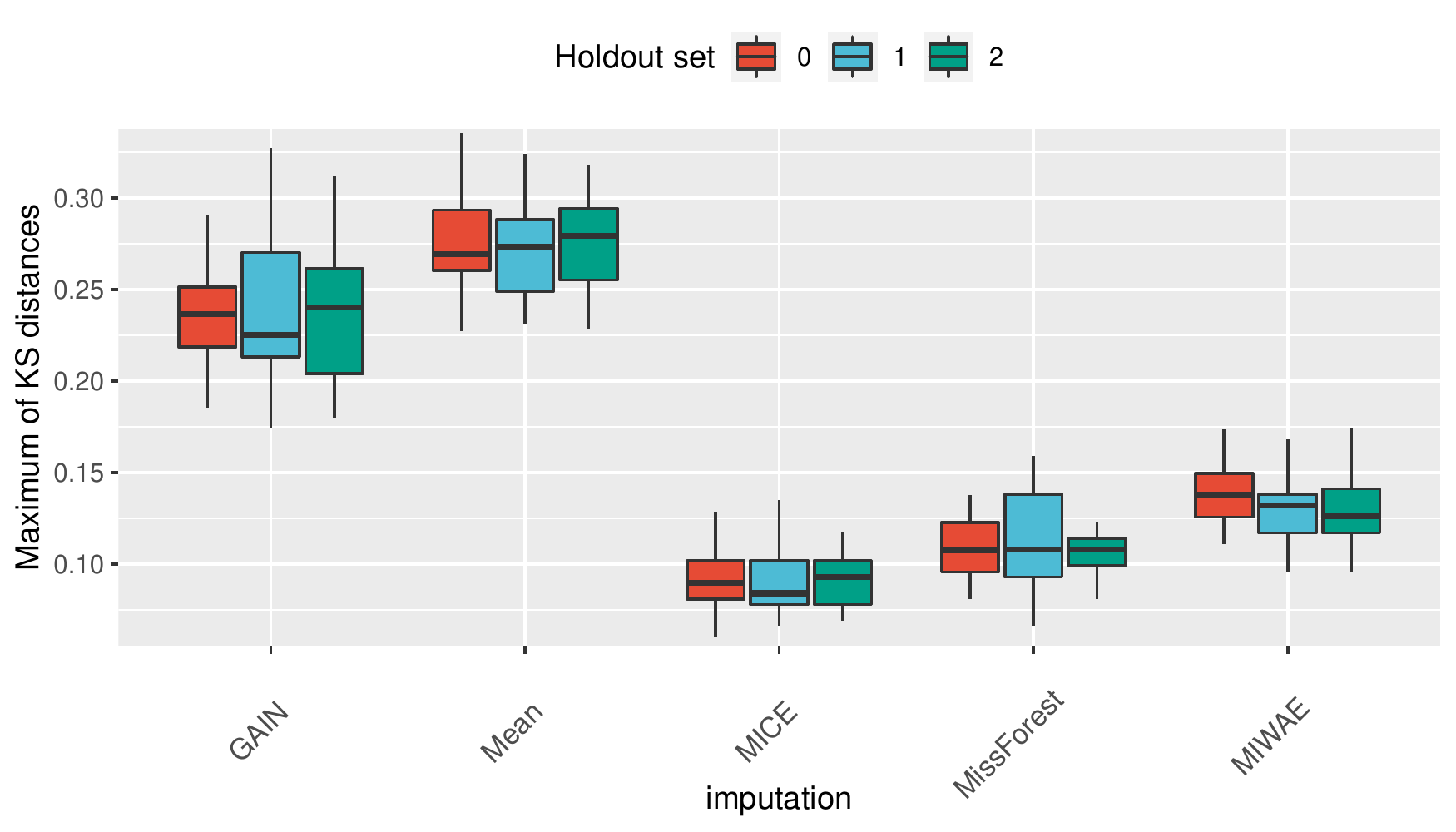}\\
      \hline
      \parbox[c][][c]{0.5in}{\rotatebox[origin=c]{90}{B3: Wasserstein}} &
      \includegraphics[height=4cm,width=5cm]{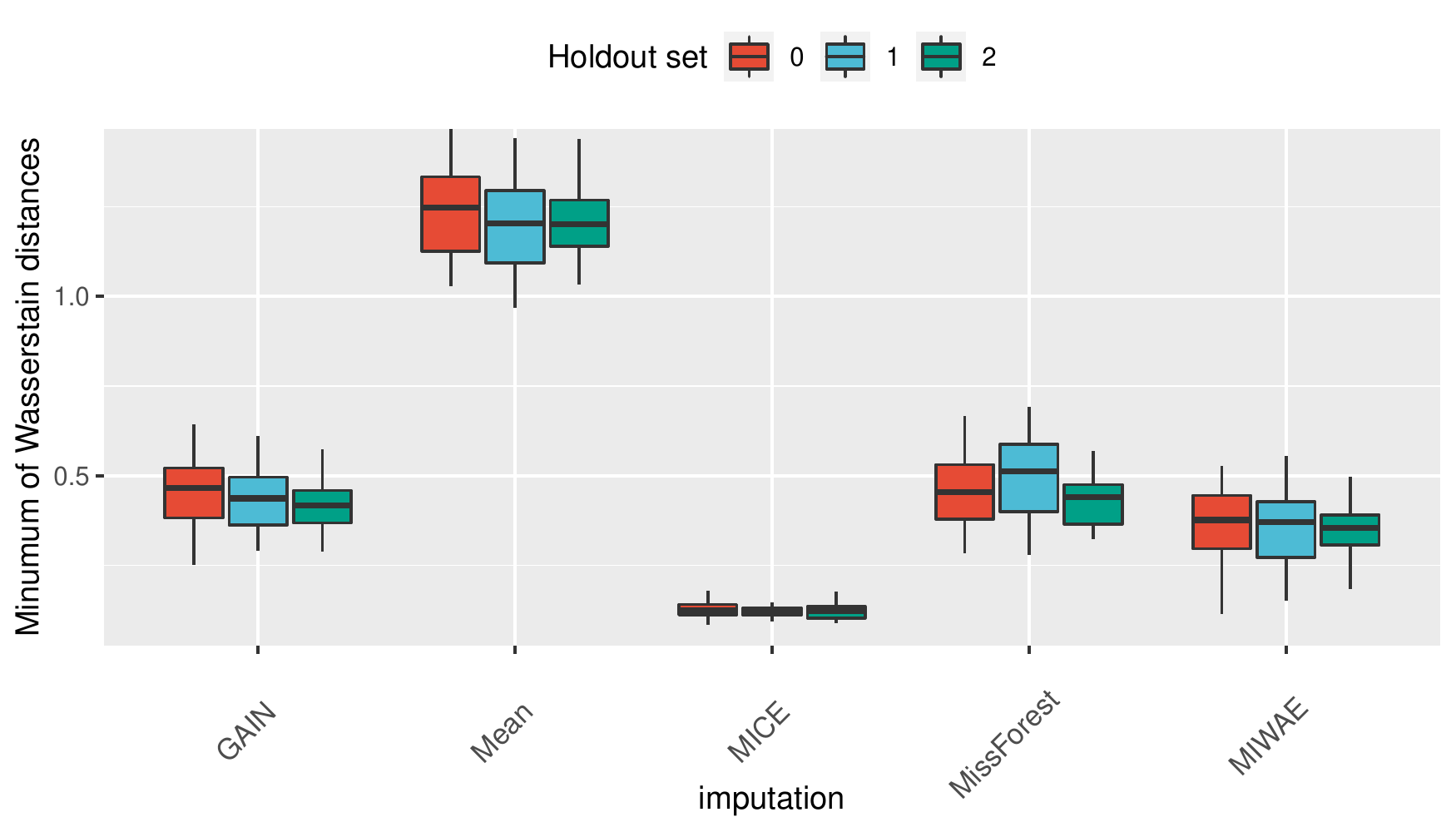}&
      \includegraphics[height=4cm,width=5cm]{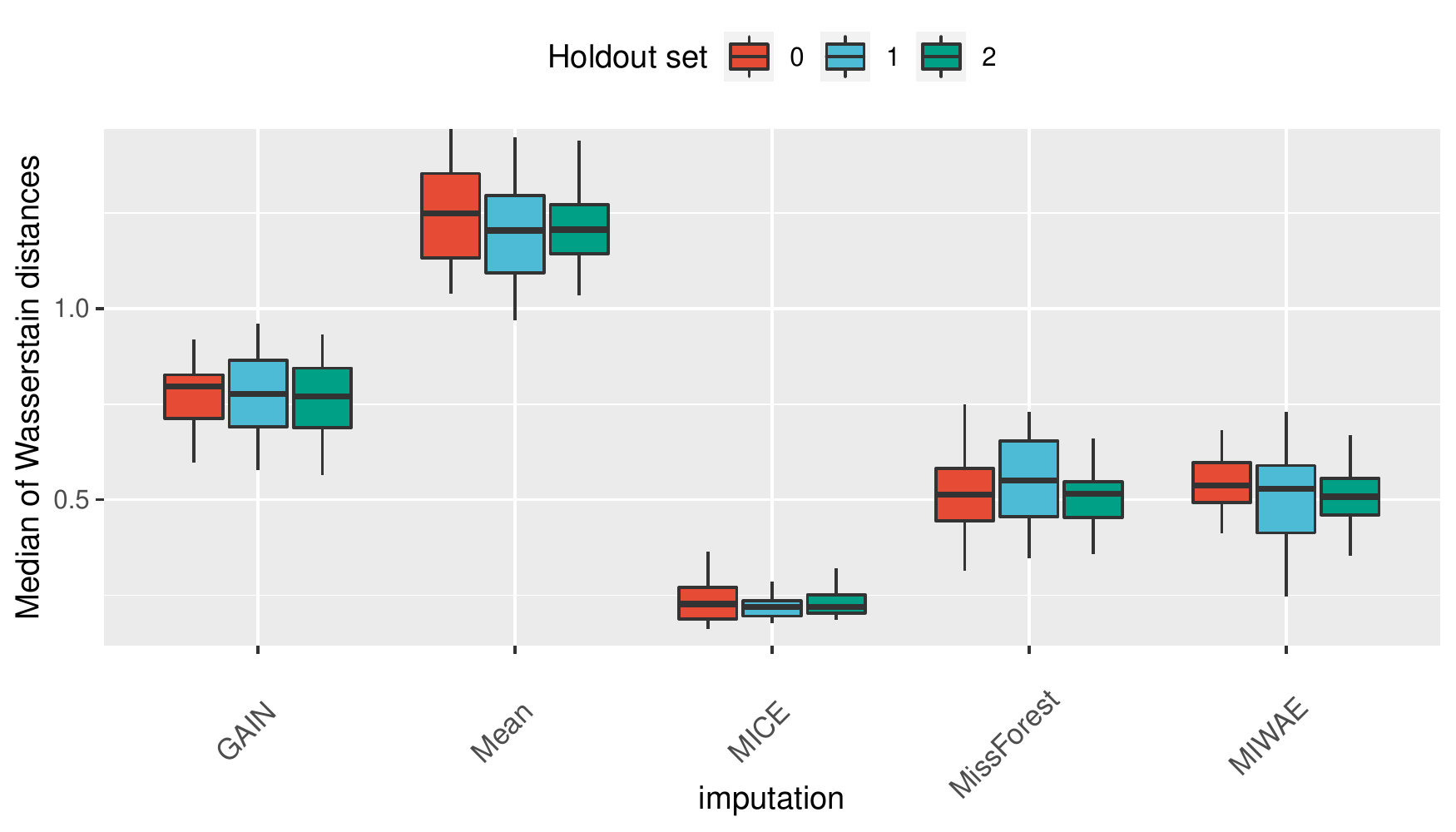}&
      \includegraphics[height=4cm,width=5cm]{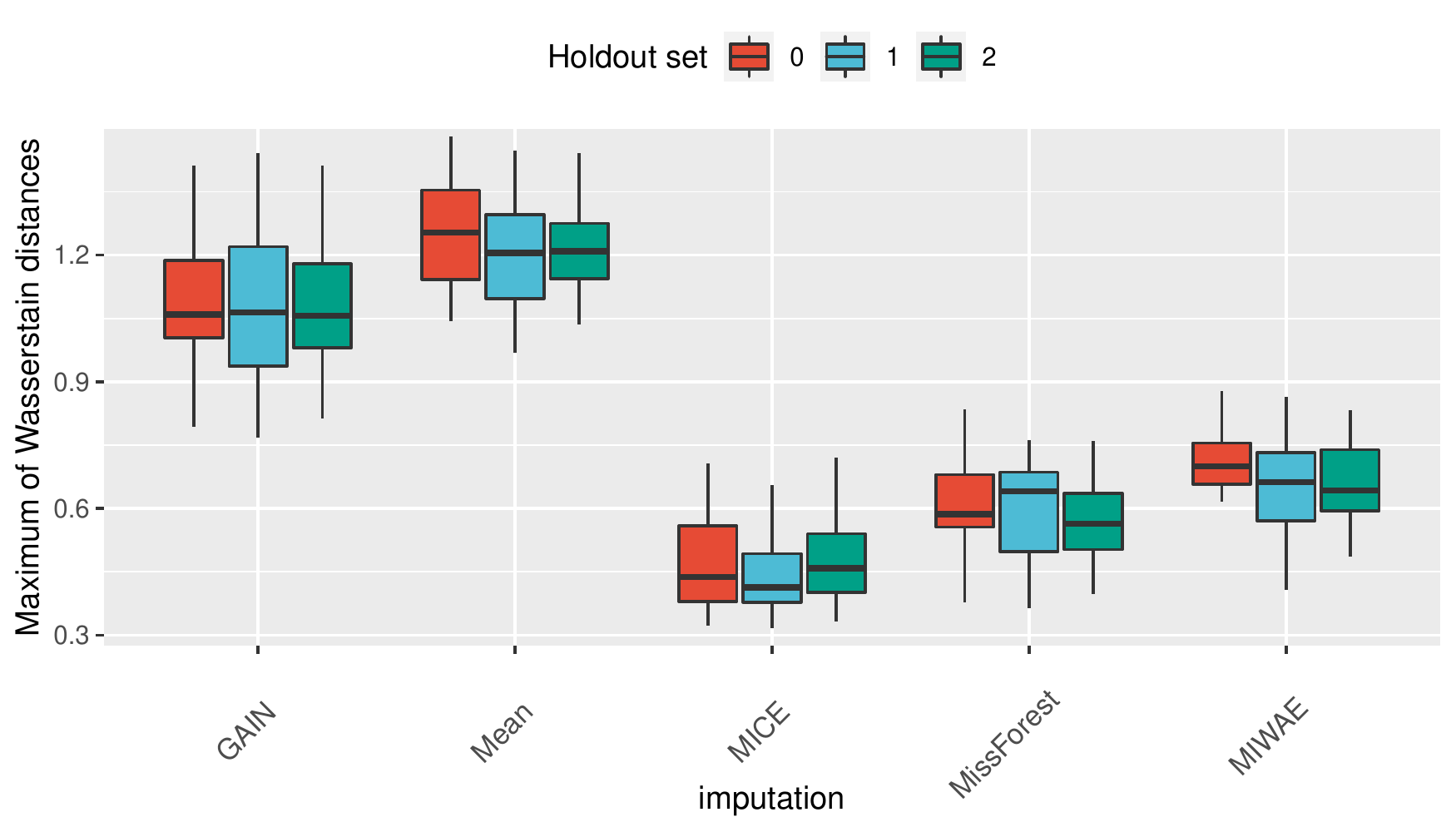}\\
      \hline
      \parbox[c][][c]{0.5in}{\rotatebox[origin=c]{90}{B3 excl. Mean}} &
      \includegraphics[height=4cm,width=5cm]{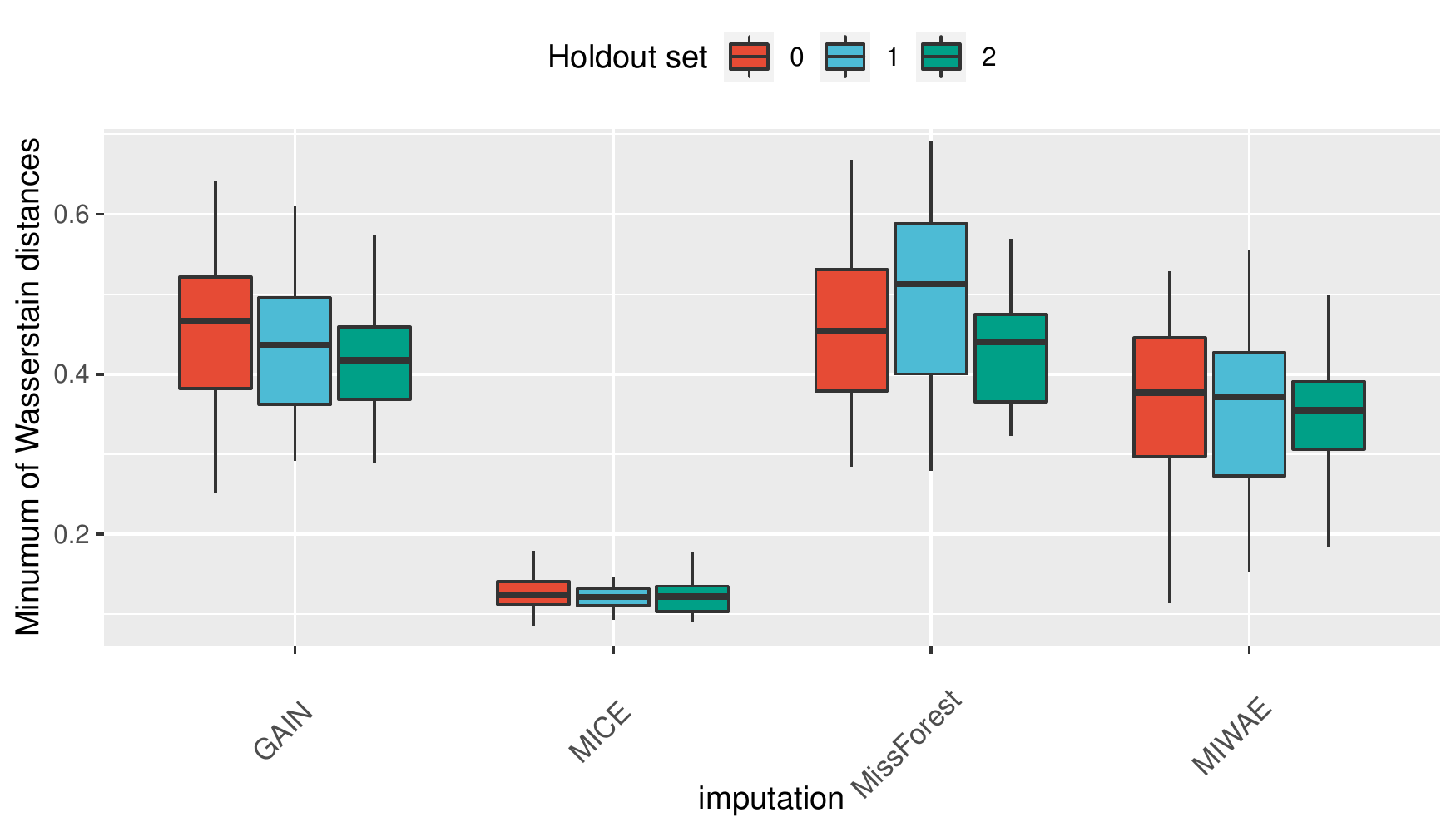}&
      \includegraphics[height=4cm,width=5cm]{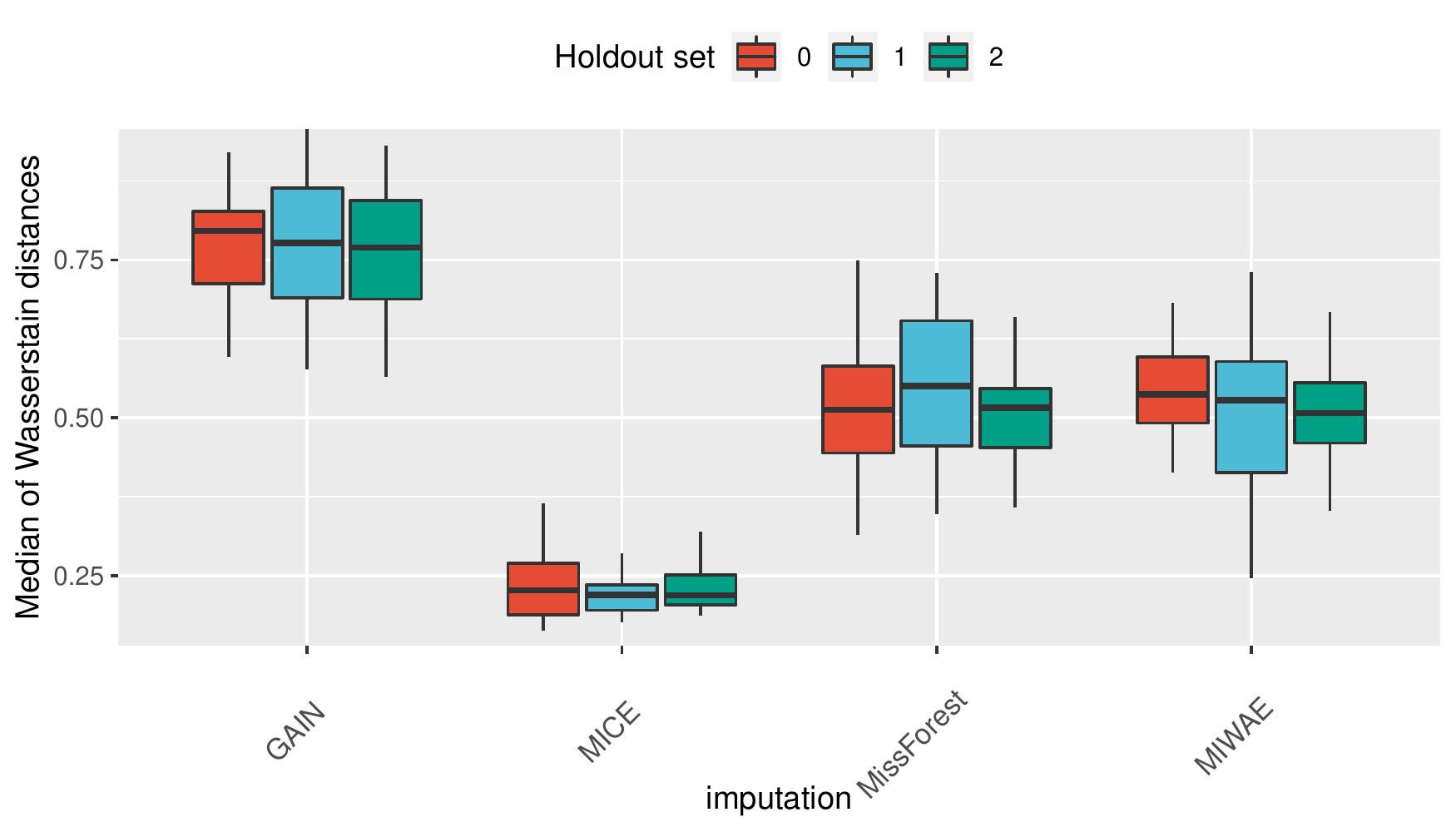}&
      \includegraphics[height=4cm,width=5cm]{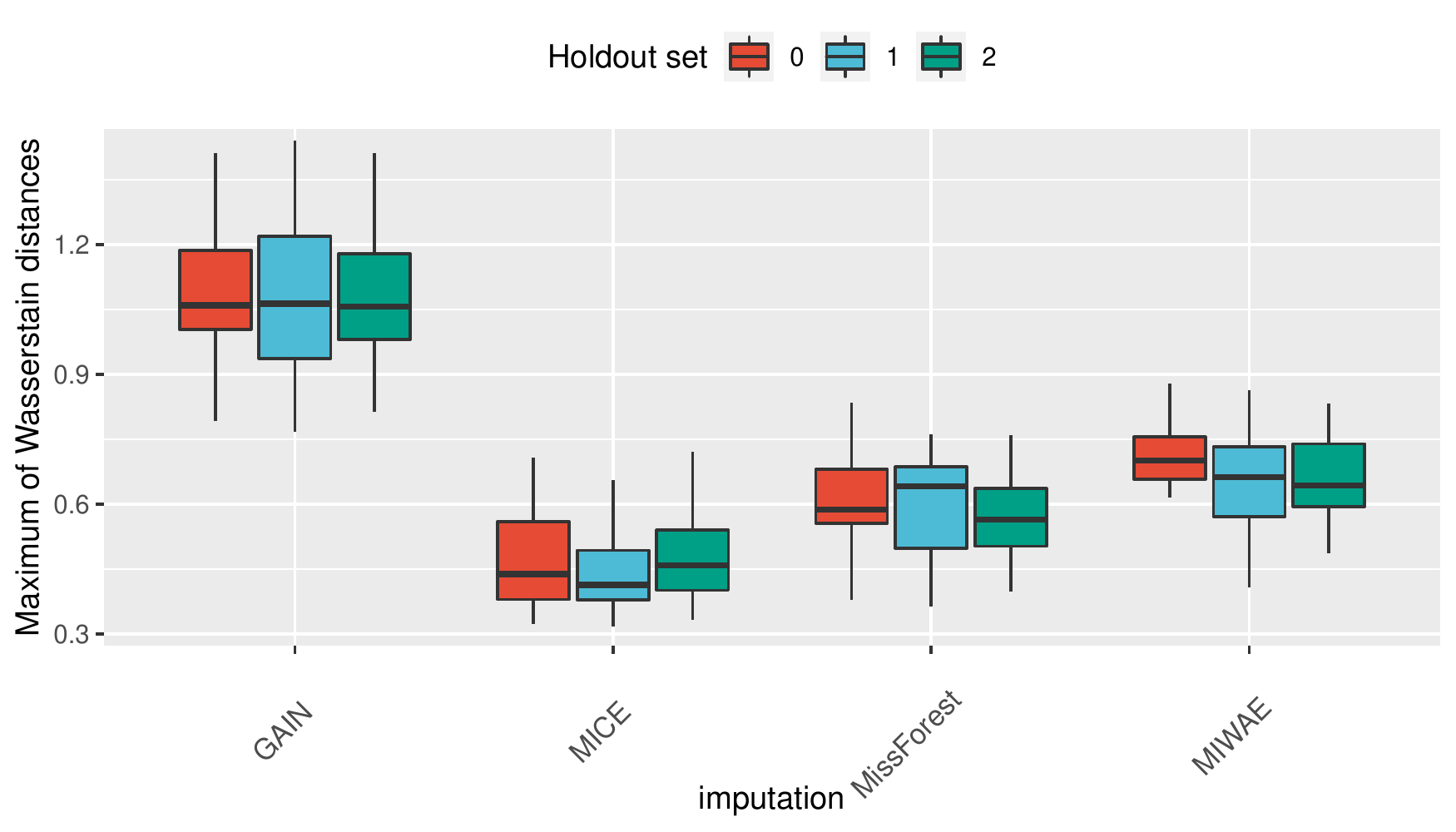}\\
    \end{tabular}
    \caption{Feature-wise 25\% train missingness and 50\% test missingness.}
    \label{fig:featurewise_syn_25_50}
\end{figure}

\begin{figure}[htb!]
    \centering
    \begin{tabular}{m{0.2in} | M{5cm} | M{5cm} | M{5cm}}
    & \textbf{Minimum} & \textbf{Median} & \textbf{Maximum} \\
    \hline
     \parbox[c][][c]{0.5in}{\rotatebox[origin=t]{90}{B1: Kullback-Leibler}} &
      \includegraphics[height=4cm,width=5cm]{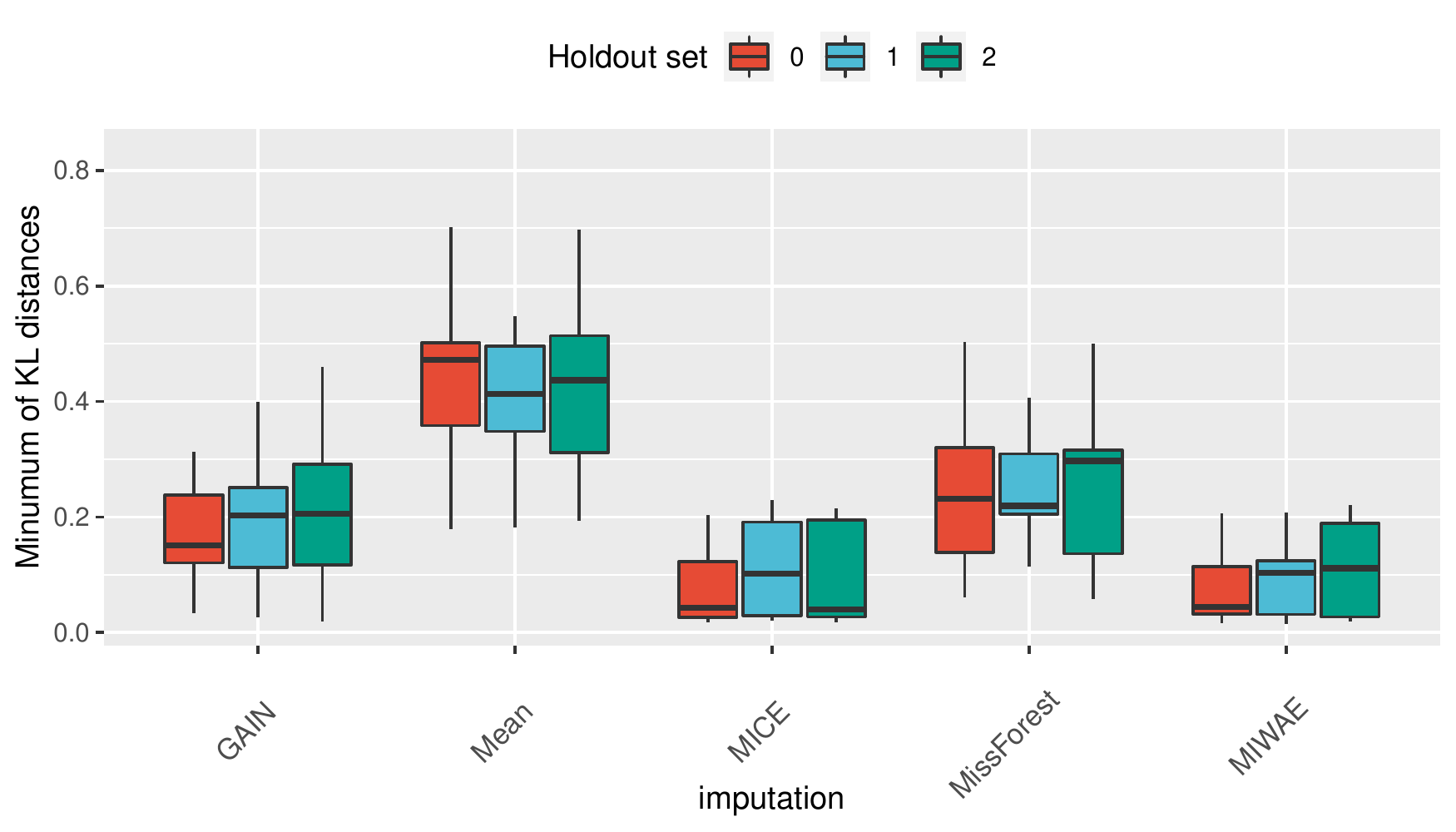}&
      \includegraphics[height=4cm,width=5cm]{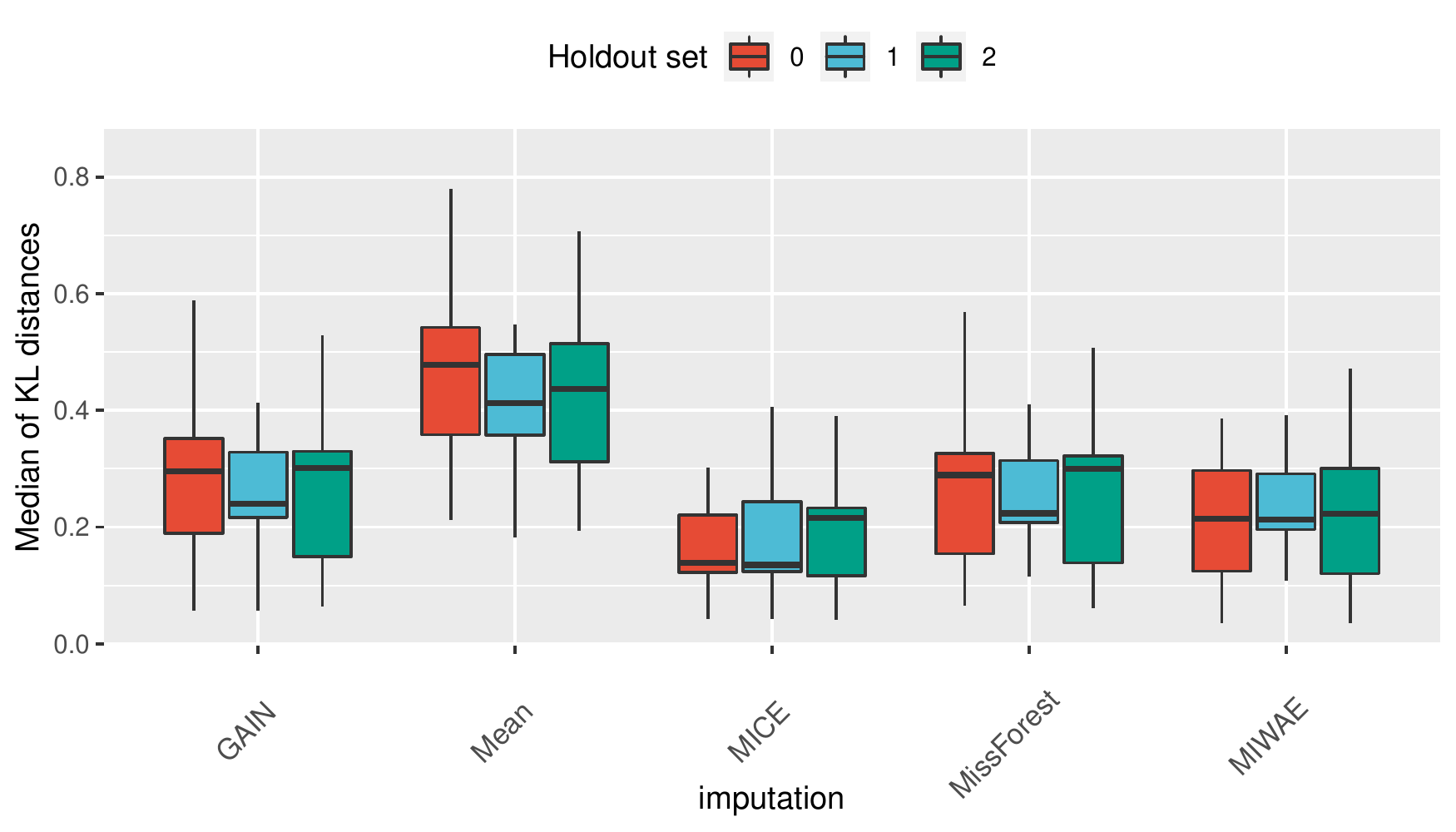}&
      \includegraphics[height=4cm,width=5cm]{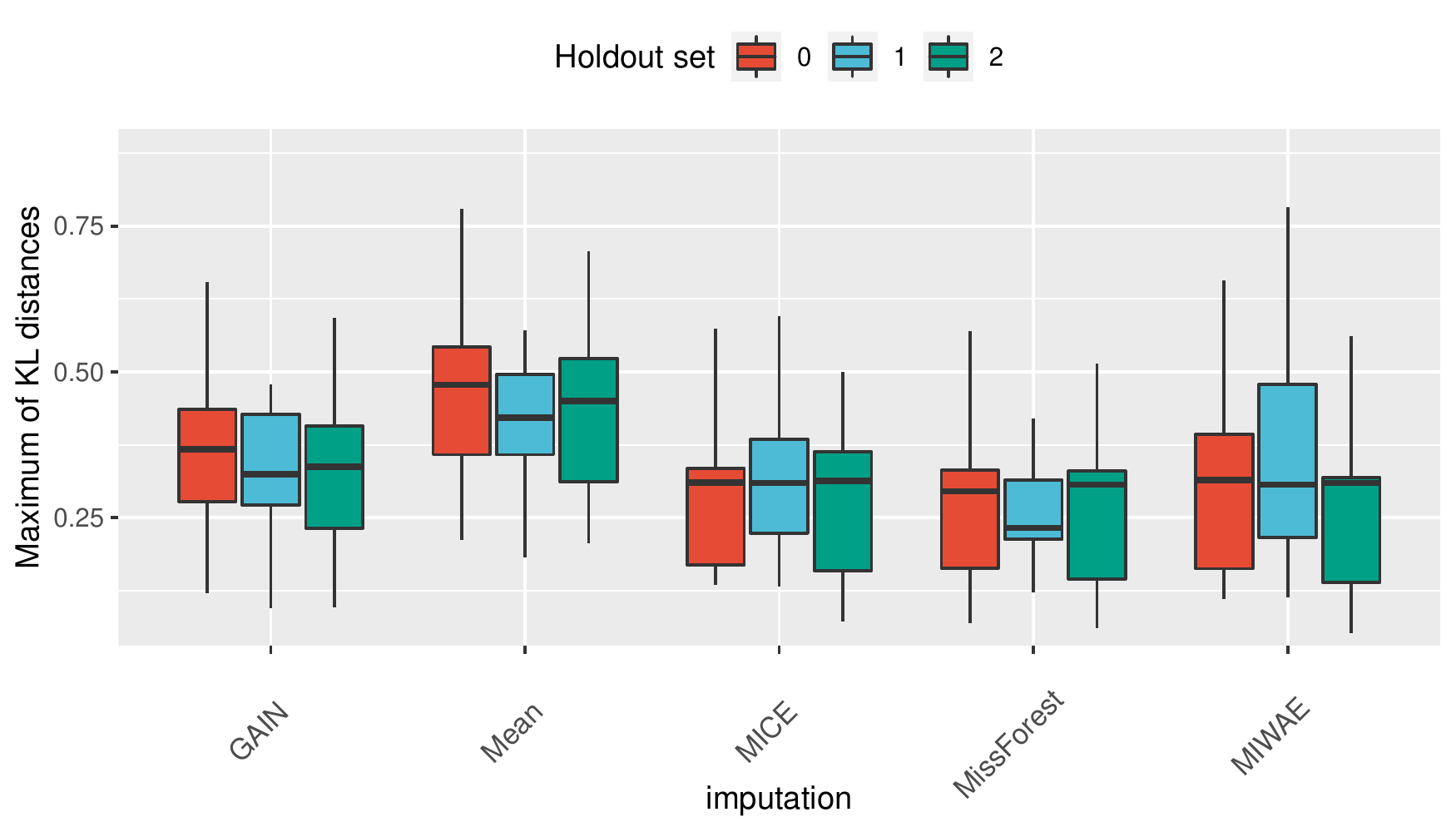}\\
      \hline
      \parbox[c][][c]{0.5in}{\rotatebox[origin=c]{90}{B2: Kolmogorov-Smirnov}} &
      \includegraphics[height=4cm,width=5cm]{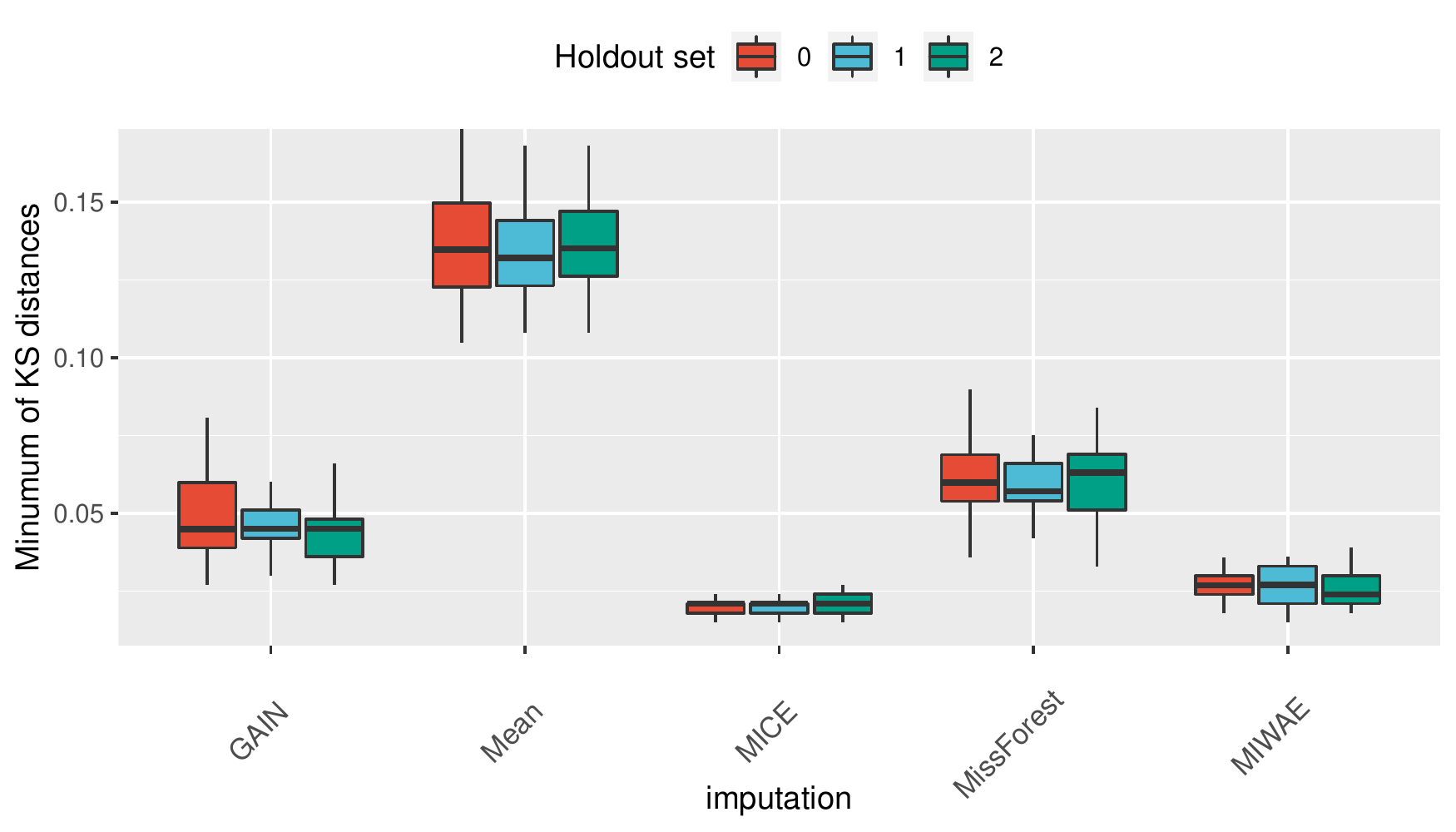}&
      \includegraphics[height=4cm,width=5cm]{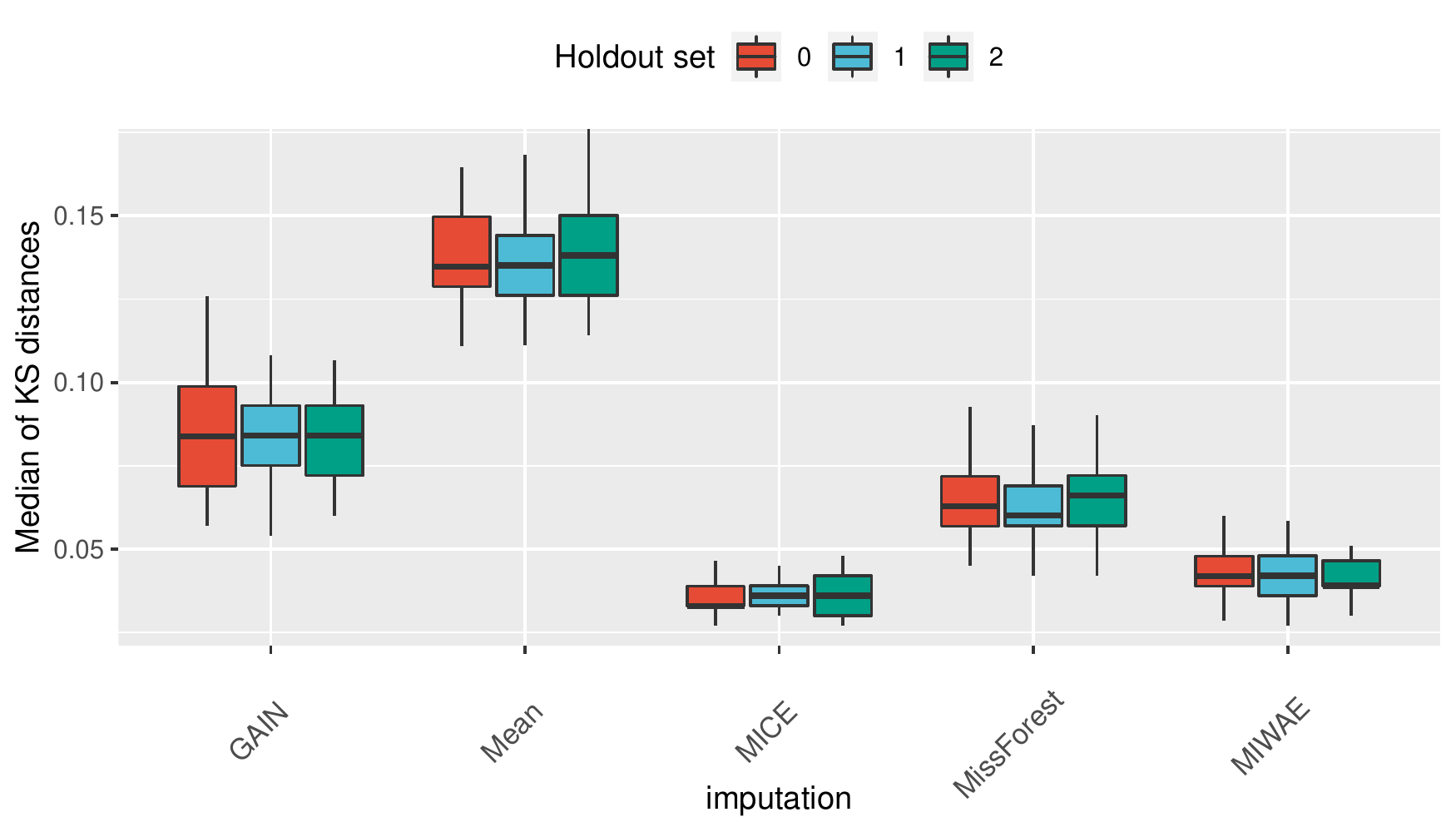}&
      \includegraphics[height=4cm,width=5cm]{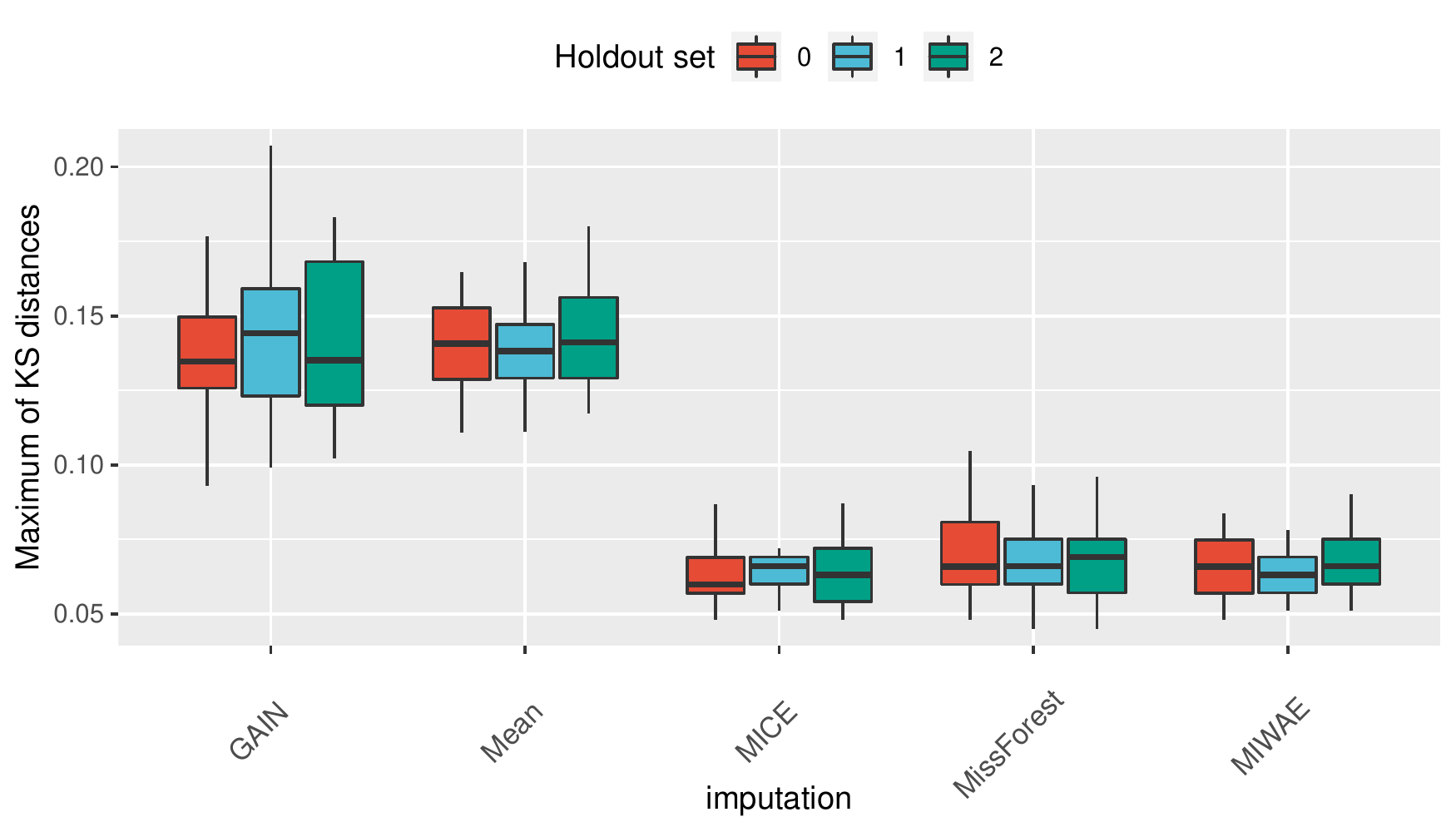}\\
      \hline
      \parbox[c][][c]{0.5in}{\rotatebox[origin=c]{90}{B3: Wasserstein}} &
      \includegraphics[height=4cm,width=5cm]{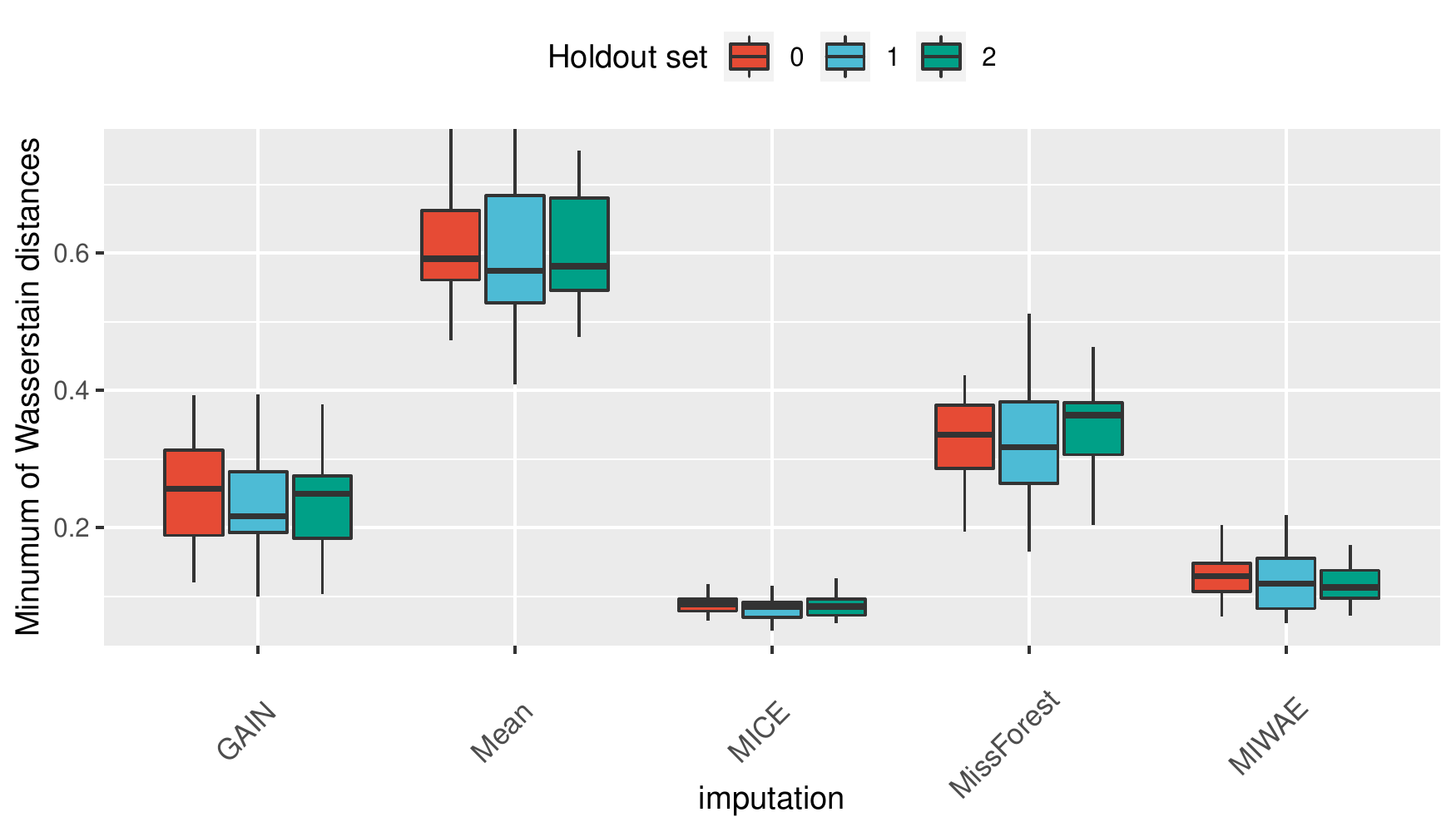}&
      \includegraphics[height=4cm,width=5cm]{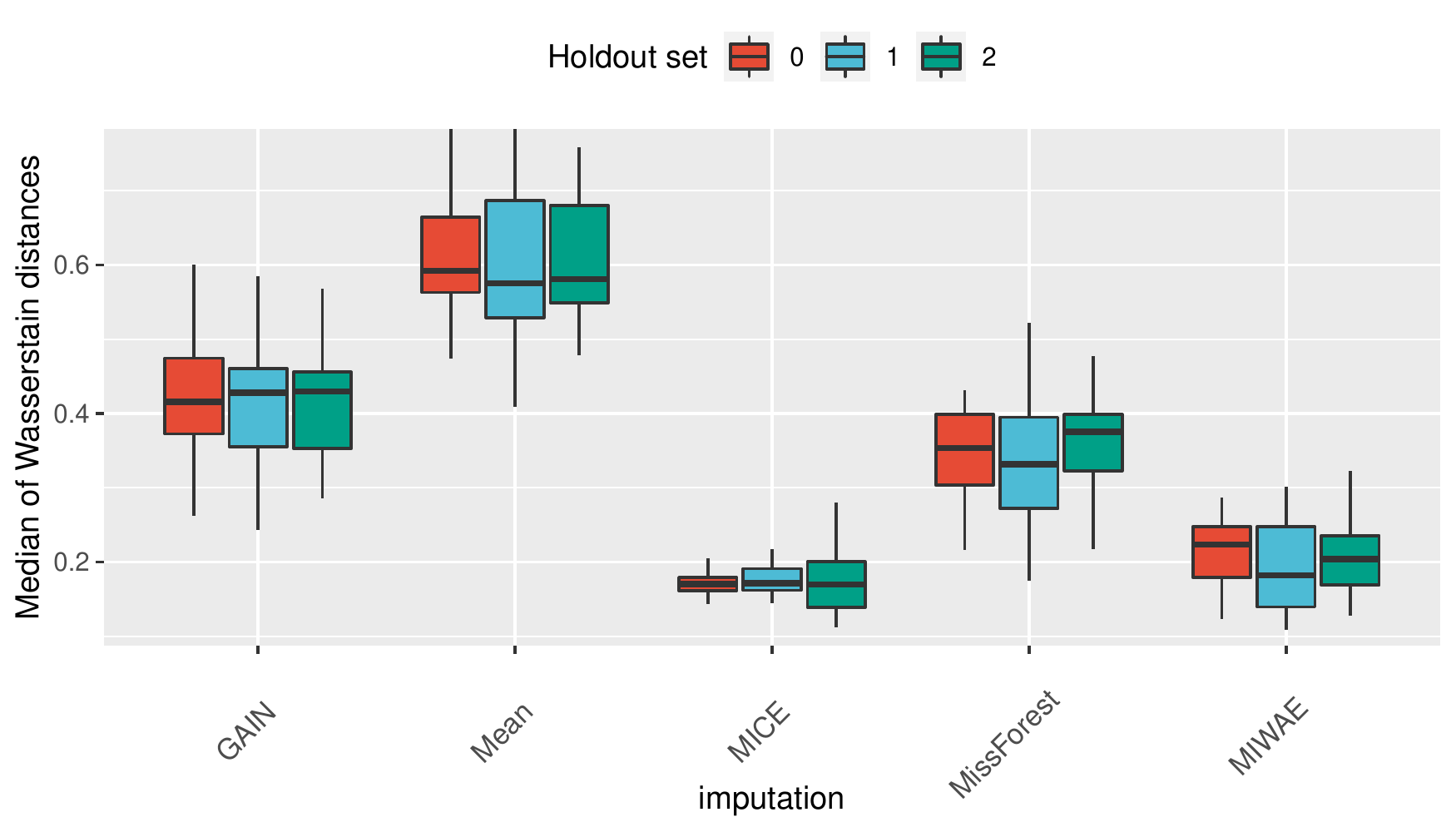}&
      \includegraphics[height=4cm,width=5cm]{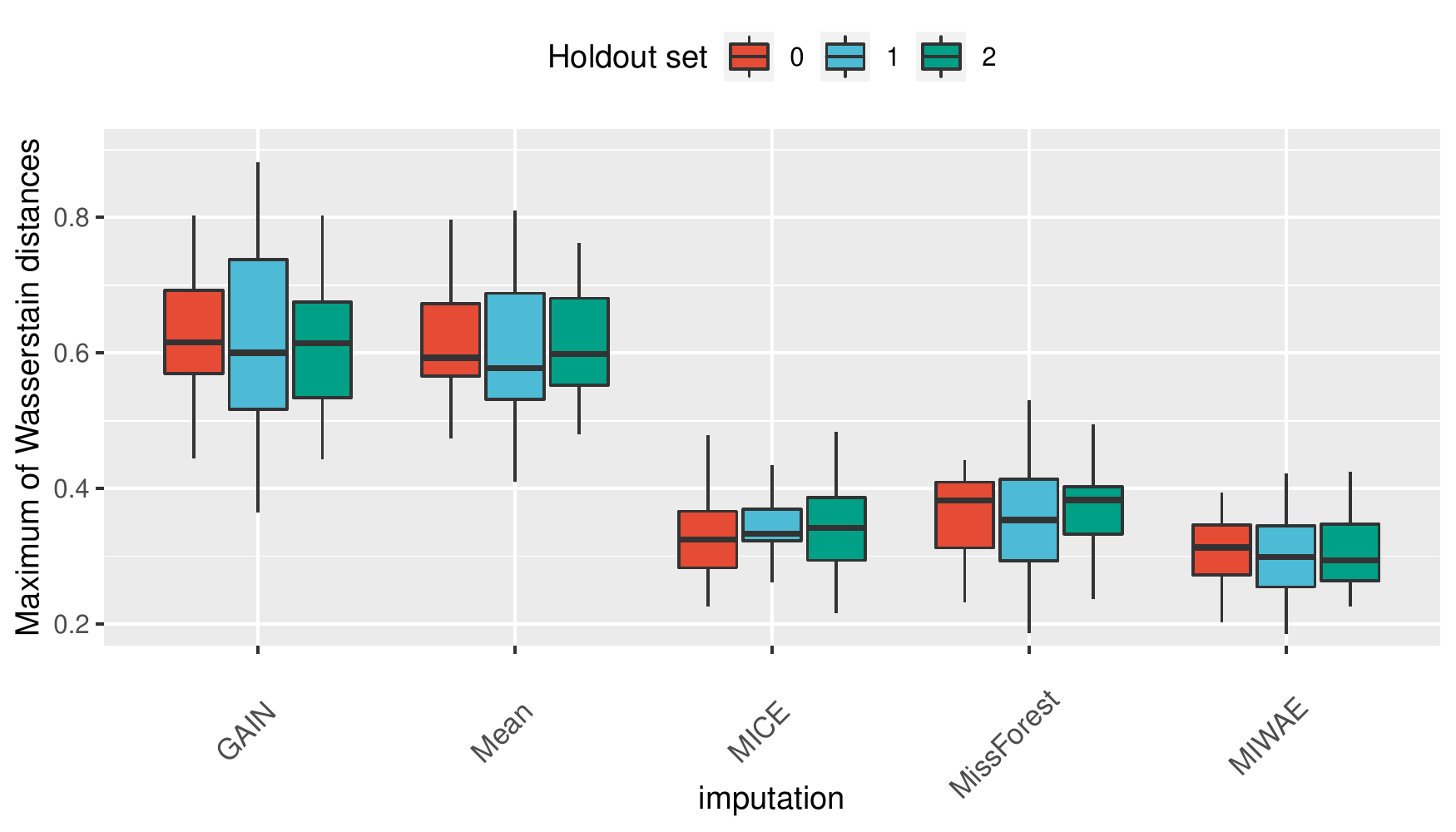}\\
      \hline
      \parbox[c][][c]{0.5in}{\rotatebox[origin=c]{90}{B3 excl. Mean}} &
      \includegraphics[height=4cm,width=5cm]{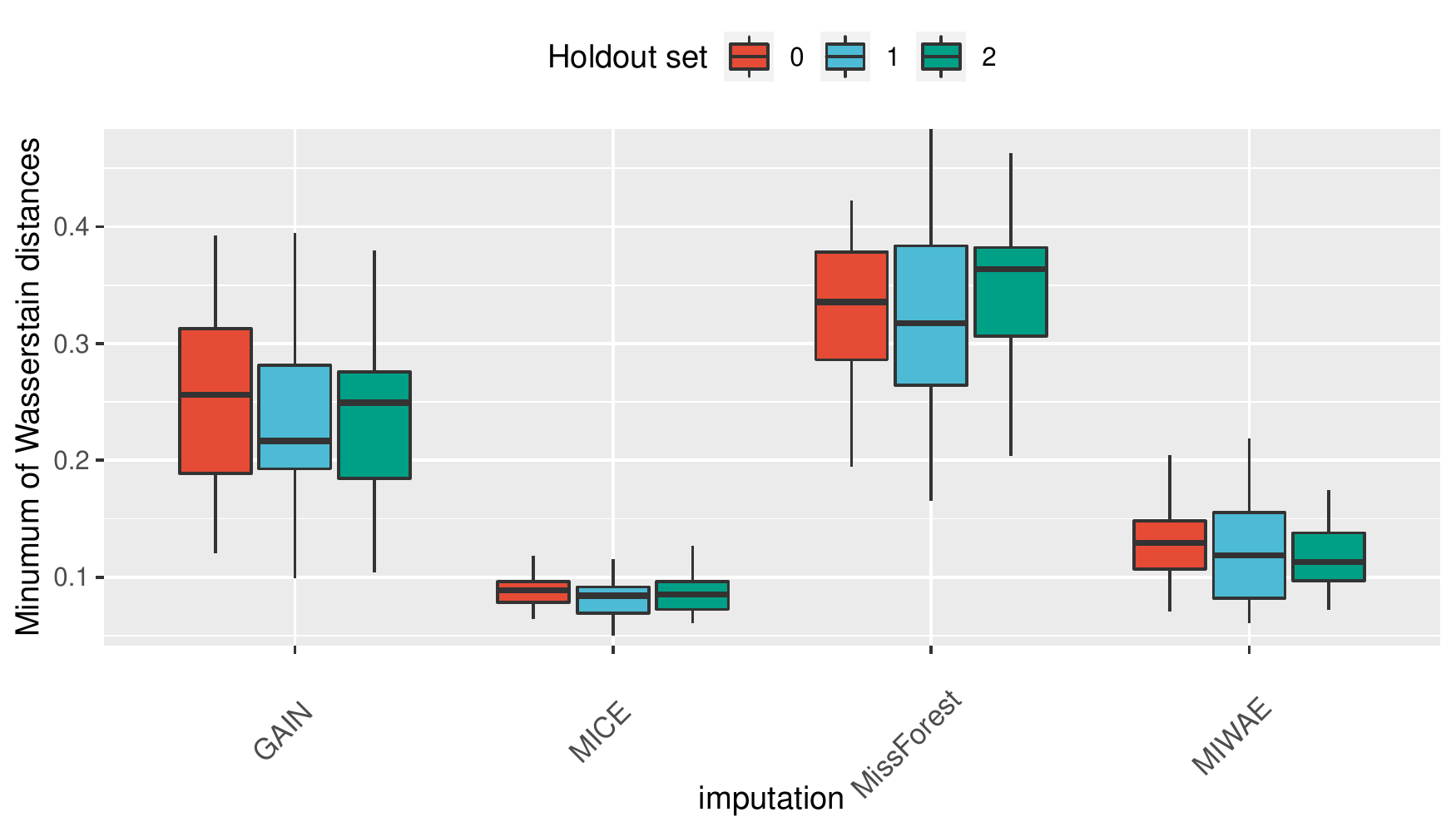}&
      \includegraphics[height=4cm,width=5cm]{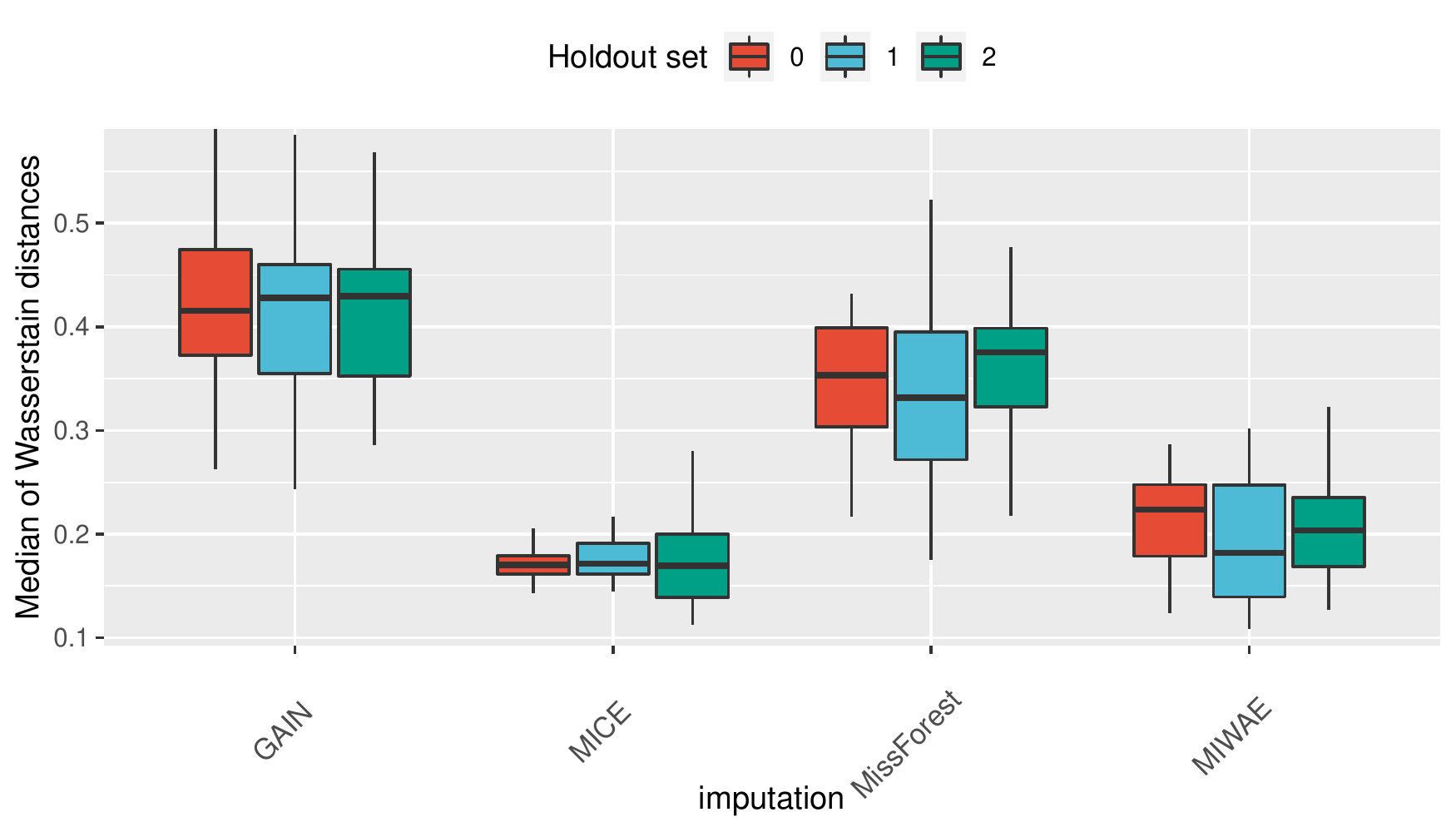}&
      \includegraphics[height=4cm,width=5cm]{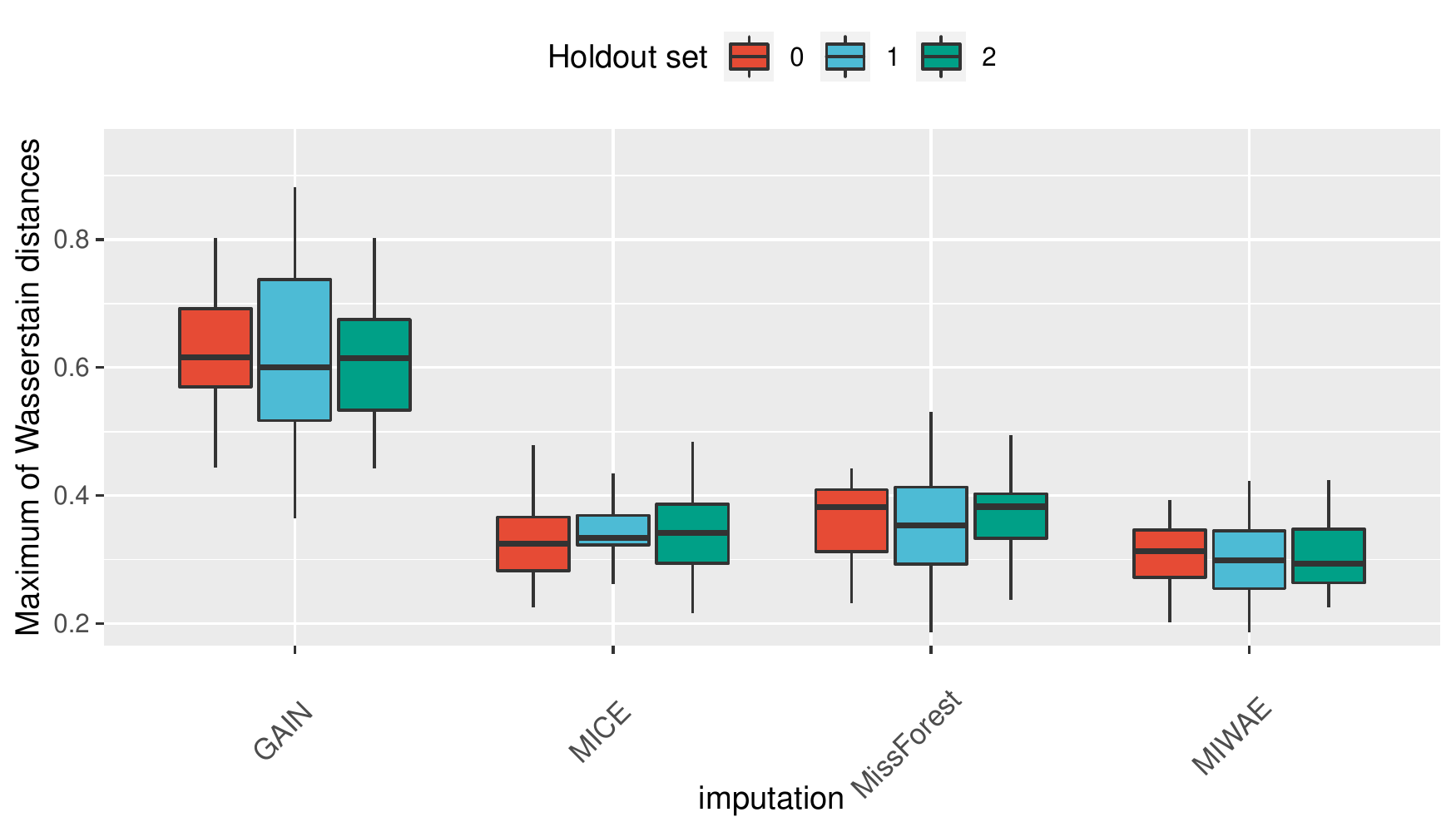}\\
    \end{tabular}
    \caption{Feature-wise 50\% train missingness and 25\% test missingness.}
    \label{fig:featurewise_syn_50_25}
\end{figure}

\begin{figure}[htb!]
    \centering
    \begin{tabular}{m{0.2in} | M{5cm} | M{5cm} | M{5cm}}
    & \textbf{Minimum} & \textbf{Median} & \textbf{Maximum} \\
    \hline
     \parbox[c][][c]{0.5in}{\rotatebox[origin=t]{90}{B1: Kullback-Leibler}} &
      \includegraphics[height=4cm,width=5cm]{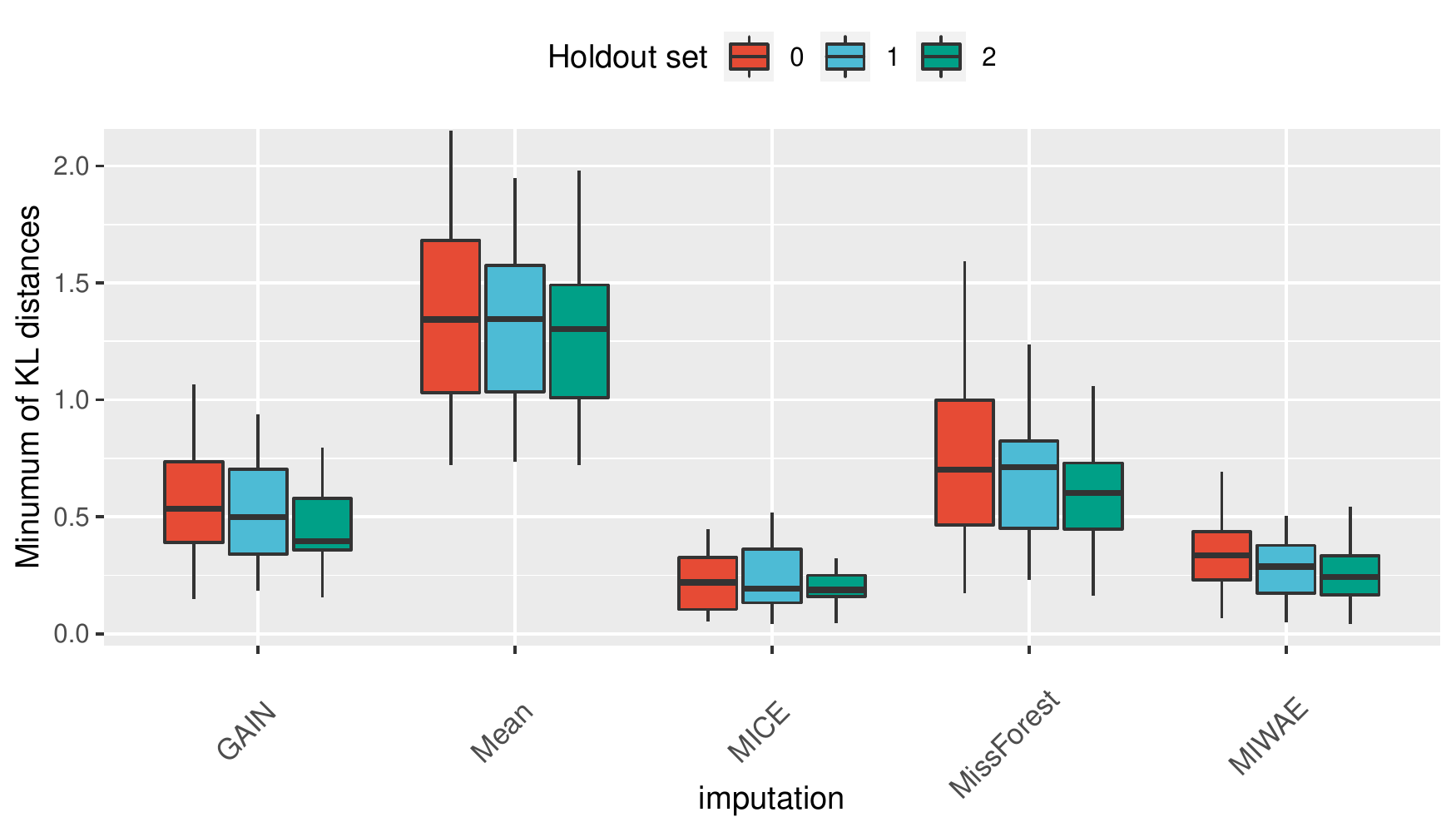}&
      \includegraphics[height=4cm,width=5cm]{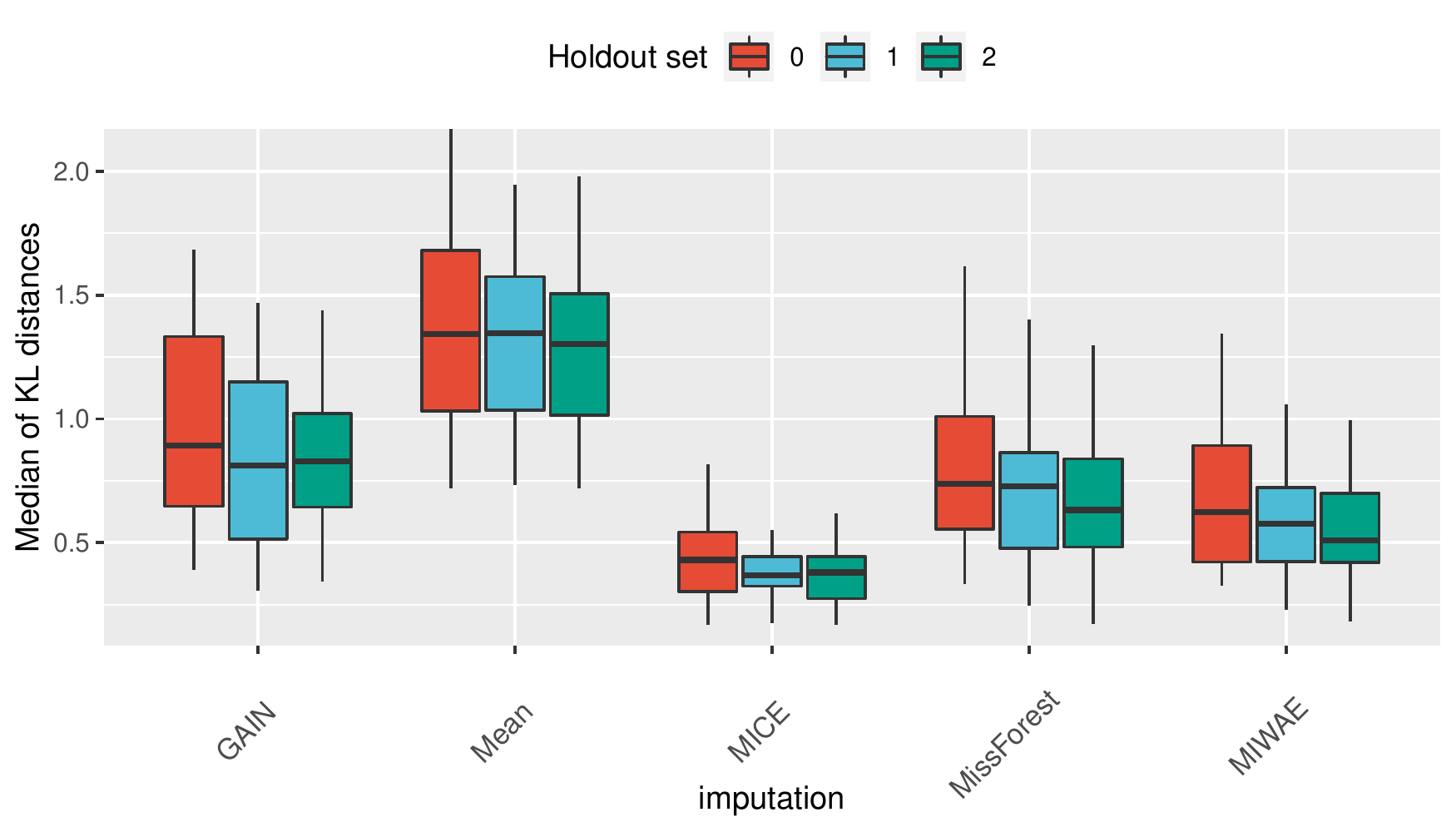}&
      \includegraphics[height=4cm,width=5cm]{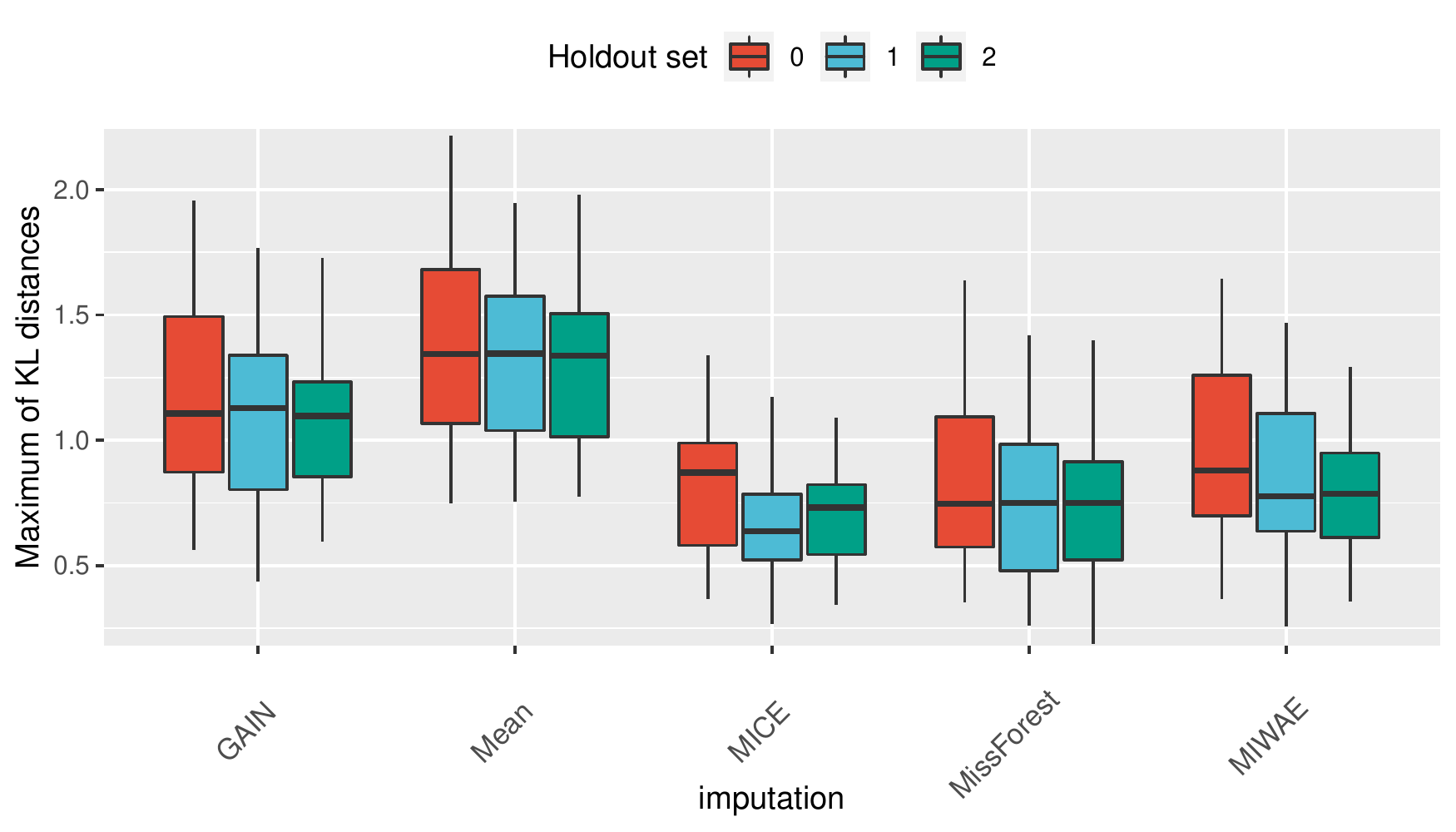}\\
      \hline
      \parbox[c][][c]{0.5in}{\rotatebox[origin=c]{90}{B2: Kolmogorov-Smirnov}} &
      \includegraphics[height=4cm,width=5cm]{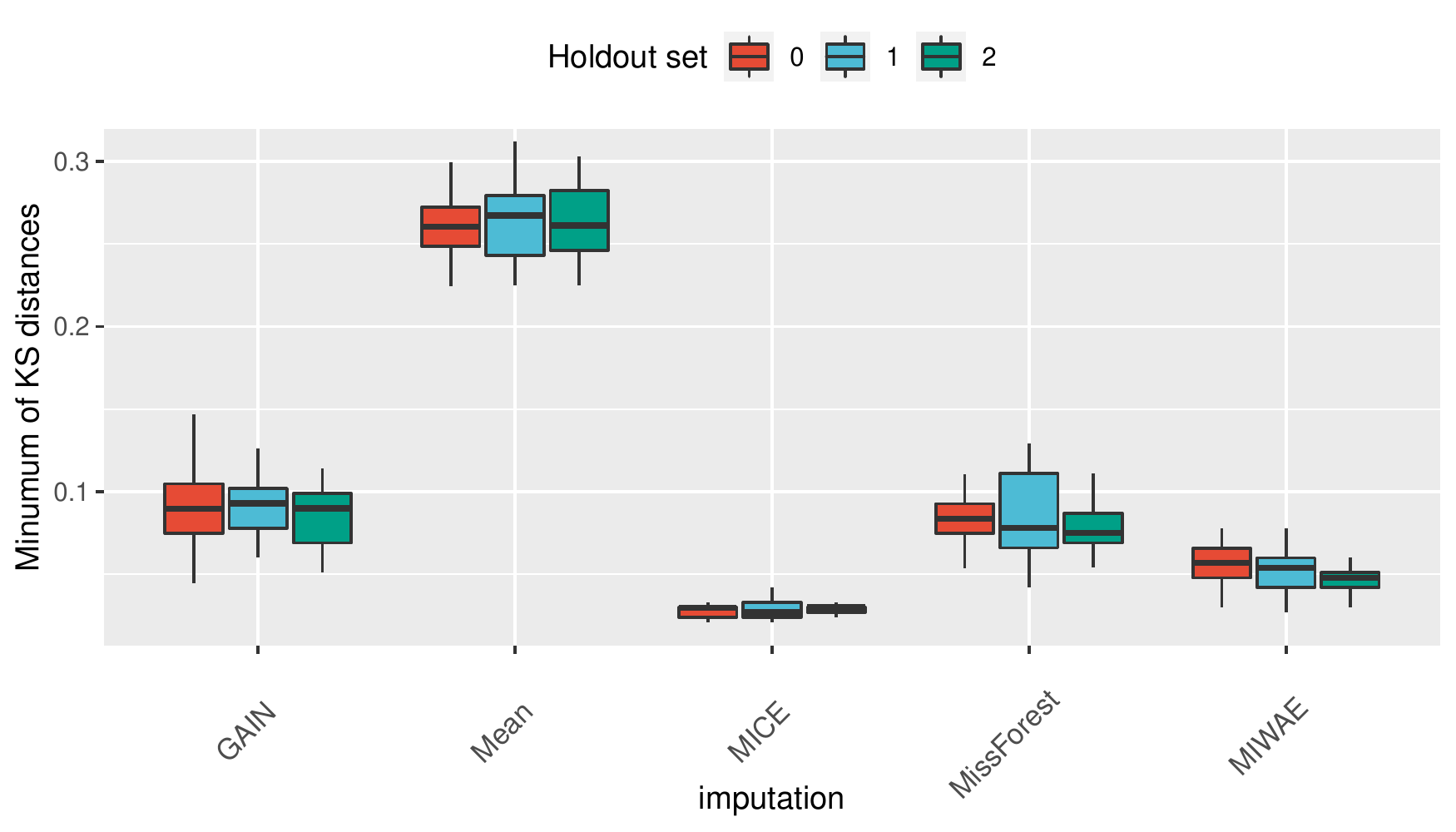}&
      \includegraphics[height=4cm,width=5cm]{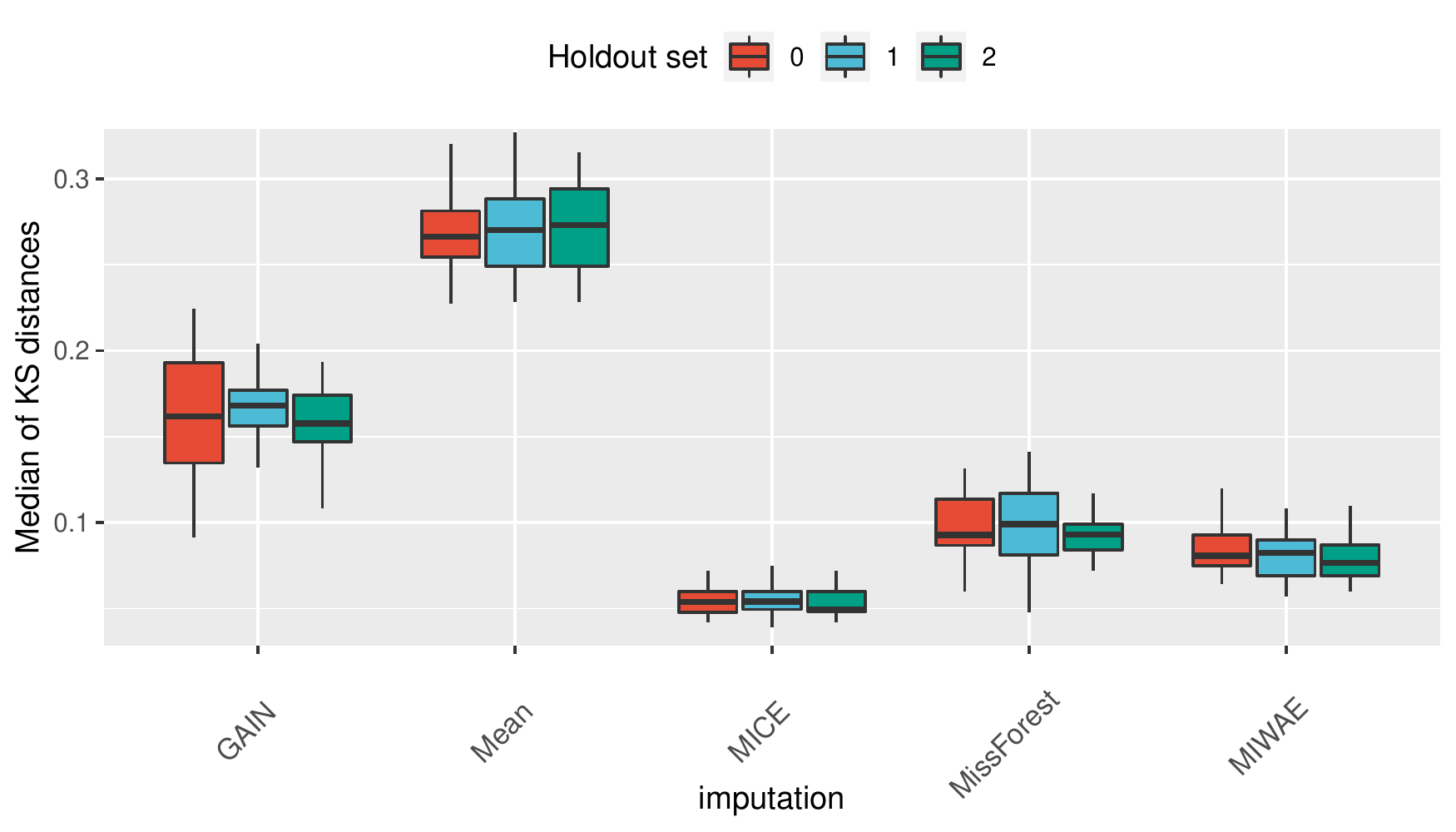}&
      \includegraphics[height=4cm,width=5cm]{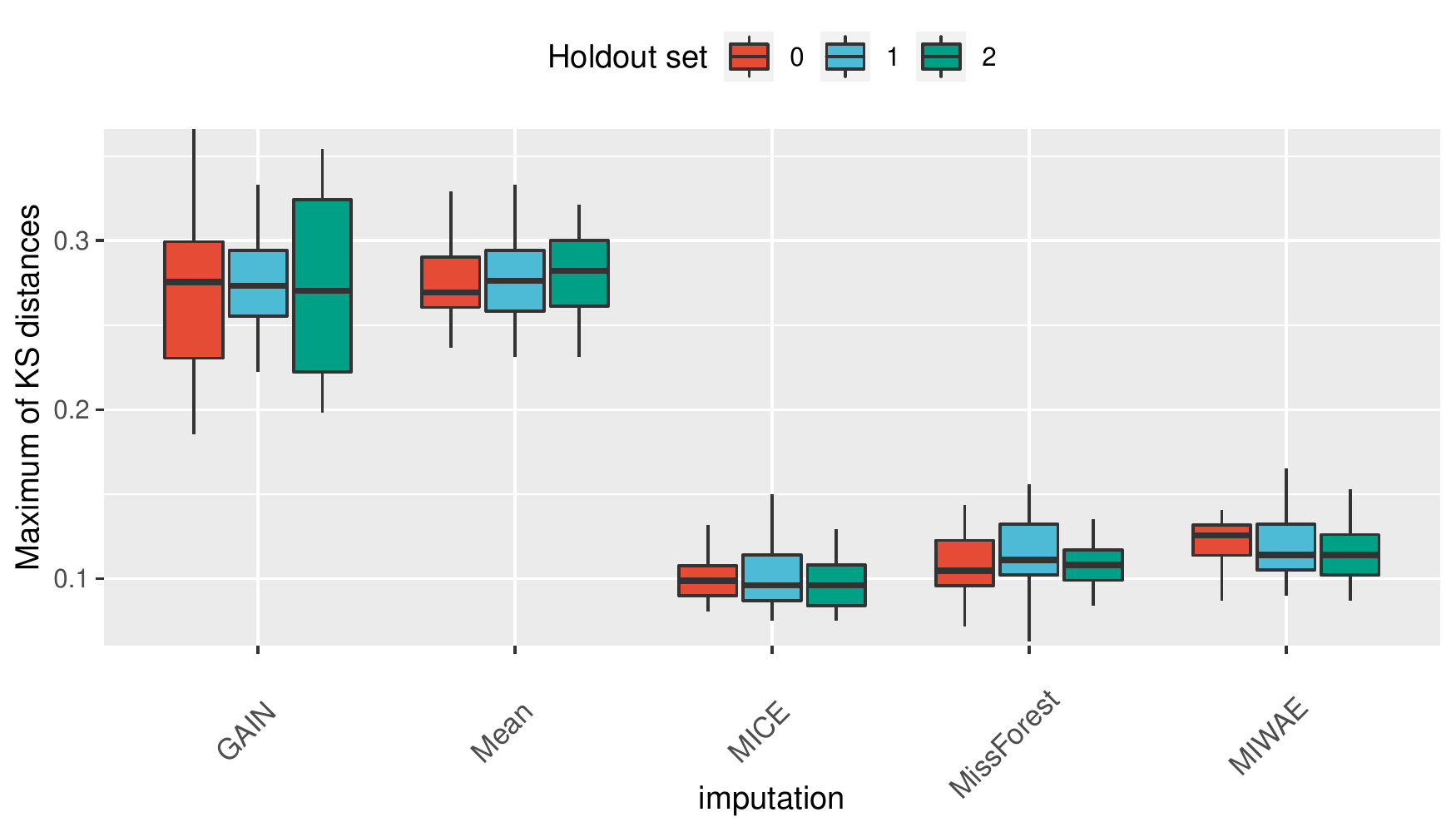}\\
      \hline
      \parbox[c][][c]{0.5in}{\rotatebox[origin=c]{90}{B3: Wasserstein}} &
      \includegraphics[height=4cm,width=5cm]{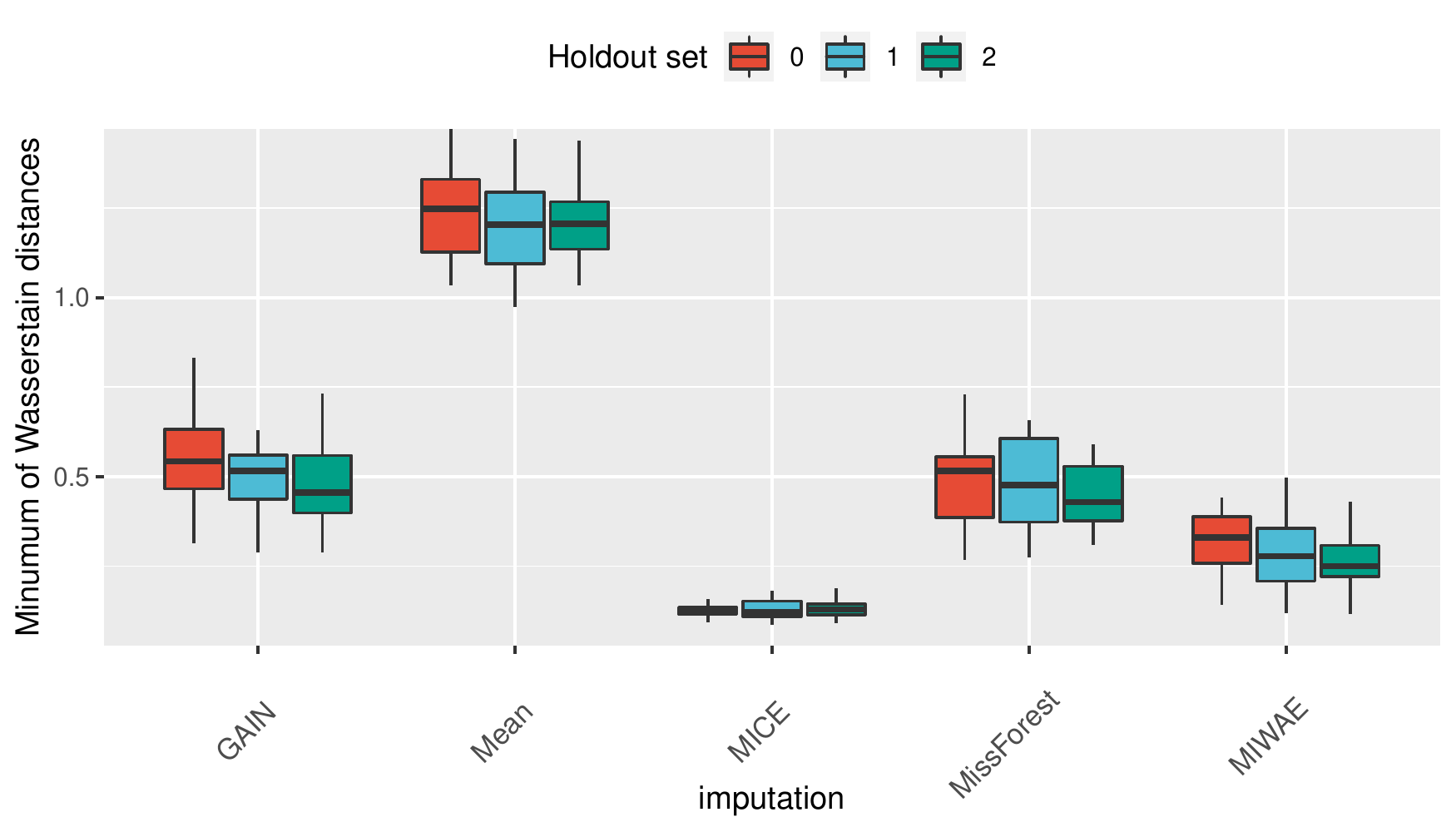}&
      \includegraphics[height=4cm,width=5cm]{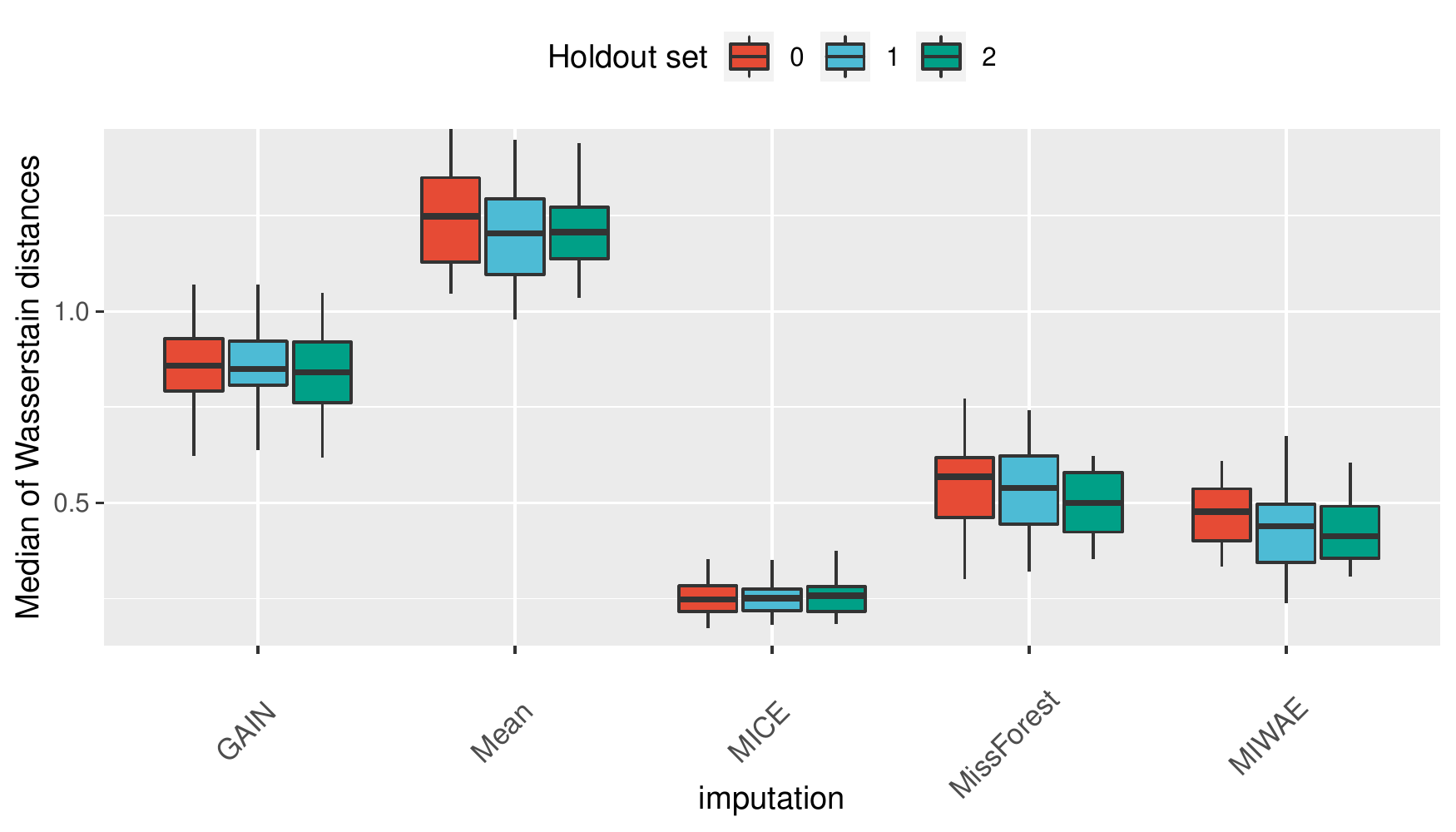}&
      \includegraphics[height=4cm,width=5cm]{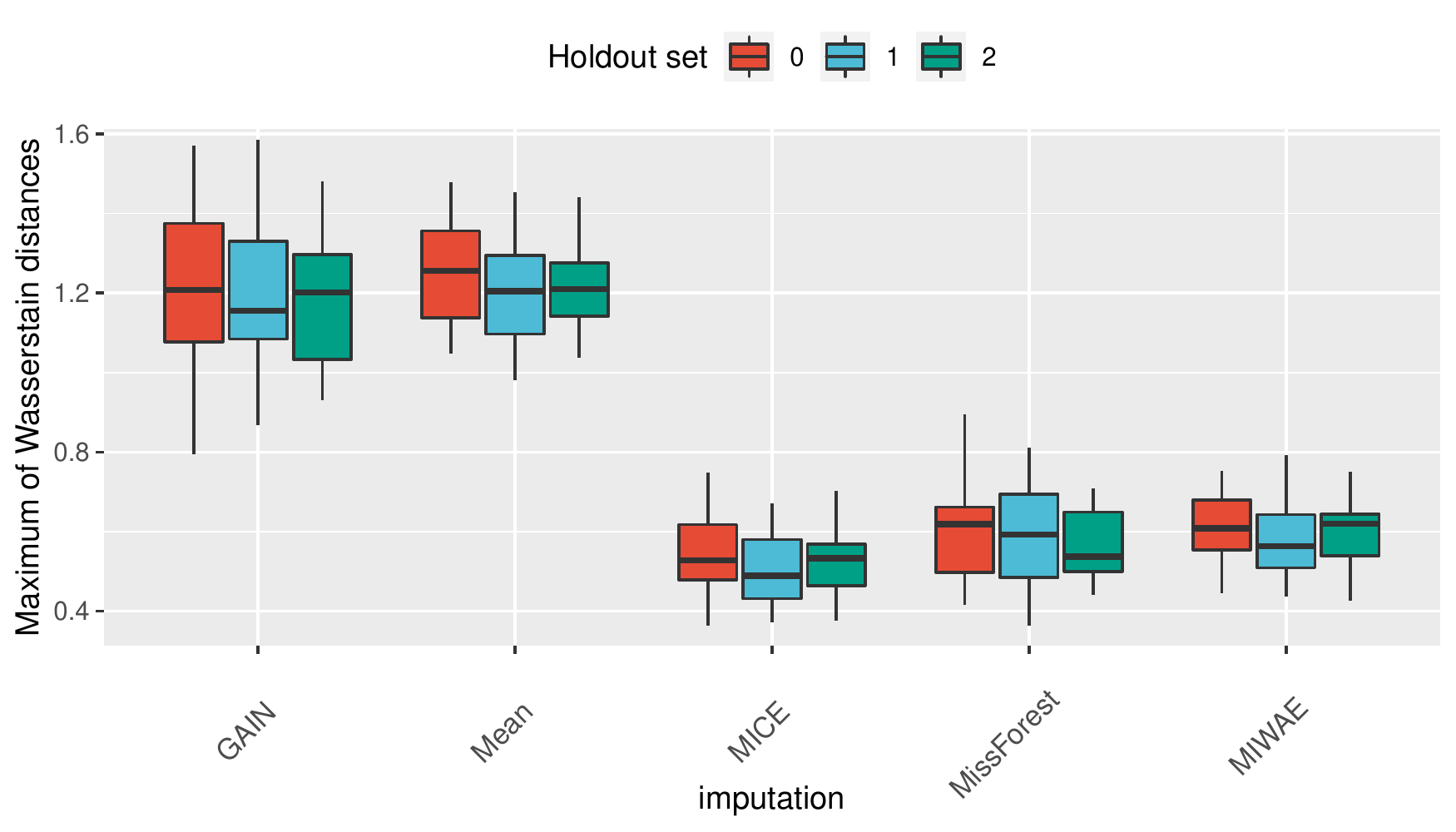}\\
      \hline
      \parbox[c][][c]{0.5in}{\rotatebox[origin=c]{90}{B3 excl. Mean}} &
      \includegraphics[height=4cm,width=5cm]{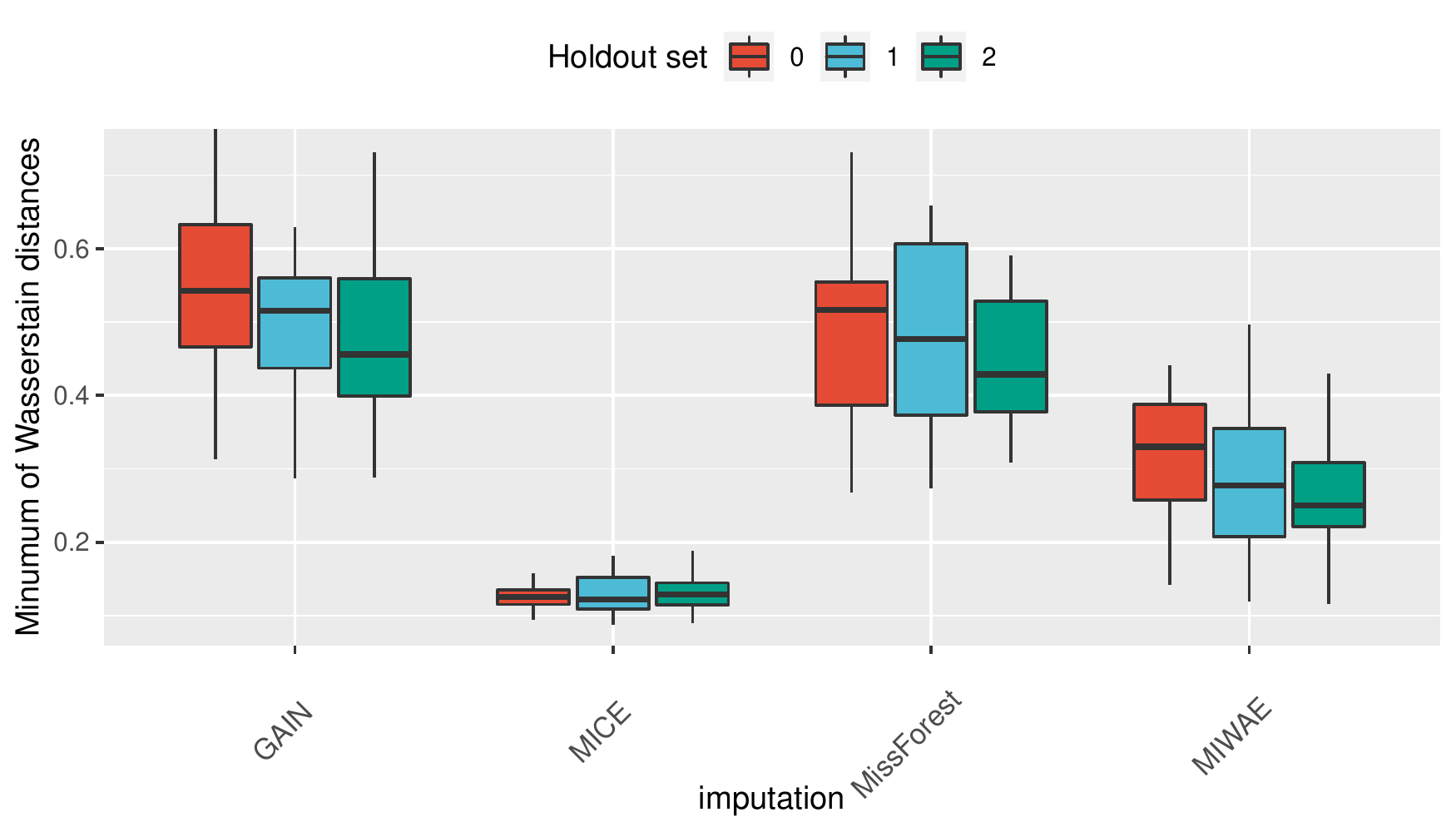}&
      \includegraphics[height=4cm,width=5cm]{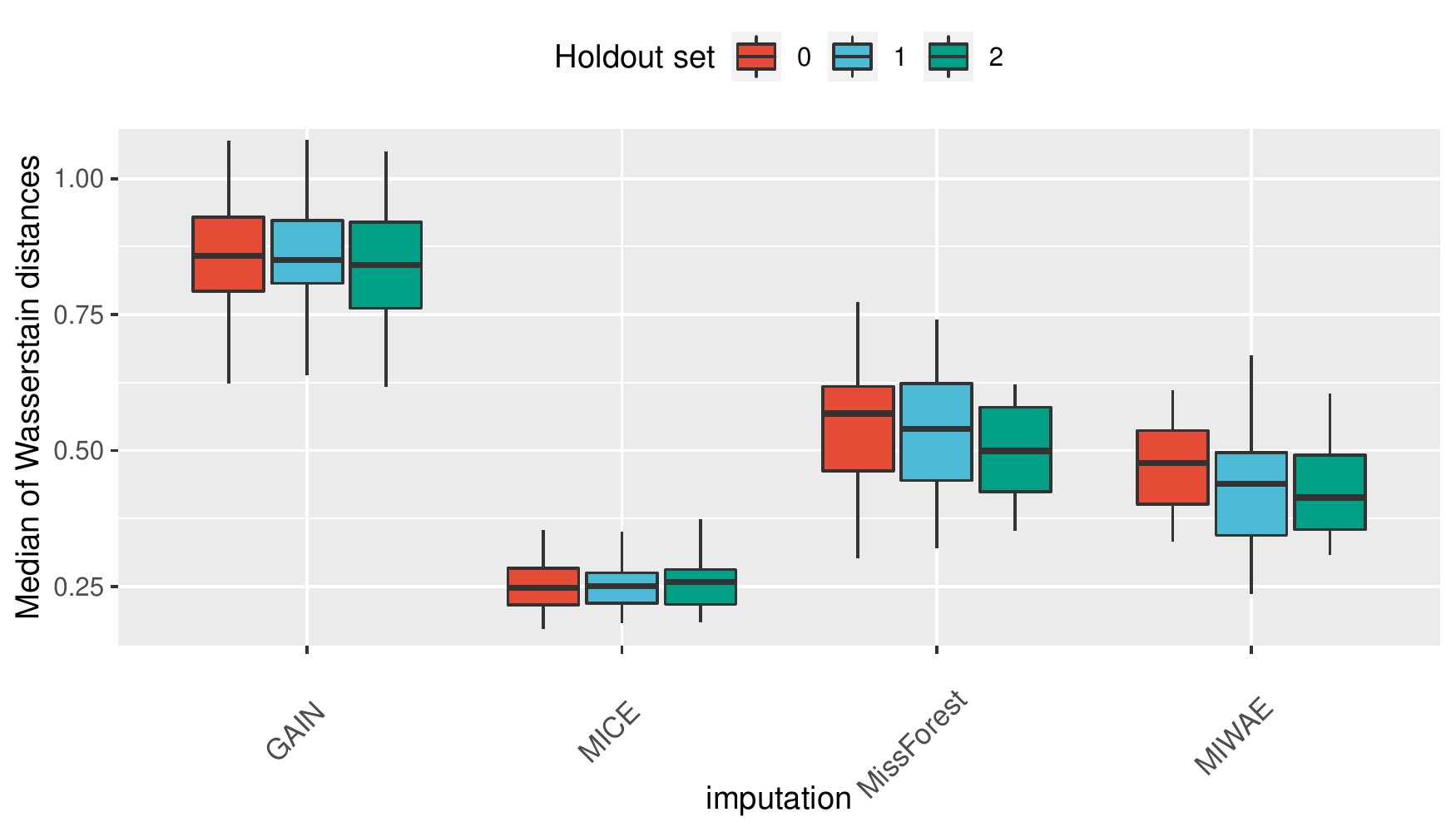}&
      \includegraphics[height=4cm,width=5cm]{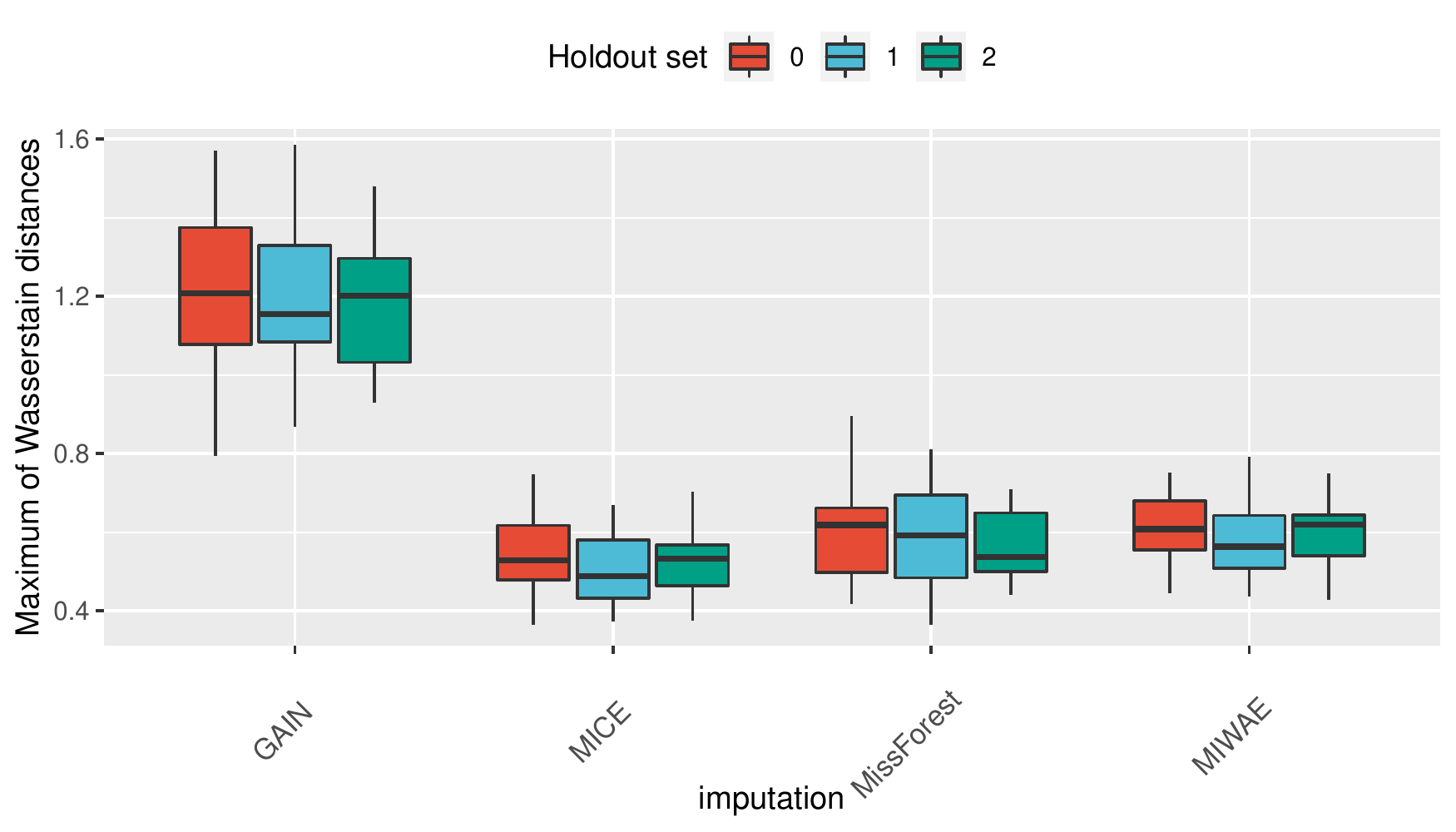}\\
    \end{tabular}
    \caption{Feature-wise 50\% train missingness and 50\% test missingness.}
    \label{fig:featurewise_syn_50_50}
\end{figure}

\noindent
{\emph{B. Feature distribution metrics.}} 
In Figure~\ref{fig:featurewise_mimic_25_25} and Supplementary Figures~\ref{fig:featurewise_mimic_25_50}--\ref{fig:featurewise_syn_50_50} we show the minimum, median and maximum feature-wise discrepancies statistics for the \textbf{MIMIC-III} and \textbf{Simulated} datasets. The mean imputation method is the worst in all metrics for minimum, median and maximum at all rates of train and test missingness in both datasets. MICE imputation, which performed worst by the sample-wise discrepancy scores, is the best performing method by the Kolmogorov-Smirnoff statistic and Wasserstein distance for all missingness rates across all the minimum, median and maximum discrepancies. It is generally the best method by Kullback-Leibler divergence, with MissForest and MIWAE also competitive. For GAIN, the minimum discrepancies are competitive with the other imputation methods. However, when considering the maximum discrepancy, we note that the difference in performance of the mean and GAIN methods narrows significantly. Increasing the test missingness rate leads to a significant increase in the feature-wise distances, whereas there is a more subtle increase in the distances with an increase in the train missingness rate from 25\% to 50\%. \\

\begin{figure}[htb!]
    \centering
    \begin{tabular}{M{5cm} | M{5cm} | M{5cm}}
     \textbf{Kullback-Leibler} & \textbf{Kolmogorov-Smirnoff} & \textbf{Wasserstein} \\
    \hline
      \includegraphics[height=4cm,width=5cm]{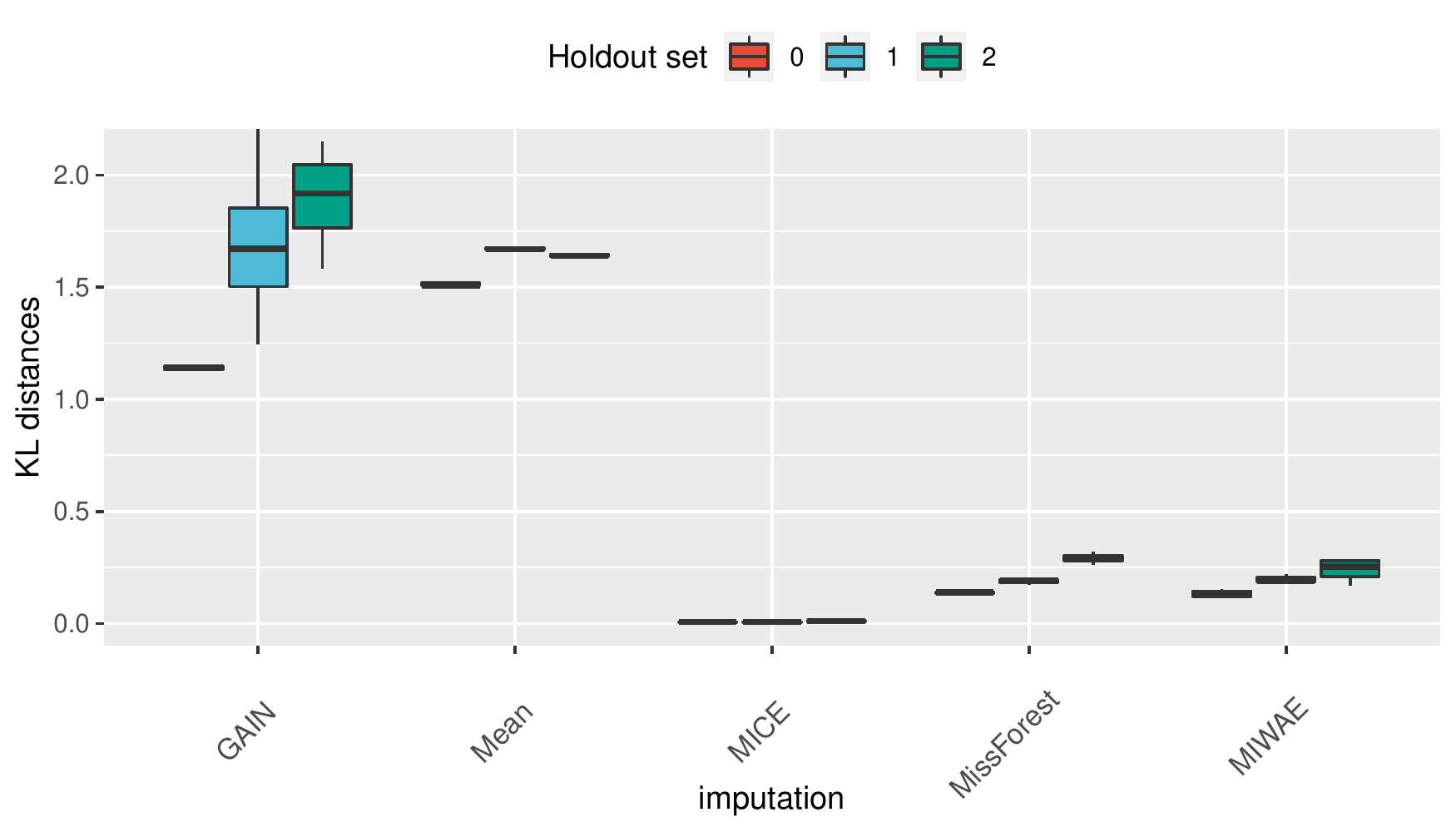}&
      \includegraphics[height=4cm,width=5cm]{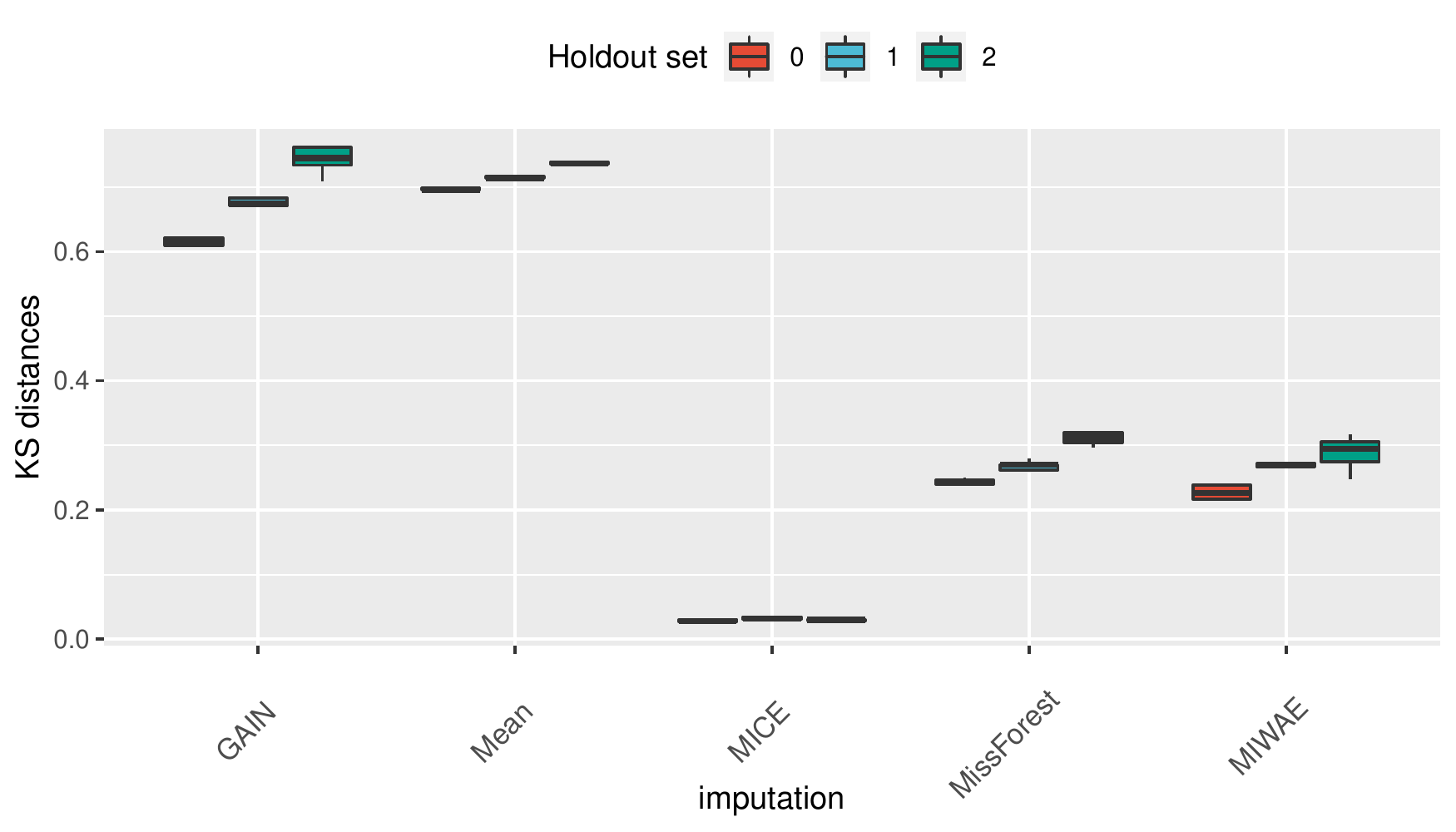}&
      \includegraphics[height=4cm,width=5cm]{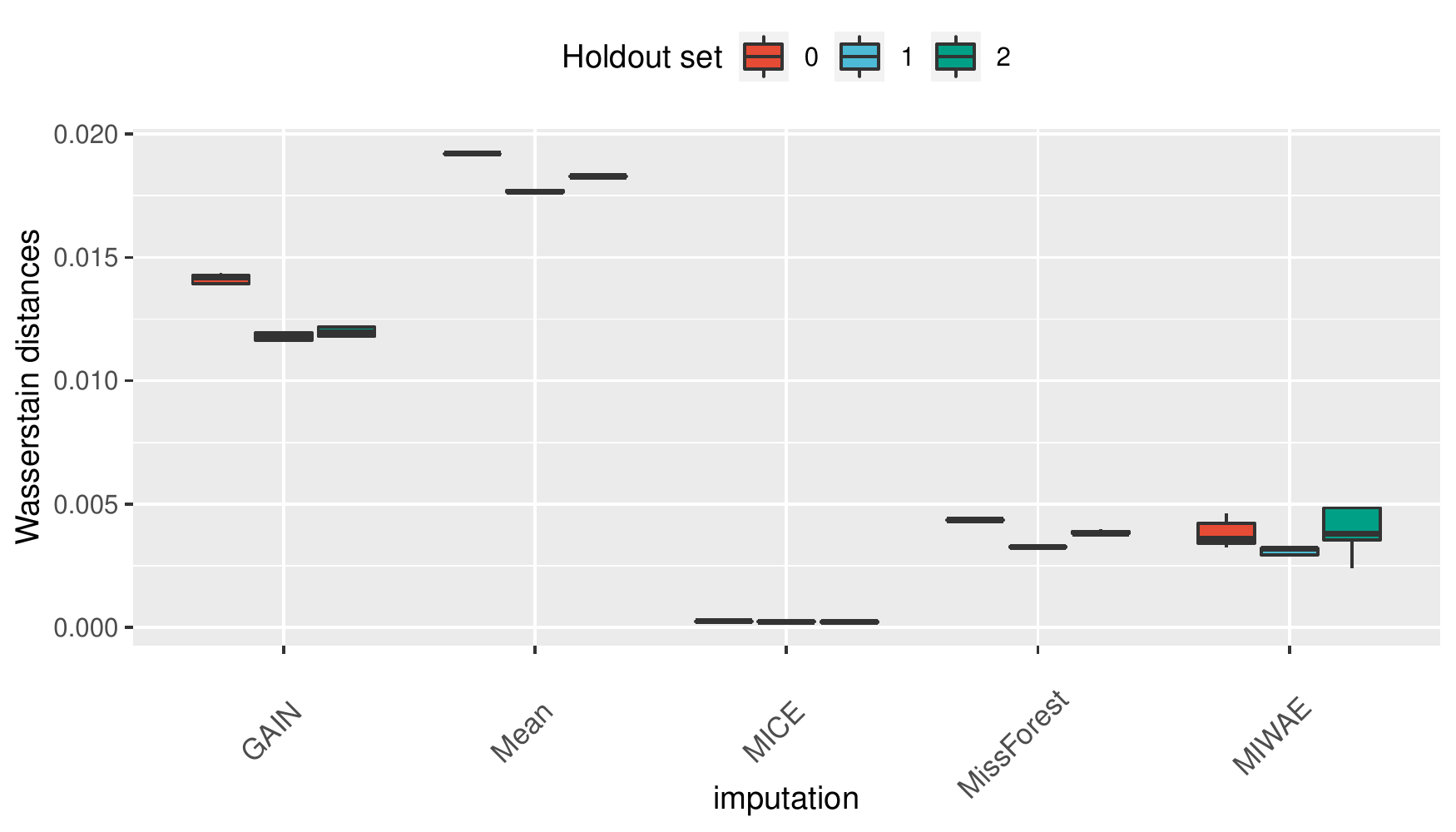}\\
      %\hline      \includegraphics[height=4cm,width=5cm]{MIMIC_proj_w_dist_dist_boxplots_logscale/MIMIC_0.25_0.25_kl.pdf}&
      %\includegraphics[height=4cm,width=5cm]{MIMIC_proj_w_dist_dist_boxplots_logscale/MIMIC_0.25_0.25_ks.pdf}&
      %\includegraphics[height=4cm,width=5cm]{MIMIC_proj_w_dist_dist_boxplots_logscale/MIMIC_0.25_0.25_W_dist.pdf}\\
    \end{tabular}
    \caption{The distribution discrepancy statistics of the sliced Wasserstein distances for the \textbf{MIMIC-III} dataset with 25\% train and test missingness rates.}
    \label{fig:datadistwise}
\end{figure}

\begin{figure}[htb!]
    \centering
    \begin{tabular}{m{0.2in} |M{5cm} | M{5cm} | M{5cm}}
     & \textbf{Kullback-Leibler} & \textbf{Kolmogorov-Smirnoff} & \textbf{Wasserstein} \\
     \hline
     \parbox[c][][c]{0.5in}{\rotatebox[origin=t]{90}{25\%: 25\%}} &
      \includegraphics[height=4cm,width=5cm]{MIMIC_proj_w_dist_dist_boxplots/MIMIC_0.25_0.25_kl.pdf}&
      \includegraphics[height=4cm,width=5cm]{MIMIC_proj_w_dist_dist_boxplots/MIMIC_0.25_0.25_ks.pdf}&
      \includegraphics[height=4cm,width=5cm]{MIMIC_proj_w_dist_dist_boxplots/MIMIC_0.25_0.25_W_dist.pdf}\\
      \hline      
      \parbox[c][][c]{0.5in}{\rotatebox[origin=t]{90}{25\%: 25\% (Log)}} &\includegraphics[height=4cm,width=5cm]{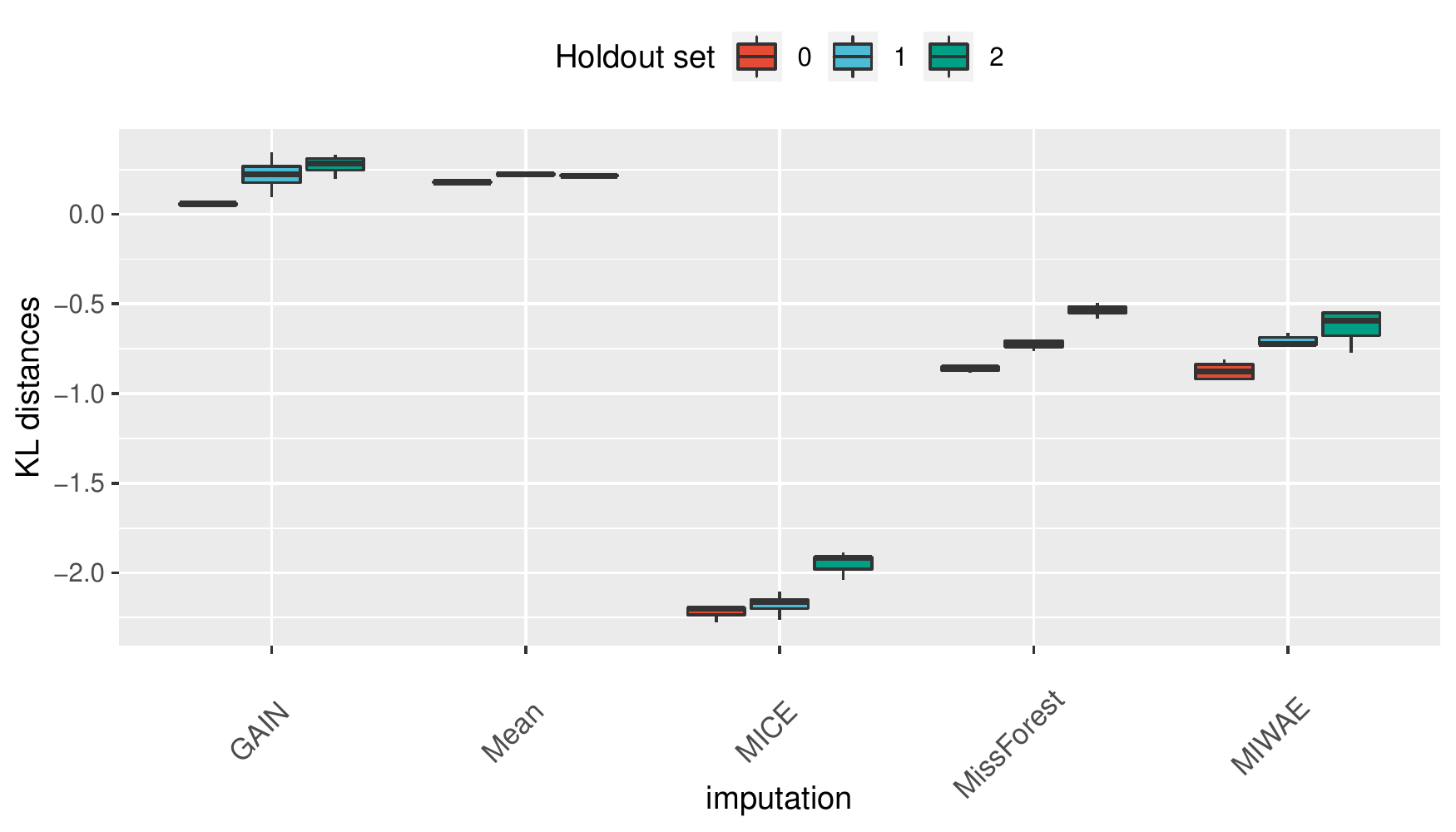}&
      \includegraphics[height=4cm,width=5cm]{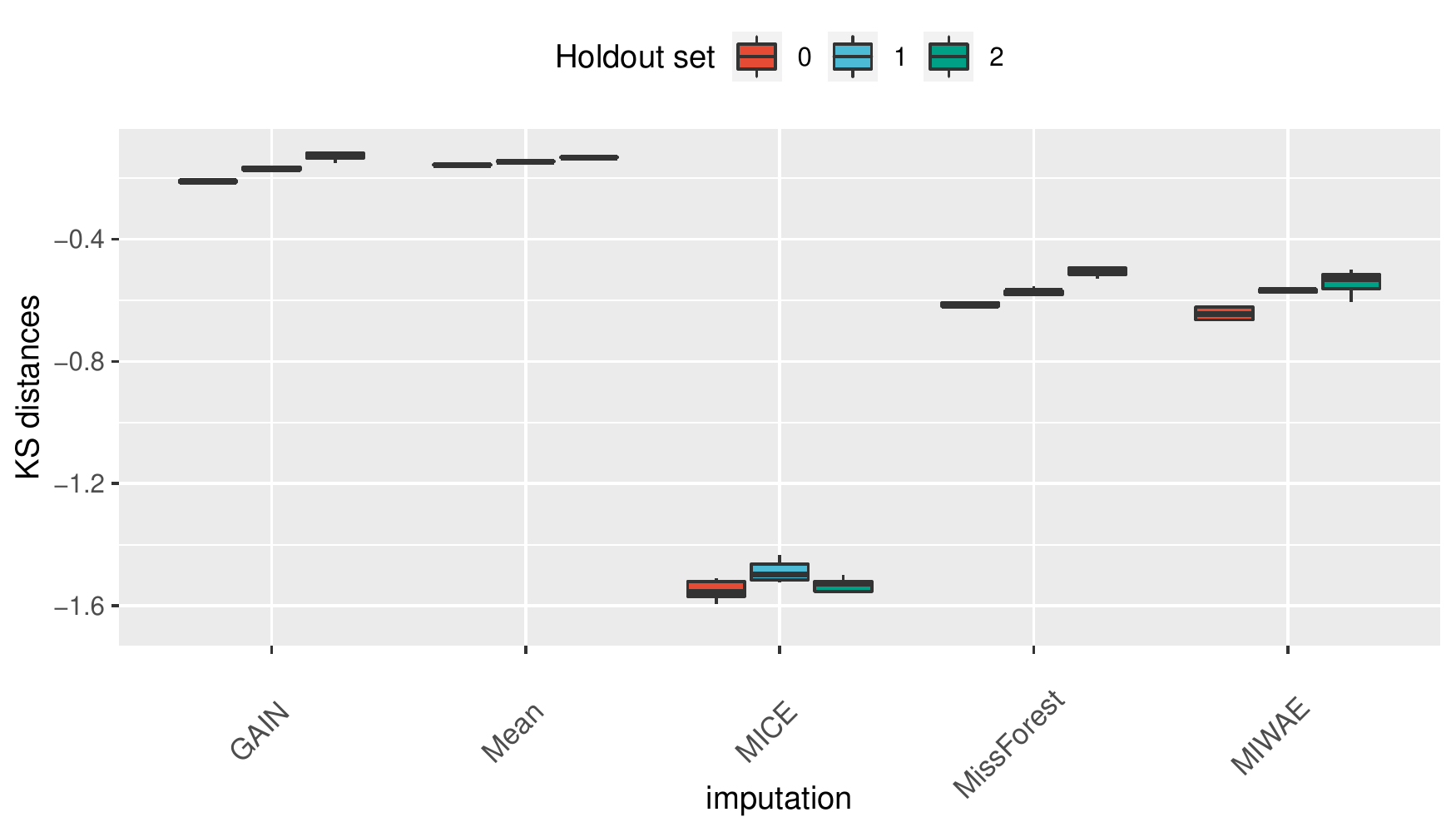}&
      \includegraphics[height=4cm,width=5cm]{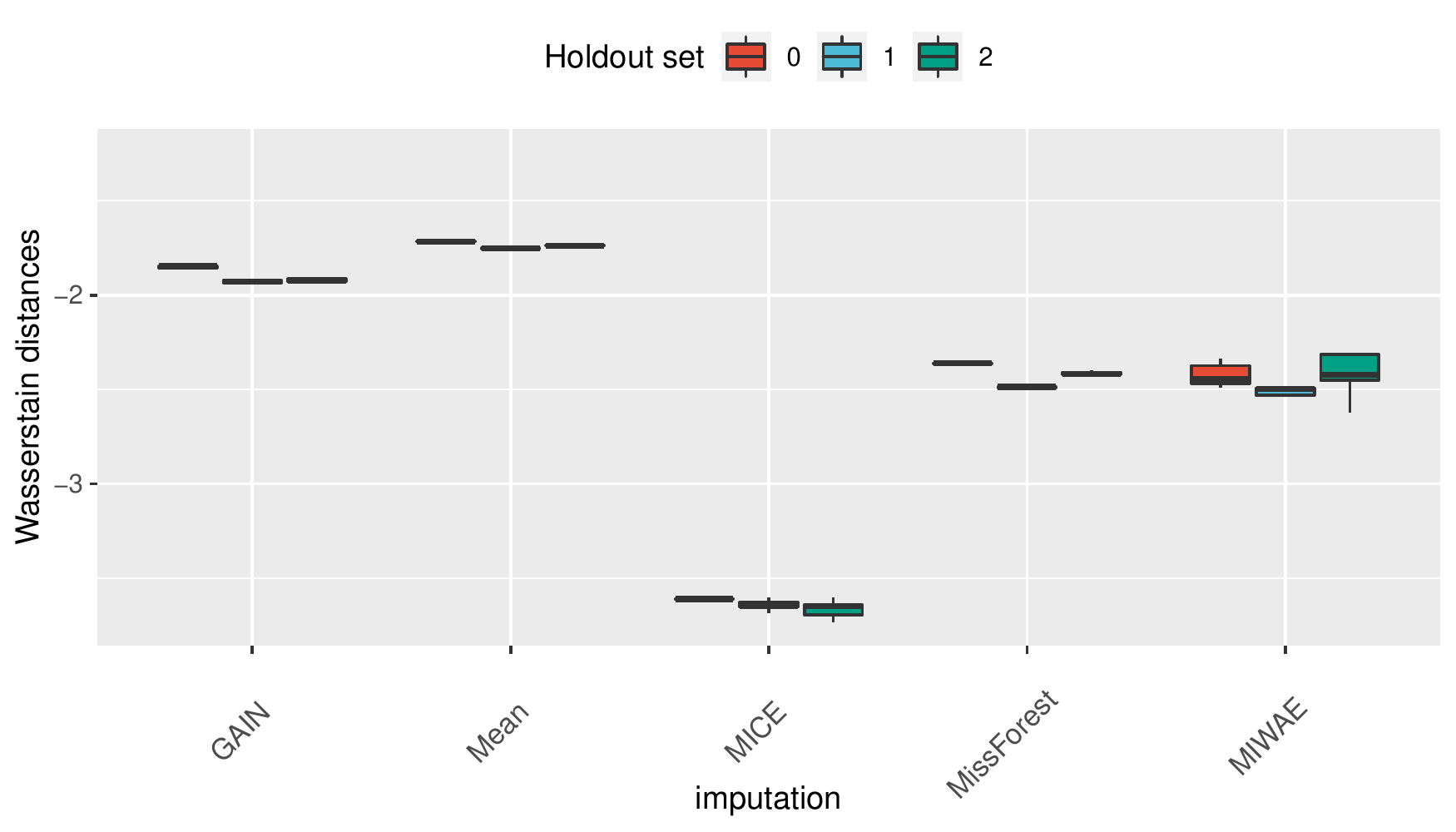}\\
     \hline
     \parbox[c][][c]{0.5in}{\rotatebox[origin=t]{90}{25\%: 50\%}} &
      \includegraphics[height=4cm,width=5cm]{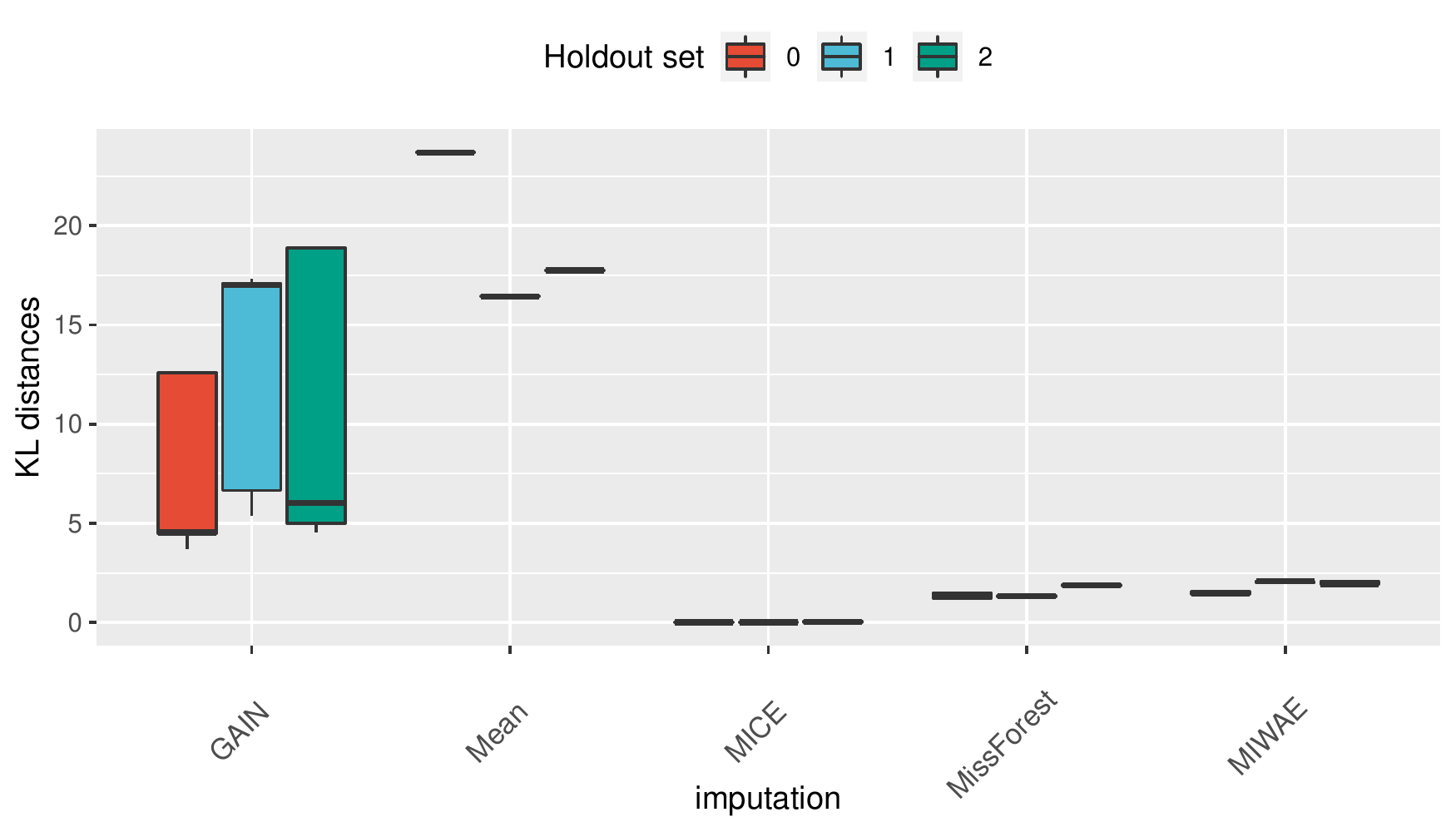}&
      \includegraphics[height=4cm,width=5cm]{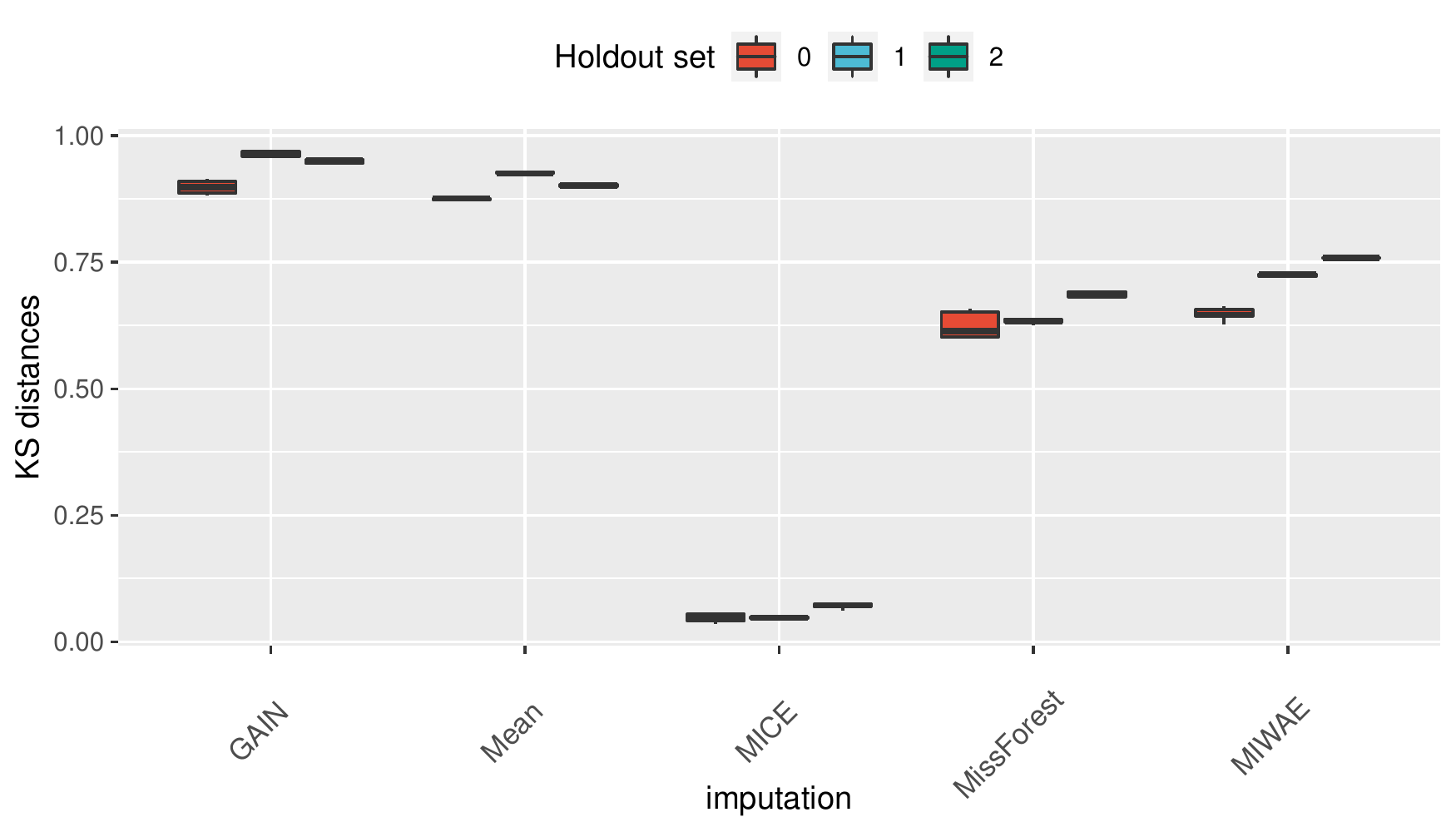}&
      \includegraphics[height=4cm,width=5cm]{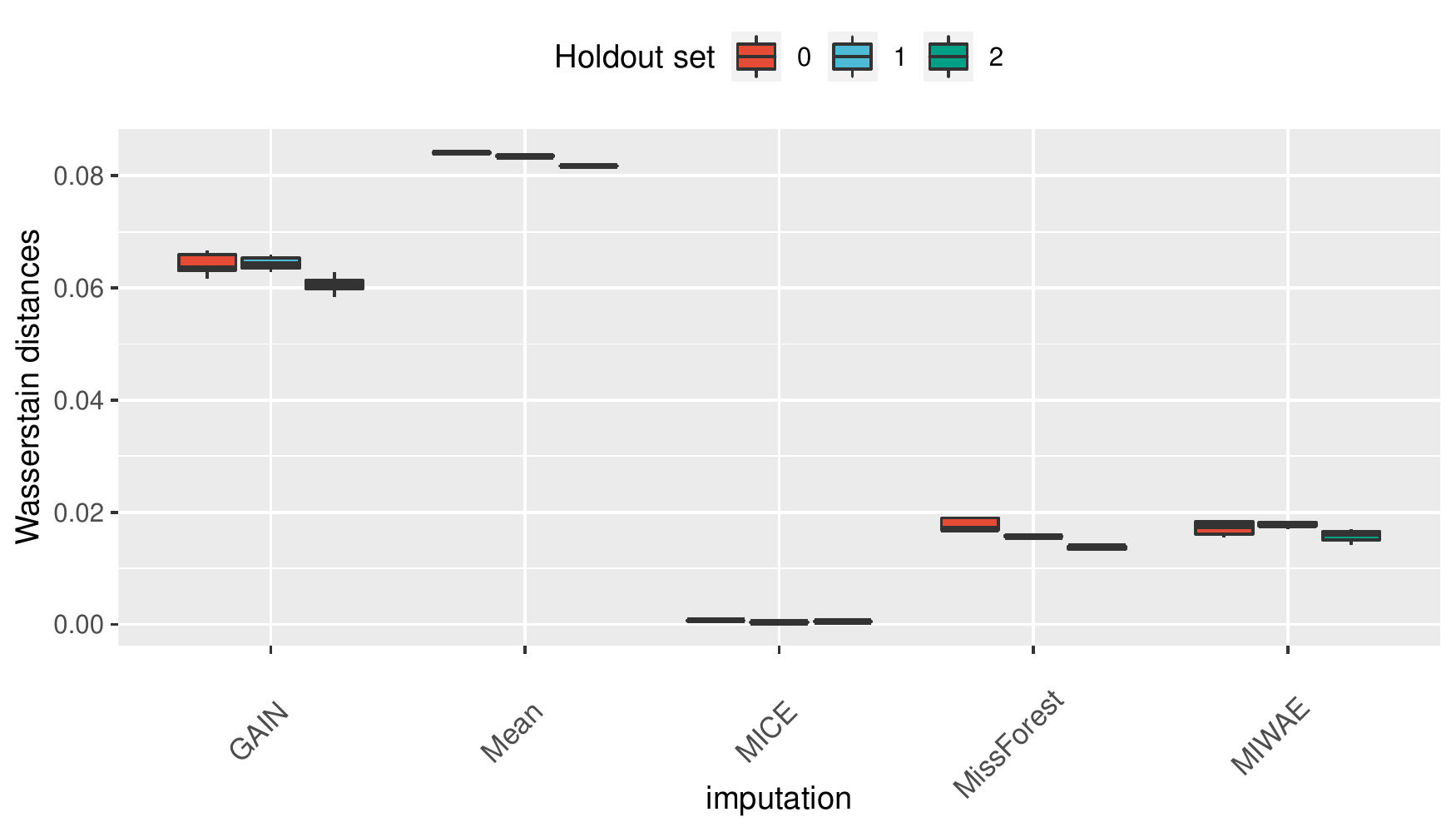}\\
      \hline      
      \parbox[c][][c]{0.5in}{\rotatebox[origin=t]{90}{25\%: 50\% (Log)}} &
      \includegraphics[height=4cm,width=5cm]{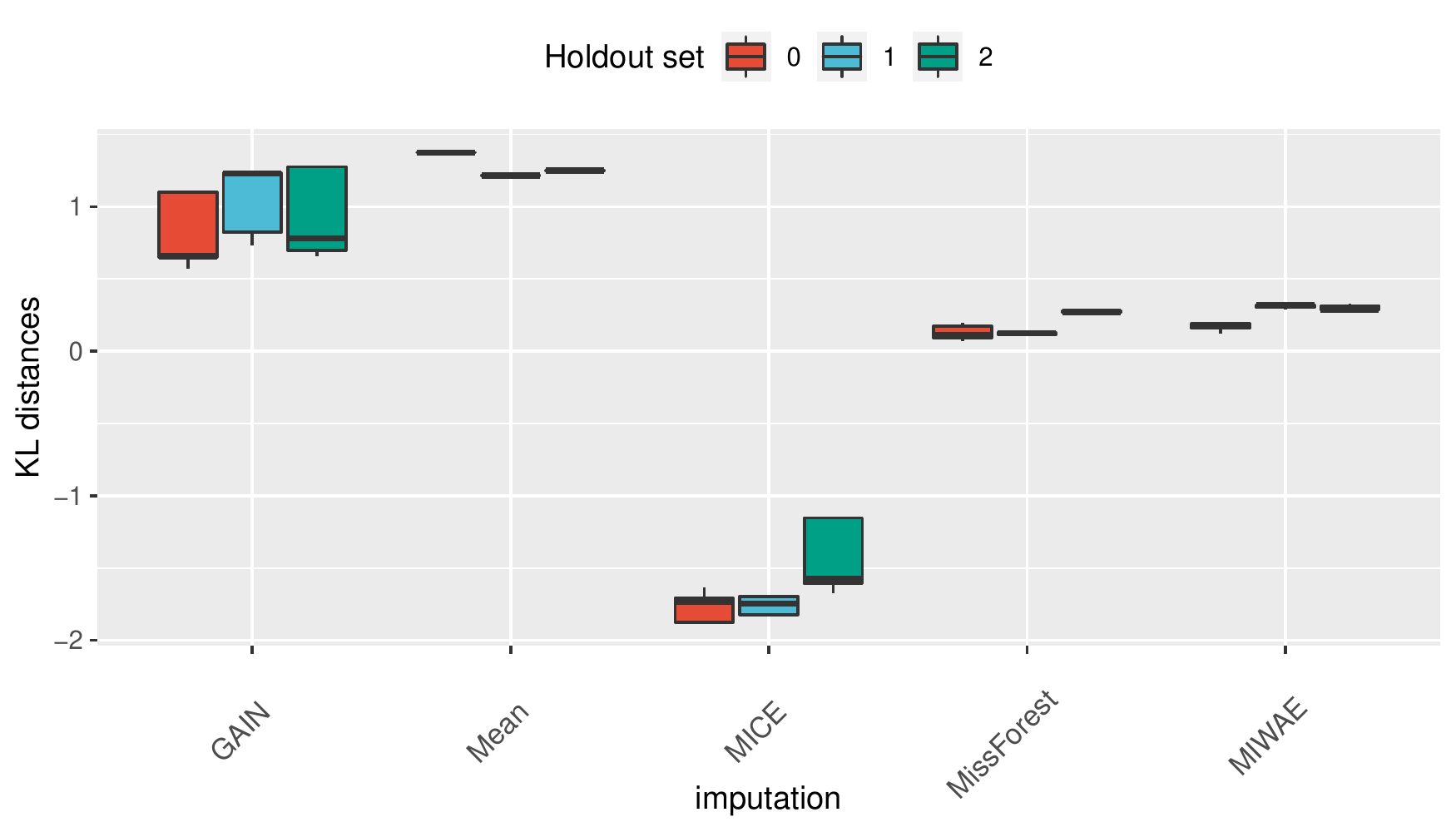}&
      \includegraphics[height=4cm,width=5cm]{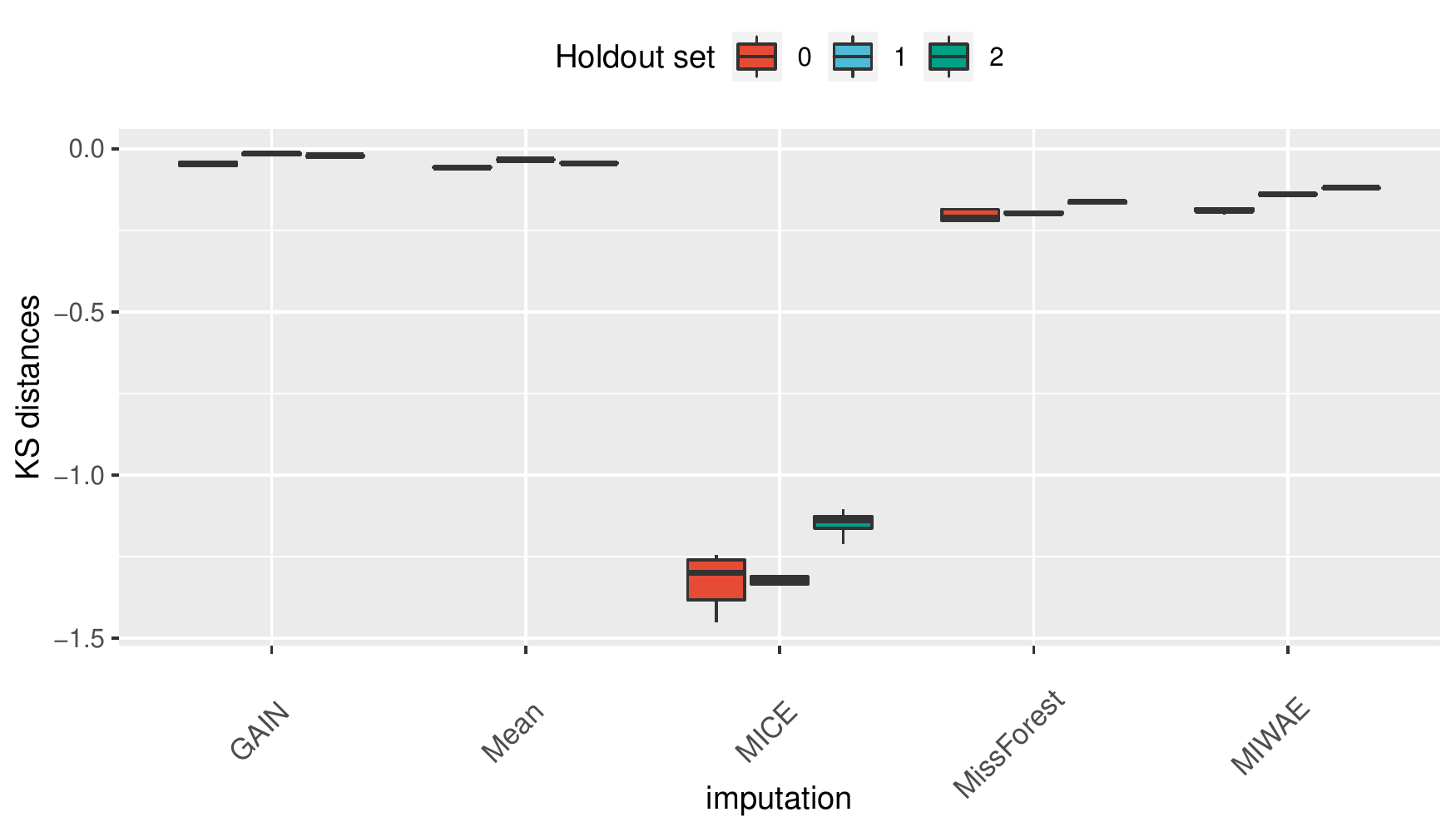}&
      \includegraphics[height=4cm,width=5cm]{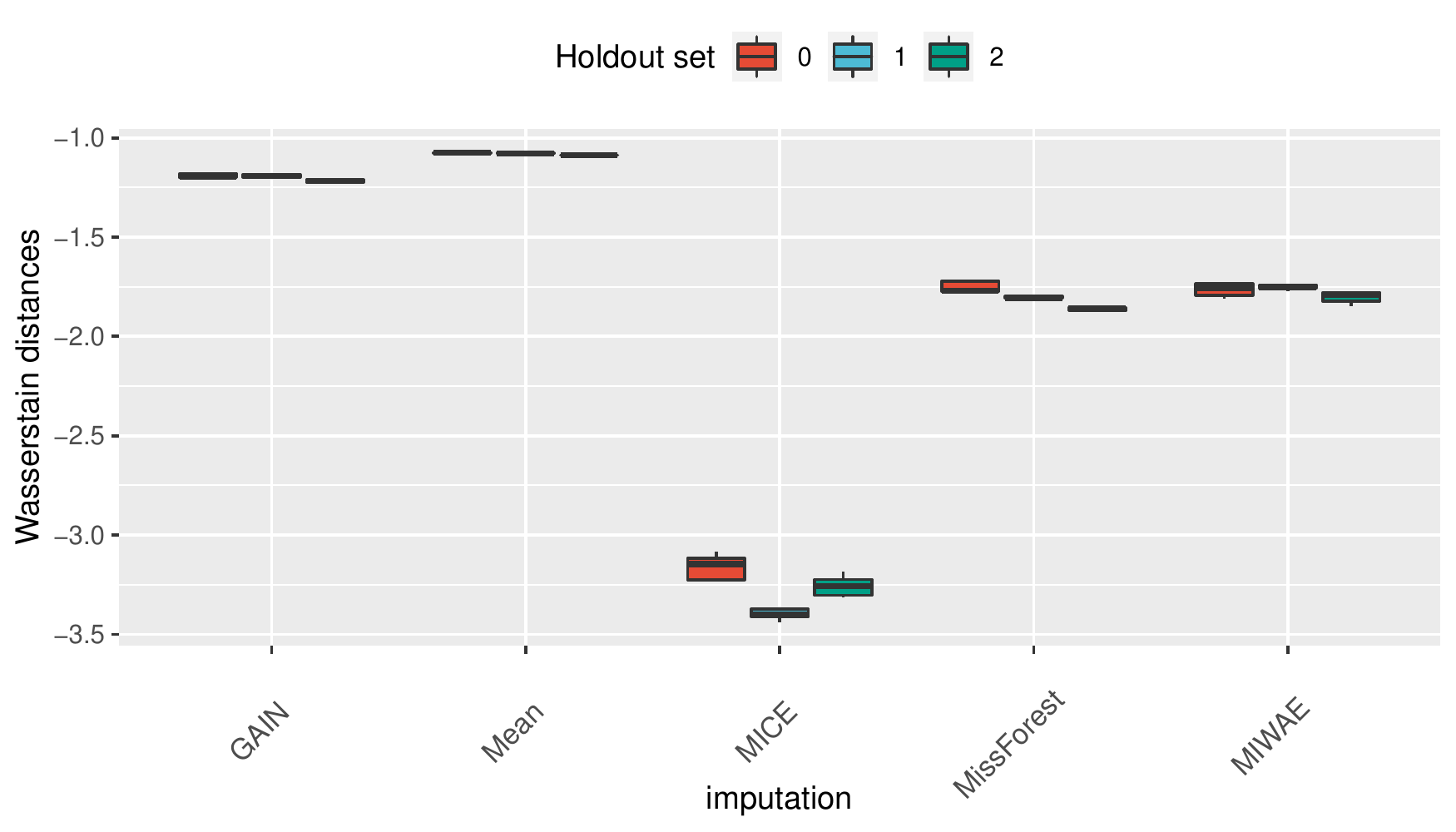}\\
    \end{tabular}
    \caption{The class C discrepancies for the sliced Wasserstein distances of the \textbf{MIMIC-III} data at the 25\% missingness rate for the training data along with 25\% and 50\% for the test data. The original values and logarithms are shown for clarity.}
    \label{fig:datadistwise_supp_25}
\end{figure}

\begin{figure}[htb!]
    \centering
    \begin{tabular}{m{0.2in} |M{5cm} | M{5cm} | M{5cm}}
     & \textbf{Kullback-Leibler} & \textbf{Kolmogorov-Smirnoff} & \textbf{Wasserstein} \\
     \hline
     \parbox[c][][c]{0.5in}{\rotatebox[origin=t]{90}{50\%: 25\%}} &
      \includegraphics[height=4cm,width=5cm]{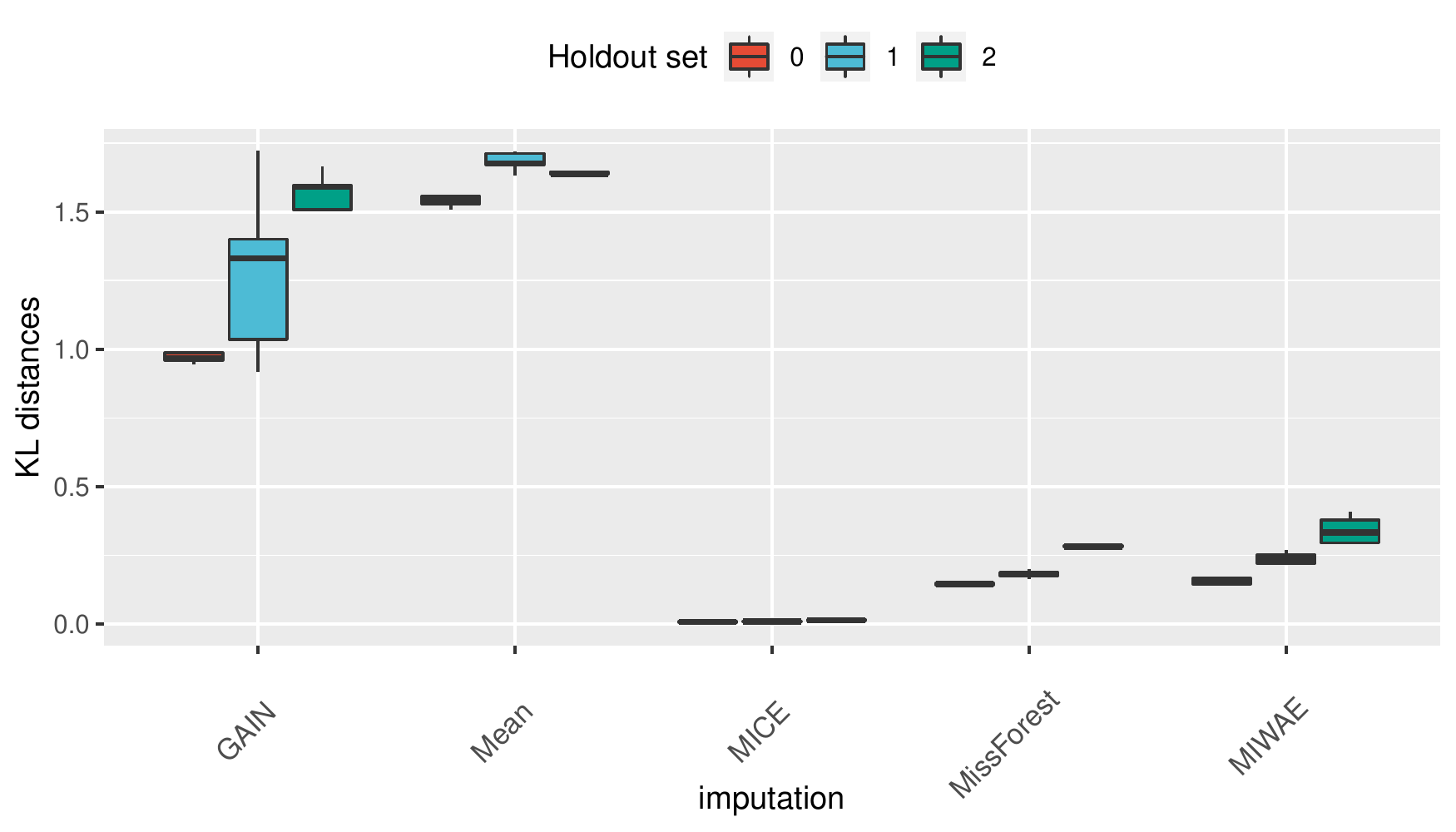}&
      \includegraphics[height=4cm,width=5cm]{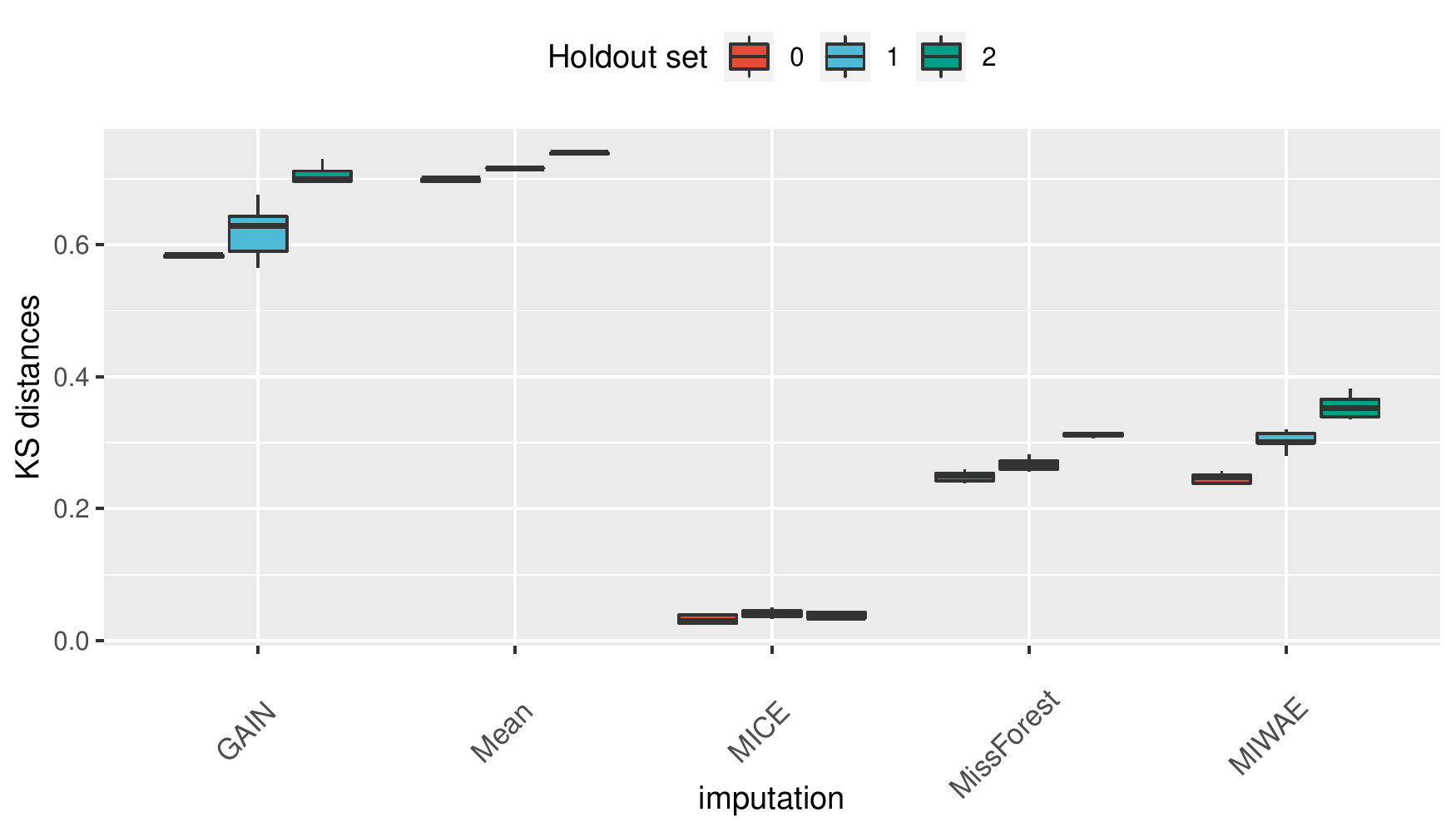}&
      \includegraphics[height=4cm,width=5cm]{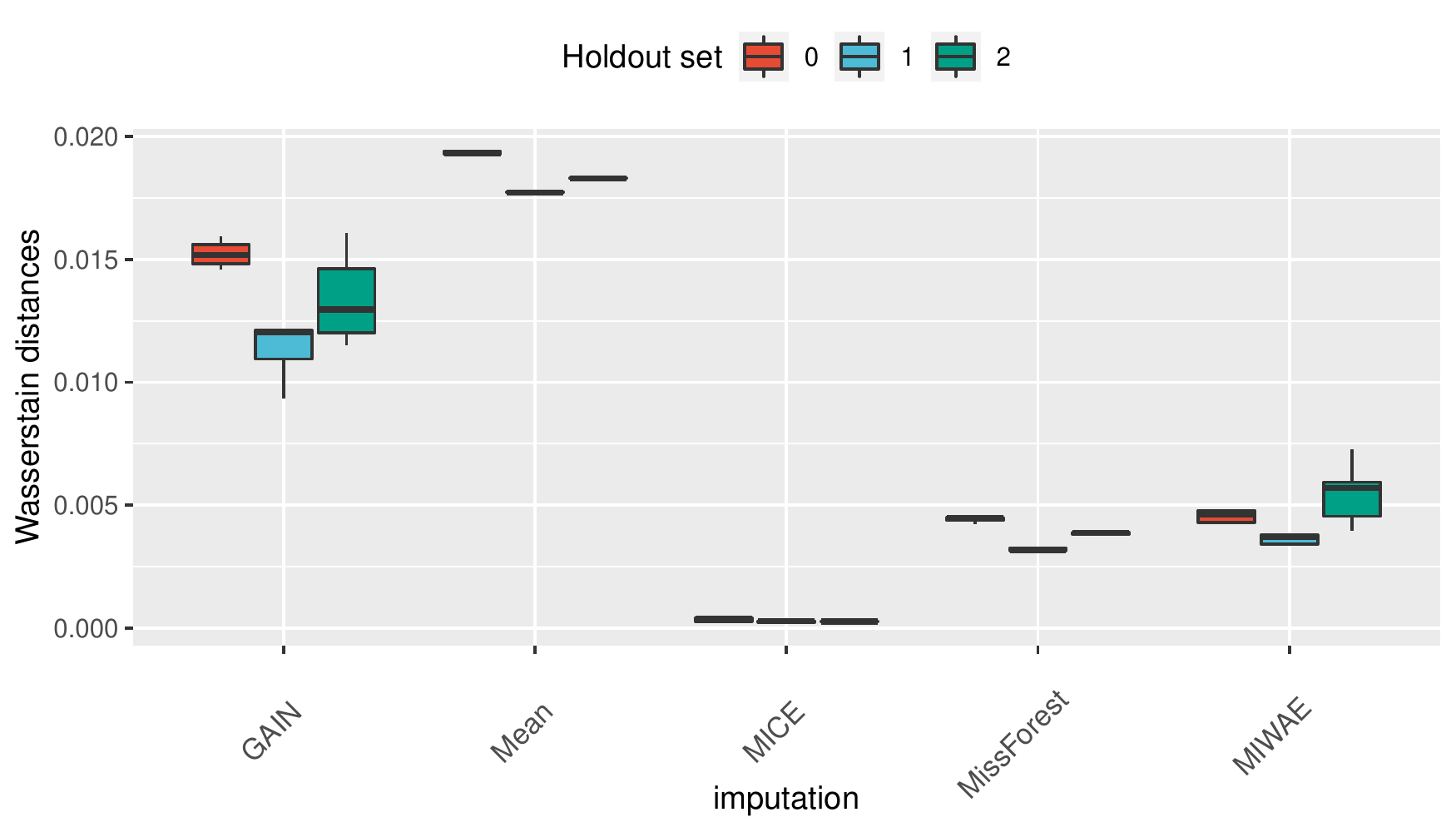}\\
      \hline      
      \parbox[c][][c]{0.5in}{\rotatebox[origin=t]{90}{50\%: 25\% (Log)}} & \includegraphics[height=4cm,width=5cm]{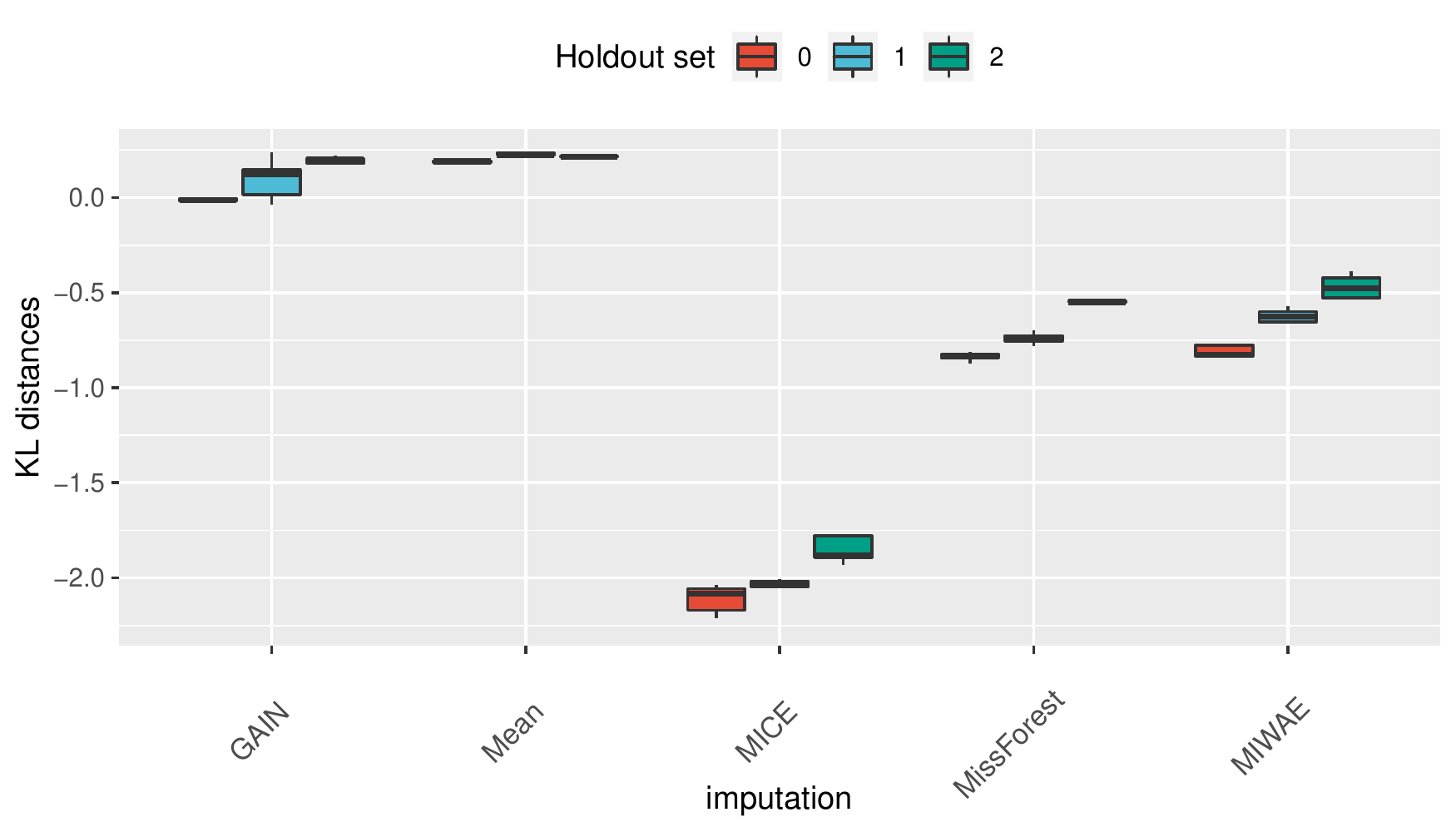}&
      \includegraphics[height=4cm,width=5cm]{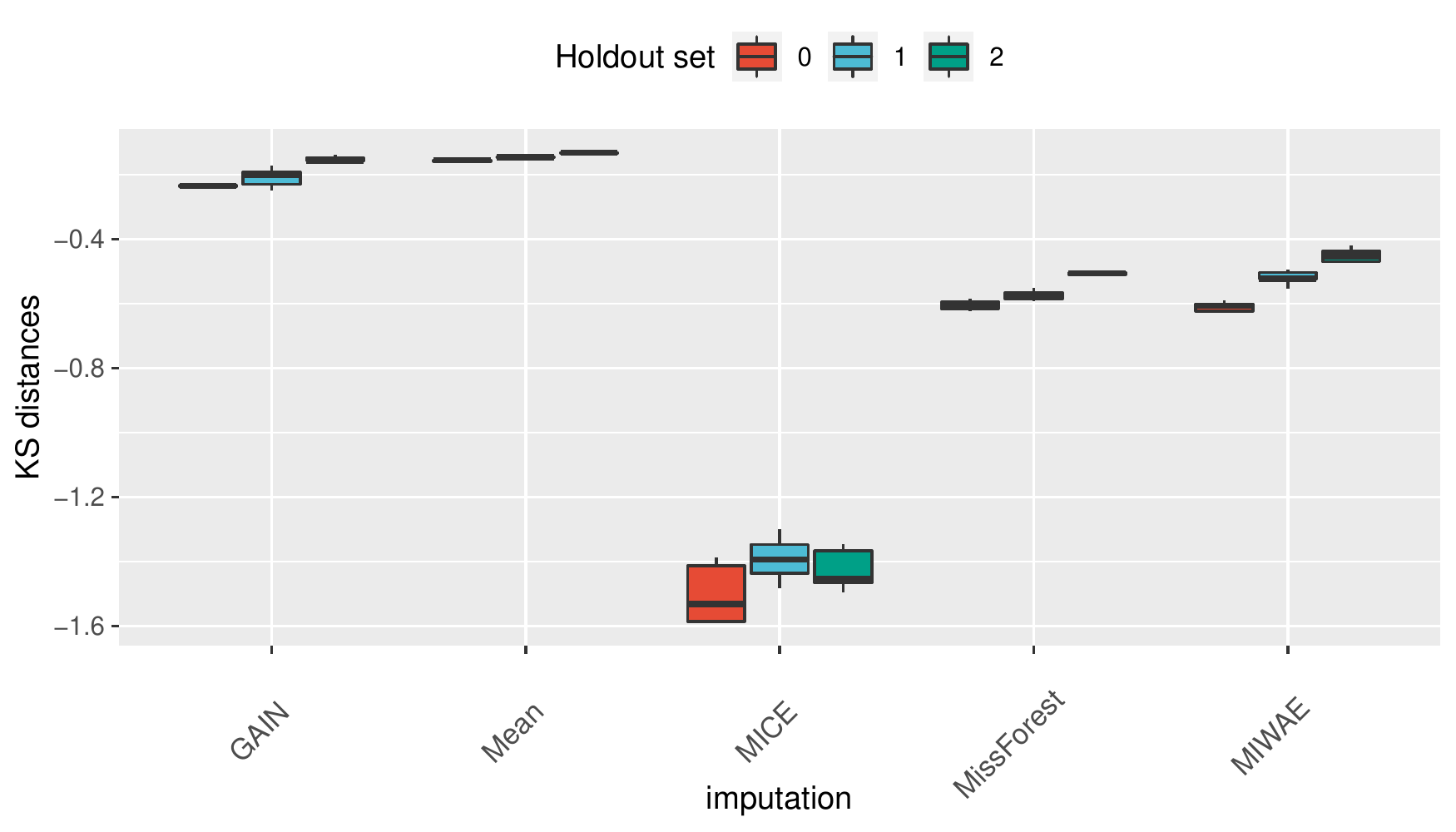}&
      \includegraphics[height=4cm,width=5cm]{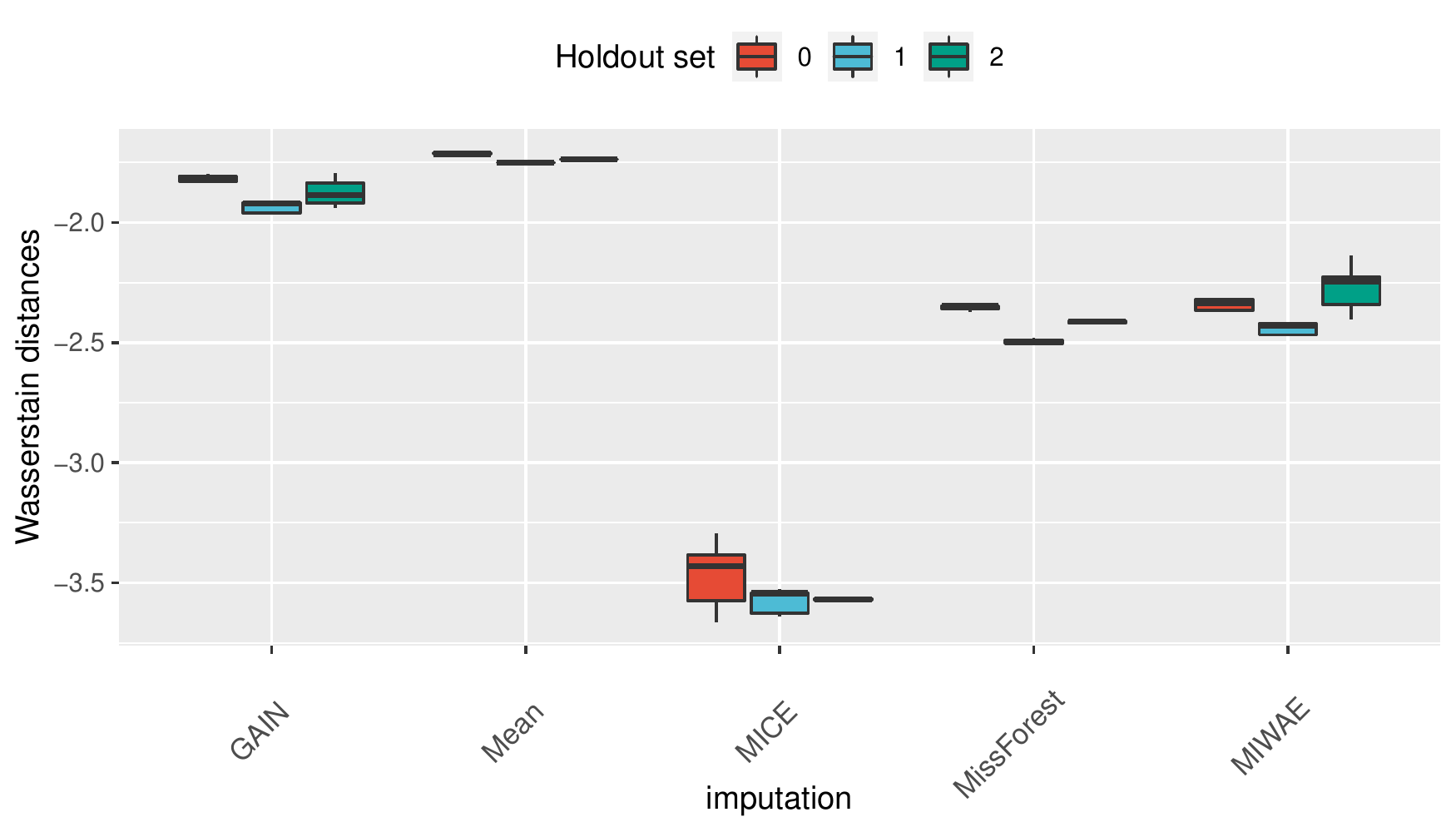}\\
     \hline
     \parbox[c][][c]{0.5in}{\rotatebox[origin=t]{90}{50\%: 50\%}} &
      \includegraphics[height=4cm,width=5cm]{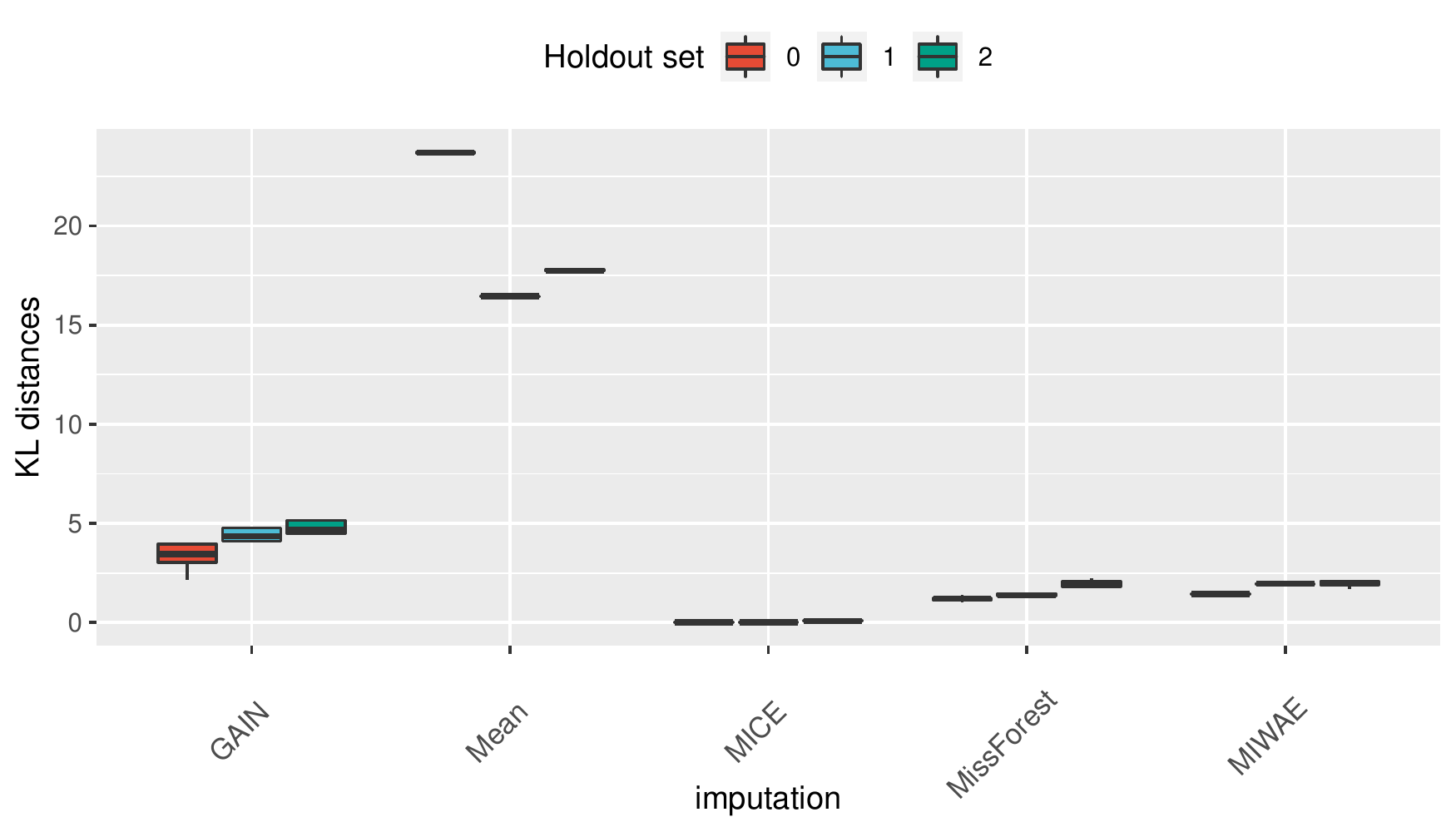}&
      \includegraphics[height=4cm,width=5cm]{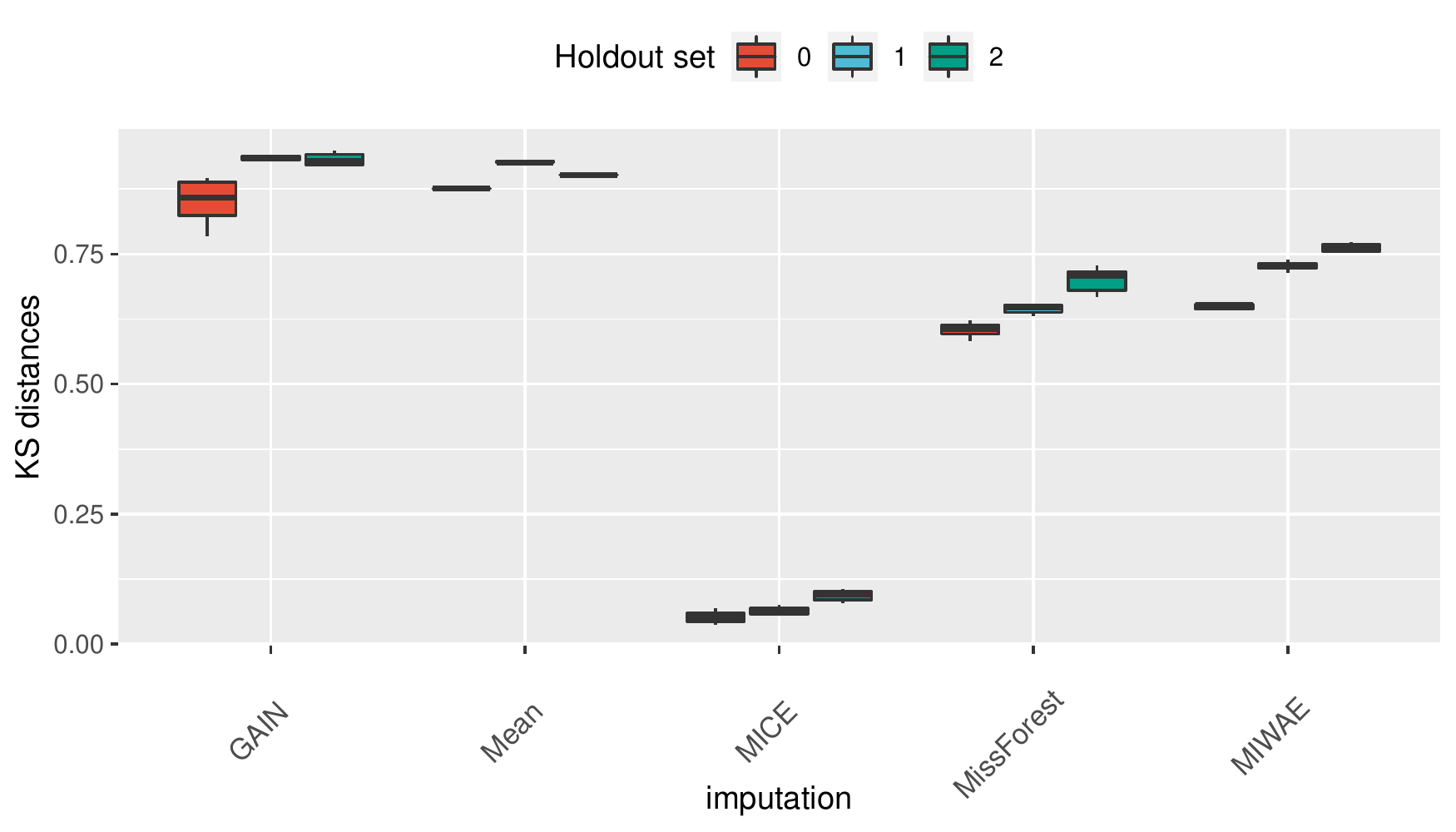}&
      \includegraphics[height=4cm,width=5cm]{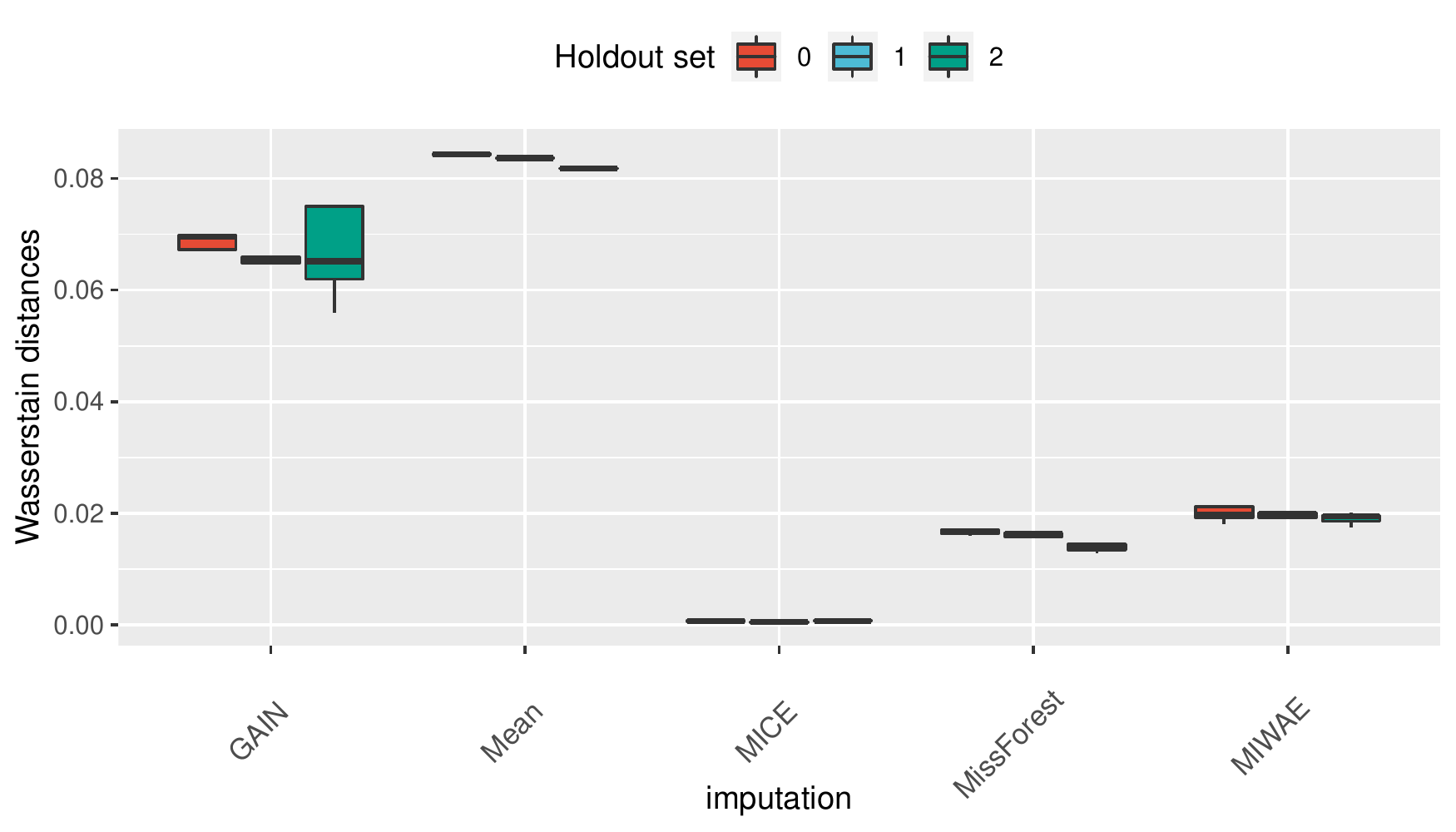}\\
      \hline      
      \parbox[c][][c]{0.5in}{\rotatebox[origin=t]{90}{50\%: 50\% (Log)}} &
      \includegraphics[height=4cm,width=5cm]{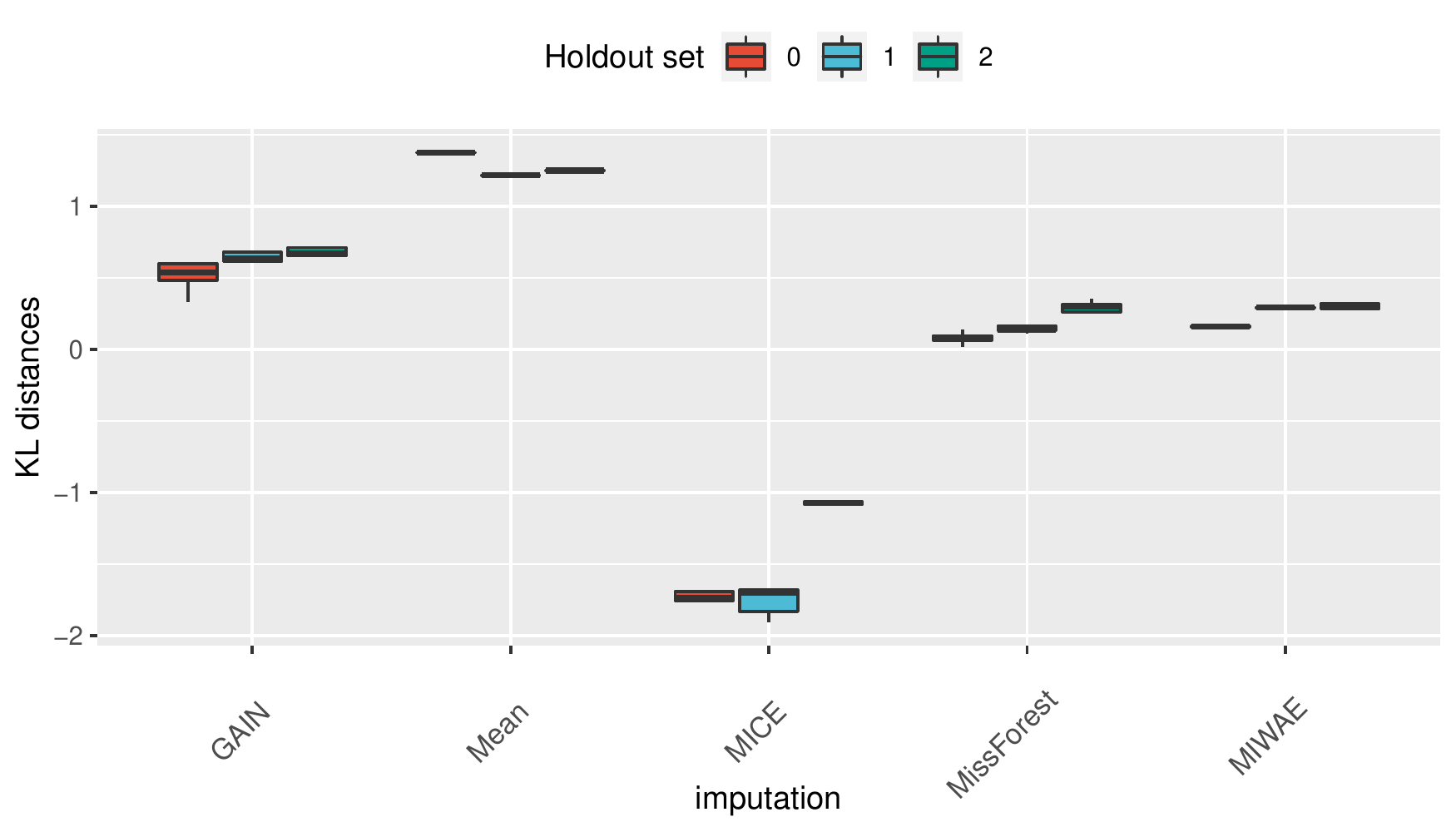}&
      \includegraphics[height=4cm,width=5cm]{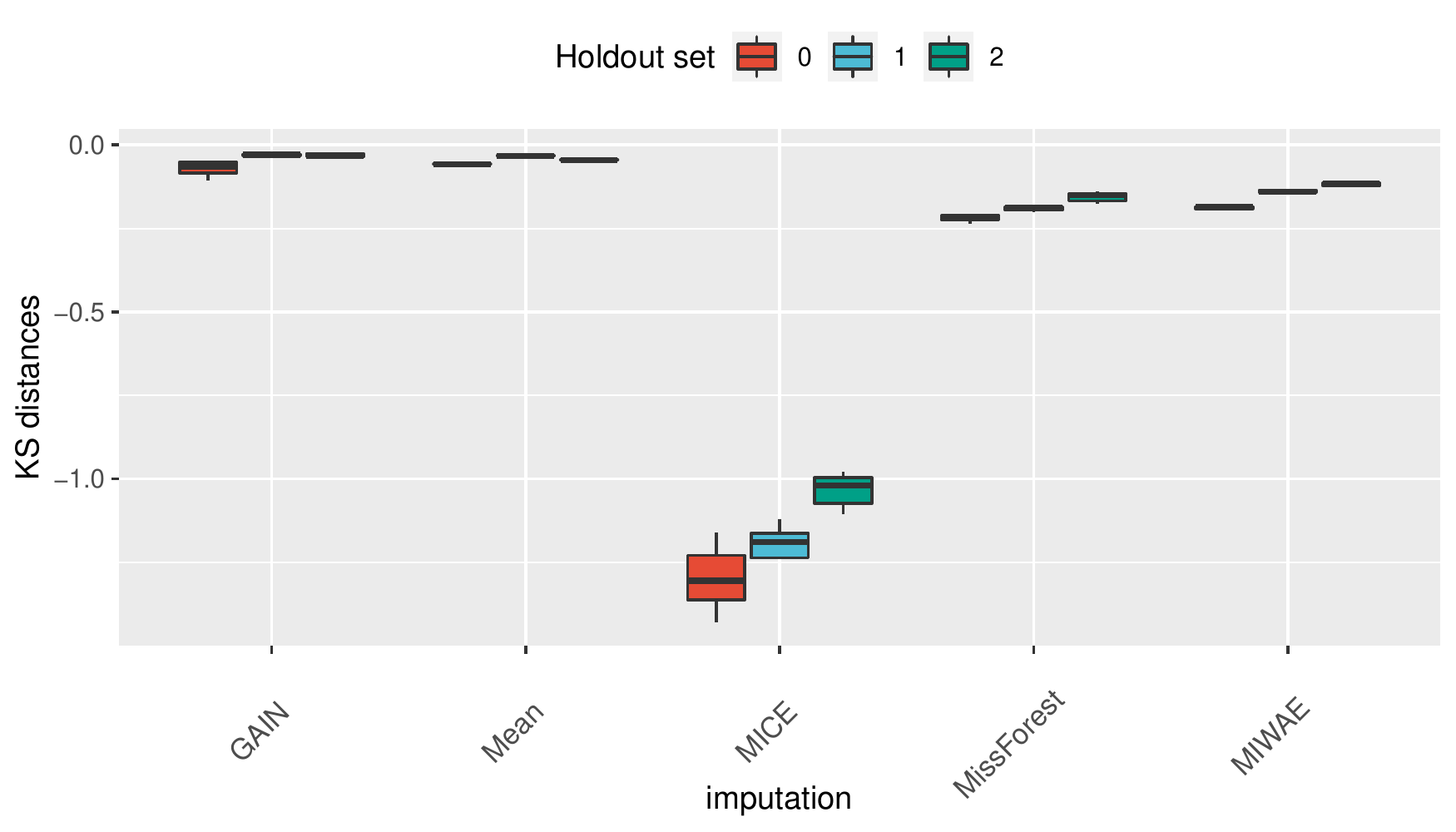}&
      \includegraphics[height=4cm,width=5cm]{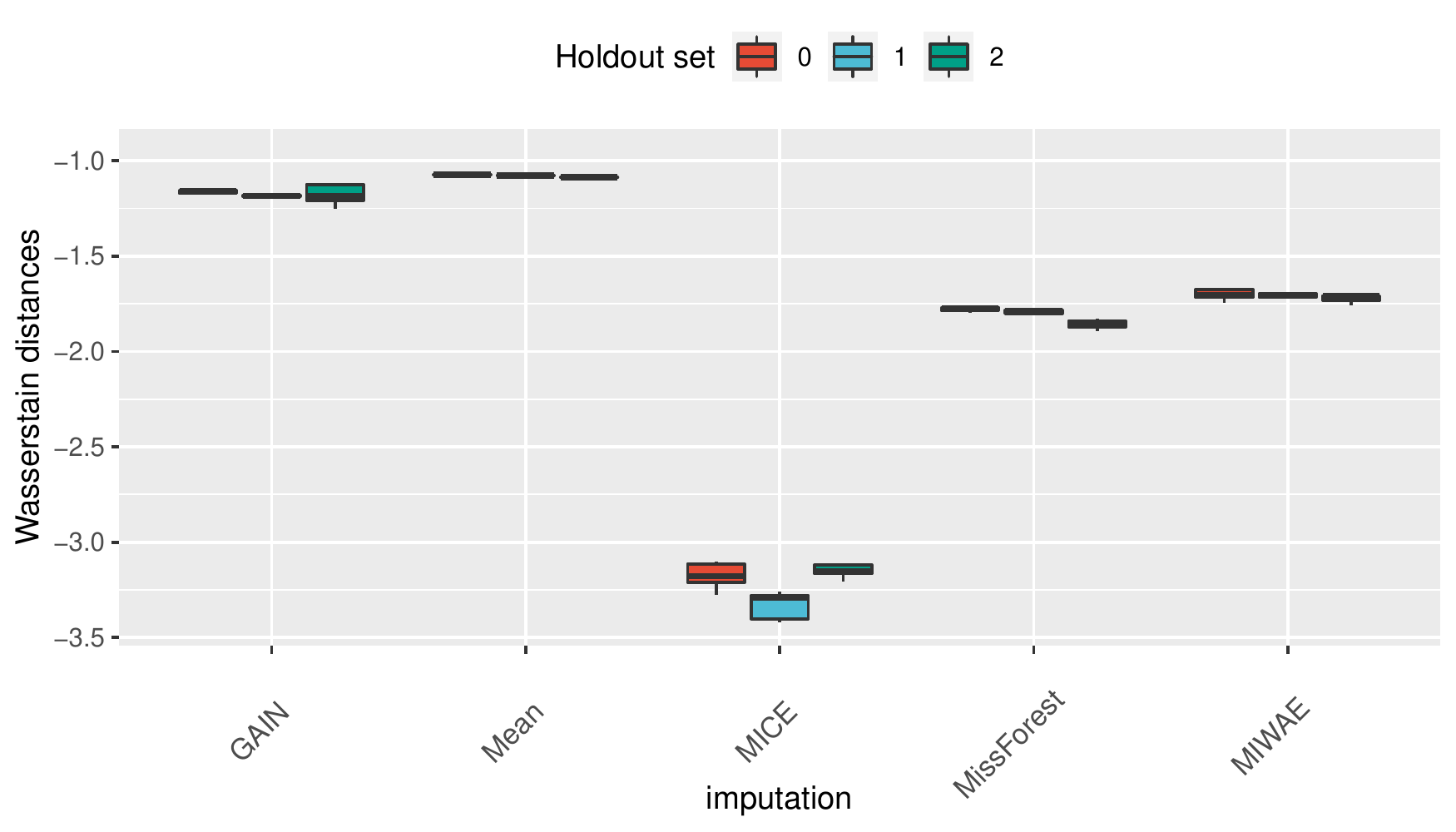}\\
    \end{tabular}
    \caption{The class C discrepancies for the sliced Wasserstein distances of the \textbf{MIMIC-III} data at the 50\% missingness rate for the training data along with 25\% and 50\% for the test data. The original values and logarithms are shown for clarity.}
    \label{fig:datadistwise_supp_50}
\end{figure}

\begin{figure}[htb!]
    \centering
    \begin{tabular}{m{0.2in} |M{5cm} | M{5cm} | M{5cm}}
     & \textbf{Kullback-Leibler} & \textbf{Kolmogorov-Smirnoff} & \textbf{Wasserstein} \\
     \hline
     \parbox[c][][c]{0.5in}{\rotatebox[origin=t]{90}{25\%: 25\%}} &
      \includegraphics[height=4cm,width=5cm]{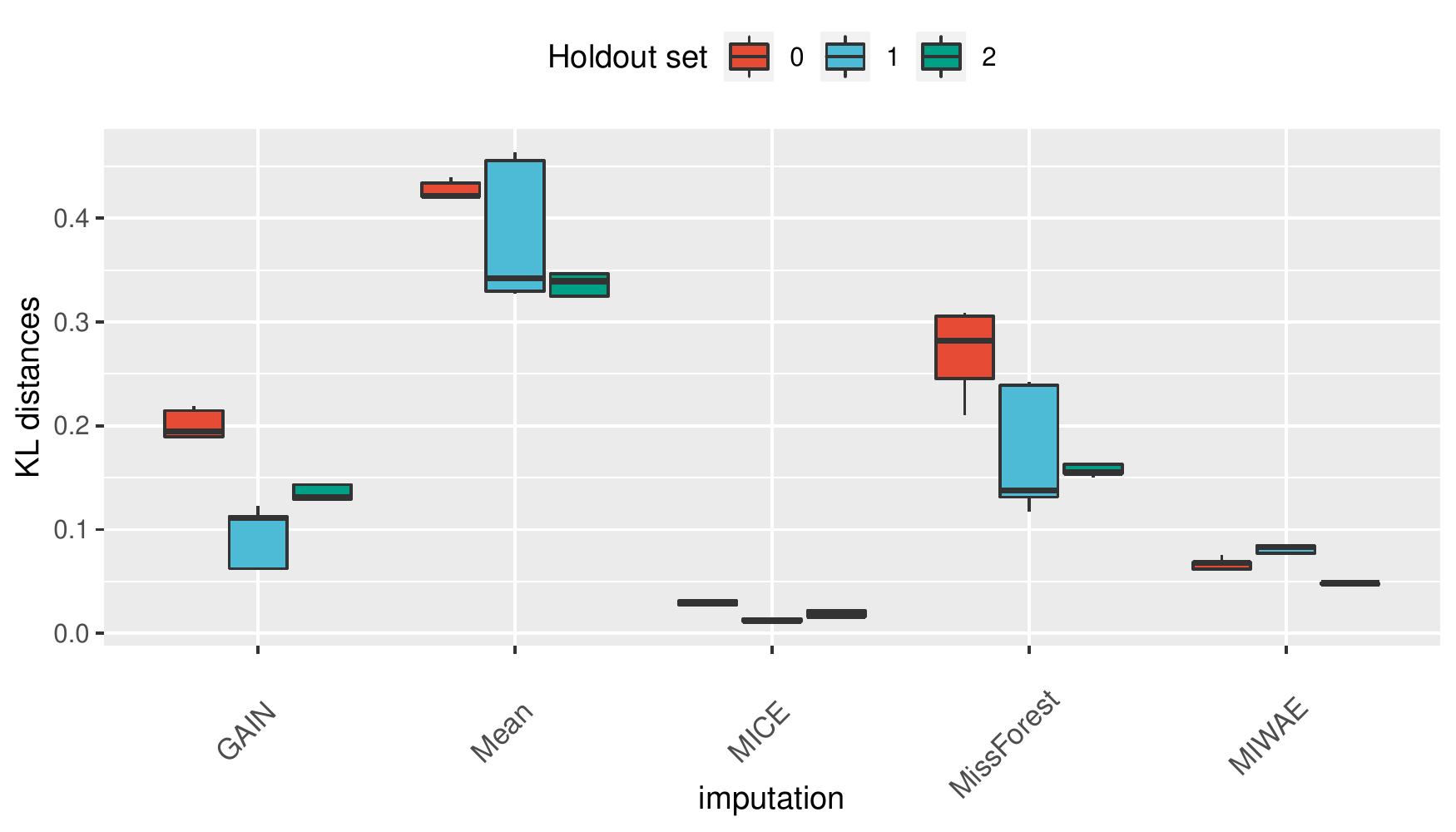}&
      \includegraphics[height=4cm,width=5cm]{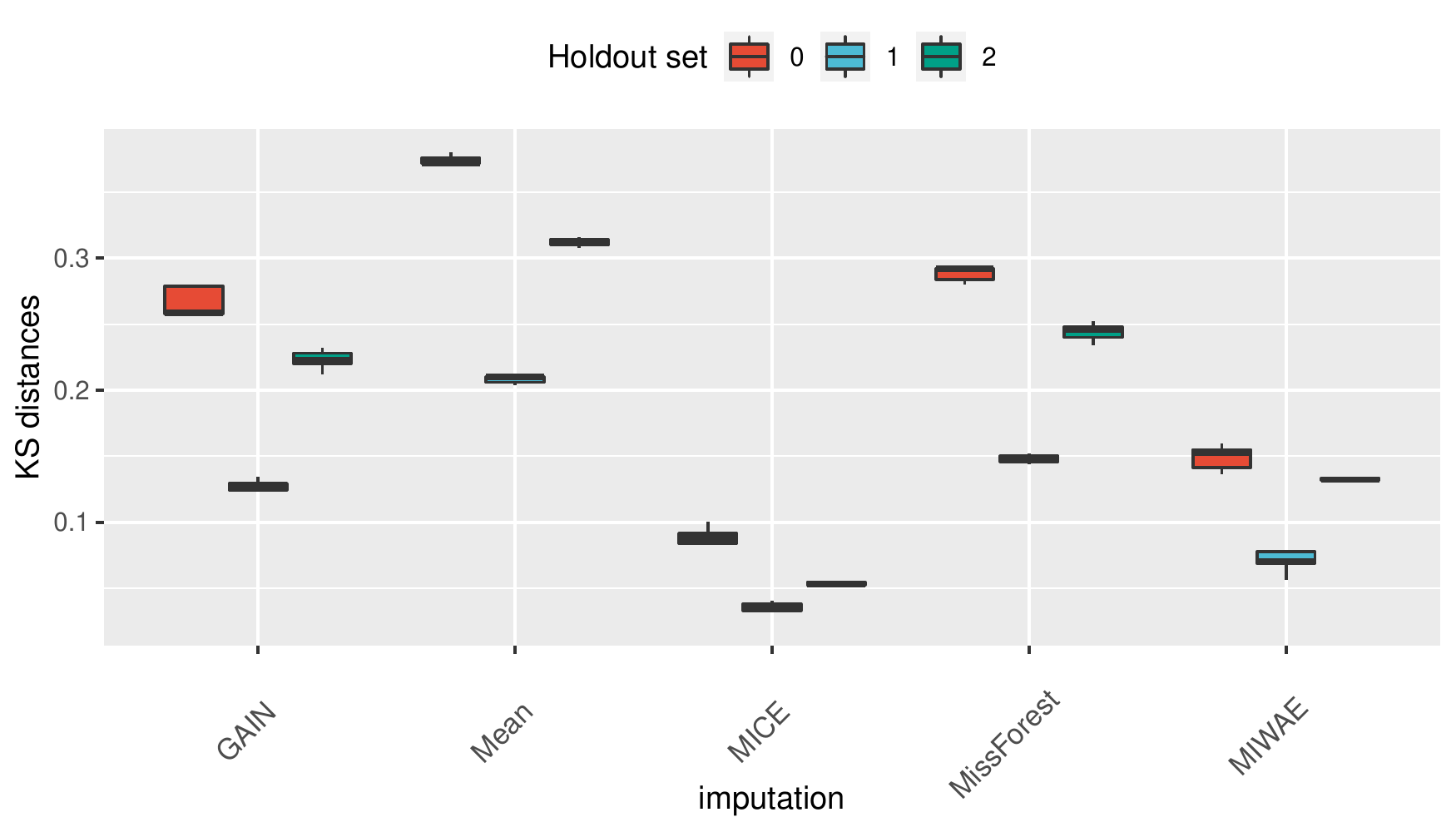}&
      \includegraphics[height=4cm,width=5cm]{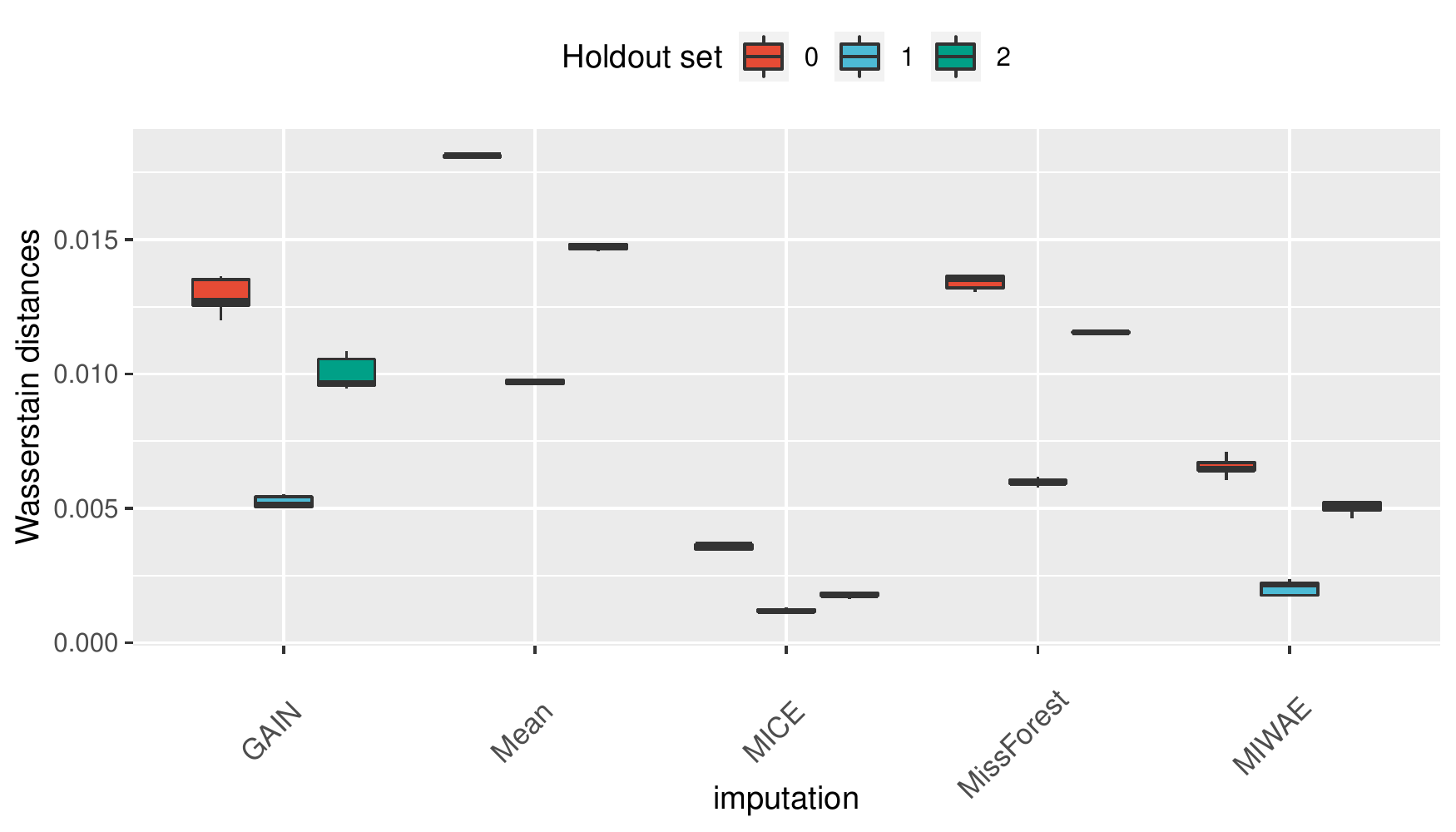}\\
      \hline      
      \parbox[c][][c]{0.5in}{\rotatebox[origin=t]{90}{25\%: 25\% (Log)}} &\includegraphics[height=4cm,width=5cm]{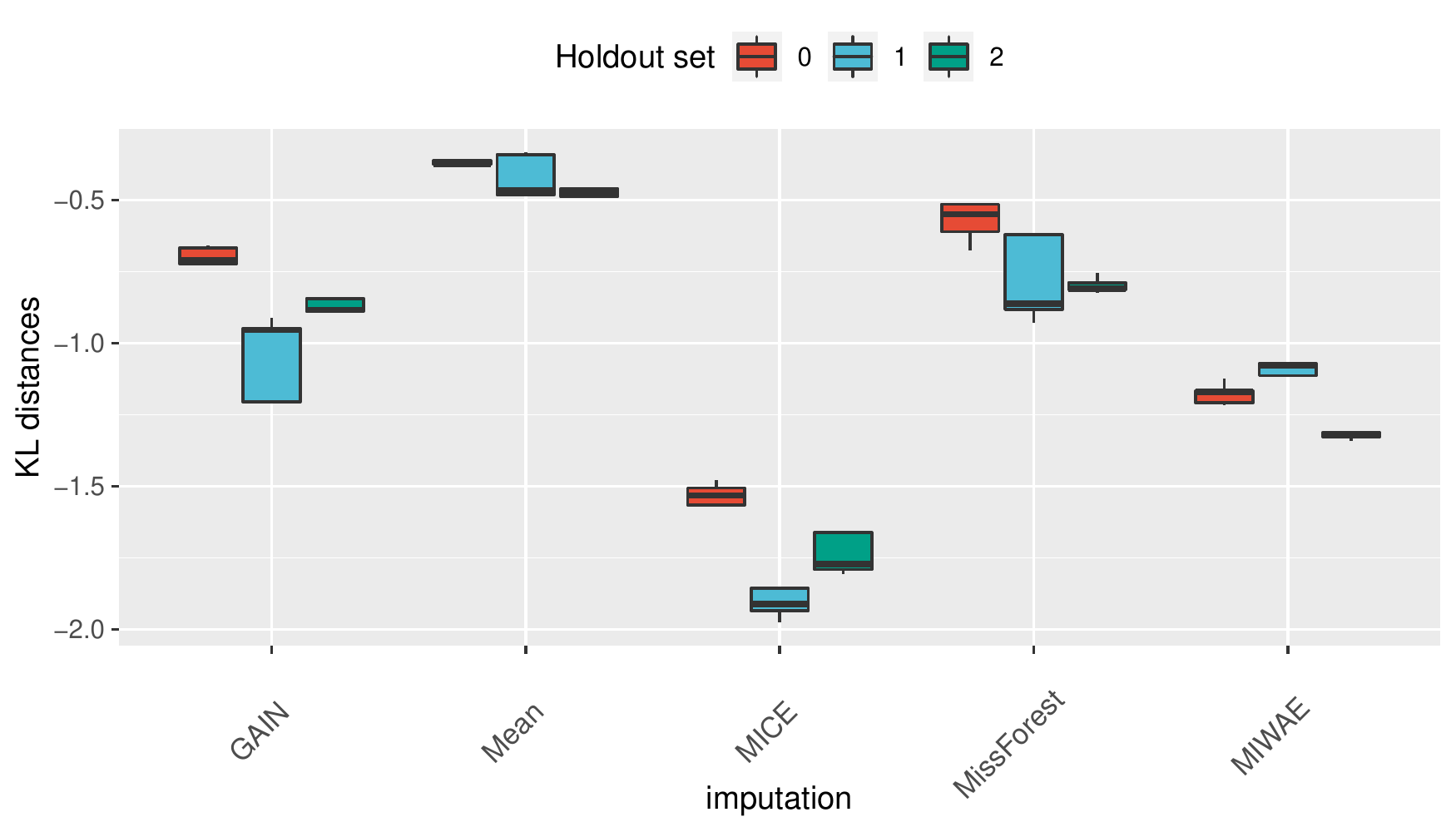}&
      \includegraphics[height=4cm,width=5cm]{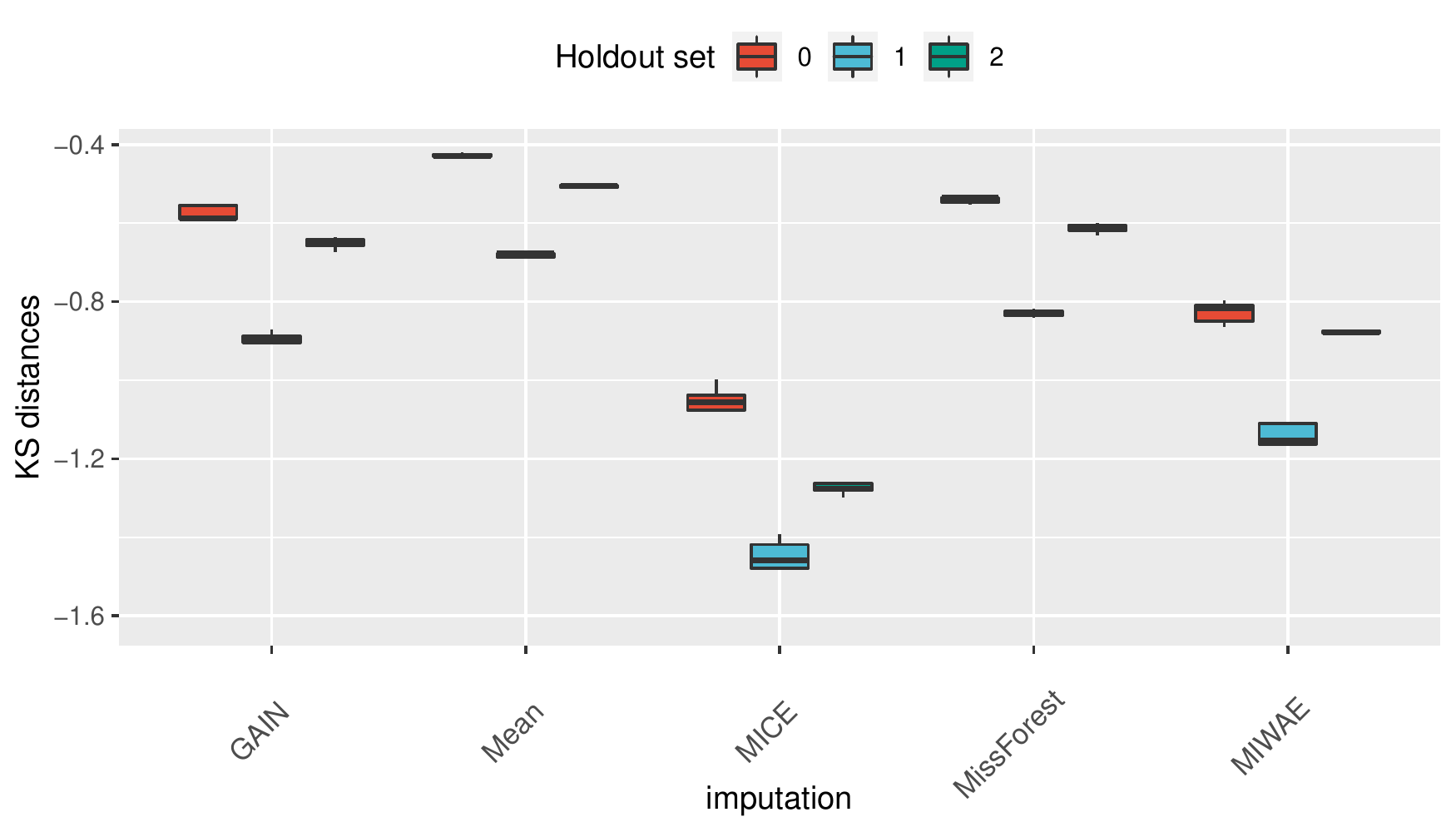}&
      \includegraphics[height=4cm,width=5cm]{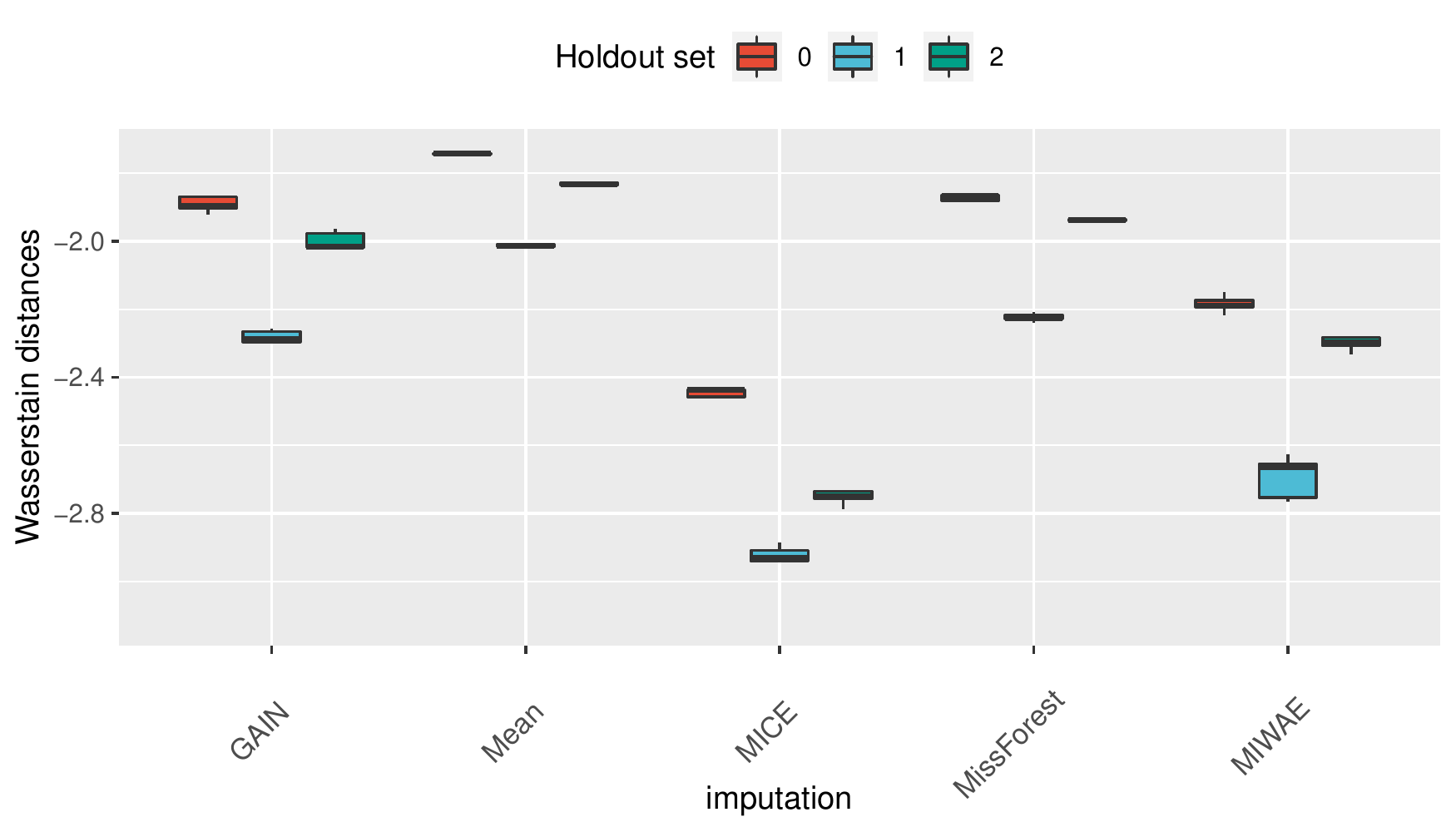}\\
     \hline
     \parbox[c][][c]{0.5in}{\rotatebox[origin=t]{90}{25\%: 50\%}} &
      \includegraphics[height=4cm,width=5cm]{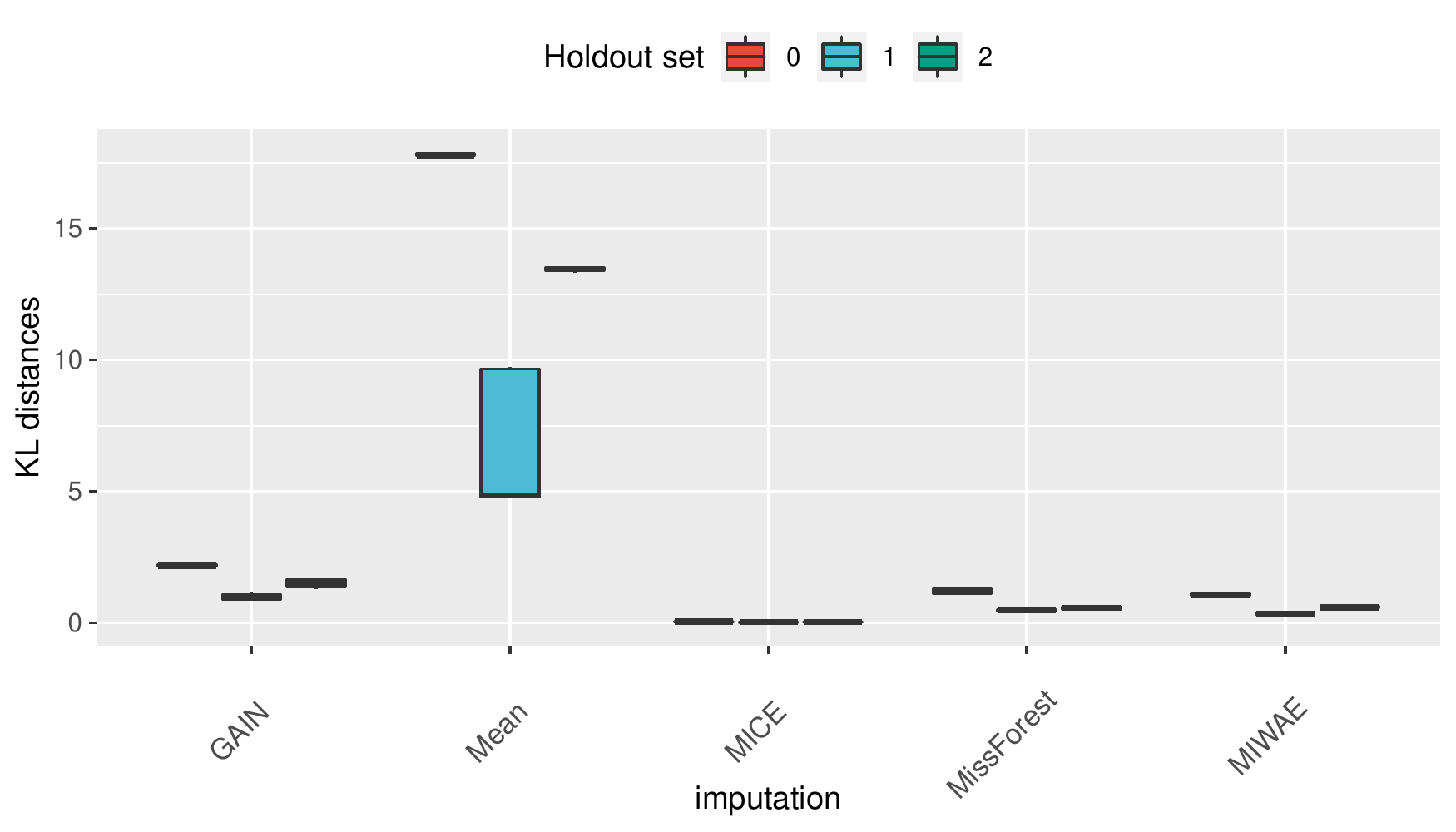}&
      \includegraphics[height=4cm,width=5cm]{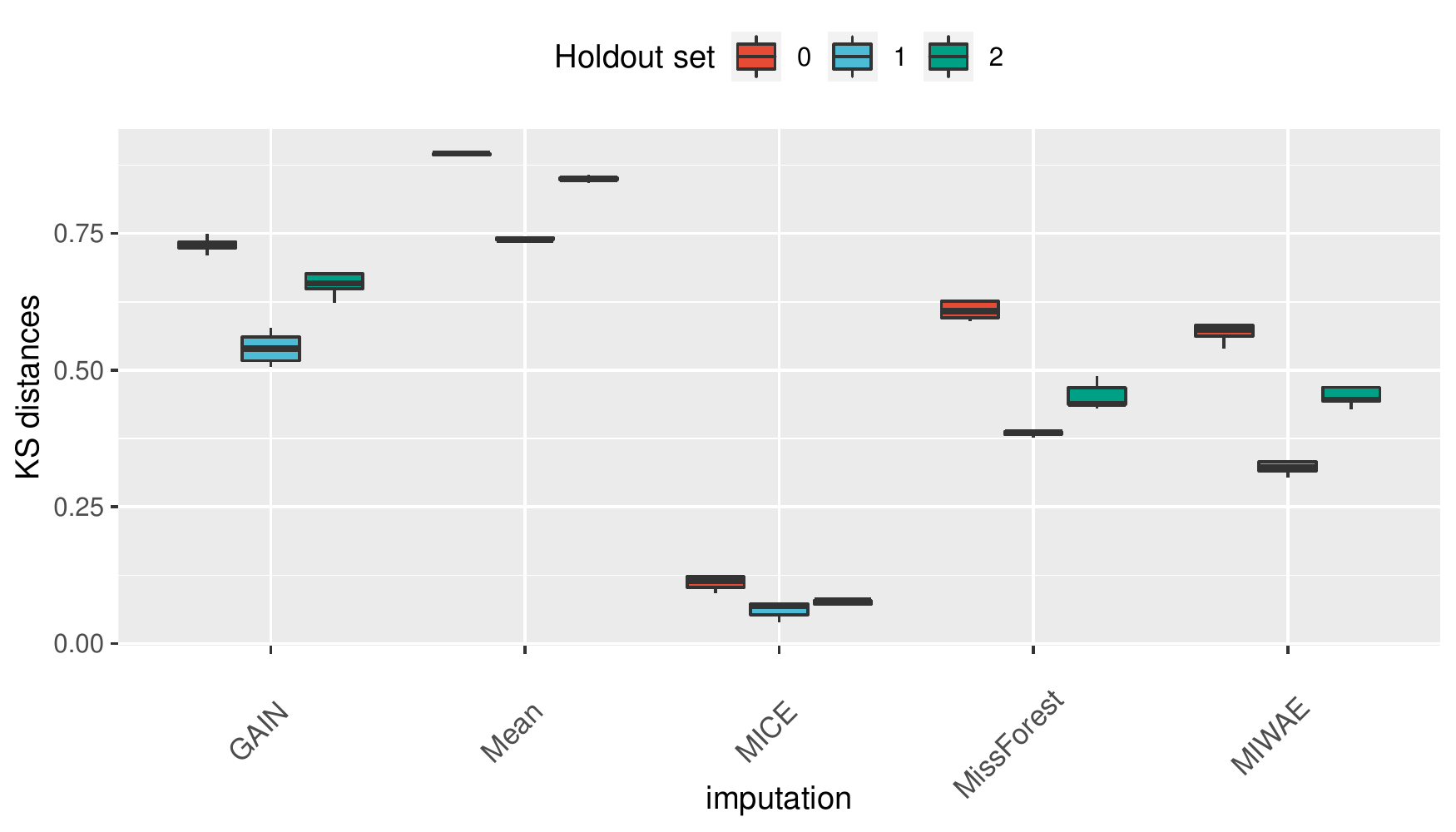}&
      \includegraphics[height=4cm,width=5cm]{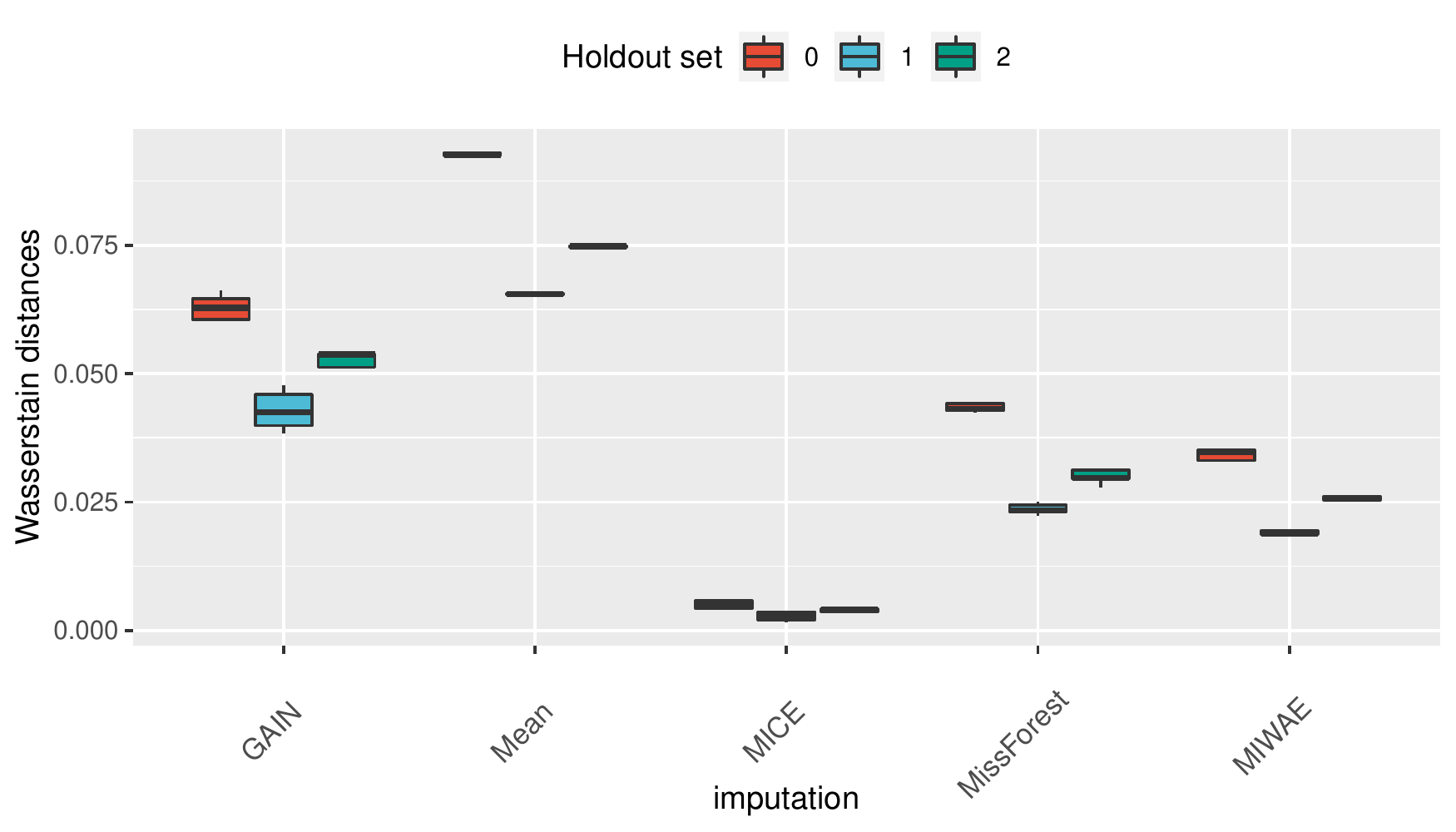}\\
      \hline      
      \parbox[c][][c]{0.5in}{\rotatebox[origin=t]{90}{25\%: 50\% (Log)}} &
      \includegraphics[height=4cm,width=5cm]{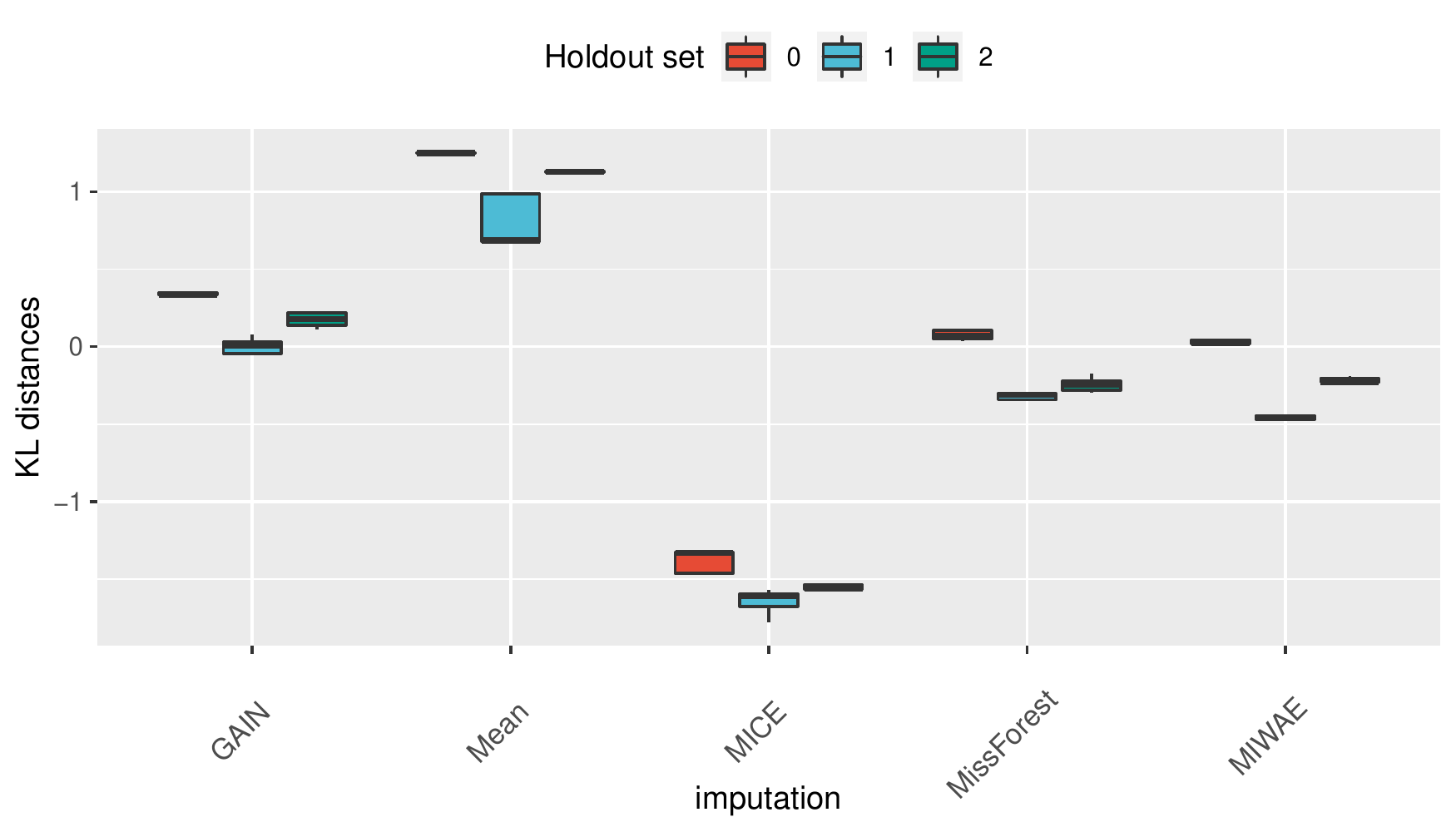}&
      \includegraphics[height=4cm,width=5cm]{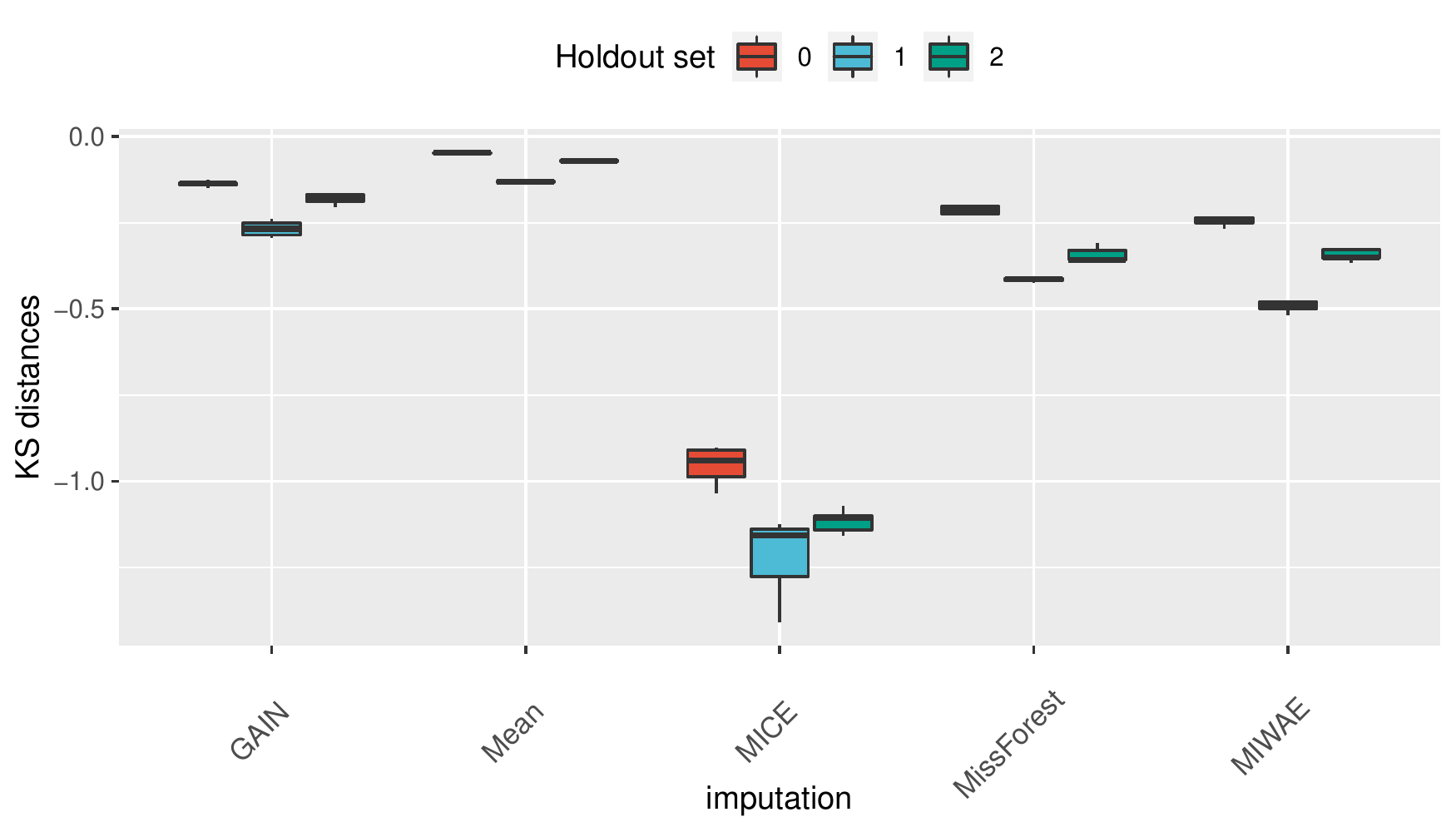}&
      \includegraphics[height=4cm,width=5cm]{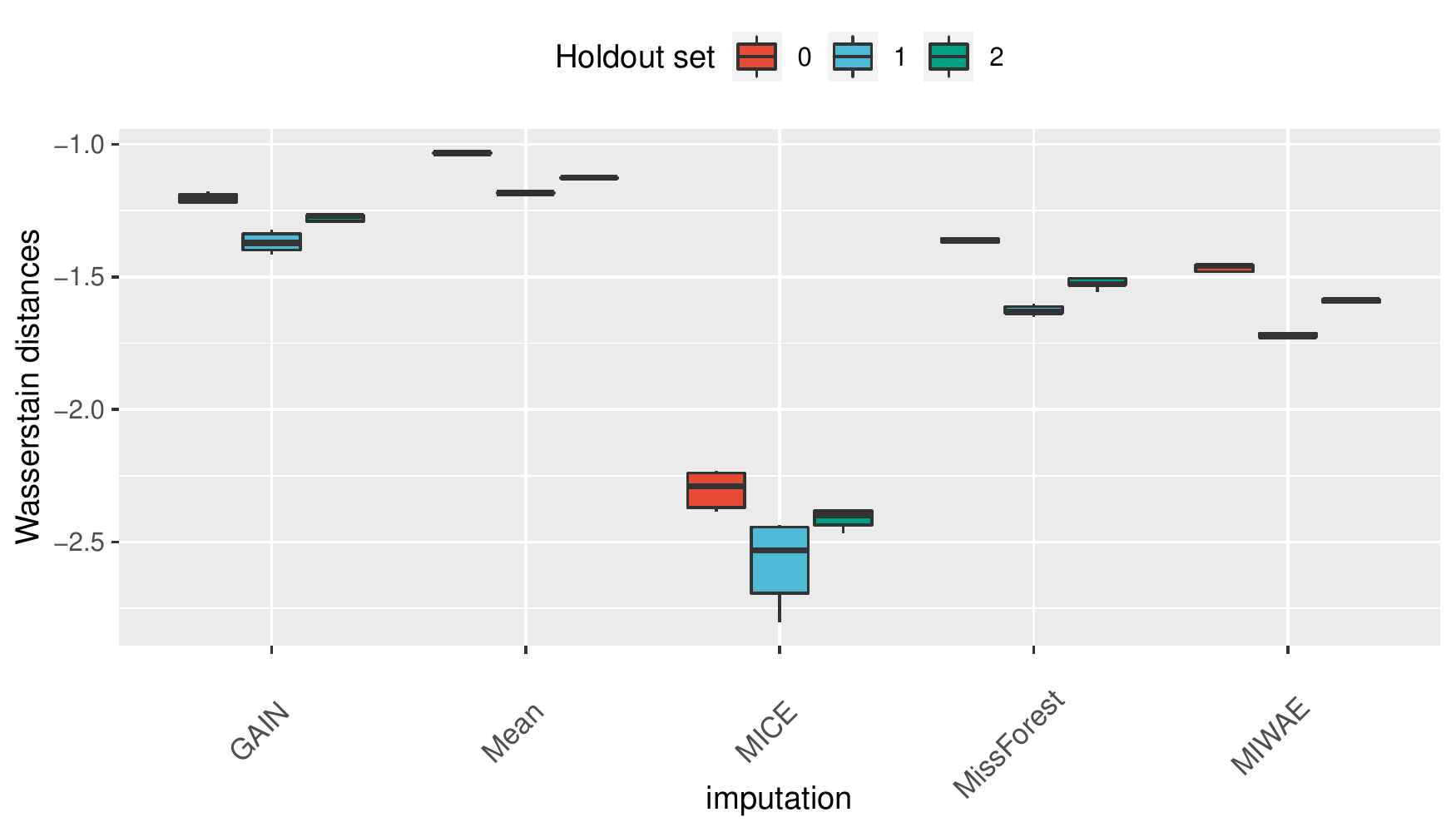}\\
    \end{tabular}
    \caption{The class C discrepancies for the sliced Wasserstein distances of the Simulated data at the 25\% missingness rate for the training data along with 25\% and 50\% for the test data. The original values and logarithms are shown for clarity.}
    \label{fig:datadistwise_supp_25_syn}
\end{figure}

\begin{figure}[htb!]
    \centering
    \begin{tabular}{m{0.2in} |M{5cm} | M{5cm} | M{5cm}}
     & \textbf{Kullback-Leibler} & \textbf{Kolmogorov-Smirnoff} & \textbf{Wasserstein} \\
     \hline
     \parbox[c][][c]{0.5in}{\rotatebox[origin=t]{90}{50\%: 25\%}} &
      \includegraphics[height=4cm,width=5cm]{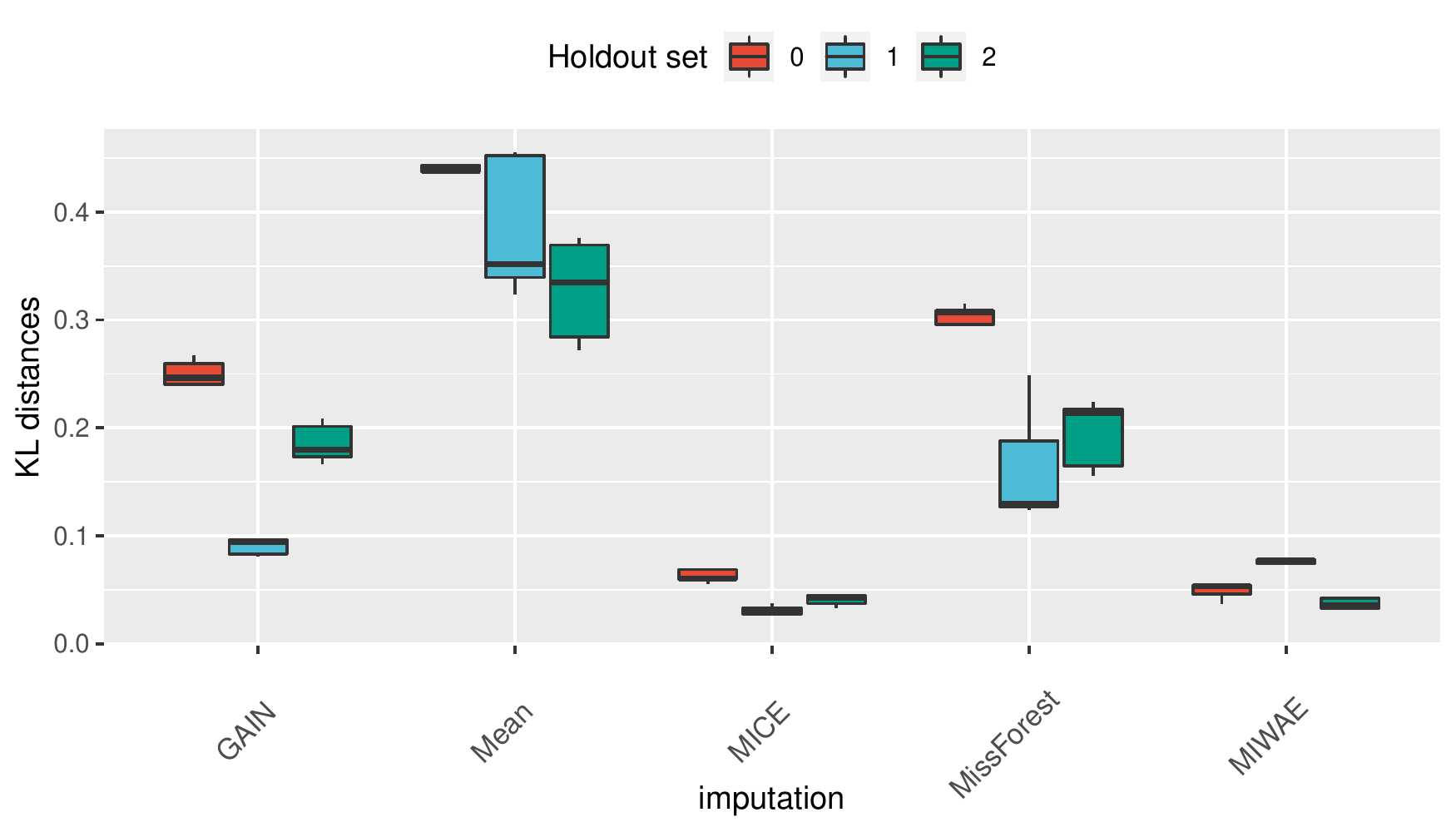}&
      \includegraphics[height=4cm,width=5cm]{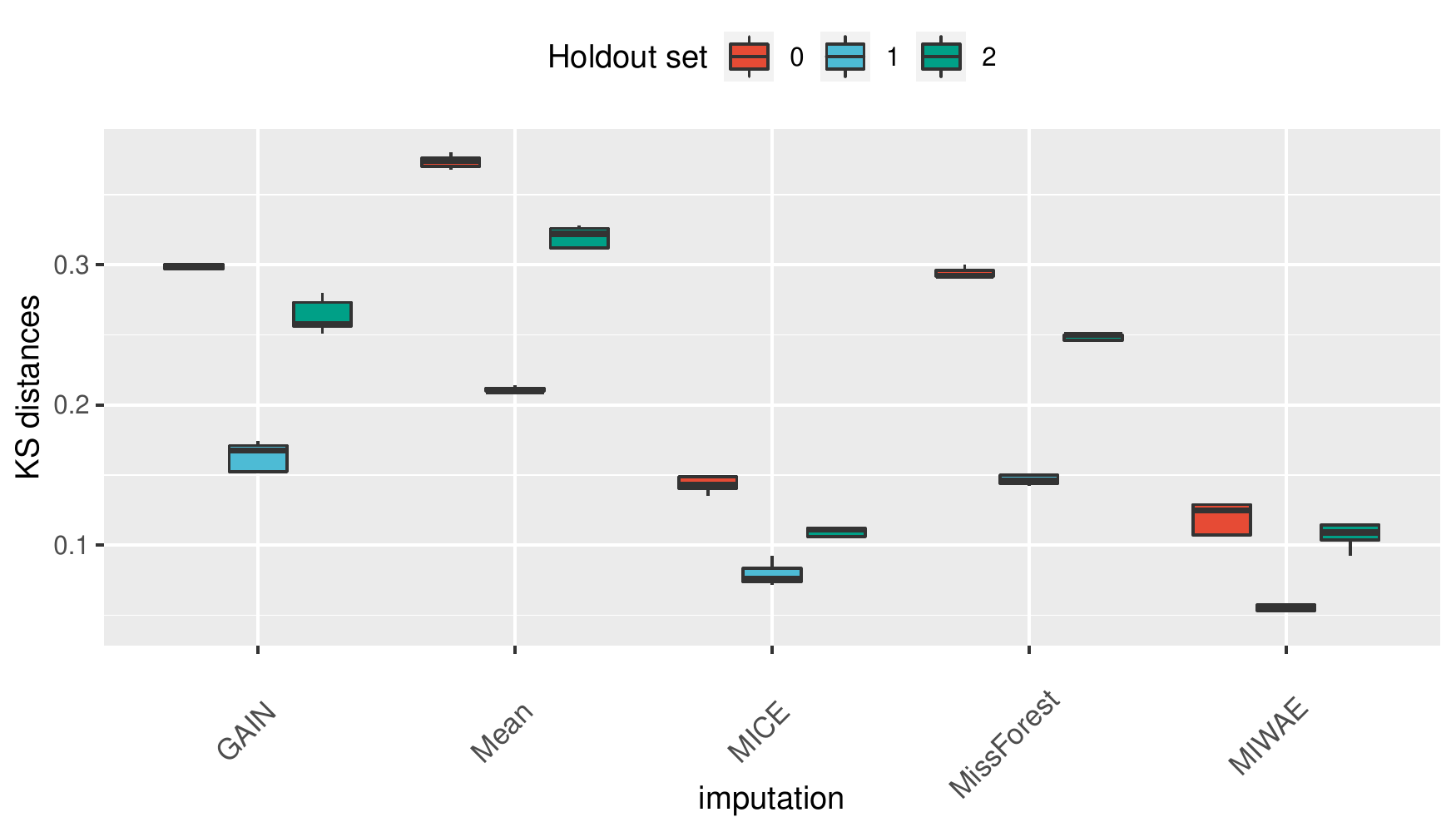}&
      \includegraphics[height=4cm,width=5cm]{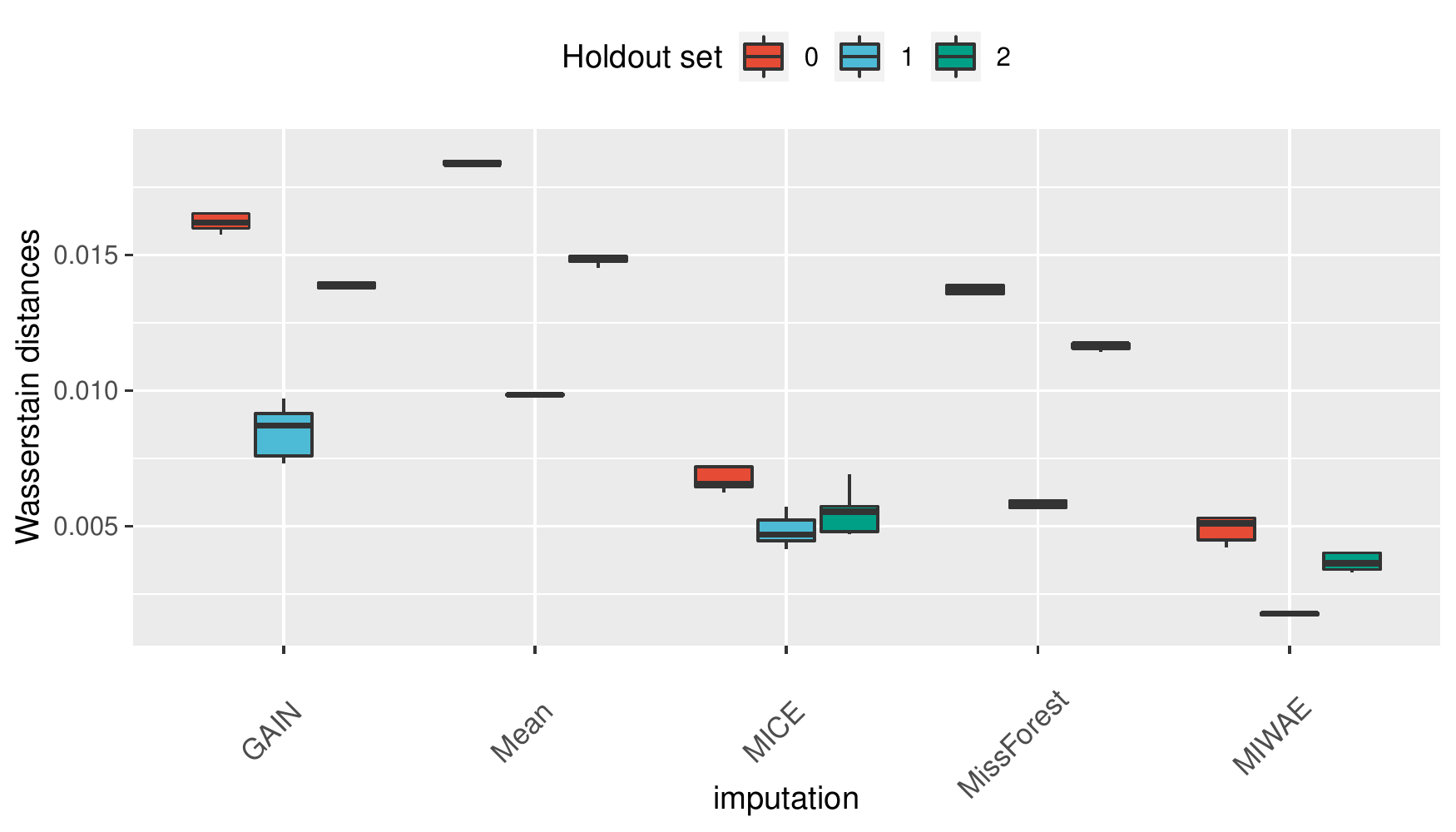}\\
      \hline      
      \parbox[c][][c]{0.5in}{\rotatebox[origin=t]{90}{50\%: 25\% (Log)}} & \includegraphics[height=4cm,width=5cm]{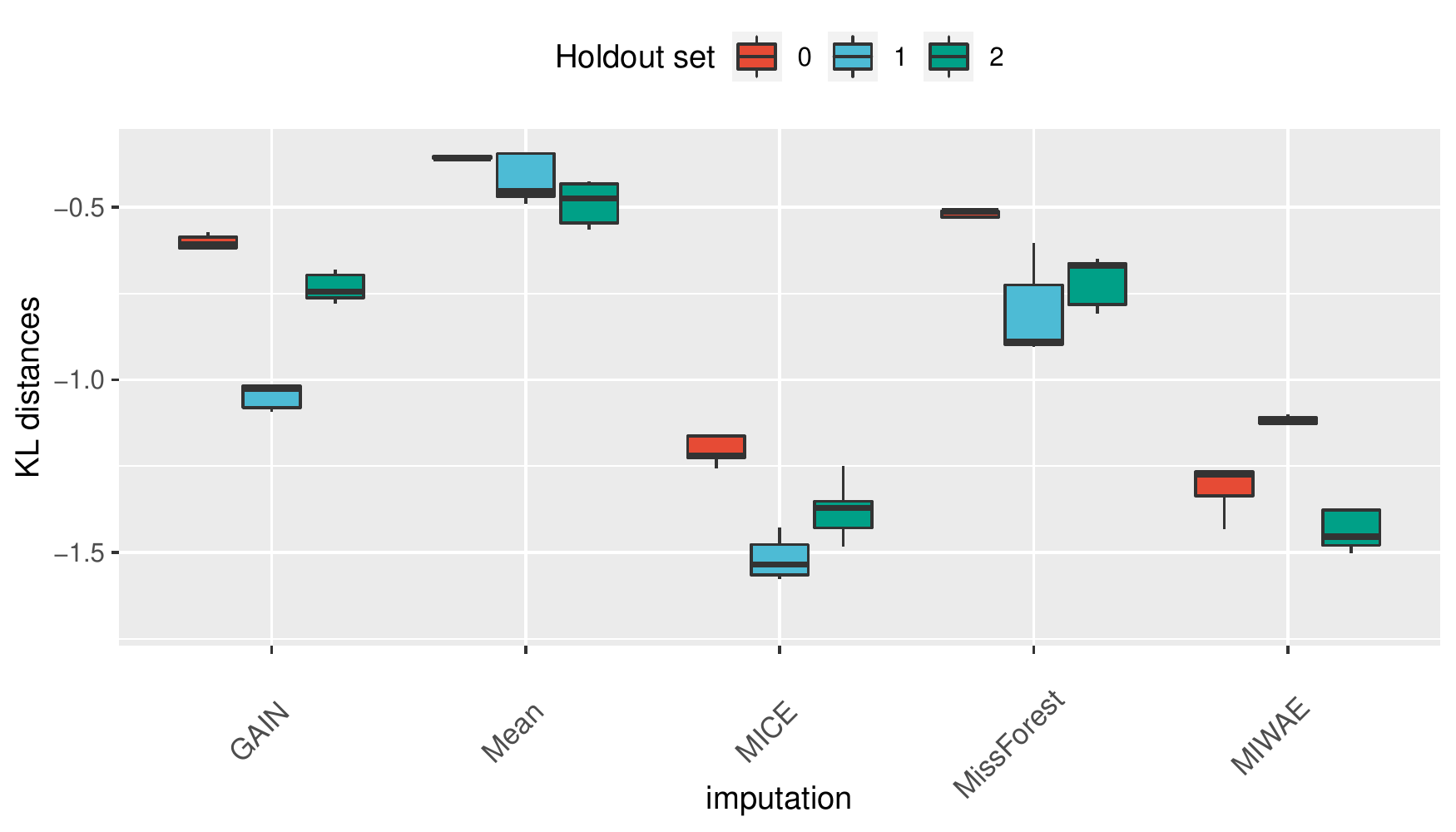}&
      \includegraphics[height=4cm,width=5cm]{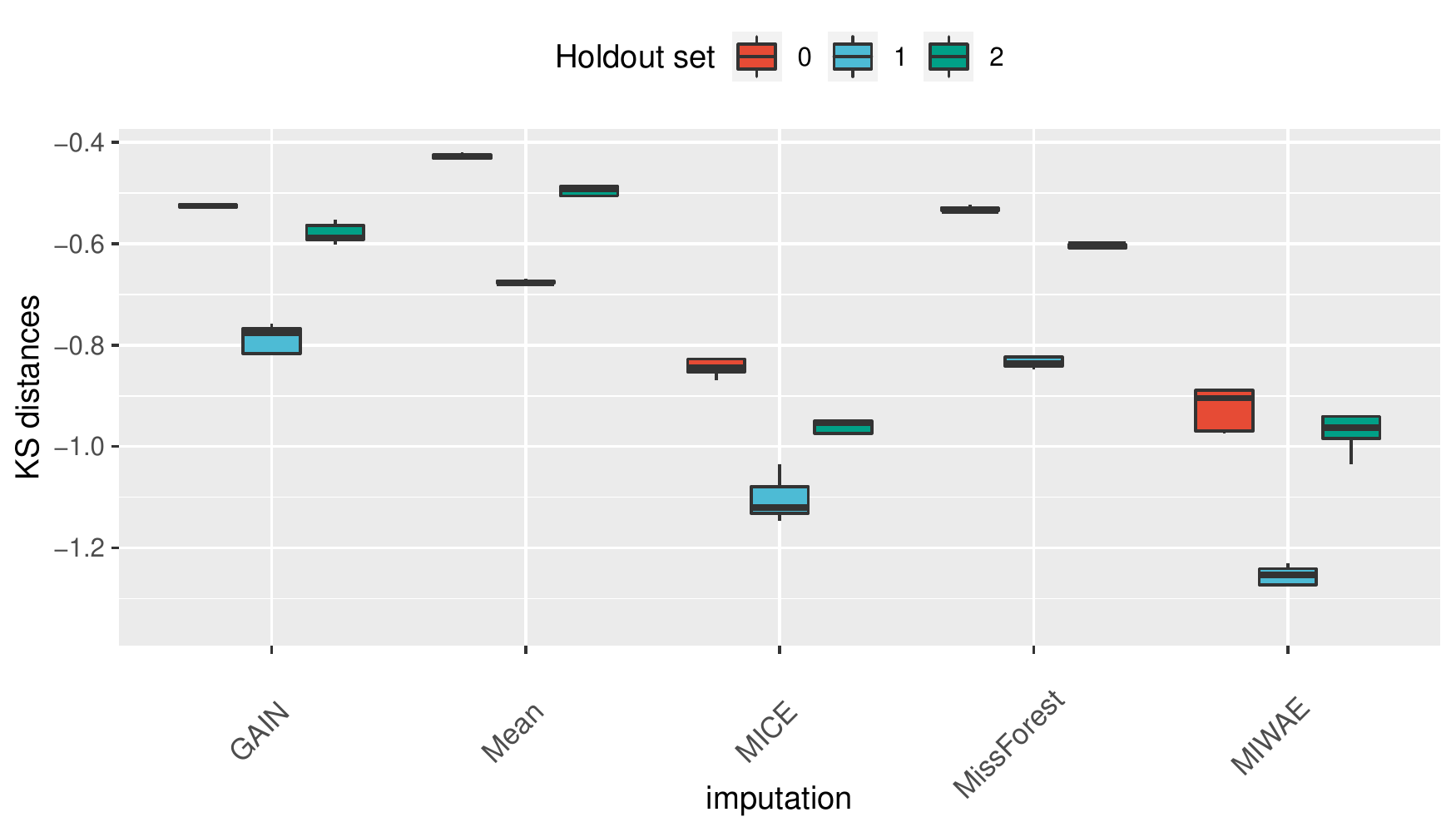}&
      \includegraphics[height=4cm,width=5cm]{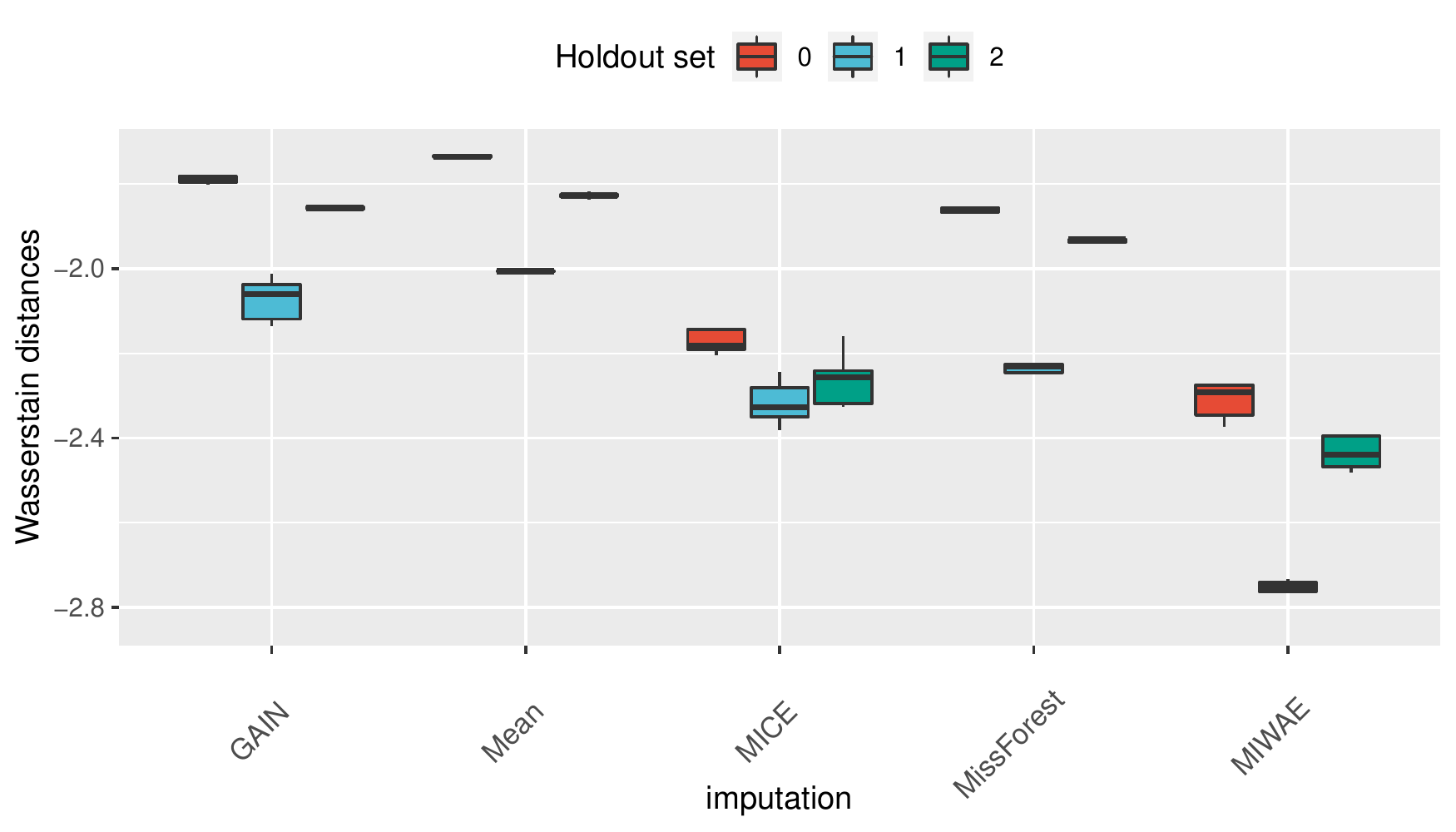}\\
     \hline
     \parbox[c][][c]{0.5in}{\rotatebox[origin=t]{90}{50\%: 50\%}} &
      \includegraphics[height=4cm,width=5cm]{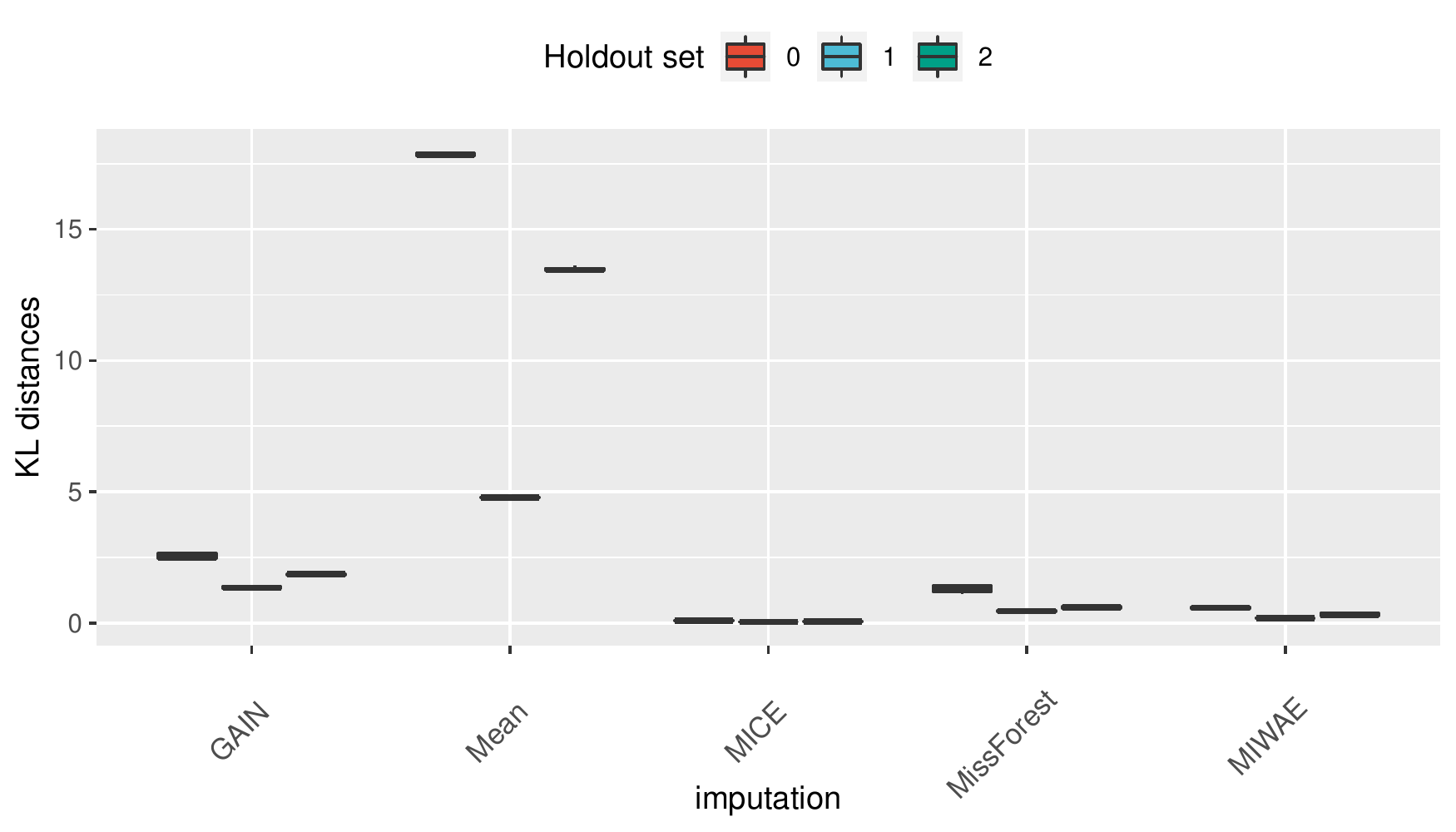}&
      \includegraphics[height=4cm,width=5cm]{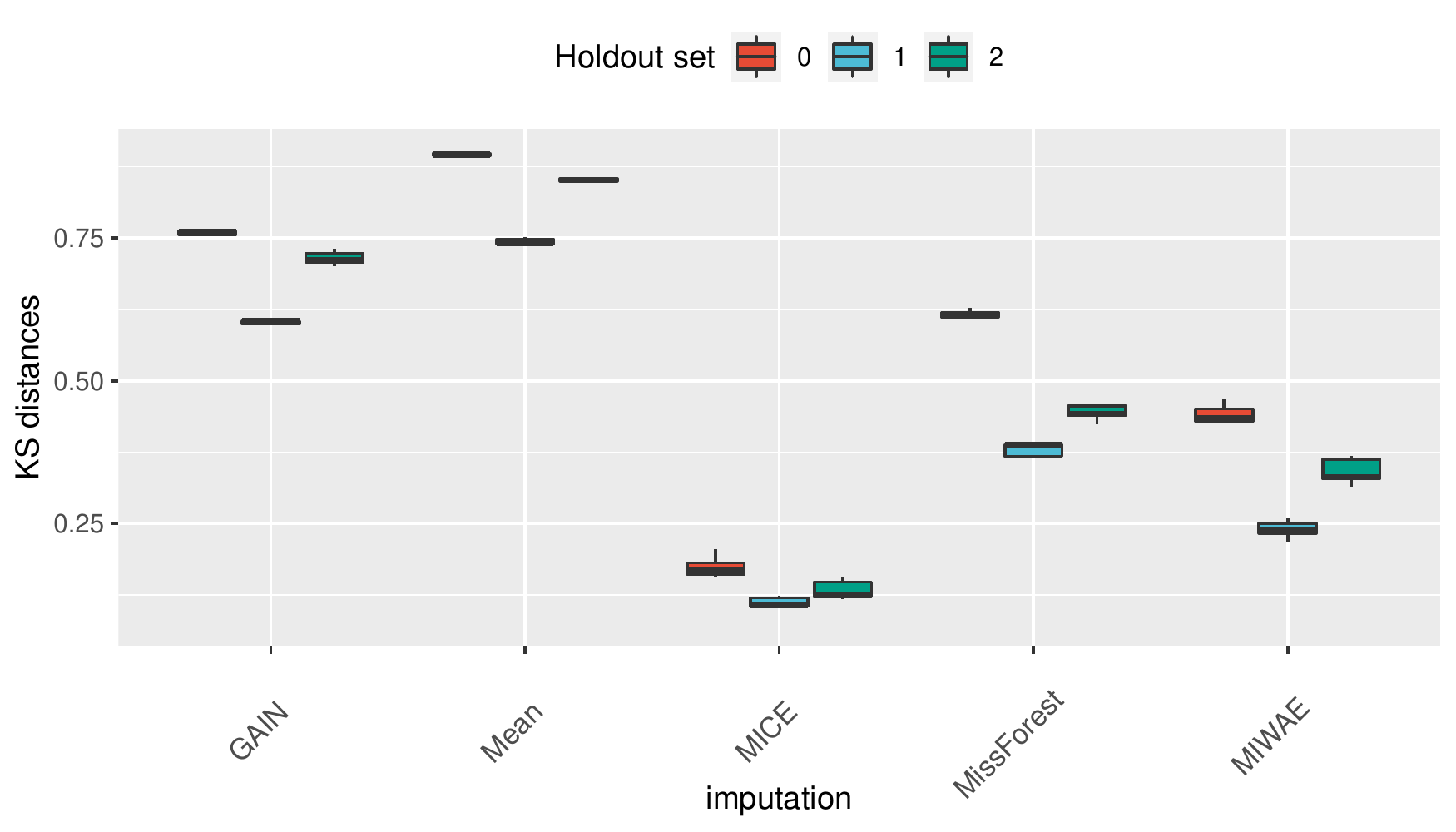}&
      \includegraphics[height=4cm,width=5cm]{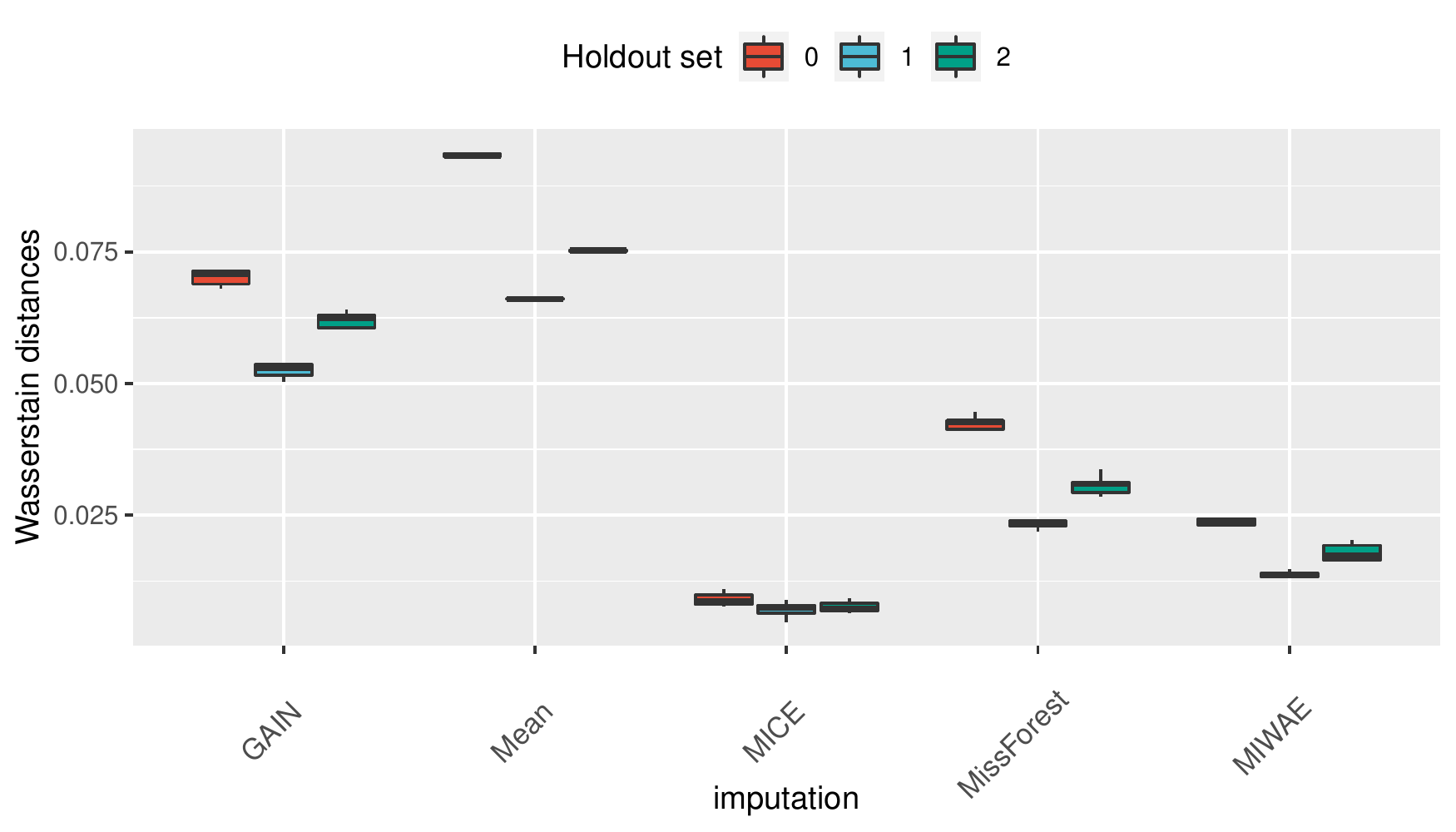}\\
      \hline      
      \parbox[c][][c]{0.5in}{\rotatebox[origin=t]{90}{50\%: 50\% (Log)}} &
      \includegraphics[height=4cm,width=5cm]{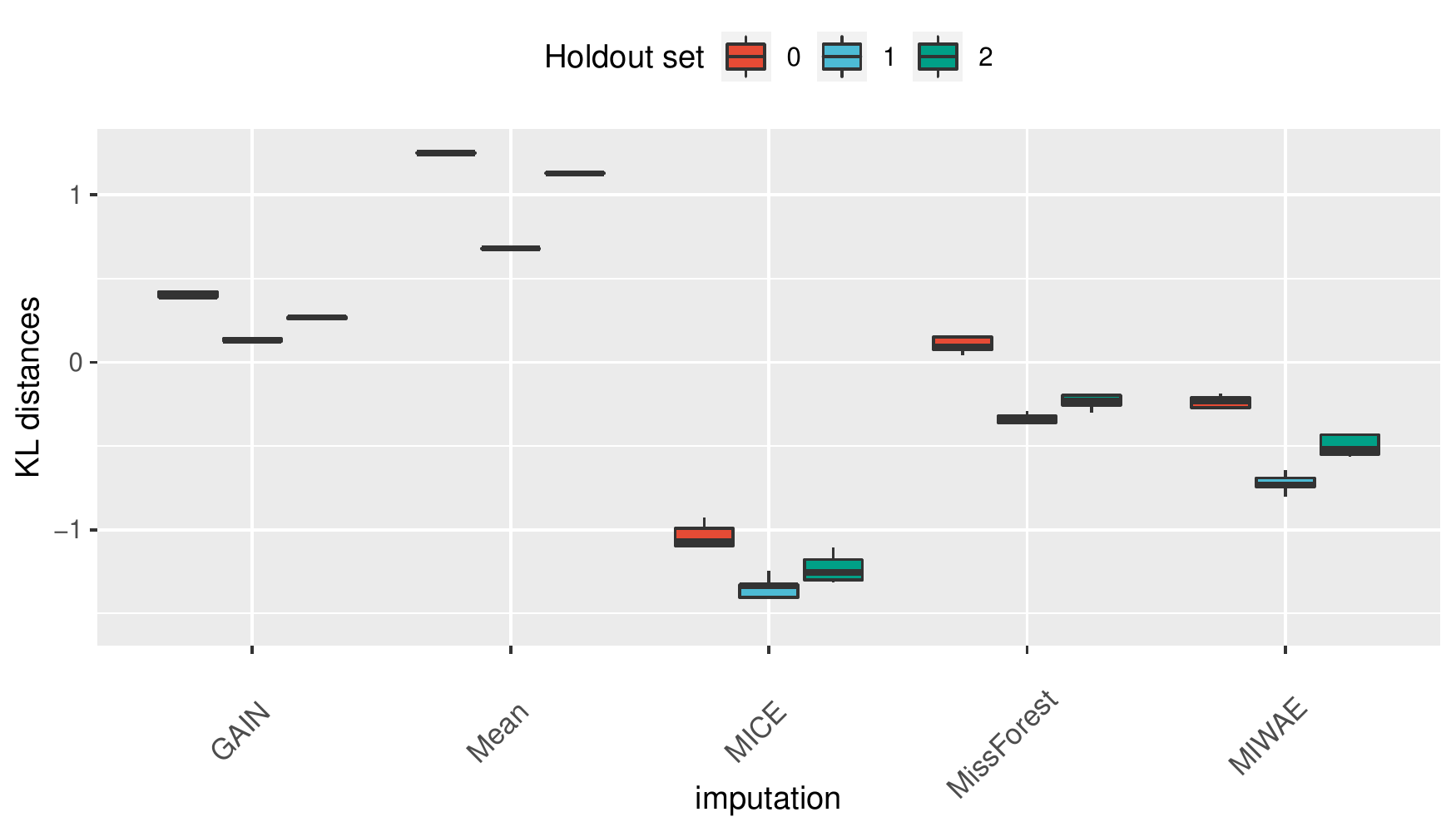}&
      \includegraphics[height=4cm,width=5cm]{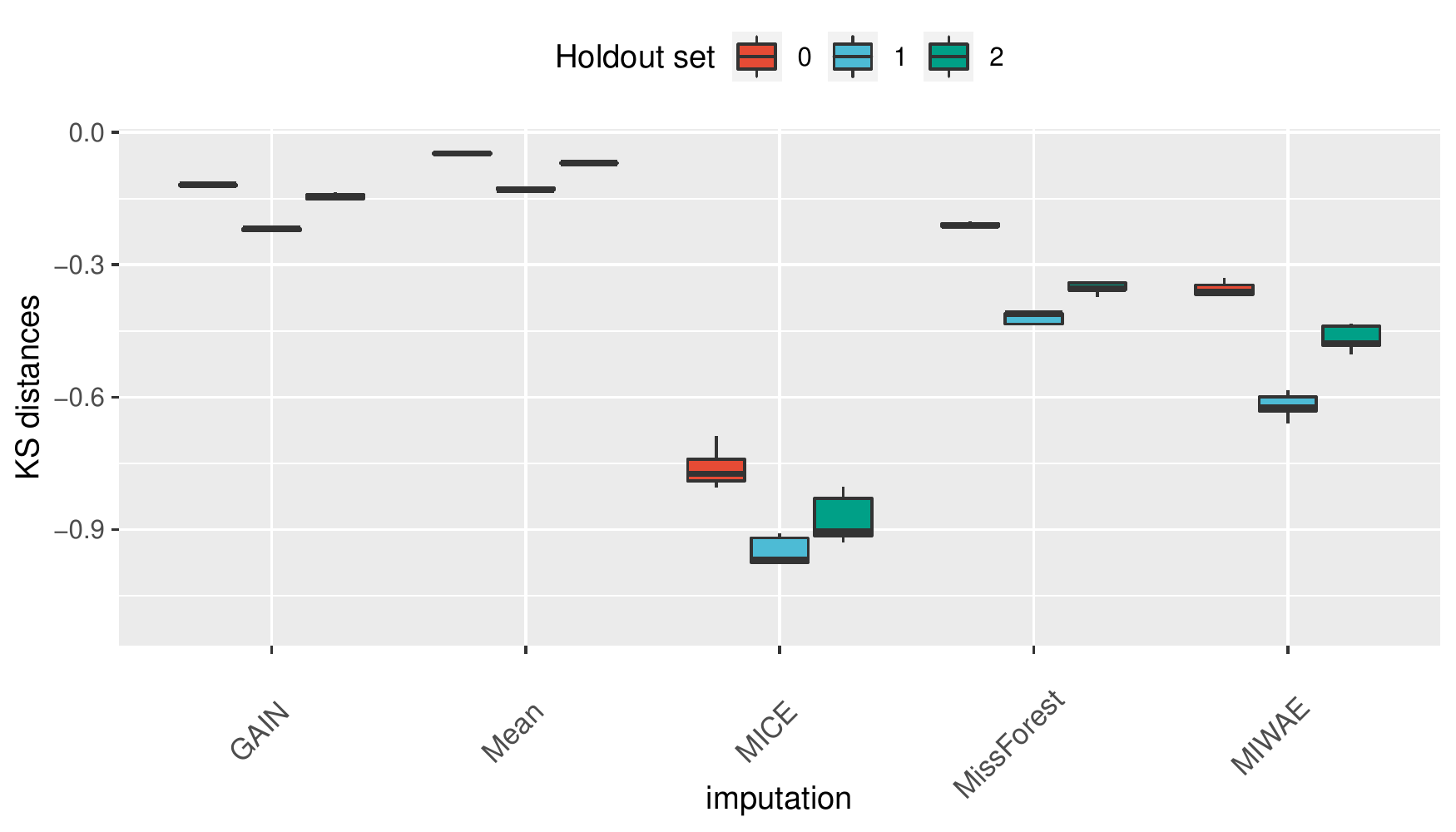}&
      \includegraphics[height=4cm,width=5cm]{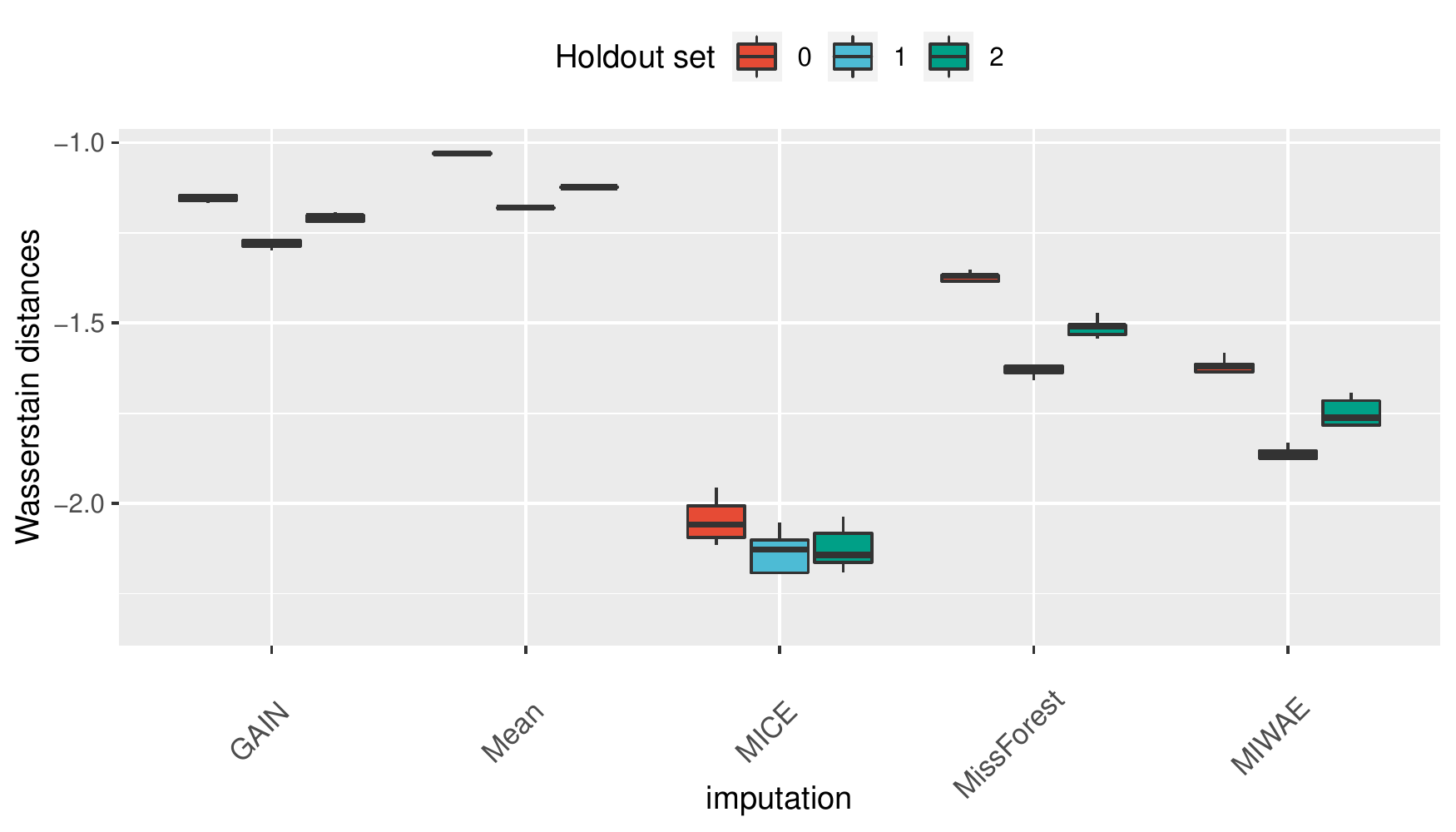}\\
    \end{tabular}
    \caption{The class C discrepancies for the sliced Wasserstein distances of the Simulated data at the 50\% missingness rate for the training data along with 25\% and 50\% for the test data. The original values and logarithms are shown for clarity.}
    \label{fig:datadistwise_supp_50_syn}
\end{figure}

\noindent
{\emph{C. Sliced Wasserstein distribution metrics.}}
In Figure~\ref{fig:datadistwise} and Supplementary Figures~\ref{fig:datadistwise_supp_25}--\ref{fig:datadistwise_supp_50_syn}, we show the discrepancies between the distributions of the sliced Wasserstein distances for the imputation methods at different train and test missingness rates for the \textbf{MIMIC-III} and \textbf{Simulated} datasets. 
For the \textbf{MIMIC-III} dataset, over all discrepancy scores and missingness rates, the MICE imputation method shows a clear dominance with the mean and GAIN methods performing poorly.
For the \textbf{Simulated} dataset, the MICE method again performs the best overall by all measures. At the 25\% test missingness rate the MIWAE imputation method is competitive with, and sometimes outperforms, MICE. Mean imputation performs the worst, with GAIN and MissForest performing similarly poorly.\\

\noindent
{\textbf{Sliced Wasserstein distance ratio analysis.}}
In Supplementary Figures~\ref{fig:distance_ratios}--\ref{fig:distance_ratios_syn} are the boxplots of the ratios of the distances from $J_{p}$ to $I_{p}$ for the imputed data compared to the original data for the \textbf{MIMIC-III} and \textbf{Simulated} datasets respectively. Firstly, we see that the MICE imputation method induces a much smaller distance ratio than any other method. Secondly, we note that with a increase in the train missingness rate the ratio of the distances remains largely consistent, however with an increase in the test missingness rate, we see a very significant increase in the ratio.\\

\begin{figure}[htb!]
    \centering
      \includegraphics[width=7cm]{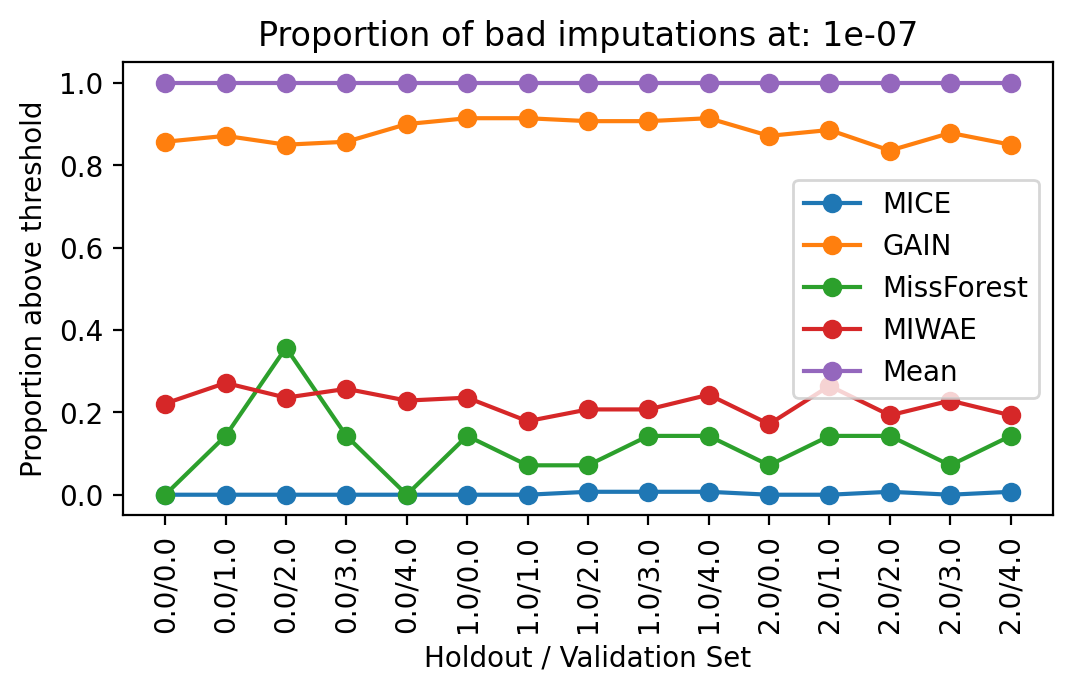}
        \caption{The proportion of repeated imputations that give outlier Wasserstein distances, at threshold $10^{-7}$, for different imputation methods.}
    \label{fig:comp_err}
\end{figure}

\noindent
{\textbf{Outlier analysis.}}
It is important to understand how stable the imputation methods are when imputation is repeated and also whether the stochastic nature of the imputation methods can lead to outlier imputed values for particular features. In Figure~\ref{fig:datadistwise} and the Supplementary Figures~\ref{fig:datadistwise_supp_25}--\ref{fig:datadistwise_supp_50} we see that some of the imputation methods can lead to distances with large variances, especially as the missingness rates in the train and test sets increase. This variance can result from either the random projections (in some cases the distributions match well and in others quite poorly) or stochasticity in the imputation algorithm. We are keen to understand the influence of each. In Figure~\ref{fig:comp_err} and Supplementary Figure~\ref{fig:comp_err_supp}, we see that the MICE method performs consistently well across all holdout and validation sets with no imputations above the distance of $10^{-7}$ and at the threshold of $1.5\times 10^{-8}$, 90\% of the MICE imputations are above this distance. This demonstrates a consistency in MICE imputations, with most imputations at a distance between $1.5\times 10^{-8}$ and $10^{-7}$ from the true values.

\begin{figure}[htb!]
    \centering
    \begin{tabular}{m{0.2in} | M{5cm} | M{5cm} | M{5cm}}
     \parbox[c][][c]{0.5in}{\rotatebox[origin=t]{90}{\multirow{2}{*}{A: Sample wise measures}}} &
      \includegraphics[height=4cm,width=5cm]{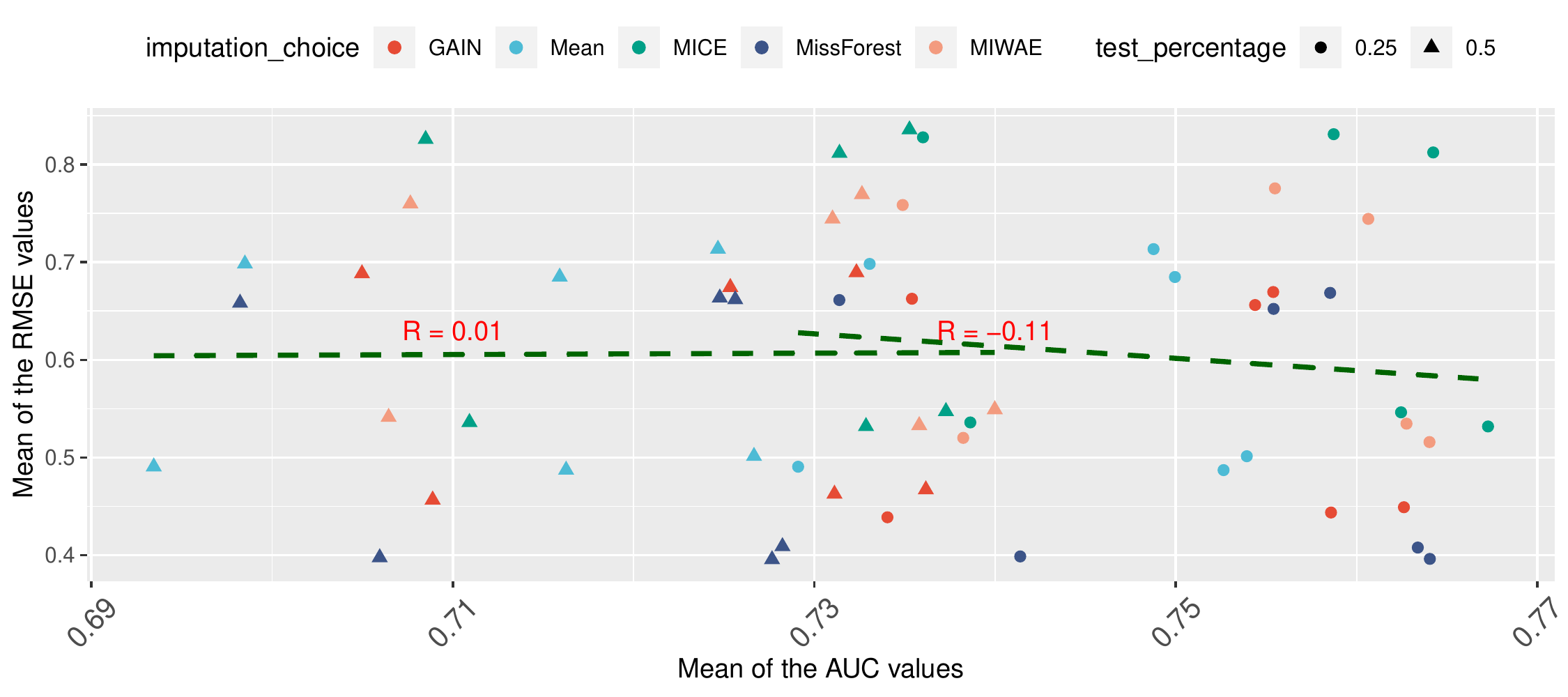}&
      \includegraphics[height=4cm,width=5cm]{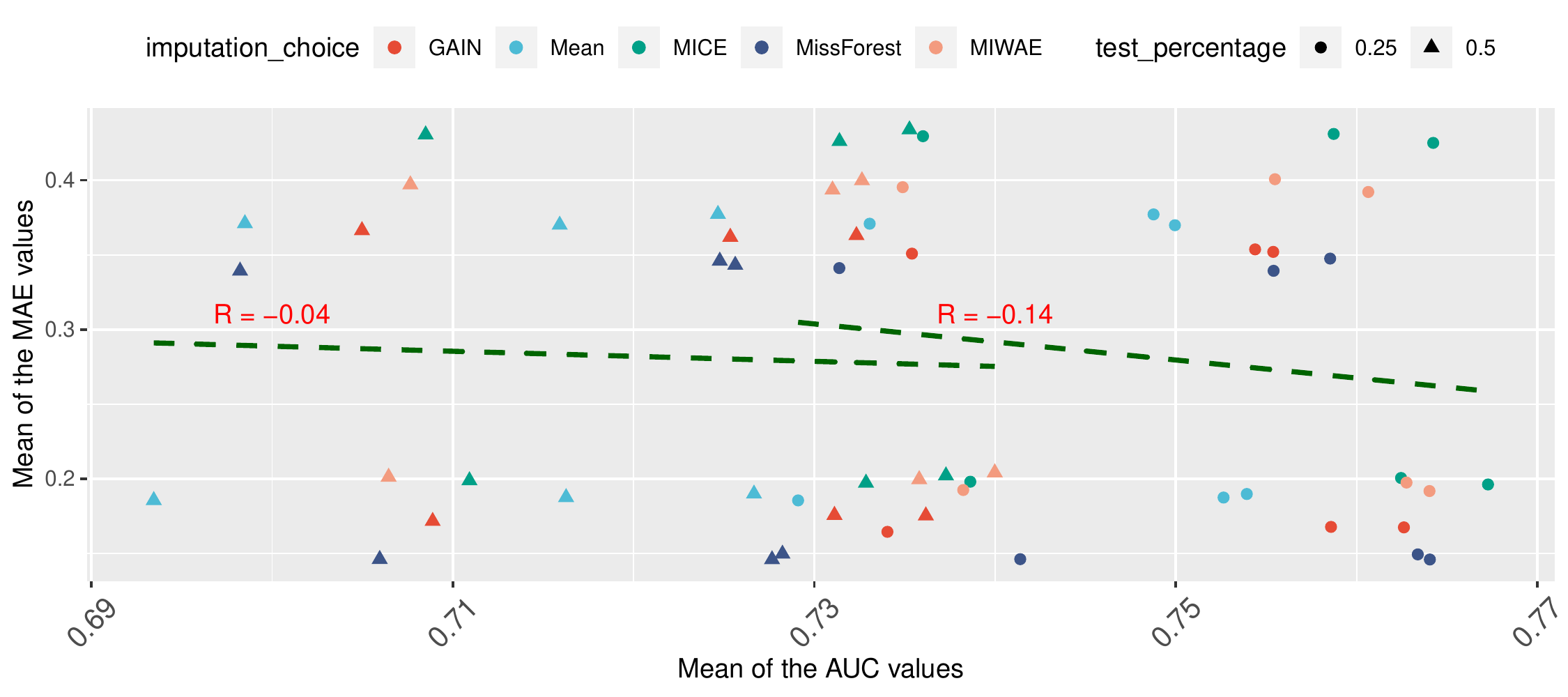}&
      \includegraphics[height=4cm,width=5cm]{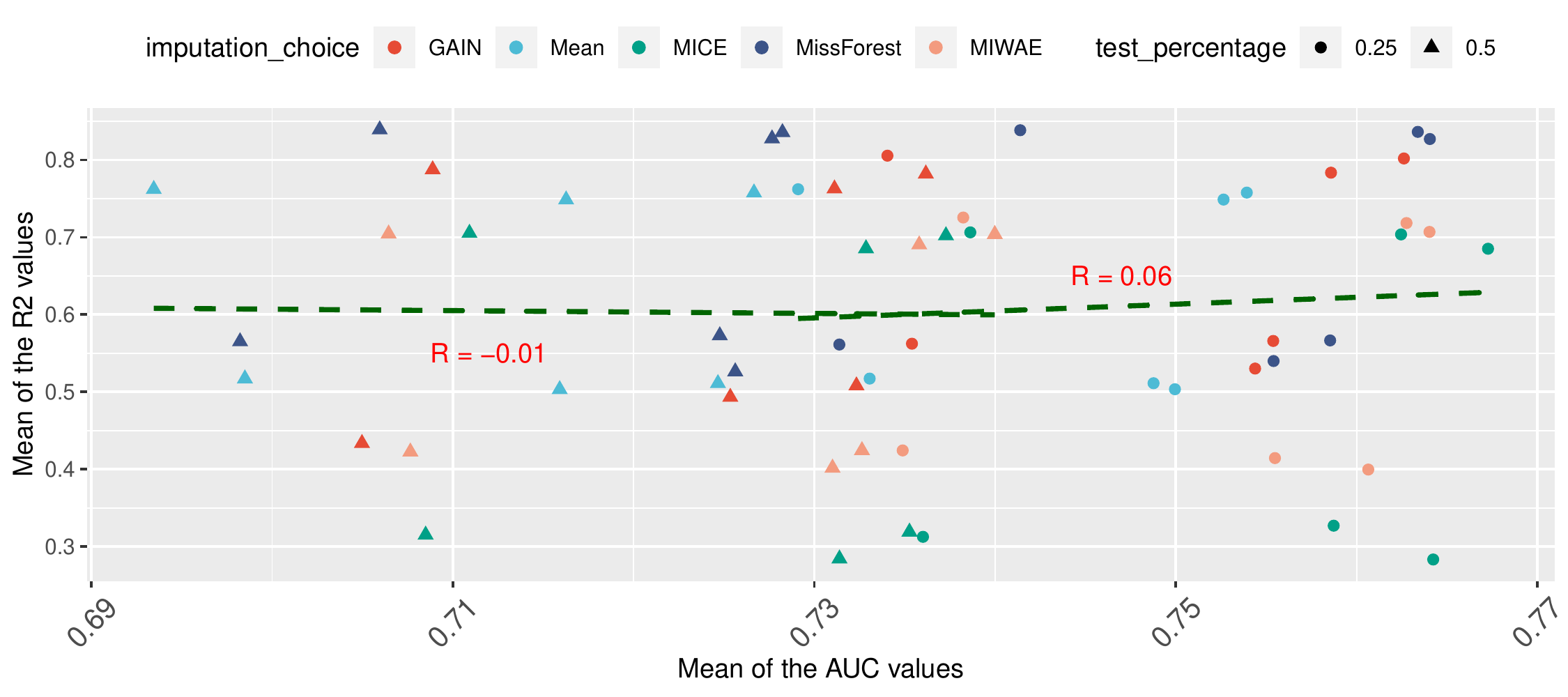}\\
      & (a) RMSE versus AUC &
      (b) MAE versus AUC &
      (c) $R^{2}$ versus AUC \\
      \hline
      \parbox[c][][c]{0.5in}{\rotatebox[origin=c]{90}{\multirow{2}{*}{B: Feature-wise Distances}}} &
      \includegraphics[height=4cm,width=5cm]{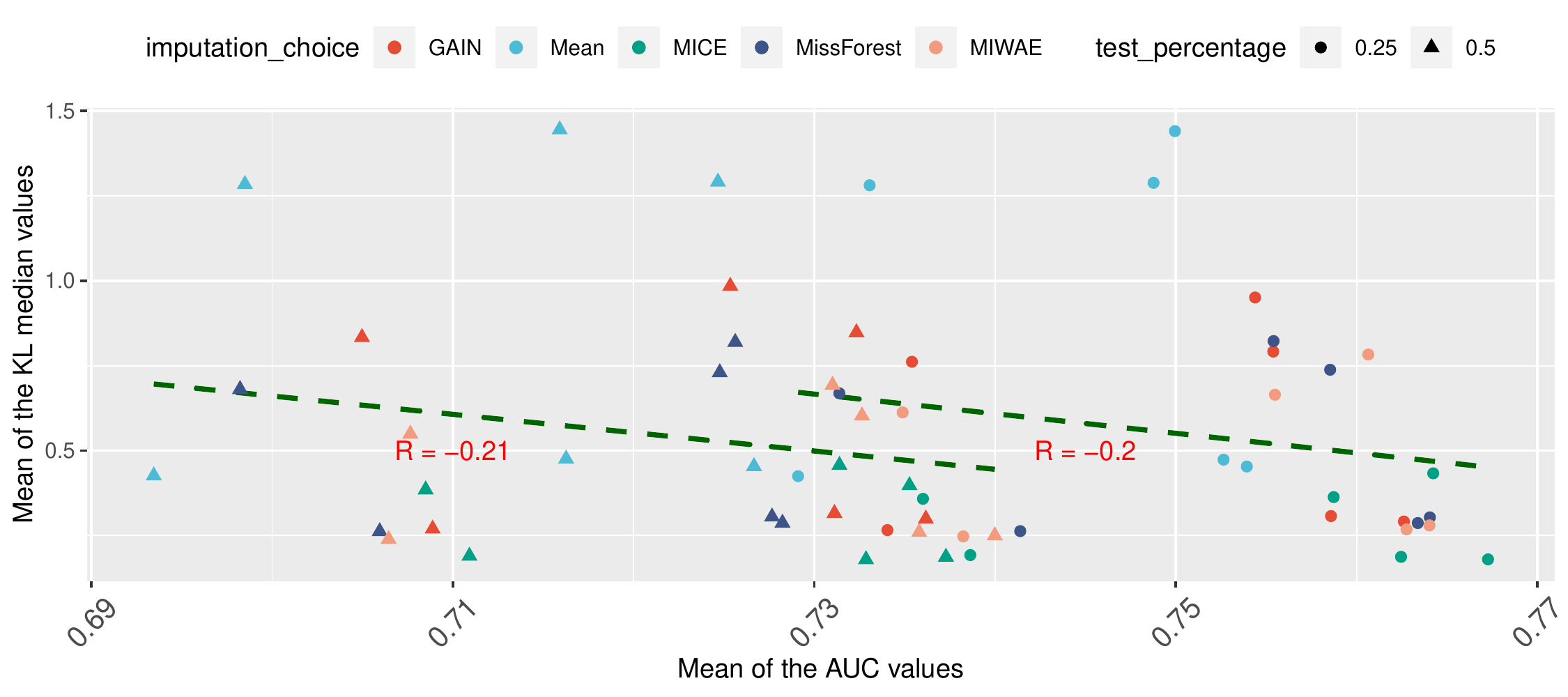}&
      \includegraphics[height=4cm,width=5cm]{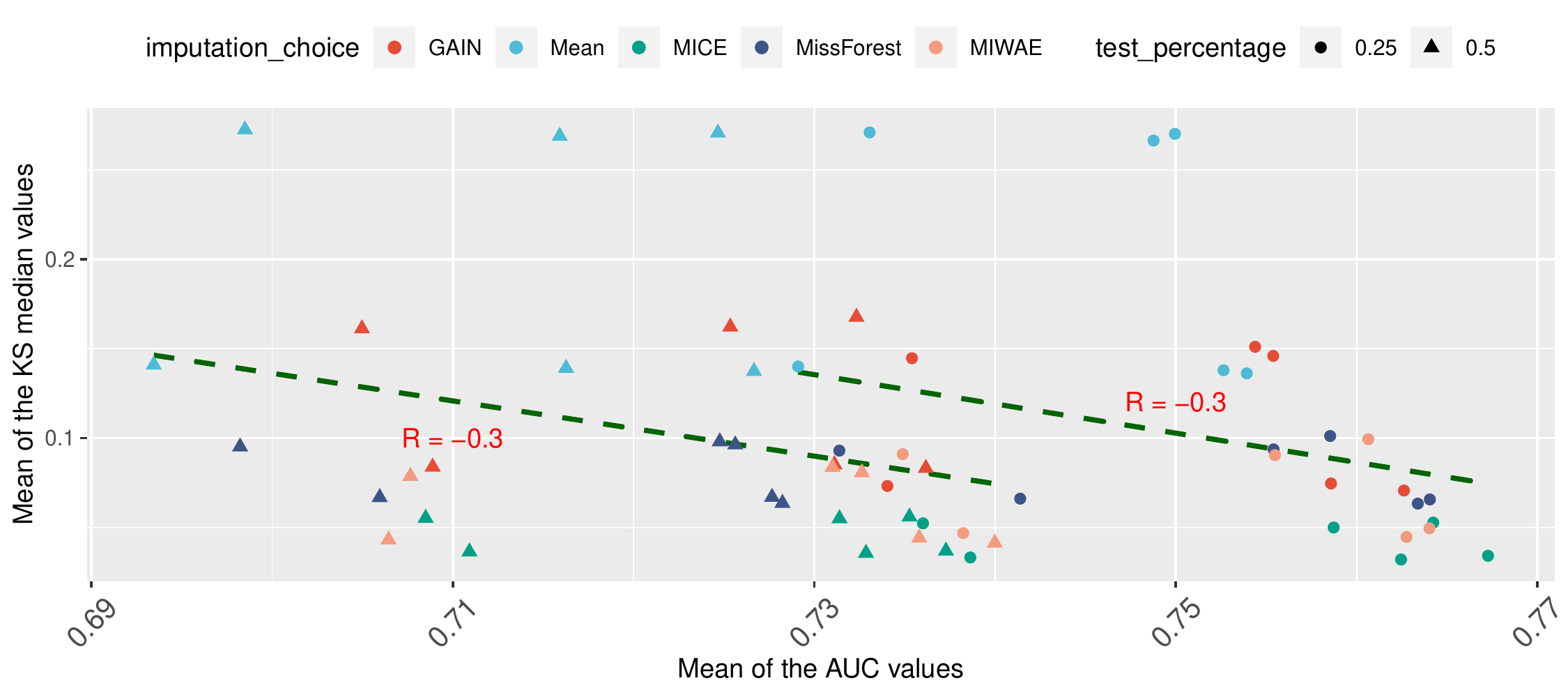}&
      \includegraphics[height=4cm,width=5cm]{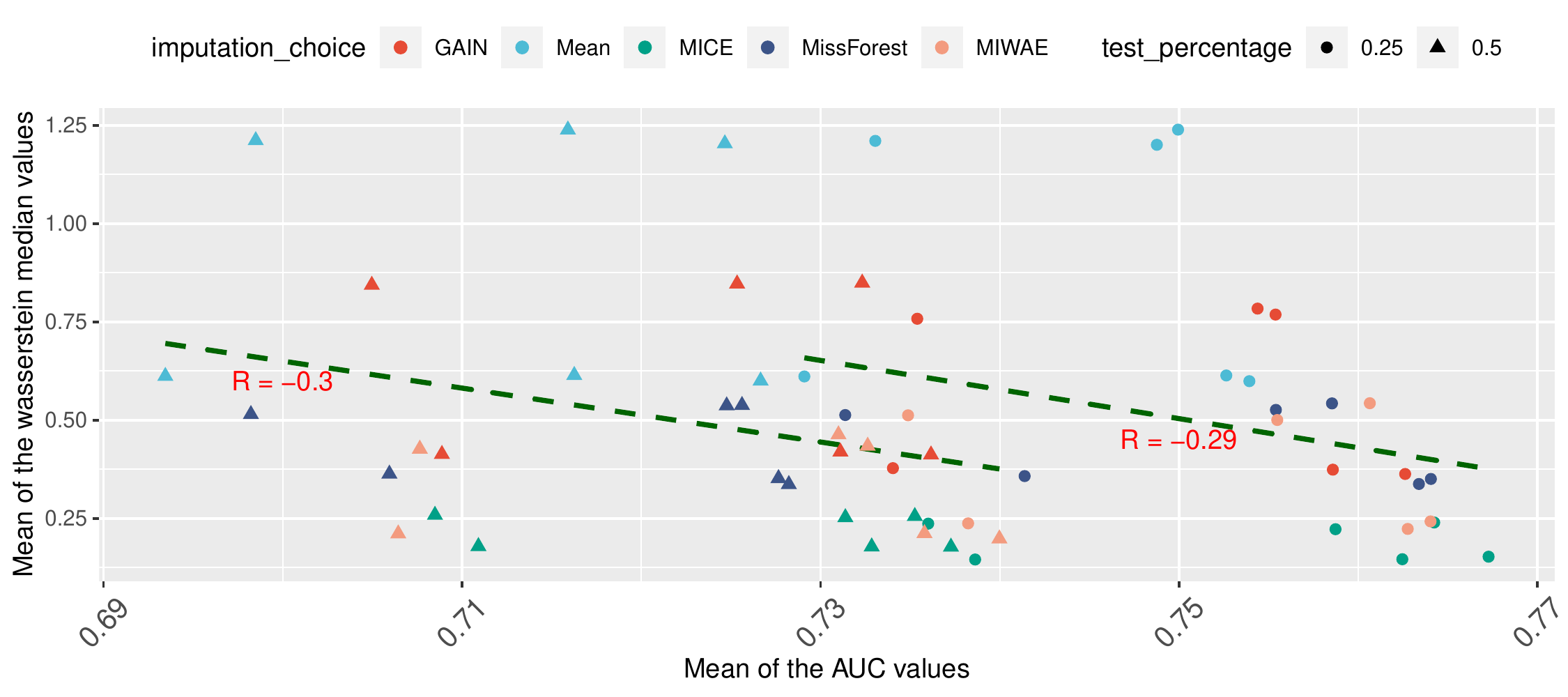}\\
      & (d) KL versus AUC &
      (e) KS versus AUC &
      (f) 2W versus AUC \\
      \hline
      \parbox[c][][c]{0.5in}{\rotatebox[origin=c]{90}{\multirow{2}{*}{C: Sliced Distances}}} &
      \includegraphics[height=4cm,width=5cm]{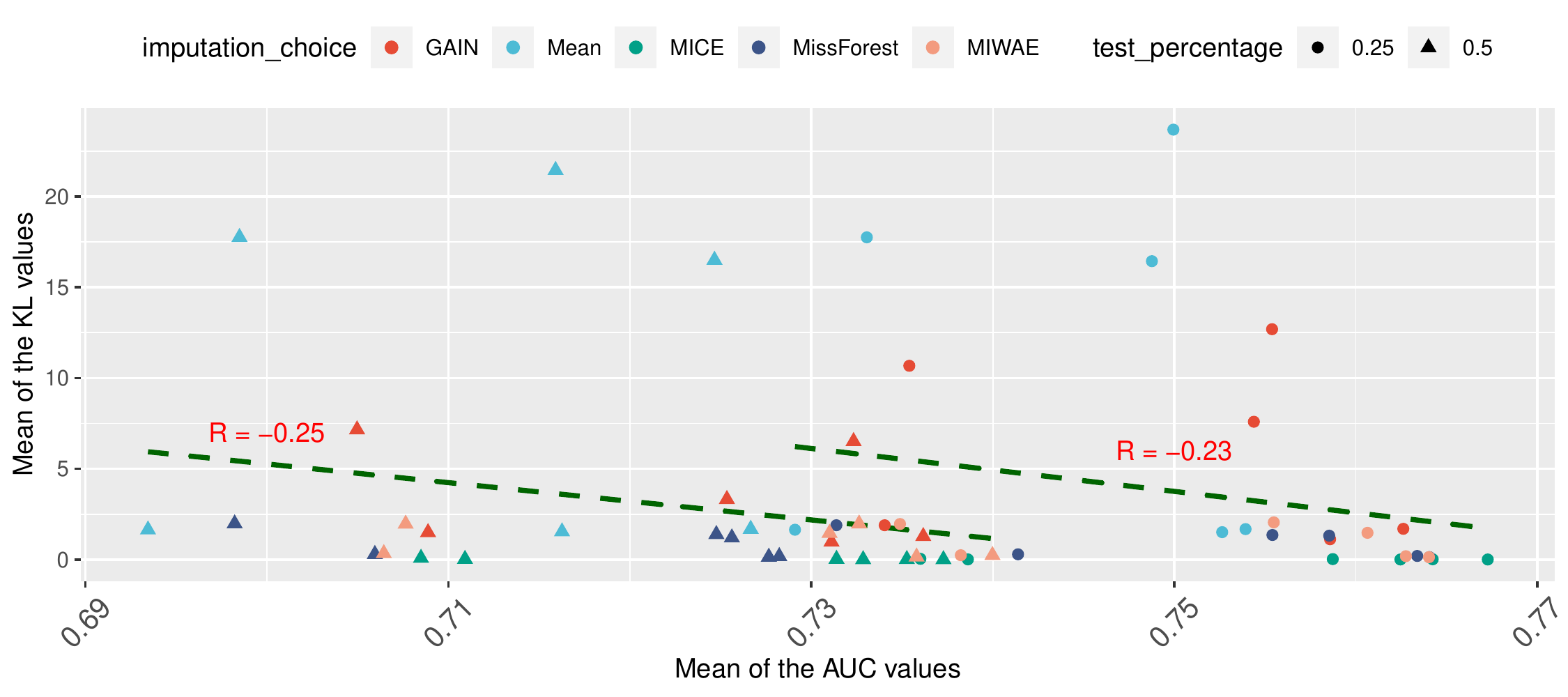}&
      \includegraphics[height=4cm,width=5cm]{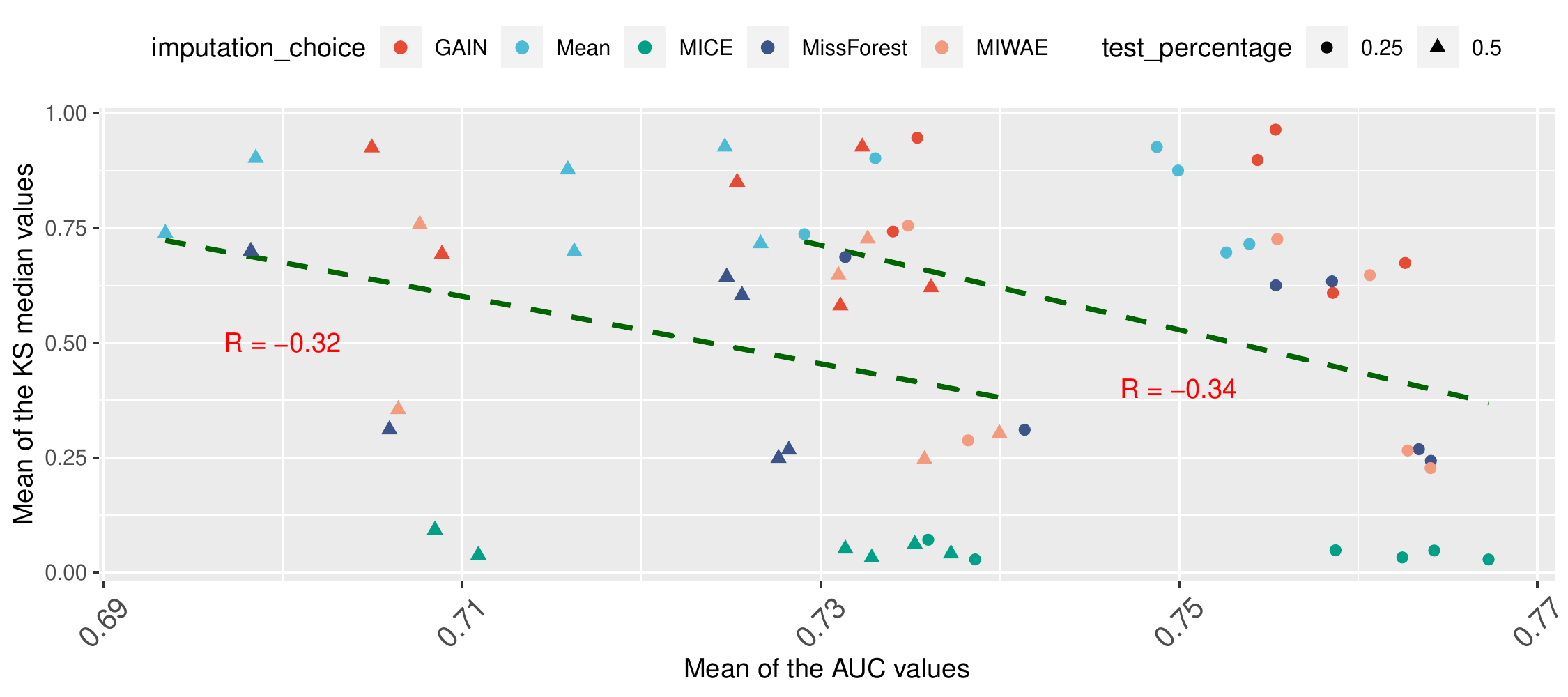}&
      \includegraphics[height=4cm,width=5cm]{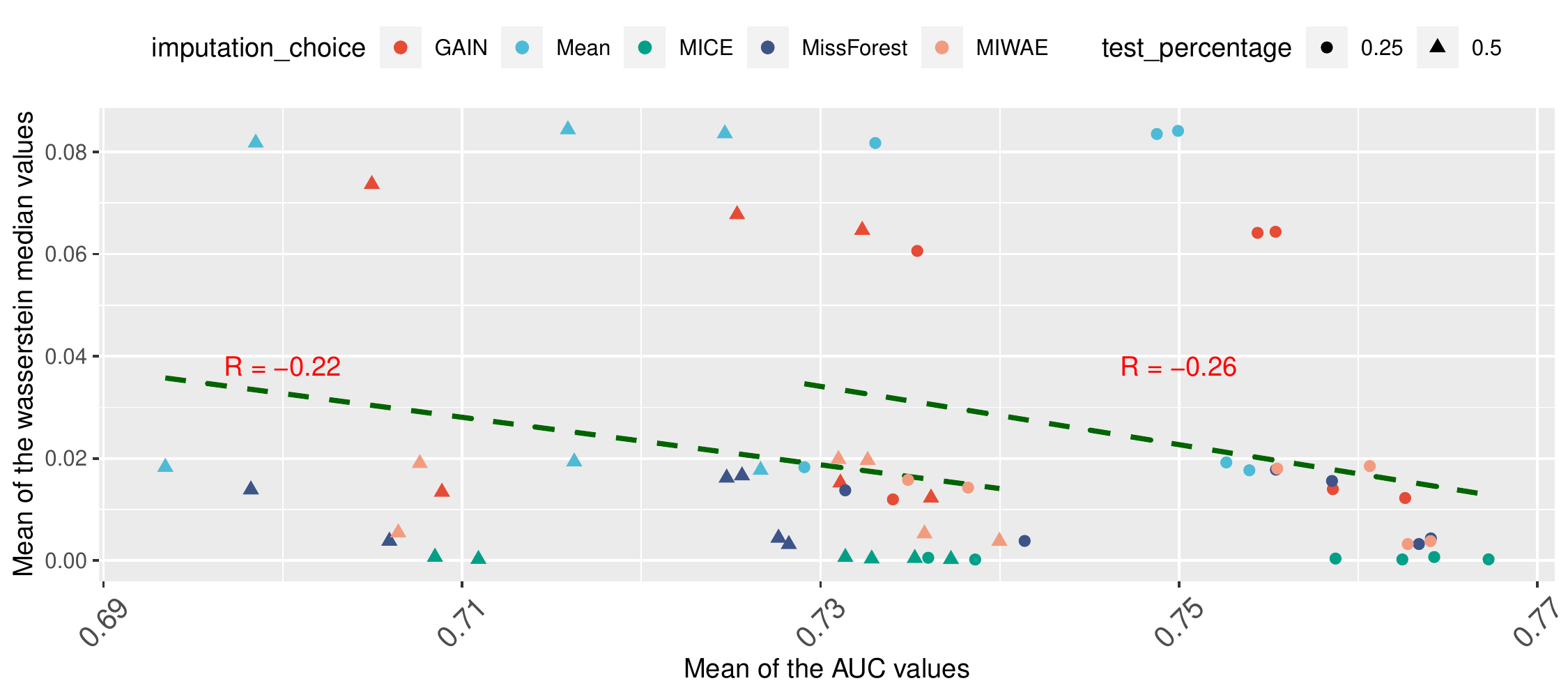}\\
      & (g) KL metric versus AUC &
      (h) KS metric versus AUC &
      (i) 2W metric versus AUC \\
    \end{tabular}
    \caption{The different imputation discrepancy metrics from classes A, B and C are shown against the downstream AUC value for the classification task of the \textbf{MIMIC-III} dataset. Trend lines are shown for 25\% and 50\% test missingness separately.}
    \label{fig:corr_quality_mimic}
\end{figure}

\begin{figure}[htb!]
     \centering
       \includegraphics[width=0.76\textwidth]{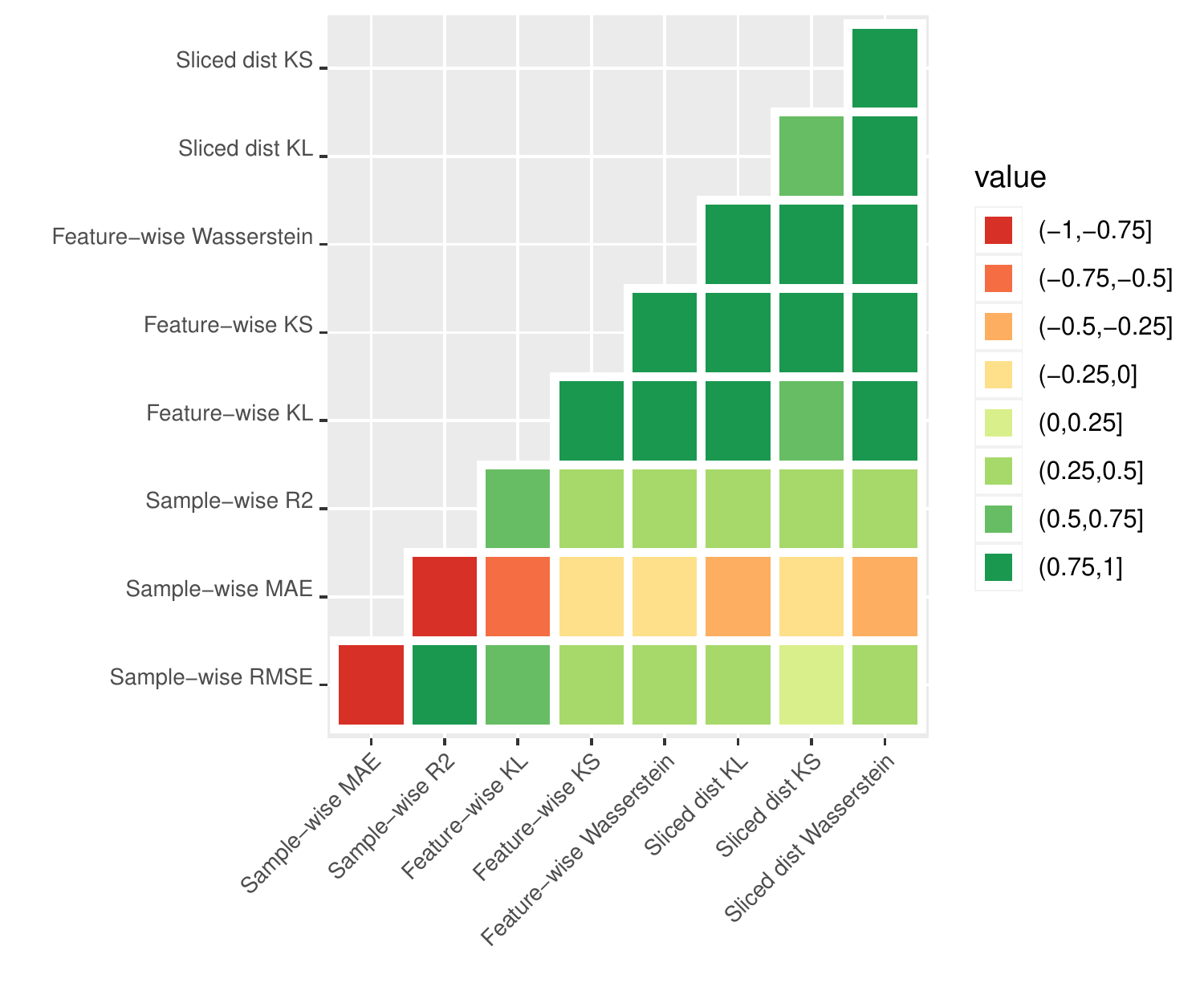}
 \caption{Correlation heatmap for all discrepancy metrics considered in this paper for the \textbf{MIMIC-III} dataset. \label{fig:conc_metrics_mimic}}
\end{figure}

\begin{figure}[htb!]
    \centering
    \begin{tabular}{m{0.2in} | M{5cm} | M{5cm} | M{5cm}}
     \parbox[c][][c]{0.5in}{\rotatebox[origin=t]{90}{\multirow{2}{*}{A: Sample wise measures}}} &
      \includegraphics[height=4cm,width=5cm]{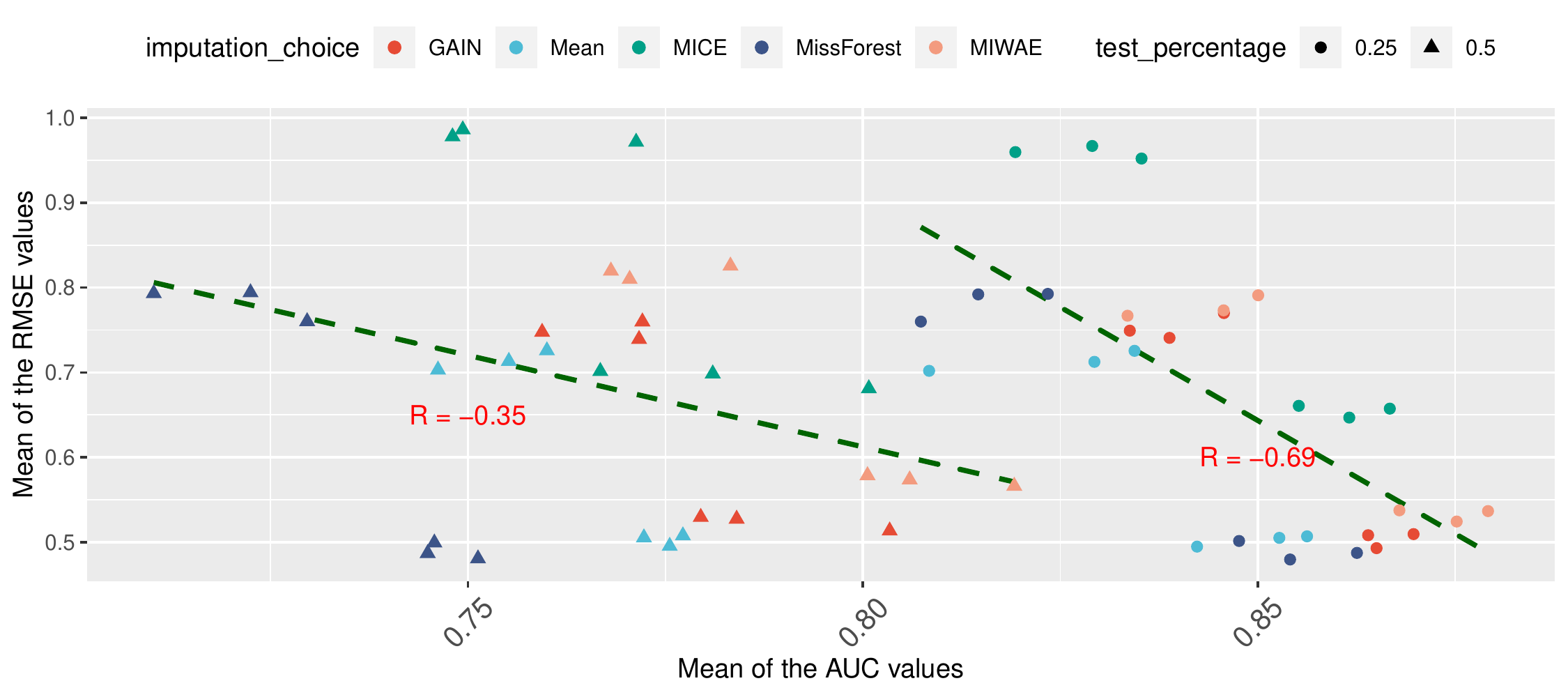}&
      \includegraphics[height=4cm,width=5cm]{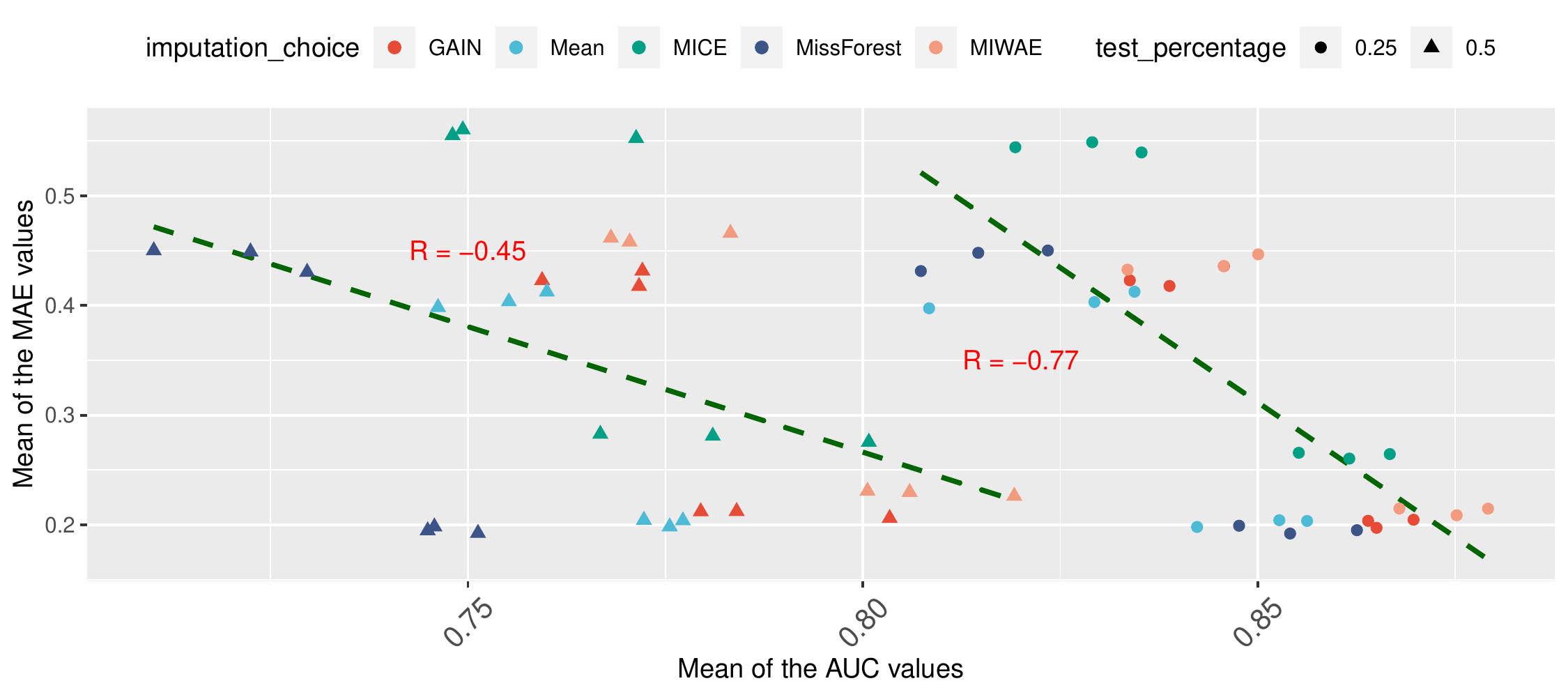}&
      \includegraphics[height=4cm,width=5cm]{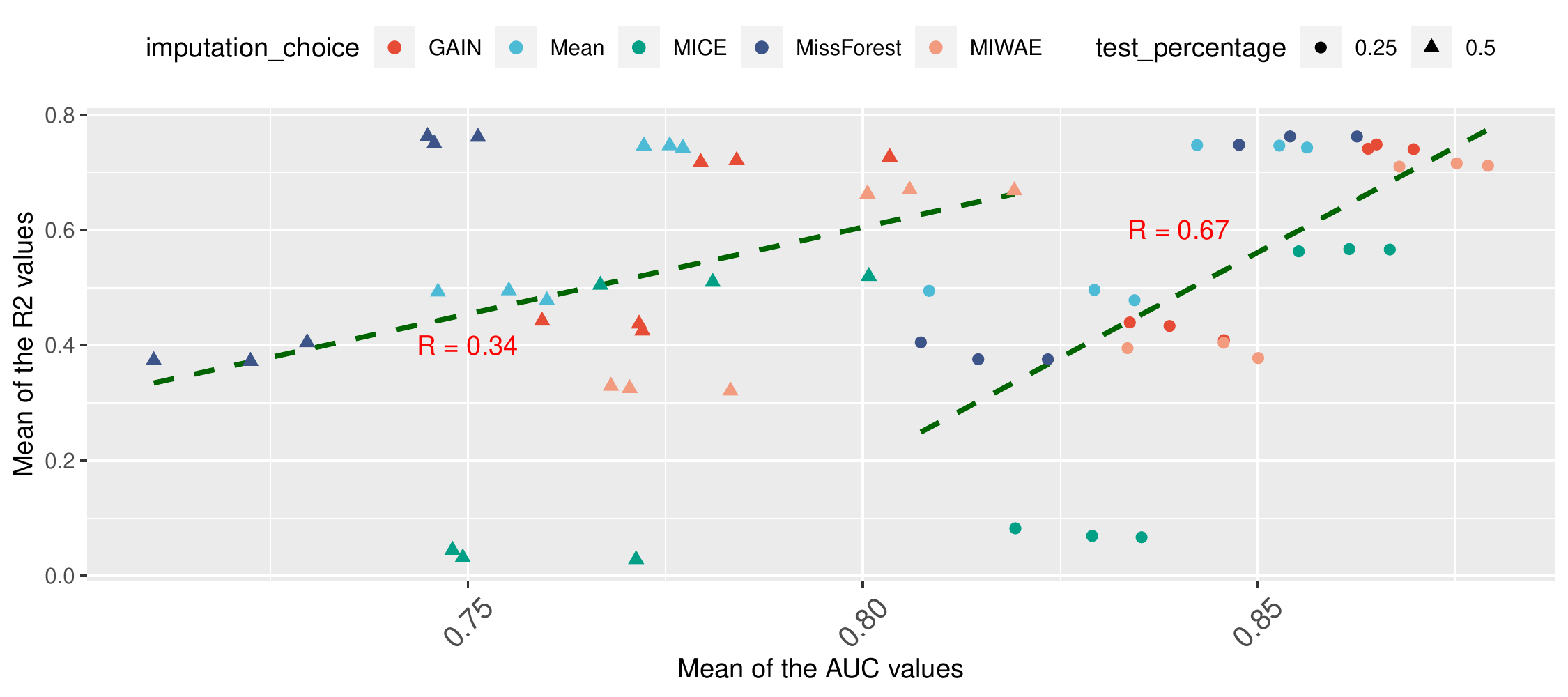}\\
      & (a) RMSE versus AUC &
      (b) MAE versus AUC &
      (c) $R^{2}$ versus AUC \\
      \hline
      \parbox[c][][c]{0.5in}{\rotatebox[origin=c]{90}{\multirow{2}{*}{B: Feature-wise Distances}}} &
      \includegraphics[height=4cm,width=5cm]{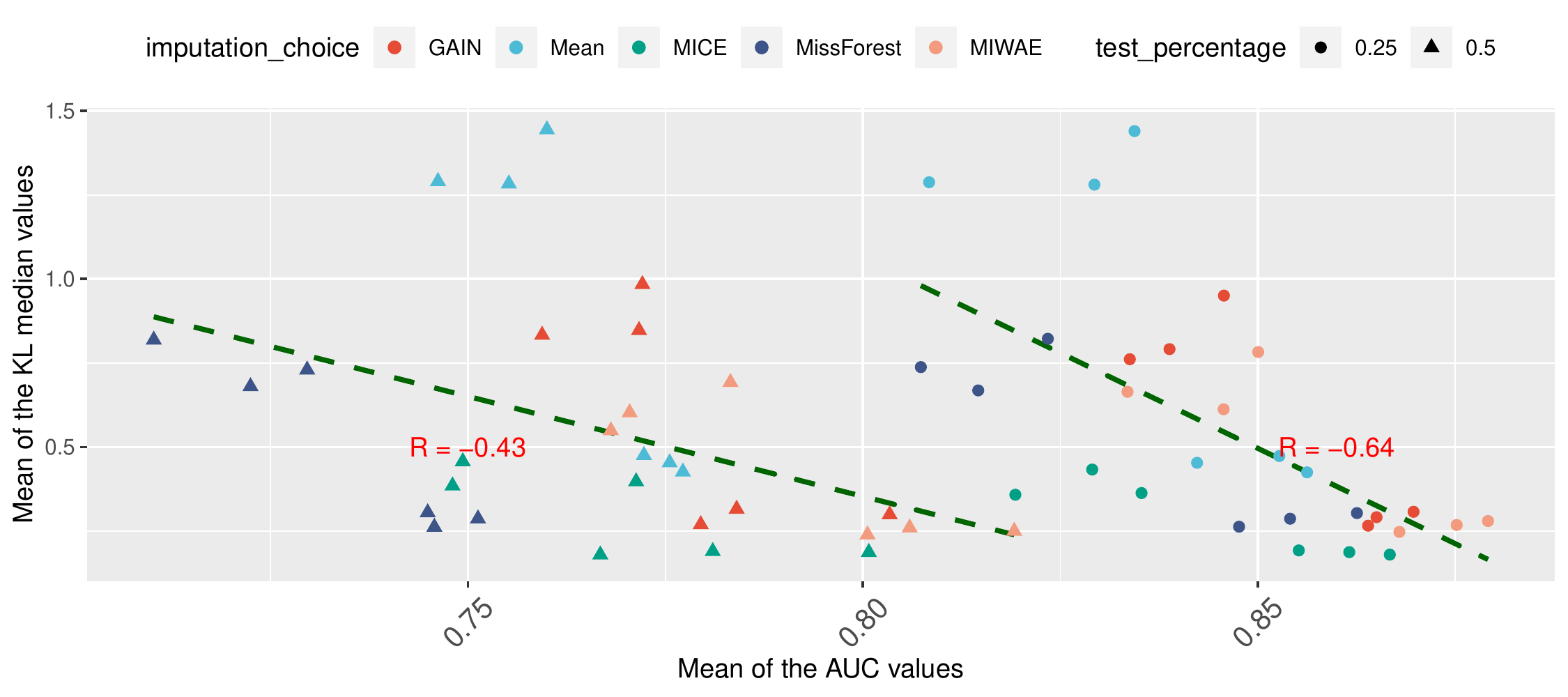}&
      \includegraphics[height=4cm,width=5cm]{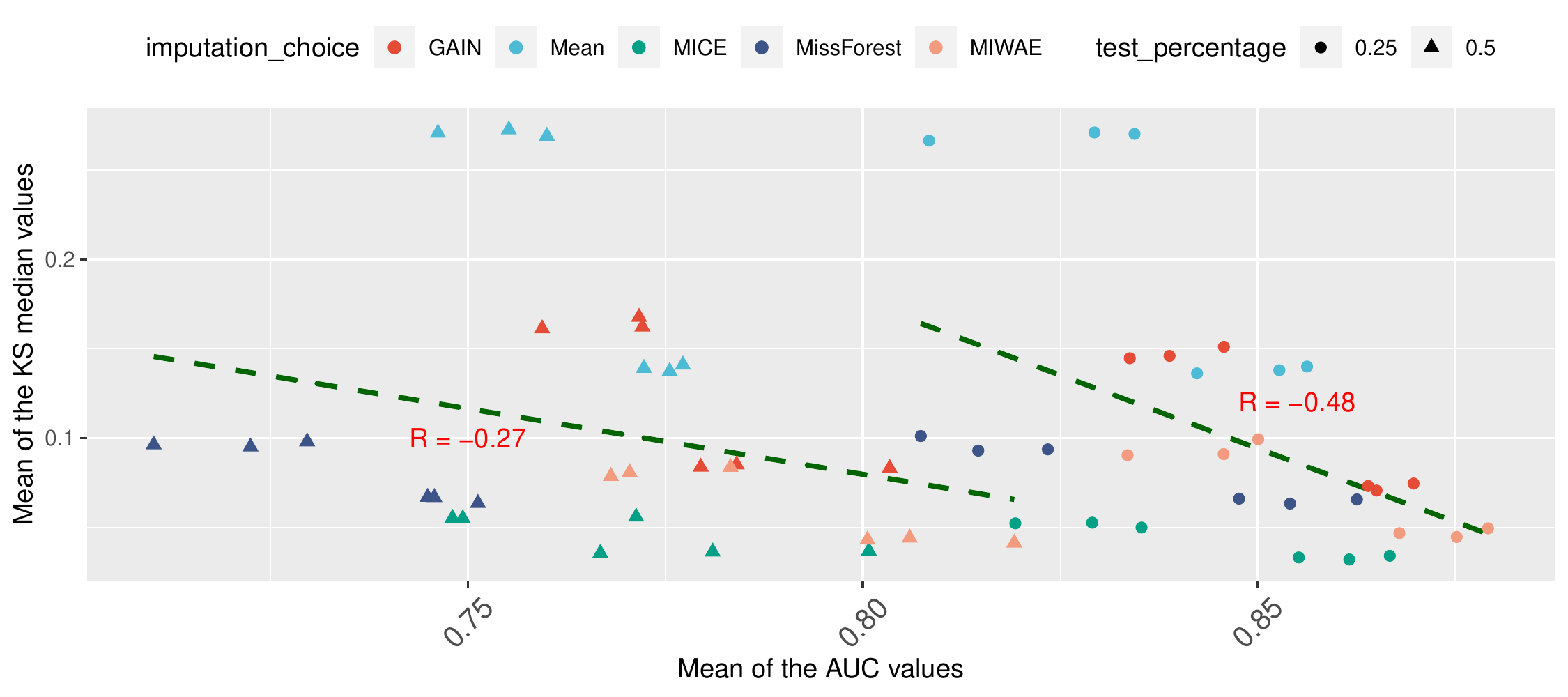}&
      \includegraphics[height=4cm,width=5cm]{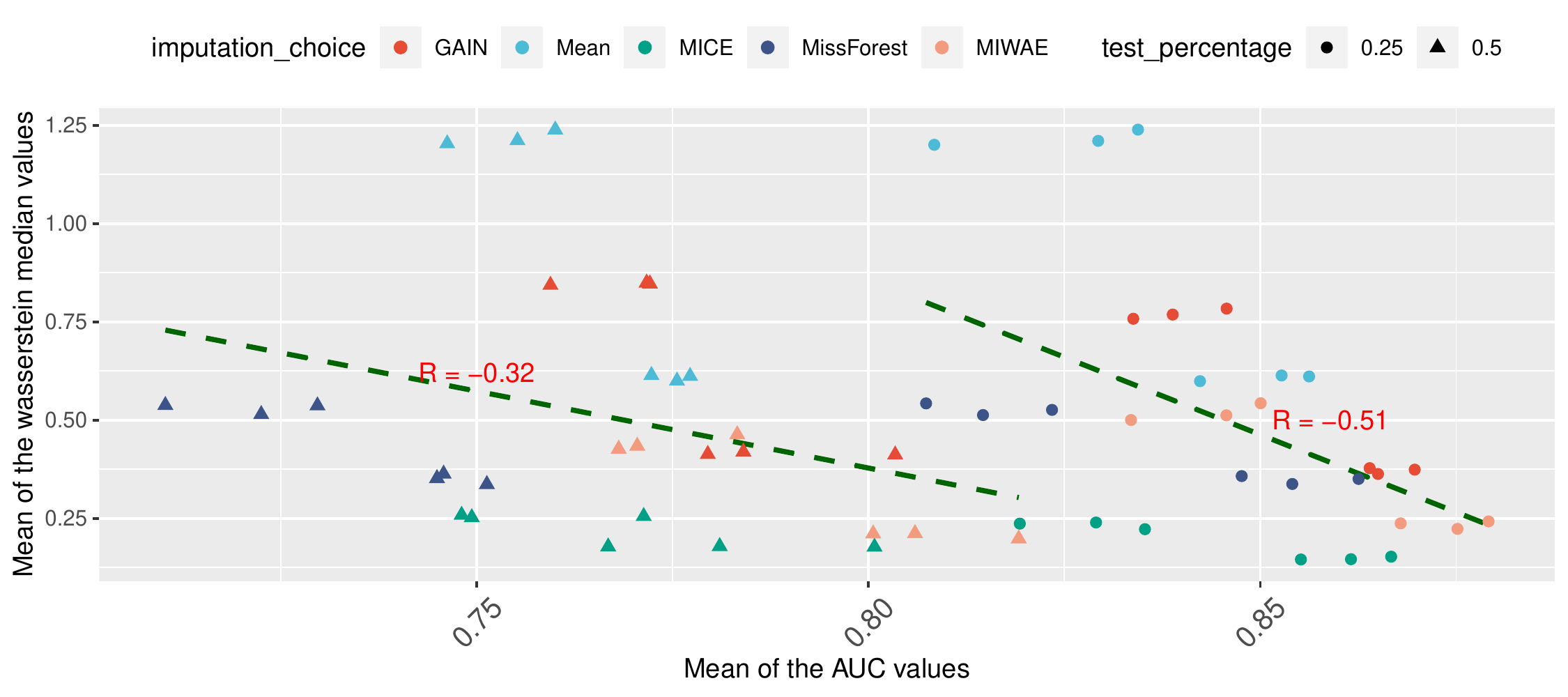}\\
      & (d) KL versus AUC &
      (e) KS versus AUC &
      (f) 2W versus AUC \\
      \hline
      \parbox[c][][c]{0.5in}{\rotatebox[origin=c]{90}{\multirow{2}{*}{C: Sliced Distances}}} &
      \includegraphics[height=4cm,width=5cm]{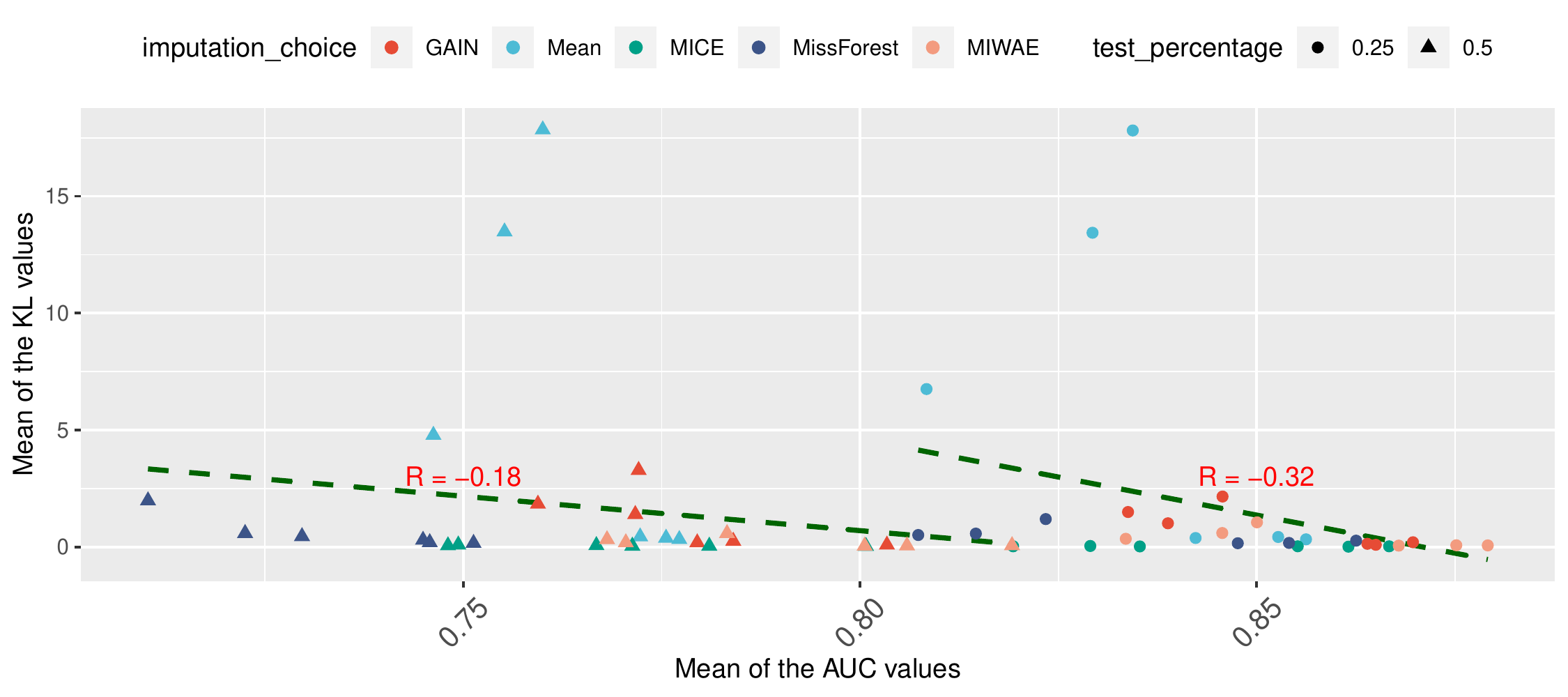}&
      \includegraphics[height=4cm,width=5cm]{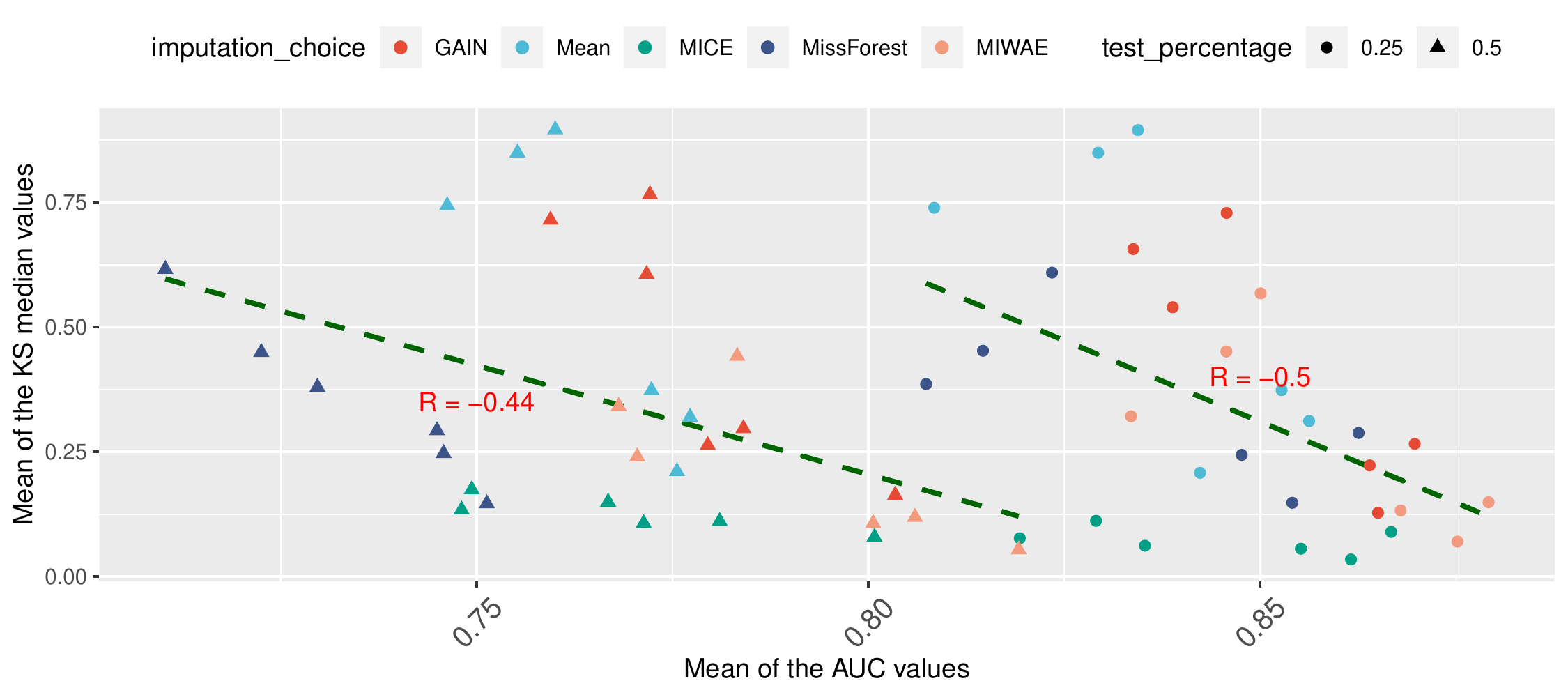}&
      \includegraphics[height=4cm,width=5cm]{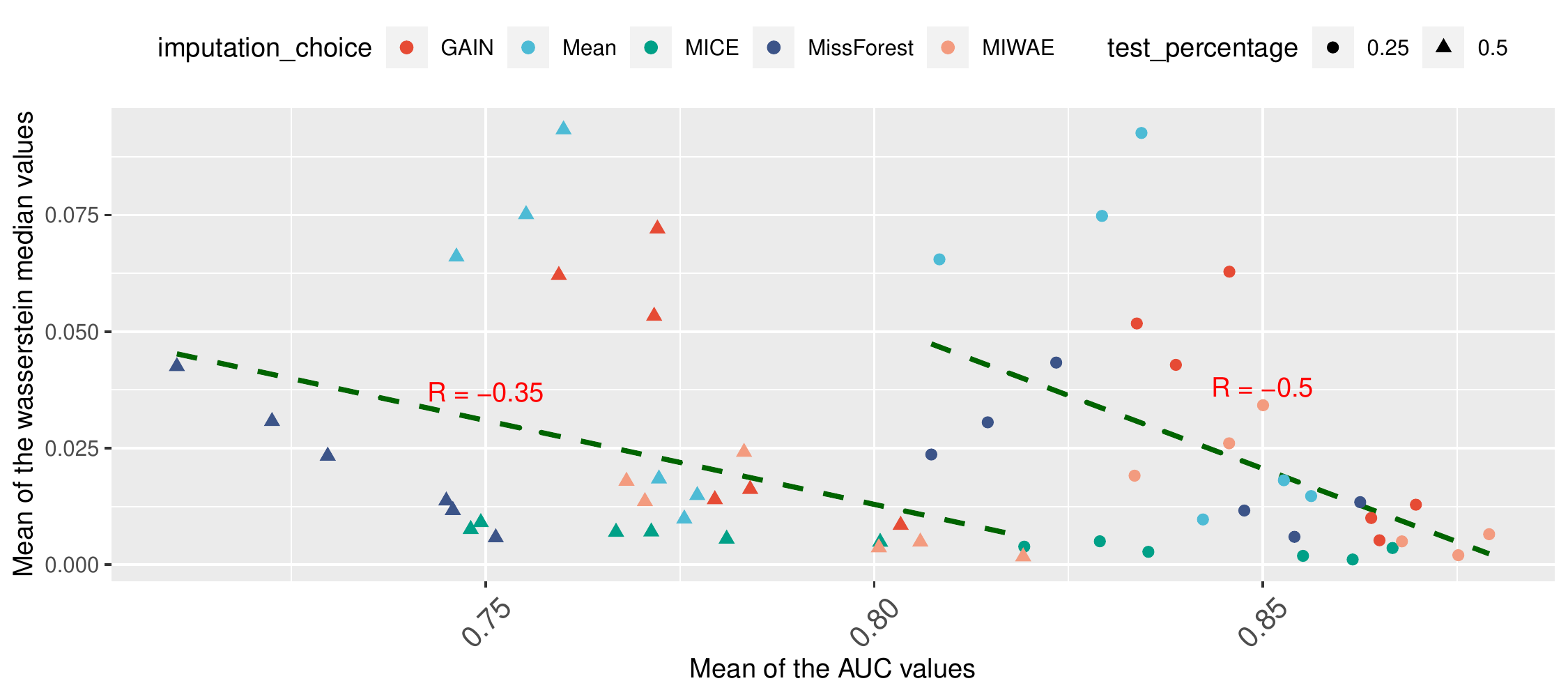}\\
      & (g) KL metric versus AUC &
      (h) KS metric versus AUC &
      (i) 2W metric versus AUC \\
    \end{tabular}
    \caption{The different imputation discrepancy metrics from classes A, B and C are shown against the downstream AUC value for the classification task of the Simulated dataset. Trend lines are shown for 25\% and 50\% test missingness separately.}
    \label{fig:corr_quality_syn}
\end{figure}

\begin{figure}[htb!]
     \centering
       \includegraphics[width=0.76\textwidth]{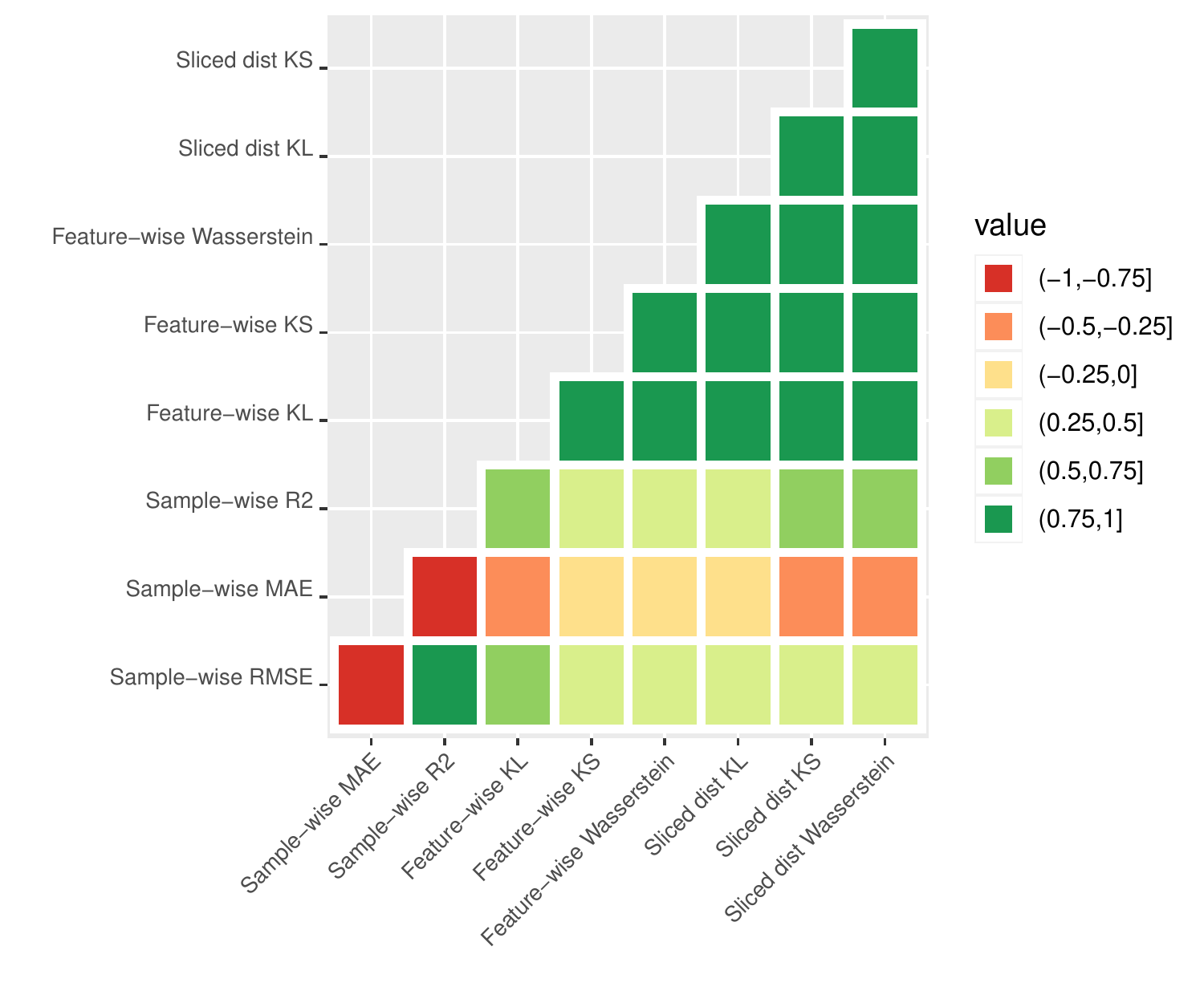}
 \caption{Correlation heatmap for all discrepancy metrics considered in this paper for the \textbf{Simulated} dataset. \label{fig:conc_metrics_syn}}
\end{figure}

\clearpage 
\noindent
\textbf{Link between imputation quality and downstream classification performance.}
In Figure~\ref{fig:corr_quality_mimic} and Supplementary Figure~\ref{fig:corr_quality_syn}, we plot the results for all class A, B and C imputation discrepancy statistics against the AUC of the downstream classification task. Our previous analysis has shown that the test missingness rate has a large influence on both imputation quality and downstream performance, so we display the correlations separately for 25\% and 50\% missingness
For the sample-wise statistics, we see a negative correlation between imputation discrepancy and downstream classifier performance only for the \textbf{Simulated} dataset. In the \textbf{MIMIC-III} dataset, we see no clear correlation between imputation quality, by any of these discrepancy scores, and the classifier performance. However, for the feature-wise and the proposed sliced Wasserstein classes, we see a negative correlation for all statistics in both datasets.
In Figure~\ref{fig:conc_metrics_mimic} and Supplementary Figure~\ref{fig:conc_metrics_syn} we give heatmaps showing the correlations between the nine different discrepancy statistics used for the \textbf{MIMIC-III} and \textbf{Simulated} datasets, respectively.
Within each class of discrepancy metric (for A, B and C), the measures are all correlated with one another, however the sample-wise metrics (class A) do not correlate significantly with any of the feature-wise (class B) or sliced Wasserstein distances (class C). There is also strong correlation between most of the class B and C metrics.\\

\noindent
\textbf{Impact of imputation quality on interpretability.}
For each classifier, we find that the best sliced Wasserstein distance ratio is always obtained for data imputed with MICE. For NGBoost and XGBoost, the worst distance ratio of the high performing models is for GAIN imputed data whereas, for Random Forest, this is for mean imputed data. 
For each of the 25 features in the \textbf{Simulated} dataset, we calculate the absolute value of the skew for the Shapley values and display these in Supplementary Figure~\ref{fig:interp_comp}. For the Random Forest classifier, we see that for 21/25 features, the absolute skew of the Shapley values for the MICE imputation is smaller than that of mean imputation. For NGBoost, we see the a smaller absolute skew in 19/25 features for MICE against GAIN. Finally, for XGBoost, we see a smaller absolute skew in 15/25 features for MICE against GAIN. 

\section{Discussion}

In this paper, we have highlighted the importance for machine learning and data science practitioners to reconsider the quality of the imputed data which is used to train a classifier. 
Each dataset we considered presents with different missingness rates and missingness types. In particular, the \textbf{NHSX COVID-19} and \textbf{Breast Cancer} datasets have data that is missing not at random, while the induced missingness to the \textbf{MIMIC- III} and \textbf{Simulated} datasets is missing completely at random (MCAR). Perhaps as a consequence of this, it was found that there is no particular classifier which performs best across all of the datasets and similarly, no particular imputation method leads to the best downstream classification performance. Even for the datasets with the same types of missingness, no optimal imputation or classification method emerges. Interestingly, classification models built on data imputed using mean imputation demonstrate the largest variance in the downstream performance, even though the imputed data is exactly the same in each of the multiple imputations.
%% anova
Using ANOVA, we found that performance is most influenced by the test missingness rate. This would suggest that when deploying a trained model to a new dataset, it is of primary importance to be conscious of the missingness rate of the new data, as classifier performance is very sensitive to it. ANOVA also highlights a minimal dependence of the downstream classification performance on the imputation method chosen. Therefore, we conclude that it is unlikely that it is possible to heuristically identify the optimal imputation and classification method for a particular dataset.

%%% imputation. quality
There is no accepted approach for evaluating the quality of an imputation method. Typically, imputation methods are evaluated using sample-wise discrepancy statistics such as RMSE, MAE and $R^{2}$, but this approach implicitly assumes that the aim of imputation is to recover the true \textit{value} of the missing datapoint, rather than recreating the correct \textit{distribution}. In our experiments, we find that the sample-wise discrepancy scores are not sufficient to assess the quality of the reconstruction of the distribution of imputed values. 
In fact, for MSE in particular, we have seen that it is optimal for imputations that give a very poor distribution match. Using MSE to evaluate the imputation quality is only statistically justified in the case of imputing data from a Gaussian distribution (minimizing MSE corresponds to maximizing the log-likelihood). Since the Gaussian assumption often doesn't hold in practice, MSE can be a very inaccurate measure of imputation quality. 
Beyond sample-wise discrepancy scores, some studies have considered the reconstruction quality of distributions for datasets on a feature-by-feature basis. However, in this study, we have shown how considering these marginal distributions alone is not sufficient to understand how well the overall data distribution has been reconstructed.
Echoing van Buuren \cite{van_buuren_flexible_2018}, \textit{``imputation is not prediction"}. The goal of an imputation method is to recover the correct distribution, rather than the exact value of each missing feature. However, the metrics used in the literature simply do not currently enforce this. In this paper, we introduced a class of discrepancy statistics, based on the sliced Wasserstein distance, which allow us to evaluate how well {\emph{all}} features have been reconstructed in a dataset.

Interestingly, although GAIN performs well in the sample-wise discrepancy scores, we see a poor performance in the discrepancy when measured feature-wise and with our proposed class of sliced Wasserstein distance-based scores. In fact, we find that the GAIN imputation method tends to perform in line with mean imputation and give a poor reconstruction of the underlying data distribution. It has been identified in \cite{vinas_deep_2021} that GAIN, which uses a generative adversarial approach for training, tends to rely mostly on the reconstruction error in the loss term (the mean square error) and the adversarial term contributes in a minimal way. 
The MICE imputation method has shown a clear dominance for recreating the distribution of the datasets as a whole. In both datasets for which it could be evaluated, and across all train and test missingness rates, it performs better than the competitor imputation methods. 
%%%% ratio
When evaluating the ratio between the sliced Wasserstein distance in the imputed data versus the original data, we found that, consistent with the earlier observations, an increased test missingness rate leads to a significant increase in the distance induced by imputation. However, an increase in the train missingness rate leads to a minimal increase in the induced distance ratio. We found that MICE outperforms the other imputation methods, inducing the smallest relative increase in distance with minimal variance. It also gave highly consistent imputation results, with minimal variance in the distances between the imputed and original values. 
%%%% outliers
Performing multiple imputations highlights that GAIN and MIWAE suffer from mode collapse in some of the rounds of imputation and can fall into outlying local minima, likely due to the highly non-convex problem they are solving. 
This is a problem common to all deep neural network architectures and serves as a reminder, for imputation methods, of the importance of using multiple imputation to overcome the risk of some poor quality imputations. 

%%% imp quality and downstream performance
We find that it is not necessarily the best quality imputation that leads to the best performing classification method. This observation could be explained by the fact that imputation does not add any information that was not present in the data set to begin with. Therefore, a powerful classifier may be able to extract all information relevant to the classification task regardless of the imputation quality. Secondly, we conjecture that an inaccurate imputation could in fact provide a form of regularization for the classifier. It has been shown that perturbing the training data with a small amount of Gaussian noise is equivalent to $L_2$ regularization \cite{bishop_training_1995} and can be beneficial to the performance in a supervised learning tasks. In the same spirit, an inaccurate imputation method resulting in noisy imputed values could provide a regularizing effect beneficial to performance in the downstream classification task.
%% corr
We also find that not only are the common discrepancy scores used by the community, i.e. the sample-wise statistics (RMSE, MAE, $R^{2}$), uncorrelated from the distributional discrepancy metrics (of classes B and C) but that they are also disconnected from the downstream classification performance of the model. This highlights how important it is to consider additional statistics when measuring imputation quality, not simply the sample-wise statistics, as the distribution discrepancy scores occupy a completely orthogonal space to the sample-wise ones.
Importantly, we find a correlation between the proposed class of, sliced Wasserstein distance derived, discrepancy scores and the downstream model performance suggesting that a link has been forged from the imputed data to the downstream model performance. Given this, we would suggest that instead of focusing on optimising performance by considering the best combination of imputation method and classifier, attention should shift towards optimising the imputation quality in terms of how well the distribution is reconstructed..
%%% interpretability
In our assessment of how the imputation quality can affect the model interpretability downstream, we find that, for comparably high performing classifiers, those fit to poorly imputed data give misleading feature importance's in the downstream classification task. Any features judged as important from these models are therefore compromised. This is crucial to appreciate, especially where models are fit to clinical data, as we may draw incorrect conclusions about the influence of particular features on patient outcomes. Overall, this suggests that the quality of imputation does feed through to the downstream model interpretability.

%%% summary
In addition to main aims of this paper, through this work, we have also identified some issues that need the attention of the imputation community, such as how categorical and ordinal variables should be correctly imputed. For example, if a categorical variable is one-hot-encoded then a valid imputation must be in $\{0, 1\}$. However, most imputation methods will give a value in the range $[0,1]$ and these must be post-processed. It is not clear how this should be performed. Similarly, if a category with multiple values is one-hot encoded, e.g.\ nationality, then only one of these variables should equal one but no imputation method enforces this.
Issues were also identified, and fixed, in the public code releases of the GAIN and MIWAE imputation methods and so, to improve the quality of future benchmarking of imputation methods, we release our codebase at \texttt{github.com/\textbf{[updated upon publication]}}. This allows for rapid computation of the results for several imputation methods across incomplete datasets with data preprocessing, partitioning and analysis all built in. This should serve as a sandbox for the development, and fair evaluation, of new imputation methods. 

It is our hope that by standardising the imputation methods, multiple imputation, classification and analysis pipelines for determining the imputation quality, future studies have all the tools necessary to fit classifiers to incomplete data that are not only achieving high performance, but are built on high quality imputed data. This study highlights how these concepts are not separable and if we do not understand the quality of the imputation, the model fit to the data is compromised.

\subsection*{Acknowledgements}
There is no direct funding for this study, but the authors are grateful for the following indirect funding: European Research Council (No. 716063) (T.S., J.T.), Academy of Finland (No. 317680) (T.S., J.T.), BusinessFinland (T.S., J.T.), The Sigrid Juselius Foundation (T.S., J.T.), the EU/EFPIA Innovative Medicines Initiative project DRAGON (101005122) (T.S., M.R., J.S. J.G., S.D., E.S., AIX-COVNET, C.-B.S.), AstraZeneca (M.R., P.T., M.P.), the Trinity Challenge (M.R., C.-B.S.), the EPSRC Cambridge Mathematics of Information in Healthcare Hub EP/T017961/1 (M.R., J.H.F.R., J.A.D.A, C.-B.S.), the Cantab Capital Institute for the Mathematics of Information (J.S., C.-B.S.). Aviva (J.S.), the European Research Council under the European Union’s Horizon 2020 research and innovation programme grant agreement no. 777826 (M.T., C.-B.S.), the Alan Turing Institute (M.T., C.-B.S.), Fundaci{\'{o}}n Rafael del Pino (R.V.), Wellcome Trust (J.H.F.R.), The Mark Foundation for Cancer Research (E.S.), Cancer Research UK Cambridge Centre (C9685/A25177) (E.S., C.-B.S.), British Heart Foundation (J.H.F.R.), the NIHR Cambridge Biomedical Research Centre (J.H.F.R.), HEFCE (J.H.F.R.), the Finnish Cancer Organizations (A.S.R.) and the Jane and Aatos Erkko Foundation (A.S.R.).
In addition, C.-B.S. acknowledges support from the Leverhulme Trust project on ‘Breaking the non-convexity barrier’, the Philip Leverhulme Prize, the EPSRC grants EP/S026045/1 and EP/T003553/1 and the Wellcome Innovator Award RG98755.
Finally, the AIX-COVNET collaboration is also grateful to Intel for financial support.

\subsection*{AIX-COVNET}

Michael Roberts$^{2,3}$, S{\"{o}}ren Dittmer$^{2,14}$, Ian Selby$^{6}$, Anna Breger$^{2,15}$, Matthew Thorpe$^{4}$, Julian Gilbey$^{2}$, Jonathan R. Weir-McCall$^{6,16}$, Effrossyni Gkrania-Klotsas$^{17}$, Anna Korhonen$^{18}$, Emily Jefferson$^{19}$, Georg Langs$^{20}$, Guang Yang$^{21}$, Helmut Prosch$^{20}$, Jacobus Preller$^{17}$, Jan Stanczuk$^{2}$, Jing Tang$^{1}$, Judith Babar$^{17}$, Lorena Escudero Sánchez$^{6}$, Philip Teare$^{3}$, Mishal Patel$^{3,7}$, Marcel Wassin$^{22}$, Markus Holzer$^{22}$, Nicholas Walton$^{23}$, Pietro Li{\'{o}}$^{5}$, Tolou Shadbahr$^{1}$, James H. F. Rudd$^{9}$, Evis Sala$^{6}$ and Carola-Bibiane Schönlieb$^{2}$.\\

\noindent
${}^{15}$ Faculty of Mathematics, University of Vienna, Austria
${}^{16}$ Royal Papworth Hospital, Cambridge, Royal Papworth Hospital NHS Foundation Trust, Cambridge, UK
${}^{17}$ Addenbrooke’s Hospital, Cambridge University Hospitals NHS Trust, Cambridge, UK. 
${}^{18}$ Language Technology Laboratory, University of Cambridge, Cambridge, UK.
${}^{19}$ Population Health and Genomics, School of Medicine, University of Dundee, Dundee, UK.
${}^{20}$ Department of Biomedical Imaging and Image-guided Therapy, Computational Imaging Research Lab Medical University of Vienna, Vienna, Austria.
${}^{21}$ National Heart and Lung Institute, Imperial College London, London, UK.
${}^{22}$ contextflow GmbH, Vienna, Austria. 
${}^{23}$ Institute of Astronomy, University of Cambridge, Cambridge, UK. 

%\newpage

\typeout{}  % just to ensure that latexmk sees the bibliography
\bibliography{imputation}

%{\color{red}{\noindent LaTeX formats citations and references automatically using the bibliography records in your .bib file, which you can edit via the project menu. Use the cite command for an inline citation, e.g.\  \cite{Hao:gidmaps:2014}.}}

%{\color{red}{For data citations of datasets uploaded to e.g.\ \emph{figshare}, please use the \verb|howpublished| option in the bib entry to specify the platform and the link, as in the \verb|Hao:gidmaps:2014| example in the sample bibliography file.}}
\newpage

\section*{Supplementary Materials}

\subsection*{Supplementary Information}\vspace{1em}

\subsubsection*{Comparison between MSE and Wasserstein distance as discrepancy scores}\vspace{1em}

In Figure~\ref{fig:AB} below, we show an example data distribution, created with the \texttt{make\_classification} function, similar to the \textbf{Simulated} dataset. We evaluate the MSE and Wasserstein Distance between true and imputed data for (b) a method which optimises for MSE and (c) using MICE. It is clear that optimising for MSE leads to a poor distribution reconstruction, however MICE reflects the underlying distribution well. This simple example demonstrates how MSE is insufficient for asessing imputation quality, as although MICE is clearly preferable in this case, it has a poor MSE.
\begin{figure}[htb!]
    \centering
    \begin{tabular}{|P{5cm} | P{5cm} |P{5cm} |}
    \hline&&\\[-0.05in]
          (a) True Data Distribution & (b) Imputation optimising mean square error & (c) Imputation with the MICE method \\[0.10in]
              \hline&&\\[-0.05in]
     \includegraphics[width=4cm]{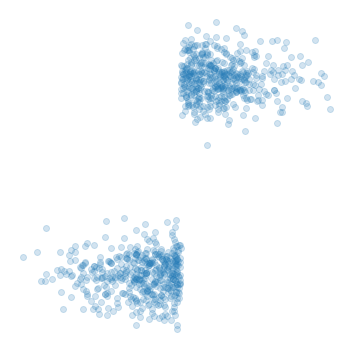}&
      \includegraphics[width=4cm]{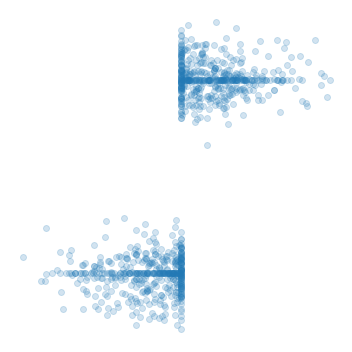}&
     \includegraphics[width=4cm]{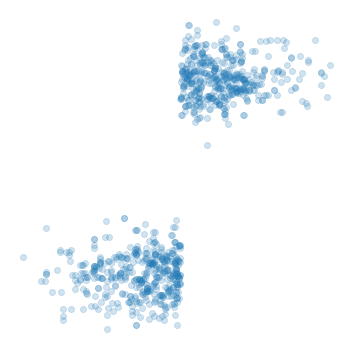}\\[\medskipamount]
         \hline&&\\[-0.05in]
        & Mean Square Error = \textbf{\textcolor{cyan}{0.9074}} & Mean Square Error = \textbf{\textcolor{red}{1.6533}} \\
         & Wasserstein Distance = \textbf{\textcolor{red}{2.68$\times 10^{-4}$}} & Wasserstein Distance = \textbf{\textcolor{cyan}{1.55$\times 10^{-4}$}} \\[0.10in]
             \hline
    \end{tabular}
    \caption{Comparing imputation for MSE and MICE. In (a) we see a simulated dataset, (b) shows the optimal imputation against the mean square error (MSE) and (c) shows the MICE imputation result. The MSE and Wasserstein distances are quoted for each.}
    \label{fig:AB}
\end{figure}

\subsubsection*{Dataset Description and Preprocessing Details} \vspace{1em}

All of the datasets used in this study are publicly available (upon reasonable request for \textbf{MIMIC-III} and \textbf{NHSX COVID-19}). For the \textbf{MIMIC-III} and \textbf{Breast Cancer} datasets, we include some details below for how the \textbf{MIMIC-III} and \textbf{Breast Cancer} datasets were preprocessed for use in this study. The code for performing this preprocessing is also available in the codebase. 

\textbf{MIMIC-III.} In preprocessing the MIMIC-III\cite{johnson_mimic-iii_2016} dataset , we only considered information for the first ICU admission of patients, and restricted to patients who are over 15 years old, spent at least 3 days in the ICU in the data collection period and were admitted as `Emergency' or `Urgent' cases. Furthermore, we extracted data on seven clinical variables: blood pressure (systolic, diastolic and mean), heart rate, oxygen saturation, respiratory rate and temperature; these were the only variables recorded for over 50\% of the patients. In our preprocessing of the dataset, we only considered data from the first 10 days in the ICU (or the whole ICU stay if shorter), and excluded any patients who had fewer than 5 observations in any of aforementioned variables. We then calculated the mean and standard deviation for each of these seven variables, giving a total of 14 numerical variables per patient. The outcome variable we use is the survival of patients in the 30 days after admission to the ICU. Our preprocessing code is based on MIMIC-Extract \cite{wang_mimic-extract_2020} and is available at \textbf{[shared upon publication]}. Due to the size of the dataset (and the number of computations we ultimately perform), we then select one third of the patients randomly, resulting in a dataset with 7214 patients.

\textbf{Breast Cancer.} This dataset is derived from an oncology dataset collected at Memorial Sloan Kettering Cancer Center between April 2014 and March 2017 \cite{razavi_genomic_2018}. The biopsy samples were collected prior or during or after treatment from different primary or metastatic sites which leads to several samples for the each patient. We only considered the first occurrence of the patients ID. The dataset, consisting of 16 different features, is assembled using data stored in three different sub-datasets: \verb+data_clinical_sample+, \verb+data_clinical_patient+, and \verb+breast_msk_2018_clinical_data+ from Razavi et al.\cite{razavi_genomic_2018}. Specifically, the features we consider are the `Fraction Genome Altered' and `Mutation Count', taken from \verb+breast_msk_2018_clinical_data+, `ER Status of the Primary', `Invasive Carcinoma Diagnosis Age', `Oncotree Code', `PR Status of the Primary' and `Stage At Diagnosis', taken from \verb+data_clinical_sample+, and `Metastatic Disease at Last Follow-up', `Metastatic Recurrence Time', `M~Stage', `N~Stage', `T~Stage', `Overall Patient HER2 Status', `Overall Patient HR Status' and `Overall Patient Receptor Status' from \verb+data_clinical_patient+. 
The features of this dataset are of several different types, `ER Status of the Primary', `Metastatic Disease at Last Follow-up', `M~Stage', `Overall Patient HER2 Status', `Overall Patient HR Status' and `PR Status of the Primary' are binary; `N~Stage', `T~Stage', `Stage At Diagnosis' and `PR Status of the Primary' are ordinal; `Oncotree Code', and `Overall Patient Receptor Status' are multilevel categorical, and `Fraction Genome Altered', `Invasive Carcinoma Diagnosis Age', `Metastatic Recurrence Time' and `Mutation Count' are numerical. 
The outcome variable we choose was `Overall Survival Status' for the classification task. Moreover, features `N~Stage' and `T~Stage' are described with 14 and 15 different levels respectively, and some of the levels only have only one or two samples. To ensure features are meaningfully represented, we only consider the parent family of each level which results in only 4 different levels for each feature.

All the ordinal variables are treated in the same manner as numeric variables if the imputation method was not capable of addressing the ordinal variables. The majority of categorical variables are binary, or have only two options (e.g.\ sex, death) and are simply encoded to zero and one and treated as numeric variables or binary (if the imputation method was capable of modelling the binary variables).

\subsubsection*{Descriptions of the imputation methods} 

Below, we give a detailed summary for each imputation method used in this paper, in particular, we detail how the categorical variables are encoded for each method.\\

\noindent
\textbf{Mean Imputation} is one of the simplest imputation methods, in which the missing values are replaced with the sample mean. This is a single imputation method, as there is no stochasticity in its computation. Mean imputation was performed using the \texttt{SimpleImputer} implementation in the Python package \texttt{scikit-learn} \cite{pedregosa_scikit-learn_2011}. Mean imputation has no special treatment for the multi-level categorical variables, therefore we one-hot encode all the multi-level categorical features as part of the preprocessing.\\

\noindent
\textbf{MissForest} is an iterative imputation method which employs a random forest (a non-parametric model) for non-linear modeling of mixed data types. Therefore, MissForest is applicable for both numerical, categorical and ordinal data, whilst making few assumptions about the structure of the data \cite{stekhoven_missforestnon-parametric_2012, breiman_random_2001}. MissForest imputation works by initially training a random forest using the observed (i.e.\ not missing) data values to predict the missing entries of a given variable. This procedure continues until either a maximum number of iterations is reached or the out-of-bag error increases \cite{stekhoven_missforestnon-parametric_2012, choudhury_missing_2020}. MissForest imputation was performed using the Python package \texttt{missingpy} \cite{bhattarai_missingpy_2018}. In our implementation, the multi-level categorical variables in \textbf{Breast Cancer} are specified to the algorithm.\\

\noindent
\textbf{Multivariate Imputation by Chained Equations (MICE)}, also known as `Fully Conditional Specification' \cite{van_buuren_multiple_2007} or `Sequential Regression Multivariate Imputation' \cite{raghunathan_multivariate_2001}, is an imputation method which iteratively imputes the missing values of each variable one at the time using a method based on the type of the corresponding variable (such as linear regression) \cite{van_buuren_flexible_1999, choudhury_missing_2020, enders_multilevel_2016}. 
In its initialisation, MICE sets the missing values using values derived from the observed values, for example the mean of the feature or a random observed value. Next, a model is fit which takes one feature as the output/target variable and all other features are used as input variables. This model is then applied to predict the target variable and those which correspond to the missing values are replaced. This process is repeated, updating all features in turn. This gives a dataset in which all missing values of the features have been updated once. This whole process can be repeated many times until the imputed values converge to within some error \cite{ding_comparison_2012, white_multiple_2011, tran_effective_2018}.
We perform our computation using the R package \texttt{mice} \cite{van_buuren_mice_2011}. In our model, all the numerical variables are initially imputed using predictive mean matching (pmm)\cite{van_buuren_mice_2011}. In the \textbf{Breast Cancer} dataset, there are variables with multi-level categorical values. These variables are one-hot encoded and imputed using logistic regression along with binary variables. Finally, the ordinal variable are imputed using a proportional odds model. \\

\noindent
\textbf{Generative Adversarial Imputation Networks (GAIN)} \cite{yoon_gain_2018} is an imputation method whose framework is based on generative adversarial networks\cite{goodfellow_generative_2014}. This approach adversarially trains a pair of neural networks, first a generator whose purpose is to generate realistic samples from the training distribution, and a discriminator which aims to identify whether a sample is from the training distribution. Training in this adversarial manner ensures that the generator creates more realistic samples as it tries to ``fool'' the discriminator. 
For the GAIN method, a binary mask is also provided to the generator function where the zeros correspond to the missing values and ones indicate an observed value. A ``hint'' matrix is generated from this mask for which some proportion of entries (decided by a parameter) are set to 0.5, i.e.\ for these entries we do not know if the value is observed or missing.
Initially, all missing values are replaced by random noise sampled from a normal distribution. The generated dataset is input to the discriminator along with the hint matrix. The task of discriminator is to predict the mask, i.e.\ identify both the observed values and imputed values \cite{wang_are_2022, yoon_gain_2018}. 
For GAIN, the official implementation provided in the paper \cite{yoon_gain_2018} is used. As GAIN cannot directly use multi-level categorical variables as input, in our implementation these are one-hot encoded. \\

\noindent
\textbf{Missing Data Importance-Weighted Autoencoder (MIWAE)} is an imputation method proposed by Mattei and Frellsen  \cite{mattei_miwae_2019} which uses a deep latent variable model (DLVM) \cite{kingma_auto-encoding_2014, rezende_stochastic_2014} for the imputation of the missing values in a given dataset. This method builds upon the importance-weighted autoencoder (IWAE) \cite{burda_importance_2016}, which aims to maximise a lower bound of the log-likelihood of the observed data. The lower bound of IWAE is a $k$-sample importance weighting estimate of the log-likelihood and is a generalization of the variational lower bound used in variational autoencoders \cite{kingma_auto-encoding_2014} (which corresponds to the case of $k=1$).
In MIWAE, the lower bound of IWAE is further generalized to the case of incomplete data (and coincides with the IWAE bound in the case of complete data). During training, the missing data is replaced with zeros before being passed into the encoder network to obtain the latent codes. The codes are passed to the decoder network and the output is compared with the observed data (only in non-missing dimensions) to compute the aforementioned lower bound. 
Once the model is trained, multiple imputation is possible via \textit{sampling importance resampling} (SIR) from the trained model. 
We used the official Python implementation \cite{mattei_miwae_2019} in our study. All the multi-level categorical variables are first one-hot encoded and the ordinal features are coded with numerical values before imputation with MIWAE. 

\subsubsection*{Descriptions of the classification methods}  

In the following, we briefly describe each of the classification methods used in this paper. Moreover, we detail the hyperparameters, including the ranges of values, that are tuned in the benchmarking exercise for obtaining the optimal performing classifier.\\

\noindent
\textbf{Logistic Regression.} This is simple and efficient classification method with high prediction performance for datasets that have linearly separable classes. This method use a logistic function $\frac{1}{1+e^{-(mx+b)}}$ to generate a binary output, where $x$ is the input and $m$ and $b$ are learned \cite{cox_regression_1958}. We used the \texttt{scikit-learn} library implementation of the logistic regression classifier. The only hyperparameter to tune is the maximum number of iterations used, we search over $\{50, 100, 150, 200, 250\}$.\\

\noindent

\textbf{Random Forest.} This classifier is one of the most popular and successful algorithms. It was proposed by Breiman \cite{breiman_random_2001} and involves creating an ensemble of randomised decision trees whose predictions are aggregated to obtain the final results. In training, for each tree $m$ samples are randomly selected by bootstrapping. Then, this subset of the dataset is used to train a randomized tree. This procedure is repeated $k$ times. The Random Forest is the aggregation of these $k$ decision trees \cite{breiman_consistency_2004}. For our study, we used the Random Forest implementation in open source \texttt{scikit-learn} library \cite{pedregosa_scikit-learn_2011} in Python. The Random Forest can be applied to variety of different prediction problems, and few parameters need to be tuned. For the number of estimators, we tried 8 different values in the range $[20,90]$ with equal step width as 10. For the maximum depth, we limited our search to $\{3,4\}$. For minimum sample split and the minimum samples in each leaf, we search over the set $\{2,3,4\}$. This results in 144 different hyperparameter combinations.  \\

\noindent
\textbf{XGBoost.} This method, proposed by Chen and Guestrin \cite{chen_xgboost_2016}, is an efficient gradient tree boosting algorithm which builds a chain of `weak learner' trees to give a high-performing classification or regression model. In this method, a base model is constructed by training an initial tree. Then, the second tree is obtained through combination with initial tree. This procedure is repeated until the maximum number of trees is reached. These additive trees are selected based on a greedy algorithm, in each step adding the tree that most minimizes the loss function. Therefore, the training of the model is in additive manner \cite{chen_xgboost_2016}. For the implementation of XGBoost, we used the official \texttt{xgboost} Python package provided in \cite{chen_xgboost_2016}. For the maximum depth of trees we consider 3 different values $\{3, 4, 5\}$, and for the number of subsamples we selected among 6 different values in the range $[0.5, 1]$ with equal step size  $0.1$. The number of the trees are selected from 8 different values, in the range $[50, 400]$ with equal step size 50. This results in 144 different hyperparameter combinations.\\

\noindent
\textbf{NGBoost.} This recent method, proposed by Duan et al. \cite{duan_ngboost_2020}, gives probabilistic predictions for the outcome variable $y$ for input $x$. It assumes there is an underlying probability distribution $P_{\theta}(y|x)$ with a parametric form, described by $\theta(x)$. This method considers natural gradient boosting, by replacing the loss function with a scoring rule. This scoring rule, compares the estimated probability distribution to the observed data, by using a predicted probability distribution $P$ and the observed value y (outcome). The proper scoring rule $S$ returns the best score for the true distribution of the outcomes. The parameter that minimize value of the score rule is obtained through natural gradient descend \cite{martens_new_2020}. For the implementation of the NGBoost method, we used provided code on the official repository at \texttt{https://github.com/stanfordmlgroup/ngboost}.
An NGBoost model has three key hyperparameters which can be tuned, namely the learning rate, the minibatch fraction and the number of estimators.
For the learning rate we used 4 different values $\{0.0001, 0.001, 0.01, 0.1\}$. For the minibatch fraction we used 6 different values in the range $[0.5, 1]$ with equal step size $0.1$. The number of the estimator is selected from 8 different values in the range $[50, 400]$ with equal step size 50. This results in 192 different hyperparameter combinations.    \\

\noindent
\textbf{(Artificial) Neural Network (NN)}.  The artificial neural network we employ is a multi-layer perceptron (MLP) \cite{hastie_elements_2009} consisting of layers of neurons. Each neuron is assigned a value, based on a weighted combination the values for neurons in the previous layer. Activation functions are employed to introduce non-linearities into the weighted sum calculations. The training of the NN is through backpropagation \cite{rumelhart_learning_1986}, where the weights of the edges between each neuron are adjusted to minimize the error between the true labels and output. We use the ReLu \cite{agarap_deep_2019} activation function and the ADAM \cite{kingma_adam_2015} optimiser method with binary cross-entropy as the loss function. We used the implementation in the open source \texttt{scikit-learn} library \cite{pedregosa_scikit-learn_2011} in Python. 
An NN has three key hyperparameters which must be tuned, specifically, the initial learning rate for the optimizer, the number of the hidden layers, and the number of the neurons in each hidden layer. The initial learning rate selected from 3 different values $\{0.001, 0.01, 0.1\}$, the number of the hidden layers is set to one of the 3 values $\{1, 2, 3\}$. 5 different values are considered for the number of the neurons in each hidden layer from the range $[20,100]$ with equal step size 20. This results in 45 different hyperparameters combinations.

\subsubsection*{Data partitioning and hyperparameter selection}

\begin{figure}[htb!]
  \includegraphics[width=\linewidth]{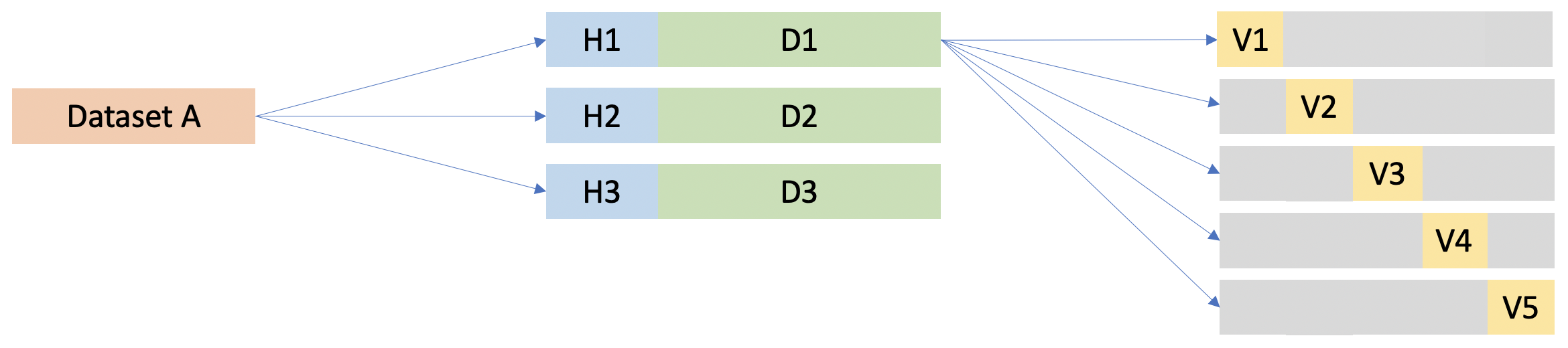}
  \caption{A schematic illustrating the hierarchical dataset split. Key: H = Holdout, D = Development, V = Validation.}
  \label{fig:dataset_splitting}
\end{figure}

We partition each dataset at two levels as shown in Figure~\ref{fig:dataset_splitting}. In the first level, we randomly partition the dataset into three holdout sets, each consisting of one third of the samples. These are used for reporting the performance of each imputation method and classifier. Each holdout set has a complementary development set and at the second level of partitioning, we divide the development set into five non-overlapping cohorts.
We use five-fold cross-validation on the development set to select the optimal hyperparameters for each combination of imputation method and classifier using the mean area under the receiver operating characteristic curve (AUC) over the five validation folds. In Table~\ref{tab:hparams}, we detail the hyperparameters combinations considered for each of the classifiers. 
\begin{table}[htb!]
\begin{center}
\begin{tabularx}{0.8\textwidth} { l|X }
\textbf{Classifier ($K$)} & \textbf{Tuned Hyperparameters} \\
\hline
Random Forest (144) & Number of trees (8), maximum depth of trees (2), minimum samples needed for splits in each individual leaf (3), minimum number of samples in each leaf (3).\\
\hline
XGBoost (144) & Number of the trees (8), maximum depth of each tree (3), subsample of the training instances used in each iteration of boosting (6).\\  
\hline
NGBoost (192) & Number of estimators (8), learning rate (4), subsample of the training instances used in each iteration of boosting (6). \\
\hline
Artificial Neural Network (45) & Number of neurons per layer (5), number of hidden layers (3), learning rate (3). \\
\hline
\end{tabularx}
\end{center}
\caption{Each classifier is fit with the $K$ hyperparameter combinations shown above. The key hyperparameters for each model were identified and the number in brackets indicates how many values were tested for each hyperparameter. \label{tab:hparams}}
\end{table}

\subsubsection*{Description of the ANOVA analysis}

In our multi-factor ANOVA analysis, for the \textbf{MIMIC-III} dataset we included the missingness rate of the development data and holdout data as separate factors to allow us to evaluate their individual effects. However, the \textbf{Breast Cancer} and \textbf{NHSX COVID-19} datasets are real datasets with their inherent missingness rates. Therefore, the missingness rate is introduced as a factor (i.e. categorical 25\%, 50\%, inherent) to the ANOVA model. 

\subsubsection*{Description of the sample-wise and feature-wise discrepancy measures}

In this paper we use nine statistics to measure the discrepancy between the imputed and true data. These fall into three classes; class A are sample-wise, class B are feature-wise and class C are derived from the sliced Wasserstein distances. For all methods, the data in the development and holdout sets are normalised to  mean zero and unit standard deviation (SD) using the development set mean and SD.\\

\noindent
\textit{Sample-wise metrics and their implementation.}
We briefly describe the sample-wise statistics used in the paper, along with giving details of the implementations used:
\begin{enumerate}
    \item[A1.]{Root MSE (RMSE). This is simply the square root of the mean square error, which is the average of the squared discrepancy errors between imputed and original samples. In this paper, we use the function \texttt{mean\_squared\_error} implemented in \texttt{sklearn} and take the square root of it.}
    \item[A2.]{Mean absolute error (MAE). This is the average absolute difference between the imputed and the original values. In this paper, we use the function \texttt{mean\_absolute\_error} implemented in \texttt{sklearn}.}
    \item[A3.]{$R^{2}$. This is the coefficient of determination, implemented in \texttt{sklearn} as \texttt{r2\_score}, which measures the proportion of the variation in the imputed values that is predictable from the original values. This is expressed as a percentage. Note that this is not a metric.}
\end{enumerate}

\noindent
\textit{Feature-wise metrics and their implementation.}
We briefly describe the distribution comparison measures used in the paper, along with giving details of the implementations used:
\begin{enumerate}
    \item[B1.] Kullback-Leibler (KL). The KL divergence measures how different two probability distributions are from one another. This is implemented in Python using the standard calculation shown in \texttt{kl.py} in our codebase.
    \item[B2.] Kolmogorov-Smirnov (KS). The KS test is used to assess whether two one-dimensional probability distributions differ. We use the function \texttt{ks\_2samp} in the Python package \texttt{scipy.stats}.
    %\item[B3.] Cram{\'{e}}r-von-Mises (CvM). The CvM criterion is used to assess how well the cumulative density functions of the imputed and original datasets fit to one another. We use the function \texttt{cramervonmises\_2samp} in the Python package \texttt{scipy.stats}.
    \item[B3.] Two-Wasserstein (2W). The 2W distance \cite{villani_optimal_2009} also measures the distance between two probability distributions using optimal transport. We use the function \texttt{emd2\_1d} in the \texttt{POT} package in Python.
\end{enumerate}

\subsubsection*{Outlier Analysis Details}

\begin{figure}[htb!]
    \centering
    \begin{tabular}{cccc}
      \includegraphics[width=4cm]{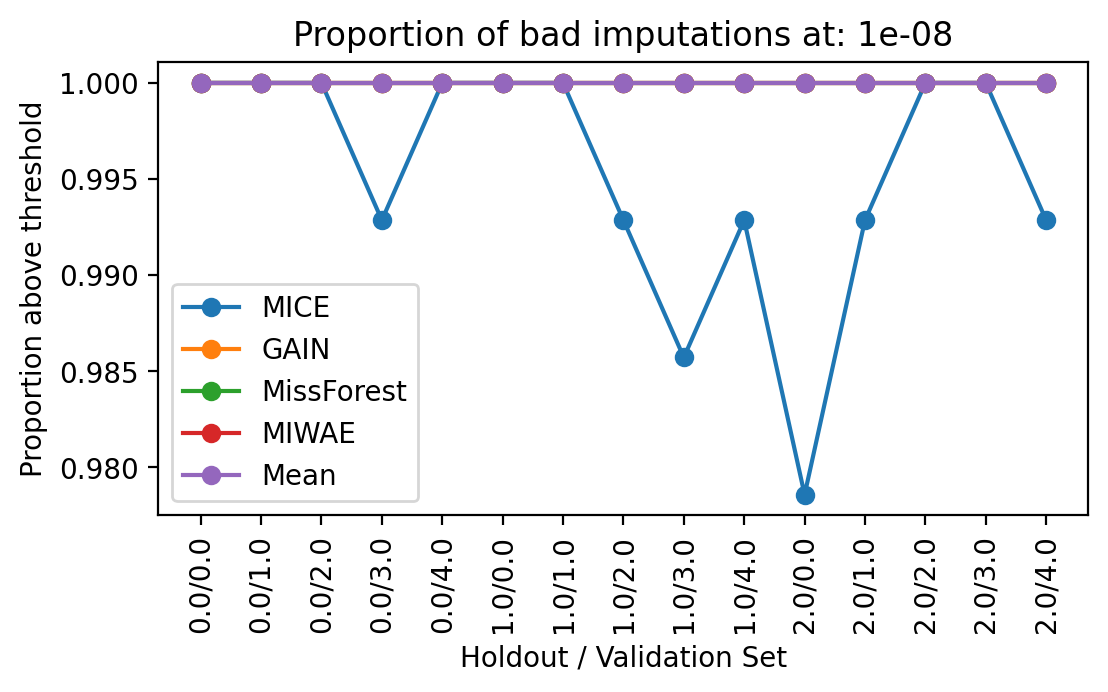}&
      \includegraphics[width=4cm]{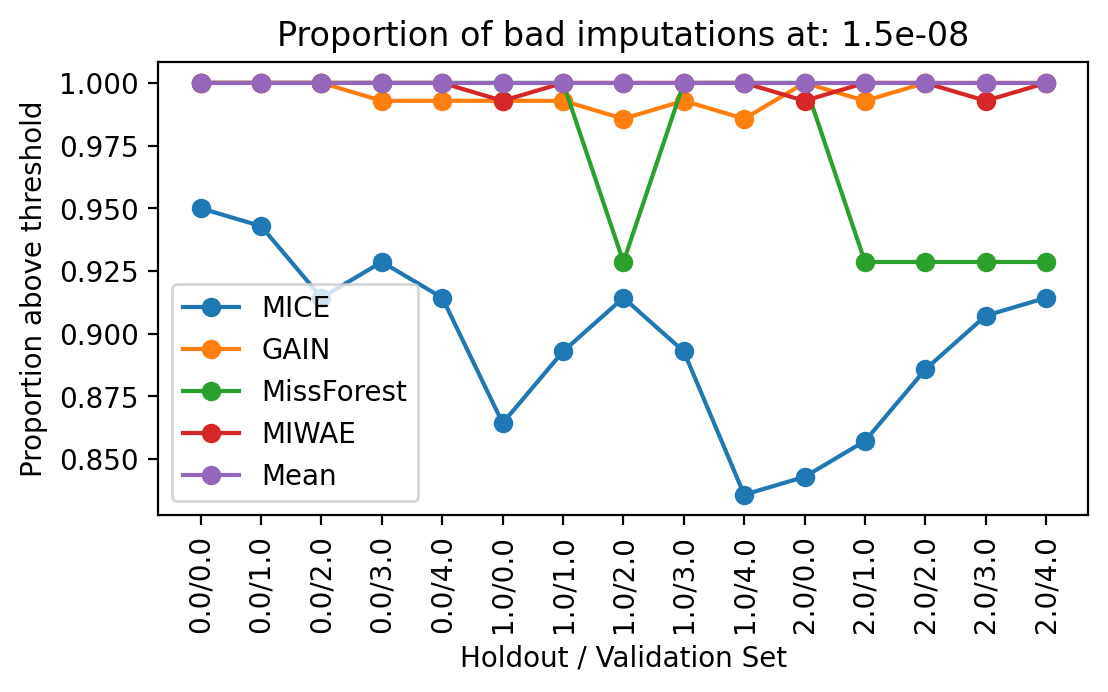}&
      \includegraphics[width=4cm]{outliers/1e-07.png}&
      \includegraphics[width=4cm]{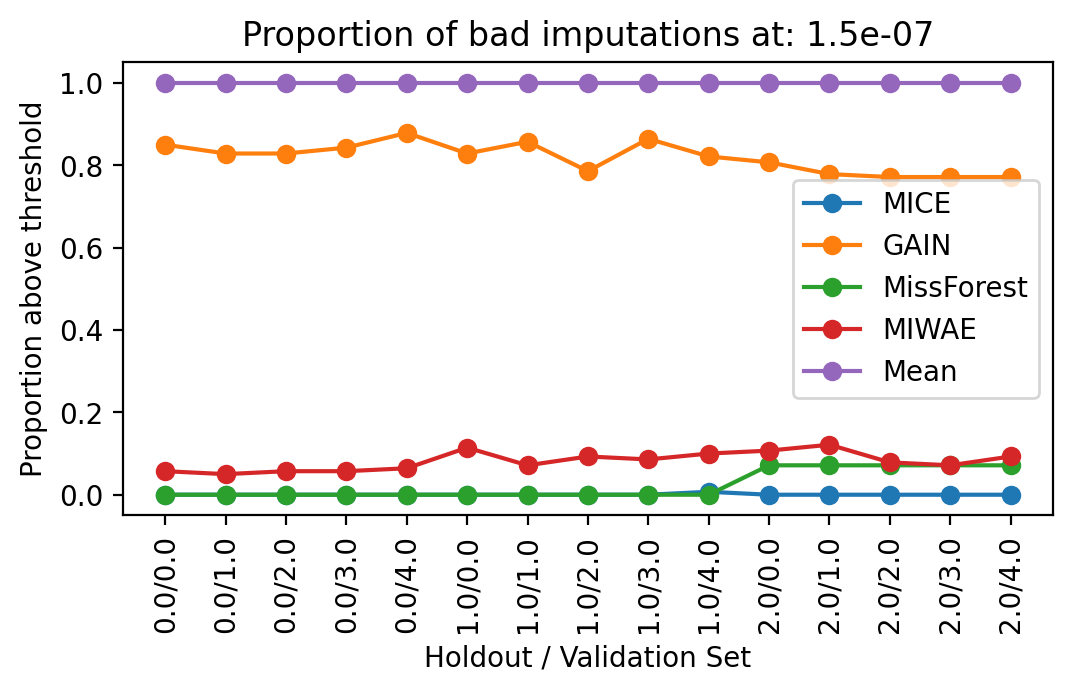} \\
      \includegraphics[width=4cm]{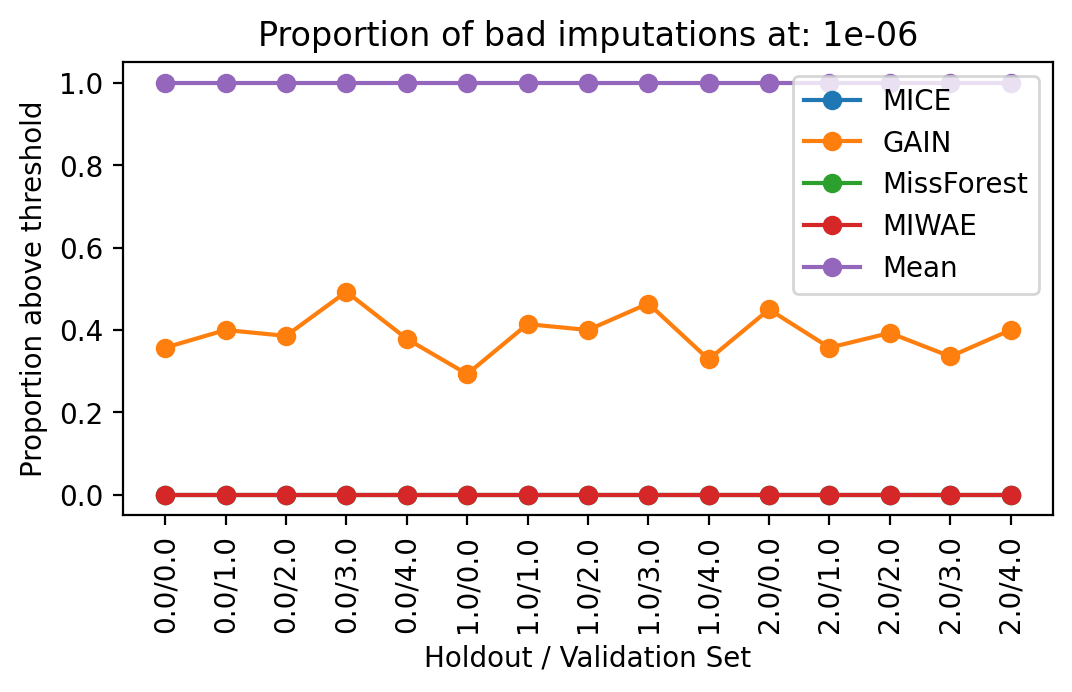}&
      \includegraphics[width=4cm]{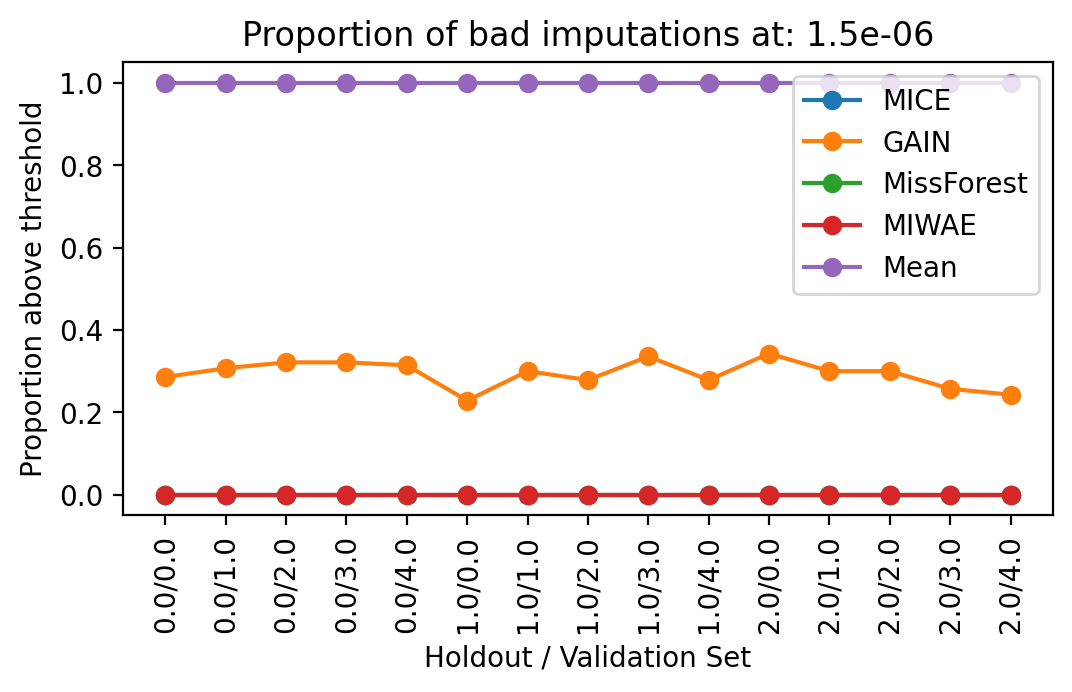}&
      \includegraphics[width=4cm]{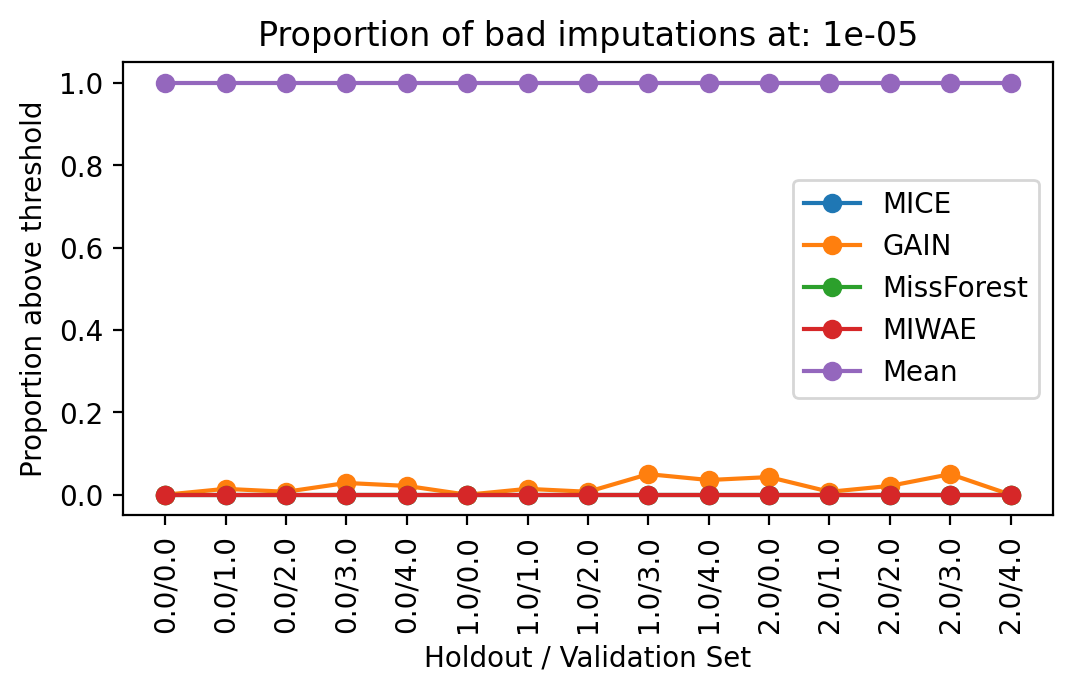}&
      \includegraphics[width=4cm]{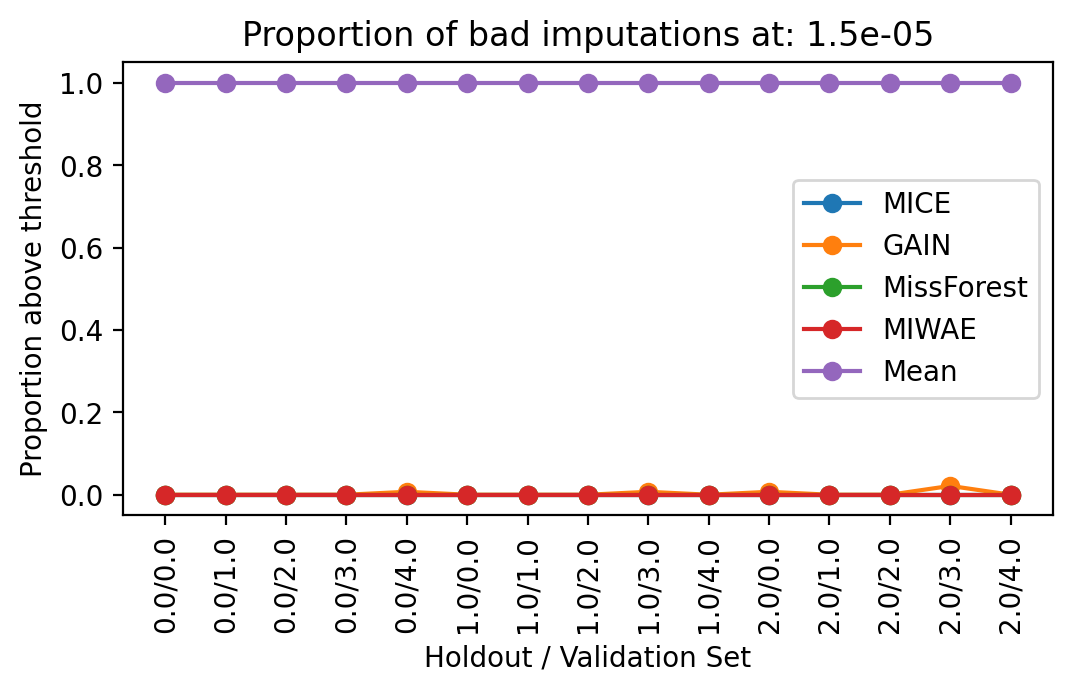} \\
      \includegraphics[width=4cm]{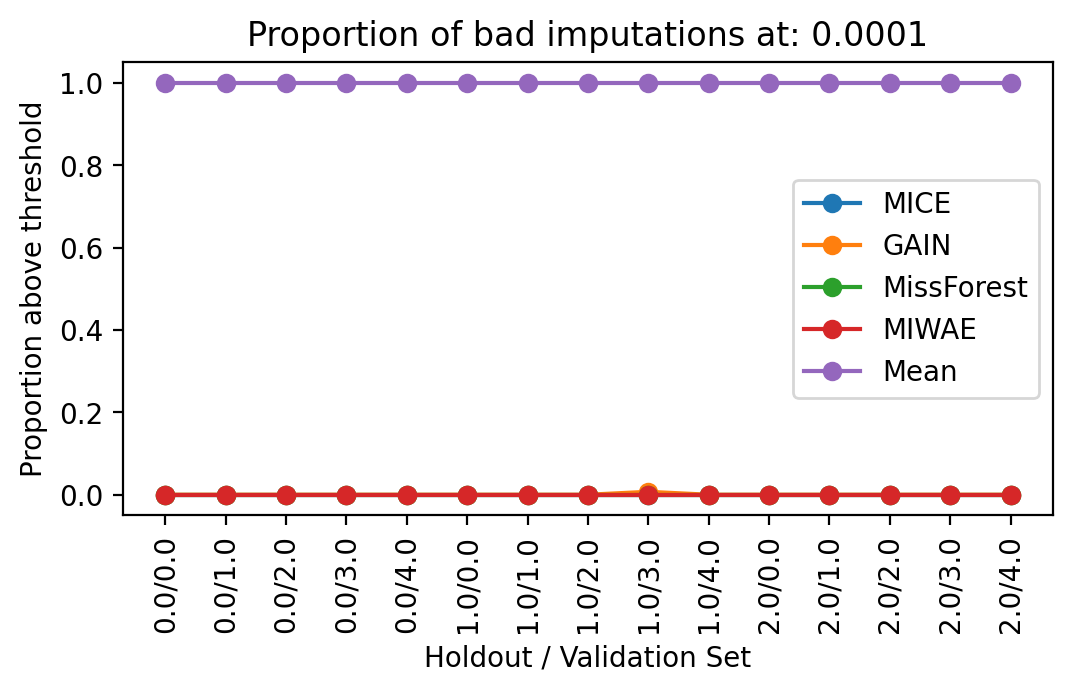}&
      \includegraphics[width=4cm]{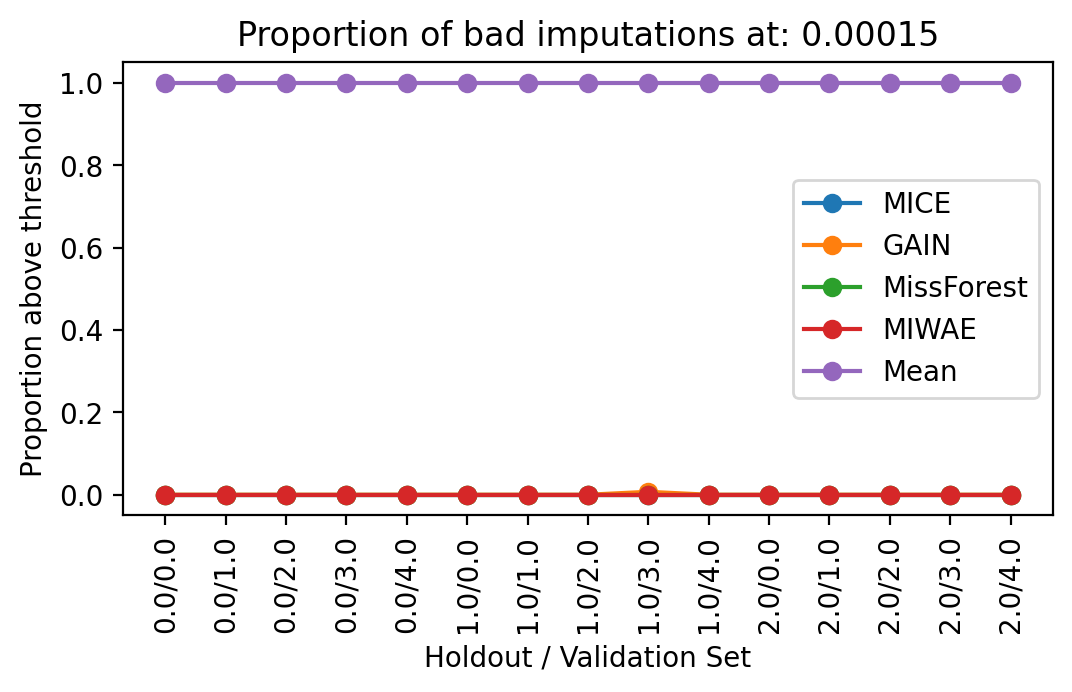}&
      \includegraphics[width=4cm]{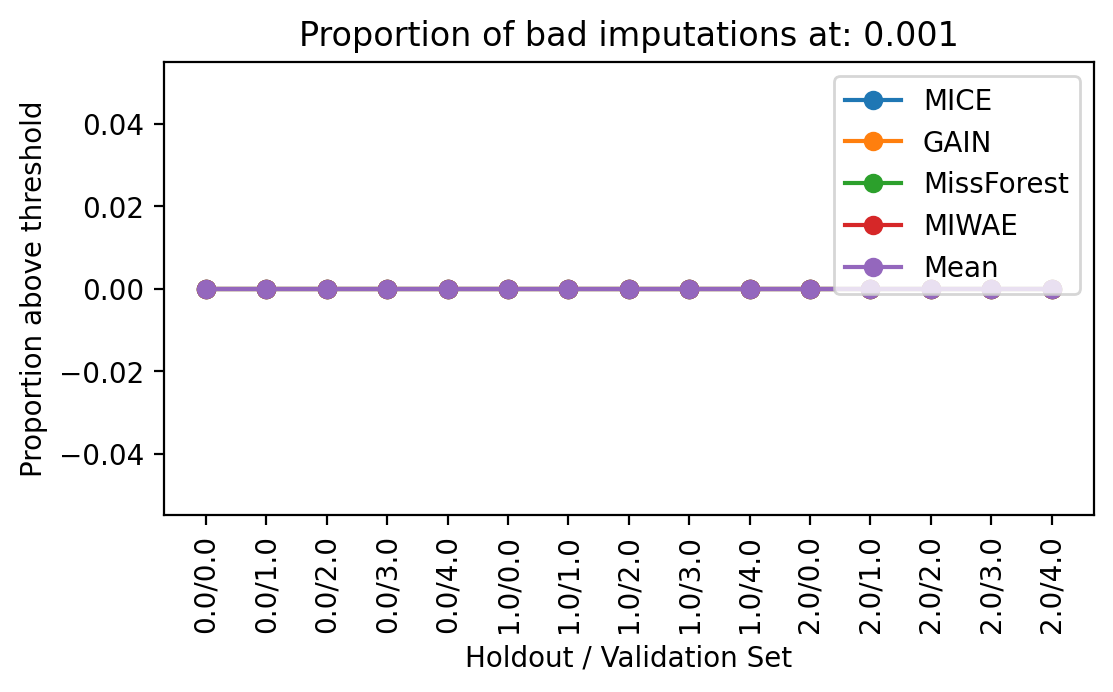}&
      \includegraphics[width=4cm]{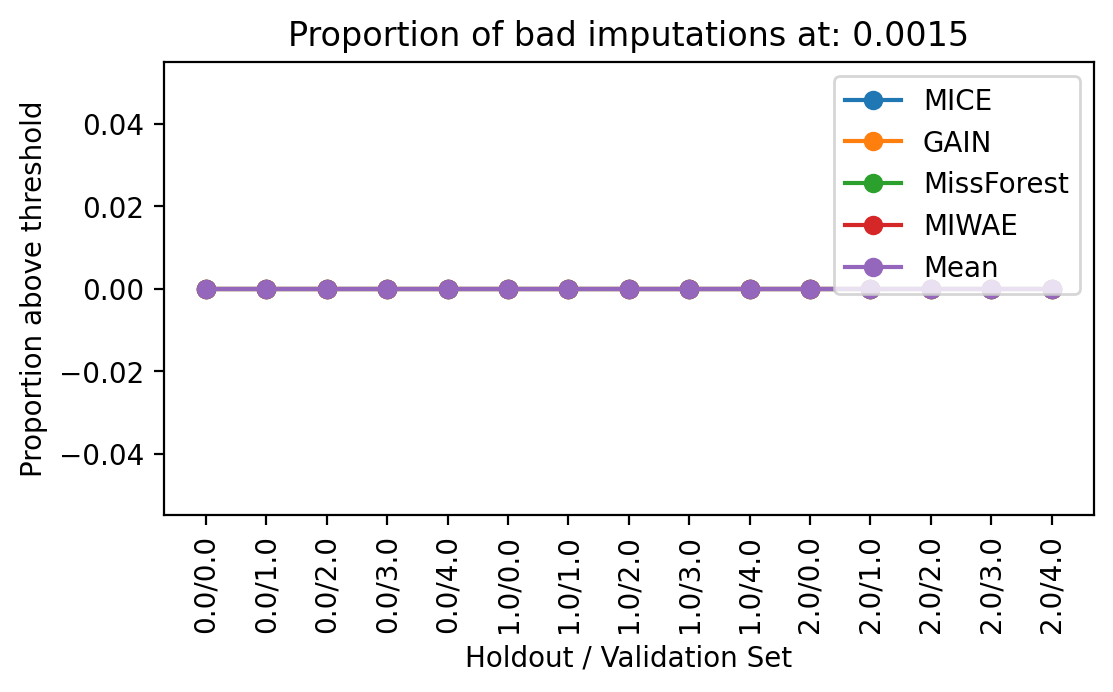} \\
    \end{tabular}
    \caption{The proportion of repeated imputations that give outlier Wasserstein distances, at the thresholds shown, for different imputation methods.}
    \label{fig:comp_err_supp}
\end{figure}

To isolate whether the large variances are due to stochasticity in the algorithms, we now go back and consider the original feature distributions, rather than the projected distributions. If an imputation algorithm occasionally imputes poorly in particular features, it will be identified here. For each holdout and validation set, we compute the Wasserstein distance between the imputed data and the true data for all features in the 10 repeats, i.e.\ for each feature we have 10 Wasserstein distances. We want to understand how often these distances are very large relative to how often they are small. In Figure~\ref{fig:comp_err}, we show the proportion of the imputations that have Wasserstein distances above a threshold of $10^{-7}$ for each holdout set and validation set. The plots for other thresholds are in Figure~\ref{fig:comp_err_supp}.
It can be seen that the Mean imputation method leads to feature imputations that always have relatively large distances from the true values. This is not surprising as it is the baseline model. It is surprising that GAIN is often imputing with high distance (80\% at $10^{-7}$ and 40\% at $10^{-6}$), indicating some stochasticity in the imputation method which causes poor imputations for some computations and better imputation for others. MIWAE demonstrates the same stochasticity to a lesser extent, followed by MissForest. This highlights the importance of performing multiple imputation runs for models which have stochasticity integral to them. For GAIN and MIWAE, this is particularly true as deep neural networks will occasionally find local minima at their optimum and generative adversarial networks are liable to mode collapse \cite{arjovsky_wasserstein_2017}.

\subsubsection*{Interpretability}

First, for each classifier, we find the top ten configurations of validation set, holdout set and imputation methods that achieve the best performance. We then rank these configurations by the distance ratio induced by the imputation method and take the model with the smallest and largest induced distance ratio for the sliced Wasserstein distance. This gives us the two models which are high performing but which are trained using data of different imputation quality (see Tables \ref{tab:interp_rf}--\ref{tab:interp_ng}). The features that are important for a model's prediction can be found using many interpretability techniques. In this paper we employ Shapley values\cite{lundberg_unified_2017} implemented in the Python package \texttt{shap}. 
\begin{figure}[htb!]
    \centering
    \begin{tabular}{ M{6.5cm} | M{6.5cm} }
     \textbf{Random Forest} & \textbf{XGBoost} \\
    \hline
      \includegraphics[width=6cm]{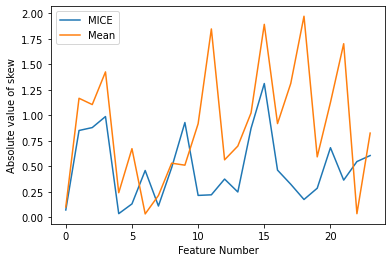}&
      \includegraphics[width=6cm]{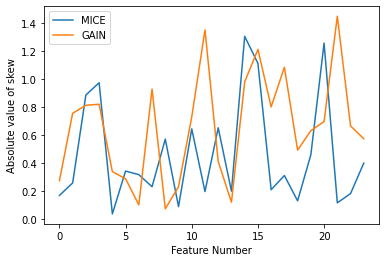}\\
           \multicolumn{2}{c}{\textbf{NGBoost}} \\
    \hline
      \multicolumn{2}{c}{\includegraphics[width=6cm]{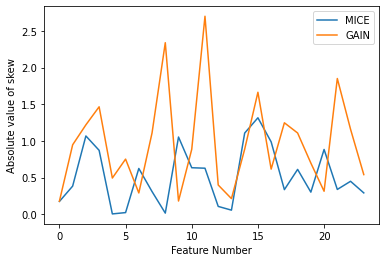}}\\
    \end{tabular}
    \caption{For each classifier, we give the absolute skew of the Shapley values for each feature for the two candidate models identified.}
    \label{fig:interp_comp}
\end{figure}

\clearpage

\subsection*{Additional performance metrics for the downstream classification task}

\begin{figure}[htb!]
\centering
\captionsetup[subfigure]{oneside,margin={-1cm,0cm}} 
         
\subfloat[Classifier dependence]{
   \includegraphics[width=0.85\textwidth]{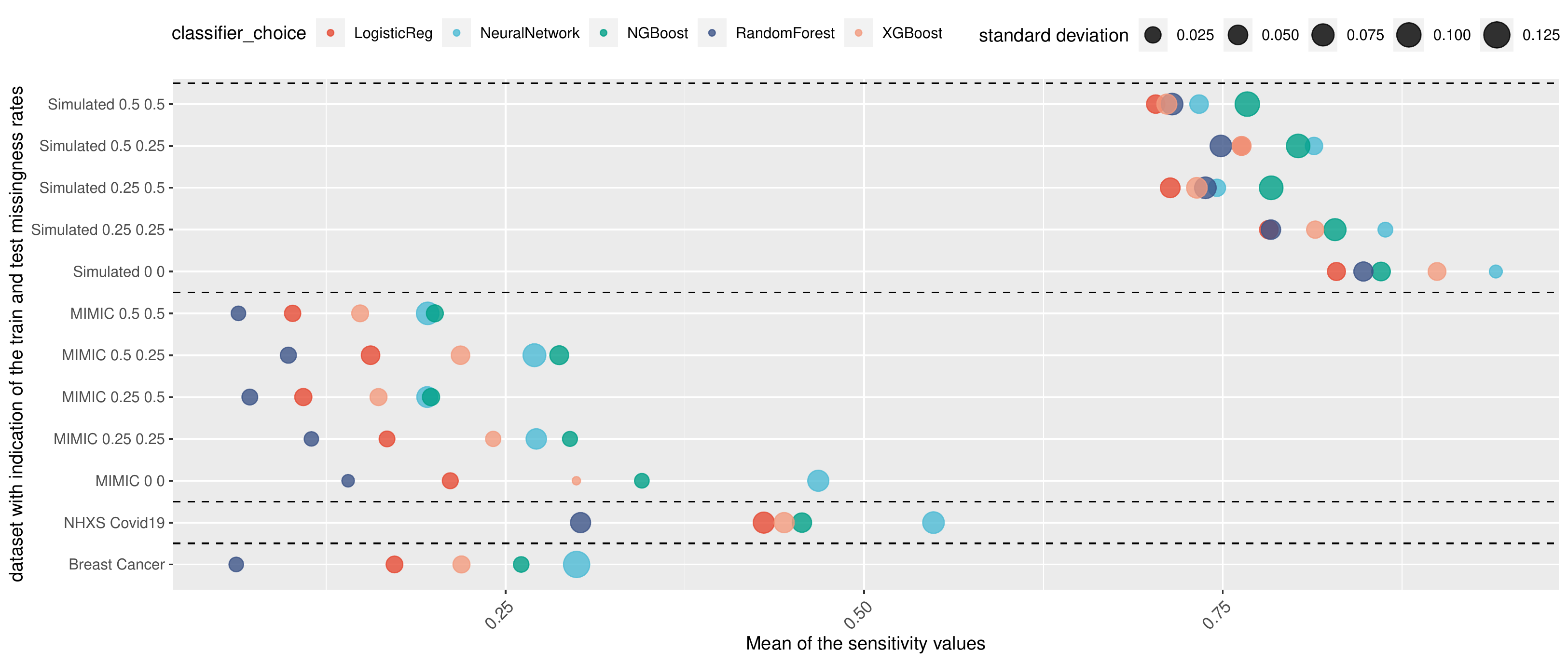}
   %{figs/buble_plt_cls_complete_dt_corrected.pdf}
         \label{supp1g}}
        % \newline
     %\end{subfigure}
     \qquad
\subfloat[Imputation dependence]{
    \centering
       \hspace{-0.1cm}\includegraphics[width=0.85\textwidth]{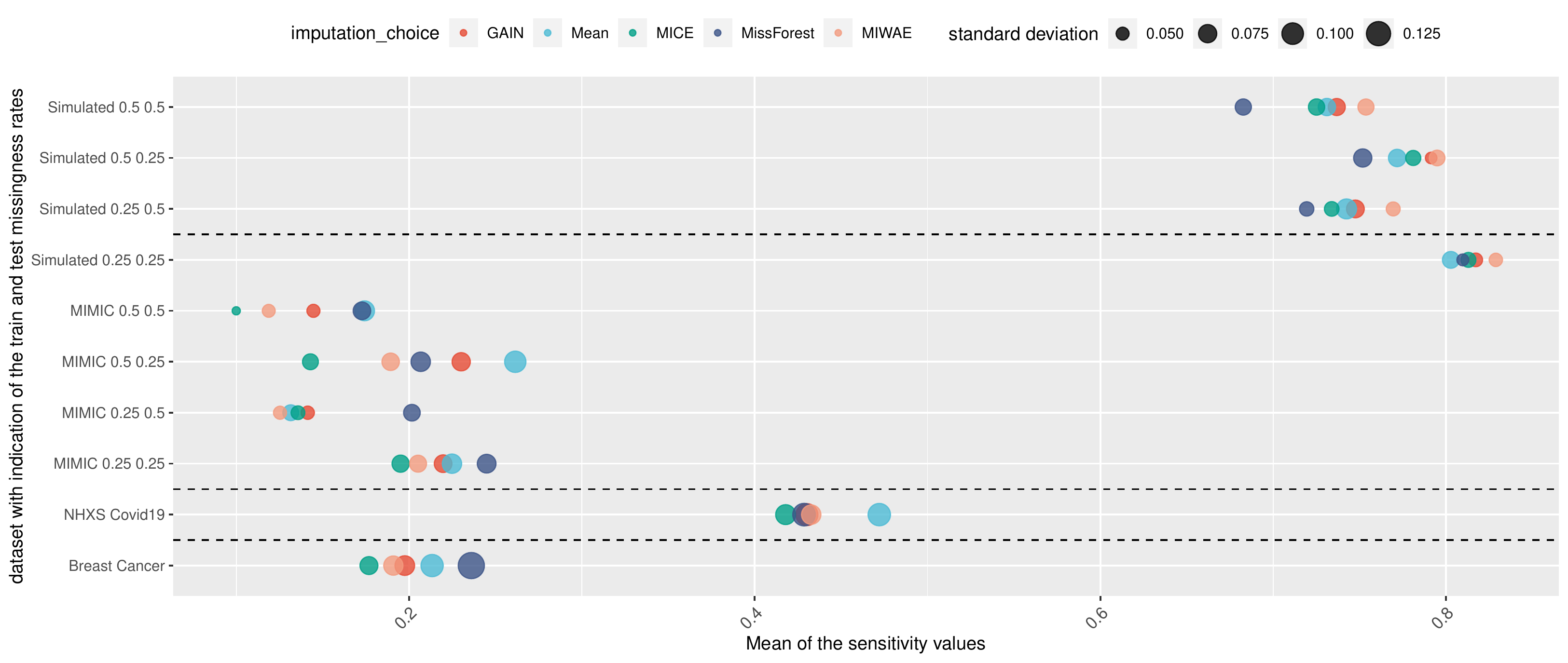}%{figs/Buble_plot_imputation.png}
         %\caption{$y=x$}
         \label{supp1h}}
         
 \caption{\textbf{Dependence of Sensitivity of downstream classification performance on classification and imputation methods.}
 \label{fig:sup4}}
\end{figure}

\begin{figure}[htb!]
\centering
\captionsetup[subfigure]{oneside,margin={-1cm,0cm}} 
         
\subfloat[Classifier dependence]{
   \includegraphics[width=0.85\textwidth]{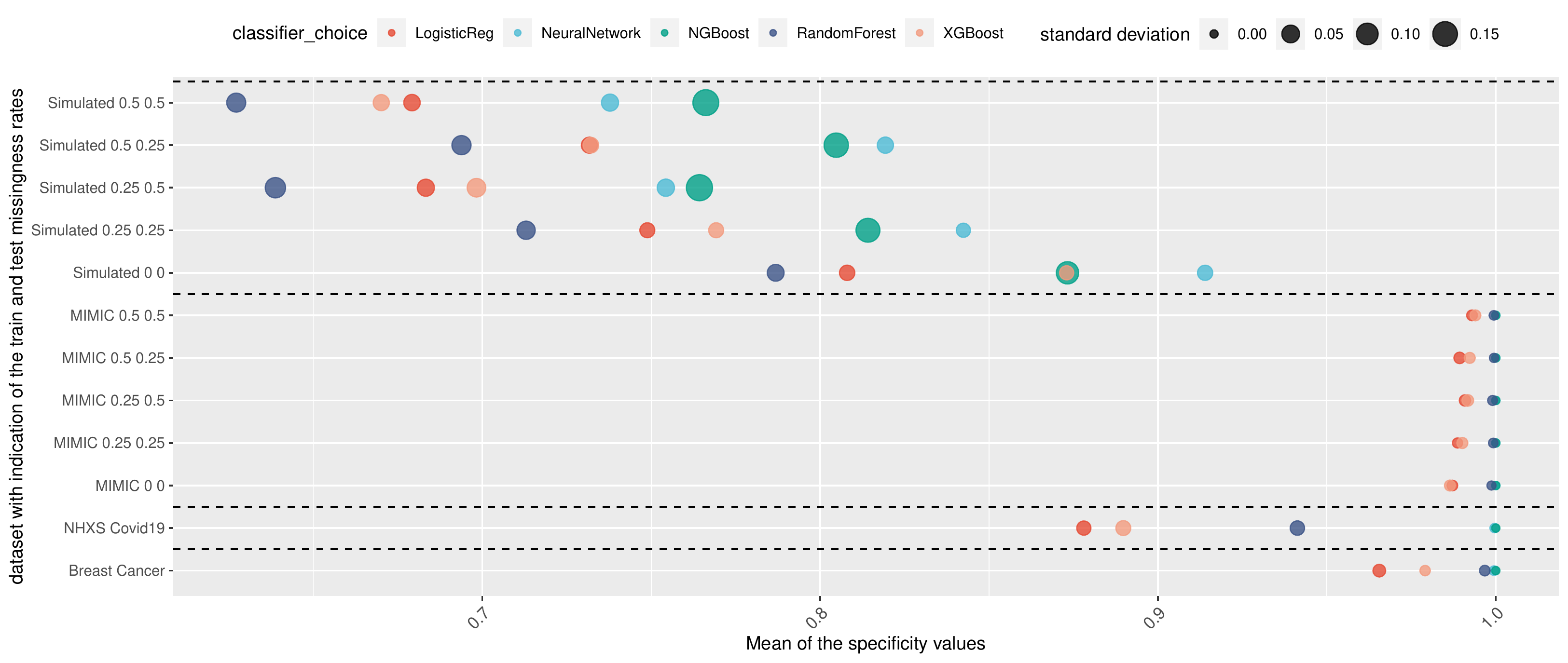}
   %{figs/buble_plt_cls_complete_dt_corrected.pdf}
         \label{supp1i}}
        % \newline
     %\end{subfigure}
     \qquad
\subfloat[Imputation dependence]{
    \centering
       \hspace{-0.1cm}\includegraphics[width=0.85\textwidth]{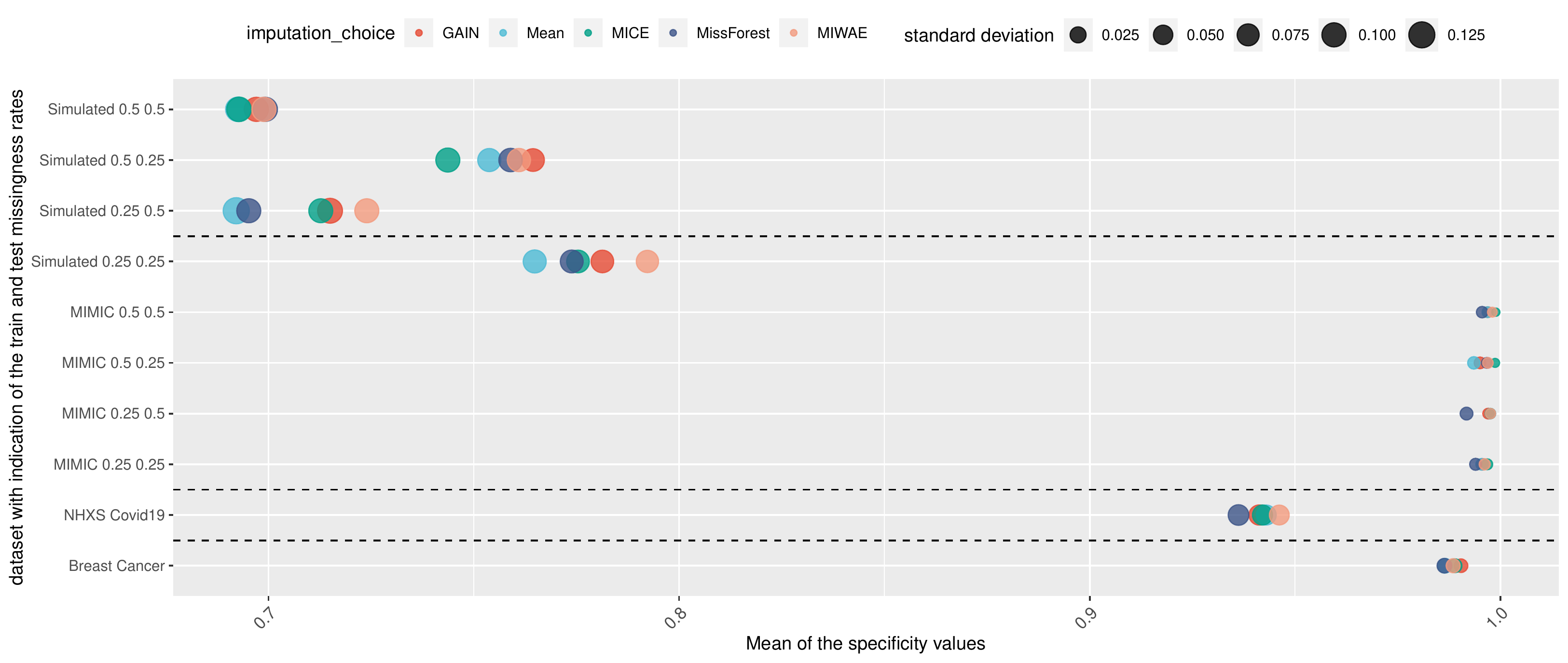}%{figs/Buble_plot_imputation.png}
         %\caption{$y=x$}
         \label{supp1j}}

 \caption{\textbf{Dependence of Specificity of downstream classification performance on classification and imputation methods.} 
 \label{fig:sup5}}
\end{figure}

In addition to the exploring the AUC values for downstream performance of classifiers, we also report the \textbf{Accuracy}, \textbf{Brier Score}, \textbf{Precision}, \textbf{Sensitivity} and \textbf{Specificity}. These results are presented in Figures~\ref{fig:sup1}--\ref{fig:sup5}. For the \textbf{Simulated} and \textbf{MIMIC-III} datasets, we show the performance for missingness rates of 25\% and 50\% in the development and test data.

\subsubsection*{Dependence of the accuracy}

\begin{figure}[h!]
\centering
\captionsetup[subfigure]{oneside,margin={-1cm,0cm}}
\subfloat[Classifier dependence]{
   \includegraphics[width=0.85\textwidth]{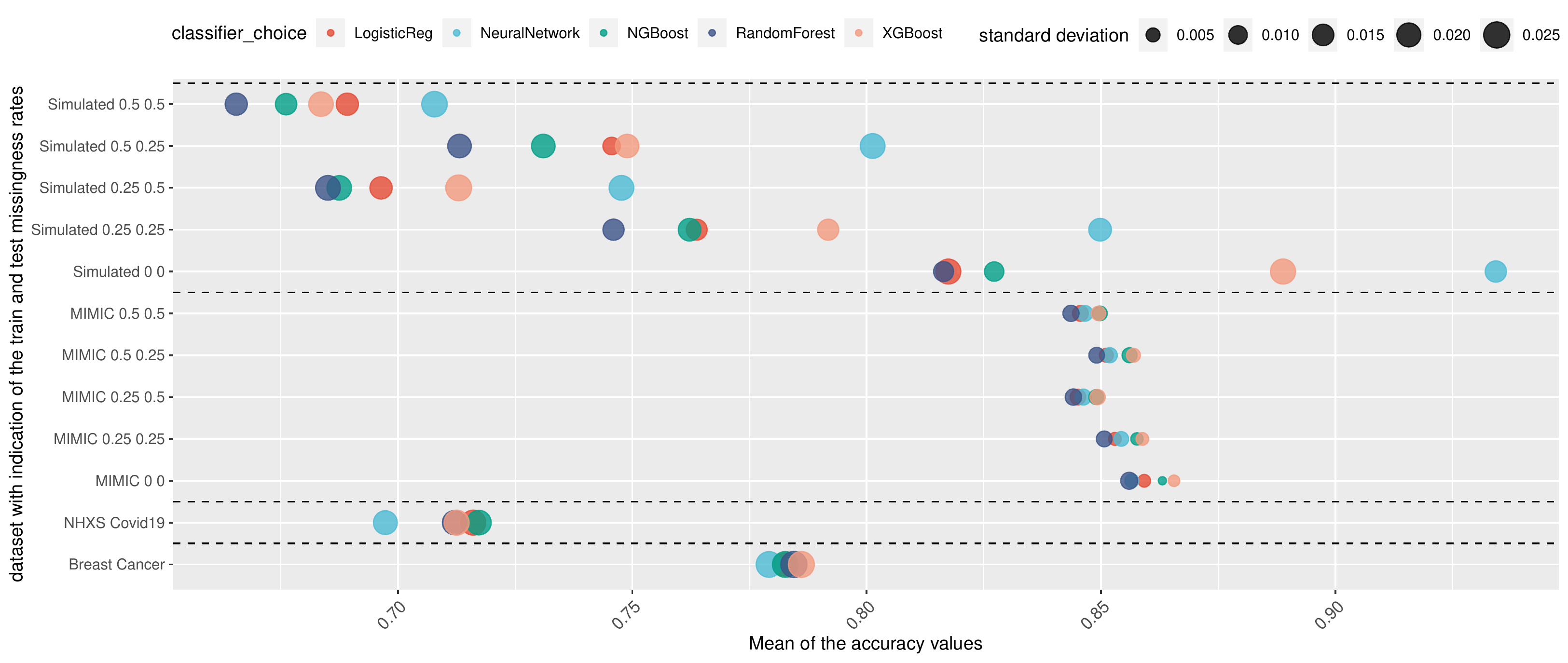}
   %{figs/buble_plt_cls_complete_dt_corrected.pdf}
         \label{supp1a}}
        % \newline
     %\end{subfigure}
     \qquad
\subfloat[Imputation dependence]{
    \centering
       \hspace{-0.1cm}\includegraphics[width=0.85\textwidth]{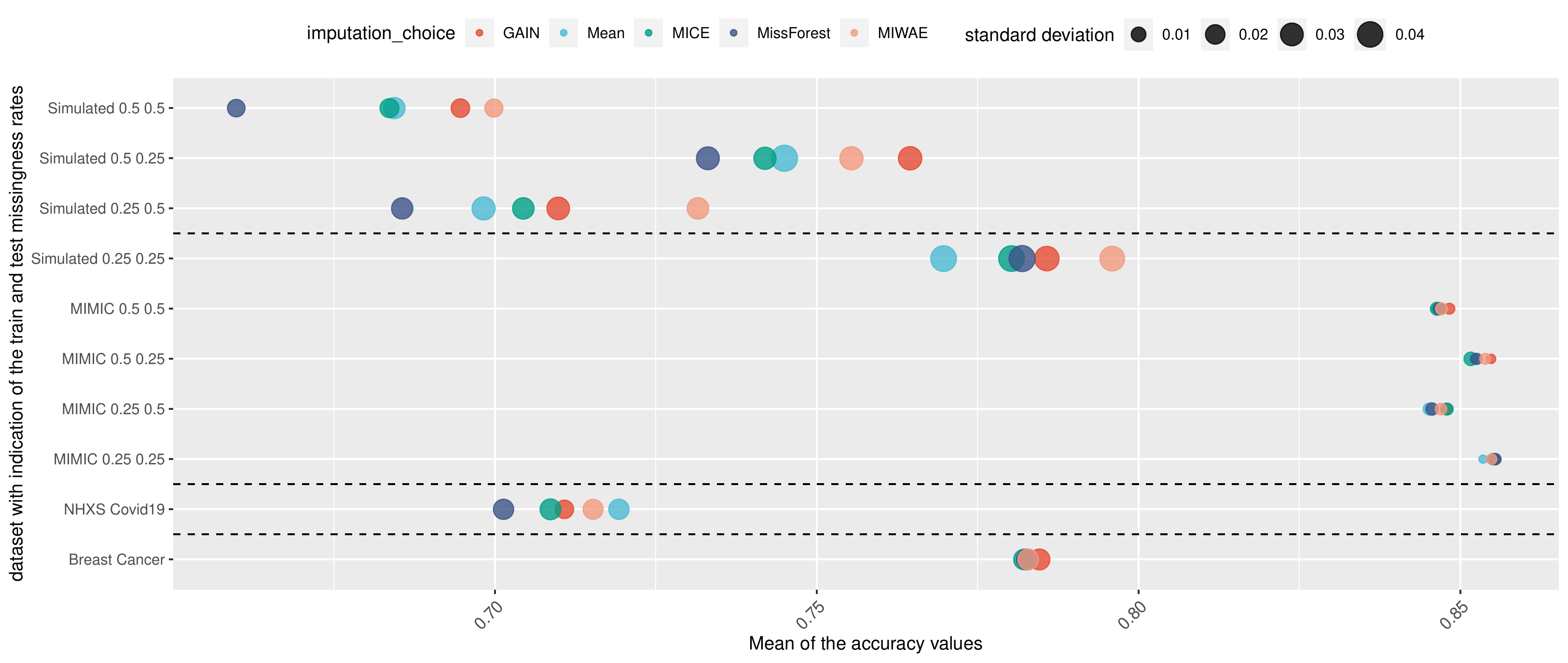}%{figs/Buble_plot_imputation.png}
         %\caption{$y=x$}
         \label{supp1b}}
         
 \caption{\textbf{Dependence of Accuracy of downstream classification performance on classification and imputation methods.} These plots show the \textbf{Accuracy} of the downstream performance on the classification (a) and imputation methods (b). The size of each marker indicates the standard deviation.
 \label{fig:sup1}}
\end{figure}

\clearpage

\subsubsection*{Dependence of the Brier score}

\begin{figure}[htb!]
\ContinuedFloat
\centering
\captionsetup[subfigure]{oneside,margin={-1cm,0cm}}         
\subfloat[Classifier dependence]{
   \includegraphics[width=0.85\textwidth]{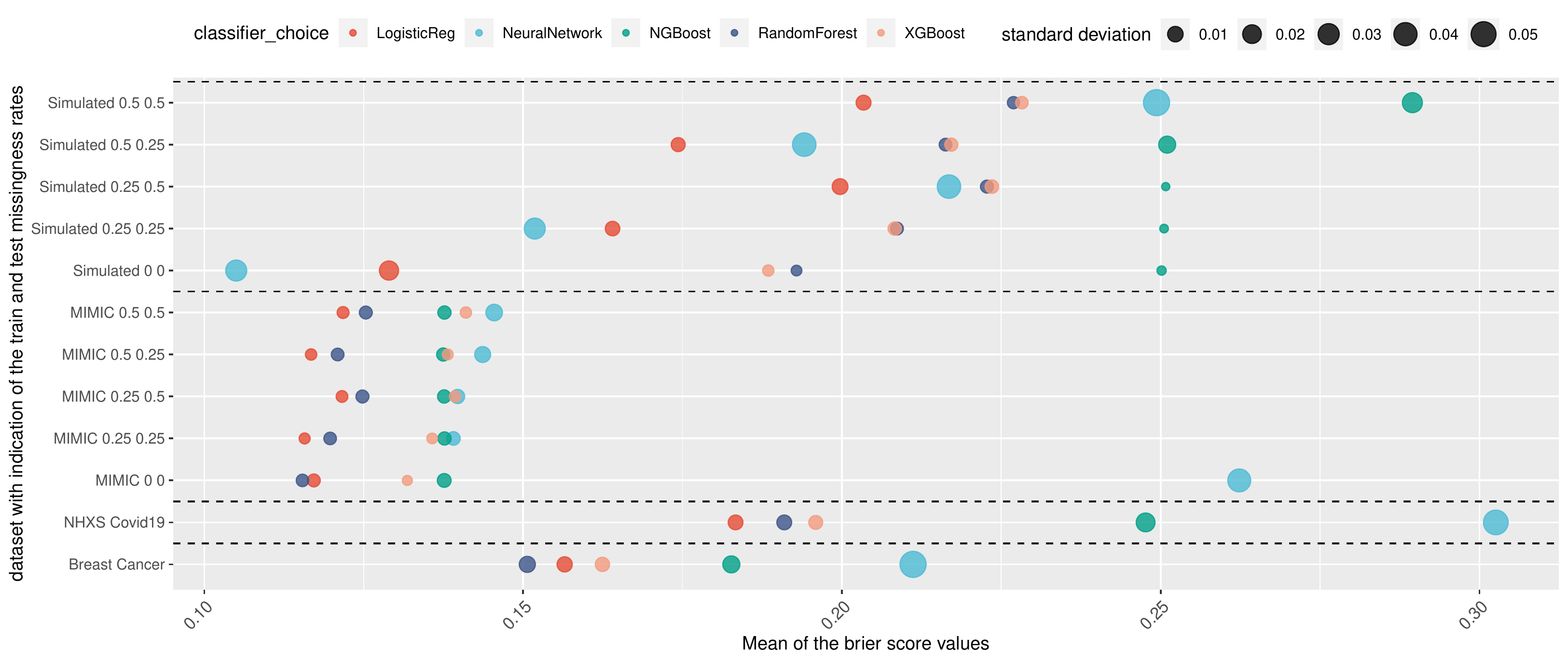}
   %{figs/buble_plt_cls_complete_dt_corrected.pdf}
         \label{supp1c}}
        % \newline
     %\end{subfigure}
     \qquad
\subfloat[Imputation dependence]{
    \centering
       \hspace{-0.1cm}\includegraphics[width=0.85\textwidth]{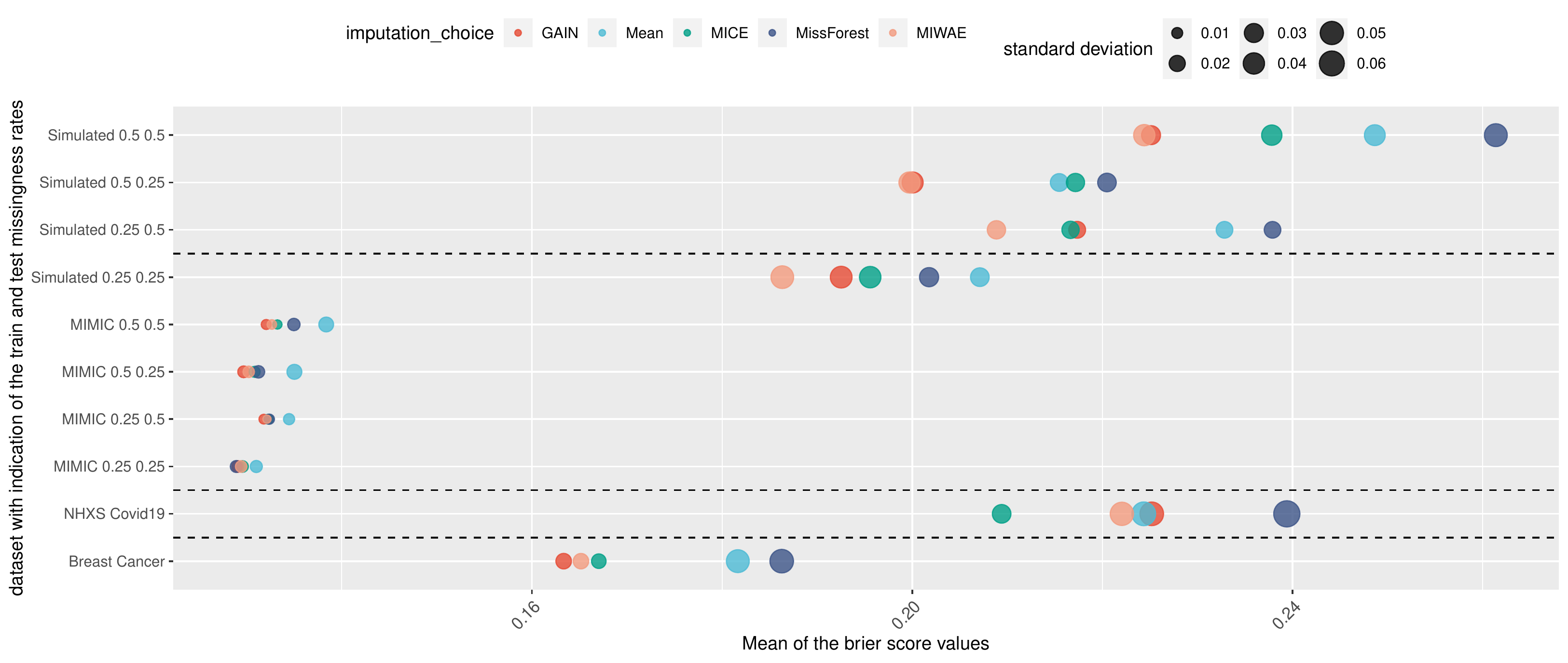}%{figs/Buble_plot_imputation.png}
         %\caption{$y=x$}
         \label{supp1d}}
 \caption{\textbf{Dependence of Brier Score of downstream classification performance on classification and imputation methods.} 
 \label{fig:sup2}}
\end{figure}

\clearpage

\subsubsection*{Dependence of the precision}

\begin{figure}[htb!]
\ContinuedFloat
\centering
\captionsetup[subfigure]{oneside,margin={-1cm,0cm}} 

\subfloat[Classifier dependence]{
   \includegraphics[width=0.85\textwidth]{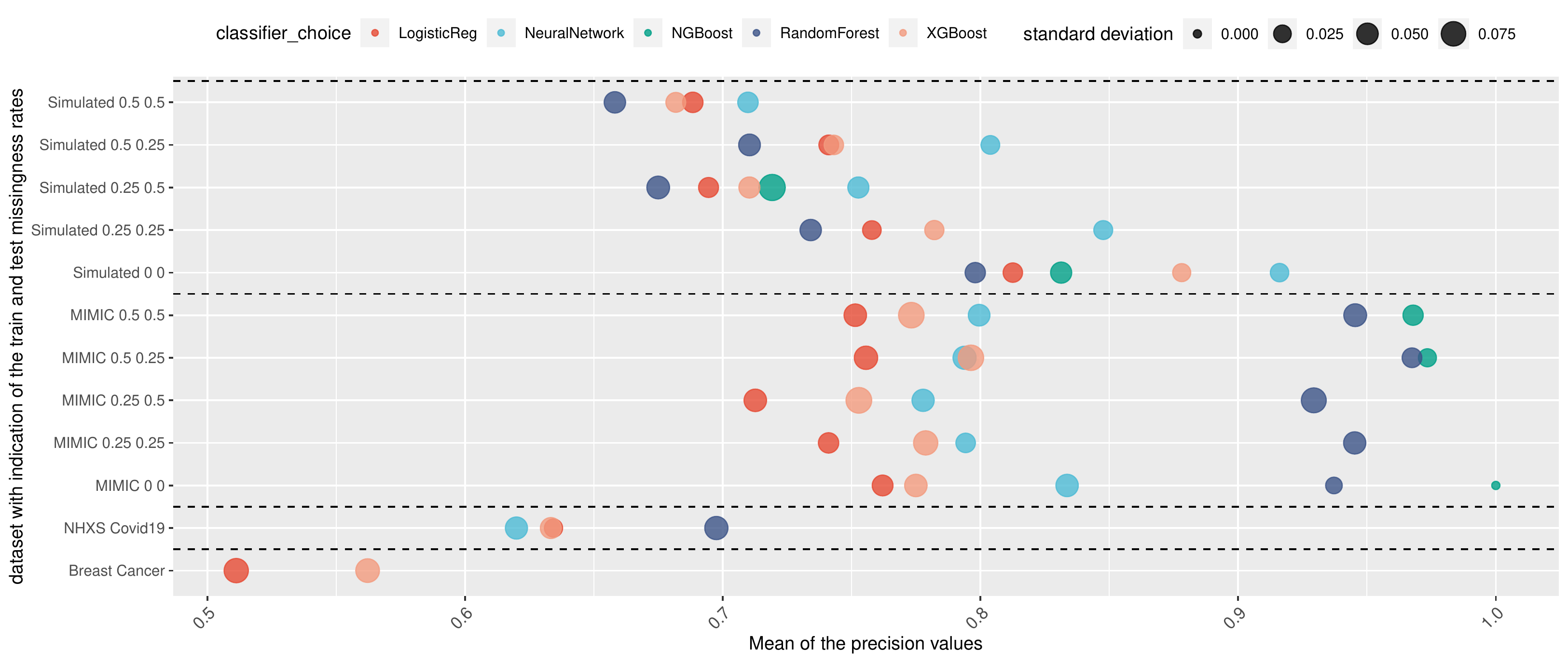}
   %{figs/buble_plt_cls_complete_dt_corrected.pdf}
         \label{supp1e}}
        % \newline
     %\end{subfigure}
     \qquad
\subfloat[Imputation dependence]{
    \centering
       \hspace{-0.1cm}\includegraphics[width=0.85\textwidth]{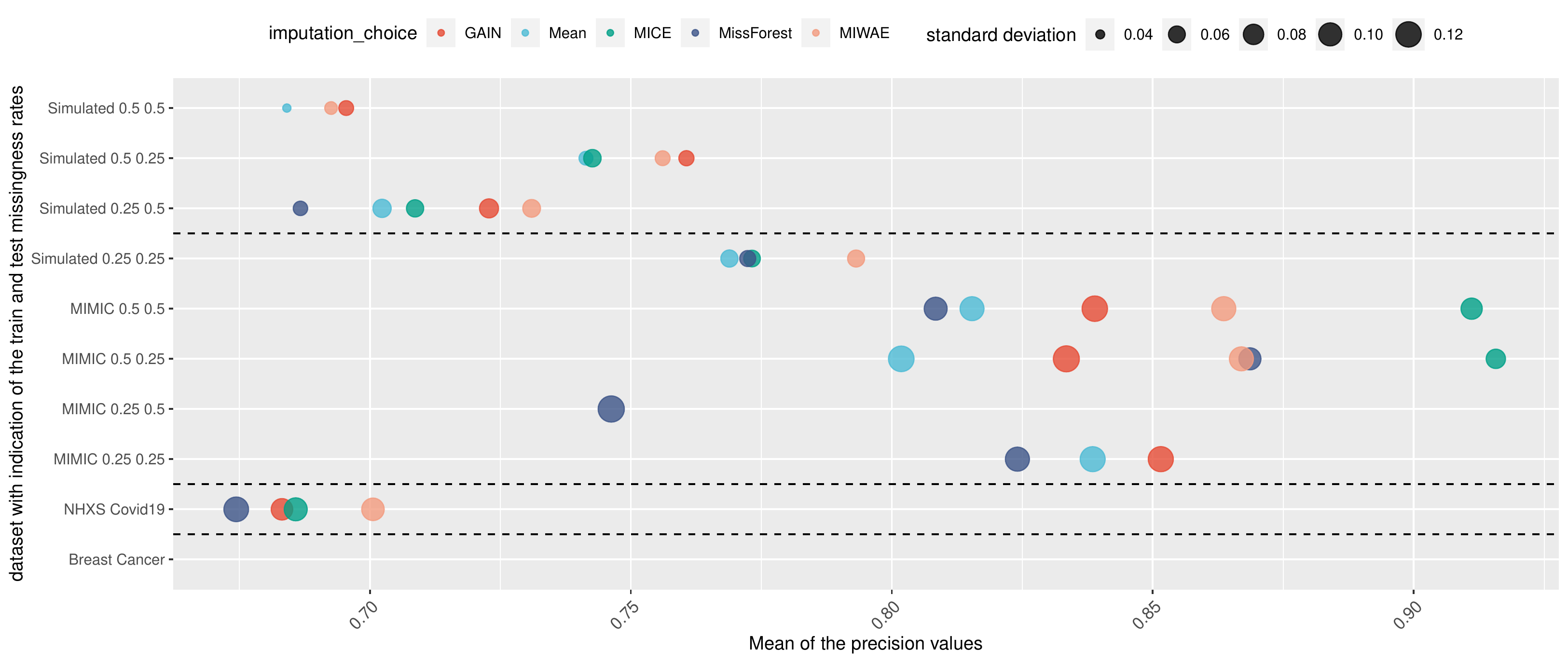}%{figs/Buble_plot_imputation.png}
         %\caption{$y=x$}
         \label{supp1f}}

 \caption{\textbf{Dependence of Precision of downstream classification performance on classification and imputation methods.}
 \label{fig:sup3}}
\end{figure}

\clearpage

\subsubsection*{Dependence of the sensitvity}

\clearpage

\subsubsection*{Dependence of the specificity}

\clearpage
\subsection*{Supplementary Figures for the Sample-wise Discrepancy Statistics}

\subsubsection*{A: Sample-wise discrepancy for the MIMIC-III dataset at different train and test missingness rates}

\clearpage

\subsubsection*{A: Sample-wise discrepancy for the Simulated dataset at different train and test missingness rates}

%\textcolor{red}{Tolou: I am checking these plots right now.}

\clearpage

\subsection*{Supplementary Figures for the Feature-wise Discrepancy Statistics}

\subsubsection*{B: Feature-wise discrepancy for the MIMIC-III dataset at the respective train and test missingness rates of 25\% and 50\%.}

\clearpage

\subsubsection*{B: Feature-wise discrepancy for the MIMIC-III dataset at the respective train and test missingness rates of 50\% and 25\%.}

\clearpage

\subsubsection*{B: Feature-wise discrepancy for the MIMIC-III dataset at the respective train and test missingness rates of 50\% and 50\%.}

\clearpage

\subsubsection*{B: Feature-wise discrepancy for the Simulated dataset at the respective train and test missingness rates of 25\% and 25\%.}

\clearpage

\subsubsection*{B: Feature-wise discrepancy for the Simulated dataset at the respective train and test missingness rates of 25\% and 50\%.}

\clearpage

\subsubsection*{B: Feature-wise discrepancy for the Simulated dataset at the respective train and test missingness rates of 50\% and 25\%.}

\clearpage

\subsubsection*{B: Feature-wise discrepancy for the Simulated dataset at the respective train and test missingness rates of 50\% and 50\%.}

\clearpage

\subsection*{Supplementary Figures for the Sliced Wasserstein Discrepancy Statistics}

\subsubsection*{C: Sliced Wasserstein discrepancy for the MIMIC-III dataset at different train and test missingness rates}

\clearpage

\subsubsection*{C: Sliced Wasserstein discrepancy for the MIMIC-III dataset at different train and test missingness rates}

\clearpage

\subsubsection*{C: Sliced Wasserstein discrepancy for the Simulated dataset at different train and test missingness rates}

\clearpage

\subsubsection*{C: Sliced Wasserstein discrepancy for the Simulated dataset at different train and test missingness rates}

\clearpage

\subsection*{Link between quality and downstream classification performance}

\clearpage

\subsection*{Supplementary Distance Ratio Figure}

\begin{figure}[htb!]
    \centering
    \begin{tabular}{ M{7.5cm} | M{7.5cm} }
     \textbf{Train 25\% Test 25\%} & \textbf{Train 25\% Test 50\%} \\
    \hline
      \includegraphics[width=7cm]{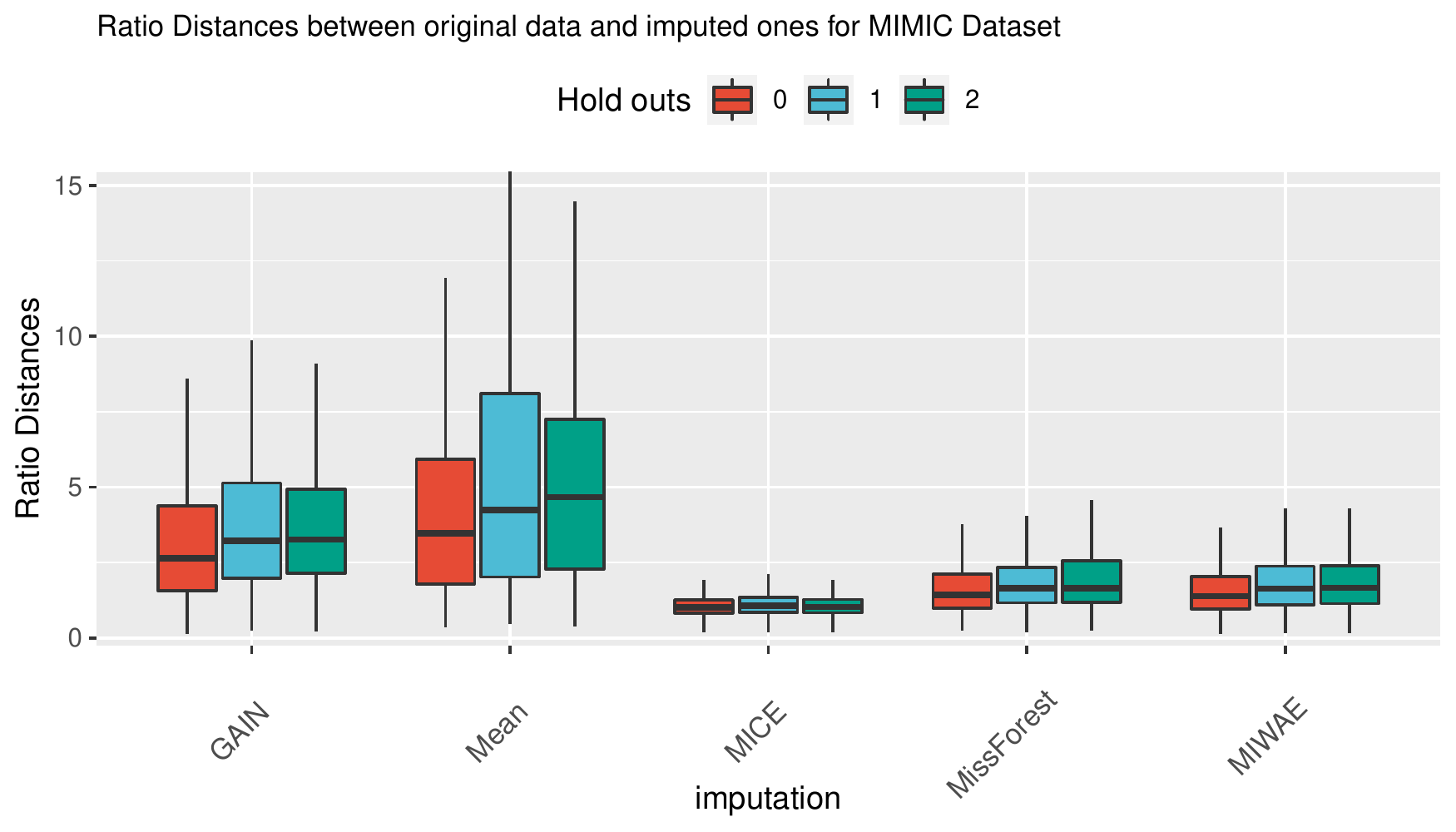}&
      \includegraphics[width=7cm]{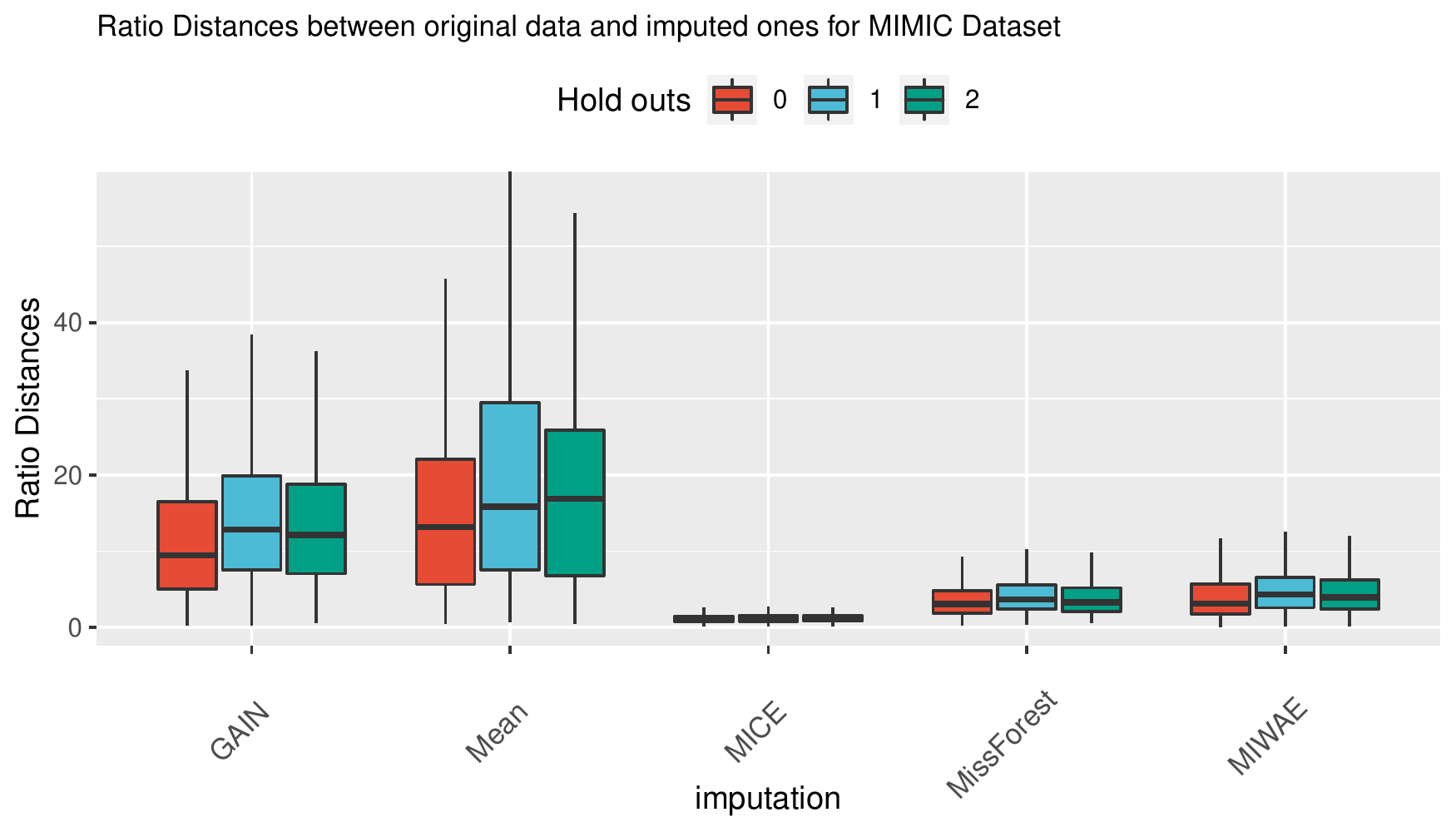}\\
           \textbf{Train 50\% Test 25\%} & \textbf{Train 50\% Test 50\%} \\
    \hline
      \includegraphics[width=7cm]{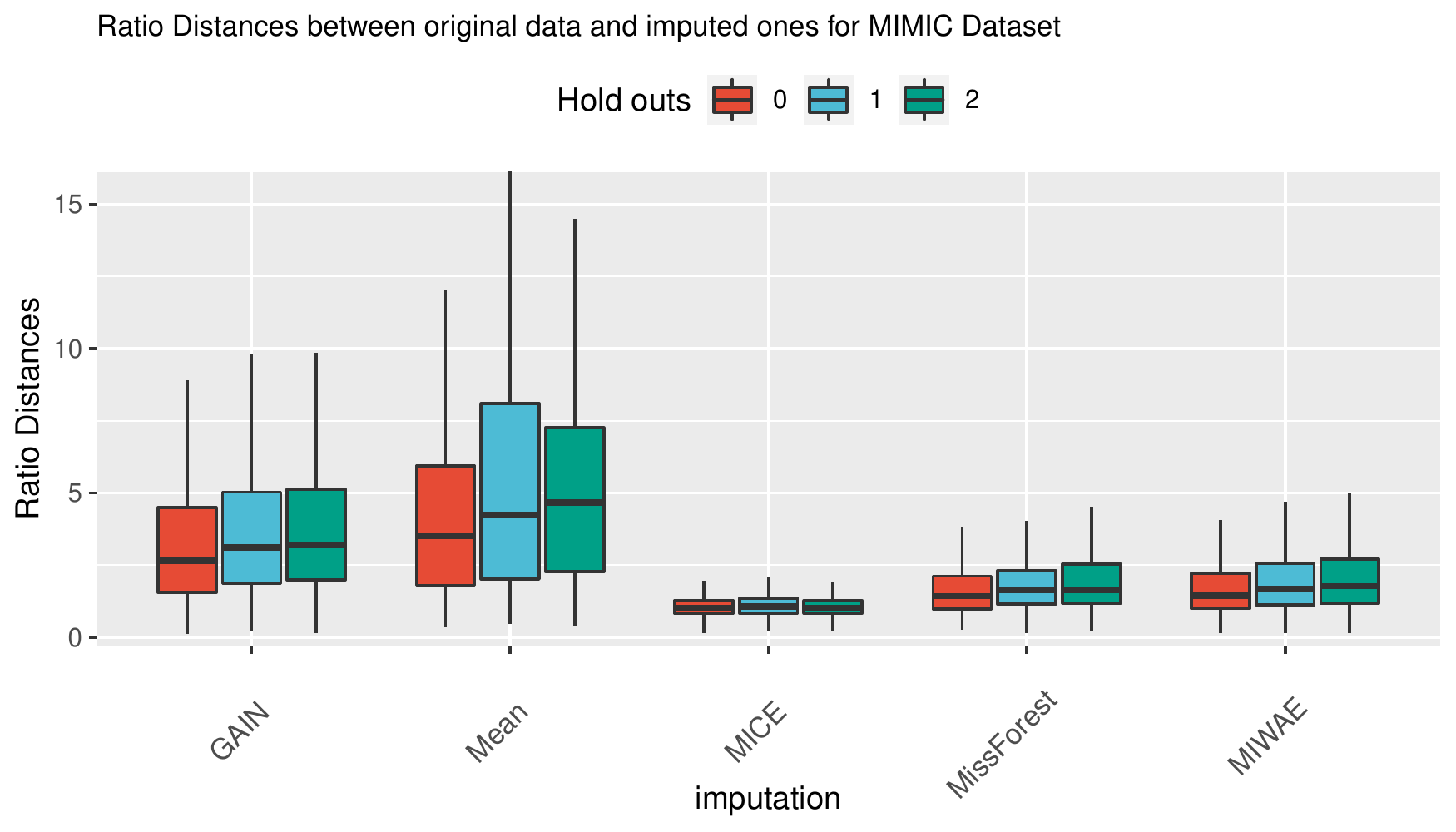}&
      \includegraphics[width=7cm]{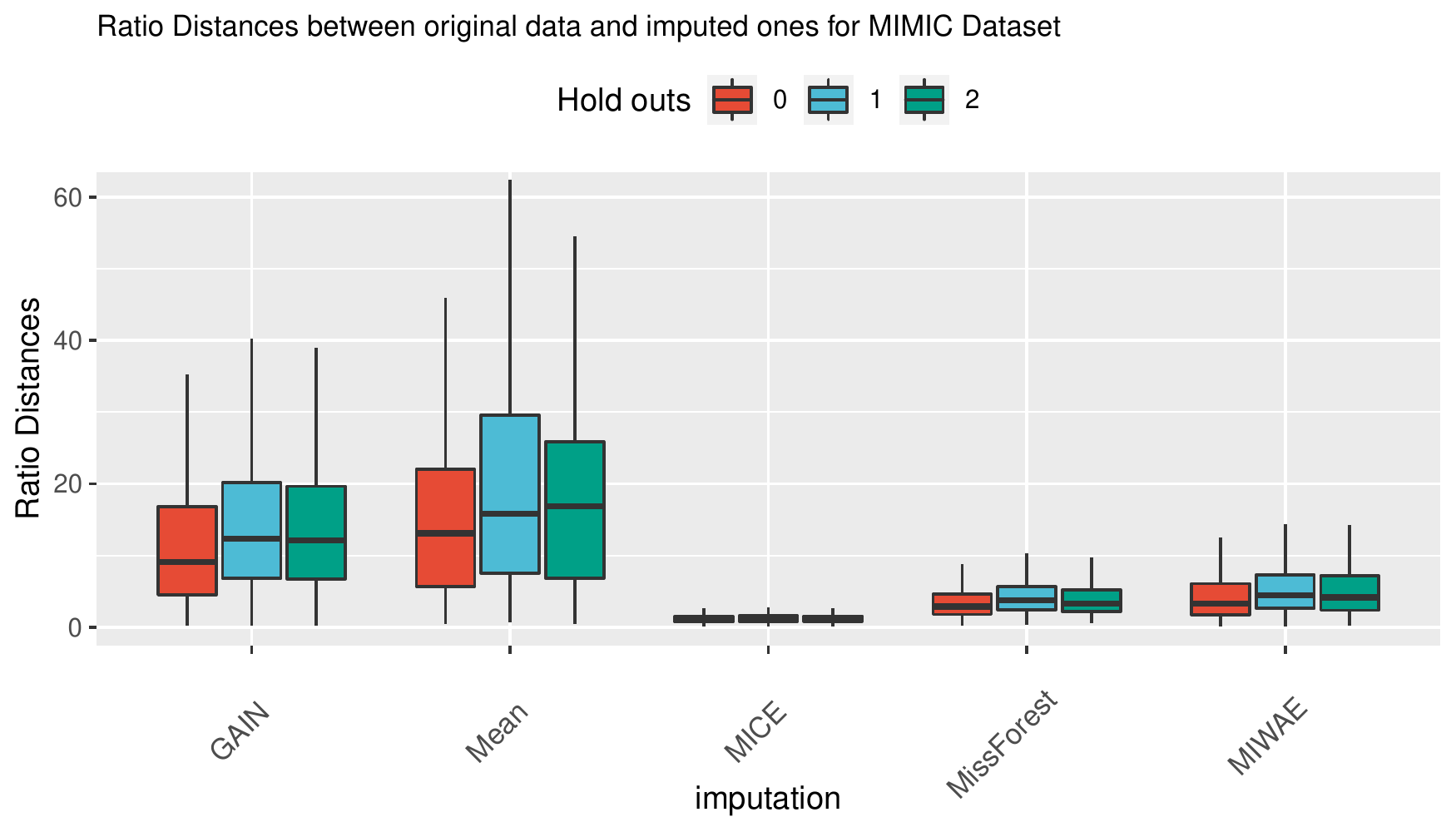}\\
    \end{tabular}
    \caption{Ratio of the Wasserstein distance for the imputed data compared to the original data for the \textbf{MIMIC-III} dataset at different train and test missingness rates.}
    \label{fig:distance_ratios}
\end{figure}

\begin{figure}[htb!]
    \centering
    \begin{tabular}{ M{7.5cm} | M{7.5cm} }
     \textbf{Train 25\% Test 25\%} & \textbf{Train 25\% Test 50\%} \\
    \hline
      \includegraphics[width=7cm]{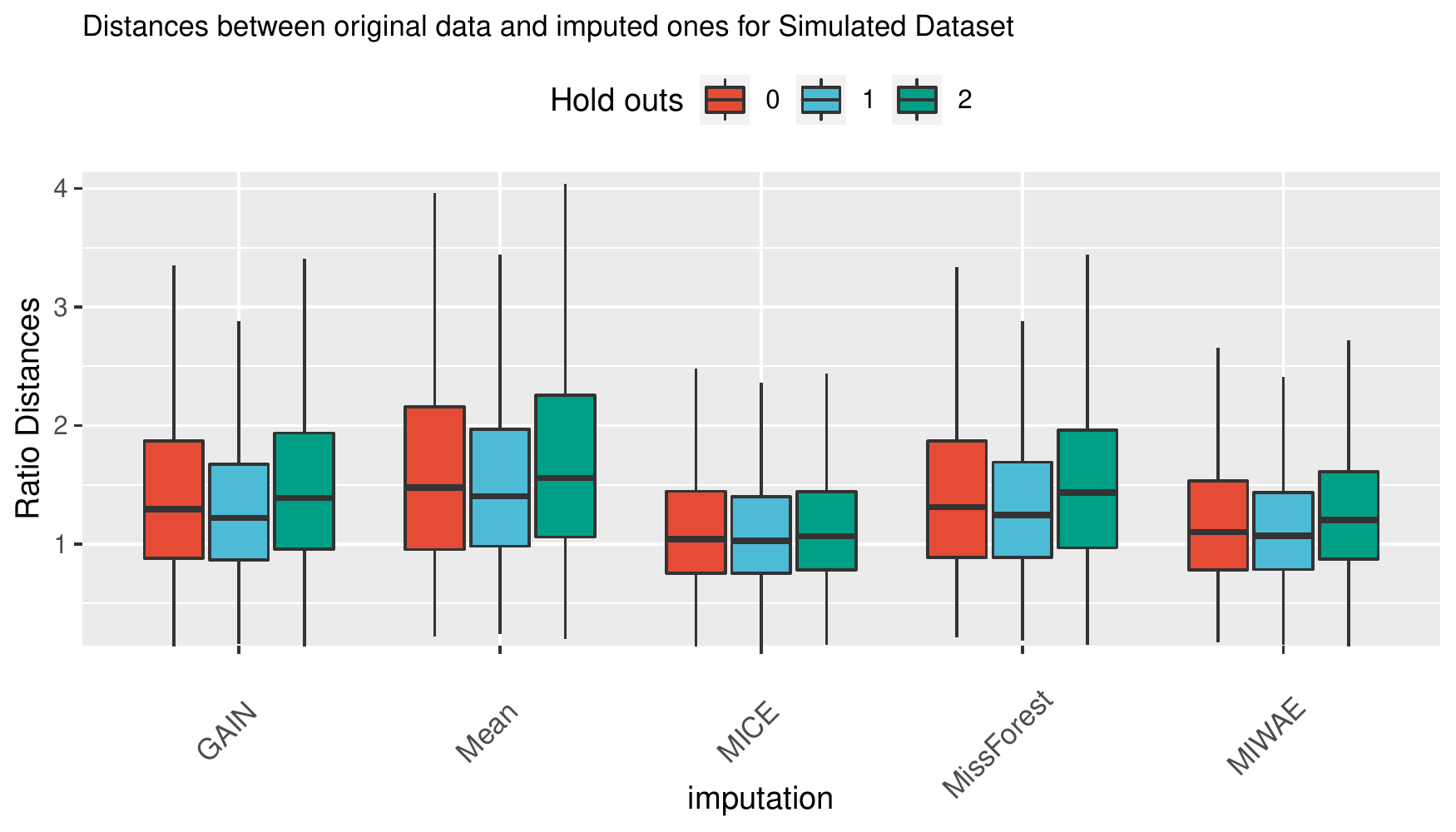}&
      \includegraphics[width=7cm]{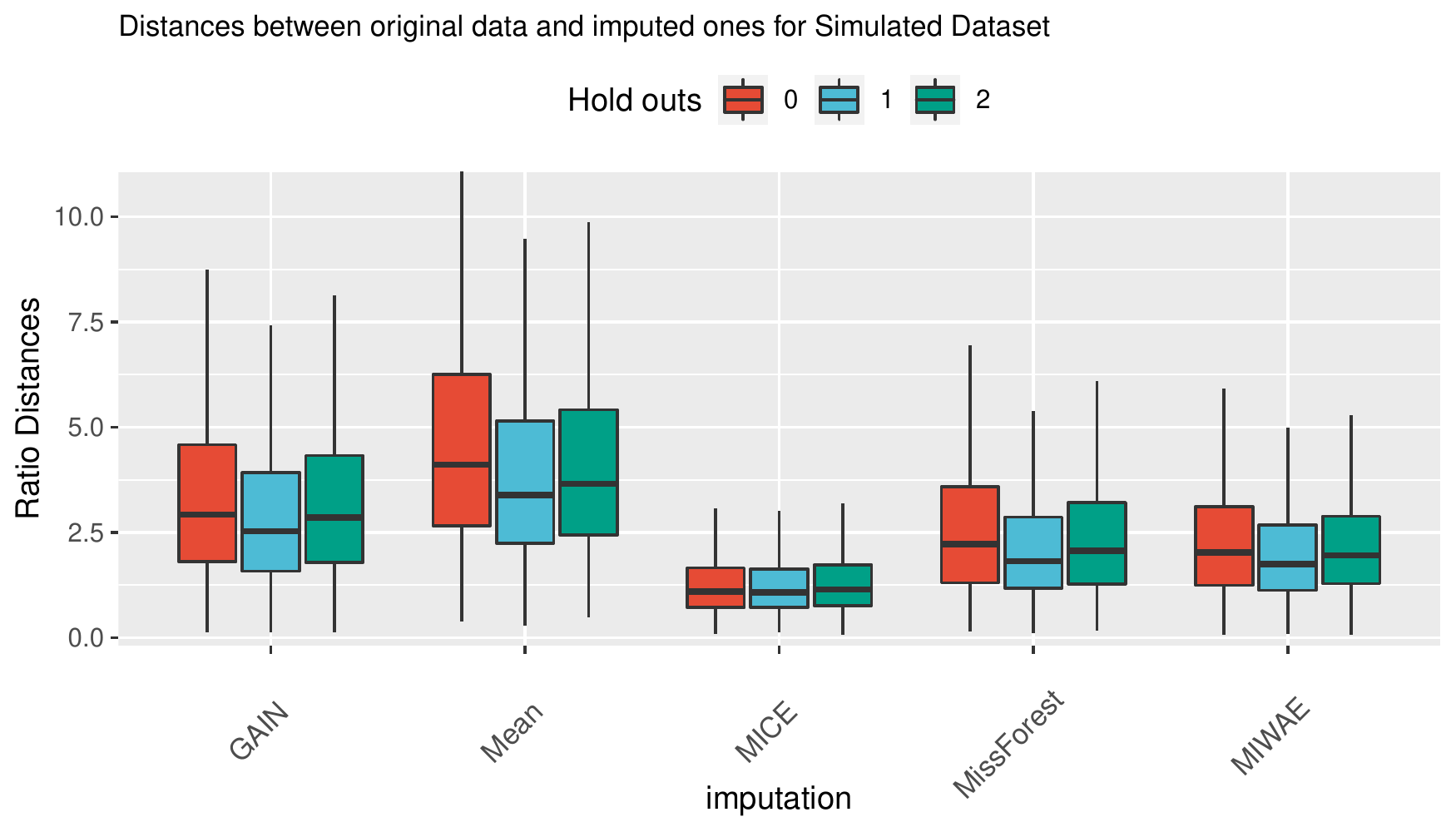}\\
           \textbf{Train 50\% Test 25\%} & \textbf{Train 50\% Test 50\%} \\
    \hline
      \includegraphics[width=7cm]{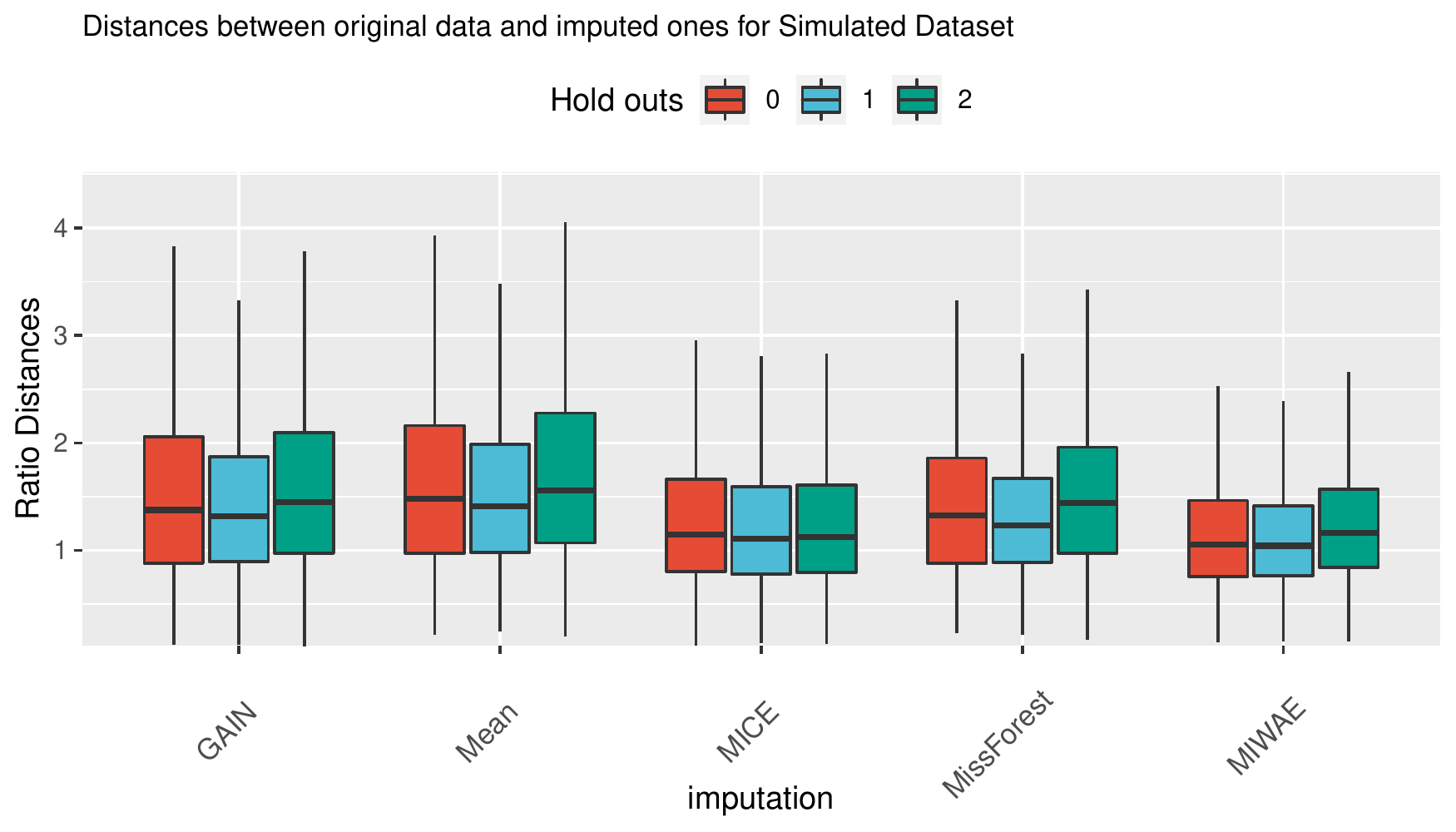}&
      \includegraphics[width=7cm]{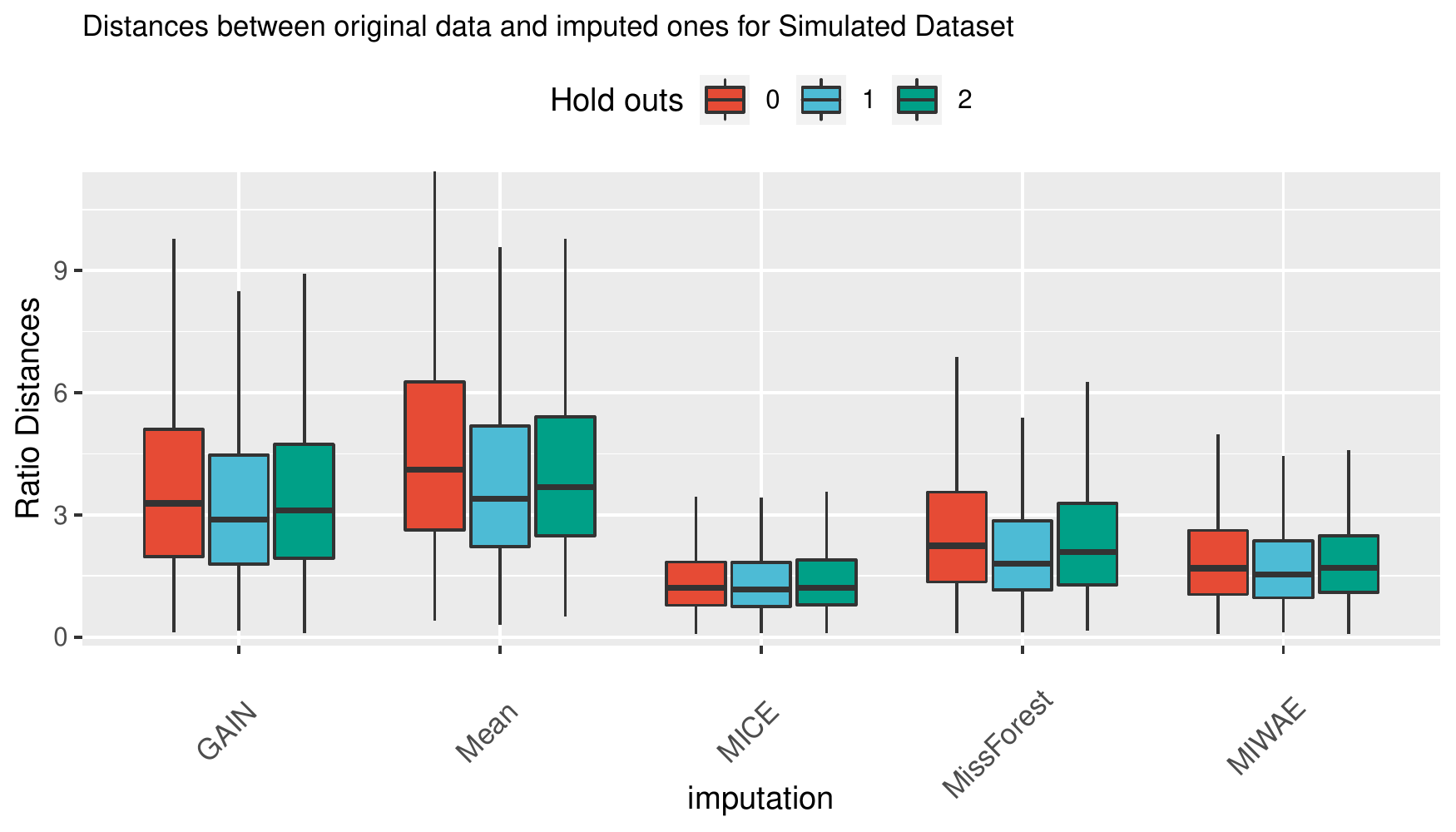}\\
    \end{tabular}
    \caption{Ratio of the Wasserstein distance for the imputed data compared to the original data for the \textbf{Simulated} dataset at different train and test missingness rates.}
    \label{fig:distance_ratios_syn}
\end{figure}

\clearpage

\subsection*{ANOVA Results for the Breast Cancer, NHSX COVID-19 and MIMIC-III datasets}

\begin{table}[ht]
\centering
\begin{tabular}{l|llllc}
  \hline
Variables & Estimate & Std. Error & $t$ value & Pr($>|t|$) \\ 
  \hline
classifier choice LogisticReg & 1.0824 & 0.0188 & 57.50 & <2e-16& *** \\ 
  classifier choice NeuralNetwork & 0.9841 & 0.0184 & 53.52 & <2e-16& *** \\ 
  classifier choice NGBoost & 1.0046 & 0.0185 & 54.37 & <2e-16& *** \\ 
  classifier choice RandomForest & 1.0475 & 0.0187 & 56.12 & <2e-16& *** \\ 
  classifier choice XGBoost & 1.0419 & 0.0186 & 55.90 & <2e-16& *** \\ 
   \hline
\end{tabular}
\caption{\label{sup_tab_1}Table of parameter coefficients for \textbf{Breast Cancer} dataset.}
\end{table}

\begin{table}[ht]
\centering
\begin{tabular}{l|llllc}
  \hline
Variables & Estimate & Std. Error & $t$ value & Pr($>|t|$) \\ 
  \hline
imputation choice GAIN & 1.1323 & 0.0340 & 33.33 & <2e-16& *** \\ 
  imputation choice Mean & 1.1689 & 0.0342 & 34.22 & <2e-16& *** \\ 
  imputation choice MICE & 1.1299 & 0.0340 & 33.27 & <2e-16& *** \\ 
  imputation choice MissForest & 1.0688 & 0.0337 & 31.74 & <2e-16& *** \\ 
  imputation choice MIWAE & 1.1764 & 0.0342 & 34.40 & <2e-16& *** \\ 
  classifier choice Neural Network & -0.0959 & 0.0354 & -2.71 & 0.0085& ** \\ 
  classifier choice NGBoost & 0.0357 & 0.0360 & 0.99 & 0.3239 \\ 
  classifier choice Random Forest & 0.0305 & 0.0359 & 0.85 & 0.3989 \\ 
  classifier choice XGBoost & 0.0749 & 0.0361 & 2.07 & 0.0423& * \\
   \hline
\end{tabular}
\caption{\label{sup_tab_2}Table of parameter coefficients for \textbf{NHSX COVID-19} dataset}
\end{table}

\begin{table}[H]
\centering
\begin{tabular}{l|rrrrc}
  \hline
Variables & Estimate & Std. Error & $t$ value & Pr($>|t|$) \\ 
  \hline
imputation choice GAIN & 1.0530 & 0.0130 & 81.03 & <2e-16& *** \\ 
  imputation choice Mean & 1.0119 & 0.0129 & 78.21 & <2e-16& *** \\ 
  imputation choice MICE & 1.0721 & 0.0130 & 82.33 & <2e-16& *** \\ 
  imputation choice MissForest & 1.0467 & 0.0130 & 80.60 & <2e-16& *** \\ 
  imputation choice MIWAE & 1.0661 & 0.0130 & 81.92 & <2e-16& *** \\ 
  classifier choice Neural Network & 0.0237 & 0.0123 & 1.93 & 0.0548& . \\ 
  classifier choice NGBoost & 0.0963 & 0.0124 & 7.78 & 1.3e-13& *** \\ 
  classifier choice RandomForest & 0.0610 & 0.0123 & 4.94 & 1.3e-06& *** \\ 
  classifier choice XGBoost & 0.1321 & 0.0124 & 10.62 & <2e-16& *** \\ 
  train percentage 0.5 & -0.0175 & 0.0079 & -2.23 & 0.0268& * \\ 
  test percentage 0.5 & -0.1531 & 0.0079 & -19.47 & <2e-16& *** \\  
   \hline
\end{tabular}
\caption{\label{sup_tab_3}Table of parameter coefficients for \textbf{MIMIC-III} dataset.}
\end{table}

\clearpage

\subsection*{ANOVA results for the Simulated dataset}

\begin{table}[H]
\centering
\begin{tabular}{l|llllc}
  \hline
Variables & Estimate & Std. Error & $t$ value & Pr($>|t|$) \\ 
  \hline
imputation choice GAIN & 1.6777 & 0.0306 & 54.79 & <2e-16& *** \\ 
  imputation choice Mean & 1.6191 & 0.0302 & 53.61 & <2e-16& *** \\ 
  imputation choice MICE & 1.6253 & 0.0302 & 53.74 & <2e-16& *** \\ 
  imputation choice MissForest & 1.6552 & 0.0302 & 54.85 & <2e-16& *** \\ 
  imputation choice MIWAE & 1.7848 & 0.0313 & 57.01 & <2e-16& *** \\ 
  classifier choice Neural Network & 0.9962 & 0.0459 & 21.69 & <2e-16& *** \\ 
  classifier choice NGBoost & -0.0190 & 0.0387 & -0.49 & 0.6241 \\ 
  classifier choice Random Forest & -0.0482 & 0.0385 & -1.25 & 0.2112 \\ 
  classifier choice XGBoost & 0.2913 & 0.0404 & 7.22 & 6.31e-12& *** \\ 
  train percentage 0.5 & -0.1057 & 0.0330 & -3.21 & 0.0015& ** \\ 
  test percentage 0.5 & -0.4593 & 0.0320 & -14.34 & <2e-16& *** \\ 
  imputation choice Mean:train percentage 0.5 & 0.0130 & 0.0269 & 0.48 & 0.6285 \\ 
  imputation choice MICE:train percentage 0.5 & -0.0301 & 0.0271 & -1.11 & 0.2675 \\ 
  imputation choice MissForest:train percentage 0.5 & -0.0334 & 0.0266 & -1.26 & 0.2105 \\ 
  imputation choice MIWAE:train percentage 0.5 & -0.0671 & 0.0276 & -2.43 & 0.0157& * \\ 
  imputation choice Mean:classifier choice Neural Network & -0.0534 & 0.0456 & -1.17 & 0.2429 \\ 
  imputation choice MICE:classifier choice Neural Network & -0.0842 & 0.0453 & -1.86 & 0.0642& . \\ 
  imputation choice MissForest:classifier choice Neural Network & -0.2190 & 0.0445 & -4.92 & 1.58e-06& *** \\
  imputation choice MIWAE:classifier choice Neural Network & -0.0732 & 0.0464 & -1.58 & 0.1159 \\ 
  imputation choice Mean:classifier choice NGBoost & -0.0507 & 0.0410 & -1.23 & 0.2182 \\ 
  imputation choice MICE:classifier choice NGBoost & 0.0408 & 0.0411 & 0.99 & 0.3226 \\ 
  imputation choice MissForest:classifier choice NGBoost & -0.0666 & 0.0407 & -1.64 & 0.1030 \\ 
  imputation choice MIWAE:classifier choice NGBoost & -0.0090 & 0.0419 & -0.22 & 0.8298 \\ 
  imputation choice Mean:classifier choice RandomForest & -0.1012 & 0.0408 & -2.48 & 0.0137& * \\ 
  imputation choice MICE:classifier choice RandomForest & 0.0109 & 0.0409 & 0.27 & 0.7907 \\ 
  imputation choice MissForest:classifier choice RandomForest & -0.1162 & 0.0405 & -2.87 & 0.0044& **\\ 
  imputation choice MIWAE:classifier choice RandomForest & -0.0416 & 0.0417 & -1.00 & 0.3196 \\ 
  imputation choice Mean:classifier choice XGBoost & -0.1364 & 0.0420 & -3.25 & 0.0013& ** \\ 
  imputation choice MICE:classifier choice XGBoost & 0.0125 & 0.0423 & 0.29 & 0.7684 \\ 
  imputation choice MissForest:classifier choice XGBoost & -0.1325 & 0.0417 & -3.18 & 0.0017& ** \\ 
  imputation choice MIWAE:classifier choice XGBoost & -0.0124 & 0.0432 & -0.29 & 0.7743 \\ 
  classifier choice NeuralNetwork:train percentage 0.5 & -0.4091 & 0.0466 & -8.79 & 2.40e-16& *** \\ 
  classifier choice NGBoost:train percentage 0.5 & -0.0428 & 0.0394 & -1.09 & 0.2786 \\ 
  classifier choice RandomForest:train percentage 0.5 & -0.0728 & 0.0389 & -1.87 & 0.0626& . \\ 
  classifier choice XGBoost:train percentage 0.5 & -0.1845 & 0.0409 & -4.52 & 9.70e-06& *** \\ 
  imputation choice Mean:test percentage 0.5 & 0.0415 & 0.0272 & 1.52 & 0.1286 \\ 
  imputation choice MICE:test percentage 0.5 & 0.0186 & 0.0274 & 0.68 & 0.4966 \\ 
  imputation choice MissForest:test percentage 0.5 & -0.0984 & 0.0269 & -3.65 & 0.0003& *** \\ 
  imputation choice MIWAE:test percentage 0.5 & 0.0305 & 0.0279 & 1.09 & 0.2753 \\ 
  classifier choice Neural Network:test percentage 0.5 & -0.5158 & 0.0442 & -11.67 & <2e-16& *** \\ 
  classifier choice NGBoost:test percentage 0.5 & -0.0235 & 0.0378 & -0.62 & 0.5335 \\ 
  classifier choice RandomForest:test percentage 0.5 & 0.0170 & 0.0375 & 0.45 & 0.6513 \\ 
  classifier choice XGBoost:test percentage 0.5 & -0.1445 & 0.0392 & -3.69 & 0.0003& *** \\ 
  train percentage 0.5:test percentage 0.5 & 0.0569 & 0.0374 & 1.52 & 0.1292 \\ 
  classifier choice Neural Network:train percentage 0.5:test percentage 0.5 & 0.1876 & 0.0591 & 3.17 & 0.0017& ** \\ 
  classifier choice NGBoost:train percentage 0.5:test percentage 0.5 & 0.0169 & 0.0524 & 0.32 & 0.7468 \\ 
  classifier choice RandomForest:train percentage 0.5:test percentage 0.5 & 0.0260 & 0.0520 & 0.50 & 0.6172 \\ 
  classifier choice XGBoost:train percentage 0.5:test percentage 0.5 & 0.0778 & 0.0539 & 1.44 & 0.1501 \\
   \hline
\end{tabular}
\caption{\label{sup_tab_4}Table of parameter coefficients for Simulated dataset.}
\end{table}

\begin{landscape}

\subsection*{Configurations of the models used in the interpretability analysis}

\begin{table}[ht]
\centering
\begin{tabular}{|l || p{0.5in} |p{0.8in} |p{0.8in} |p{0.5in} |p{0.5in} |p{0.8in} |p{0.5in} |p{0.8in} |p{0.8in} |p{0.5in} |}
  \hline
Classifier & Imputation & Train Missingness & Test Missingness & Holdout Set & Distance Ratio & Mean Test AUC & Max Depth & Min Samples Split & Min Samples Leaf & Estimators\\
  \hline
Random Forest & MICE & 0.25 & 0.25 & 1 & 1.16 & 0.83 & 4 & 2 & 4 & 60 \\
Random Forest & Mean & 0.25 & 0.25 & 0 & 1.68 & 0.83 & 4 & 2 & 3 & 90\\
   \hline
\end{tabular}
\caption{\label{tab:interp_rf} The model configurations used in the interpretability analysis for the Random Forest.}
\end{table}

\begin{table}[ht]
\centering
\begin{tabular}{|l || p{0.5in} |p{0.8in} |p{0.8in} |p{0.5in} |p{0.5in} |p{0.8in} |p{0.5in} |p{0.8in} |p{0.8in}|}
  \hline
Classifier & Imputation & Train Missingness & Test Missingness & Holdout Set & Distance Ratio & Mean Test AUC & Max Depth & Estimators & Subsampling\\
  \hline
XGBoost & MICE & 0.25 & 0.25 & 2 & 1.11 & 0.87 & 5 & 400 & 0.6 \\
XGBoost & GAIN & 0.25 & 0.25 & 0 & 1.51 & 0.88 & 5 & 400 & 0.6 \\
   \hline
\end{tabular}
\caption{\label{tab:interp_xg} The model configurations used in the interpretability analysis for XGBoost.}
\end{table}

\begin{table}[ht]
\centering
\begin{tabular}{|l || p{0.5in} |p{0.8in} |p{0.8in} |p{0.5in} |p{0.5in} |p{0.8in} |p{0.5in} |p{0.8in} |p{0.8in}|}
  \hline
Classifier & Imputation & Train Missingness & Test Missingness & Holdout Set & Distance Ratio & Mean Test AUC & Estimators & Learning Rate & Minibatch Fraction\\
  \hline
NGBoost & MICE & 0.25 & 0.25 & 1 & 1.16 & 0.84 & 400 & 0.01 & 0.5 \\
NGBoost & GAIN & 0.25 & 0.25 & 0 & 1.51 & 0.85 & 350 & 0.01 & 0.5 \\
   \hline
\end{tabular}
\caption{\label{tab:interp_ng} The model configurations used in the interpretability analysis for NGBoost.}
\end{table}

\end{landscape}

\clearpage

\subsection*{Correlation between discrepancy statistics for the Simulated dataset}

\end{document}